\documentclass[twoside,11pt]{report}
\usepackage{marginnote}
\usepackage[left=1.5in,right=1.5in,top=1.5in,bottom=1.5in]{geometry}
\usepackage[usenames,dvipsnames]{xcolor}
\usepackage{graphicx,amsmath,multirow,booktabs,color,verbatim,url,placeins,framed}\usepackage{amsthm}\usepackage{amssymb}
\usepackage{array,wrapfig}
\usepackage[normalem]{ulem}
\usepackage{tgpagella}
\usepackage{epigraph}
\usepackage{epsfig}
\usepackage{wasysym}
\usepackage{xcolor} 
\usepackage{bbm}
\usepackage{mathtools}
\usepackage{wrapfig}
\usepackage{authblk}
\usepackage{wasysym}
\usepackage{algorithm}
\usepackage{algorithmicx}
\usepackage{algpseudocode}
\usepackage{wrapfig}
\usepackage{caption}
\usepackage{tabularx}
\usepackage{makecell}
\usepackage{gensymb}
\usepackage{algpseudocode}
\usepackage{fancyhdr}
\usepackage{txfonts}

\usepackage{listings}
\usepackage{subcaption}
\usepackage[square,sort,comma,numbers]{natbib}

\usepackage{graphics, graphicx, caption,float,subcaption,booktabs,xcolor,multirow,array,color,ifthen,tabu,colortbl,dblfloatfix,url,amsfonts,bm,ragged2e,tikz,stackengine,etoolbox,xpatch,enumerate,xstring,setspace,tabularx,makecell,changepage,cuted,titlesec,enumitem,wrapfig,tcolorbox, wrapfig, capt-of, amssymb,xspace,nicefrac,microtype,amsmath,amsthm, xparse,mathtools,patchcmd,longtable}
\usepackage[pagebackref=true,breaklinks=true,colorlinks,bookmarks=false]{hyperref}
\usepackage{wrapfig}
\usepackage[capitalise,noabbrev,nameinlink]{cleveref}
\usepackage[english]{babel}

\crefname{section}{Sec.}{Secs.}
\Crefname{section}{Section}{Sections}
\Crefname{table}{Table}{Tables}
\crefname{table}{Tab.}{Tabs.}
\Crefname{equation}{Equation}{Equations}
\crefname{equation}{eq.}{eqs.}
\crefname{chapter}{Chapter}{Chapters}

\renewcommand{\epigraphsize}{\small}
\renewcommand{\textflush}{flushright}
\renewcommand{\sourceflush}{flushright}
\newcommand{\epitextfont}{\itshape}
\newcommand{\episourcefont}{\scshape}

\makeatletter
\newsavebox{\epi@textbox}
\newsavebox{\epi@sourcebox}
\newlength\epi@finalwidth
\renewcommand{\epigraph}[2]{%
  \vspace{\beforeepigraphskip}
  {\epigraphsize\begin{\epigraphflush}
   \epi@finalwidth=\z@
   \sbox\epi@textbox{%
     \varwidth{\epigraphwidth}
     \begin{\textflush}\epitextfont#1\end{\textflush}
     \endvarwidth
   }%
   \epi@finalwidth=\wd\epi@textbox
   \sbox\epi@sourcebox{%
     \varwidth{\epigraphwidth}
     \begin{\sourceflush}\episourcefont#2\end{\sourceflush}%
     \endvarwidth
   }%
   \ifdim\wd\epi@sourcebox>\epi@finalwidth 
     \epi@finalwidth=\wd\epi@sourcebox
   \fi
   \leavevmode\vbox{
     \hb@xt@\epi@finalwidth{\hfil\box\epi@textbox}
     \vskip1.75ex
     \hrule height \epigraphrule
     \vskip.75ex
     \hb@xt@\epi@finalwidth{\hfil\box\epi@sourcebox}
   }%
   \end{\epigraphflush}
   \vspace{\afterepigraphskip}}}
\makeatother

\usepackage{amsmath,amsfonts,bm}

\def\eqref#1{equation~\ref{#1}}

\def\1{\bm{1}}

\DeclareMathAlphabet{\mathsfit}{\encodingdefault}{\sfdefault}{m}{sl}
\SetMathAlphabet{\mathsfit}{bold}{\encodingdefault}{\sfdefault}{bx}{n}

\newcommand{\E}{\mathbb{E}}

\newcommand{\clearemptydoublepage}{\newpage{\pagestyle{empty}\cleardoublepage}}

\newcommand{\kw}[1]{\textbf{#1}}
\newcommand{\pddl}[1]{{\texttt{#1}}} %

\usepackage{color, colortbl}
\definecolor{Gray}{gray}{0.9}

\begin{document}

\thispagestyle{empty}
\date{}

\begin{center}

{\Huge Unlocking Generalization for Robotics via Modularity and Scale} \\
\vspace{1cm}
{\Large Murtaza Dalal} \\
\vspace{1cm} 

\vspace{-0.5cm}
{\Large CMU-RI-TR-25-02}
\vspace{0.5cm}

{\Large January 2025} \\
\vspace{1cm}

{\large
The Robotics Institute\\
Carnegie Mellon University\\
Pittsburgh, Pennsylvania 15213\\
}
\vspace{1cm}
{\Large
{\bf Thesis Committee}\\
\begin{table}[h!]
    \centering
    \resizebox{0.75\linewidth}{!}{
    \begin{tabular}{rl}
        Ruslan Salakhutdinov & Carnegie Mellon University (\textit{Chair}) \\
        Deepak Pathak & Carnegie Mellon University \\
        David Held & Carnegie Mellon University \\
        Shuran Song & Stanford University \\
        Ankur Handa & NVIDIA
    \end{tabular}}
\end{table}

}
\par ~ \\
{\it Thesis submitted in partial fulfillment of the \\
    requirements for the degree of Doctor of Philosophy in Robotics}
\vfill

\vspace{0.5cm}
{\small \copyright~ Murtaza Dalal, 2025. All rights reserved.}
\end{center}

\clearpage

\pagenumbering{Roman}
\clearemptydoublepage
\setcounter{page}{1}

\begin{centering} \section*{Acknowledgements} 
\end{centering}
A PhD, as we all know, is a long journey, a marathon, not a sprint. While it is often a lonely one, with long hours spent in the lab, toiling away at experiments, writing code, deploying robots, it is by no means a solitary one. Not at all. Rather it is the people around us, the people in our lives, our families, our friends, our mentors, our collaborators and all those who want the best for us that help make this incredible achievement possible. I want to spend a bit of time here to acknowledge the people in my life that helped make this happen for me. 

Any discussion of how I got here begins with the most important people in my life, my family. Mom and Dad, you raised me to be the ambitious, high achieving, well-rounded individual I am today. You have pushed me to achieve my dreams at every stage of my life, from school, to college and beyond. You have been my rock when I have struggled, when I have been stressed and when I have broken down. You have been the proud audience members at each of my achievements and are the people I always strive hardest to impress. Without you, this thesis would not have been possible, so with all my heart I thank you for always being there, for being \textit{my} parents and constantly supporting and encouraging me to pursue my dreams. Beyond just my parents, I want to thank my brother Taykhoom, my constant companion and friend since childhood, for always being there for me, listening to my rants and crazy ideas and giving me the most sage advice, even as the younger child. Finally, I want to thank my grandparents, Nana, Nani, Dada, and Dadi, and of course my entire extended family, your love and support has gotten me to where I am today!

Now I would like to thank the people who helped turn me into the researcher I am today, my advisors and mentors. Ruslan, I knew I wanted to work with you from our first meeting over Skype all the way back in 2020. You sold me on your vision of embodied agents being the future of machine learning research and I just knew that we had to work together. It's been a blast working with you ever since. You have always supported my ideas, work and directions while nudging and guiding me along the way to be the best researcher I can be. Not only in just research, but in career and life advice, you have given me invaluable tokens that I will constantly cherish, appreciate and use. I have always felt that you have my best interests at heart, and to me that means more than anything else in the world. Thank you for taking me on as your student and I look forward to our relationship moving forward!

Deepak, my advisor in all but name, your impact on my research career is quite literally incalculable. You trained me to be ambitious in my research, yet practical in my execution, to think big but also get things to work, to read, read and read, anything and everything I could possibly get my hands on and to always make sure things work in the real world, as quickly as possible. I have deeply ingrained these lessons and I believe they have been critical to my success as a robotics researcher. Beyond work though, you, through sheer force of will, fostered an incredible lab environment and community from which I count some of my closest friends and collaborators. I am incredibly fortunate to have worked so closely with you during my PhD - thank you for making me a part of your lab, your community and your vision, it has been an experience of a lifetime!

During my PhD, I was fortunate to intern/collaborate at NVIDIA and Apple, experiences which heavily shaped and guided the direction of my PhD. Ankur, Ajay and Caelan, I had a fantastic summer working with you at the Seattle Robotics Lab. I learned so much about imitation learning, TAMP, and large scale learning from the three of you. Your investment in my project, all the way from high-level ideation to low-level coding, made my experience all the more enjoyable and productive. The work I did at NVIDIA had a lasting impact on my PhD and how I view and value distillation, sim2real transfer and large scale learning for robotics. Jian, Walter and Chen, working with you in the final year of my PhD was an absolute pleasure. ManipGen, the work that I am most proud of during my PhD, would not have been possible without your advice and involvement. Your unique perspectives helped shape a lot of my thinking and provided the best of sounding boards for my ideas. I am incredibly fortunate to have worked with the three of you.

Finally, on the topic of critical mentors that shaped my research, I cannot complete this discussion without mentioning my experience at UC Berkeley. Sergey, you took a chance on a hungry, yet inexperienced freshman who had only just learned linear algebra and thrust him into the world of research. Thank you for giving me that opportunity, I would not be here today without you! Vitchyr, my first ever research mentor, you shaped my core research identity and gave me the tools to succeed. Thank you for putting up with all of my questions and struggles, your mentorship is truly a model for how to enable undergraduate researchers to succeed and be motivated to pursue PhDs. 

While advisors and mentors are critical to the success of your PhD, so are your friends and community - the ones that help keep you going and make the entire experience a fun one.

Russell, outside of my advisors, you have perhaps shaped my research career more than anyone else. You introduced me to machine learning and I introduced you to Sergey and RL research. The rest is of course, history. We have had an incredible journey together, from freshman year roommates  to working in the same lab together during PhD and beyond. We have had countless discussions on research, AI, the future and everything in between. It is difficult to fit everything into a single paragraph, so I will just say this: You made this thesis happen, you are the person I turn to whenever I get stuck on a problem, the person I discuss with when I want to generate ideas, the person who motivates me to get back at it when I'm struggling. You have had a hand in every project I've ever done, from being a sounding board on ideas to a direct collaborator. If I have one regret from my PhD, it is that we did not directly work together enough. But I genuinely thank you Russell, for always being there for me, not only as a researcher and lab-mate, but as one of my best and lifelong friends.

PhD students are extremely ambitious, driven people who are truly passionate about their work. That often means that when they spend time together, that is what they mostly talk about. With you, Alex, however, we explicitly never talked shop and I deeply, deeply valued that. You were the friend that separated me from work which is so incredibly critical to maximizing your mental and emotional health during a journey as long and intense as a PhD. Our weekly Game of Thrones / cooking nights kept me going, both through the light and dark times of my PhD. Your genuine kindness and supportiveness as a friend always touched me and I always knew I could come to you for anything I needed. Thank you Alex, for being there for me and for being one of my closest friends.

Sometimes, you create unexpected communities that have an outsized impact on your PhD experience. Weekly Novelty Dinner was that for me. What started as just a way for my friend Gaurav and I to explore Pittsburgh, turned into a weekly experience with some of my fondest PhD memories. From devouring the entire menu at Apteka to the flaming rice at Caspian Corner, Novelty produced some of the greatest fun I had during my PhD. More often than that, Novelty was just a source of stability and something to look forward to each week. Some folks even said ``it was the highlight of their week". I want to thank Gaurav for co-organizing this with me and the entire Novelty crew (too many to list here) for coming every week and making the experience as great as it was!

My final year and a half of my PhD was perhaps my most intense and challenging, yet it would not have been the same with this group of friends also known as The Biweekly Breakfast Club. Satyam, Srujana and Vallari, I still cannot believe the incredible circumstances under which we met, but I am incredibly thankful that we did and that we quickly became as close as we are now. I cherish our time together in Pittsburgh quite a bit - from waking up super early to sample the entire Pittsburgh brunch scene to attending the Indian CMU events and so much more. This group helped me stabilize and balance when work was at its most challenging and got me through some of my most difficult times. 

Finally, I want to acknowledge all the friends I made throughout my PhD. Shikhar, my first ever research partner from undergrad, idea sounding board, close friend and lab mate, Ananye, Lili and Rohan with whom I made some incredibly fun memories hanging out over the years, particularly going to the gym together. Mononito, my friend who will walk anywhere with me, especially when it would be easier to drive instead. Alex, Christina, Suvansh and Thomas, with whom I shared some great times playing boardgames together. Gokul, with whom I always have the most insightful conversations, over the best food. Rishi, of course, for putting up with me as a housemate for all these years. Mansi, who almost became my mentee, but instead one of my good friends with whom I've shared some great memories at RI events. Adithya, Albert, Michael, and Yang who I met and became close with during my internship at NVIDIA. Sudeep, who dragged me to so many football games that I enjoyed despite my complaints! Gina, my physical therapist who also became one of my close friends and fellow ice skater. Andrew, for re-introducing me to pickle ball and so much more! We had so many great times at Novelty's and formals. I want to also thank everyone in both of my labs, LEAP Lab and Russ Lab for all of the fun memories, socials and insightful discussions we had over all these years. The RI and GSA communities, for being so vibrant and organizing so many fun events that made my time in Pittsburgh during my PhD so much more enjoyable. If there is anyone who I have not mentioned, please know you are all in my heart and I genuinely appreciate everyone who has come into my life over these last four and a half years.

Overall, this journey would not have been the same without any of you. So thank you all so much for being there for me, you made this incredible accomplishment possible!

\clearpage

\begin{centering} \section*{Abstract} \end{centering}
How can we build generalist robot systems? Looking at fields such as vision and language, the common theme has been large scale end-to-end learning with massive, curated datasets. 
In robotics, on the other hand, scale alone may not be enough due to the significant multi-modality of robotics tasks, lack of easily accessible data and the safety and reliability challenges of deploying on physical hardware. Meanwhile, some of the most successfully deployed robotic systems today are inherently modular and can leverage the independent generalization capabilities of each module to perform well. Inspired by these qualities, this thesis seeks to tackle the task of building generalist robot agents by integrating these components into one: combining modularity with large scale learning for general purpose robot control. 

We begin by exploring these two aspects independently. The first question we consider is: how can we build modularity and hierarchy into learning systems? Our key insight is that rather than having the agent learn hierarchy and low-level control end-to-end, we can explicitly enforce modularity via planning to enable significantly more efficient and capable robot learners. Next, we come to the role of scale in building generalist robot systems. To effectively scale, neural networks require vast amounts of diverse data, expressive architectures to fit the data and a source of supervision to generate the data. To that end, we leverage a powerful supervision source: classical planning algorithms, which can generalize broadly, but are expensive to run and require access to perfect, privileged information to perform well in practice. We use these planning algorithms to supervise large-scale policy learning in simulation to produce generalist agents.

Finally, we consider how to unify modularity with large-scale policy learning to build autonomous real-world robot systems capable of performing zero-shot long-horizon manipulation. We propose to do so by tightly integrating key ingredients of modular high and mid-level planning, learned local control, procedural scene generation and large-scale policy learning for sim-to-real transfer. We demonstrate that this recipe can produce powerful results: a single, generalist agent can solve challenging long-horizon manipulation tasks in the real world, solely from text instruction.
{
  \hypersetup{linkcolor=black}
  \tableofcontents
}
\clearpage

{
  \hypersetup{linkcolor=black}
  \listoffigures
}
\clearpage

{
  \hypersetup{linkcolor=black}
  \listoftables
}
\cleardoublepage

\pagenumbering{arabic}
\setcounter{page}{1}

\chapter{Introduction}
\label{chap:introduction}
\epigraph{\textit{A jack of all trades is a master of none, but oftentimes better than a master of one.}}{\textit{William Shakespeare}}

\section{Motivation}
Of what use is generalization in robotics? While humans have built machines to automate work by eliminating a significant portion of repetitive, dangerous labor, these automatons can only complete a limited set of tasks in a limited set of regimes. Move things slightly out of place, change the task by a small amount, experience a bit too much wear and tear and these systems tend to fall apart. To effectively automate the future of manual labor, robots must be able to operate as humans do: in general, unstructured environments, with under-specified instructions and perhaps unseen tasks or circumstances. In effect, robots must be able to \textit{generalize}. 

Naturally, then the question becomes generalize to what? For robots to operate autonomously in-the-wild, they need to be able to robustly solve unseen tasks, in varying environments, and be invariant to poses, arrangements, objects, distractors and backgrounds. General purpose robots must be able to generalize across semantics, geometry and the low-level skills required to solve a large set of tasks. Cast in this manner, the problem of generalization appears quite daunting: there are a large number of axes upon which we must generalize, each posing a unique challenge. 

In recent years, a simple recipe has been shown to be quite capable when tackling questions of generalization along complex axes and tasks: large-scale data-driven learning. By simplifying the problem to learning from large sources of data with sufficiently expressive models, the machine learning community has demonstrated impressive generalization results across vision and language tasks. When applied to robotics, however, several key issues crop up. First and foremost, there is the issue of data, or lack thereof for robotics: datasets must be manually collected and curated. Second, there is the sheer number of axes to generalize across, which can make the data requirement impractical, particularly when learning to map from visual observation to low-level control. Finally, robotics is fundamentally a hardware problem, and thus issues such as safety, reliability and deployability are of paramount importance, yet challenging to handle with learned black-box controllers.

To address these challenges, we take inspiration from the human brain, which is divided into different cortices that handle different components of behavior. Among the various modules that make up our brain, we have components for handling motor control, sensing information as well as planning and reasoning. For robotics, this naturally lends itself to a modular approach in which different modules are responsible for generalizing across independent axes. 
As a result, we can decompose the overall generalization problem into sub-problems: 1) semantics and planning, 2) vision and geometry and 3) low-level robot control. In this thesis, our key insight is to combine explicit hierarchies with large-scale learning, using advances in vision, language and planning to handle sub-problems 1 and 2, while scaling robot learning in simulation via planner distillation for problem 3, in order to produce generalist robot agents.

To do so, we first explore how modularity can be used to accelerate and improve learning performance. In Chapter~\ref{chap:raps}, we structure the learning problem as discovering high-level control strategies over manually engineered skills. While this paradigm dramatically improves learning performance, its capabilities are bounded by the expressiveness of the hand-defined skills. At the same time, we note that classical and Large Language Model (LLM) planning algorithms generalize well over high-level reasoning tasks yet struggle to perform low-level control, which is a strength of learning methods. Building on these insights, in Chapter~\ref{chap:psl}, we develop an algorithm for efficiently learning to solve long-horizon robotics tasks.

Next, we aim to understand how to scale learning in order to produce generalist policies for robot control. Our key insight is planning algorithms can supervise large-scale policy learning in simulation. We can perform distillation by generating large amounts of diverse, procedurally generated scenes in simulation and fitting models to the expert. We first explore this framework in Chapter~\ref{chap:optimus} for end-to-end learning of long-horizon manipulation and show strong generalization results across simulated pick and place as well as articulated object manipulation tasks. In Chapter~\ref{chap:nmp}, we then focus on applications to real robots via sim-to-real transfer, demonstrating that large-scale policies trained in simulation to perform motion planning can effectively generalize to real world motion generation tasks.

Finally, in Chapter~\ref{chap:manipgen}, we unify the principles of modularity and scale into a single framework for enabling robots to generalize to a large set of unseen tasks. We build a robust Visual Language Model (VLM) planning system for combinatorial generalization over a set of pre-trained skills. We ground VLM plans in real-world scenes via fast, reactive geometric motion planning trained at scale. Lastly, and most importantly, we train skills using a new policy class for sim-to-real transfer, one that is focused on ease of transfer, pose invariance and long-horizon chaining. We demonstrate that this agent is broadly capable, solving a wide array of real-world robotic manipulation tasks solely from text instruction. 

\section{Contributions}
The key questions we study in this thesis are 1) How do we integrate modularity into learning; what should the hierarchy be and what should we plan vs. learn? 2) Can we create a simple, yet practical recipe for scaling robot learning? and 3) How can we unify modularity and large scale learning to create generalist robot agents?  

\subsection{Integrating Modularity into Learned Control}
In order to integrate modularity into learning algorithms, we consider 1) How should we construct the hierarchy? We explore two forms: high-level and low-level split, high-level, mid-level and low-level split. 2) What to learn vs. what to plan? We explore learning the high-level, and planning the low-level, and vice versa.

\textbf{Robot Action Primitives} In Chapter~\ref{chap:raps}, we design modular RL method that involves specifying a library of robot action primitives (RAPS), parameterized with arguments that are learned by an RL policy. These parameterized primitives are expressive, simple to implement, enable efficient exploration and can be transferred across robots, tasks and environments. In this case, we have high/low-level split hierarchy in which we learn the high-level policy while planning the low-level control. We perform a thorough empirical study across challenging tasks in three distinct domains with image input and a sparse terminal reward. We find that our simple change to the action interface substantially improves both the learning efficiency and task performance irrespective of the underlying RL algorithm, significantly outperforming prior methods which learn skills from offline expert data. However, this decomposition fails to scale well to more challenging, contact-rich tasks with longer-horizons.

\textbf{Plan-Sequence-Learn} As a result, in Chapter~\ref{chap:psl} we instead use the internet-scale knowledge from LLMs for high-level policies, guiding reinforcement learning (RL) policies to efficiently solve robotic control tasks online without requiring a pre-determined set of skills. Specifically, we propose a modular approach that uses motion planning to bridge the gap between abstract language and learned low-level control for solving long-horizon robotics tasks from scratch. PSL solves long-horizon tasks from raw visual input spanning four benchmarks, out-performing language-based, classical, and end-to-end approaches. Although this approach is capable of solving a particular long-horizon task end-to-end efficiently, our true goal is generalization, so for that we first explore large-scale learning for specific skills.

\subsection{Scaling via Procedural Scene Generation and Planner Distillation}
Broadly, in this section we propose a simple recipe for large-scale learning: 1) We can use simulation as a means of generating vast amounts of data, 2) use procedural scene generation as a way of making the data as diverse as possible, 3) To generate expert data itself, we will use planning algorithms, which can already solve a wide array of tasks in simulation, 4) Distill planners into closed-loop visuomotor policies that can operate in the real world using solely perceptual information

\textbf{TAMP Imitation at Scale} In Chapter~\ref{chap:optimus} we show that the combination of large-scale datasets generated by TAMP supervisors and flexible Transformer models to fit them is a powerful paradigm for robot manipulation. To this end, we present an imitation learning system called OPTIMUS that trains large-scale visuomotor Transformer policies by imitating a TAMP agent. OPTIMUS introduces a pipeline for generating TAMP data that is specifically curated for imitation learning and can be used to train performant transformer-based policies. We demonstrate that OPTIMUS can solve a wide variety of challenging vision-based manipulation tasks with a wide variety of objects, ranging from long-horizon and pick-and-place tasks, to shelf and articulated object manipulation. While OPTIMUS achieves great results and the recipe is powerful when applied to TAMP imitation, it is hard to scale end-to-end manipulation on its own due to its inherent complexity. 

\textbf{Neural Motion Planning} We can instead stress test generalization instead by focusing on one skill: collision free motion generation, but maximize the diversity on objects, obstacles and scene configurations. In Chapter~\ref{chap:nmp} we do so by building a large number of complex scenes in simulation, collecting expert data from a motion planner, then distilling it into a reactive generalist policy. We then combine this with lightweight optimization to obtain a safe path for real world deployment. We evaluate our method on 64 real-world motion planning tasks across four diverse environments with randomized poses, scenes and obstacles, in the real world, demonstrating significant improvements over motion planning success rate over state of the art sampling, optimization and learning based planning methods.

\subsection{A Unifying Framework for Building Generalist Agents by Combining Modularity and Scale}

In Chapter~\ref{chap:manipgen}, we unify large scale learning with modularity to enable a single agent to broadly generalize to a large swath of tasks. The key-idea builds on Plan-Seq-Learn and transforms it into a generalist approach for manipulation: First to perform real world task planning, we require the agent to perceive the environment, for which use Visual Language Models.
Next, for fast, reactive, real-world motion planning, we train a motion planner that can operate on noisy point cloud input: Neural Motion Planner. Lastly, to solve new tasks in a zero-shot manner, we require pre-trained skills for which we scale up learning in simulation and transfer sim-to-real. To do so, we propose a new class of policies for sim-to-real transfer: local policies. Locality enables a variety of appealing properties including invariance to absolute robot and object pose, skill ordering, and global scene configuration, producing broad generalization at deployment. When deployed, this single agent is capable of solving a wide array of real-world robotic manipulation tasks entirely zero shot, from natural language instructions.

\part{Integrating Modularity into Learned Control}

\chapter{Accelerating Robotic Reinforcement Learning via Parameterized Action Primitives}
\label{chap:raps}
\section{Introduction}
\label{intro}
\begin{figure}[t]
    \centering
    \includegraphics[width=.95\linewidth]{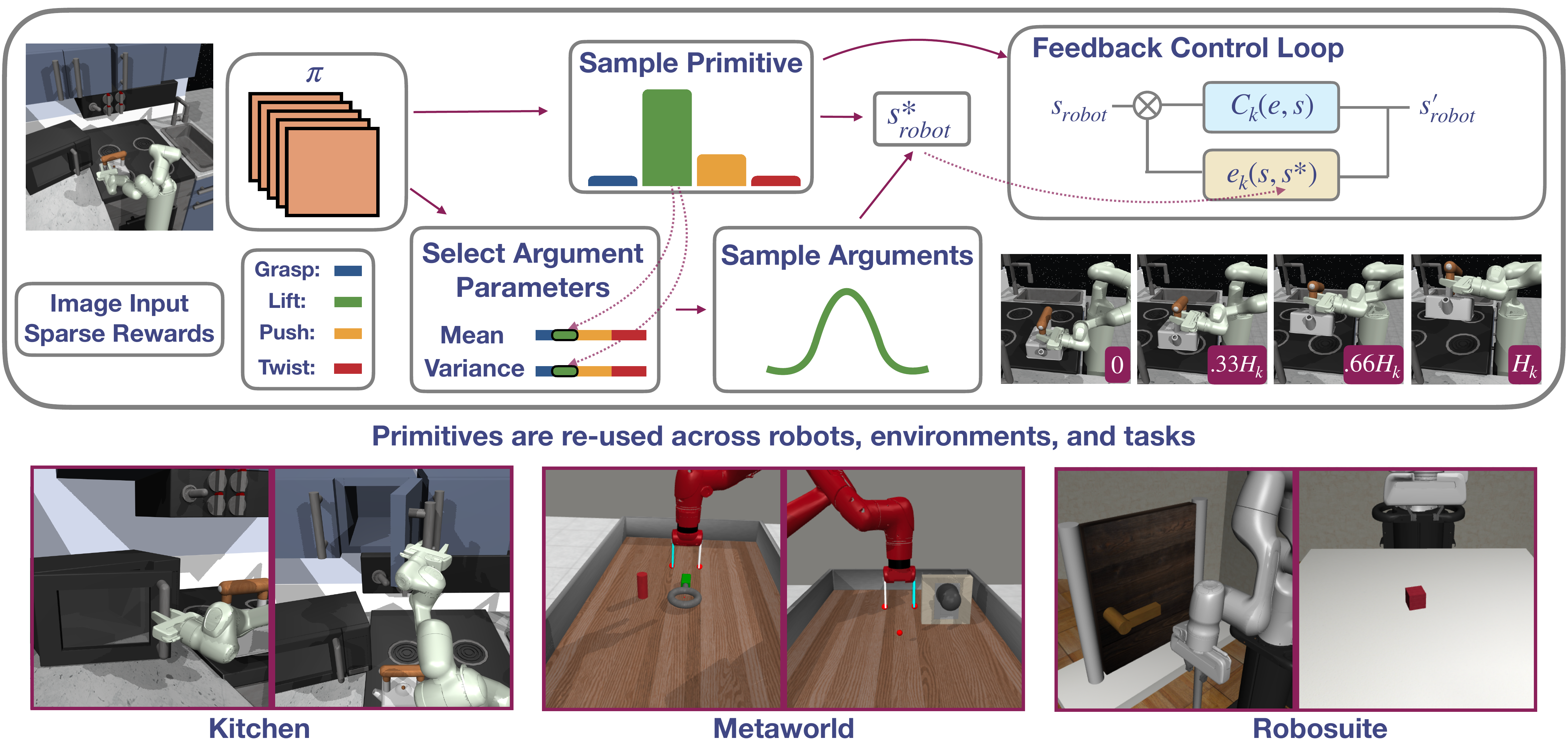}
    \caption{\footnotesize Visual depiction of RAPS\xspace, outlining the process of how a primitive is executed on a robot. Given an input image, the policy outputs a distribution over primitives and a distribution over all the arguments of all primitives, samples a primitive and selects its corresponding argument distribution parameters, indexed by which primitive was chosen, samples an argument from that distribution and executes a controller in a feedback loop on the robot for a fixed number of timesteps ($H_k$) to reach a new state. 
    We show an example sequence of executing the \texttt{lift} primitive after having grasped the kettle in the Kitchen environment. The agent observes the initial ($0$) and final states ($H_k$) and receives a reward equal to the reward accumulated when executing the primitive. Below we visualize representative tasks from the three environment suites that we evaluate on. 
    }
    \vspace{-0.19in}
    \label{fig:method_figure}
\end{figure}

Meaningful exploration remains a challenge for robotic reinforcement learning systems. For example, in the manipulation tasks shown in Figure~\ref{fig:method_figure}, useful exploration might correspond to picking up and placing objects in different configurations. However, random motions in the robot's joint space will rarely, if ever, result in the robot touching the objects, let alone pick them up. Recent work, on the other hand, has demonstrated remarkable success in training RL agents to solve manipulation tasks ~\citep{levine2016end, kalashnikov2018qt, andrychowicz2020learning} by sidestepping the exploration problem with careful engineering.
\citet{levine2016end} use densely shaped rewards, while \citet{kalashnikov2018qt} leverage a large scale robot infrastructure and \citet{andrychowicz2020learning} require training in simulation with engineered reward functions in order to transfer to the real world.
In general, RL methods can be prohibitively data inefficient, require careful reward development to learn, and struggle to scale to more complex tasks without the aid of human demonstrations or carefully designed simulation setups. 

An alternative view on why RL is difficult for robotics is that it requires the agent to
learn both \textit{what} to do in order to achieve the task and \textit{how} to control the robot to execute the desired motions. For example, in the kitchen environment featured at the bottom of Figure~\ref{fig:method_figure}, the agent would have to learn how to accurately manipulate the arm to reach different locations as well as how to grasp different objects, while also ascertaining what object it has to grasp and where to move it. Considered independently, the problems of controlling a robot arm to execute particular motions and figuring out the desired task from scalar reward feedback, then achieving it, are non-trivial. Jointly learning to solve both problems makes the task significantly more difficult. 

In contrast to training RL agents on raw actions such as torques or delta positions, a common strategy is to decompose the agent action space into higher (i.e., \textit{what}) and lower (i.e., \textit{how}) level structures. A number of existing methods have focused on designing or learning this structure, from manually architecting and fine-tuning action hierarchies \citep{vezhnevets2017feudal,nachum2018data,frans2017meta,li2019sub}, to organizing agent trajectories into distinct skills~\citep{hausman2018learning,sharma2019dynamics,allshire2021laser, xie2020latent} to more recent work on leveraging large offline datasets in order to learn skill libraries~\citep{lynch2020learning,shankar2019discovering}. 
While these methods have shown success in certain settings, many of them are either too sample inefficient, do not scale well to more complex domains, or lack generality due to dependence on task relevant data. 

In this chapter, we investigate the following question: instead of learning low-level primitives, what if we were to design primitives with minimal human effort, enable their expressiveness by parameterizing them with arguments and learn to control them with a high-level policy?
Such primitives have been studied extensively in task and motion planning (TAMP) literature~\citep{kaelbling2011hierarchical} and implemented as
parameterized actions~\citep{hausknecht2015deep} in RL. We apply primitive robot motions to redefine the policy-robot interface in the context of robotic reinforcement learning. These primitives include manually defined behaviors such as \texttt{lift}, \texttt{push}, \texttt{top-grasp}, and many others. The behavior of these primitives is parameterized by arguments that are the learned outputs of a policy network. For instance, \texttt{top-grasp} is parameterized by four scalar values: grasp position (x,y), how much to move down (z) and the degree to which the gripper should close.
We call this application of parameterized behaviors, Robot Action Primitives for RL (RAPS\xspace).
A crucial point to note is that these parameterized actions are \textit{easy} to design, need only be defined \textit{once} and can be \textit{re-used} without modification across tasks.

The main contribution of this work is to support the effectiveness of RAPS\xspace via a thorough empirical evaluation across several dimensions:
\begin{itemize}[leftmargin=1.5em,noitemsep,topsep=0pt]
\item How do parameterized primitives compare to other forms of action parameterization?
\item How does RAPS\xspace compare to prior methods that learn skills from offline expert data?
\item Is RAPS\xspace agnostic to the underlying RL algorithm?
\item Can we stitch the primitives to perform multiple complex manipulation tasks in sequence? 
\item Does RAPS\xspace accelerate exploration even in the absence of extrinsic rewards?
\end{itemize}
We investigate these questions across complex manipulation environments including Kitchen Suite, Metaworld and Robosuite domains. We find that a simple parameterized action based approach outperforms prior state-of-the-art by a significant margin across most of these settings\protect\footnotemark. 

\footnotetext{Please view our website for performance videos and links to our code: \href{https://mihdalal.github.io/raps/}{https://mihdalal.github.io/raps/}}

\section{Related Work}
\label{related-work}
\textbf{Higher Level Action and Policy Spaces in Robotics} $\quad$ In robotics literature, decision making over primitive actions that execute well-defined behaviors has been explored in the context of task and motion planning~\citep{cambon2009hybrid,kaelbling2011hierarchical,kaelbling2017learning,simeonov2020long}. However, such methods are dependent on accurate state estimation pipelines to enable planning over the argument space of primitives. One advantage of using reinforcement learning methods instead is that the agent can learn to adjust its implicit state estimates through trial and error experience. Dynamic Movement Primitive and ensuing policy search approaches~\citep{ijspeert2002learning,schaal2006dynamic,kober2009learning,peters2010relative,daniel2016hierarchical} leverage dynamical systems to learn flexible, parameterized skills, but are sensitive to hyper-parameter tuning and often limited to the behavior cloning regime. Neural Dynamic Policies~\citep{bahl2020neural} incorporate dynamical structure into neural network policies for RL, but evaluate in the state based regime with dense rewards, while we show that parameterized actions can enable RL agents to efficiently explore in sparse reward settings from image input.

\textbf{Hierarchical RL and Skill Learning} $\quad$
Enabling RL agents to act effectively over temporally extended horizons is a longstanding research goal in the field of hierarchical RL. Prior work introduced the options framework~\citep{sutton1999between}, which outlines how to leverage lower level policies as actions for a higher level policy. In this framework, parameterized action primitives can be viewed as a particular type of fixed option with an initiation set that corresponds to the arguments of the primitive. Prior work on options has focused on discovering ~\citep{eysenbach2018diversity, achiam2018variational,sharma2019dynamics} or fine-tuning options \citep{bacon2017option,frans2017meta,li2019sub} in addition to learning higher level policies. Many of these methods have not been extended beyond carefully engineered state based settings. More recently, research has focused on extracting useful skills from large offline datasets of interaction data ranging from unstructured interaction data~\citep{whitney2020dynamicsaware}, play~\citep{lynch2020learning,lynch2020grounding} to demonstration data~\citep{shankar2019discovering,shankar2020learning,zhou2020plas,pertsch2020accelerating,singh2020parrot,ajay2021opal,tanneberg2021skid}. While these methods have been shown to be successful on certain tasks, the learned skills are only relevant for the environment they are trained on. New demonstration data must be collected to use learned skills for a new robot, a new task, or even a new camera viewpoint. Since RAPS\xspace uses manually specified primitives dependent only on the robot state, RAPS\xspace can re-use the same implementation details across robots, tasks and domains.

\textbf{Parameterized Actions in RL} $\quad$ The parameterized action Markov decision process (PAMDP) formalism was first introduced in~\citet{masson2016reinforcement}, though there is a large body of earlier work in the area of hybrid discrete-continuous control, surveyed in  ~\citep{branicky1998unified, bemporad1999control}. Most recent research on PAMDPs has focused on better aligning policy architectures and RL updates with the nature of parameterized actions and has largely been limited to state based domains \citep{xiong2018parametrized,fan2019hybrid}. A number of papers in this area have focused on solving a simulated robot soccer domain modeled as either a single-agent \citep{hausknecht2015deep,masson2016reinforcement,wei2018hierarchical} or multi-agent~\citep{fu2019deep} problem. In this paper, we consider more realistic robotics tasks that involve interaction with and manipulation of common household objects.
While prior work~\citep{sharma2020learning} has trained RL policies to select hand-designed behaviors for simultaneous execution, we instead train RL policies to leverage more expressive, \textit{parameterized} behaviors to solve a wide variety of tasks. 
Closely related to this work is~\citet{chitnis2020efficient}, which develops a specific architecture for training policies over parameterized actions from \textit{state} input and sparse rewards in the context of bi-manual robotic manipulation. Our work is orthogonal in that we demonstrate that a higher level policy architecture is sufficient to solve a large suite of manipulation tasks from image input. We additionally note that there is concurrent work~\citep{nasiriany2021augmenting} that also applies engineered primitives in the context of RL, however, we consider learning from image input and sparse terminal rewards. 

\vspace{-.1in}
\section{Robot Action Primitives in RL}
\label{sec:main-method}
To address the challenge of exploration and behavior learning in continuous action spaces, we decompose a desired task into the \textit{what} (high level task) and the \textit{how} (control motion). The \textit{what} is handled by the \textit{environment-centric} RL policy while the \textit{how} is handled by a fixed, manually defined set of \textit{agent-centric} primitives parameterized by continuous arguments. This enables the high level policy to reason about the task at a high level by choosing primitives and their arguments while leaving the low-level control to the parameterized actions themselves.
\subsection{Background} 
Let the Markov decision process (MDP) be defined as $ (\mathcal{S}, \mathcal{A}, \mathcal{R}(s,a, s'), \mathcal{T}(s'|s,a),  p(s_0),\gamma,)$ in which $\mathcal{S}$ is the set of true states, $\mathcal{A}$ is the set of possible actions, $\mathcal{R}(s,a, s')$ is the reward function, $\mathcal{T}(s'|s,a)$ is the transition probability distribution, $p(s_0)$ defines the initial state distribution, and $\gamma$ is the discount factor. The agent executes actions in the environment using a policy $\pi(a|s)$ with a corresponding trajectory distribution $ p(\tau=(s_0, a_0, ... a_{t-1}, s_T)) = p(s_0) \Pi_t \pi(a_t|s_t)\mathcal{T}(s_{t+1}|s_t,a_t)$.
The goal of the RL agent is to maximize the expected sum of rewards with respect to the policy: $ \mathbb{E}_{s_0, a_0, ... a_{t-1}, s_T,  \sim p(\tau)} \left[ \sum_t \gamma^t \mathcal{R}(s_t, a_t) \right]$.
In the case of vision-based RL, the setup is now a partially observed Markov decision process (POMDP); we have access to the true state via image observations. In this case, we include an observation space $\mathcal{O}$ which corresponds to the set of visual observations that the environment may emit, an observation model $p(o|s)$ which defines the probability of emission and policy $\pi(a|o)$ which operates over observations. In this chapter, we consider various modifications to the action space $\mathcal{A}$ while keeping all other components of the MDP or POMDP the same.
\vspace{-.15in}
\subsection{Parameterized Action Primitives}
We now describe the specific nature of our parameterized primitives and how they can be integrated into RL algorithms (see Figure~\ref{fig:method_figure} for an end-to-end visualization of the method). In a library of $K$ primitives, the k-th primitive is a function $f_k(s, \operatorname{args})$ that executes a controller $C_k$ on a robot for a fixed horizon $H_k$, $s$ is the robot state and $\operatorname{args}$ is the value of the arguments passed to $f_k$. $\operatorname{args}$ is used to compute a target robot state $s^*$ and then $C_k$ is used to drive $s$ to $s^*$. A primitive dependent error metric $e_k(s, s^*)$ determines the trajectory $C_k$ takes to reach $s^*$. $C_k$ is a general purpose state reaching controller, e.g. an end-effector or joint position controller; we assume access to such a controller for each robot and it is straightforward to define and tune if not provided. In this case, the same primitive implementation can be re-used across any robot. In this setup, the choice of controller, error metric and method to compute $s^*$ define the behavior of the primitive motion, how it uniquely forms a movement in space.
We refer to Procedure~\ref{alg:p3} for a general outline of a parameterized primitive. To summarize, each skill is a feedback control loop with end-effector low level actions. The input arguments are used to define a target state to achieve and the primitive executes a loop to drive the error between the robot state and the target robot state to zero.

As an example, consider the ``lifting`` primitive, which simply involves lifting the robot arm upward. For this action, $\operatorname{args}$ is the amount to lift the robot arm, e.g. by 20cm., the robot state for this primitive is the robot end-effector position, k is the index of the lifting primitive in the library, $C_k$ is an end-effector controller, $e_k(s, s^*)=s^*-s$, and $H_k$ is the end-effector controller horizon, which in our setting ranges from 100-300. The target position $s^*$ is computed as $s+[0, 0, \operatorname{args}]$. $f$ moves the robot arm for $H_k$ steps, driving $s$ towards $s^*$. The other primitives are defined in a similar manner; see the appendix for a precise description of each primitive we define. 
\begin{wrapfigure}[10]{R}{0.6\textwidth}
    \begin{minipage}{0.6\textwidth}
    \vspace{-.3in}
    \begin{algorithm}[H]
       	\footnotesize
       	\caption{Parameterized Action Primitive}
       	\label{alg:p3}
       	\begin{algorithmic}[1]
        \Require primitive dependent argument vector $\operatorname{args}$, primitive index $k$, robot state $s$
        \State compute $s^* (\operatorname{args}, s)$
            \For{$i=1,...,H_k$ low-level steps}
                \State $e_i = e_k(s_i,s^*)$ \algorithmiccomment{compute state error}
                \State $a_i = C_k(e_i, s_i)$ \algorithmiccomment{compute torques}
                \State execute $a_i$ on robot
            \EndFor
       	\end{algorithmic}
    \end{algorithm}
    \end{minipage}
\end{wrapfigure}

Robot action primitives are a function of the robot state, not the world state. The primitives function by reaching set points of the robot state as directed by the policy, hence they are \textit{agent-centric}. This design makes primitives agnostic to camera view, visual distractors and even the underlying environment itself. The RL policy, on the other hand, is \textit{environment centric}: it chooses the primitive and appropriate arguments based on environment observations in order to best achieve the task.
A key advantage of this decomposition is that the policy no longer has to learn \textit{how} to move the robot and can focus directly on \textit{what} it needs to do. Meanwhile, the low-level control need not be perfect because the policy can account for most discrepancies using the arguments. 

One issue with using a fixed library of primitives is that it cannot define all possible robot motions. As a result, we include a dummy primitive that corresponds to the raw action space. The dummy primitive directly takes in a delta position and then tries to achieve it by taking a fixed number of steps. This does not entirely resolve the issue as the dummy primitive operates on the high level horizon for $H_k$ steps when called. Since the primitive is given a fixed goal for $H_k$ steps, it is less expressive than a feedback policy that could provide a changing argument at every low-level step. For example, if the task is to move in a circle, the dummy primitive with a fixed argument could not provide a target state that would directly result in the desired motion without resorting to a significant number of higher level actions, while a feedback policy could iteratively update the target state to produce a smooth motion in a circle. Therefore, it cannot execute every trajectory that a lower level policy could; however, the primitive library as a whole performs well in practice. 

In order to integrate these parameterized actions into the RL setting, we modify the action space of a standard RL environment to involve two operations at each time step: (a) choose a primitive out of a fixed library (b) output its arguments. As in \citet{chitnis2020efficient}, the policy network outputs a distribution over one-hot vectors defining which primitive to use as well as a distribution over all of the arguments for all of the primitives, a design choice which enables the policy network to have a fixed output dimension. After the policy samples an action, the chosen parameterized action and its corresponding arguments are indexed from the action and passed to the environment. The environment selects the appropriate primitive function $f$ and executes the primitive on the robot with the appropriate arguments. After the primitive completes executing, the final observation and sum of intermediate rewards during the execution of the primitive are returned by the environment. We do so to ensure if the task is achieved mid primitive execution, the action is still labelled successful. 

\begin{figure}
    \centering
    \includegraphics[width=\textwidth]{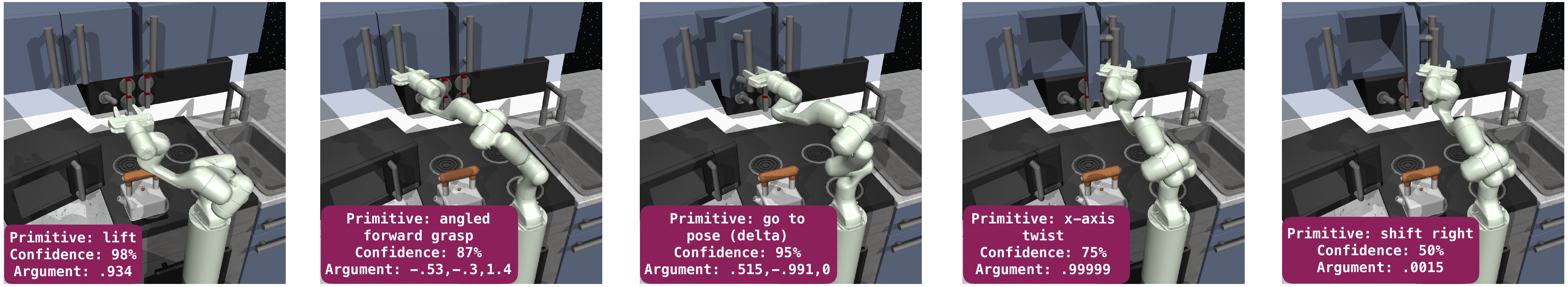}
    \vspace{-0.18in}
    \caption{We visualize an execution of an RL agent trained to solve a cabinet opening task from sparse rewards using robot action primitives. At each time-step, we display the primitive chosen, the policy's confidence in the action choice and the corresponding argument passed to the primitive in the bottom left corner.  }
    \label{fig:example}
    \vspace{-0.1in}
\end{figure}

We describe a concrete example to ground the description of our framework. If we have 10 primitives with 3 arguments each, the higher level policy network outputs 30 dimensional mean and standard deviation vectors from which we sample a 30 dimensional argument vector. It also outputs a 10 dimensional logit vector from which we sample a 10 dimensional one-hot vector. Therefore in total, our action space would be 40 dimensional. The environment takes in the 40 dimensional vector and selects the appropriate argument (3-dimensional vector) from the argument vector based on the one-hot vector over primitives and executes the corresponding primitive in the environment. Using this policy architecture and primitive execution format, we train standard RL agents to solve manipulation tasks from sparse rewards. See Figure~\ref{fig:example} for a visualization of a full trajectory of a policy solving a hinge cabinet opening task in the Kitchen Suite with RAPS\xspace.

\section{Experimental Setup}
\label{sec:experiments}
In order to perform a robust evaluation of robot action primitives and prior work, we select a set of challenging robotic control tasks, define our environmental setup, propose appropriate metrics for evaluating different action spaces, and summarize our baselines for comparison.

\textbf{Tasks and Environments}: $\quad$ We evaluate RAPS\xspace on three simulated domains: Metaworld~\citep{gupta2019relay}, Kitchen~\citep{yu2020meta} and Robosuite~\citep{zhu2020robosuite}, containing 16 tasks with varying levels of difficulty, realism and task diversity (see the bottom half of Fig.~\ref{fig:method_figure}). We use the Kitchen environment because it contains seven different subtasks within a single setting, contains human demonstration data useful for training learned skills and contains tasks that require chaining together up to four subtasks to solve. In particular, learning such temporally-extended behavior is challenging~\citep{gupta2019relay,pertsch2020accelerating,ajay2021opal}. Next, we evaluate on the Metaworld benchmark suite due to its wide range of manipulation tasks and established presence in the RL community. We select a subset of tasks from Metaworld (see appendix) with different solution behaviors to robustly evaluate the impact of primitives on RL.
Finally, one limitation of the two previous domains is that the underlying end-effector control is implemented via a simulation constraint as opposed to true position control by applying torques to the robot.
In order to evaluate if primitives would scale to more realistic learning setups, we test on Robosuite, a benchmark of robotic manipulation tasks which emphasizes realistic simulation and control. We select the block lifting and door opening environments which have been demonstrated to be solvable in prior work~\citep{zhu2020robosuite}. We refer the reader to the appendix for a detailed description of each environment.

\textbf{Sparse Reward and Image Observations} $\quad$ We modify each task to use the environment success metric as a sparse reward which returns $1$ when the task is achieved, and $0$ otherwise.
We do so in order to establish a more realistic and difficult exploration setting than dense rewards which require significant engineering effort and true state information to compute. Additionally, we plot all results against the mean task success rate since it is a directly interpretable measure of the agent's performance.
We run each method using visual input as we wish to bring our evaluation setting closer to real world setups. The higher level policy, primitives and baseline methods are not provided access to the world state, only camera observations and robot state depending on the action.

\begin{figure}[t]
    \centering
    \includegraphics[width=.24\textwidth]{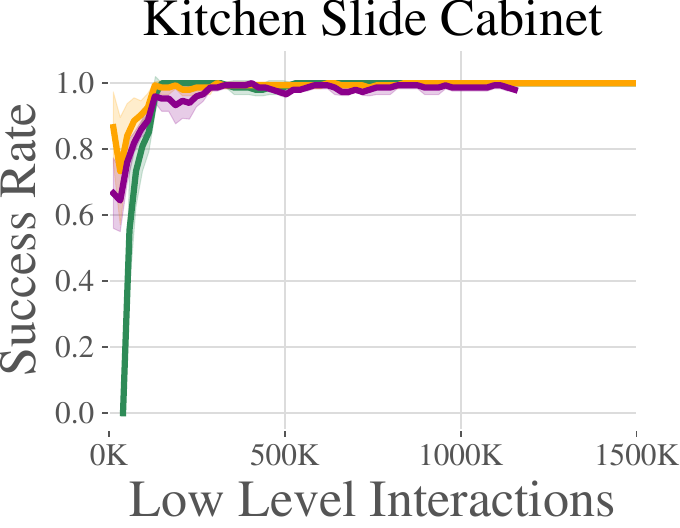}
    \includegraphics[width=.24\textwidth]{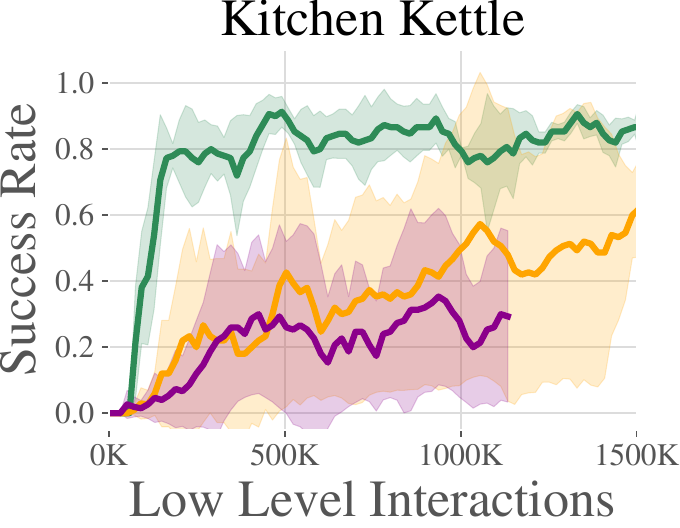}
    \includegraphics[width=.24\textwidth]{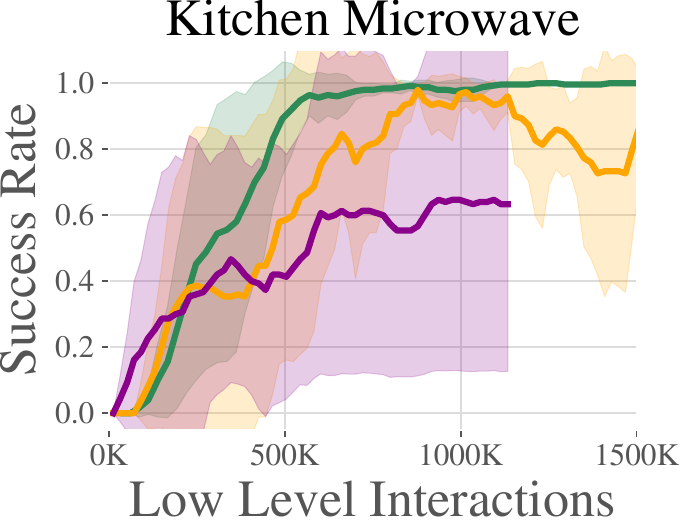} \includegraphics[width=.24\textwidth]{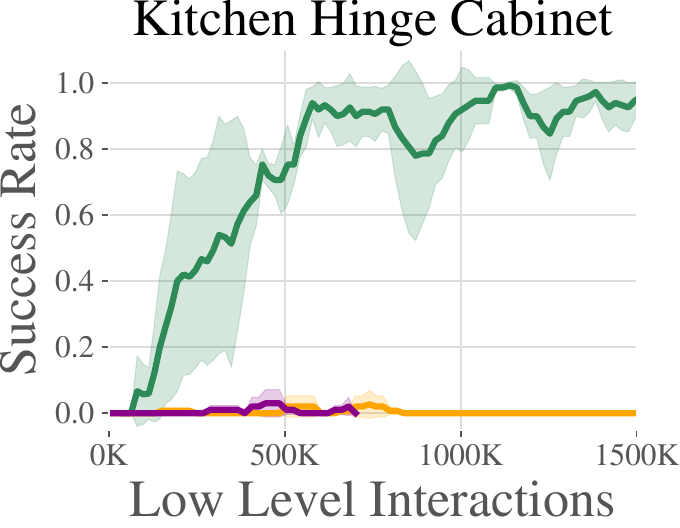}
    \includegraphics[width=\textwidth]{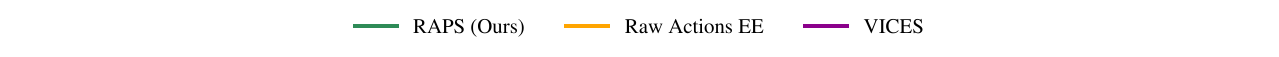}
    \vspace{-0.2in}
    \caption{In the Kitchen environment suite, we run comparisons logging the number of low level interactions of RAPS, Raw actions and VICES. While the methods appear closer in efficiency with respect to low-level actions, RAPS still maintains the best performance across every task. We note that on a real robot, RAPS runs significantly faster than the raw action space in terms of wall-clock time.}
    \vspace{-0.1in}
    \label{fig:low-level-steps}
\end{figure}

\vspace{0.01in}
\textbf{Evaluation Metrics} $\quad$ 
One challenge when evaluating hierarchical action spaces such as RAPS\xspace alongside a variety of different learned skills and action parameterizations, is that of defining a fair and meaningful definition of sample efficiency. We could define one sample to be a forward pass through the RL policy. For low-level actions this is exactly the sample efficiency, for higher level actions this only measures how often the policy network makes decisions, which favors actions with a large number of low-level actions without regard for controller run-time cost, which can be significant. Alternatively, we could define one sample to be a single low-level action output by a low-level controller. This metric would accurately determine how often the robot itself acts in the world, but it can make high level actions appear deceptively inefficient. Higher level actions execute far fewer forward passes of the policy in each episode which can result in faster execution on a robot when operating over visual observations, a key point low-level sample efficiency fails to account for. We experimentally verify this point by running RAPS\xspace and raw actions on a real xArm 6 robot with visual RL and finding that RAPS\xspace executes each trajectory \textbf{32x} times faster than raw actions. We additionally verify that RAPS is efficient with respect to low level steps in Figure \ref{fig:low-level-steps}.

To ensure fair comparison across methods, we instead propose to perform evaluations with respect to two metrics, namely, (a) \textbf{Wall-clock Time}: 
the amount of total time it takes to train the agent to solve the task, both interaction time and time spent updating the agent, and (b) \textbf{Training Steps}: the number of gradient steps taken with a fixed batch size. Wall clock time is not inherently tied to the action space and provides an interpretable number for how long it takes for the agent to learn the task. 
To ensure consistency, we evaluate all methods on a single RTX 2080 GPU with 10 CPUs and 50GB of memory. 
However, this metric is not sufficient since there are several possible factors that can influence wall clock time which can be difficult to disambiguate, such as the effect of external processes, low-level controller execution speed, and implementation dependent details. 
As a result, we additionally compare methods based on the number of training steps, a proxy for data efficiency. The number of network updates is only a function of the data; it is independent of the action space, machine and simulator, making it a non-transient metric for evaluation.
The combination of the two metrics provides a holistic method of comparing the performance of different action spaces and skills operating on varying frequencies and horizons.

\textbf{Baselines} $\quad$  The simplest baseline we consider is the default action space of the environment, which we denote as \textbf{Raw Actions}. 
One way to improve upon the raw action space is to train a policy to output the parameters of the underlying controller alongside the actual input commands. This baseline, \textbf{VICES}~\citep{martinvices}, enables the agent to tune the controller automatically depending on the task.
Alternatively, one can use unsupervised skill extraction to generate higher level actions which can be leveraged by downstream RL.
We evaluate one such method, \textbf{Dyn-E}~\citep{whitney2020dynamicsaware}, 
which trains an observation and action representation from random policy data such that the subsequent state is predictable from the embeddings of the previous observation and action.
A more data-driven approach to learning skills involves organizing demonstration data into a latent skill space. Since the dataset is guaranteed to contain meaningful behaviors, it is more likely that the extracted skills will be useful for downstream tasks. We compare against \textbf{SPIRL}~\citep{pertsch2020accelerating}, a method that ingests a demonstration dataset to train a fixed length skill VAE $z=e(a_{1:H}),a_{1:H} = d(z)$ and prior over skills $p(z|s)$, which is used to guide downstream RL.
\begin{figure}
    \centering
    \includegraphics[width=.24\textwidth]{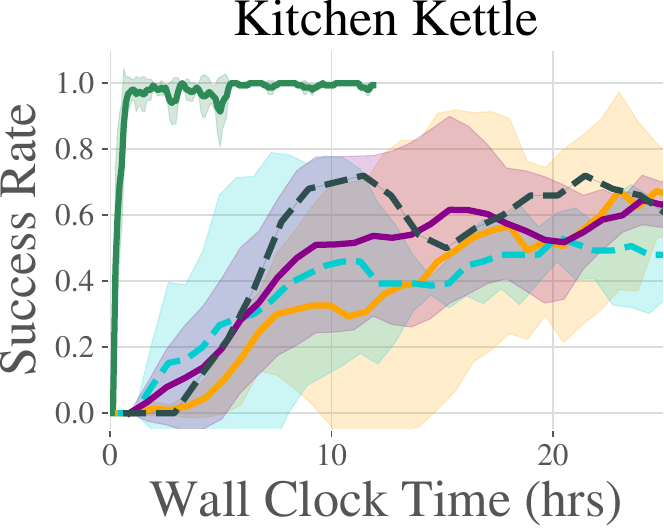}
    \includegraphics[width=.24\textwidth]{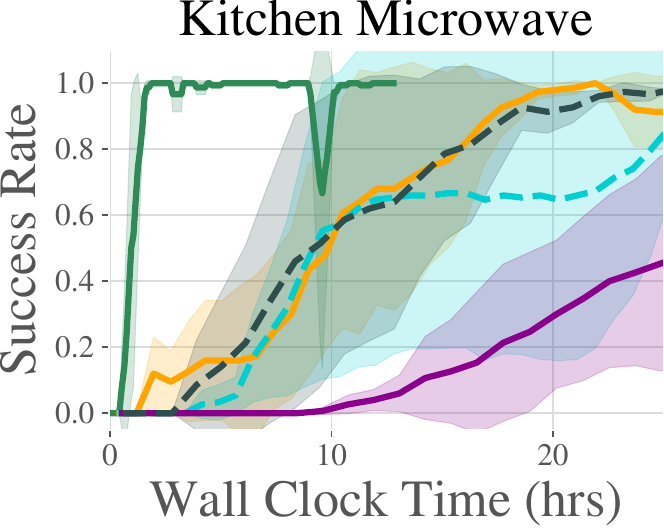}
    \includegraphics[width=.24\textwidth]{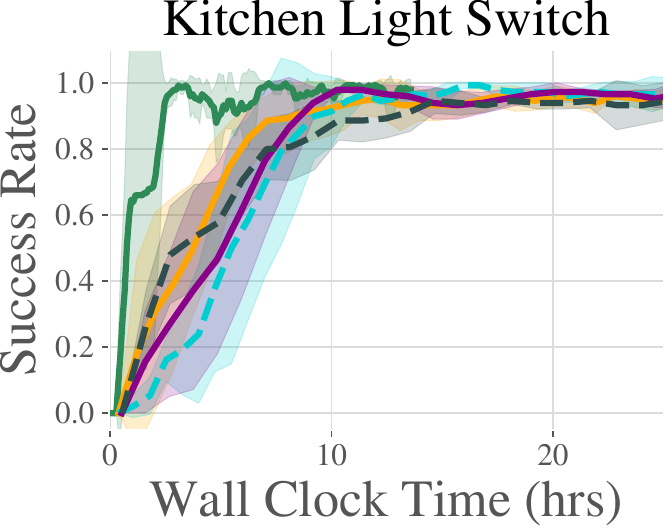}
    \includegraphics[width=.24\textwidth]{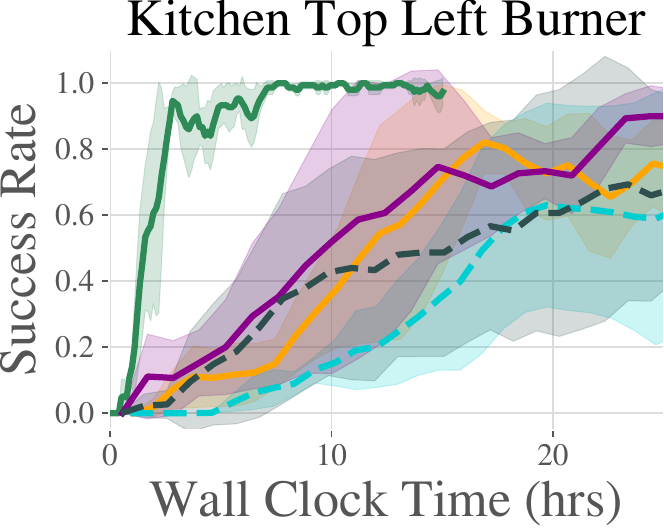}
    \vspace{.2cm}
    \\
    \includegraphics[width=.24\textwidth]{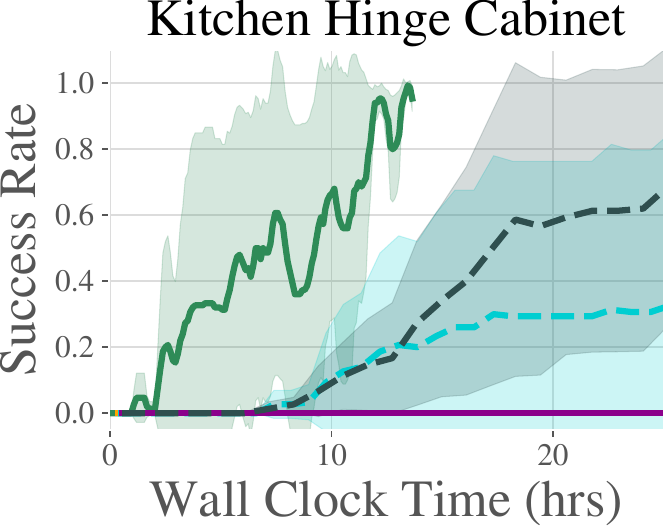}
    \includegraphics[width=.24\textwidth]{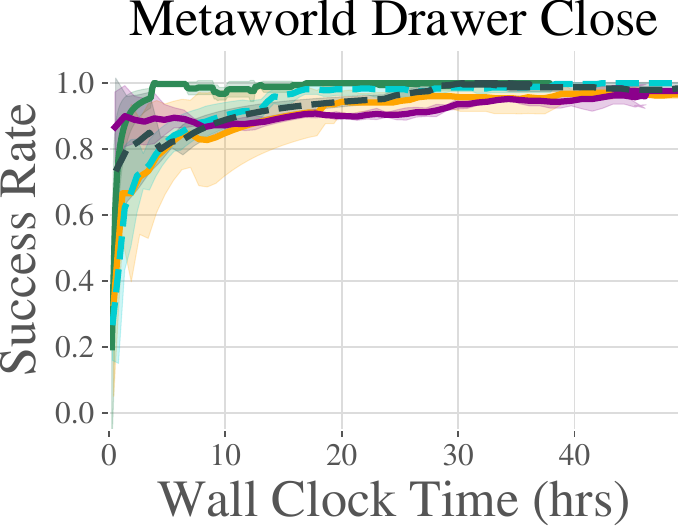}
    \includegraphics[width=.24\textwidth]{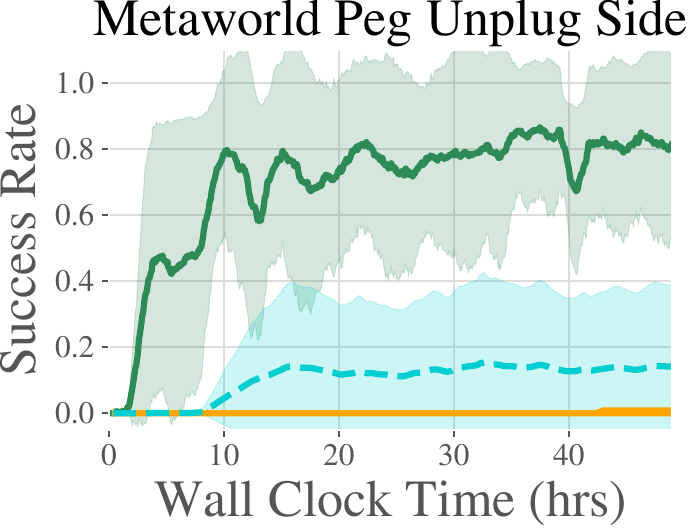}
    \includegraphics[width=.24\textwidth]{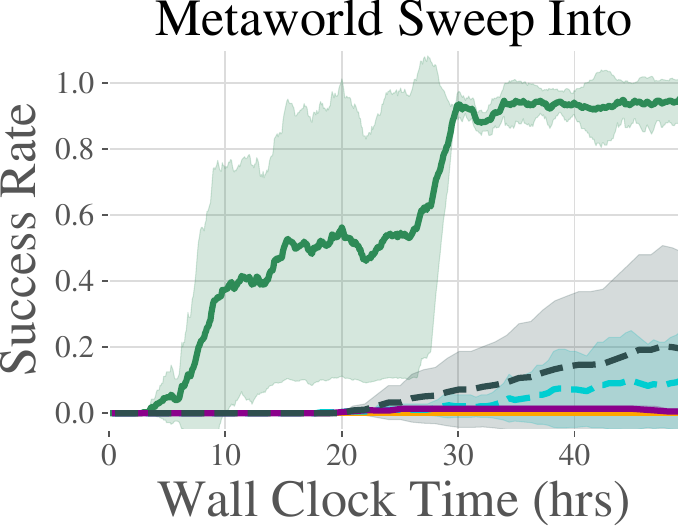}
    \vspace{.2cm}
    \\
    \includegraphics[width=.24\textwidth]{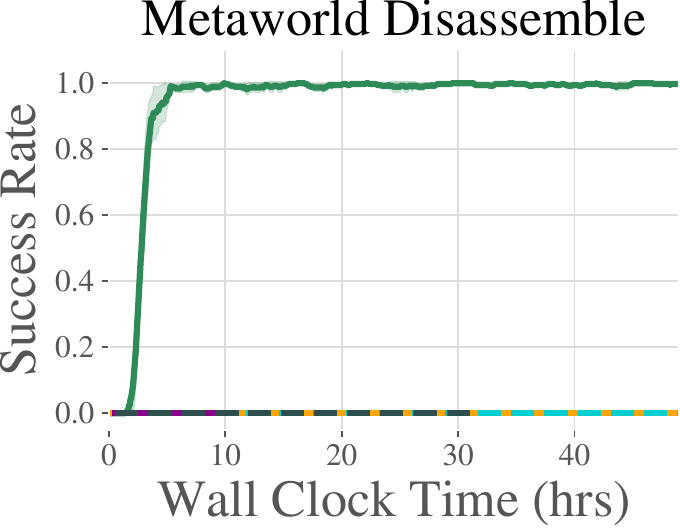}
    \includegraphics[width=.24\textwidth]{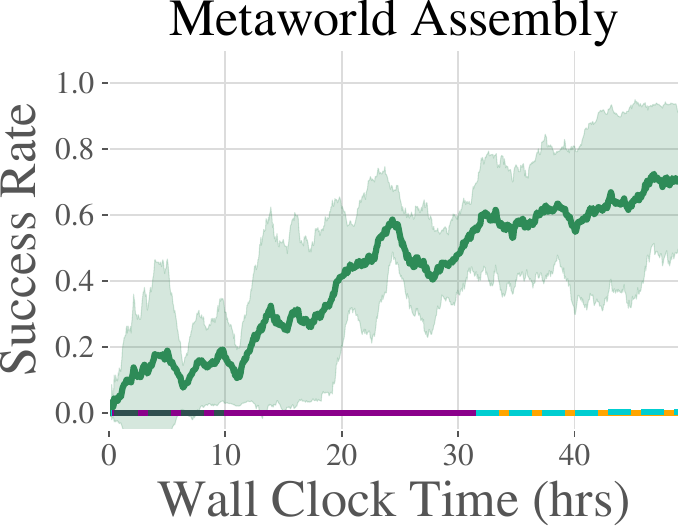}
    \includegraphics[width=.24\textwidth]{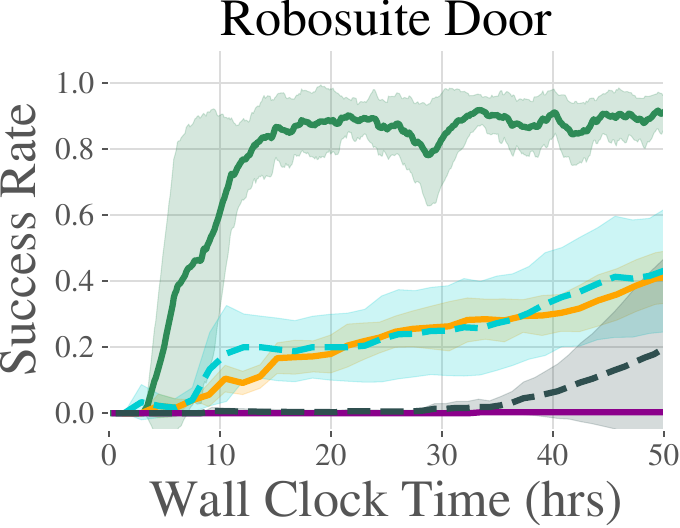}
    \includegraphics[width=.24\textwidth]{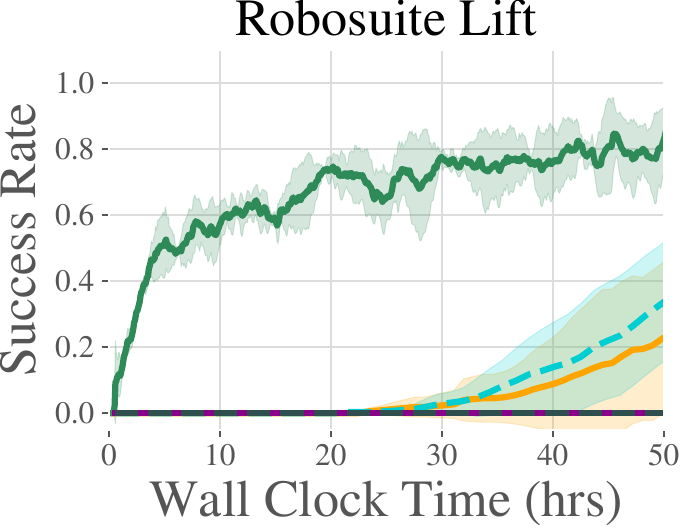}
    \\
    \includegraphics[width=\textwidth]{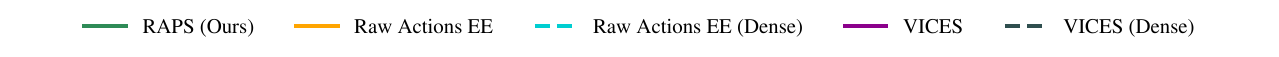}
    \vspace{-0.2in}
    \caption{Comparison of various action parameterizations and RAPS\xspace across all three environment suites\protect\footnotemark using Dreamer as the underlying RL algorithm. RAPS\xspace (green), with sparse rewards, is able to significantly outperform all baselines, particularly on the more challenging tasks, even when they are augmented with dense reward. See the appendix for remaining plots on the \texttt{slide-cabinet} and \texttt{soccer-v2} tasks.}
    \label{fig:action-param-results}
    \vspace{-0.1in}
\end{figure}
\footnotetext{In all of our results, each plot shows a 95\% confidence interval of the mean performance across three seeds.}
Additionally, we compare against \textbf{PARROT}~\citep{singh2020parrot}, which trains an observation conditioned flow model on an offline dataset to map from the raw action space to a latent action space. 
In the next section, we demonstrate the performance of our RAPS\xspace against these methods across a diverse set of sparse reward manipulation tasks. 
\begin{figure}[t]
    \centering
    \includegraphics[width=.24\textwidth]{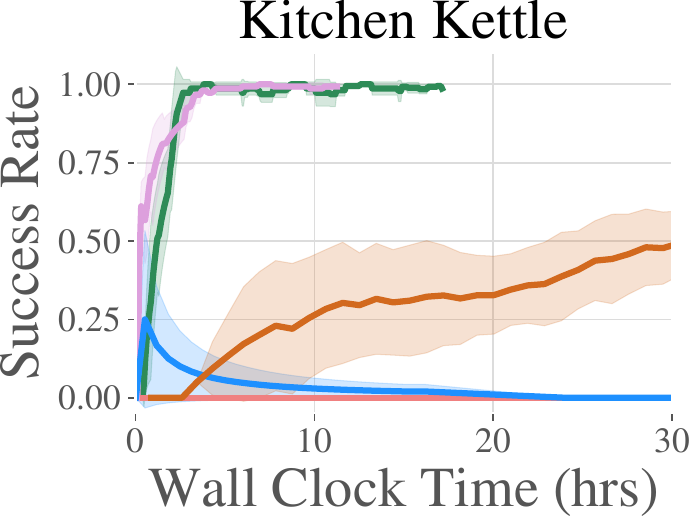}
    \includegraphics[width=.24\textwidth]{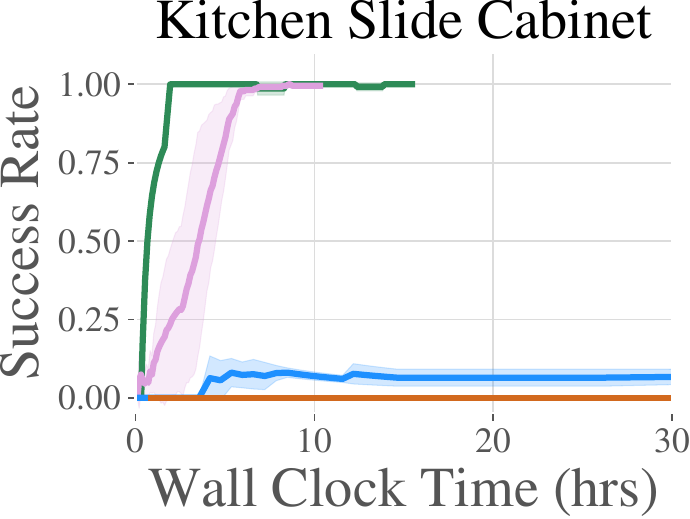}
    \includegraphics[width=.24\textwidth]{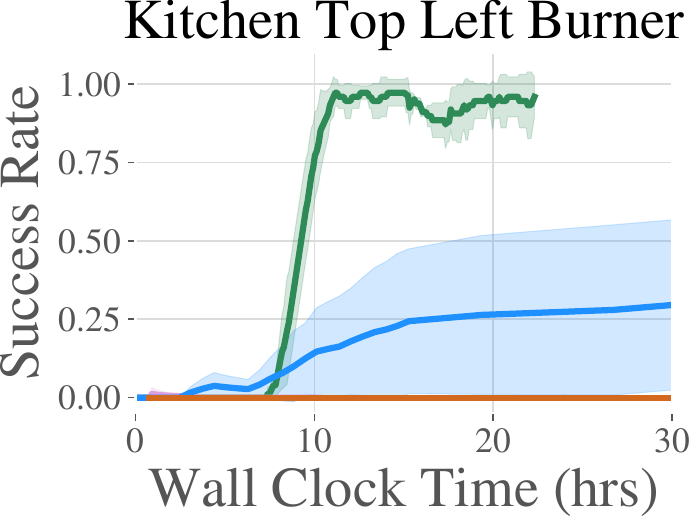}
    \includegraphics[width=.24\textwidth]{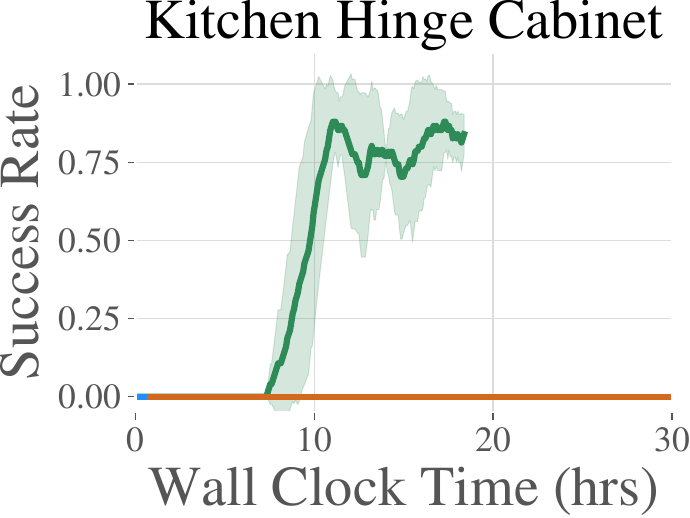}
    \\
    \includegraphics[width=\textwidth]{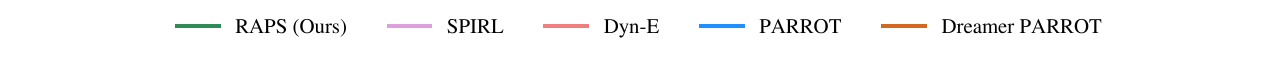}
    \vspace{-0.25in}
    \caption{Comparison of RAPS\xspace and skill learning methods on the Kitchen domain using SAC as the underlying RL algorithm. While SPIRL and PARROT are competitive or even improve upon RAPS\xspace's performance on easier tasks, only RAPS\xspace (green) is able to solve \texttt{top-left-burner} and \texttt{hinge-cabinet}.}
    \label{fig:offline-skill-learn-results}
    \vspace{-0.05in}
\end{figure}

\begingroup
\renewcommand{\arraystretch}{1.25} 
\begin{table}[]
\resizebox{\textwidth}{!}{%
  \begin{tabular}{|l|l|l|l|l|l|l|l|l|l|l|l|l|}
    \hline
    \multirow{2}{*}{RL Algorithm} &
      \multicolumn{2}{c}{Kettle} &
      \multicolumn{2}{c}{Slide Cabinet} &
      \multicolumn{2}{c}{Light Switch} &
      \multicolumn{2}{c}{Microwave} &
      \multicolumn{2}{c}{Top Burner} &
      \multicolumn{2}{c|}{Hinge Cabinet} \\
    & \small{Raw} & \small{RAPS\xspace} & \small{Raw} & \small{RAPS\xspace} & \small{Raw} & \small{RAPS\xspace} & \small{Raw} & \small{RAPS\xspace} & \small{Raw} & \small{RAPS\xspace} & \small{Raw} & \small{RAPS\xspace}  \\
    \hline
    Dreamer & 0.8 & \textbf{.93} & 1.0 & 1.0 & 1.0 & 1.0 & .53 & \textbf{0.8} & .93 & \textbf{1.0} & 0.0 & \textbf{1.0} \\
    \hline
    SAC & .33 & \textbf{0.8} & .67 & \textbf{1.0} & \textbf{.86} & .67 & .33 & \textbf{1.0} & .33 & \textbf{1.0} & 0.0 & \textbf{1.0} \\
    \hline
    PPO & .33 & \textbf{1.0} & .66 & \textbf{1.0} & .27 & \textbf{1.0} & 0.0 & \textbf{.66} & .27 & \textbf{1.0} & 0.0 &\textbf{1.0}\\
    \hline
  \end{tabular}}
  \caption{Evaluation of RAPS\xspace across RL algorithms (Dreamer, PPO, SAC) on Kitchen. We report the final success rate of each method on five evaluation trials trained over three seeds from sparse rewards. While raw action performance (left entry) varies significantly across RL algorithms, RAPS\xspace (right entry) is able to achieve high success rates on \textit{every} task with \textit{every} RL algorithm.}
  \label{tab:rl-algos-results}
  \vspace{-0.1in}
\end{table}
\endgroup
\vspace{-.2in}
\section{Results}
We evaluate the efficacy of RAPS\xspace on three different settings: single task reinforcement learning across Kitchen, Metaworld and Robosuite, as well as hierarchical control and unsupervised exploration in the Kitchen environment. We observe across all evaluated settings, RAPS\xspace is robust, efficient and performant, in direct contrast to a wide variety of learned skills and action parameterizations.
\subsection{Accelerating Single Task RL using RAPS\xspace}
\label{sec:supervised-single}
In this section, we evaluate the performance of RAPS\xspace against fixed and variable transformations of the lower-level action space as well as state of the art unsupervised skill extraction from demonstrations. 
Due to space constraints, we show performance against the number of training steps in the appendix.

\textbf{Action Parameterizations} $\quad$ We compare RAPS\xspace against Raw Actions and VICES using Dreamer~\citep{hafner2020dream} as the underlying algorithm across all three environment suites in Figure~\ref{fig:action-param-results}. Since we observe weak performance on the default action space of Kitchen, joint velocity control, we instead modify the suite to use 6DOF end-effector control for both raw actions and VICES.
We find Raw Actions and VICES are able to make progress on a number of tasks across all three domains, but struggle to execute the fine-grained manipulation required to solve more difficult environments such as \texttt{hinge-cabinet},  \texttt{assembly-v2} and \texttt{disassembly-v2}. The latter two environments are not solved by Raw Actions or VICES even when they are provided dense rewards.
In contrast, RAPS\xspace is able to quickly solve every task from sparse rewards. 

On the kitchen environment, from sparse rewards, no prior method makes progress on the hardest manipulation task: grasping the hinge cabinet and pulling it open to 90 degrees, while RAPS\xspace is able to quickly learn to solve the task. In the Metaworld domain, \texttt{peg-unplug-side-v2}, \texttt{assembly-v2} and \texttt{disassembly-v2} are difficult environments which present a challenge to even dense reward state based RL~\citep{yu2020meta}. However, RAPS\xspace is able to solve all three tasks with \textit{sparse rewards} directly from image input. 
We additionally include a comparison of RAPS\xspace against Raw Actions on all 50 Metaworld tasks with final performance shown in Figure \ref{fig:mt50-bar-plot}
as well as the full learning performance 
in the Appendix.
RAPS\xspace is able to learn to solve or make progress on \textbf{43 out of 50} tasks purely from sparse rewards.
Finally, in the Robosuite domain, by leveraging robot action primitives, we are able to learn to solve the tasks more rapidly than raw actions or VICES, with respect to wall-clock time and number of training steps, demonstrating that RAPS\xspace scales to more realistic robotic controllers.

\textbf{Offline Learned Skills} $\quad$ An alternative point of comparison is to leverage offline data to learn skills and run downstream RL. We train SPIRL and PARROT from images using the kitchen demonstration datasets in D4RL~\citep{fu2021d4rl}, and Dyn-E with random interaction data. We run all agents with SAC as the underlying RL algorithm and extract learned skills using joint velocity control, the type of action present in the demonstrations. See Figure~\ref{fig:offline-skill-learn-results} for the comparison of RAPS\xspace against learned skills. Dyn-E is unable to make progress across any of the domains due to the difficulty of extracting useful skills from highly unstructured interaction data.
In contrast, SPIRL and PARROT manage to leverage demonstration data to extract useful skills; they are competitive or even improve upon RAPS\xspace on the easier tasks such as \texttt{microwave} and \texttt{kettle}, but struggle to make progress on the more difficult tasks in the suite. 
PARROT, in particular, exhibits a great deal of variance across tasks, especially with SAC, so we include results using Dreamer as well. 
We note that both SPIRL and PARROT are limited by the tasks which are present in the demonstration dataset and unable to generalize their extracted skills to other tasks in the same environment or other domains. 
In contrast, parameterized primitives are able to solve \textit{all} the kitchen tasks and are re-used across domains as shown in Figure~\ref{fig:action-param-results}.

\textbf{Generalization to different RL algorithms} $\quad$
A generic set of skills should maintain performance regardless of the underlying RL algorithm. In this section, we evaluate the performance of RAPS\xspace against Raw Actions on three types of RL algorithms: model based (Dreamer), off-policy model free (SAC) and on-policy model free (PPO) on the Kitchen tasks. We use the end-effector version of raw actions as our point of comparison on these tasks.
As seen in Table~\ref{tab:rl-algos-results}, unlike raw actions, RAPS\xspace is largely agnostic to the underlying RL algorithm. 
\begin{figure}[t]
    \centering
    \includegraphics[width=\textwidth]{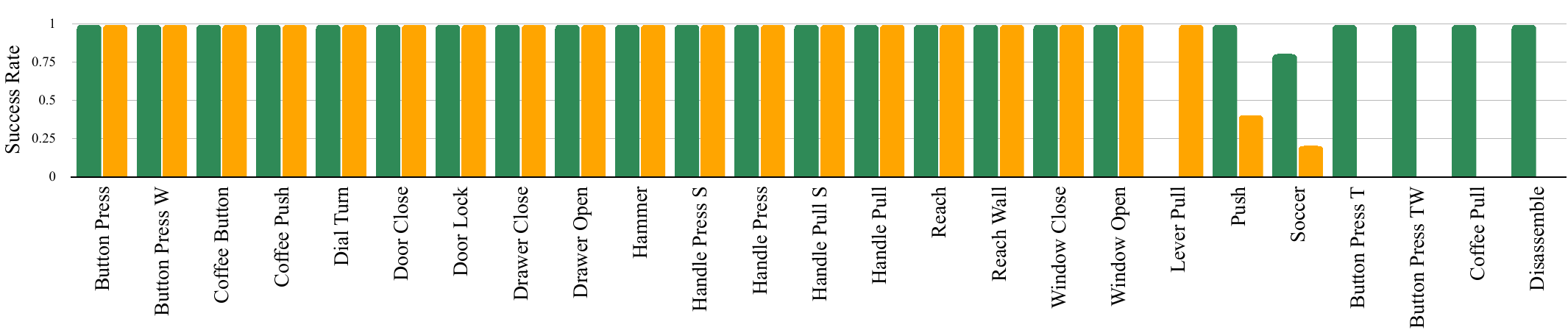}
    \vspace{-.125in}
    \includegraphics[width=\textwidth]{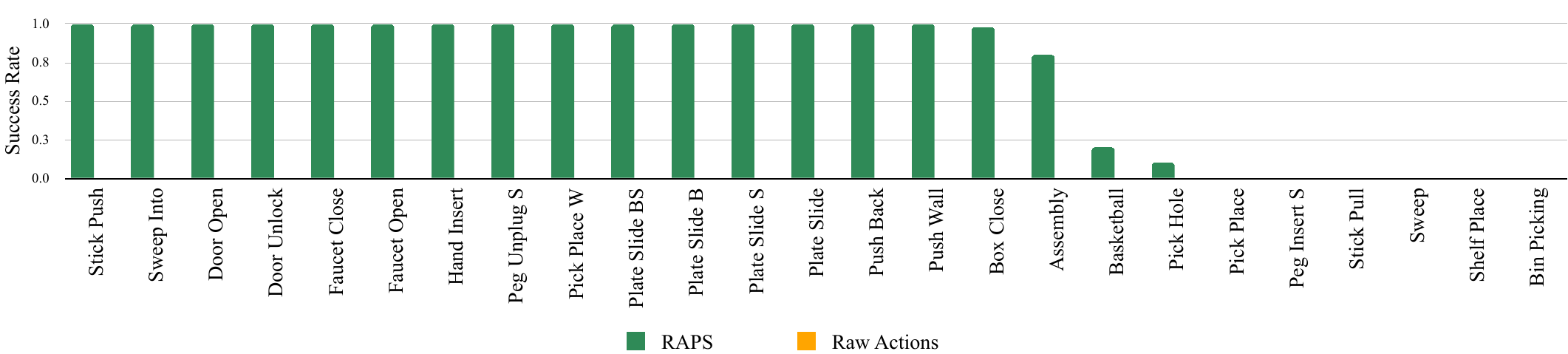}
    \caption{Final performance results for single task RL on the Metaworld domain after 3 days of training using the Dreamer base algorithm. RAPS\xspace is able to successfully learn most tasks, solving 43 out of 50 tasks while Raw Actions is only able to solve 21 tasks.}
    \label{fig:mt50-bar-plot}
\end{figure}
\begin{figure}[t]
    \centering
    \rotatebox[origin=c]{90}{\small{SAC}}
    \begin{subfigure}{.97\textwidth}
         \includegraphics[width=.24\textwidth]{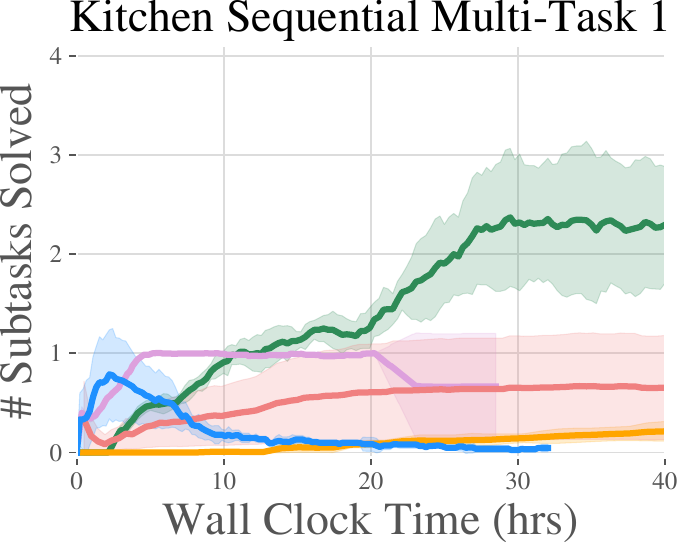}
        \includegraphics[width=.24\textwidth]{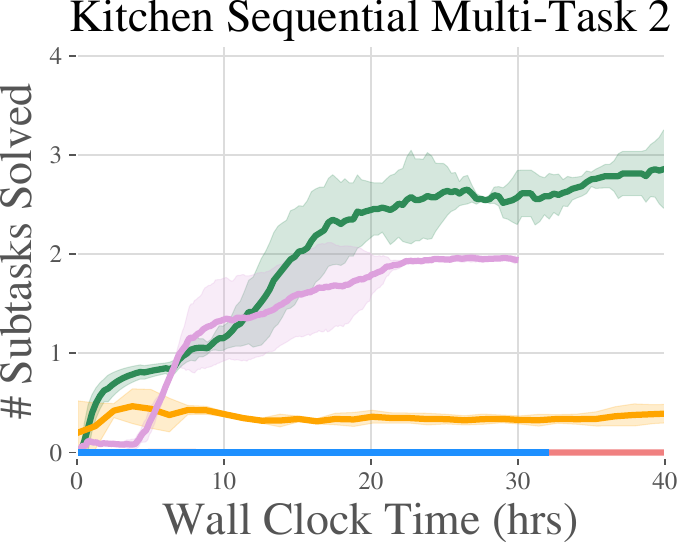}
        \includegraphics[width=.24\textwidth]{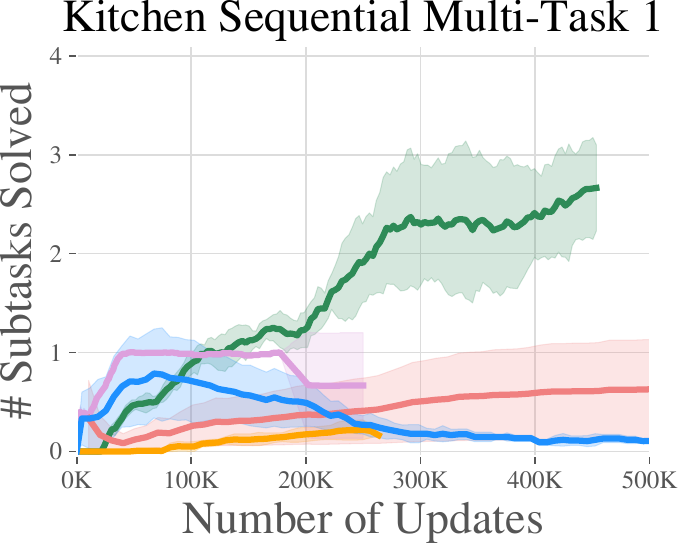}
        \includegraphics[width=.24\textwidth]{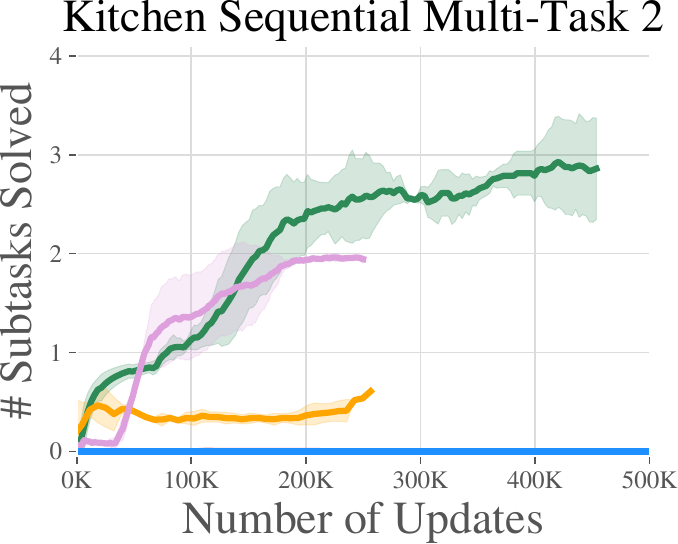}
    \end{subfigure}
    \vspace{.2cm}
    \\
    \rotatebox[origin=c]{90}{\small{Dreamer}}
    \begin{subfigure}{.97\textwidth}
        \includegraphics[width=.24\textwidth]{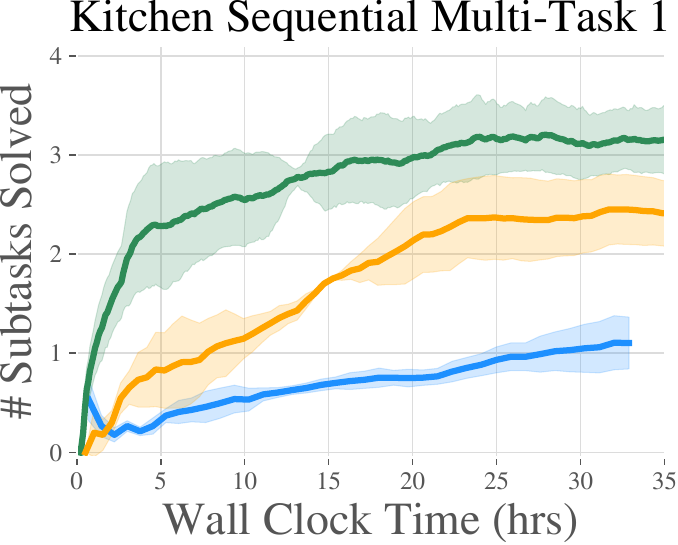}
        \includegraphics[width=.24\textwidth]{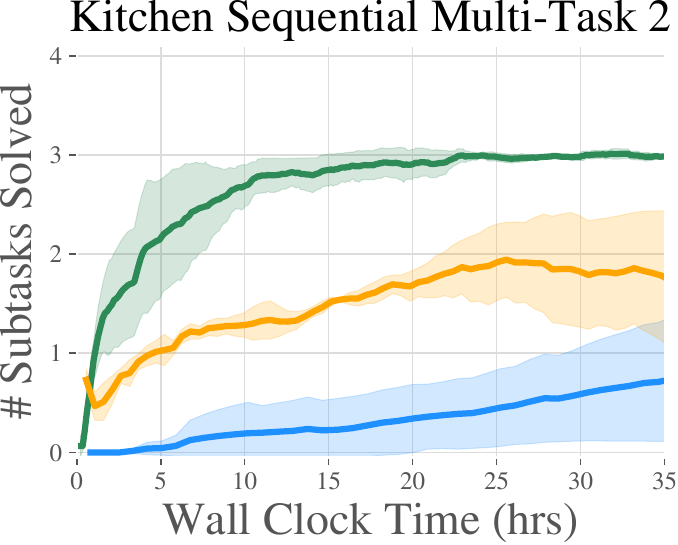}
        \includegraphics[width=.24\textwidth]{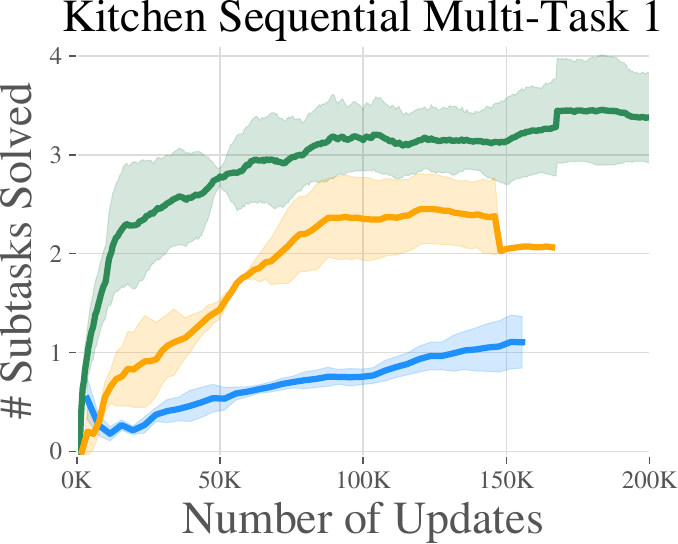}
        \includegraphics[width=.24\textwidth]{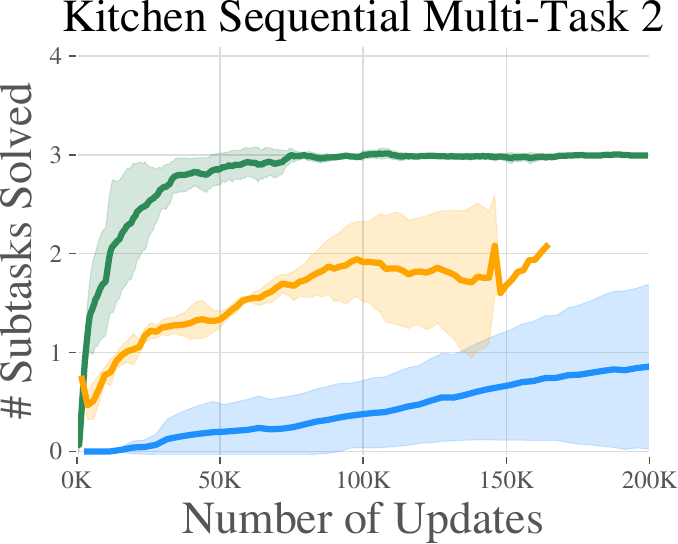}
    \end{subfigure}
    \vspace{.2cm}
    \\
    \rotatebox[origin=c]{90}{\small{PPO}}
    \begin{subfigure}{.97\textwidth}
        \includegraphics[width=.24\textwidth]{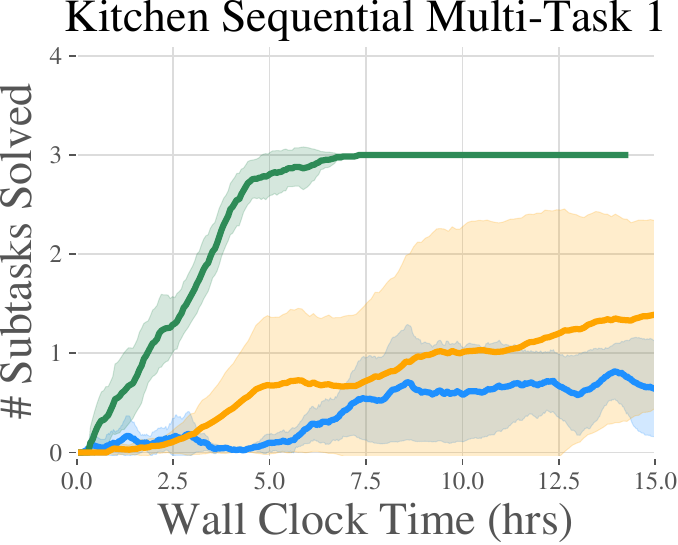}
        \includegraphics[width=.24\textwidth]{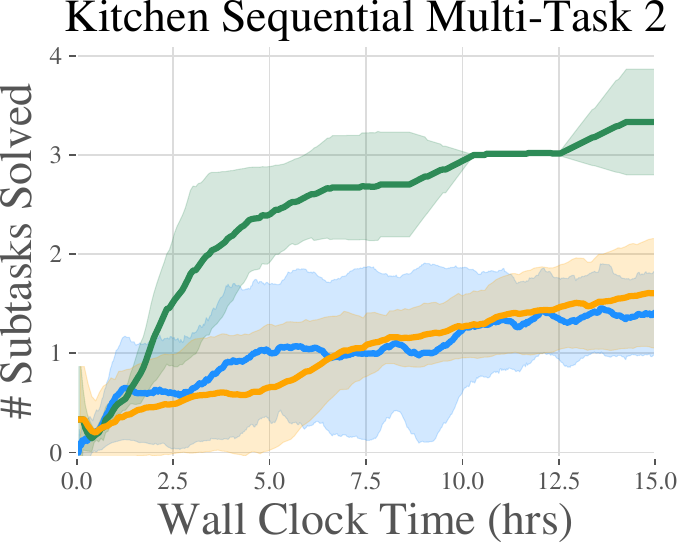}
        \includegraphics[width=.24\textwidth]{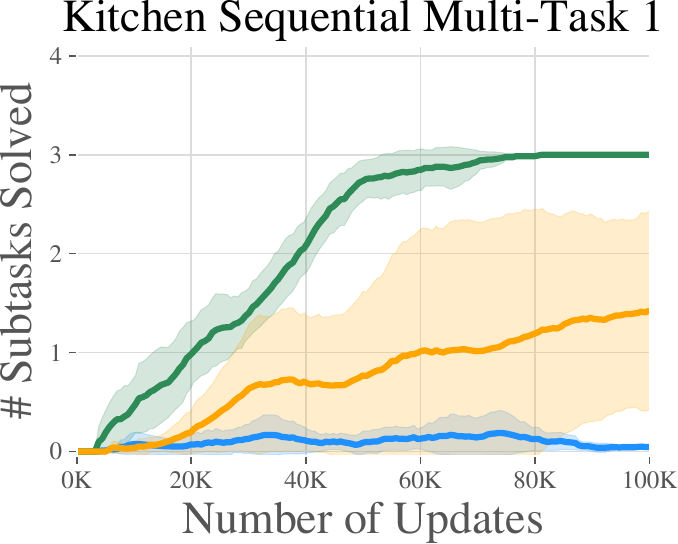}
        \includegraphics[width=.24\textwidth]{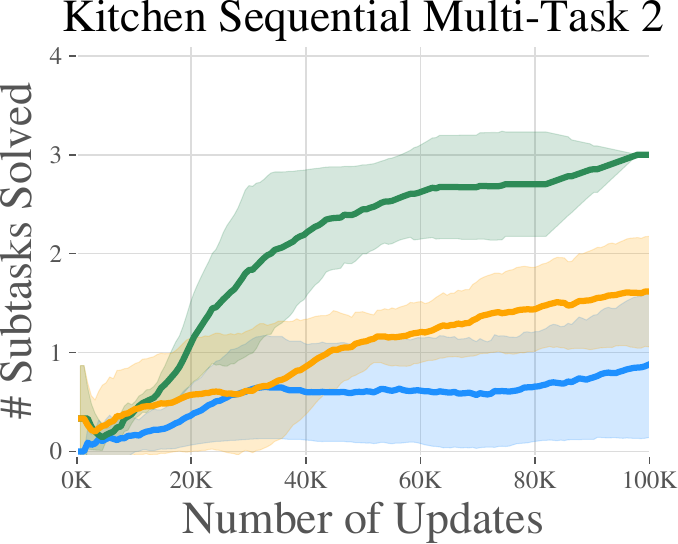}
    \end{subfigure}
    \\
    \includegraphics[width=\textwidth]{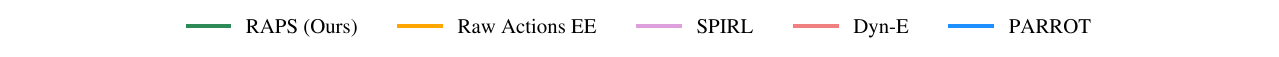}
    \vspace{-0.25in}
    \caption{Learning performance of RAPS\xspace on sequential multi-task RL. Each row plots a different base RL algorithm (SAC, Dreamer, PPO) while the first two columns plot the two multi-task environment results against wall-clock time and the next two columns plot against number of updates, i.e. training steps. RAPS\xspace consistently solves at least three out of four subtasks while prior methods generally fail to make progress beyond one or two.}
    \vspace{-.1in}
    \label{fig:kitchen-multi-task}
\end{figure}
\vspace{-.1in}
\subsection{Enabling Hierarchical Control via RAPS\xspace}
We next apply RAPS\xspace to a more complex setting:  sequential RL, in which the agent must learn to solve multiple subtasks within a single episode, as opposed to one task. We evaluate on the Kitchen Multi-Task environments and plot performance across SAC, Dreamer, and PPO in Figure~\ref{fig:kitchen-multi-task}. 
Raw Actions prove to be a strong baseline, eventually solving close to three subtasks on average, while requiring significantly more wall-clock time and training steps. SPIRL initially shows strong performance but after solving one to two subtasks it then plateaus and fails to improve. PARROT is less efficient than SPIRL but also able to make progress on up to two subtasks, though it exhibits a great deal of sensitivity to the underlying RL algorithm. For both of the offline skill learning methods, they struggle to solve any of the subtasks outside of \texttt{kettle}, \texttt{microwave}, and \texttt{slide-cabinet} which are encompassed in the demonstration dataset. Meanwhile, with RAPS\xspace, across all three base RL algorithms, we observe that the agents are able to leverage the primitive library to rapidly solve three out of four subtasks and continue to improve. This result demonstrates that RAPS\xspace can elicit significant gains in hierarchical RL performance through its improved exploratory behavior.
\subsection{Leveraging RAPS\xspace to enable efficient unsupervised exploration}
\label{sec:unsupervised}

\begin{figure}
    \centering
          \includegraphics[width=.24\textwidth]{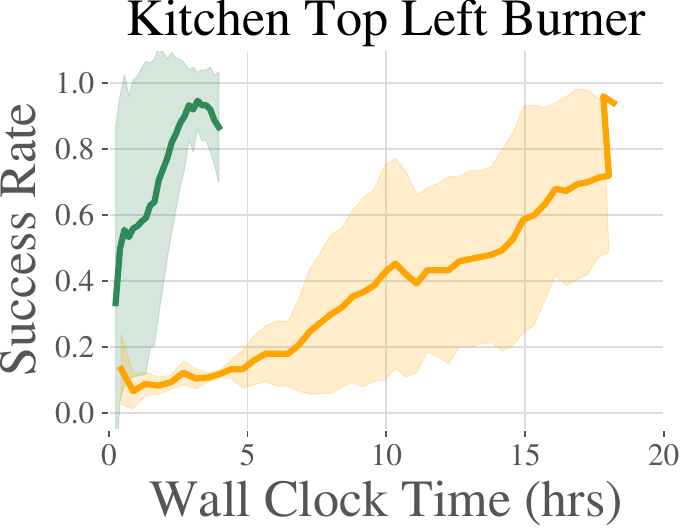}
    \includegraphics[width=.24\textwidth]{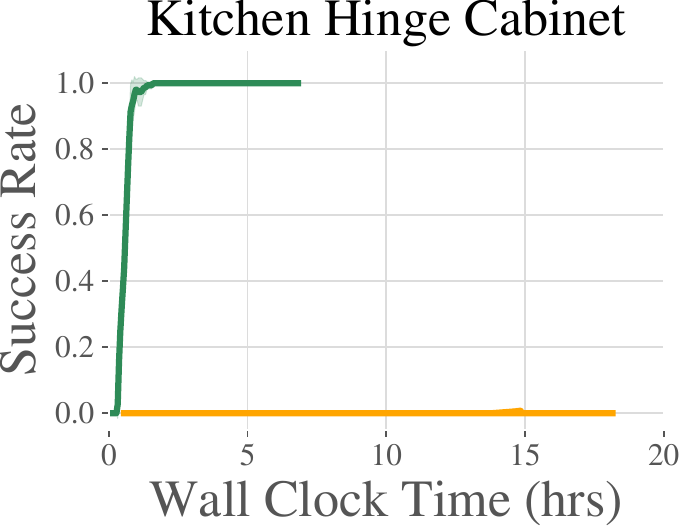}
    \includegraphics[width=.24\textwidth]{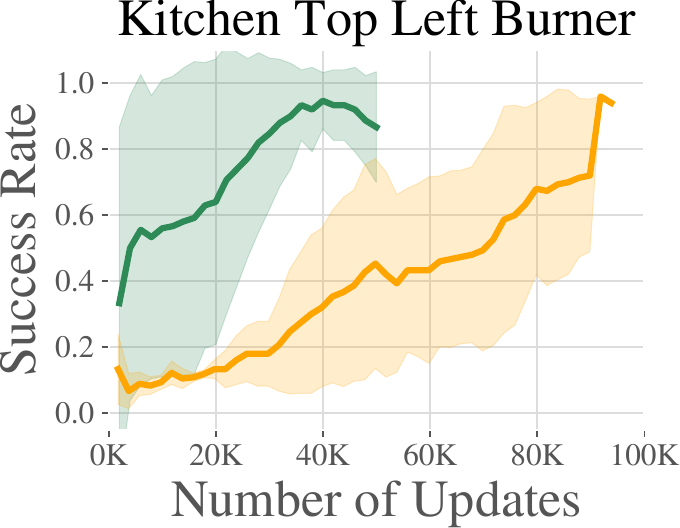}
    \includegraphics[width=.24\textwidth]{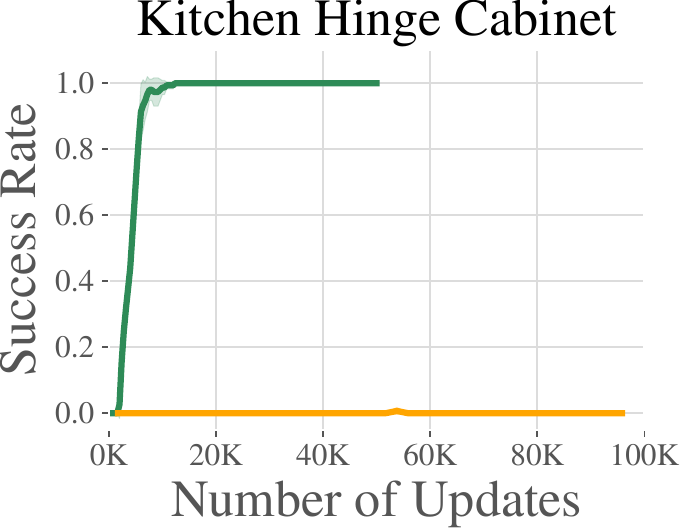}
   
    \includegraphics[width=.3\textwidth]{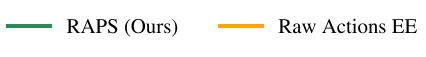}

    \caption{RAPS\xspace significantly outperforms raw actions in terms of total wall clock time and number of updates when fine-tuning initialized from reward free exploration.
    }
    \label{fig:p2expresult}
    \vspace{-0.15in}
\end{figure}
In many settings, sparse rewards themselves can be hard to come by. Ideally, we would be able to train robot without train time task rewards for large periods of time and fine-tune to solve new tasks with only a few supervised labels.
We use the kitchen environment to test the efficacy of primitives on the task of unsupervised exploration. We run an unsupervised exploration algorithm, Plan2explore~\citep{sekar2020planning}, for a fixed number of steps to learn a world model, and then  
fine-tune the model and train a policy using Dreamer to solve specific tasks. 
We plot the results in Figure~\ref{fig:p2expresult} on the \texttt{top-left-burner} and \texttt{hinge-cabinet} tasks. 
RAPS\xspace enables the agent to learn an effective world model that results in rapid learning of both tasks, requiring only \textbf{1 hour of fine-tuning} to solve the \texttt{hinge-cabinet} task. Meanwhile, the world model learned by exploring with raw actions is unable to quickly finetune as quickly. We draw two conclusions from these results, a) RAPS\xspace enables more efficient exploration than raw actions, b) RAPS\xspace facilitates efficient model fitting, resulting in rapid fine-tuning.

\section{Discussion and Limitations}
\label{sec:discussion}
\textbf{Limitations and Future Work} $\quad$
While we demonstrate that RAPS\xspace is effective at solving a diverse array of manipulation tasks from visual input, there are several limitations that future work would need to address. One issue to consider is that of dynamic, fluid motions. Currently, once a primitive begins executing, it will not stop until its horizon is complete, which prevents dynamic behavior that a feedback policy on the raw action space could achieve. 
In the context of RAPS\xspace, integrating the parameterization and environment agnostic properties of robot action primitives with standard feedback policies could be one way to scale RAPS\xspace to more dynamic tasks. Another potential concern is that of expressivity: the set of primitives we consider in this work cannot express all possible motions that robot might need to execute. 
As discussed in Section~\ref{sec:main-method}, we do combine the base actions with primitives via a dummy primitive so that the policy can fall back to default action space if necessary. Future work could improve upon our simple solution.
Finally, more complicated robot morphologies may require significant domain knowledge in order to design primitive behaviors. In this setting, we believe that learned skills with the agent-centric structure of robot action primitives could be an effective way to balance between the difficulty of learning policies to control complex robot morphologies~\citep{andrychowicz2020learning, nagabandi2020deep} and the time needed to manually define primitives.

\textbf{Conclusion} $\quad$
In this chapter, we present an extensive evaluation of RAPS\xspace, which leverages parameterized actions to learn high level policies that can quickly solve robotics tasks across three different environment suites. We show that standard methods of re-parameterizing the action space and learning skills from demonstrations are environment and domain dependent. In many cases, prior methods are unable to match the performance of robot action primitives. While primitives are not a general solution to every task, their success across a wide range of environments illustrates the utility of incorporating an agent-centric structure into the robot action space. 

Moving forward, we aim to address one of the core limitations of RAPS, its expressivity. We do so by flipping the decomposition. Instead of learning high-level information and using structured low-level control, we develop an explicit hierarchical agent that uses high and mid-level planning to perform a majority of the reasoning offline and then learns low-level control rapidly in the loop. As we show in the next chapter, this new framework leads to a dramatic speedup in learning performance and enables RL agents to efficiently solve long-horizon manipulation tasks from visual input.

\clearpage

\chapter{Plan-Seq-Learn: Language Model Guided RL for Solving Long Horizon Robotics Tasks}
\label{chap:psl}

\vspace{-10pt}
\section{Introduction}
\label{sec:intro}
\vspace{-5pt}

In recent years, the field of robot learning has witnessed a significant transformation with the emergence of Large Language Models (LLMs) as a mechanism for injecting internet-scale knowledge into robotics. One paradigm that has been particularly effective is LLM planning over a predefined set of skills~\citep{ahn2022can,singh2023progprompt,huang2022inner,wu2023tidybot}, producing strong results across a wide range of robotics tasks. These works assume the availability of a pre-defined skill library that abstracts away the robotic control problem. They instead focus on designing methods to select the right sequence skills to solve a given task. However, for robotics tasks involving contact-rich robotic manipulation (Fig.~\ref{fig:reel}), such skills are often not available, require significant engineering effort to design or train a-priori or are simply not expressive enough to address the task. 
How can we move beyond pre-built skill libraries and enable the application of language models to general purpose robotics tasks with as few assumptions as possible? 
Robotic systems need to be capable of \textbf{online improvement} over \textit{low-level} control policies while being able to \textbf{plan} over long horizons.

End-to-end reinforcement learning (RL) is one paradigm that can produce complex low-level control strategies on robots with minimal assumptions~\citep{akkaya2019solving,rlscale2023arxiv,handa2022dextreme,kalashnikov2018scalable,kalashnikov2021mt,chen2022visual,agarwal2023legged}. Unlike hierarchical approaches which impose a specific structure on the agent which may not be applicable to all tasks, end-to-end learning methods can, in principle, learn a better representation directly from data.
However, RL methods are traditionally limited to the short horizon regime due to the significant challenge of exploration in RL, especially in high-dimensional continuous action spaces characteristic of robotics tasks. RL methods struggle with longer-horizon tasks in which high-level reasoning and low-level control must be learned simultaneously; effectively decomposing tasks into sub-sequences and accurately achieving them is challenging in general~\citep{sutton1999between,parr1997reinforcement}. 

Our key insight is that LLMs and RL have \textit{complementary} strengths and weaknesses. 
Prior work~\citep{ahn2022can,huang2022language,wu2023tidybot,singh2023progprompt,song2023llm} has shown that when appropriately prompted, language models are capable of leveraging internet scale knowledge to break down long-horizon tasks into achievable sub-goals, but lack a mechanism to produce low-level robot control strategies~\cite{wang2023prompt}, while RL can discover complex control behaviors on robots but struggles to simultaneously perform long-term reasoning~\citep{nachum2018data}. 
However, directly combining the two paradigms, for example, via training a language conditioned policy to solve a new task, does not address the exploration problem. The RL agent must now simultaneously learn language semantics and low-level control.
Ideally, the RL agent should be able to follow the guidance of the LLM, enabling it to learn to efficiently solve each predicted sub-task online. 
How can we connect the abstract language space of an LLM with the low-level control space of the RL agent in order to address the long-horizon robot control problem?
In this chapter, we propose a learning method to solve long-horizon robotics tasks by tracking language model plans using motion planning and learned low-level control.
Our approach, called \textbf{P}lan-\textbf{S}eq-\textbf{L}earn (PSL\xspace), is a modular framework in which a high-level language plan given by an LLM (\textbf{Plan}) is interpreted and executed using motion planning (\textbf{Seq}), enabling the RL policy (\textbf{Learn}) to rapidly learn short-horizon control strategies to solve the overall task. This decomposition enables us to effectively leverage the complementary strengths of each module: language models for abstract planning, vision-based motion planning for task plan tracking as well as achieving robot states and RL policies for learning low-level control. 
Furthermore, we improve learning speed and training stability by sharing the learned RL policy across all stages of the task, using local observations for efficient generalization, and introducing a simple, yet scalable curriculum learning strategy for tracking the language model plan.
To our knowledge, ours is the first work enabling language guided RL agents to efficiently learn low-level control strategies for long-horizon robotics tasks.

Our contributions are: 1) A novel method for long-horizon robot learning that tightly integrates large language models for high-level planning, motion planning for skill sequencing and RL for learning low-level robot control strategies;
 2) Strategies for efficient policy learning from high-level plans, which include policy observation space design for locality, shared policy network and reward function structures, and curricula for stage-wise policy training;
 3) An extensive experimental evaluation demonstrating that PSL\xspace can solve over \textbf{25} long-horizon robotics tasks with up to \textbf{10} stages, outperforming SOTA baselines across four benchmark suites at success rates of \textbf{over 85\%} purely from visual input. PSL\xspace produces agents that solve challenging long-horizon tasks such as NutAssembly at \textbf{96}\% success rate.

\begin{figure}
    \centering
    \vspace{-0.3in} 
    \includegraphics[width=1\linewidth]{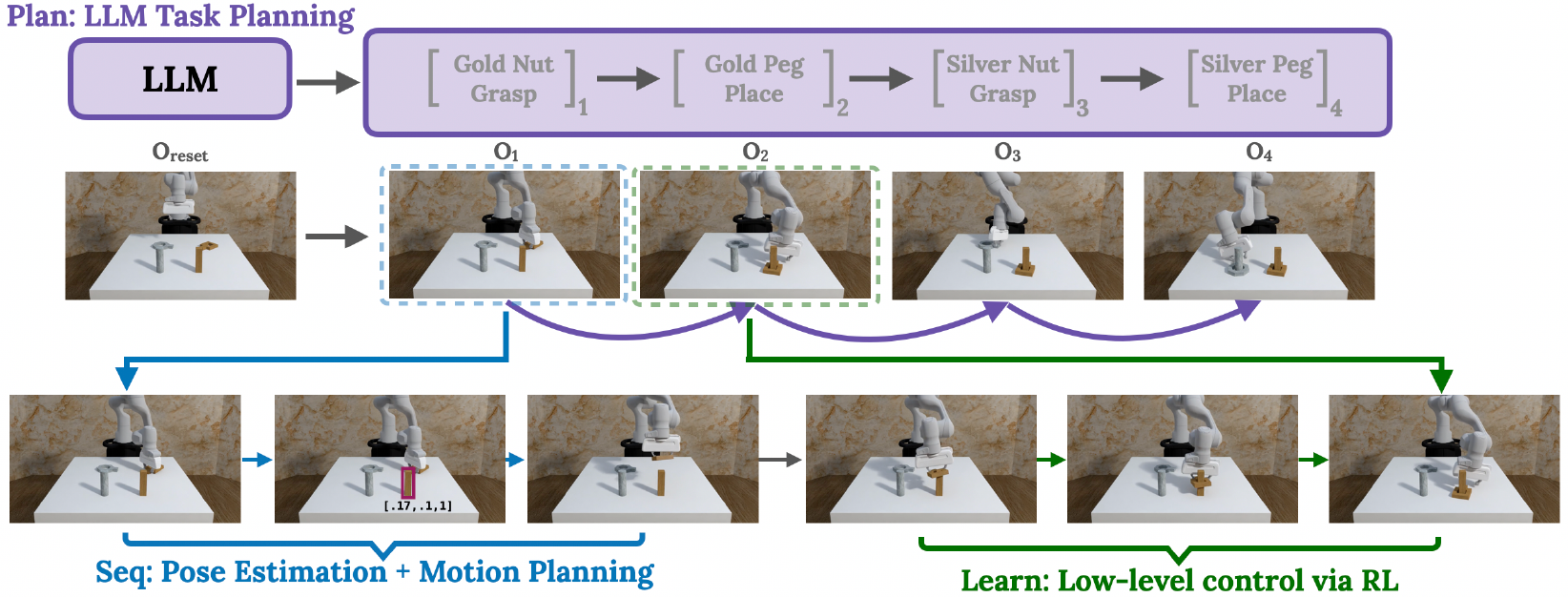}
    \vspace{-5pt}
    \caption{\small \textbf{Long horizon task visualization.} We visualize PSL\xspace solving the NutAssembly task, in which the goal is to put both nuts on their respective pegs. After predicting the high-level plan using an LLM, PSL\xspace computes a target robot pose, achieves it using motion planning and then learns interaction via RL (\textit{third row}).}
    \label{fig:reel}
\end{figure}

\vspace{-7pt}
\section{Related Work}
\label{sec:related work}
\vspace{-10pt}
\textbf{Classical Approaches to Long Horizon Robotics:}
Historically, robotics tasks have been approached via the Sense-Plan-Act (SPA) pipeline~\citep{paul1981robot,whitney1972mathematics,vukobratovic1982dynamics,kappler2018real,murphy2019introduction}, which requires comprehensive understanding of the environment (sense), a model of the world (plan), and a low-level controller (act). Traditional approaches range from manipulation planning~\citep{lozano1984automatic,taylor1987sensor}, grasp analysis~\cite{miller2004graspit}, and Task and Motion Planning (TAMP)~\cite{Garrett2021}, to modern variants incorporating learned vision~\citep{mahler2016dex,mousavian20196,sundermeyer2021contact}. 
Planning algorithms enable long horizon decision making over complex and high-dimensional action spaces. However, these approaches can struggle with contact-rich interactions~\citep{mason2001mechanics,whitney2004mechanical}, experience cascading errors due to imperfect state estimation~\citep{kaelbling2013integrated}, and require significant manual engineering and systems effort to setup~\citep{garrett2020online}. Our method leverages learning at each component of the pipeline to sidestep these issues: it handles contact-rich interactions using RL, avoids cascading failures by learning online, and sidesteps manual engineering effort by leveraging pre-trained models for vision and language. 

\textbf{Planning and Reinforcement Learning:} 
Recent work has explored the integration of motion planning and RL to combine the advantages of both paradigms~\citep{lee2020guided,yamada2021motion,cheng2022guided,xia2020relmogen,james2022q,james2022coarse,liu2022distilling}. GUAPO~\citet{lee2020guided} is similar to the Seq-Learn components of our method, yet their system considers the single-stage regime and is focused on keeping the RL agent in areas of low pose-estimator uncertainty. Our method instead considers long-horizon tasks by encouraging the RL agent to follow a high-level plan given by an LLM using vision-based motion planning. MoPA-RL~\cite{yamada2021motion} also bears resemblance to our method, yet it opts to learn when to use the motion planner via RL, requiring the RL agent to discover the right decomposition of planner vs. control actions on its own. Furthermore, roll-outs of trajectories using MoPA can result in the RL agent choosing to motion plan multiple times in sequence, which is inefficient - one motion planner action is sufficient to reach any position in space. In our method, we instead explicitly decompose tasks into sequences of contact-free reaching (motion planner) and contact-rich environment interaction (RL).

\textbf{Language Models for RL and Robotics}
LLMs have been applied to RL and robotics in a wide variety of ways, from planning~\cite{ahn2022can,singh2023progprompt,huang2022language,huang2022inner,wu2023tidybot,liu2023llm+p,rana2023sayplan,lin2023text2motion}, reward definition~\cite{kwon2023reward,yu2023language}, generating quadrupedal contact-points~\cite{tang2023saytap}, producing tasks for policy learning~\cite{du2023guiding, colas2020language} and controlling simulation-based trajectory generators to produce diverse tasks~\cite{ha2023scalingup}. Our work instead focuses on the online learning setting and aims to leverage language model driven planning to guide RL agents to solve new robotics tasks in a sample efficient manner.
BOSS~\citet{zhangboss} is closest to our overall method; this concurrent work also leverages LLM guidance to learn new skills via RL. Crucially, their method depends on the existence of a skill library and learns skills that are combination of high-level actions. Our method instead efficiently learns \textit{low-level} robot control skills without depending on a pre-defined skill library, by taking advantage of motion planning to track an LLM plan.
\vspace{-8pt}
\section{Plan-Seq-Learn}
\label{sec:method}
\vspace{-10pt}

\begin{figure}
    \centering
    \vspace{-20pt}
    \includegraphics[width=\linewidth]{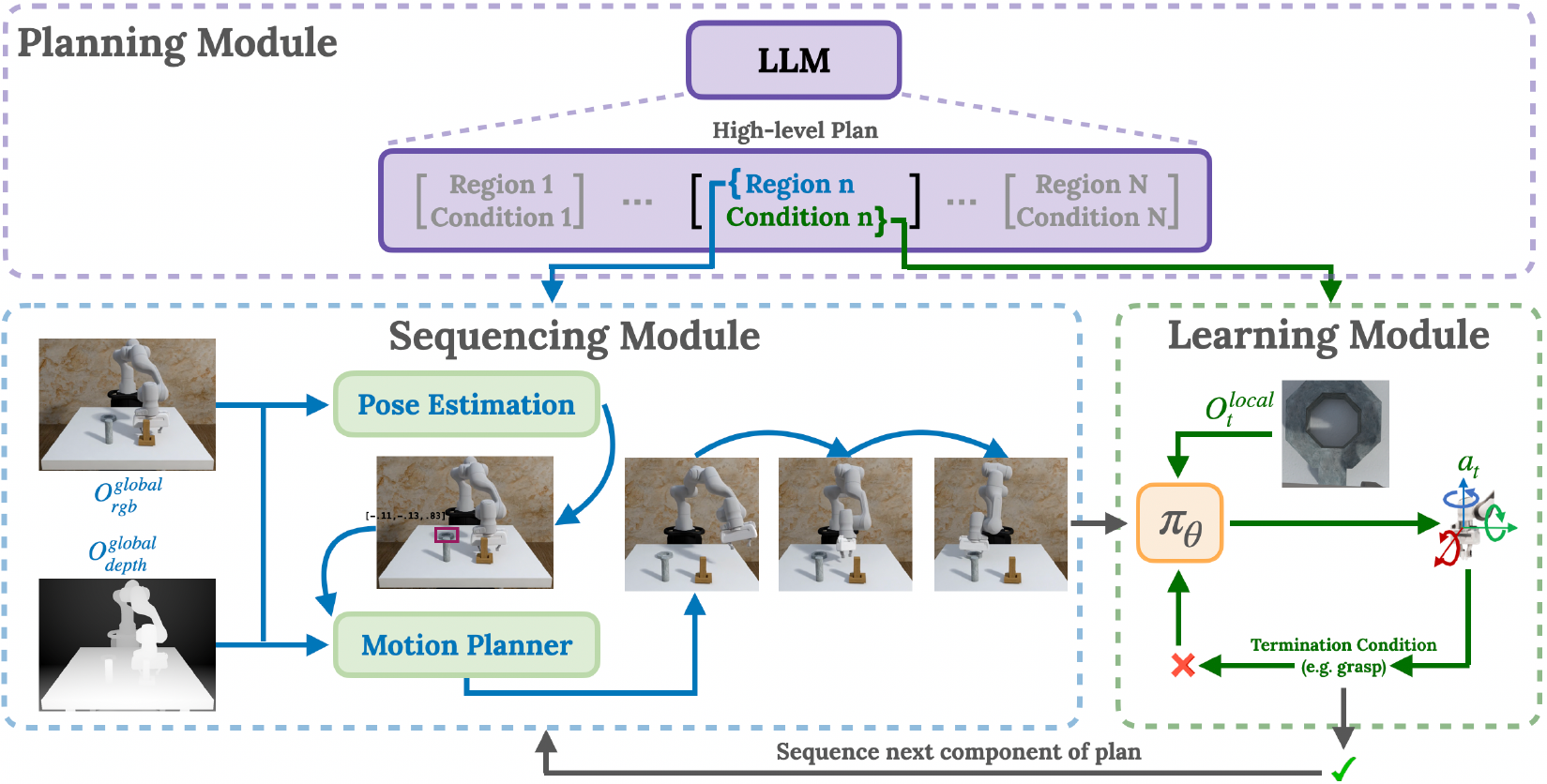}
    \vspace{-15pt}
    \caption{\small \textbf{Method overview.} 
    PSL\xspace decomposes tasks into a list of regions and stage termination conditions using an LLM (\textit{top}), sequences the plan using motion planning (\textit{left}) and learns control policies using RL (\textit{right}). 
    }
    \label{fig:method-overview}
\vspace{-0.1in}
\end{figure} 
In this section, we describe our method for solving long-horizon robotics tasks, PSL\xspace, outlined in Fig.~\ref{fig:method-overview}. Given a text description of the task, our method breaks up the task into meaningful sub-sequences (\textbf{Plan}), uses vision and motion planning to translate sub-sequences into initialization regions (\textbf{Seq}) from which we can efficiently train local control policies using RL (\textbf{Learn}). 

\vspace{-5pt}
\subsection{Problem Setup}
\vspace{-5pt}

We consider Partially Observed Markov Decision Processes (POMDP) of the form $(\mathcal{S}, \mathcal{A}, \mathcal{T}, \mathcal{R}, p_0, \mathcal{O}, p_O, \gamma)$. $\mathcal{S}$ is the set of environment states, $\mathcal{A}$ is the set of actions, $\mathcal{T}(s' \mid s,a)$ is the transition probability distribution, $\mathcal{R}(s,a,s')$ is the reward function, $p_0$ is the distribution over the initial state $s_0 \sim p_0$, $\mathcal{O}$ is the set of observations, $p_O$ is the distribution over observations conditioned on the state $O \sim p_O(O|s)$ and $\gamma$ is the discount factor. In our case, the observation space is the set of all RGB-D (RGB and depth) images. The reward function is defined by the environment. The agent's goal is to maximize the expected sum of rewards over the trajectory, $ \mathbb{E}\left[ \sum_t \gamma^t \mathcal{R}(s_t, a_t, s_{t+1}) \right]$. In our work, we consider POMDPs that describe an embodied robot agent interacting with a scene. We assume that a text description of the task, $g_l$, is provided to the agent in natural language.

\vspace{-5pt}
\subsection{Overview}
\vspace{-5pt}

\begin{wrapfigure}{R}{0.5\textwidth}
    \begin{minipage}{0.5\textwidth}
      \vspace{-23pt}
      \begin{algorithm}[H]
        \caption{Plan-Seq-Learn Overview}
        \label{alg:method}
        \begin{algorithmic}
            \Require LLM, Pose Estimator P, task description $g_l$, Motion Planner MP, low-level horizon $H_l$
            \Statex \textit{Planning Module} 
            \State High-level plan $\mathcal{P}\leftarrow$ Prompt(LLM, $g_l$)
            \For{$p \in \mathcal{P}$}
                \Statex \textit{Sequencing Module}
                \State target region ($t$), termination condition $\leftarrow p$
                \State Compute pose $q_{target} = P(O_t^{global}, t)$  
                \State Achieve pose MP($q_{target}$, $O_t^{global}$)
                \Statex \textit{Learning Module}
                \For{$i=1,...,H_l$}
                    \State Get action $a_t \sim \pi_{\theta}(O^{local}_t)$
                    \State Get next state $O^{local}_{t+1} \sim p(· | s_t, a_t)$.
                    \State Store $(O^{local}_t, a_t, O^{local}_{t+1}, r)$ into $\mathcal{R}$
                    \State update $\pi_{\theta}$ using RL
                    \If{$f_{stage}(O^{global})$}
                        \State break
                    \EndIf
                \EndFor
            \EndFor
        \end{algorithmic}
        \label{algo:full_method}
        \end{algorithm}
        \vspace{-20pt}
    \end{minipage}
\end{wrapfigure}
To solve long-horizon robotics tasks, we need a module capable of bridging the gap between zero-shot language model planning and learned low-level control. 
Observe that many tasks of interest can be decomposed into alternating phases of contact-free motion and contact-rich interaction. One first approaches a target region and then performs interaction behavior, prior to moving to the next sub-task. Contact-free motion generation is exactly the motion planning problem. For estimating the position of the target region, we note that state-of-the-art vision models are capable of accurate language-conditioned state estimation~\citep{kirillov2023segment,zhou2022detecting,liu2023grounding,bahl2023affordances,ye2023affordance,labbe2022megapose}.
As a result, we propose a Sequencing Module which uses off-the-shelf vision models to estimate target robot states from the language plan and then achieves these states using a motion planner. From such states, we train interaction policies that optimize the task reward using RL. See Alg.~\ref{algo:full_method} and Fig.~\ref{fig:method-overview} for an overview of our method. 
\vspace{-6pt}
\subsection{Planning Module: Zero-Shot High-level Planning}
\label{sec:plan}
\vspace{-6pt}
Long-horizon tasks can be broken into a series of stages to execute. Rather than discovering these stages using interaction or using a task planner~\cite{fikes1971strips} that may require privileged information about the environment, we use language models to produce natural language plans zero shot without access to the environment. Specifically, given a task description $g_l$ by a human, we prompt an LLM to produce a plan. Designing the plan granularity and scope are crucial; we need plans that can be interpreted by the Sequencing Module, a vision-based system that produces and achieves robot poses using motion planning. As a result, the LLM predicts a target region (a natural language label of an object/receptacle in the scene, e.g. ``silver peg") which can be translated into a target pose to achieve at the beginning of each stage of the plan. 

When the RL policy is executing a step of the plan, we propose to add a stage termination condition (e.g. \textit{grasp}, \textit{place}, \textit{turn}, \textit{open}, \textit{push}) to know the stage is complete and to move onto the next stage. This condition is defined as a function $f_{stage}(O^{global})$ that takes in the current observation of the environment and evaluates a binary success criteria as well as a natural language descriptor of the condition for prompting the LLM (e.g. ‘grasp’ or ‘place’). These stage termination conditions are estimated using visual pose estimates. We describe the stage termination conditions in greater detail in Sec.~\ref{sec:learn} and Appendix~\ref{app:our impl details}. The LLM prompt consists of the task description $g_l$, the list of supported stage termination conditions (which we hold constant across all environments) and additional prompting strings for output formatting. We format the language plans as follows: (``Region 1", ``Termination Condition 1"), ... (``Region N", ``Termination Condition N"), assuming the LLM predicts N stages. Given the prompt, the LLM outputs a natural language plan in the format listed above. Below, we include an example prompt and plan for the Nut Assembly task.
\begin{framed}
\small \textbf{Prompt:} Stage termination conditions: (grasp, place). 
Task description: The gold nut goes on the gold peg and the silver nut goes on the silver peg. Give me a simple plan to solve the task using only the stage termination conditions. Make sure the plan follows the formatting specified below and make sure to take into account object geometry. Formatting of output: a list in which each element looks like: ($<$object/region$>$, $<$stage termination condition$>$). Don't output anything else.
\\
\small \textbf{Plan:} [(``gold nut",``grasp"), (``gold peg", ``place"), (``silver nut", ``grasp"), (``silver peg", ``place")]
\end{framed}

While any language model can be used to perform this planning process, we found that of a variety of publicly available LLMs (via weights or API), only GPT-4~\cite{openai2023gpt4} was capable of producing correct plans across all the tasks we consider. We sample from the model with temperature $0$ for determinism. We also delete components of the plan that contain LLM hallucinations (if present). We provide additional details in Appendix~\ref{app:our impl details} and example prompts in Appendix~\ref{app:llm prompts}. 
\vspace{-6pt}
\subsection{Sequencing Module: Vision-based Plan Tracking}
\label{sec:seq}
\vspace{-6pt}
Given a high-level language plan, we now wish to step through the plan and enable a learned RL policy to solve the task, using off-the-shelf vision to produce target poses for a motion planning system to achieve. At stage X of the high-level plan, the Sequencing Module takes in the corresponding step high-level plan (``Region Y", ``Termination Condition Z") as well as the current global observation of the scene $O^{global}$ (RGB-D view(s) that cover the whole scene), predicts a target robot pose $q_{target}$ and then reaches the robot pose using motion planning. 

\textbf{Vision and Estimation:} 
Using a text label of the target region of interest from the high-level plan and observation $O^{global}$, we need to compute a target robot state $q_{target}$ for the motion planner to achieve. In principle, we can train an RL policy to solve this task (learn a policy $\pi_v$ to map $O^{global}$ to $q_{target}$) given the environment reward function. However, observe that the 3D position of the target region is a reasonable estimate of the optimal policy $\pi_v^*$ for this task: intuitively, we wish to initialize the robot nearby to the region of interest so it can efficiently learn interaction. Thus, we can bypass learning a policy for this step by leveraging a vision model to estimate the 3D coordinates of the target region. We opt to use Segment Anything~\cite{kirillov2023segment} to perform segmentation, as it is capable of recognizing a wide array of objects, and use calibrated depth images to estimate the coordinates of the target region. We convert the estimated region pose into a target robot pose $q_{target}$ for motion planning using inverse kinematics. 

\textbf{Motion Planning:}
Given a robot start configuration $q_0$ and a robot goal configuration $q_{target}$ of a robot, the motion planning module aims to find a trajectory of way-points $\tau$ that form a collision-free path between $q_0$ and $q_{target}$. For manipulation tasks, for example, $q$ represents the joint angles of a robot arm. We can use motion planning to solve this problem directly, such as search-based planning~\cite{Cohen2010}, sampling-based planning~\cite{KuffnerLaValleRRT} or trajectory optimization~\cite{schulman2013finding}. In our implementation, we use AIT*~\cite{strub2020adaptively}, a sampling-based planner, due to its minimal setup requirements (only collision-checking) and favorable performance on planning. For implementation details, please see Appendix~\ref{app:our impl details}. 

Overall, the Sequencing Module functions as the connective tissue between language and control by moving the robot to regions of interest in the plan, enabling the RL agent to quickly learn short-horizon interaction behaviors to solve the task.

\vspace{-6pt}
\subsection{Learning Module: Efficiently Learning Local Control}
\label{sec:learn}
\vspace{-6pt}
Once the agent steps through the plan and achieves states near target regions of interest, it needs to train an RL policy $\pi_{\theta}$ to learn low-level control for solving the task. We train $\pi_{\theta}$ using DRQ-v2~\cite{yarats2021mastering}, a SOTA visual model-free RL algorithm, to produce low-level control actions (joint control or end-effector control) from images. Furthermore, we propose three modifications to the learning pipeline in order to further improve learning speed and stability. 

First, we train a \textit{single} RL policy across all stages, stepping through the language plan via the Sequencing Module, to optimize the task reward function. The alternative, training a separate policy per stage, would require designing stage specific reward functions per task. Instead, our design enables the agent to solve the task using a single reward function by sharing the policy and value functions across stages. This simplifies the training setup and allowing the agent to account for future decisions as well as inaccuracies in the Sequencing Module. For example, if $\pi_{\theta}$ is initialized at a sub-optimal position relative to the target region, $\pi_{\theta}$ can adapt its behavior according to its value function, which is trained to model the full task return $ \mathbb{E}\left[ \sum_t \gamma^t \mathcal{R}(s_t, a_t, s_{t+1}) \right]$.

Second, instead of executing $\pi_{\theta}$ for a fixed number of steps per stage $H_l$, we predict a stage termination condition $f_{stage}(O^{global})$ using the language model and evaluate the condition at every time-step to test if a stage is complete, otherwise it times out after $H_l$ steps. For most conditions, $f_{stage}$ is evaluated by computing the pose estimate of the relevant object and thresholding. This process functions as a form of curriculum learning: only once a stage is completed is the agent allowed to progress to the next stage of the plan. As we ablate in Sec.~\ref{sec:results}, stage termination conditions enable the agent to learn more performant policies by preventing dithering behavior at each stage. As an example, in the nut assembly task shown in Fig.~\ref{fig:reel}, once $\pi_{\theta}$ places the silver nut on the silver peg, the placement condition triggers (by compare the pose of the nut to the peg pose) and the Sequencing Module moves the arm to near the gold peg. 

Finally, as opposed to training the policy using the global view of the scene ($O^{global}$), we train using \textit{local} observations $O^{local}$, which can only observe the scene in a small region around the robot (\textit{e.g.} wrist camera views for robotic manipulation). This design choice affords several unique properties that we validate in Appendix~\ref{app:additional exps}, namely: 1) improved learning efficiency and speed, 2) ease of chaining pre-trained policies. 
Our policies are capable of leveraging local views because of the decomposition in PSL\xspace: the RL policy simply has to learn interaction behaviors in a small region, it has no need for a global view of the scene, in contrast to an end-to-end RL agent that would need to see a global view of the scene to know where to go to solve a task. 
For additional details in regarding the structure and training process of the Learning Module, see Appendix~\ref{app:our impl details}.

\vspace{-5pt}
\section{Experimental Setup}
\label{sec:setup}
\vspace{-7pt}

\subsection{Tasks}
\vspace{-5pt}
We conduct experiments on single and multi-stage robotics tasks across four simulated environment suites (\textbf{Meta-World}, \textbf{Obstructed Suite}, \textbf{Kitchen} and \textbf{Robosuite}) which contain obstructed settings, contact-rich setups, and sparse rewards (Fig.~\ref{fig:task-list}). 
See Appendix~\ref{app:tasks} for additional details. 
    
    \textbf{Meta-World:} ~\cite{yu2020meta} is an RL benchmark with a rich source of tasks. From Meta-World, we select four tasks: \texttt{MW-Disassemble} (removing a nut from a peg), \texttt{MW-BinPick} (picking and placing a cube), \texttt{MW-Assembly} (putting a nut on a peg), \texttt{MW-Hammer} (hammering a nail). \\ \\
\textbf{ObstructedSuite:}~\citet{yamada2021motion} contains tasks that evaluate our agent's ability to plan, move and interact with the environment in the presence of obstacles. It consists of three tasks: \texttt{OS-Lift} (cube lifting in a tall box), \texttt{OS-Push} (push a block surrounded by walls), and \texttt{OS-Assembly} (avoiding obstacles to place table leg at target). \\ \\
    \textbf{Kitchen:}~\cite{gupta2019relay,fu2020d4rl} tests two aspects of our agent: its ability to handle sparse terminal rewards and its long-horizon manipulation capabilities. The single-stage kitchen tasks include \texttt{K-Slide} (open slide cabinet), \texttt{K-Kettle} (push kettle forward), \texttt{K-Burner} (turn burner), \texttt{K-Light} (flick light switch), and \texttt{K-Microwave} (open microwave). The multi-stage Kitchen tasks denote the number of stages in the name and include combinations of the aforementioned single tasks. \\ \\
    \textbf{Robosuite:}~\cite{zhu2020robosuite} contains a wide array of robotic manipulation tasks ranging from single stage (\texttt{RS-Lift}: cube lifting, \texttt{RS-Door}: door opening) to multi-stage (\texttt{RS-NutRound},\texttt{RS-NutSquare}, \texttt{RS-NutAssembly}: pick-place nut(s) onto target peg(s) and \texttt{RS-Bread}, \texttt{RS-Cereal}, \texttt{RS-Milk}, \texttt{RS-Can}, \texttt{RS-CerealMilk}, \texttt{RS-CanBread}: pick-place object(s) into appropriate bin(s)). Robosuite emphasizes realism and fidelity to real-world control, enabling us to highlight the potential of our method to be applied to real systems.
\vspace{-7pt}
\subsection{Baselines} 
\vspace{-5pt}

We compare against two types of baselines, methods that learn from data and methods that perform offline planning. We include additional details in Appendix~\ref{app:our impl details}. 

\textbf{Learning Methods:}
\begin{itemize}
    \item \textbf{E2E:}~\cite{yarats2021mastering} DRQ-v2 is a SOTA model-free visual RL algorithm.
    \item \textbf{RAPS:}~\cite{dalal2021accelerating} is a hierarchical RL method that modifies the action space of the agent with engineered subroutines (primitives). RAPS greatly accelerates learning speed, but is limited in expressivity due to its action space, unlike PSL\xspace.
    \item \textbf{MoPA-RL:}~\cite{yamada2021motion} is similar to PSL\xspace in its integration of motion planning and RL but differs in that it does not leverage an LLM planner; it uses the RL agent to decide when and where to call the motion planner. 
\end{itemize}

\textbf{Planning Methods:}
\begin{itemize}
    \item \textbf{TAMP:}~\cite{garrett2020pddlstream} is a classical baseline that uses a privileged view of the world to perform joint high-level (task planning) and low-level planning (motion planning with primitives) for solving long-horizon robotics tasks. 
    \item \textbf{SayCan:} a re-implementation of SayCan~\cite{ahn2022can} using publicly available LLMs that performs LLM planning with a fixed set of pre-defined skills.
    Following the SayCan paper, we specify a skill library consisting of object picking and placing behaviors 
    using pose-estimation, motion-planning and heuristic action primitives. 
    We do not learn the pick skill as done in SayCan because our setup does not contain a separate set of train and evaluation environments. In this chapter, we evaluate the single-task RL regime in which the agent is tested with held out poses, not held out environments. 
\end{itemize}

\begin{figure}[t]
    \centering
    \vspace{-0.3in} 
    \begin{tabular}{cc}
    \includegraphics[width=.24\textwidth]{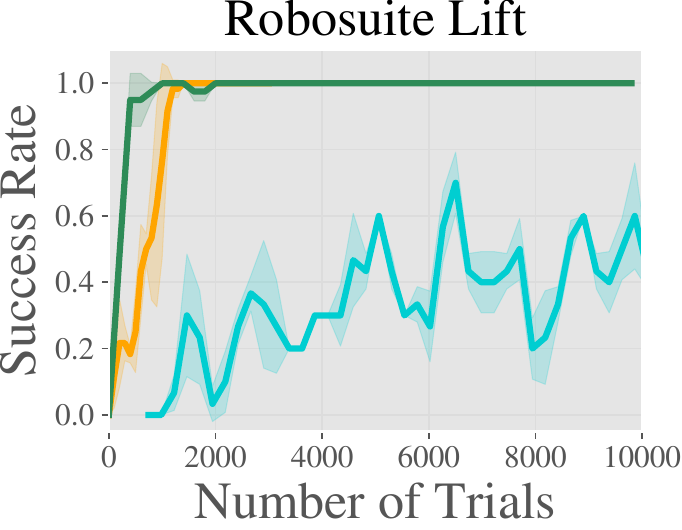}
    \includegraphics[width=.24\textwidth]{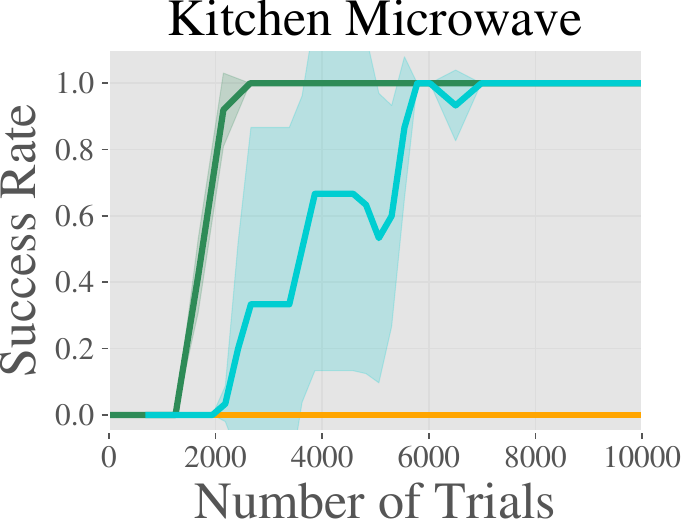} &
    \includegraphics[width=.24\textwidth]{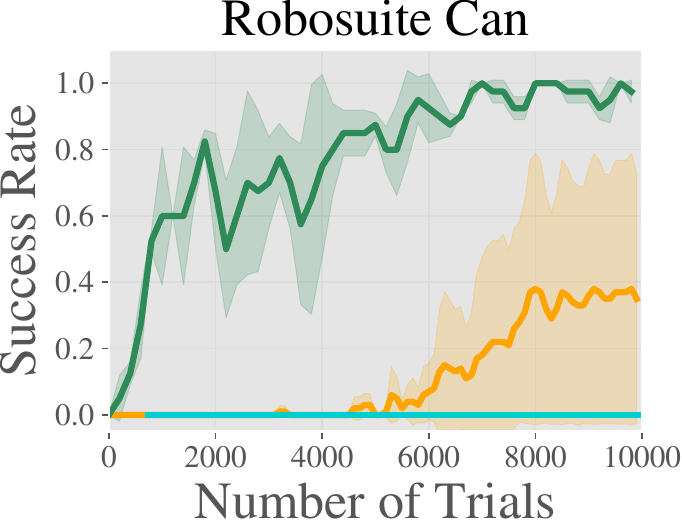}
    \includegraphics[width=.24\textwidth]{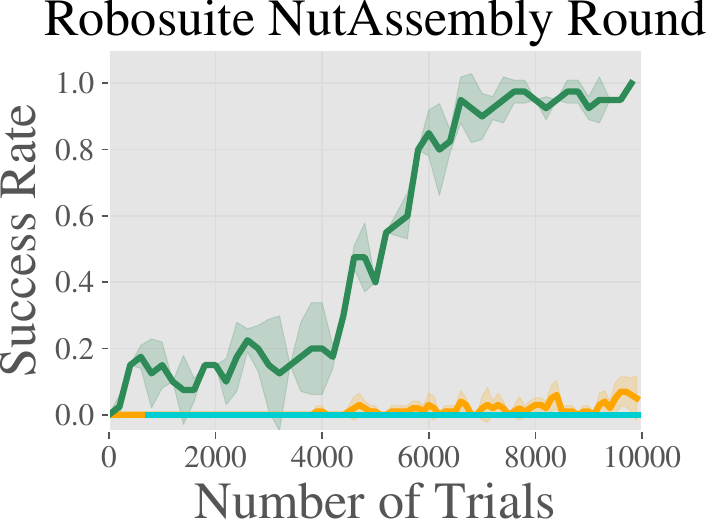}
    \vspace{.2cm}
    \\
    \includegraphics[width=.24\textwidth]{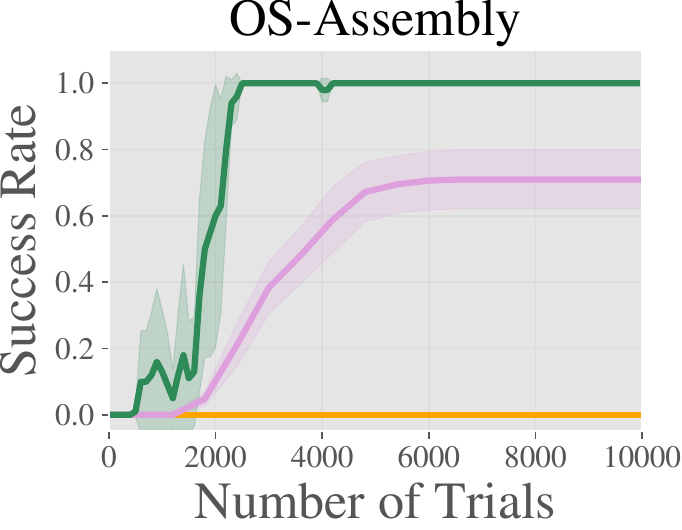}
    \includegraphics[width=.24\textwidth]{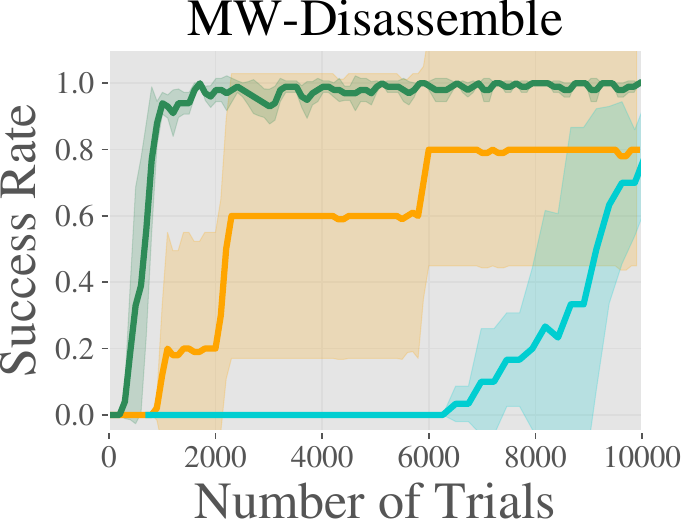} &
    \includegraphics[width=.24\textwidth]{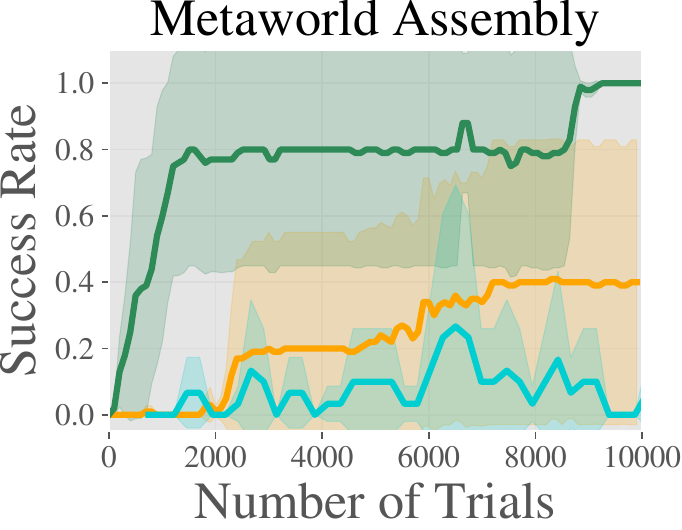}
    \includegraphics[width=.24\textwidth]{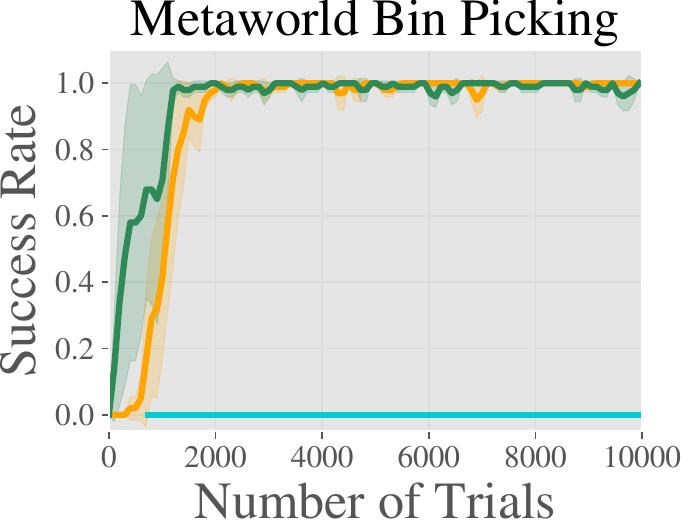}
    \end{tabular}

     \includegraphics[width=.4\textwidth]{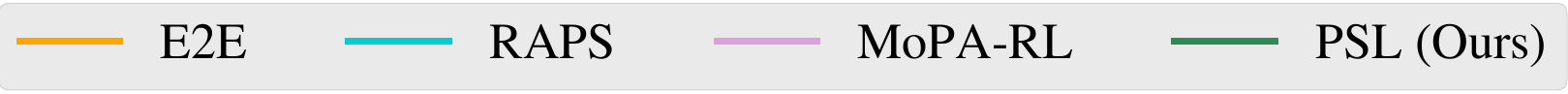}
    \vspace{-5pt}
    \caption{\small \textbf{Sample Efficiency Results.} We plot task success rate as a function of the number of trials. PSL\xspace improves on the sample efficiency of the baselines across each task in Robosuite, Kitchen, Meta-World, and Obstructed Suite. PSL\xspace is able to do so because it initializes the RL policy near the region of interest (as predicted by the Plan and Sequence Modules) and leverages local observations to efficiently learn interaction. Additional learning curves in Appendix~\ref{app:additional exps}.}
    \label{fig: main curves}
    \vspace{-7pt}
\end{figure}

\vspace{-10pt}
\subsection{Experiment details}
\vspace{-5pt}
We evaluate all methods aside from TAMP and MoPA-RL (which use privileged simulator information) using visual input. SayCan and PSL\xspace use $O^{global}$ and $O^{local}$. For E2E and RAPS, we provide the learner access to a single global fixed view observation from $O^{global}$ for simplicity and speed of execution; we did not find including $O^{local}$ improves performance (Fig.~\ref{fig:baseline camera abl})
We measure performance in terms of task success rate with respect to the number of trials. We do so to provide a fair metric for evaluating a variety of different control implementations across PSL\xspace, RAPS, and E2E. Each method is trained for 10K episodes total. We train on each task using the default environment reward function. For each method, we run 7 seeds on every task and average across 10 evaluations. 
\vspace{-5pt}
\section{Results}
\label{sec:results}
\vspace{-10pt}

We begin by evaluating PSL\xspace on a variety of single stage tasks across Robosuite, Meta-World, Kitchen and ObstructedSuite.
Next, we scale our evaluation to the long-horizon regime in which we show that PSL\xspace can leverage LLM task planning to efficiently solve multi-stage tasks. Finally, we perform an analysis of PSL\xspace, evaluating its sensitivity to pose estimation error and stage termination conditions. 
\vspace{-5pt}
\subsection{Learning Results}
\vspace{-6pt}
\setlength{\tabcolsep}{2.4pt}

\begin{table}[t]
\scriptsize
\centering
\begin{tabular}{cccccccc}
\toprule
  & \texttt{RS-Bread}                & \texttt{RS-Can}                  & \texttt{RS-Milk}                 & \texttt{RS-Cereal}                                       & \texttt{RS-NutRound}             & \texttt{RS-NutSquare}  \\
\midrule
\textbf{E2E}                        & .52 $\pm$ .49       & 0.32 $\pm$ .44 & .02 $\pm$ .04      & 0.0 $\pm$ 0.0                          & .06 $\pm$ .13  & 0.02 $\pm$ .045     \\
\rowcolor{Gray}
\textbf{RAPS}                       & 0.0 $\pm$ 0.0       & 0.0 $\pm$ 0.0    & 0.0 $\pm$ 0.0      & 0.0 $\pm$ 0.0                               & 0.0 $\pm$ 0.0     & 0.0 $\pm$ 0.0     \\
\textbf{TAMP}          & 0.9 $\pm$ .01     & 1.0 $\pm$ 0.0  & .85 $\pm$ .06  & \textbf{1.0 $\pm$ 0.0 }                            & 0.4 $\pm$ 0.3   & .35 $\pm$ .2  \\
\rowcolor{Gray}
\textbf{SayCan}                    & .93 $\pm$ .09 & 1.0 $\pm$ 0.0  & 0.9 $\pm$ .05 & .63 $\pm$ .09 & .56 $\pm$ .25 & .27 $\pm$ .21  \\
\midrule
\textbf{PSL\xspace} & \textbf{1.0 $\pm$ 0.0}       & 1.0 $\pm$ 0.0    & \textbf{1.0 $\pm$ 0.0 }     & \textbf{1.0 $\pm$ 0.0 }                              & \textbf{.98 $\pm$ .04} & \textbf{.97 $\pm$ .02} \\
\bottomrule
\end{tabular}
\vspace{-5pt}
\caption{\small \textbf{Robosuite Two Stage Results.} Performance is measured in terms of success rate on two-stage (\textit{2 planner actions}) tasks. SayCan is competitive with PSL\xspace on pick-place style tasks, but SayCan's performance drops considerably ($86.5\%$ to $41.5\%$ on average) on contact-rich tasks involving assembling nuts due to cascading failures. Online learning methods (E2E and RAPS) make little progress on the long-horizon tasks in Robosuite. On the other hand, PSL\xspace is able to solve each task with at least 97\% success rate.}

\label{table:two stage results}
\vspace{-10pt}
\end{table}

\textbf{PSL\xspace accelerates learning efficiency on a wide array of single-stage benchmark tasks.}
For single-stage manipulation, (in which the LLM predicts only a single step in the plan), the Sequencing Module motion plans to the specified region, then hands off control to the RL agent to complete the task. In this setting, we solely evaluate the learning methods since the planning problem is trivial (only one step). We observe improvements in learning efficiency (with respect to number of trials) as well as final performance in comparison to the learning baselines E2E, RAPS and MoPA-RL, across 11 tasks in Robosuite, Meta-World, Kitchen and ObstructedSuite (Fig.~\ref{fig: main curves}, left). For all learning curves, please see the Appendix~\ref{app:additional exps}.
PSL\xspace especially performs well on sparse reward tasks, such as in Kitchen, for which a key challenge is figuring out which object to manipulate and where it is. Additionally, we observe qualitatively meaningful behavior using PSL\xspace: PSL\xspace learns to use the gripper to grasp and turn the burner knob, unlike E2E or RAPS which end up using other joints to flick the burner. 

\textbf{PSL\xspace efficiently solves tasks with obstructions by leveraging motion planning.} 
We now consider three tasks from the Obstructed Suite in order to highlight PSL\xspace's effectiveness at learning control in the presence of obstacles. As we observe in Fig.~\ref{fig: main curves} and Fig.~\ref{fig:single stage results}, PSL\xspace is able to do so efficiently, solving each task within 5K episodes, while E2E fails to make progress. PSL\xspace is able to do so because the Sequencing Module handles the obstacle avoidance implicitly via motion planning and initializes the RL policy in advantageous regions near the target object. In contrast, E2E spends a significant amount of time attempting to reach the object in spite of the obstacles, failing to learn the task. While MoPA-RL is also able to solve many of the tasks, it requires more trials than PSL\xspace even though it operates over \textit{privileged} state input, as the agent must simultaneously learn \textit{when} and \textit{where} to motion plan as well as \textit{how} to manipulate the object.

\textbf{PSL\xspace enables visuomotor policies to learn long-horizon behaviors with up to 10 stages.} 
Two-stage results across Robosuite and Meta-World are shown in Table~\ref{table:two stage results} and Table~\ref{table:metaworld two stage results}, with learning curves in Fig.~\ref{fig: main curves} (right) and Fig.~\ref{fig:metaworld_two_stage_curves}.
On the Robosuite tasks, E2E and RAPS fail to make progress: while they learn to reach the object, they fail to consistently grasp it, let alone learn to place it in the target location. On the Meta-World tasks, the learning baselines perform well on most tasks, achieving similar performance to PSL\xspace due to shaped rewards, simplified low-level control (no orientation changes) and small pose variations. However, PSL\xspace is significantly more sample-efficient than E2E and RAPS as shown in Fig.~\ref{fig:metaworld_two_stage_curves}. TAMP and SayCan are able to achieve high performance across each PickPlace variant of the Robosuite tasks ($93.75\%,86.5\%$ averaged across tasks), as the manipulation skills do not require significant contact-rich interaction, reducing failure skill failure rates. Cascading failures still occur due to the baselines' open-loop nature of execution, imperfect state estimation (SayCan), planner stochasticity (TAMP). Only PSL\xspace is able to achieve perfect performance across each task, avoiding cascading failures by learning from online interaction. 

On multi-stage tasks (involving 3-10 stages), we find that TAMP and SayCan performance drops significantly in comparison to PSL\xspace ($38\%$ and $37\%$ vs. $95\%$ averaged across tasks). For multiple stages, the cascading failure problem becomes all the more problematic, causing all three baselines to fail at intermediate stages, while PSL\xspace is able to learn to adapt to imperfect Sequencing Module behavior via RL. See Table~\ref{table:multi stage results} for a detailed breakdown of the results.

\textbf{PSL\xspace solves contact-rich, long-horizon control tasks such as NutAssembly.}
In these experiments, we show that PSL\xspace can learn to solve contact-rich tasks (\texttt{RS-NutRound}, \texttt{RS-NutSquare}, \texttt{RS-NutAssembly}) that pose significant challenges for classical methods and LLMs with pre-trained skills due to the difficulty of designing manipulation behaviors under continuous contact. By learning an interaction policy whose purpose is to produce locally correct contact-rich behavior, we find that PSL\xspace is effective at performing contact-rich manipulation over long horizons (Table~\ref{table:two stage results}, Table~\ref{table:multi stage results}), outperforming SayCan by a wide margin ($97\%$ vs. $35\%$ averaged across tasks). Our decomposition into contact-free motion generation and contact-rich interaction decouples the \textit{what} (target nut) and \textit{where} (peg) from the \textit{how} (precision grasp and contact-rich place), allowing the RL agent to simply focus on the aspect of the problem that is challenging to estimate a-priori: how to interact with the objects in the appropriate manner.
\vspace{-5pt}
\subsection{Analysis}
\vspace{-6pt}
We now turn to analyzing PSL\xspace, evaluating its robustness to pose estimates and the importance of our proposed stage termination conditions. We include additional analysis of PSL\xspace in Appendix~\ref{app:additional exps}.

\begin{table}
\centering
\scriptsize
\begin{tabular}{@{}cccccccc@{}}
\toprule
                      & \texttt{RS-CerealMilk}   & \texttt{RS-CanBread}  & \texttt{RS-NutAssembly}  & \texttt{K-MS-3} & \textcolor{black}{\texttt{K-MS-5}} & \textcolor{black}{\texttt{K-MS-7}} & \textcolor{black}{\texttt{K-MS-10}}\\
Stages                     & 4           & 4         & 4               & 3            & \textcolor{black}{5}                                    & \textcolor{black}{7} & \textcolor{black}{10}            \\
\midrule
\textbf{E2E}     & 0.0 / 0.0 & 0.0 / 0.0 & 0.0 / 0.0  & 0.0 / 0.0 & \textcolor{black}{0.0 / 0.0} & \textcolor{black}{0.0 / 0.0} & \textcolor{black}{0.0 / 0.0} \\
\rowcolor{Gray}
\textbf{RAPS}    & 0.0 / 0.0 & 0.0 / 0.0 & 0.0 / 0.0  & .89 / 0.1  & \textcolor{black}{0.0 / 0.0} & \textcolor{black}{0.0 / 0.0} & \textcolor{black}{0.0 / 0.0} \\
\textbf{TAMP}    & .71 / .05 & .72 / .25 & 0.2 / 0.3  & \textbf{1.0 / 0.0} & \textcolor{black}{0.0 / 0.0} & \textcolor{black}{0.0 / 0.0} & \textcolor{black}{0.0 / 0.0} \\
\rowcolor{Gray}
\textbf{SayCan} & .73 / .05 & .63 / .21 & .23 / .21 & \textbf{1.0 / 0.0} & \textcolor{black}{0.0 / 0.0} & \textcolor{black}{0.0 / 0.0} & \textcolor{black}{0.0 / 0.0} \\
\midrule
\textbf{PSL\xspace} & \textbf{.85 $\pm$ .21}   & \textbf{0.9 $\pm$ 0.2} & \textbf{.96 $\pm$ .08}   & \textbf{1.0 $\pm$ 0.0}        & \textcolor{black}{\textbf{1.0 $\pm$ 0.0}}                            & \textcolor{black}{\textbf{1.0 $\pm$ 0.0}} & \textcolor{black}{\textbf{1.0 $\pm$ 0.0}} \\
\bottomrule     
\end{tabular}
\vspace{-5pt}
\caption{\small \textbf{Multistage (Long-horizon) results.} Performance is measublack in terms of mean task success rate at convergence. PSL\xspace is the consistently solves each task, outperforming planning methods by over 70\% on challenging contact-intensive tasks such as NutAssembly.}
\label{table:multi stage results}
\vspace{-10pt}
\end{table}

\begin{table}[h]
\centering
\scriptsize
\begin{tabular}{@{}cccccc@{}}
\toprule
& $\sigma=0$ & $\sigma=0.01$ & $\sigma=0.025$ & $\sigma=0.1$  & $\sigma=0.5$ \\
\midrule 
\textbf{SayCan}  & 1.0 $\pm$ 0.0    & .93 $\pm$ .05   & .27 $\pm$ .12    & 0.0 $\pm$ 0.0       & 0.0 $\pm$ 0.0       \\
\textbf{PSL\xspace} & 1.0 $\pm$ 0.0    & \textbf{1.0 $\pm$ 0.0}       & \textbf{1.0 $\pm$ 0.0 }       & \textbf{.75 $\pm$ .07} & 0.0 $\pm$ 0.0     \\ 
\bottomrule 
\end{tabular}
\vspace{-5pt}
\caption{\small \textbf{Noisy Pose Ablation Results.} We add noise sampled from $\mathcal{N}(0, \sigma)$ to the pose estimates and evaluate SayCan and PSL\xspace. PSL\xspace is able to handle noisy poses by training online with RL, only observing performance degradation beyond $\sigma=0.1$.}
\label{table:pose_sigma_abl}
\vspace{-10pt}
\end{table}

\textbf{PSL\xspace leverages stage termination conditions to learn faster.}
While the target object sequence is necessary for PSL\xspace to plan to the right location at the right time,
in this experiment we evaluate if knowledge of the stage termination conditions is necessary. Specifically, on the \texttt{RS-Can} task, we evaluate the use of stage termination condition checks in PSL\xspace to trigger the next step in the plan versus simply using a timeout of 25 steps. We find that it is in fact critical to use stage termination condition checks to enable the agent to effectively sequence the plan; use of a timeout results in dithering behavior which slows down learning. After 10K episodes we observe a performance improvement of 31\% ($100\%$ vs. $69\%$) by including plan stage termination conditions in our pipeline.

\textbf{PSL\xspace produces policies that are robust to noisy pose estimates.}
In real world settings, there is often noise in pose estimation due to noisy depth values, imperfect camera calibration or even network prediction errors. Ideally, the agent should be adapt to such potential failure modes: open-loop planning methods such as TAMP and SayCan are not well-suited to do so because they do not improve online. In this experiment we evaluate the PSL\xspace's ability to handle noisy/inaccurate poses by leveraging online interaction via RL. On the \texttt{RS-Can} task, we add zero-mean Gaussian noise to the pose, with $\sigma \in {0.01, 0.025, .1, .5}$ and report our results in Table ~\ref{table:pose_sigma_abl}. While SayCan struggles to handle $\sigma > 0.01$, PSL\xspace is able to learn with noisy poses at $\sigma = .1$, at the cost of slower learning performance. 
Neither method performs well at $\sigma=0.5$, however at that point the poses are not near the object and the effect is similar to resetting to a random robot pose in the workspace every episode.

\vspace{-3pt}
\section{Discussion and Limitations}
\label{sec:discussion}
\vspace{-10pt}

\textbf{Limitations} There are several limitations of PSL\xspace which leave scope for future work. 1) We impose a specific structure on the language plans and task solution (go to location X, interact there, so on). While this assumption covers a broad set of tasks as well illustrate in our experimental evaluation, tasks that involve interacting with multiple objects simultaneously or continuous switching between interaction and movement in a fluid manner may not be directly applicable. Future work can explore integrating a more expressive plan structure with the Sequencing Module. 2) Use of motion-planning makes application to dynamic tasks challenging. To that end, research on motion-planner distillation, such as Motion Policy Networks~\cite{fishman2022motion} could enable much faster, reactive behavior. 3) Although the RL agent is capable of adapting pose estimation errors, in the current formulation, there is not much the Learning Module can do if the high-level plan itself is entirely incorrect, or if the Sequencing module misinterprets the language instruction and moves the robot to the wrong object. One extension to address this limitation would be to fine-tune the Plan and Seq Modules online using RL as well, to adapt the large models to the specific environment and reward function. 

\textbf{Conclusions} In this chapter, we propose PSL\xspace, a method that integrates the long-horizon reasoning capabilities of language models with the dexterity of learned RL policies via a skill sequencing module. At the heart of our method lies the decomposition of robotics tasks into sequential phases of contact-free motion generation (using language model planning) and environment interaction. We solve these phases using motion planning (informed by visual pose-estimation) and model-free RL respectively, an approach which we validate via an extensive experimental evaluation. We outperform state-of-the-art methods for end-to-end RL, hierarchical RL, classical planning and LLM planning on over 20 challenging vision-based control tasks across four benchmark environment suites. 

Overall, while this section demonstrates two distinct techniques for integrating modularity with learning. The key idea is to form explicit hierarchies that decompose the agent into learned and non-learned (planner) components. While these methods greatly accelerate the speed of learning, they are still limited largely to the single-task regime. In the next part of this thesis, we explore how to appropriately leverage scale to tackle the questions of generalization and multi-task learning for low-level control. Finally, in the proposed work we will unify these approaches to produce a system capable of high and low-level control generalization.

\clearpage

\clearpage

\part{Scaling via Procedural Scene Generation and Planner Distillation}

\chapter{Imitating Task and Motion Planning with Visuomotor Transformers}
\label{chap:optimus}

\section{Introduction}
\vspace{-5pt}
Large-scale data-driven learning, powered by the Transformer architecture~\cite{vaswani2017attention}, has transformed the fields of natural language processing (NLP) and computer vision (CV). Large models at the scale of billions of parameters, trained on massive corpi~\cite{brown2020language, sun2017revisiting, schuhmann2022laion}
exhibit powerful capabilities such as writing coherently~\citep{brown2020language,chowdhery2022palm}, answering questions~\cite{thoppilan2022lamda}, and image classification~\cite{dosovitskiy2020image,liu2021swin} and generation~\cite{saharia2022photorealistic}.
Although there is recent work applying large Transformers to robot learning~\cite{rt12022arxiv,saycan2022arxiv,reed2022generalist}, the recipe of large-scale data-driven learning and Transformers has not yet achieved the same level of widespread success in robotic manipulation. 
One significant bottleneck 
is a lack of useful data -- data collection is especially challenging because it requires the robot to interact in \textbf{real-time} with the world. Furthermore, not all data is useful: the collected interactions should be \textbf{relevant} for solving manipulation tasks of interest. Finally, for learned policies to be broadly applicable, they require access to a \textbf{diverse} set of task instances, which necessitates a \textbf{scalable} data collection pipeline. 

Prior work has used human teleoperation~\citep{hokayem2006bilateral,argall2009survey,mandlekar2018roboturk,lynch2020learning,lynch2020language,cui2022play,rosete2022latent} to collect large robot manipulation datasets, enabling training large scale models~\citep{jang2022bc,rt12022arxiv}. However, this can require significant human time and labor -- RT-1~\cite{rt12022arxiv} required 1.5 years of data collection.
Other works have used reinforcement learning (RL) -- this has the potential to scale more efficiently via autonomous data collection, but it is prohibitively expensive to run in terms of robot time due to its sample inefficiency~\citep{kalashnikov2018scalable,kalashnikov2021mt,zhu2020ingredients,pong2019skew}, and requires significant computation time and human reward engineering 
\cite{akkaya2019solving,handa2022dextreme}. 
In this chapter, we consider an alternative form of supervision, Task and Motion Planning (TAMP)~\cite{Garrett2021}, which addresses some key limitations of prior data-collection techniques. 
TAMP plans a discrete sequence of objects to interact with and how to manipulate them, and continuous motions that safely and correctly facilitate these interactions.
TAMP supervision is beneficial because it: 1) collects data \textit{autonomously} and 2) \textit{efficiently} generates demonstrations by leveraging privileged information. TAMP can generate supervision on a wide distribution of task instances, producing \textbf{task relevant}, \textbf{diverse}, \textbf{large-scale} datasets for robot-learning. 

However, TAMP on its own requires accurate estimation of the scene geometry and its state, is not reactive, and can spend significant time on planning. 
Instead, we propose to imitate TAMP across a wide range of tasks using closed-loop, visuomotor Transformer policies. As a result, we obtain \textbf{fast-to-execute}, \textbf{reactive} agents that can solve long horizon manipulation tasks \textbf{without state estimation}. 
Furthermore, by training on large, diverse datasets of successful trajectories, we show in our experimental evaluation that large Transformer policies have the capability of improving beyond TAMP performance. 
Finally, we note that while McDonald et al.~\cite{mcdonald2022guided} have also learned closed-loop policies from TAMP supervision, we perform an extensive study of the challenges in imitating TAMP, evaluate models across a wide range of tasks, and demonstrate novel capabilities including high-frequency end-to-end visuomotor control, task plan adaptation and scene generalization. 

However, there existf challenges in imitating TAMP include learning from decisions made based on privileged information and multimodal demonstrations~\citep{robomimic2021,mandlekar2018roboturk}. To address these challenges, we propose \textbf{O}ffline \textbf{P}retrained \textbf{T}AMP \textbf{Im}itation \textbf{S}ystem, or OPTIMUS\xspace, a system for training visuomotor Transformer policies via imitation learning.
\textbf{Our contributions are:} 
\newline\noindent $\bullet$ a novel framework for training visuomotor Transformer policies for high-frequency (30-50Hz) low-level control by taking advantage of TAMP supervision
\newline\noindent $\bullet$ an empirically validated data-generation pipeline and study of the insights required to imitate TAMP
\newline\noindent $\bullet$ strong results demonstrating that our trained policies can solve \textbf{over 300} long-horizon manipulation tasks involving up to \textbf{8 stages} and \textbf{72} different objects, achieving success rates of \textbf{over 70\%}

\begin{figure}
    \centering
    \includegraphics[width=.99\linewidth]{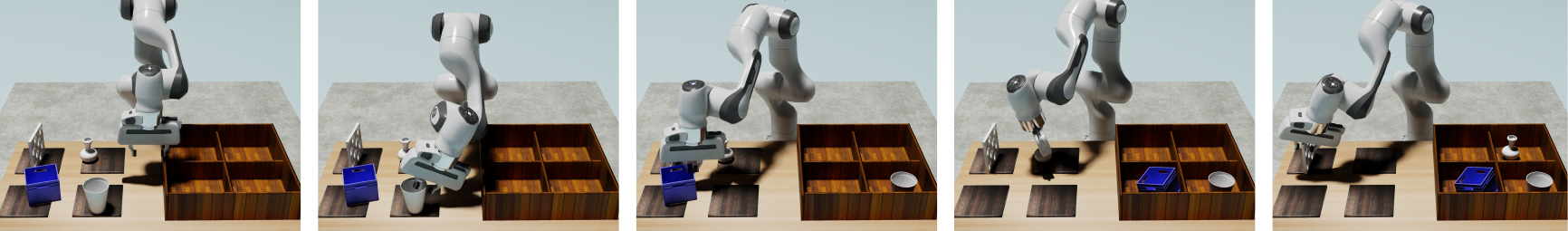}
    \caption{\small
    \textbf{Long-horizon task visualization.} We visualize the initial state and each intermediate pick state for the pick-and-place task.
    Note there is significant variation in geometry across each object, requiring the agent to perform a diverse series of grasps to complete the task.}
    \label{fig:reel}
    \vspace{-7pt}
\end{figure}
\vspace{-10pt}

\section{Related Work}
\label{app:related}
\subsection{Offline Learning from Demonstrations}
Imitation Learning (IL) is a paradigm for training robots to perform manipulation tasks by leveraging a set of expert demonstrations. In this work, we focus on offline learning, in which a policy learns a dataset of demonstrations, without any additional interaction. This is typically done through Behavior Cloning (BC)~\cite{pomerleau1989alvinn}, in which a policy is trained to imitate the actions in the dataset through supervised learning. While this is a simple approach, it has proved incredibly effective for robotic manipulation~\citep{robomimic2021, Ijspeert2002MovementIW, Finn2017OneShotVI, Billard2008RobotPB, Calinon2010LearningAR, zhang2017deep, mandlekar2020learning, wang2021generalization}, particularly when coupled with a large number of demonstrations~\cite{rt12022arxiv, jang2022bc, ahn2022can, ebert2021bridge}. Concurrent work has proposed leveraging Diffusion Models~\cite{sohl2015deep} to train policies via BC~\cite{chi2023diffusion} in order to handle multi-modality of demonstrations. Our work instead focuses on how to best imitate TAMP with Transformers;  Diffusion Policies, in particular their Transformer variants, could be straightforwardly integrated into OPTIMUS\xspace. 

Human supervision is a common source of demonstrations. Several prior works use kinesthetic teaching~\citep{kormushev2011imitation, hersch2008dynamical, niekum2013incremental, akgun2012trajectories}, in which a human manually guides an arm through a task, but this does not scale. 
Many works have leveraged teleoperation systems~\citep{hokayem2006bilateral, argall2009survey, zhang2017deep, mandlekar2018roboturk, mandlekar2019scaling, tung2020learning, wong2022error, jang2022bc, ahn2022can, ebert2021bridge}, in which a human  
remote controls a robot arm to guide it through a task. However, scaling teleoperation is costly because it can require months of data collection and numerous human operators~\citep{rt12022arxiv,jang2022bc,mandlekar2019scaling}. This has motivated the development of intervention-based systems, in which humans provide smaller corrective behaviors to an agent~\citep{kelly2019hg, mandlekar2020human, zhang2016query, hoque2021thriftydagger, hoque2021lazydagger, dass2022pato}, enabling more sample-efficient learning and less operator burden. 
Instead of relying on human operators for supervision, we learn policies from demonstrations provided by a TAMP supervisor, which can generate large, diverse datasets without human supervision.

\subsection{Transformers for Robot Control}

Recent work explores the application of Transformers to controlling robot manipulators. Transformer-based policy architectures such as Gato \cite{reed2022generalist}, PerAct \cite{shridhar2022peract}, VIMA \cite{jiang2022vima}, RT-1 \cite{rt12022arxiv}, ~\citet{dasari2021transformers}, and Behavior Transformer \cite{shafiullah2022behavior} have demonstrated impressive results across a range of robotic manipulation tasks, yet make use of discretization of the input observations and output actions, limiting their applicability to tasks requiring precise manipulation. Additionally, PerAct~\cite{shridhar2022peract} and VIMA~\cite{jiang2022vima} use abstracted actions to ease the learning burden at the cost of expressivity and execution speed. HiveFormer~\cite{guhur2022instruction} is closest to our method in terms of architecture and training protocol but 
also assumes temporally-extended motion planner actions. 
As a result, these systems require privileged knowledge of the geometry of the environment to ensure safety.
In contrast, OPTIMUS\xspace uses a Transformer architecture that is efficient to train and scale, fast-to-execute, consumes raw observations, and outputs low-level control actions.

\subsection{Task and Motion Planning}
Task and Motion Planning (TAMP)~\cite{Garrett2021} addresses controlling a hybrid system through planning a sequence of discrete of manipulation types ({\em task planning}) realized through continuous motions ({\em motion planning}).
TAMP approaches consume kinematic or dynamic models~\cite{toussaint2018differentiable} of individual manipulation types and search over combining them in a manner that achieves a goal.
Classically, these models are engineered; however, recently, they have been learned using methods such as Gaussian Processes~\cite{wang2021learning} or Deep Neural Networks~\cite{qureshi2021nerp,curtis2022long,driess2022learning}.
These mixed engineering-learning TAMP techniques can be quite effective, but they impose a strong human design bias, capping policy performance.
Also, they are too computationally expensive to be run in real-time, preventing them from quickly reacting to new observations.

There has been recent interest in approaches that imitate planning \cite{bhardwaj2017learning,qureshi2019motion,fishman2022motion}; however, these approaches generally focus on single-step motion generation.
The exception is \cite{mcdonald2022guided}, which recently proposed an approach, Guided TAMP, that directly imitates TAMP. 
Our work builds on this direction in several ways.
First, Guided TAMP primarily addresses control from privileged state, while we focus exclusively on visuomotor learning, which requires fewer assumptions.
Second, Guided TAMP proposes a hierarchical policy that first predicts a discrete task-level action and then, conditioned on that action, predicts the next control.
In order for the learner to predict a task-level action, they require a fixed set of ground actions, preventing the same policy from being deployed in tasks, for example, with varying numbers of objects.
In contrast, our Transformer architecture does not explicitly reason about task-level actions and thus does not require grounding and fixing the objects in the scene.
Finally, we identify new considerations when using TAMP as a data generation pipeline.

\begin{figure}
    \centering
    \includegraphics[width=\linewidth]{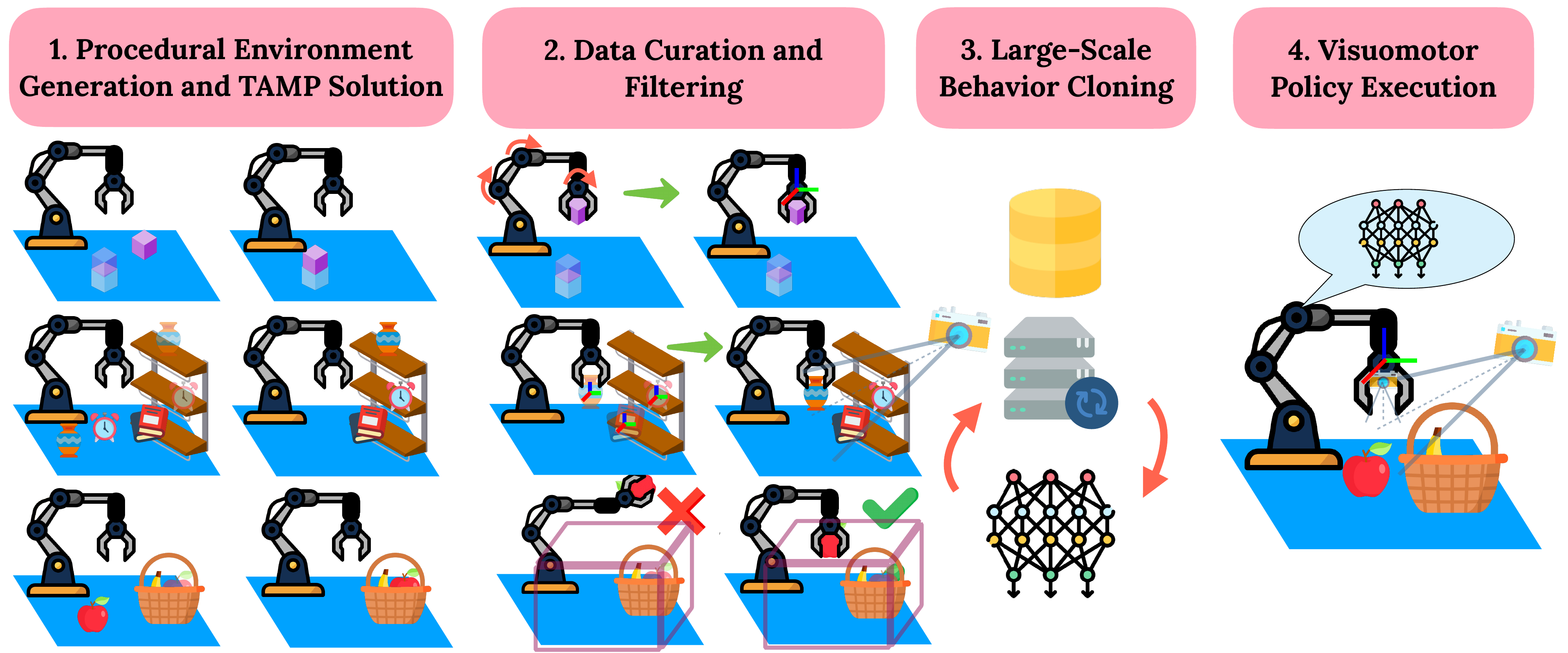}
    \caption{\small \textbf{OPTIMUS\xspace system}. {\em Column 1}: We generate a variety of tasks with differing initial configurations ({\em left}) and goals ({\em right}). {\em Column 2}: We transform TAMP joint space demonstrations to task space ({\em top}), go from privileged scene knowledge in TAMP to visual observations ({\em middle}) and prune TAMP demonstrations based on workspace constraints. {\em Columns 3 and 4}: We perform large-scale behavior cloning using a Transformer-based architecture and execute the visuomotor policies.}
    \label{fig:main-figure}
    \vspace{-.1in}
\end{figure}
\vspace{-5pt}
\section{Designing a TAMP Imitation System}
\label{sec:data pipeline}
\vspace{-5pt}
In this section, we motivate and describe 
our TAMP imitation system, OPTIMUS\xspace. 
We distill a privileged TAMP policy into a neural network in order to obtain policies that do not require access to state information, are fast to execute, and react instantaneously.
To design OPTIMUS\xspace, we apply a TAMP supervisor 
to a procedural problem generator
to produce demonstrations across a diverse range of tasks. However, trajectories produced by TAMP are not necessarily straightforward for an 
agent to imitate, especially when the agent must learn without access to privileged state information.
Consequently, we carefully create a data curation pipeline and couple it with agent design decisions that maximize its ability to learn from TAMP trajectories and solve challenging manipulation tasks.

 We consider tasks with significant variation across 
 objects, poses, and configurations. 
We design four environments: 1) block stacking, 2) single and multi-step pick and place, 3) shelf pick and place, and 4) articulated object manipulation with microwaves. To obtain object diversity, we load objects from the ShapeNet dataset~\cite{chang2015shapenet}.
With a TAMP supervisor and diverse task distribution in place, we describe the data collection pipeline and how we 
use it for policy learning. We begin with a discussion of background knowledge and notation for our method.

\textbf{Background:} We address Partially Observable Markov Decision Processes (POMDP)
$\langle \mathcal{S}, \mathcal{A}, \mathcal{T}, \mathcal{R}, p_0, \Omega, \mathcal{O}, \gamma \rangle$, where $\mathcal{S}$ is the set of environment states, $\mathcal{A}$ is the set of actions, $\mathcal{T}(s' \mid s,a)$ is the transition probability distribution, 
$\mathcal{R}(s,a)$ is the reward function, 
$p_0$ defines the distribution of the initial state $s_0 \sim p_0$,
$\Omega$ is the set of observations,
$\mathcal{O}(o \mid s)$ is the observation distribution,
and $\gamma$ is the discount factor. 
We consider sparse reward POMDPs where ${\cal R}(s, a) \equiv -\mathbbm{1}_{s \notin {\cal S}_*}$ is zero at terminal, goal states ${\cal S}_* \subseteq {\cal S}$ and elsewhere negative one.
Solutions are {\em policies}
$\pi_\theta(o_t, h_t)$ that operate on the history $h_t = (o_{1}, a_{1}, ..., o_{t-1}, a_{t-1})$ of observations $o \in \Omega$ and actions $a \in \mathcal{A}$, outputting the next action $a_t = \pi_\theta(o_t, h_t)$.
The objective is to find a policy $\pi_\theta(o_t, h_t)$ that maximizes the expected policy return $\E[\sum_{t = 1}^\infty {\cal R}(s_t, a_t)]$.
In this context, for behavior cloning, $\pi_\theta(o_t, h_t)$ is trained to regress $a_t$ from $(o_t, h_t)$ from a dataset $\mathcal{D}$ 
consisting of trajectories $\tau_i^n = (o^i_1, a^i_1, ..., o^i_{T_i}, a^i_{T_i})$ produced 
by the expert, in which $i$ is the i-th trajectory in the dataset, $T_i$ is its length and $n$ is the n-th MDP. 

\textbf{Task and Motion Planning:} TAMP algorithms address deterministic and observable, but hybrid, control problems~\cite{Garrett2021}.
In order to apply them to the POMDP for data collection, we grant them observability to the system state $s$.
In simulation, this can be done through providing them access to the underlying simulator state.
As a result, a TAMP policy $\pi_p(s_t)$ need only be a function of the state $s_t$, which is a sufficient statistic for the history $\langle h_t, o_t \rangle$.
To construct this policy, we approximate the now observable POMDP with a deterministic model that 
can be effectively planned with~\cite{garrett2020online}.
Then, a TAMP algorithm uses this approximate model to plan a sequence of object interactions, the constraints present in each interaction ({\it e.g.} grasps and placements), and finally safe joint motions that realize 
them.
An automated policy is built around the TAMP algorithm
by tracking plans with a high-frequency feedback controller that outputs actions $a$ and periodically replanning~\cite{garrett2020online}.

Consider an example TAMP problem in which the goal is to place a \kw{cup} on a \kw{shelf} ({\it i.e.} the {\bf Shelf} task).
The TAMP model has the following parameterized actions:
$\pddl{move}(q_1, \tau, q_2)$ moves the robot from configuration $q_1$ to configuration $q_2$ via trajectory $\tau$,
$\pddl{pick}(o, g, p, q)$ picks object $o$ at placement pose $p$ with grasp pose $g$ when the robot is at configuration $q$,
and $\pddl{place}(o, g, p, q, o_2)$ places object $o$ at placement pose $p$ on object $o_2$ with grasp pose $g$ when the robot is at configuration $q$.
An example TAMP plan $p$ for the Shelf task is:
    $p = [\pddl{move}(\bm{q_0}, \tau_1, q_1), \pddl{pick}(\kw{cup},  g, \bm{p_0}, q_1), \pddl{move}(q_1, \tau_2, q_2), \pddl{place}(\kw{cup}, g, p, q_2, \kw{shelf})]$
The values in bold, the initial configuration $\bm{q_0}$ and cup placement $\bm{p_0}$, are constants.
The other values are free parameters.
A TAMP algorithm searches to find both the {\em plan skeleton}, the sequence of parameterized actions, as well as values for grasp $g$, placement $p$, configurations $q_1, q_2$, and trajectories $\tau_1, \tau_2$ that satisfy grasp, stability, kinematic, and collision constraints.

\vspace{-5pt}
\subsection{Cost-Minimizing TAMP}
\vspace{-5pt}
We use the PDDLStream planning framework~\cite{garrett2020pddlstream} to model the TAMP domain and the {\em adaptive} algorithm, a sampling-based algorithm, to plan.
Our formulation makes use of samplers for grasp generation, placement sampling, inverse kinematics, and motion planning.
The samplers can produce a large, if not infinitely large, set of diverse values.
We implement the grasp generator using the ACRONYM grasp dataset~\cite{eppner2021acronym} for ShapeNet objects.
We use TRAC-IK~\cite{Beeson-humanoids-15} for inverse kinematics (IK), and bidirectional Rapidly-Exploring Random Trees (BiRRT)~\cite{KuffnerLaValle} for motion planning.

When using TAMP solutions for imitation learning, it is essential to train on high-quality plan traces.
Behavior cloning techniques typically are adverse to multi-modal policy behavior, so a TAMP demonstrator that takes several different actions at a particular state produces data is challenging to imitate.
One way to reduce TAMP policy variability is to optimize for low-cost plans.
Although a TAMP problem is not guaranteed to have a unique minimum cost solution, 
this strategy biases solutions to a consistent family of low-cost plans.

We propose a two-stage approach to producing low-cost TAMP solutions.
First, we use cost-sensitive PDDLStream planning that minimizes the joint-space distance traveled.
Specifically, we define costs for $\pddl{move}(q_1, \tau, q_2)$ actions that limit $\infty$-norm (max) of the distance $||q_1 - q_2||_\infty$ between configurations $q_1$, $q_2$.
The straight-line distance between two configurations is a lower bound on the length of the shortest collision-free path between them.
We optimize this lower bound before performing motion planning which is computationally expensive due to continuous collision checking.
This PDDLStream algorithm is asymptotically optimal~\cite{vega2016asymptotically,garrett2020pddlstream}, but 
it might take arbitrary long to find a plan below a target cost bound.
In practice, we run the planner in an anytime mode with a computation budget of five seconds and return the best plan identified.
In the second stage, we perform motion planning using BiRRT; however, 
it can produce motions that are jagged and locally sub-optimal.
To smooth these trajectories, we post-process them using cubic spline short cutting with velocity and acceleration limits~\cite{hauser2010fast}, which converges to a locally time-optimal trajectory.

Finally, we aim to limit the variability in IK solutions.
This \textcolor{red}{is} also advantageous for task-space control, which lacks the control authority to reach all IK solutions.
We seed TRAC-IK's optimization-based IK from a single configuration seed, the initial configuration, and optimize for the closest solution to the initial configuration within a 10 millisecond timeout.
This also biases TAMP toward plans that stay near the initial configuration, typically accelerating the search for low-cost plans.
By intentionally not exploiting the redundancy to explore diverse IK solutions, we limit the completeness of the TAMP algorithm for the benefit of downstream learning.
\vspace{-5pt}
\subsection{Generating Imitation Data from TAMP}
\vspace{-5pt}
Directly training on datasets collected by 
TAMP is a challenge for imitation learning, as the TAMP system operates with access to information unavailable to the learner, controls the robot in 
joint space, which can be difficult to learn in,
and generates demonstrations that may not necessarily take the shortest path in task-space. To address these issues, we highlight design decisions regarding the observations and actions we produce from the TAMP data-generation process as well as how we select which demonstrations to train on.

\textbf{Imitating a Privileged Expert:}
TAMP operates over a privileged view of the world. 
It has access to information that is difficult to obtain from a perception system, 
such as environment geometry and object state.
To address these issues, OPTIMUS\xspace operates over image observations
by using multiple camera views in each task (1-2 fixed cameras and 1 wrist-mounted camera). We find that multiple views, in particular the wrist camera, help the agent to better perceive scene geometry~\cite{hsu2022vision} and align its actions with the privileged expert. 
By training over multi-view RGB observations, we provide the network with an observation space that is invariant to object symmetry, encodes 3D information, is efficient to train over, and enables simplicity of the architecture.

\textbf{Learning from TAMP Generated Actions:}
The TAMP system plans arm motions in configuration space, in which it can fully control each robot degree of freedom.
However, training vision-based policies in joint-space is difficult due to the challenge of learning the camera projection from pixels to poses and then the redundant inverse kinematics mapping from pixels to joint angles~\citep{martin2019variable,zhu2020robosuite,robomimic2021}.
Additionally, for robots with more than six degrees of freedom, joint space is higher dimensional than task space.
Thus, in OPTIMUS\xspace, we instead use task-space control.
We generate task space trajectories by performing forward kinematics on joint-space way-points given by the TAMP planner, then execute 
an operational-space (task-space) controller~\cite{khatib1987unified}
to achieve them. 
Appendix~\ref{app:appendix ablations} conducts an experiment comparing the trained policy success rate with joint-space actions versus task-space actions.
Fig.~\ref{fig:action and data curation ablations} shows that task-space actions enable higher success rates.

\textbf{Filtering Demonstrations:}
    \begin{figure}
    \includegraphics[width=1\linewidth]{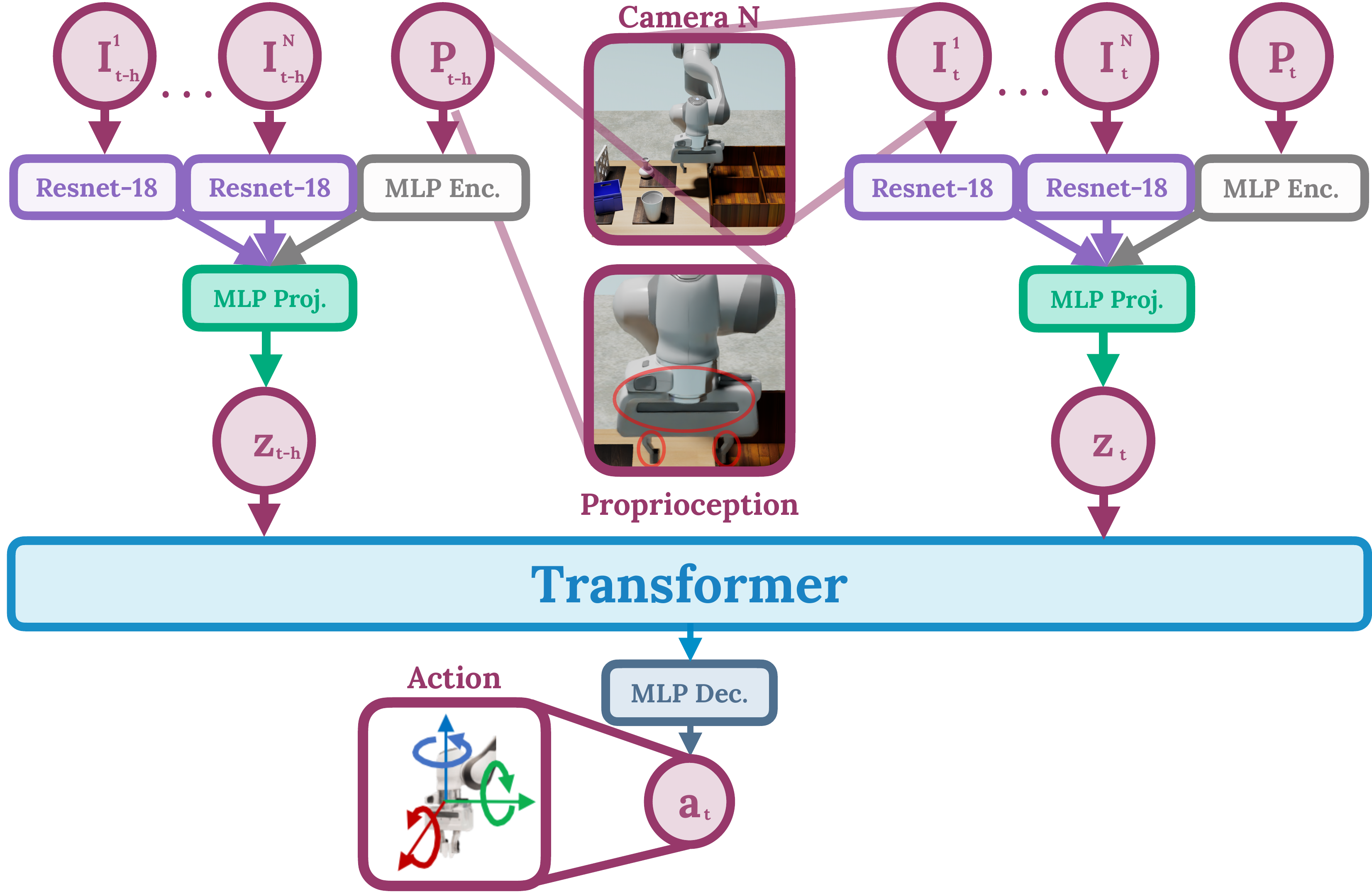}
    \vspace{-20pt}
    \caption{\small \textbf{OPTIMUS\xspace policy architecture}. The model takes as input multiple images and proprioception information per time-step, with a context of $h$. We encode the input using Resnet-18 for images and a MLP for the low-dimensional observations. We concatenate the embeddings, project them into the Transformer embedding dimension and pass them to the Transformer, which predicts an embedding that is decoded into an action.}
    \label{fig:Transformer-figure}
    \end{figure}
Since there is variance in run-time due to random sampling and the TAMP system is not guaranteed to converge in plan cost within the fixed time limit, some plans and thus behaviors may be sub-optimal.
This data can often hamper policy learning by operating outside of the space of nominal solution trajectories. 
Training on this data from the TAMP system increases the likelihood of the agent leaving areas of high state space coverage, which produces policies that exhibit heightened compounding error. To ease the burden on the policy, we curate the data using several trajectory pruning rules. During data collection, we employ joint-space path smoothing.
However, straight-line paths through joint-space are non-linear in task space, 
resulting in longer motions in the learner's action space.
Therefore, we propose two data pruning rules (Fig.~\ref{fig:main-figure} \textit{column 2}) to filter TAMP demonstrations.
First, we remove outlier trajectories that have task-space length greater than two standard deviations away from the mean trajectory length, which can be viewed as randomly restarting TAMP episodes to reduce plan variance.
Second, we impose a containment constraint in the form of a bounding box in visible workspace
and prune out trajectories in which the end-effector pose exits the box. 
Appendix~\ref{app:appendix ablations} and Fig.~\ref{fig:action and data curation ablations} illustrate that the combination of these rules does improve performance by comparing the trained policy success rate with and without filtering.
\vspace{-5pt}
\subsection{Training Imitation Policies at Scale}
\vspace{-5pt}
\label{sec:imitation pipeline}

We now describe the imitation pipeline in OPTIMUS\xspace. Given large, diverse datasets from TAMP, we perform offline behavior cloning to distill the TAMP expert into a visuomotor policy. 

\textbf{OPTIMUS\xspace Architecture:}
Our policy must operate over a history of multiple camera views and proprioception, output low-level task space actions, and execute in real time. To that end, we design our policy $\pi$, visualized in Fig.~\ref{fig:Transformer-figure}, as a Transformer operating over a history of observations $h$, in which each token corresponds to a single observation time-step. As a result, the Transformer can efficiently attend to all observations as the Transformer context length is set to $h$. To produce a single input token for a time-step $t$, we first embed each input, images from cameras $1,...,N$ ($I_1^t$ through $I_N^t$) as well as proprioception $p_t$, into fixed dimensional vector spaces. For proprioception, we pass in the end-effector pose (xyz position and quaternion orientation) and gripper joint position (dual finger positions), encoded by an MLP. For embedding images, we use the vision backbone from~\citet{robomimic2021}: ResNet-18~\cite{he2016deep} with a spatial softmax~\cite{levine2016end} output activation. We then fuse the inputs for a single time-step to produce $z_t$, a vector matching the Transformer embedding dimension, by concatenating and performing an MLP projection. The Transformer attends to each token $z_t$ and outputs a distribution over action $a_t$ corresponding to the current time-step. 

The data distribution outputted by the TAMP supervisor is heavily multi-modal, from the diversity in planned paths to the variety of grasps and placements per object. As a result, we use a Gaussian Model Mixture (GMM) output distribution with $K=5$ components for the policy from~\citet{robomimic2021} and train the model using log likelihood. 
As in~\cite{robomimic2021}, we find that this loss function provides a significant improvement over the standard MSE loss, which produces a unimodal policy.

\begin{figure*}[t]
\centering
\begin{subfigure}[b]{0.24\linewidth}
    \includegraphics[width=\linewidth]{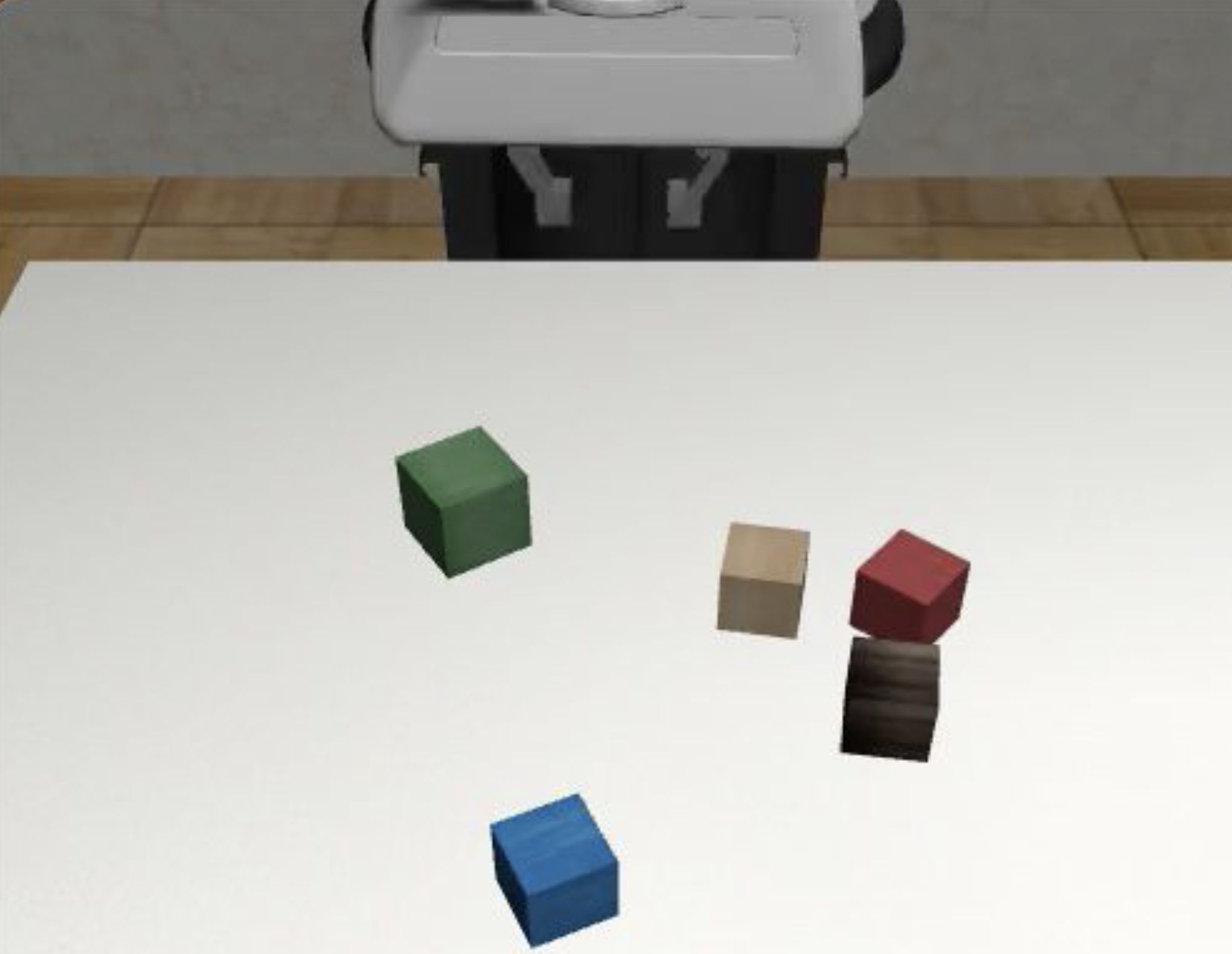}
    \caption{\small StackFive}
\end{subfigure}
\begin{subfigure}[b]{0.24\linewidth}
    \includegraphics[width=\linewidth]{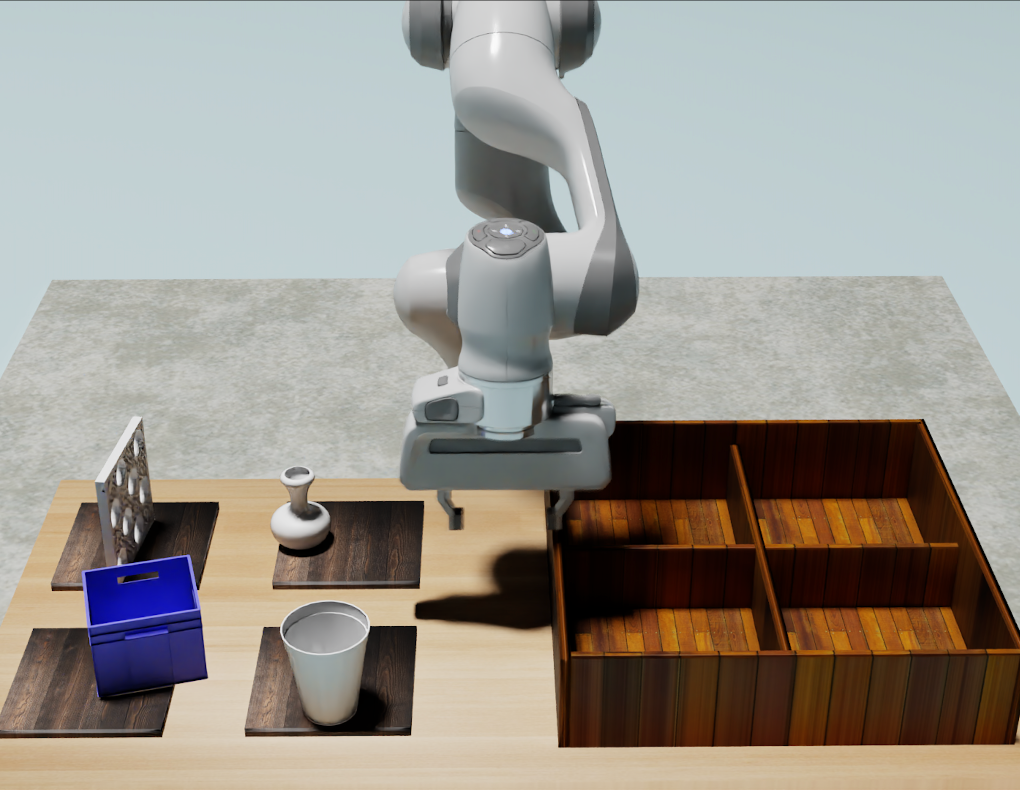}
    \caption{\small PickPlaceFour}
\end{subfigure}
\begin{subfigure}[b]{0.24\linewidth}
    \includegraphics[width=\linewidth]{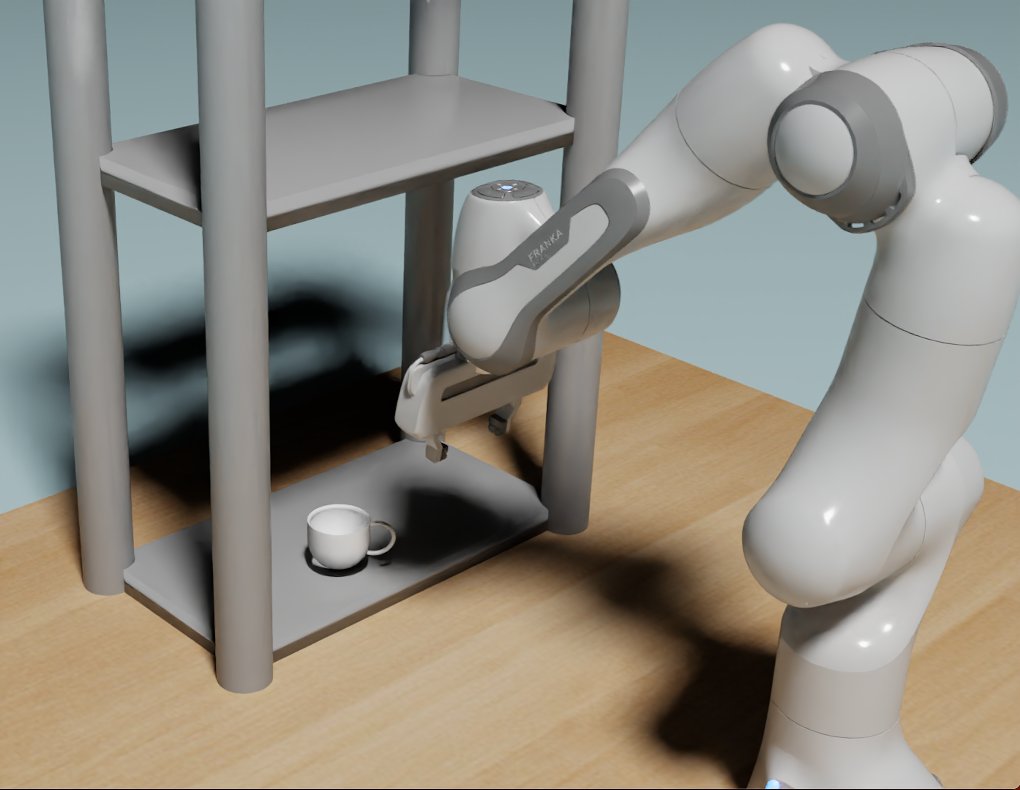}
    \caption{\small Shelf}
\end{subfigure}
\begin{subfigure}[b]{0.24\linewidth}
    \includegraphics[width=\linewidth]{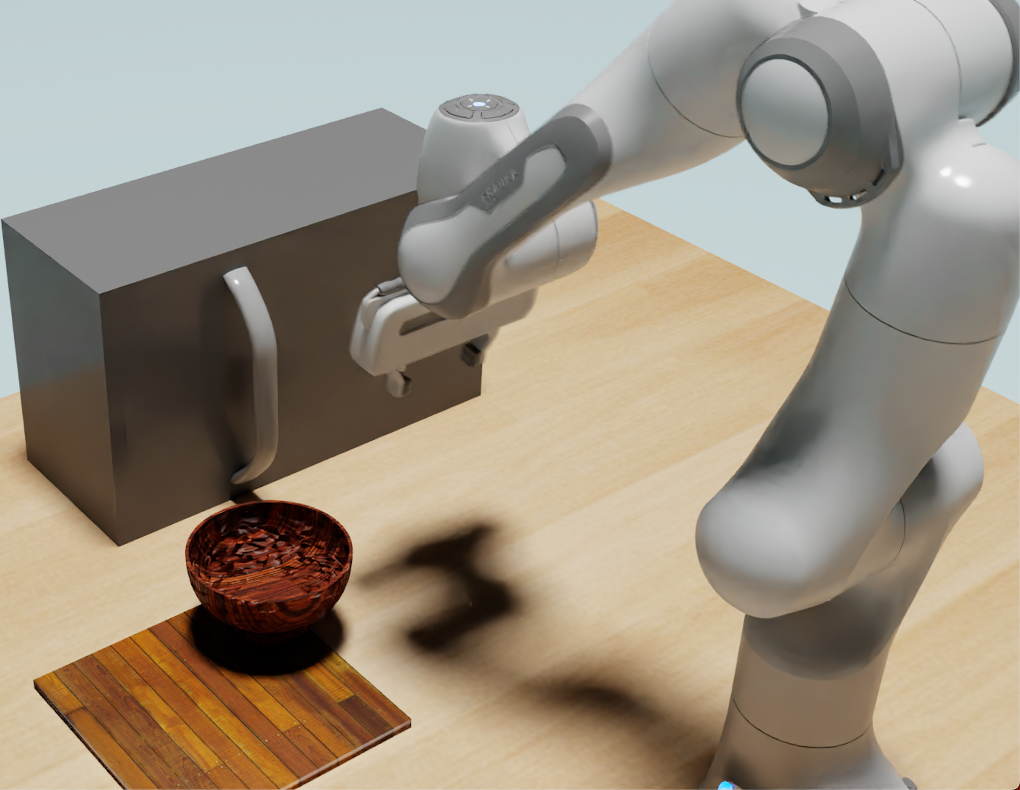}
    \caption{\small Microwave}
\end{subfigure}
\caption{\small \textbf{Environment Visualizations}. We evaluate OPTIMUS\xspace on long-horizon block stacking (a), multi-step pick-place (b), shelf object manipulation (c), and articulated object manipulation (d).}
\vspace{-.1in}
\label{fig:env figure}
\end{figure*}
\vspace{-10pt}
\section{Experimental Setup}
\label{sec:experiments}
\vspace{-7pt}
For our experiments, we describe the datasets, tasks, and protocols that we use to evaluate OPTIMUS\xspace.

\textbf{Datasets and Tasks:} We evaluate OPTIMUS\xspace across block stacking, pick and place, shelf manipulation and articulated object manipulation. (Fig.~\ref{fig:env figure}). Our block stacking tasks have two (Stack), three (StackThree), four (StackFour) or five (StackFive) blocks, with 1K, 2K, 5K, and 7K demonstrations respectively. For pick and place, we have: PickPlace-1, pick-place with a single object using 1K demos, and pick-place with two (PickPlaceTwo), three (PickPlaceThree), and four objects (PickPlaceFour) into separate bins. Finally we have two tasks in which the goal is for the agent to move the object to the target location while maneuvering in tight spaces (Shelf-1) or first pulling open a microwave door (Microwave-1), for which we generate 1K demonstrations each. 
For PickPlace, Shelf, and Microwave, we additionally evaluate two multi-task variants, in which we sample a set of 19 and 72 objects from ShapeNet. We collect a 1K demonstrations per object, with 19K and 72K total trajectories, resulting in the following datasets: Pickplace-19, PickPlace-72, Shelf-19, Shelf-72, Microwave-19, Microwave-72. See Appendix Sec.~\ref{app:envs} for complete task descriptions and details.

\textbf{Evaluation Protocol:} We evaluate BC-MLP~\cite{zhang2018deep} and BC-RNN~\cite{robomimic2021}, which consist of a Resnet-18 backbone followed by an MLP and an LSTM~\cite{hochreiter1997long}. Additionally, we compare against Behavior Transformer (BeT)~\cite{shafiullah2022behavior}, which discretizes the dataset into clusters using K-Means and uses a Transformer model to predict a cluster center and an offset, in order to handle multi-modal data. Each method uses on the order of magnitude of 30M parameters, uses the same architecture as OPTIMUS\xspace for the vision-backbone, and leverages the data-pipeline we propose in Sec.~\ref{sec:data pipeline}.
For each task, we evaluate on a dataset of unseen initial environment states. For single-task results, we evaluate using 50 problems and average across 3 random seeds per run. For multi-task results, we evaluate using 10 problems per task, with a single seed per run.
We use task success rate as our evaluation metric, which is 1 if all objects are in a goal arrangement and 0 otherwise. 
\vspace{-5pt}
\section{Results}
\label{subsec:learning results}
\vspace{-5pt}
In our experimental evaluation of OPTIMUS\xspace, we aim to answer the following questions: 1) Can imitating TAMP enable end-to-end policies to acquire long-horizon behaviors? 2) Does TAMP allow networks to solve complex manipulation tasks involving 3D obstacles and articulated objects? 3) Does diverse environment generation along with TAMP data-collection enable large-scale behavior learning?

We first show that OPTIMUS\xspace can imitate the TAMP system to high fidelity on simple, shorter horizon tasks. We then extend our evaluation to the long-horizon regime in which the task complexity grows significantly with the number of objects. Next, we move beyond the table-top manipulation setting and train policies to solve tasks involving a shelf and microwave. Finally, we demonstrate that OPTIMUS\xspace can enable multi-task policies that can manipulate a wide range of objects. Please see Appendix~\ref{app:appendix ablations} for a detailed analysis and ablation of OPTIMUS\xspace.

\begin{figure}[t!]
\centering
\begin{minipage}[t]{0.47\textwidth}
\centering
\resizebox{1\linewidth}{!}{
\begin{small}
\begin{tabular}{lcccc}
\toprule
\textbf{Dataset} & 
\begin{tabular}[c]{@{}c@{}}\textbf{BC-MLP}\end{tabular} & 
\begin{tabular}[c]{@{}c@{}}\textbf{BC-RNN}\end{tabular} &
\begin{tabular}[c]{@{}c@{}}\textbf{BeT}\end{tabular} &
\begin{tabular}[c]{@{}c@{}}\textbf{OPTIMUS\xspace}\end{tabular} \\ 
\midrule
Stack & $100$ & $100$ & $100$ & $100$ \\
\rowcolor{Gray}
StackThree & $98$ & $88$ & $76$ & $\textbf{100}$ \\
StackFour & $83$ & $77$ & $61$ & $\textbf{96}$ \\
\rowcolor{Gray}
StackFive & $57$ & $57$ & $45$ & $\textbf{70}$ \\
\bottomrule
\end{tabular}
\end{small}
}
\vspace{-.1in}
\label{table:stacking}

\end{minipage}
\hfill
\begin{minipage}[t]{0.51\textwidth}
\centering
\resizebox{1\linewidth}{!}{
\begin{small}
\begin{tabular}{lcccc}
\toprule
\textbf{Dataset} & 
\begin{tabular}[c]{@{}c@{}}\textbf{BC-MLP}\end{tabular} & 
\begin{tabular}[c]{@{}c@{}}\textbf{BC-RNN}\end{tabular} &
\begin{tabular}[c]{@{}c@{}}\textbf{BeT}\end{tabular} &
\begin{tabular}[c]{@{}c@{}}\textbf{OPTIMUS\xspace}\end{tabular} \\ 
\midrule
PickPlaceTwo & $96$ & $\textbf{98}$ & $80$ & $96$ \\
\rowcolor{Gray}
PickPlaceThree & $62$ & $81$ & $46$ & $\textbf{91}$ \\
PickPlaceFour & $33$ & $38$ & $22$ & $\textbf{60}$ \\
\bottomrule
\end{tabular}
\end{small}
}
\vspace{-.1in}

\end{minipage}
\caption{\small \textbf{Long Horizon Manipulation Results.} 
(left) Performance is shown in terms of task success rate. While all methods are able to solve single-step block stacking, only OPTIMUS\xspace is able to solve longer-horizon variants. 
(right) 
For long-horizon manipulation, while the baselines are competitive with OPTIMUS\xspace on PickPlaceTwo, OPTIMUS\xspace demonstrates significant improvement in success rate as the number of objects increases.}
\label{fig:long-horizon-manip}
\end{figure}

\textbf{OPTIMUS\xspace imitates TAMP to high fidelity on simple pick-and-place tasks.}
On Stack (Fig.~\ref{fig:long-horizon-manip}), we find that OPTIMUS\xspace and the baselines are all able to achieve 100\% performance on the task. On the other hand, on PickPlace-1, (Fig.~\ref{fig:long-horizon-manip}), while the baselines achieve high success rates of up to 97\%, only our method is able to solve the task at 100\% success rate. These results demonstrate that on simple tasks, OPTIMUS\xspace can fit well to the output of the TAMP system, even though OPTIMUS\xspace does not have access to any privileged information. We note that even with significant tuning, BeT struggles to fit to TAMP data on most of our tasks. We hypothesize that this may be due to the difficulty of fitting K-Means as the dataset size increases, especially as TAMP generated datasets contain on the order of 1-100K trajectories depending on the task. As a result, the cluster centers can be highly inaccurate, increasing the burden on the transformer to fit appropriate offsets.

\textbf{OPTIMUS\xspace enables visuomotor policies to solve manipulation tasks with up to 8 stages.}
We first evaluate on long-horizon block stacking, a task that is difficult 
because the stack of blocks becomes more unstable as its height grows. We train visuomotor policies across StackThree, StackFour, and StackFive, and visualize the results in Fig.~\ref{fig:long-horizon-manip}. OPTIMUS\xspace outperforms the baseline methods while achieving near-perfect performance across each task. Multi-step pick-place is even more difficult as the network must learn to fit a variety of different grasps for each object. 
We plot the results for the multi-step pickplace tasks in Fig.~\ref{fig:long-horizon-manip}. We find that while BC-RNN outperforms OPTIMUS\xspace on PickPlaceTwo, OPTIMUS\xspace exhibits a large performance improvement on PickPlaceThree and Four. 
These results demonstrate that with either primitive or general-purpose rigid objects, it is possible to train policies to perform long-horizon behaviors consisting of up to 8 pick and place operations or 40 TAMP high-level actions, with high success rates of \textbf{70\%} and \textbf{60\%} respectively. An important take-away from these results is that for longer-horizon tasks, the Transformer policy architecture we develop in OPTIMUS\xspace greatly outperforms MLPs and RNNs. 

\begin{wrapfigure}{l}{0.3\textwidth}
\vspace{-0.3in}
\begin{minipage}{.3\textwidth}
\begin{table}[H]
\resizebox{1\textwidth}{!}{
\begin{small}
\begin{tabular}{cc}
\toprule
\begin{tabular}[c]{@{}c@{}}\textbf{Guided TAMP}\end{tabular} & 
\begin{tabular}[c]{@{}c@{}}\textbf{OPTIMUS\xspace}\end{tabular} \\ 
\midrule
 $88$ & $\textbf{90}$ \\
\bottomrule
\end{tabular}
\end{small}
}
\vspace{-.1in}
\caption{\small \textbf{Comparison against Guided TAMP.} Results are in terms of task success rate. 
}
\label{table:guided-tamp-compare}
\end{table}
\end{minipage}
\vspace{-0.15in}
\end{wrapfigure}
We additionally compare against prior work on imitating TAMP~\citep{mcdonald2022guided} on the Robosuite~\cite{zhu2020robosuite} PickPlace task, which involves picking and placing four fixed objects: a milk carton, a soda can, a cereal box and a piece of bread, into separate bins. 
In contrast to PickPlaceFour, Robosuite PickPlace can be solved with top-down, axis-aligned grasps due to the simplicity of the object geometry. However, the initial configurations are more challenging as all the objects are placed together in the same bin. We generate 25K demonstrations of the task using our system. As we show in Table~\ref{table:guided-tamp-compare}, OPTIMUS\xspace achieves favorable results to Guided TAMP (90\% vs. 88\%) without requiring access to privileged state information, a fixed set of ground actions or online supervision.

\textbf{OPTIMUS\xspace can also solve tasks requiring obstacle awareness and skills beyond pick-and-place.}
On Shelf-1, OPTIMUS\xspace is able to grasp then place the object in the middle rung of the shelf at high success rates of 91\% shown in Table~\ref{table:all multitask}. On Microwave-1 OPTIMUS\xspace outperforms the baselines by nearly 10\%, achieving 86\% success rate overall. This is likely because OPTIMUS\xspace is able to better fit the data in the multi-step manipulation regime, as noted in the prior section. The results on the Shelf and Microwave tasks demonstrate that OPTIMUS\xspace can learn to solve difficult manipulation tasks that require obstacle awareness and the ability to manipulate articulated objects. 

\textbf{OPTIMUS\xspace can learn to adapt its behavior based on the scene configuration.} 
As we describe in the Appendix, OPTIMUS\xspace is able to learn to adapt its task plan to produce additional stacking operations (StackAdapt) or clear the area in front of the microwave (MicrowaveAdapt) achieving 96\% and 75\% success. OPTIMUS\xspace is able to generalize to unseen receptacle sizes, achieving 80\% and 70\% success rate on held out shelves and microwaves. 

We next evaluate the ability of our TAMP generation pipeline to collect diverse datasets in order to train large-scale policies. We add variety in the form of objects with differing geometries, requiring a single network to learn a range of manipulation behaviors end-to-end. 

\begin{wrapfigure}{l}{0.5\textwidth}
\vspace{-0.3in}
\begin{minipage}{0.5\textwidth}
\begin{table}[H]
\centering
\resizebox{1\linewidth}{!}{
\begin{small}
\begin{tabular}{lcccccccccccccc}
\toprule
\textbf{Dataset} & 
\begin{tabular}[c]{@{}c@{}}\textbf{BC-MLP}\end{tabular} & 
\begin{tabular}[c]{@{}c@{}}\textbf{BC-RNN}\end{tabular} &
\begin{tabular}[c]{@{}c@{}}\textbf{BeT}\end{tabular} &
\begin{tabular}[c]{@{}c@{}}\textbf{OPTIMUS\xspace}\end{tabular} \\ 
\midrule
PickPlace-1 & $94$ & $97$ & $85$ & $\textbf{100}$ \\
\rowcolor{Gray}
PickPlace-19 & $61$ & $58$ & $50$ & $\textbf{85}$ \\
PickPlace-72 & $50$ & $49$ & $41$ & $\textbf{75}$ \\
\midrule
Shelf-1 & $\textbf{91}$ & $88$ & $70$ & $\textbf{91}$ \\
\rowcolor{Gray}
Shelf-19 & $48$ & $31$ & $26$ & $\textbf{66}$ \\
Shelf-72 & $30$ & $36$ & $13$ & $\textbf{48}$ \\
\midrule
Microwave-1 & $73$ & $77$ & $51$ & $\textbf{86}$ \\
\rowcolor{Gray}
Microwave-19 & $24$ & $41$ & $31$ & $\textbf{61}$ \\
Microwave-72 & $23$ & $29$ & $16$ & $\textbf{47}$ \\
\bottomrule
\end{tabular}
\end{small}
}
\vspace{-.1in}
\caption{\small \textbf{Single and Multitask Results across PickPlace, Shelf, Microwave.} Performance is shown in terms of task success rate. 
While the baselines are competitive with OPTIMUS\xspace on the single task variants of each task, OPTIMUS\xspace greatly outperforms the baselines as the number of objects increases across all tasks.
}
\label{table:all multitask}
\end{table}

\end{minipage}
\vspace{-0.225in}
\end{wrapfigure}

\textbf{OPTIMUS\xspace achieves high success rates on vision-based manipulation tasks with up to 72 objects.} For each task: PickPlace, Shelf, and Microwave, we evaluate on their 19 and 72 object variants (Table~\ref{table:all multitask}). On the 19 object tasks, OPTIMUS\xspace achieves 85\%, 66\%, and 61\% in greatly outperforming the best baseline for each task: 61\%, 48\%, and 41\%. Similarly on the 72 object tasks, we find that OPTIMUS\xspace obtains 75\%, 48\% and 47\% success rates, in comparison to 50\%, 36\% and 29\% for the best baseline. 
From these results, we note two important points: 1) Transformer-based architectures such as OPTIMUS\xspace are highly effective for multi-task imitation learning: they greatly outperform MLPs and RNNs. 2) While the single task variants of these tasks are solved at high success rates, performance drops significantly in the multi-task case, particularly for more challenging manipulation tasks such as Shelf and Microwave, indicating further work remains to bridge that gap.

Finally, we highlight three advantages of OPTIMUS\xspace over the TAMP system: 1) success rate improvement over the TAMP supervisor, 2) faster run-time, 3) operation from image instead of state input. 
We evaluate TAMP and OPTIMUS\xspace on all of the single task datasets and find that on average, OPTIMUS\xspace almost doubles the performance of the TAMP supervisor (87\% vs. 52\%). Additionally, we evaluate the run-time of OPTIMUS\xspace against TAMP by computing the average time per step for both systems across 100 trials. Overall, OPTIMUS\xspace is 5-7.5x faster than TAMP (21/31ms vs. 150ms per action). 
See Appendix Sec.~\ref{app:additional results} for analysis of scene adaptation and TAMP comparison results.

\vspace{-10pt}
\section{Discussion and Limitations}
\vspace{-5pt}
In this chapter, we propose an approach for distilling privileged TAMP experts into large-scale visuomotor policies.
We generate large, diverse datasets and train high-capacity Transformer models to solve challenging, long-horizon manipulation tasks without task, state, or environment knowledge. 
Even so, there are limitations to OPTIMUS\xspace and scope for future work. First, for OPTIMUS\xspace to be able to solve a task, TAMP needs to be capable of solving it at training time, which could prevent OPTIMUS\xspace from being applied to tasks that require considering dynamics or tasks involving contact-rich manipulation, which can be challenging for traditional TAMP. However, TAMP can be applied to such tasks, e.g. pouring, scooping, stirring, peg insertion, or coffee making, by leveraging integrated task planning and skill learning approaches ~\cite{wang2021learning,curtis2022long,cheng2022guided,mandlekar2023hitltamp}, which OPTIMUS\xspace can leverage for supervision as well. Second even with OPTIMUS\xspace, there is a significant drop in success rate with increasing task difficulty and number of objects (Sec.~\ref{subsec:learning results}), which suggests further work on multi-task learning is necessary to bridge that gap.

Inspired by OPTIMUS\xspace, in the next chapter we explore the question of how to train broadly capable multi-task policies that can be directly deployed on robots. We do so in the context of the problem of motion planning, which poses largely a perceptual challenge for sim2real transfer, and thus is a natural testbed for scaling policies in simulation. As we show in the following chapter, the simple recipe of generating diverse scenes in simulation, producing vast amounts of data with a planner expert and an expressive policy to fit the data produces capable, multi-task policies that generalize a large set of real-world tasks.

\clearpage

\chapter{Data Driven Neural Motion Planning}
\label{chap:nmp}
\begin{figure}[h]
    \includegraphics[width=1\linewidth]{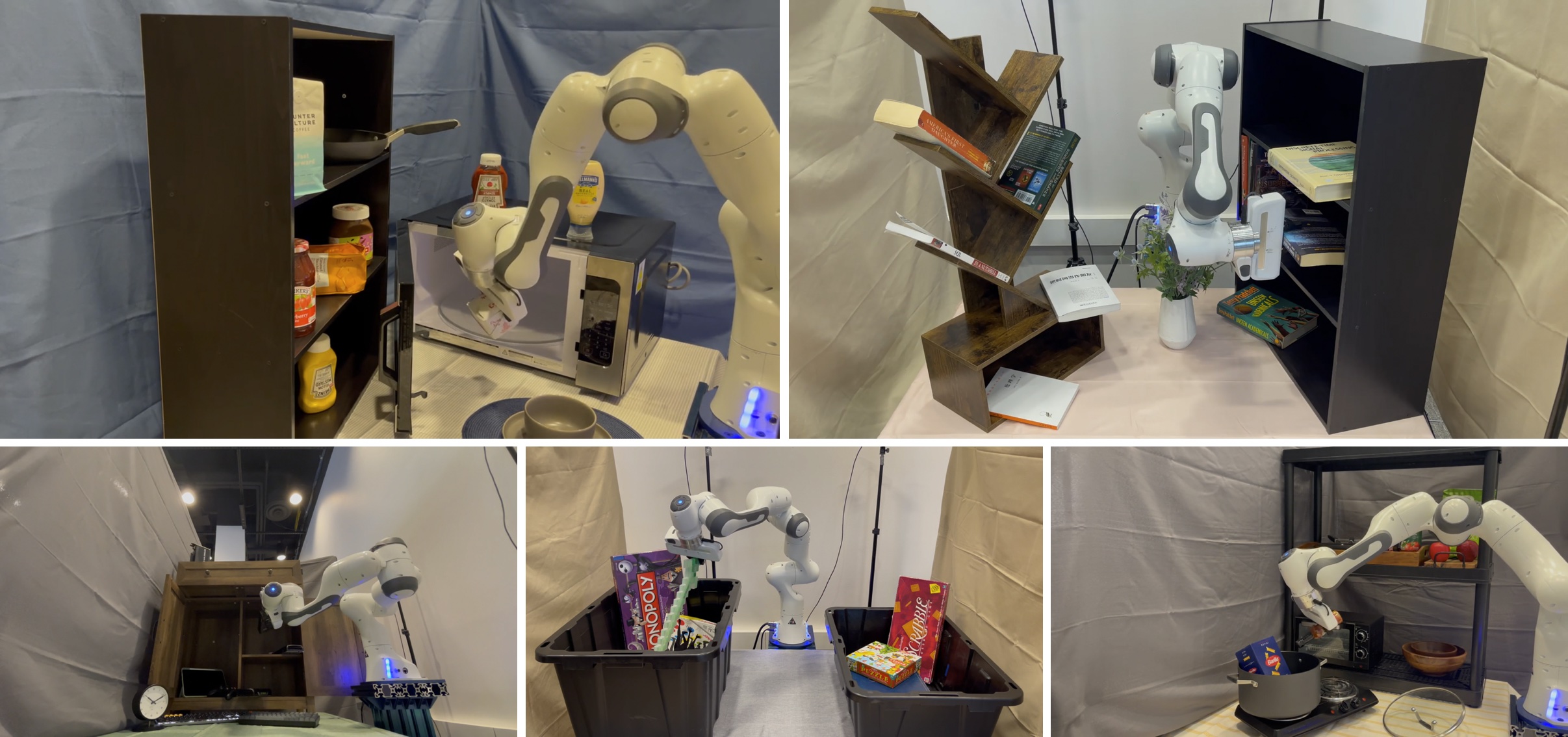}
      \centering
      \vspace{-5pt}
      \caption{\small \textbf{Neural Motion Planning at Scale in the Real World} Our approach enables a \textit{single}, \textit{generalist} neural network policy to solve motion planning problems across diverse setups; Neural MP can generate collision free motions for a wide array of unseen tasks significantly faster and with higher success than traditional as well as learning-based motion planning approaches.}
    \label{fig:teaser}
\end{figure}
\section{Introduction}

Motion planning is a longstanding problem of interest in robotics, with
previous approaches ranging from potential fields~\citep{khatib1986real,warren1989global,quinlan1993elastic}, sampling~\citep{kavraki1996probabilistic,lavalle2001rapidly,lazyprm,karaman2011sampling,kuffner2000rrt,bit*,strub2020adaptively}, search~\citep{hart1968formal,likhachev2003ara,koenig2006new} and trajectory optimization~\citep{ratliff2009chomp,schulman2014motion,dragan2011manipulation,sundaralingam2023curobo}. 
Despite being ubiquitous, these methods are often slow at producing solutions since they 
largely plan from scratch at test time, re-using little to no information outside of the current problem and what is engineered by a human designer. Since motion-planning is a core component of the robotics stack for manipulation, its speed, capability and ease of use form a core bottleneck to developing efficient and reliable manipulation systems.

On the other hand, humans can generate motions in a closed loop manner, move quickly, react to various dynamic obstacles, and generalize across a wide distribution of problem instances. 
Rather than planning open loop from scratch, people draw on their vast amounts of experience moving and interacting with their environment while reactively adjusting their movements in order to quickly and efficiently move about the world. How can we create motion planners with similar properties? 
In this work, we argue that \textit{distillation} at scale is the answer: we can \textit{distill} the planning process into a \textbf{reactive, generalist} neural policy.

The primary challenge in training data-driven motion planning is the data collection itself, as scaling robotic data collection in real-world requires significant human time and effort. Recently, there has been a concerted effort to scale up data collection for robot tasks~\citep{padalkar2023open, khazatsky2024droid}. However, the level of diversity of scenes and arrangement of objects is still limited, especially for learning obstacle avoidance behavior that scales to the real world. Constructing such setups with diverse obstacle arrangements with numerous objects is prohibitively expensive in terms of cost and labor.
\setcounter{figure}{1}
\begin{figure*}
    \centering
    \includegraphics[width=\textwidth]{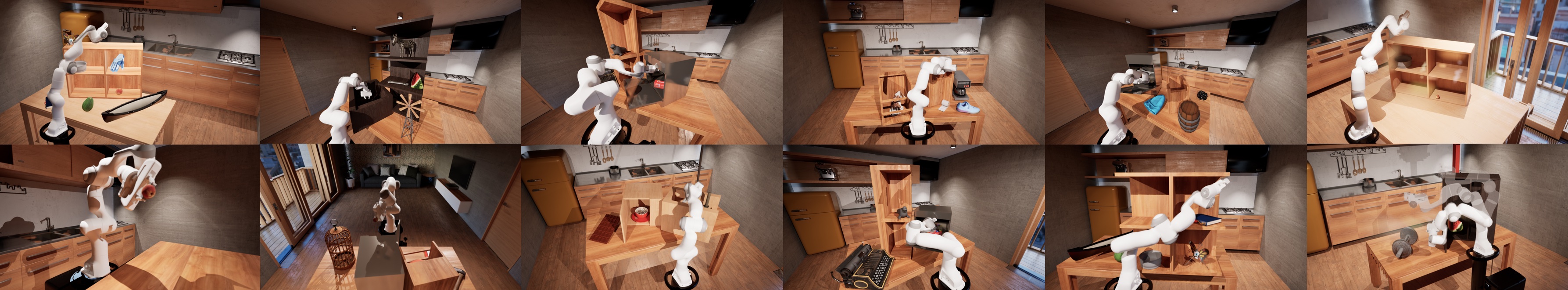} 
     \vspace{-15pt}
    \caption{\small \textbf{Visualization of Diverse Simulation Training Environments}: We train Neural MP on a wide array of motion planning problems generated in simulation, with significant pose, procedural asset, and mesh configuration randomization to enable generalization. }
    \label{fig:sim vis}
\end{figure*}

Instead, we leverage simulation, which makes it cheap and easy to obtain diverse data, is highly scalable via parallelization, and runs significantly faster than real world. Recent approaches have shown great promise in enabling policy learning for high-dof robots~\citep{lee2020learning, zhuang2023robot, cheng2023extreme, kumar2021rma, haarnoja2024learning,agarwal2023legged}. We build a large number of complex environments by combining procedural, programmatic assets with models of everyday objects sampled from large 3D datasets. These are used to collect expert data from state-of-the-art (SOTA) motion planners~\citep{strub2020adaptively}, which we then \emph{distill} into a reactive, generalist policy. Since this policy has seen data from \textbf{1 million} scenes, it is capable of generalizing to novel obstacles and scene configurations that it has never seen before. 
However, deploying neural policies in the real world might be unsafe for the system due to the potential of collisions. We mitigate this by using a linear model to predict future states the robot will end up in and run optimization to ensure a safe path.

Our core contribution is a SOTA motion planner that runs zero-shot on any environment, with more accuracy and in orders of magnitude less execution time. We demonstrate that large scale data generation in simulation can enable training generalist policies that can be successfully deployed for real-world motion planning tasks.
To our knowledge, Neural MP is the first work to demonstrate that such a neural policy can generalize to a broad set of out-of-distribution of real-world environments, generalizing across tasks with significant variation across poses, objects, obstacles, backgrounds, scene arrangements, in-hand objects, and start/goal pairs. 
Specifically, we propose a simple, scalable approach for training and deploying fast, general purpose neural motion planners: 1) \textbf{large-scale procedural scene generation} with diverse environments in realistic configurations, 2) \textbf{multi-modal sequence modeling} for fitting to sampling-based motion planning data and 3) \textbf{lightweight test-time optimization} to ensure fast, safe, and reliable deployment in the real world. 
We execute a thorough real-world empirical study of motion-planning methods, evaluating our approach on \textbf{64} real world motion planning tasks across \textbf{four} diverse environments, demonstrating a motion planning success rate improvements of \textbf{23\%} over sampling-based, \textbf{17\%} over optimization-based and \textbf{79\%} over neural motion planning methods.

\section{Related Work}

\noindent \textbf{Approaches for Training General-Purpose Robot Policies}
Prior work on large scale imitation learning using expert demonstrations~\citep{brohan2022rt,brohan2023rt,padalkar2023open,dalal2023optimus,shridhar2023perceiver,khazatsky2024droid} has shown that large models trained on large datasets can demonstrate strong performance on challenging tasks and some varying levels of generalization. On the other hand, sim2real transfer of RL policies trained with procedural scene generation has demonstrated strong capabilities for producing generalist robot policies for locomotion~\citep{kumar2021rma,zhuang2023robot,agarwal2023legged,cheng2023extreme}. 
In this work, we combine the strengths of these two approaches by proposing a method for procedural scene generation in simulation and combine it with large scale imitation learning and test-time optimization to produce generalist neural motion planners.

\noindent \textbf{Procedural Scene Generation for Robotics}
Automatic scene generation and synthesis has been explored in vision and graphics~\citep{wang2021sceneformer,chang2015text,chang2014learning,ritchie2019fast} while more recent work has focused on embodied AI and robotics settings~\citep{deitke2022️,wang2023robogen,katara2023gen2sim,dalal2023optimus}.
In particular, methods such as Robogen~\citep{wang2023robogen} and Gen2sim~\citep{katara2023gen2sim} use LLMs to propose tasks and build scenes using existing 3D datasets~\citep{deitke2023objaverse} or text-to-3D~\citep{poole2022dreamfusion,wang2023score} and train policies limited to simulation. Our method is instead rule-based rather than LLM-based, is designed for generating data to train neural motion planners, demonstrating strong real-world transfer results. MotionBenchmaker~\citep{chamzas2021motionbenchmaker}, on the other hand, is similar to our data generation method in that it also autonomously generates scenes using programmatic assets. However, the datasets generated by MotionBenchmaker are not realistic: floating robots, a single major obstacle per scene and primitive objects that are spaced far apart. By comparison, the scenes and data generated by our work (Fig.~\ref{fig:sim vis}) are more diverse, containing programmatic assets that incorporate articulations (microwave, dishwasher), multiple large obstacles per scene, complex meshes sampled from Objaverse~\citep{deitke2023objaverse}, and tightly packed obstacles.

\noindent \textbf{Neural Motion Planning}
Finally, there is a large body of recent work~\citep{qureshi2019motion,fishman2023motion,qureshi2020neural,carvalho2023motion,ichter2020broadly,saha2023edmp,huang2023diffusion} focused on imitating motion planners. MPNet~\citep{qureshi2019motion,johnson2020dynamically,qureshi2020neural} trains a network to imitate a planner, then integrates this model into a search procedure at deployment. Our method leverages large scale scene generation and sequence modeling to train a stronger base model, enabling it to use a faster optimization process at test time while obtaining strong results across a diverse set of tasks. 
M$\pi$Nets~\citep{fishman2023motion} trains the SOTA neural motion planning policy using procedural scene generation and demonstrates transfer to the real world. 
Our approach is similar, albeit with 1) increased diversity via programmatic asset generation and complex real-world meshes, 2) more expressive architecture for policy learning and 3) test-time optimization, enabling significantly improved performance over M$\pi$Nets.

\section{Neural Motion Planning}
Our approach enables generalist neural motion planners, by leveraging large amounts of training data generated in simulation via expert planners. The policies can generalize to out-of-distribution settings by using powerful deep learning architectures along with diverse, large-scale training data. To further improve the performance of these policies at deployment, we leverage test time optimization to select the best path out of a number of options. 

\subsection{Large-scale Data Generation}
\label{subsec:datagen}
We leverage simulation to generate vast datasets for training robot policies. Our approach generates assets using programmatic generation of primitives and by sampling from diverse meshes of common objects. These assets are combined to create complex scenes resembling real world scenarios (Fig.~\ref{fig:sim vis}), as described in Alg.~\ref{alg:scene_gen}.

\noindent \textbf{Procedural Generation From Primitives}
To automatically generate a vast set of scenes for training generalist neural motion planners, we take the approach of \emph{procedural} scene generation, using a set of six \emph{parametrically variable} categories - shelves, cubbies, microwaves, dishwashers, open boxes, and cabinets. These categories are representative of a large set of objects in everyday scenarios that robots encounter and have to avoid colliding with. Each category instance is constructed using a combination of primitive cuboid objects and is parameterized by category specific parameters which define the asset. 
Specifically a category instance $g$ is comprised of N cuboids $g = \{ x_0..x_i...x_N\}$, which satisfy the category level constraint  given by $C(g)$. For controlled variation within each category, we make use of \emph{parametric} category specific generation functions
 $X(\textbf{p}) = \{ x_0..x_i.x_N\}, \text{s.t. } C(X(\textbf{p}))$, where $p$ specifies the size and scale of each of the cuboids, their relative positions, and specific axes of articulation. C(.) relates to the relative positions, scales and orientations of the different cuboids, \textit{e.g} for the microwave category the constraint ensures each of the walls are the same height, and that the microwave has a hinge door.

\begin{figure*}
    \centering
    \vspace{10pt}
    \includegraphics[width=\textwidth]{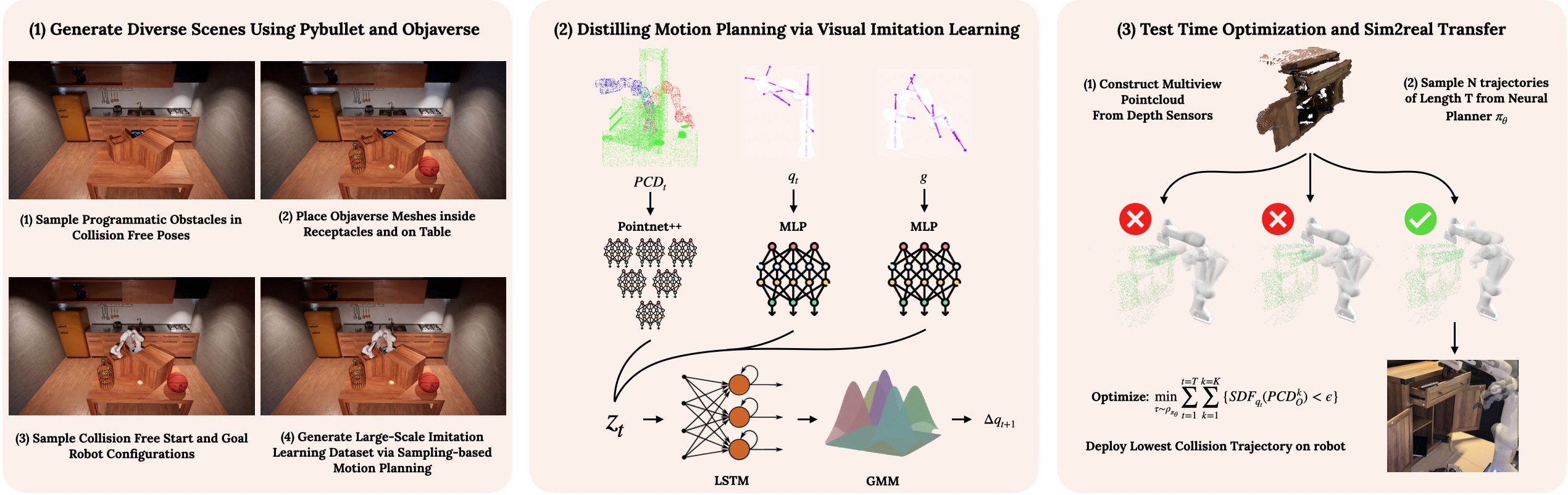} 
    \vspace{-17pt}
    \caption{\small \textbf{Method Overview}: We present Neural Motion Planners, which consists of 3 main components. \textbf{Left}: Large Scale data generation in simulation using expert planners \textbf{Middle}: Training deep network models to perform fast reactive motion planning \textbf{Right}: Test-time optimization at inference time to improve performance. }
    \label{fig:method}
\end{figure*}
\noindent \textbf{Complex Scene Generation} 
To enable trained policies to generalize to complex, real world scenes, we augment our dataset with everyday objects such as comic books, jars, record players, caps, etc. sampled from a large-scale 3D object dataset, Objaverse~\citep{deitke2023objaverse}. We sample these assets in the task-relevant sampling location of the programmatic asset(s) in the scene, such as between shelf rungs, inside cubbies or within cabinets.
For constructing realistic scenes using these objects, a naive approach is to iteratively sample assets on a surface and re-sample those that are in collision. However, as the number, size and type of objects increases, so does the probability of sampling assets that are in collision, making such a process prohibitively expensive and biased towards simple scenes with few assets, which is not ideal for training generalist policies. Instead, we our approach iteratively adds assets to a scene by adjusting their position using the effective collision normal vector, computed from the existing assets in the scene (Alg.~\ref{alg:scene_gen}).

\noindent \textbf{Motion Planning Experts}: 
To collect expert data in the diverse generated scenes, we leverage SOTA sampling-based motion planners due to their ease of application to a wide array of tasks. Specifically, we use Adaptively Informed Trees~\citep{strub2020adaptively} (AIT*) to produce expert plans using privileged information, namely access to a perfect collision checker in simulation. 
We generate challenging training (tight-space) configurations by sampling end-effector poses from task-relevant regions (\textit{e.g.}, inside a cubby or microwave) and using inverse kinematics (IK) to derive the joint pose. Tight-space configurations are sampled 50\% of the time, to ensure that we collect trajectories in which the robot has to move carefully around obstacles.
Additionally, we spawn objects grasped in the end-effector, with significant randomization including boxes, cylinders, spheres or even Objaverse meshes.
Finally, we found that naively imitating the output of the planner performs poorly as the planner output is not well suited for learning. Plans produced by AIT* often result in way-points that are far apart, creating large action jumps and sparse data coverage, making it difficult for networks to fit the data. To address this issue, we perform smoothing using cubic spline interpolation while enforcing velocity and acceleration limits, ensuring high data coverage via small action size limits for each transition.
\begin{algorithm}[h]
\small
\caption{Procedural Scene Generation}
\label{alg:scene_gen}
\begin{algorithmic}[1]
\Require Asset category generators $ \{ X_i(\textbf{p}) \}_{0,1..G}$  
\Require Number of scenes $N$
\Require Max objects per scene $K$
\Require Collision checker $Q$
\For {scene 1: N}
    \State Initialize scene $S = \{ \}$
    \State Sample number of assets $k \sim [1,...K]$
    \For {asset 1:k}
        \State Sample asset category $g \sim [0,..N]$
        \State Sample asset parameter $p$
        \State Sample asset $x \sim X_g(p)$
        \While {$Q(S, x)$}
            \For {each asset $s_i$ in $S$}
                \State $n_i$ = collision normal b/n $x$ and $s_i$
            \EndFor
            \State Effective collision normal $n = \sum n_i$
            \State Update $p$ so $X_g(p)$ center is shifted along $n$
        \EndWhile
        \State Add asset $x$ to scene $S$
    \EndFor
    \State yield scene S
\EndFor
\end{algorithmic}
\end{algorithm}

\subsection{Generalist Neural Policies}
We would like to obtain agents that can use diverse sets of experiences to plan efficiently in new settings.
In order to build generalist neural motion planning policies, we need an observation space amenable to sim2real transfer, and utilize an architecture capable of absorbing vast amounts of data.

\noindent \textbf{Observations}: We begin by addressing the sim2real transfer problem, which requires considering the observation and action spaces of the trained policy. With regards to observation, point-clouds are a natural representation of the scene for transfer~\citep{fishman2023motion,christen2023handoversim2real,jiang2024transic,chen2023visual,wu2023sim2real}, as they are 3D points grounded in the base frame of the robot and therefore view agnostic, and largely consistent between sim and real. 
We include proprioceptive and goal information in the observations, consisting of the current joint angles $q_t$, the target joint angles $g$, in addition to the point-cloud $PCD$.

\noindent \textbf{Network Architecture}: 
We design our policy $\pi$ (Fig.~\ref{fig:method}) to be a sequence model to imitate the expert using history which is useful for fitting privileged experts using partially observed data~\citep{dalal2023optimus}. In principle, any sequence modeling architecture could be used; we opt for LSTMs for their fast inference time and comparable performance to Transformers on our datasets. We operate the LSTM policy over joint embeddings of $PCD_t$, $q_t$, and $g$ with a history length of 2. 
We encode point-clouds using PointNet++~\citep{qi2017pointnet++}, while we use MLPs to encode $q_t$ and $g_t$. We follow the design decisions from M$\pi$Nets regarding point-cloud observations: we segment the robot point-cloud, obstacle point-cloud and the target robot point-cloud before passing it to PointNet++. For each time-step, we concatenate the embeddings of each of the observations together into one vector and then pass them into the LSTM for action prediction.
For the output of the model, note that sampling-based motion planners such as AIT* are multi-modal: for the same scene they may give wildly different plans for each run. As a result, we require an expressive, multi-modal distribution to capture such data
To that end, Neural MP predicts the parameters of a Gaussian Mixture Model (GMM) distribution over delta joint angles ($\Delta q_{t+1}$), which are used to sample the next target joint way-point during deployment: $q_{t+1} = q_t + \Delta q_{t+1}$. 
As we show in our experiments, for fitting to sampling-based motion planning, minimizing the negative log-likelihood of the GMM outperforms the PointMatch loss from M$\pi$Nets, Diffusion~\citep{saha2023edmp} and Action-chunking~\citep{zhao2023learning} (Sec.~\ref{sec:result}).

\subsection{Deploying Neural Motion Planners}
~\label{subsec:deployment}
\noindent \textbf{Test time Optimization} While our base neural policy is capable of solving a wide array of challenging motion planning problems, we would still like to ensure that these motions are safe to be deployed in real environments. We enable this property by combining our learned policy with a simple light-weight optimization procedure at inference time.
This relies on a simple model that 
assumes the obstacles do not move and the controller can accurately reach the target way-points. Given world state $s = [q, e]$ (e is the environment state), the predicted world state is $s' = [q + \hat{a}$, e] where $\hat{a}$ is the policy prediction. With this forward model, we can sample N trajectories from the policy using the initial scene point-cloud to provide the obstacle representation and estimate the number of scene points that intersect the robot
using the robot Signed Distance Function (SDF). We then optimize for the path with the least robot-scene intersection in the environment:
$\min_{\tau \sim \rho_{\pi_{\theta}}} \sum_{t=1}^{t=T} \sum_{k=1}^{k=K} \mathbbm{1} \{ SDF_{q_t}(PCD_{O}^{k}) < \epsilon\}$
in which $\rho_{\pi_{\theta}}$ is the distribution of trajectories under policy $\pi_{\theta}$ with a linear model as described above, $PCD_{O}^{k}$ is the kth point of the obstacle point-cloud (with max $K=4096$ points) and $SDF_{q_t}$ is the SDF of the robot at the current joint angles. In practice, we optimize this objective with finite samples in a single step, computing the with minimal objective value by selecting the path with minimal objective value.

\noindent \textbf{Sim2real and Deployment}
During execution, we linearly interpolate predicted joint waypoints and execute using a joint space controller. Our setup includes four calibrated Intel RealSense depth cameras positioned at the table's corners. 
To produce the segmented point cloud for input to the model, we compute a point-cloud using the 4 cameras, segment out the partial robot cloud using a mesh-based representation of the robot to exclude points. 
We generate the current robot and target robot point clouds by sampling points from the robot mesh and place them into the scene. For real-world vision-based collision checking, we calculate the SDF between the point cloud and the spherical representation of the robot, enabling fast SDF calculation (0.01-0.02s).

\section{Experimental Setup}
\begin{figure}
    \centering
    \vspace{4pt}
    \begin{subfigure}[b]{1\linewidth}
        \centering
        \includegraphics[width=1.\linewidth]{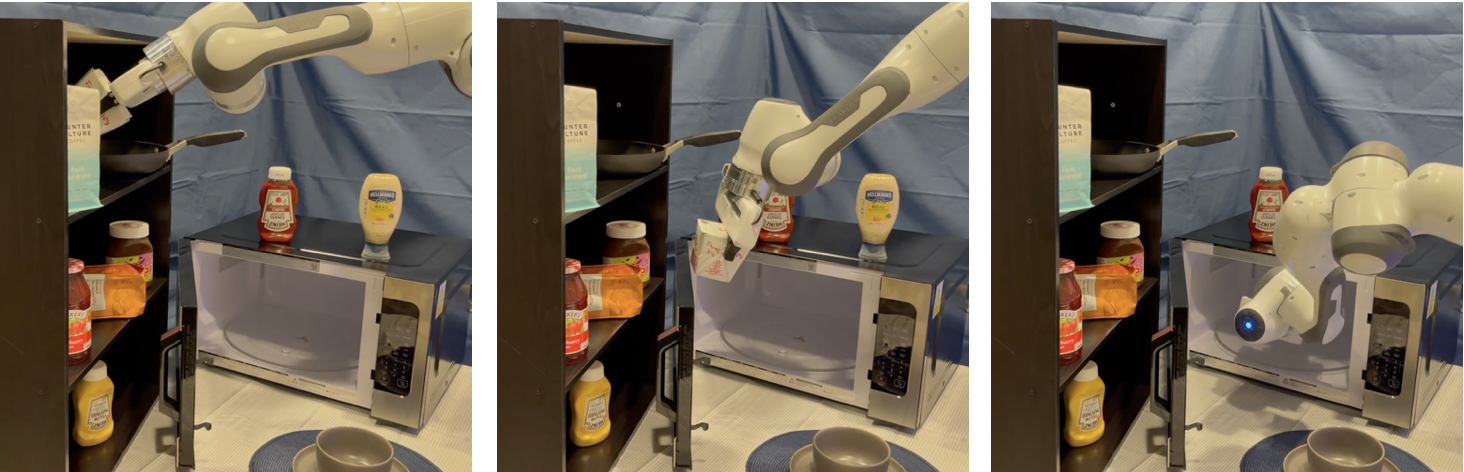}
        \vspace{-15pt}
        \caption{\small Sampling-based planners struggle with tight spaces, 
        a regime in which Neural MP performs well.
        }
        \label{fig:sub1}
    \end{subfigure}
    \hfill
    \begin{subfigure}[b]{1.0\linewidth}
        \centering
        \includegraphics[width=1.\linewidth]{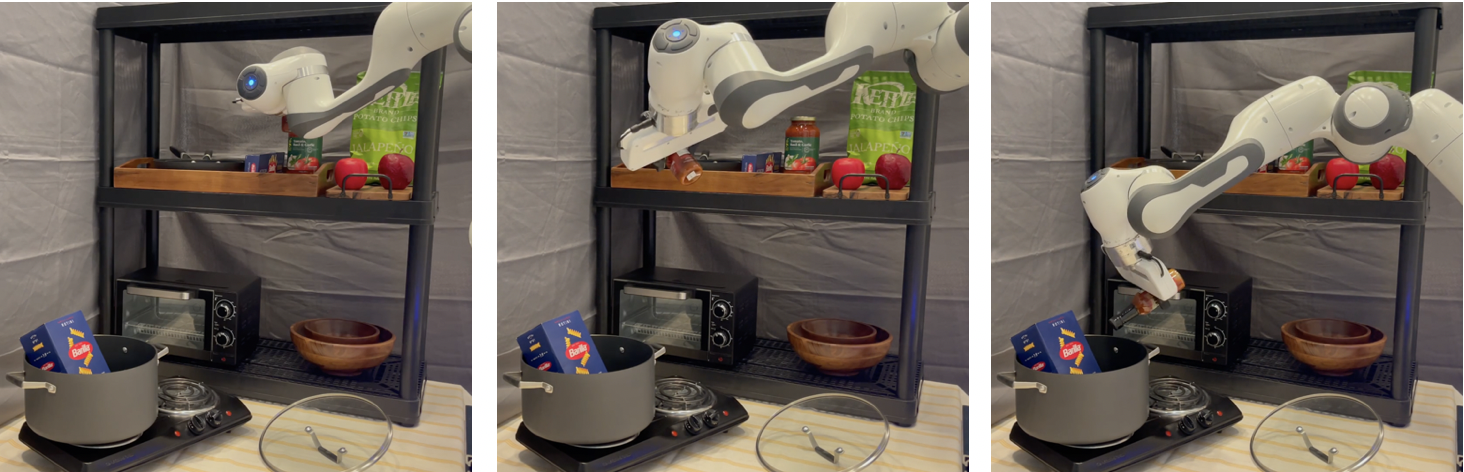}
        \vspace{-15pt}
        \caption{\small Our method is able to motion plan with objects in-hand, a crucial skill for manipulation.}
        \label{subfig: in hand rollout}
    \end{subfigure}
    \hfill
    \begin{subfigure}[b]{1\linewidth}
        \centering
        \includegraphics[width=1.\linewidth]{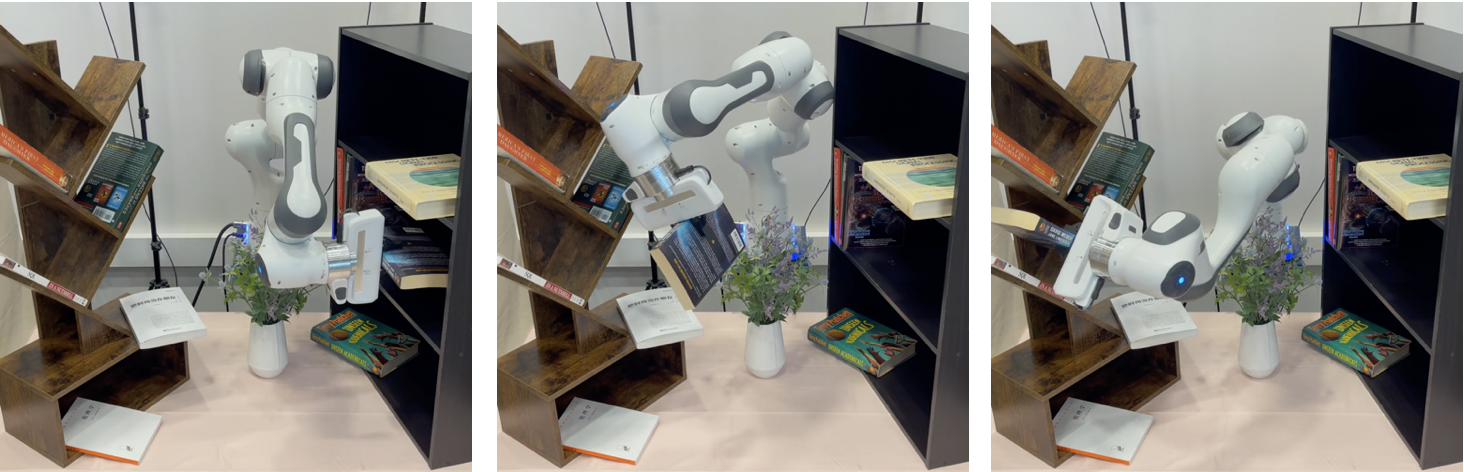}
        \vspace{-15pt}
        \caption{\small Our policy has not been trained on this bookcase, yet it is able to insert the book into the correct location.}
        \label{fig:bookcase rollout}
    \end{subfigure}
    \vspace{-15pt}
    \caption{\small Emergent Capabilities of Neural MP}
    \vspace{-15pt}
    \label{fig:rollouts}
\end{figure}
In our experiments, we consider motion planning in four different real world environments containing obstacles. Importantly, these are not included as part of the training set, and thus the policy needs to generalize to perform well on these settings. We begin by describing our environment design, then each of the environments, and finally our evaluation protocol and comparisons.

\noindent \textbf{Environment Design}
Our four environments are 1) \textbf{Bins}: moving in-between, around and inside two different industrial bins 2) \textbf{Shelf}: moving in-between and around the rungs of a black shelf 3) \textbf{Articulated}: moving inside and within cubbies, drawers and doors 4) \textbf{in-hand}: moving between rungs of a shelf while holding different objects. We evaluate on 16 start-goal robot configuration pairs in each environment, in each randomizing the poses of obstacles and objects, the table height and the type of objects used.

\noindent \textbf{Evaluation Protocol}
For fairness,we evaluate all methods on open loop planning, though our method and M$\pi$Nets, are capable of executing trajectories in a closed loop manner. For neural planners, this involves generating an open loop path by passing the agent's predictions back into itself. We then execute the plans on the robot, recording the success rate of the robot in reaching the goal, its collision rate and the time taken to reach the goal. We follow M$\pi$Nets' definition of success rate: reaching within 1cm and 15 degrees of the goal end-effector pose of the target goal configuration while also not colliding with anything in the scene. In practice, our policy achieves orientation errors significantly below this threshold, 2 degrees or less.

\noindent \textbf{Comparisons}
We propose three baselines for real-world comparisons to evaluate different aspects of our method's capabilities. We compare against sampling-based motion planning, which is expensive to run but has strong guarantees on performance. The first baseline is the expert we use to train our model, AIT* with 80 seconds of planning time. We run this planner with the same vision-based collision checker used by our method in the real world.
AIT*-80s is impractical to deploy in most settings due to its significant planning time. Thus, we compare to a faster variant of AIT* with 10 seconds of planning time, which uses comparable time to our method (Note: AIT*-3s is unable to find a plan for any real world task). 
Next, we compare against Curobo~\citep{sundaralingam2023curobo}, a SOTA motion-generation method which performs GPU-parallelized optimization and is orders of magnitude faster than AIT*. We run this baseline with a voxel-based collision checker and optimize its voxel resolution per task due to its sensitivity to that parameter. Finally, we compare against the SOTA neural motion planning approach, M$\pi$Nets.

\begin{table}[]
    \centering
    \begin{tabular}{lcccc}
    \toprule
     & Bins & Shelf & Articulated & Average \\
    
    \multicolumn{2}{l}{\textit{Sampling-based Planning}:}\vspace{0.4em}\\
    \texttt{AIT*-80s~\citep{strub2020adaptively}} & 93.75 & 75.0 & 50.0 & 72.92\\
    \texttt{AIT*-10s (fast)~\citep{strub2020adaptively}} & 75.0 & 37.5 & 25.0 & 45.83\\
    \multicolumn{2}{l}{\textit{Optimization-based Planning}:}\vspace{0.4em}\\
    \texttt{Curobo~\citep{sundaralingam2023curobo}} & 93.75 & 81.25 & 62.5 & 79.17\\
    \midrule
    \multicolumn{2}{l}{\textit{Neural}:}\vspace{0.4em}\\
    \texttt{M$\pi$Nets~\citep{fishman2023motion}} & 18.75 & 25.0 & 6.25 & 16.67  \\
    \midrule
    \texttt{\textbf{Ours}-Base Policy} & 81.25 &	75.0 &	43.75 & {66.67} \\
    \textbf{\texttt{Ours}} & \textbf{100} & \textbf{100} & \textbf{87.5} & \textbf{95.83}  \\ 
    \bottomrule
    \end{tabular}
    \caption{\small 
    Neural MP achieves SOTA motion planning results, demonstrating greater improvement as the task complexity grows.}
    \label{tab: main results}
    \vspace{-10pt}
\end{table}
\section{Results}
\label{sec:result}
To guide our evaluation, we pose a set of experimental questions. 1) Can a single policy trained in simulation learn to solve complex motion planning tasks in the real world? 2) How does Neural MP compare to SOTA neural planning, sampling-based and trajectory optimization planning approaches? 3) How well does Neural MP extend to motion planning tasks with objects in-hand? 4) Can Neural MP perform dynamic obstacle avoidance? 5) What are the impacts key ingredients of Neural MP have on its performance? 

\noindent \textbf{Free Hand Motion Planning}
In this set of experiments, we evaluate motion planning the robot's hand is empty (Table~\ref{tab: main results}). We find that our base policy on its own performs comparably to AIT*-80s (66.67\% vs. 72.92\%) while only using 1s of planning time. When we include test-time optimization (3s of planning), we find that across all three tasks, Neural MP achieves the best performance with a 95.83\% success rate. In general, we find that Bins is the easiest task, with the sampling/optimization-based methods performing well, Shelf is a bit more difficult as it requires simultaneous vertical and horizontal collision avoidance, while Articulated is the most challenging task as it contains a diverse set of obstacles and tight spaces. Neural MP performs well across each task as it is trained with a diverse set of parametric objects that cover the types of real-world obstacles we encounter and it also incorporates complex meshes which cover the irregular geometries of the additional objects we include. 

In our experiments, M$\pi$Nets performs poorly across the board. We attribute this finding to 1) M$\pi$Nets is only trained on data in which the expert goes from tight spaces to tight spaces, which means the fails to generalize well to start/goal configurations in free space and 2) the end-effector point matching loss in M$\pi$Nets fails to distinguish between 0 and 180 degree rotations of the end-effector, so the network has not learned how to match ambiguous target end-effector poses. Note, even if we change the success rate metric for M$\pi$Nets to count 180 degree flipped end-effector poses as successes as well, the average success rate of M$\pi$Nets only improves from 16.67\% to 29.17\% - it is still far below the other methods. Meanwhile, failure cases for AIT* and Curobo are tight spaces for which vision-based collision checking is inaccurate and the probability of sampling/optimizing for a valid path is low. In contrast, our method performs well on each task, generalizing to 48 different unseen environment, scene, obstacle and joint configuration combinations.

\noindent \textbf{In-Hand Motion Planning}
In this experiment, we extend our evaluation to motion planning with objects in-hand, a crucial capability for manipulation. We evaluate Neural MP against running the neural policy without test time optimization and without including any Objaverse data, achieving 81\% performance vs. 31\% and 44\%. We visualize an example trajectory in Fig.~\ref{fig:rollouts}. Our method performs well on in-distribution objects such as the book and board game, but struggles on out of distribution objects such as the toy sword, which is double the size of objects at training time. 
We additionally deploy our method on significantly out of distribution objects such as the bookcase (Fig.~\ref{fig:bookcase rollout}) and find that Neural MP generalizes well to in-hand motion planning tasks such as inserting the book in the right rung. 

This experiment also serves as an ablation of our method, demonstrating the importance of test time optimization on out of distribution scenarios. For these tasks, the base policy performance results in a large number of collisions as two of the in-hand objects are out of distribution (sword and board game), but the optimization step is able to largely remove them and produce clean behavior that reaches the target without colliding. Additionally, this experiment demonstrates that the Objaverse data is crucial for the success of our method in the real world. Models trained only on cuboid-based parametric assets fail to generalize to the complexity of the real world (43.75\%) while those trained on Objaverse perform well (81.25\%), highlighting the importance of incorporating Objaverse meshes into scene generation.

\noindent \textbf{Dynamic Motion Planning}
In many real-world scenarios, the environment may be changing as the motion planner is acting.
We test how well Neural MP can motion plan in such settings by introducing obstacles into the environment while the motion planner is moving to a goal. We evaluate the motion planner on four different goals with three different added obstacles (drawer, monitor and pot). To handle dynamic obstacles, we run the neural motion planner closed loop and perform single-step test-time optimization. We compare against M$\pi$Nets and find that Neural MP performs $53\%$ better ($63.33\%$ vs. $10\%$), performing particularly well on the drawer and pot object while struggling on the monitor object which is significantly taller. We also qualitatively evaluate Neural MP on two significantly more challenging motion planning tasks in which we continuously move the obstacle into the robot's path and demonstrate that it can adjust its behavior to avoid collisions while reaching the goal.

\noindent \textbf{Comparisons to Learning-based Motion Planners}
We next evaluate how Neural MP compares to two additional learning-based methods, MPNets~\citep{qureshi2019motion} and EDMP~\citep{saha2023edmp} (a Diffusion-based neural motion planner) as well as M$\pi$Nets~\citep{fishman2023motion} in simulation. We compare these three neural motion planning methods in simulation trained on the same dataset (from M$\pi$Nets) of 3.27 million trajectories. We train policies on the Global expert data and the Hybrid datasets and then evaluate on 5400 test problems across the Global, Hybrid and Both solvable subsets. We include numerical results Tab.~\ref{tab:sim results}, with numbers for the baselines taken from the EDMP and M$\pi$Nets papers. We find that across the board, Neural MP is the best learning-based motion planning method, outperforming both EDMP and M$\pi$Nets. We attribute this to the use of sequence modelling, the ability of the GMM to fit multimodal data and test-time optimization to prune out any collisions.

\begin{table}
    \vspace{5pt}
    \centering
        \begin{tabular}{lcccc}
        \toprule
         & Global & Hybrid & Both & Average \\
        \midrule
        \multicolumn{2}{l}{\texttt{MPNet~\citep{qureshi2019motion}}}\vspace{0.4em}\\
        \textit{Hybrid Expert} & 41.33 & 65.28 & 67.67 & 58.09\\
        \multicolumn{2}{l}{\texttt{M$\pi$Nets~\citep{fishman2023motion}}}\vspace{0.4em}\\
        \textit{Global Expert} & 75.06 & 80.39 & 82.78 & 79.41\\
        \textit{Hybrid Expert} & 75.78 & 95.33 & 95.06 & 88.72\\
        \multicolumn{2}{l}{\texttt{EDMP~\citep{saha2023edmp}}}\vspace{0.4em}\\
        \textit{Global Expert} & 71.67 & 82.84 & 82.79 & 79.10\\
        \textit{Hybrid Expert} & 75.93 & 86.13 & 85.06 & 82.37\\
        \midrule
        \multicolumn{2}{l}{\texttt{Ours}}\vspace{0.4em}\\
        \textit{Global Expert} & \textbf{77.93} & 85.50 & 87.67 & 83.70\\
        \textit{Hybrid Expert} & 76.33 & \textbf{97.28} & \textbf{96.78} & \textbf{90.13}\\
        \bottomrule
        \end{tabular}
    \caption{\small Performance comparison of neural motion planning methods across 5400 test problems in the M$\pi$Nets dataset in simulation. Neural MP achieves the SOTA results on these tasks.}
    \label{tab:sim results}
    \vspace{-10pt}
\end{table}

\noindent \textbf{Ablations}
We run ablations of components of our method (training objective, observation composition) in simulation to evaluate which have the most impact, evaluating performance on held out scenes. For training objective, we find that GMM (ours) outperforms L2 loss, L1 loss, and PointMatch Loss (M$\pi$Nets) by (7\%, 12\%, and 24\%) respectively. We find that including both $q$ and $g$ vectors is crucial for performance as we observe a 62\%, 65\%, and 75\% performance drop when using only $g$, only $q$ and neither $q$ nor $g$ respectively. 
\section{Discussion and Limitations}
\label{sec:conclusion}
In this work, we present Neural MP, a method that builds a data-driven policy for motion planning by scaling procedural scene generation, distilling sampling-based motion planning and improving at test-time via refinement. Our model demonstrably improves over the sampling-based planning in the real world, operating 2.5x-20x faster than AIT* while improving by over 20\% in terms of motion planning success rate. Notably, our model generalizes to a wide distribution of task instances and demonstrates favorable scaling properties. 

In summary, this section proposes large-scale training in simulation as a mechanism for training general purpose policies for robotics. The core ingredients are diverse data generation via procedural scene generation, traditional planner experts for supervision and large-scale policy learning with expressive, multimodal outputs. While we demonstrate strong real-world generalization results for motion planning, this framework is limited in its ability to generalize combinatorially, leverage common-sense knowledge for reasoning and operate in regimes that the planner experts fails to supervise. To that end, in the next part, we design a framework for unifying modularity with scale to produce generalist robot systems that will be capable of zero-shot long-horizon manipulation.
\clearpage

\part{A Unifying Framework for Building Generalist Agents by Combining Modularity and Scale}
\chapter{Local Policies Enable Zero-shot Long-horizon Manipulation}
\label{chap:manipgen}
\begin{figure}[h]
    \includegraphics[width=1\linewidth]{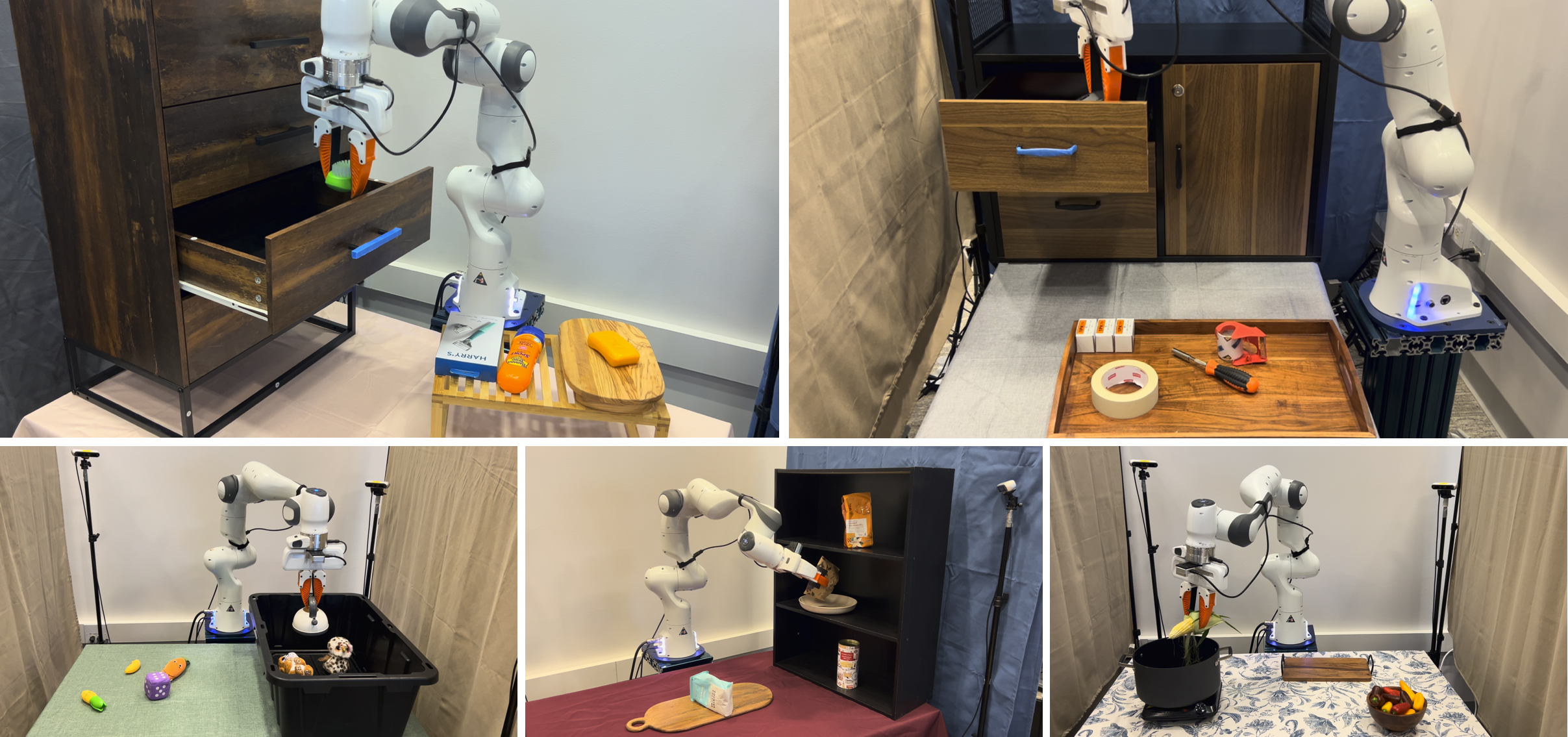}
      \centering
      \vspace{-5pt}
      \caption{\small \textbf{Zero-shot Long-horizon Manipulation} Our approach trains a library of generalist manipulation skills in simulation and transfers them zero-shot to long-horizon manipulation tasks. We show a single, text-conditioned agent can manipulate unseen objects, in arbitrary poses and scene configurations, across long-horizons in the real world, solving challenging manipulation tasks with complex obstacles. }
      \label{fig:teaser}
\end{figure}
\section{Introduction}

\label{sec:intro}

How can we develop generalist robot systems that plan, reason, and interact with the world like humans? Tasks that humans solve during their daily lives, such as those shown in Figure~\ref{fig:teaser}, are incredibly challenging for existing robotics approaches. Cleaning the table, organizing the shelf, putting items away inside drawers, etc. are complex, long-horizon problems that require the robot to act capably and consistently over an extended period of time. 
Furthermore, such a generalist robot should be able to do so without requiring task-specific engineering effort or demonstrations.
Although large-scale data-driven learning has produced generalists for vision and language~\cite{openai2023gpt4}, such models don't yet exist in robotics due to the challenges of scaling data collection. It often takes significant manual labor cost and years of effort to just collect datasets on the order of 100K-1M trajectories~\cite{collaboration2023open, khazatsky2024droid,ebert2021bridge,walke2023bridgedata}. Consequently, generalization is limited, often to within centimeters of an object's pose for complex tasks~\cite{zhao2023learning,fu2024mobile}.

Instead, we seek to use a large-scale approach via simulation-to-reality (sim2real) transfer, a cost-effective technique for generating vast datasets that has enabled training generalist policies for locomotion which can traverse complex, unstructured terrain~\cite{lee2020learning,kumar2021rma,agarwal2023legged,zhuang2023robot,cheng2023parkour,hoeller2024anymal}. While sim2real transfer has shown success in industrial manipulation tasks~\cite{tang2023industreal,tang2024automate,jiang2024transic}, including with high-dimensional hands~\cite{akkaya2019solving,handa2022dextreme,lum2024dextrah,chen2022visual}, these efforts often involve training and testing on the same task in simulation. Can we extend sim2real to open-world manipulation, where robots need to solve any task from text instruction?
\setcounter{figure}{1}
\begin{figure*}[t]
\includegraphics[width=\textwidth]{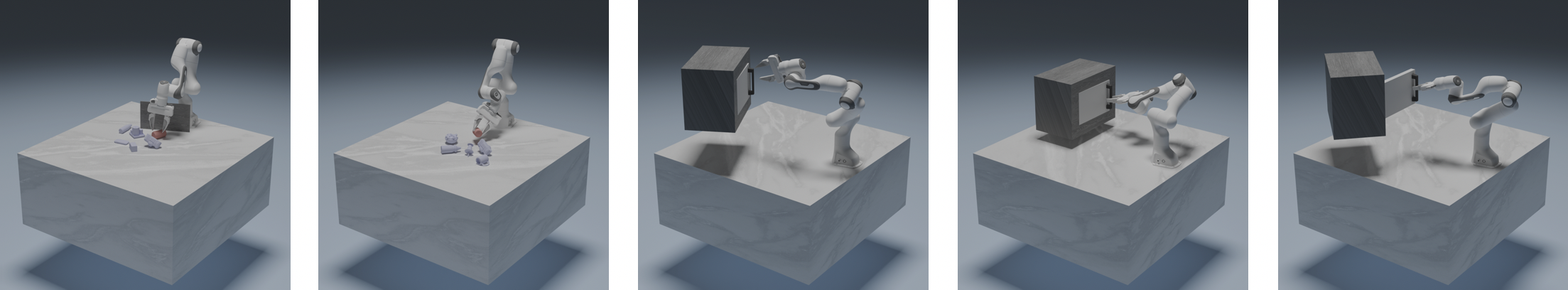}
\vspace{-15pt}
\caption{\small \textbf{Training Environments} We train local policies (left to right) on picking, placing, handle grasping, opening and closing.}
\label{fig:sim vis}
\end{figure*}
The core bottlenecks are: 1) accurately simulating contact dynamics~\cite{todorov2012mujoco} - for which strategies such as domain randomization~\cite{akkaya2019solving,andrychowicz2020learning}, SDF  contacts~\cite{narang2022factory,tang2023industreal,tang2024automate}, and real world corrections~\cite{jiang2024transic} have shown promise, 2) generating all possible scene and task configurations to ensure trained policies generalize and 3) acquiring long-horizon behaviors themselves, which may require potentially intractable amounts of data for as the horizon grows.

To address points 2) and 3), our solution is to note that for many manipulation tasks of interest, the skill can be simplified to two steps: achieving a pose near a target object, then performing manipulation. The key idea is that of \textit{locality of interaction}. 
Policies that observe and act in a region local to the target object of interest are by construction: 
\newline $\bullet$ \textbf{absolute pose invariant}: they reason about a much smaller set of relative poses between the objects and the robot.
\newline $\bullet$ \textbf{skill order invariant}: transition from the termination set of one policy and initiation set of the next via motion planning.
\newline $\bullet$ \textbf{scene configuration invariant}: they solely observe the local region around the point of interaction. 

We propose a novel approach that leverages the strong generalization capabilities of existing foundation models such as Visual Language Models (VLMs) for decomposing tasks into sub-problems~\cite{openai2023gpt4}, processing and understanding scenes~\cite{ren2024grounded} and planning collision-avoidant motions~\cite{dalal2024neuralmp}. 
Specifically, given a text prompt, our approach outputs a plan to solve the task (using a VLM), estimates where to go and moves the robot accordingly (using motion planning) and deploys local policies to perform interaction. As a result, a simple scene generation approach (Fig.~\ref{fig:sim vis}) can produce strong transfer results across many manipulation tasks (Fig.~\ref{fig:teaser}).

\looseness=-1
Our contribution is an approach to training agents at scale solely in simulation that are capable of solving a vast set of long-horizon manipulation tasks in the real world \textit{zero-shot}. Our method generalizes to unseen objects, poses, receptacles and skill order configurations. To do so, our method, ManipGen, 1) introduces a novel policy class for sim2real transfer 2) proposes techniques for training policies at scale in simulation 3) and deploys policies via integration with VLMs and motion planners. 
We perform a thorough, real world evaluation of ManipGen on \textbf{50} long-horizon manipulation tasks in \textbf{five} environments with up to \textbf{8} stages, achieving a success rate of \textbf{76\%}, outperforming SayCan, OpenVLA, LLMTrajGen and VoxPoser by \textbf{36\%}, \textbf{76\%}, 
 \textbf{62\%} and \textbf{60\%}.

\begin{figure*}[h!]
    \centering
    \vspace{5pt}
    \includegraphics[width=\linewidth]{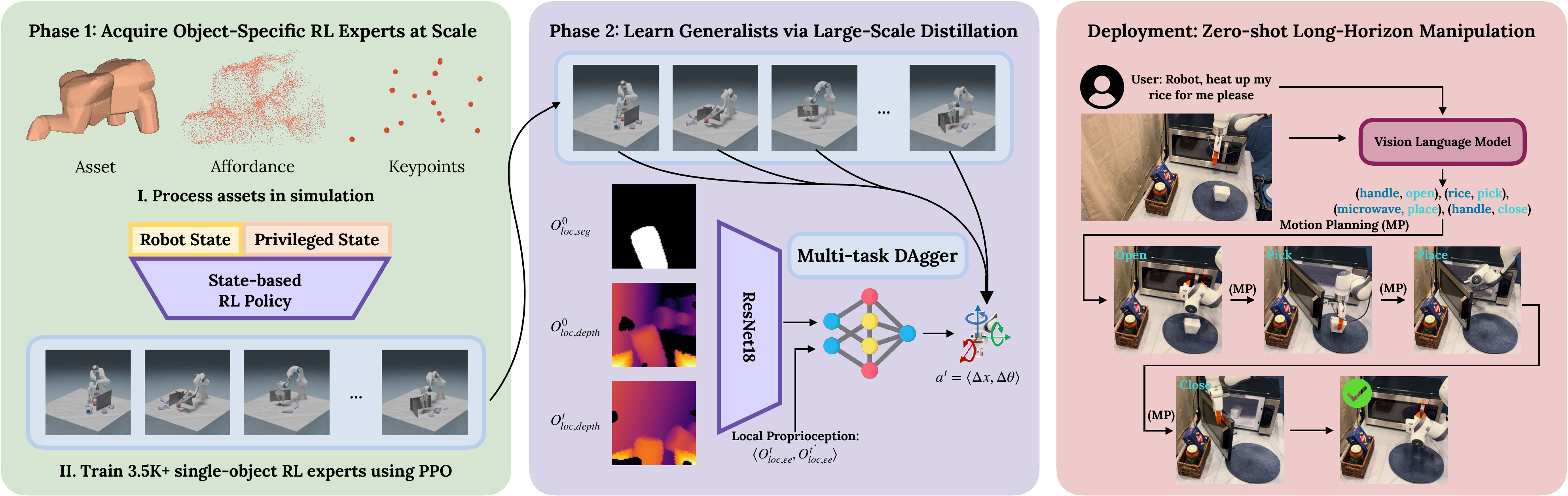}
    \vspace{-15pt}
    \caption{\small \textbf{ManipGen Method Overview} (\textit{left}) Train 1000s of RL experts in simulation using PPO (\textit{middle}) Distill single-task RL experts into generalist visuomotor policies via DAgger (\textit{right}) Text-conditioned long-horizon manipulation via task decomposition (VLM), pose estimation and goal reaching (Motion Planning) and sim2real transfer of local policies}
    \label{fig:main fig}
\end{figure*}
\section{Related Work}
\label{sec:related work}
\noindent \textbf{Long-horizon Robotic Manipulation}
Sense-Plan-Act (SPA) has been explored extensively over the past 50 years~\cite{center1984shakey,paul1981robot,whitney1972mathematics,vukobratovic1982dynamics,kappler2018real,murphy2019introduction}. Traditionally, SPA assumes access to accurate state estimation, a well-defined model of the environment and low-level control primitives. SPA, while capable of generalizing to a broad set of tasks, can require manual engineering and systems effort to set up~\cite{garrett2020online}, struggles with contact-rich interactions~\cite{mason2001mechanics,whitney2004mechanical} and fails due to state-estimation errors~\cite{kaelbling2013integrated}. By contrast, our method can be deployed to new tasks using generalist models which have minimal setup cost, train polices for contact-rich interactions and handle state-estimation issues by training with significant local randomization.

\noindent \textbf{Zero/Few-shot Manipulation Using Foundation Models}
The robotics community has begun to investigate VLM's capabilities for controlling robots in a zero/few-shot manner~\cite{ahn2022say,huang2022inner,huang2022language,wu2023tidybot,lin2023text2motion,huang2023voxposer,zhangboss,liu2023llm+p,dalal2024psl}. Work such as SayCan~\cite{ahn2022say} and TidyBot~\cite{wu2023tidybot} are similar to our own. They behavior clone / design a library of skills and use LLMs to perform task planning over the set of skills. Our work focuses primarily on designing the structure of skills for low-level control, decomposing them into motion planning and sim2real local policies. On the other hand, works such as LLMTrajGen~\cite{kwon2024language} and CoPa~\cite{huang2024copa} directly prompt VLMs to output sequences of end-effector poses, but are limited to short horizon tasks. Finally, PSL~\cite{dalal2024psl} and Boss~\cite{zhangboss} use LLMs to accelerate the RL training process for long-horizon tasks, yet must train on the test task, unlike our method which can solve a wide array of manipulation tasks zero-shot.

\noindent \textbf{Sim2real approaches in robotics}
Transfer of RL policies trained with procedural scene generation has produced generalist robot policies for locomotion~\cite{lee2020learning,kumar2021rma,zhuang2023robot,agarwal2023legged,cheng2023parkour}. However, the robot is often trained for a single skill, such as walking, or a limited set of similar skills, such as walking at different velocities or headings. Sim2real transfer has also been explored for transferring dexterous manipulation skills~\cite{akkaya2019solving,andrychowicz2020learning,handa2022dextreme,agarwal2023dexterous,lum2024dextrah} and contact-rich manipulation~\cite{tang2023industreal,tang2024automate,jiang2024transic}. In our work, we train a variety of skills for manipulation and demonstrate zero-shot capabilities on a large set of unseen tasks. We outperform methods that use end-to-end sim2real transfer~\cite{peng2017simtoreal} as well as real world corrections~\cite{jiang2024transic}, ManipGen is orthogonal to human correction approaches, and can benefit from real-world data as well.
  
\section{Building a Generalist Manipulation Agent}
\label{sec:methods}
To build agents capable of generalizing to a wide class of long-horizon robotic manipulation tasks, we propose a novel approach (ManipGen) that hierarchically decomposes manipulation tasks, takes advantage of the generalization capabilities of foundation models for vision and language and uses large-scale learning with our proposed policy class to learn manipulation skills. We begin by describing our framework (Fig.~\ref{fig:main fig}) and formulate local policies. We then discuss how to train local policies for sim2real transfer. 
Finally, we outline deployment: integrating VLMs, Motion Planning and sim2real policy learning to foster broad generalization. 

\subsection{Framework}
We can decompose any task the robot needs to complete into a problem of learning a set of temporally abstracted actions (skills) as well as a policy over those skills~\cite{SUTTON1999181}. Given a language goal $g$, and observation $O$, we can select our high-level policy, $p_{\theta}(g_k | g, O)$ to be a pre-trained VLM, where $g_k$ is the $k$'th language subgoal. The choice of skill will be extracted from $g_k$ below. State-of-the-art VLMs have been shown to be capable of decomposing robotics tasks into high-level language subgoals~\cite{ahn2022say,huang2022inner,huang2022language,wu2023tidybot} because they are trained using a vast corpus of internet-scale data and have captured powerful, visually grounded semantic priors for what various real world tasks look like. 

Any policy class can be used to define the skills, denoted as $p_{\phi_k}(a^t|g_k, O^t)$, which take in the kth sub-goal $g_k$ and current observation $O^t$.
However, note that many manipulation skills (\textit{e.g.} picking, pushing, turning, etc.) can be decomposed into a policy $\pi_{reach}$ to achieve target poses near objects, $X_{targ,k}$, followed by policy $\pi_{loc}$ for  contact-rich interaction. Accordingly, $p_{\phi_k}(a^t|g_k, O^t) = \pi_{reach}(\tau_{reach} | g_k, O^t) \pi_{loc}(a_{loc}^t | O_{loc}^t)$. To implement $\pi_{reach}$, we need to interpret language sub-goals $g_k$ to take the robot from its current configuration $q_{k,i}$ to some target configuration $q_{k,f}$ such that $X_{ee}$ (the end-effector pose) is close to $X_{targ,k}$. Thus, we structure the VLM's sub-goal predictions, $g_k$, as tuples containing the following information (object, skill). We then interpret these plans into robot poses by pairing any language conditioned pose estimator or affordance model (to predict $X_{targ,k}$) with an inverse kinematics routine (to compute $q_{k,f}$). Motion planning is used to implement $\pi_{reach}$ by predicting trajectories $\tau_{reach}$ to achieve the target configuration $q_{k,f}$ while avoiding collisions. 

Finally, we instantiate local policies ($\pi_{loc}$) to be invariant to robot poses as well as object poses, order of skill execution and scene configurations with: 1) initialization region $s_{init}$ near a target region/object of interest which has pose $X_{targ,k}$, 2) local observations $O_{loc}^t$, independent of the absolute configuration of the robot and scene and only observing the environment around the interaction region and 3) actions $a_{loc}^t$relative to the local observations. Overall:
\begin{align*}
\pi_{loc}(a_{loc}^t | O_{loc}^t), s_{init} = \{s \mid ||X_{ee} - X_{targ,k} ||^2 < \epsilon\}
\end{align*}

\subsection{Training Local Policies for Sim2Real Manipulation}
To train local policies, we adapt the standard two-phase training approach~\cite{agarwal2023dexterous,cheng2023parkour,zhuang2023robot,uppal2024spin,lum2024dextrah,jiang2024transic} in which we first train state-based expert policies using RL, then distill them into visuomotor policies for transfer. 
Although local policies can generalize automatically across scene arrangements, robot configurations, and object poses, they must be trained across a wide array of objects to foster object-level generalization. To do so, we train a vast array of \textit{single-object} state-based RL policies and then distill them into \textit{generalist} visuomotor policies per skill.
 
While such local policies can cover a broad set of manipulation skills (pick and place, articulated/deformable object manipulation, assembly, etc.), in this work, we focus on training the following skills $\pi_{loc}$: \textbf{pick}, \textbf{place}, \textbf{grasp handle}, \textbf{open} and \textbf{close} (Fig.~\ref{fig:sim vis}) as a minimal skill library to demonstrate generalist manipulation capabilities for a specific class of tasks. \textbf{Pick} grasps any free rigid objects. \textbf{Place} sets the object down near the initial pose. \textbf{Grasp Handle} grasps the handle of any door or drawer. \textbf{Open and Close} pull or push doors and drawers to open or close them. 

To train robust local policies via RL, they require a diverse set of training environments, carefully designed observations and action spaces and well-defined reward functions enabling them to acquire behaviors in a manner that will transfer to the real world. We describe how to in this section.
 
\noindent \textbf{Data Generation}
We need to first specify a set of objects to manipulate, an environment, and an initial local state distribution. 
For pick/place, we train on 3.5K objects from UnidexGrasp~\cite{xu2023unidexgrasp}, randomly spawned on a table top. To ensure local policies can learn obstacle avoidance and constrained manipulation, we spawn clutter objects and obstacles in the scene. We sample initial poses in a half-sphere, with the gripper pointing toward the object (for picking) and near the placement location (for placing). For local articulated object manipulation, the region of interaction only contains the handle (2.6K objects of Partnet~\cite{mo2019partnet}) and door/drawer surface (designed as cuboids). We randomize the size, shape, position, orientation, joint range, friction and damping coefficients, covering a wide set of real world articulated objects. We sample initial poses in a half-sphere around the handle (for grasp handle) and a randomly sampled initial joint pose (open/close). Finally we collect valid pre-grasp poses (antipodal sampling~\cite{sundermeyer2021contact}) for picking and grasping handles and rest poses (from UnidexGrasp) for learning placing.

\noindent \textbf{Observations} 
We use a single observation space for all RL experts, accelerating learning by incorporating significant amounts of privileged information.
Blind local policies can struggle to learn to manipulate objects with complex geometries as it is often necessary to have some notion of object shape to know how to manipulate. Thus, we propose to use a low-dimensional representation of the object shape by performing Farthest Point Sampling (FPS) on the object mesh with a small set number of desired key-points K (16). 
Furthermore, to ease the burden of credit assignment and thereby accelerate learning, we incorporate the individual reward components $\{\mathbf{r}\}$ and an indicator for the final observation $\mathbbm{1}\{t=T\}$.
RL observations are $O^t=\langle X_{ee}^t, \dot{X_ee}^t, X_{obj}^t, \{FPS_{obj}^t\}_{k=1}^K, \{\mathbf{r}\}^t, \mathbbm{1}\{t=T\}\rangle$

\noindent \textbf{Actions} We use the action space from Industreal~\cite{tang2023industreal} which has been shown to successfully transfer manipulation policies from sim2real for precise assembly tasks. Our policies predict delta pose targets for a Task Space Impedance (TSI) controller, where $a = [\Delta x; \Delta \theta]$, where $\Delta x$ is a position error and $\Delta \theta$ is a axis-angle orientation error. 

\noindent \textbf{Rewards} We train RL policies ($\pi_{loc_k}$) in simulation using reward functions we design to elicit the desired behavior per skill $k$.
We propose a reward framework that encompasses our local skills: $\mathbf{r} = c_{1}r_{ee} + c_{2}r_{obj} + c_{3}r_{ee,obj} + c_{4}r_{action} + c_{5}r_{succ}$. $\mathbf{r}$ specifies behavior for a broad range of manipulation tasks which involve moving the end-effector to specific poses (often right before contact) as well as a target object to desired poses and need to do so while maintaining certain constraints on the relative motion between the end-effector and the object as well as pruning out undesirable actions. 
$r_{ee}$ encourages reaching/maintaining specific end-effector poses, $r_{obj}$ restricts/encourages specific object poses or joint configurations, $r_{ee,obj}$ constrains the end-effector motion relative to the object(s) in the scene, $r_{action}$ restricts or penalizes undesirable actions and $r_{succ}$ is a binary success reward.  Please see the website for detailed descriptions of the task specific reward functions.

\subsection{Generalist Policies via Distillation}
In order to convert single-object, privileged policies into real world deployable skills, we distill them into multi-object, generalist visuomotor policies using DAgger~\cite{ross2011reduction}.

\noindent \textbf{Multitask Online Imitation Learning}
Empirically the standard, off-policy version of DAgger with interleaved behavior cloning (to convergence) and large dataset collection does not perform well. The policy ends up modeling data from policies whose state visitation distributions deviate significantly from the current policy. On the other hand, on-policy variants of DAgger, which take a single gradient step per environment step~\cite{agarwal2023legged,agarwal2023dexterous,uppal2024spin,lum2024dextrah}, can produce unstable results in the multi-task regime since the policy only gets data from a single object in a batch. We introduce a simple variant of DAgger which smoothly trades off between the two extremes by incorporating a replay buffer of size $K$ that holds the last $K*B$ trajectories in memory.  Training alternates between updating the agent for a single epoch on this buffer and collecting a batched set of trajectories (size $B$) from the environment for the current object. 

\noindent \textbf{Observation Space Design for Locality}
For local policies to transfer effectively to the real robot, the observation space and augmentations must be designed with transfer in mind. To imitate a privileged expert, our observation space must be expressive - providing as much information as possible to the agent. The observations must also be local to enable all of the properties of locality, and augmentations must ensure the policy is robust to noisy real world vision.

Local observations use wrist camera depth maps. Depth maps transfer well from sim2real for locomotion~\cite{agarwal2023legged,zhuang2023robot,cheng2023parkour,uppal2024spin}, and wrist views are inherently local and improve manipulation performance~\cite{hsu2022vision,dalal2023optimus,robomimic2021}.
To further enforce locality, we clamp depth values to a max depth of $30cm$ and then normalize the values to between $0$ and $1$. 
Since local wrist-views often get extremely close to the object during execution, it can become difficult for the agent to understand the overall object shape. Thus, we include the initial local observation $O_{loc,depth}^0$ at every step with a segmentation mask of the target object ($O_{loc,seg}^0$) so that the local policy is aware of which object to manipulate.
We transform absolute proprioception into local by computing observations relative to the first time-step ($O_{loc,ee} = [X_{ee,t}^0-X_{ee}^0]$) and incorporate velocity information ($\dot{O_{loc,ee,t}}$), which improves transfer.
Our observation space is $\mathbf{O_{loc}^t} = \langle O_{loc,depth}^t, O_{loc,seg}^0, O_{loc,depth}^0, O_{loc,ee}^t, \dot{O_{loc,ee}^t} \rangle$.

\noindent \textbf{Augmentations} To enable robustness to noisy real world observations, namely edge artifacts and irregular holes, we augment the clean depth maps we obtain in simulation. For edge artifacts, in which we observe dropped pixels and noisiness along edges, we use the correlated depth noise via bi-linear interpolation of shifted depth from ~\cite{barron2013intrinsic} which tends to model this effect well. We also observe that real world depth maps tend to have randomly placed irregular holes (pixels with depth 0). As a result, we compute random pixel-level masks and Gaussian blur them to obtain irregularly shaped masks that we then apply to the depth image.  We also use random camera cropping augmentations which has been shown to improve visuomotor learning performance~\cite{robomimic2021}. 

\subsection{Zero-shot Text Conditioned Manipulation}
Given our framework and trained local policies, how do we now deploy them in the real-world, to solve a wide array of manipulation tasks in a zero shot manner? 

To enable our system to solve long-horizon tasks, $p_{\theta}(g_k | g, O)$, decomposes the task into a skill chain to execute given task prompt $g$. We implement $p_{\theta}$ as GPT-4o, a SOTA VLM. Given the task prompt $g$, descriptions of the pre-trained local skills and how they operate, and images of the scene $O$, we prompt GPT-4o to give a plan for the task structured as a list of (object, skill) tuples. For example, for the task shown in Fig.~\ref{fig:main fig}, GPT outputs ((handle, open), (rice pick), (microwave, place), (handle, close)). We then need a language conditioned pose estimator (to compute $X_{targ,k}$) that generalizes broadly; we opt to use Grounded SAM~\cite{ren2024grounded} due to its strong open-set segmentation capabilities. To estimate $X_{targ,k}$, we can segment the object pointcloud, average it to get a position and use its surface normals to select a collision-free orientation. One issue is that Grounding Dino~\cite{liu2023grounding}, used in Grounded SAM, is very sensitive to the prompt. As a result, we pass its predictions back into GPT-4o to adjust the object prompts to capture the correct object. 

For predicting $\tau_{reach}$, while any motion planner can be used, we select Neural MP~\cite{dalal2024neuralmp} due to its fast planning time (2s) and strong real-world planning performance. Given $X_{targ,k}$, we compute target joint state $q_{k,f}$, plan with Neural MP open-loop and execute the predicted $\tau_{reach}$ on the robot using a PID joint controller. We then execute the appropriate local policy (as predicted by the VLM) on the robot to perform manipulation. We alternate between motion planning and deploying local policies until the task is complete. Finally, we note that the particular choice of models is orthogonal to our method. For additional details regarding all aspects of our method, please see the Appendix.

\section{Experimental Setup}
In our experiments, we evaluate ManipGen on simulated and real-world benchmarks as well as a set of challenging long-horizon manipulation tasks we created to stress-test ManipGen's generalization capabilities. We provide experiment details in this section and additional descriptions in the corresponding Appendix.

\noindent \textbf{Architecture and Training}
We train all RL policies at scale using PPO~\cite{schulman2017proximal} in GPU-parallelized simulation~\cite{makoviychuk2021isaac}. We train for 500 epochs, with an environment batch size of 8192 and max episode length of 120 steps per skill. To learn visuomotor policies to perform high-frequency (60 Hz) end-effector control, we pair Resnet-18~\cite{he2024learning} and Spatial Soft-max~\cite{finn2016deep} with a two layer MLP decoder (4096 hidden units). 
Finally, for training, minimizing Mean Squared Error loss is sufficient for learning multitask policies via DAgger. In early experiments, we found that our architecture performs comparably to using LSTMs~\cite{hochreiter1997long}, Transformers~\cite{vaswani2017attention}, and ACT~\cite{zhao2023learning} and is faster to train (5-10x) and deploy (2x). 

\noindent \textbf{Simulation Benchmarks} We evaluate against the long-horizon manipulation tasks used in PSL~\cite{dalal2024psl} from the Robosuite benchmark~\cite{zhu2020robosuite} in simulation which has a set of challenging long-horizon manipulation tasks (\textbf{PickPlace\{Bread, Milk, Cereal, CanBread, CerealMilk\}}). 
We compare to end-to-end RL methods~\cite{yarats2021mastering}, hierarchical RL~\cite{dalal2021accelerating,dalal2024psl}, task and motion planning~\cite{garrett2020pddlstream} and LLM planning~\cite{ahn2022say}. 

\noindent \textbf{Hardware Setup}
We use the Franka Panda robot arm with the UMI~\cite{chi2024universal} gripper fingertips and a wrist-mounted Intel Realsense d405 camera for obtaining local observations (84x84 resolution). 
Note for local observations, using depth sensing that is accurate at short range (such as the d405) is crucial to obtaining high quality local depth maps. 
We perform hole-filling and smoothing to clean the depth maps. 
Following Transic~\cite{jiang2024transic}, we do not model the compliance of the UMI gripper in simulation, but instead transfer policies trained with rigid fingertips to the real world, which performs well in practice. 
For real world control, we use a TSI end-effector controller at 60 Hz with (Leaky) Policy Level Action Integration (PLAI)~\cite{tang2023industreal}. We use Leaky PLAI with $.001$ position action scale, $.05$ rotation action scale for pick and $.005$ rotation action scale for all other skills. Finally, we use 4 calibrated Intel Realsense d455 cameras for global view observations (640x480). 

\noindent \textbf{Furniture Bench} To evaluate the sim2real capabilities of local policies (Tab.~\ref{table:transic results}), we deploy ManipGen on FurnitureBench~\cite{heo2023furniturebench}, comparing against a wide array of direct-transfer~\cite{peng2017simtoreal}, imitation learning~\cite{NIPS1988_812b4ba2, mandlekar2021matters}, offline RL~\cite{kostrikov2021offline} and human-in-the-loop methods~\cite{jiang2024transic, kelly2018hgdagger, mandlekar2020humanintheloop} from Transic~\cite{jiang2024transic}. These tasks are single stage; we train local policies to perform pushing (\textbf{Stabilize}), picking (\textbf{Reach and Grasp}) and insertion (\textbf{Insert}). We predict a start pose to initialize the local policy from and deploy the simulation-trained policies.

\noindent \textbf{Zero-shot Long-horizon Manipulation}
To test the generalization capabilities of our method, we propose 5 diverse long-horizon manipulation tasks (Fig.~\ref{fig:teaser}) which involve pick and place, obstacle avoidance and articulated object manipulation. \textbf{Cook}: put food into a pot on a stove (2 stages), \textbf{Replace}: take a pantry item out of the shelf, put it on a tray and take an object from the tray and put it in the shelf (4 stages), \textbf{CabinetStore}: open a drawer in the cabinet, put an object inside and close it (4 stages). \textbf{DrawerStore}: open a drawer, put two personal care items inside and close the drawer (6 stages) and \textbf{Tidy}: clean up the table by putting all the toys into a bin (8 stages). Each task has a unique object set (5 objects), receptacle (pot, shelf, etc.) and text description. We run 10 evaluations per task, randomizing which objects are present and their poses, receptacle poses, and target poses. All poses are randomized over the table and we select a diverse set of evaluation objects. 

\textbf{Evaluation Criteria} For each task we identify the stages required for completion. A trial is considered successful if the final state meets the task's goal as specified. Additionally, we track the number of stages completed in each trial. We conduct 10 trials per task, reporting the success rate and average number of stages completed.

\noindent \textbf{Comparisons}
We evaluate SOTA text-conditioned manipulation approaches: SayCan~\cite{ahn2022say}, LLMTrajGen~\cite{kwon2024language} and VoxPoser~\cite{huang2023voxposer}. For SayCan, we use our VLM and motion planning system with engineered primitives for interaction; testing the importance of training local policies. We additionally compare against a pre-trained model for manipulation, OpenVLA~\cite{kim2024openvla}. For each task, we collect 25 demonstrations on held out objects in held out poses and scene configurations and fine-tune OpenVLA per task. 
We pass in a text prompt specifying the task, recording the task success rate and number of stages completed.
\section{Results}
\label{sec:setup}

We pose the following experimental questions that guide our evaluation: 1) Can an autonomous agent control a robot to perform a wide array of \textit{long-horizon} manipulation tasks zero-shot?
2) How does our approach compare to methods that learn from online interaction? 
3) For direct sim2real transfer, how do Local Policies compare against end-to-end learning and other transfer techniques that leverage human correction data?
4) To what degree do the design decisions made in ManipGen affect the performance of the method?

\subsection{Simulation Comparisons and Analysis}
\noindent \textbf{Robosuite Benchmark Results}
In these experiments, we \textit{zero-shot} transfer our trained policies to Robosuite and evaluate their performance against methods that use task specific data (Tab.~\ref{table:psl results}). ManipGen outperforms or matches PSL, the SOTA method on these tasks, across the board, achieving an average success rate of $97.33\%$ compared to $95.83\%$. These results demonstrate that ManipGen can outperform methods that are trained on the task of interest~\cite{dalal2024psl,dalal2021accelerating,yarats2021mastering} as well as planning methods that have access to privileged state info~\cite{garrett2020pddlstream,ahn2022say}.
\begin{table}[t]
\vspace{5pt}
\resizebox{\linewidth}{!}{%
\begin{tabular}{cccccccc}
\toprule
\multicolumn{1}{l}{} & Bread      & Can        & Milk       & Cereal     & CanBread   & CerealMilk & Average \\
\midrule
\textit{Stages}      & \textit{2} & \textit{2} & \textit{2} & \textit{2} & \textit{4} & \textit{4} & \\
\midrule
\textit{Online Learning:} \\
DRQ-v2                  & $52\%        $ & $32  \%    $   & $2  \%    $    & $0   \%     $  &$ 0\%   $      & $0  \%   $ &  14\%\\
RAPS                 & $0 \%       $  & $0    \%  $    & $0   \%  $     & $0    \%   $   & $0  \%  $      &$ 0  \%  $     & 0\% \\
PSL                 & $100  \% $     & $100   \% $    & $100   \%  $   & $100   \%   $  & $90   \%  $    & $85   \%   $ &  96\% \\
\midrule
\textit{Zero-Shot:} \\
TAMP                 & $90 \%     $   & $100  \%   $   & $85   \%   $   & $100  \%    $  & $72  \%  $     & $71  \%   $   & 86\% \\
SayCan               & $93  \%   $    & $100   \%  $   & $90   \%   $   & $63   \%   $   & $63   \%  $    & $73   \%  $   & 80\% \\
\textbf{Ours}               & $100  \%$      & $100   \%  $   & $99    \%  $   & $97    \%  $   & $97   \%  $    & $91   \%   $  & \textbf{97\%} \\
\bottomrule
\end{tabular}
}
\vspace{-5pt}
\caption{
\small \textbf{Robosuite Benchmark Results.} ManipGen zero-shot transfers to Robosuite, outperforming end-to-end and hierarchical RL methods as well as traditional and LLM planning methods.}
\label{table:psl results}
\end{table}

\noindent \textbf{ManipGen Analysis and Ablations}.
We study design decisions proposed in our method by training single object pick policies on 5 objects (remote, can, bowl, bottle, camera) and testing on held out poses. We begin with our observation space design choices: ManipGen achieves $97.44\%$ success rate in comparison to (94.33\%, 96.64\%, 97.25\%) for removing key-point observations, success observation and reward observations respectively. Incorporating key-point observations is the most impactful change, enabling the agent to perceive the shape of the target object. Next, we evaluate how the level of locality (the size of the region around the target object that we initialize over) affects learning performance. At convergence, we find that ManipGen (8cm max distance from target) achieves $97.44\%$ success rate while performance diminishes with increasing distance ($95.65\%$, $89.55\%$, $72.52\%$) for $16$cm, $32$cm and $64$cm respectively. 

For DAgger, we analyze our observation design choices and find that including velocity information, the first observation, and changing proprioception to be relative to the first frame are crucial to the success of our method. While ManipGen gets $94.3\%$ success, removing velocity info and using absolute proprioception hurt significantly ($89.92\%$ and $90.94\%$) while removing the first observation drops performance to $93.13\%$. We also vary the DAgger buffer size, from $1$ (on-policy), $10$, $100$, and $1000$ (off-policy) for multitask training (with 3.5K objects, not 5). We find that 100 performs best, achieving 85\% in simulation averaged across 100 held out objects, out performing (78\%, 82\% and 75\%) for 1, 10 and 1000 respectively. 

\subsection{Real World Evaluation}
\noindent \textbf{FurnitureBench Results}
ManipGen matches or outperforms end-to-end direct transfer methods ($75\%$, $53.3\%$), imitation methods ($55\%$, $82.7\%$, $65\%$, $75\%$, $86.7\%$) and sim2real methods that leverage additional correction data~\cite{jiang2024transic} across the FurnitureBench suite. 
For Insert, local policies are able to outperform Transic without using any real world data, achieving 80\% while Transic achieves 45\%. These experiments demonstrate ManipGen improves over end-to-end learning and is capable of handling challenging initial states, contact-rich interaction and precise motions.

\begin{table}
\centering
\vspace{2pt}
\resizebox{\linewidth}{!}{%
\begin{tabular}{@{}cccccccc@{}}
\toprule
Tasks  & Ours  & Transic & \begin{tabular}[c]{@{}c@{}}Direct\\ Transfer\end{tabular} & \begin{tabular}[c]{@{}c@{}}DR. \& Data\\ Aug.~\cite{peng2017simtoreal}\end{tabular} & \begin{tabular}[c]{@{}c@{}}HG-Dagger~\cite{kelly2018hgdagger}\end{tabular} & \begin{tabular}[c]{@{}c@{}}IWR~\cite{mandlekar2020humanintheloop}\end{tabular} & BC~\cite{NIPS1988_812b4ba2} \\ \midrule
Stabilize       & $95\%$        & \textbf{100\%} & 10\%           & 35\%            & 65\%           & 65\%       & 40\% \\
Reach and Grasp & \textbf{95\%} & \textbf{95\%}  & 35\%           & 60\%            & 30\%           & 40\%       & 25\% \\
Insert          & \textbf{80\%} & 45\%           & 0\%            & 15\%            & 35\%           & 40\%       & 10\% \\
\midrule
Avg             & \textbf{90\%} & 80\%           & 15\%           & 36.7\%          & 43.3\%         & 48.3\%     & 25\% \\
\bottomrule
\end{tabular}
}
\vspace{-5pt}
\caption{
\small \textbf{Transic Benchmark Results} ManipGen achieves SOTA results on the Transic~\cite{jiang2024transic} benchmark in terms of task success rate without using any real world data, outperforming direct transfer, imitation learning and human-in-the-loop methods.
}
\label{table:transic results}
\end{table}

\noindent \textbf{Zero-shot Long-horizon Manipulation}
Across all 5 tasks (Tab.~\ref{table:main results}), we find that ManipGen outperforms all methods, achieving \textbf{$76\%$ zero-shot success rate} overall. Note that we have not trained our local policies on \textit{any of these specific objects} or in \textit{these specific configurations}; there is \textit{no adaptation} in the real world. ManipGen is able to avoid obstacles while performing manipulation of unseen objects in arbitrary poses and configurations. Failure cases for our method resulted from 1) vision failures as open-set detection models such as Grounding Dino~\cite{liu2023grounding} detected the wrong object, 2) imperfect motion planning, resulting in collisions with the environment during execution which dropped objects sometimes and 3) local policies failing to manipulate from sub-optimal initial poses. In general, DrawerStore and Tidy are the most challenging tasks due to their horizon, and consequently all methods, including our own perform worse (60\% for ours, 20\% for best baseline). 

SayCan is the strongest baseline (40\% success), achieving non-zero success on every task by leveraging the generalization capabilities of vision-language foundation models in a structured manner. However, when initial poses are not ideal or the task requires contact-rich control, pre-defined primitives fall apart (10-20\% success).
LLMTrajGen, while capable of performing top-down unconstrained pick and place (Cook: $70\%$), only makes partial progress on tasks requiring obstacle avoidance (Replace) or articulated object manipulation (Store) as its prompts struggle to cover those cases well. 
VoxPoser achieves similar performance on top-down unconstrained pick and place (Cook); however, it struggles to achieve desired rotation on articulated objects and in tight spaces (Replace and Store). Moreover, VoxPoser frequently generates incorrect plans in long-horizon tasks (Tidy).
Finally, OpenVLA failed to solve any task, failing to generalize to held out objects and poses even though it was the only method that was given few-shot data. We attempted to evaluate it on its training objects and it still performs poorly with strong pose randomization. 
For additional details regarding evaluation protocol, experiment conditions and deployment, please see the Appendix.

\begin{table}[t]
\vspace{2pt}
\centering
\resizebox{\linewidth}{!}{%
\begin{tabular}{lcccccc}
\toprule
 & Cook & Replace & CabinetStore & DrawerStore  & Tidy & Avg \\
\midrule
\textit{Stages}      & \textit{2} & \textit{4} & \textit{4} & \textit{6} & \textit{8} & \textit{4.8} \\
\midrule
OpenVLA              & 0\% (0.1)      & 0  (0.0)     & 0\% (0.0)      & 0 (0.0)     & 0  (0.0)    & 0\% (.02)   \\
SayCan               & 80\%  (1.7)    & 10\% (1.3)      & 70\% (3.5)      & 20\%  (3.6)    & 20\% (4.8)      & 40\% (3.0)   \\
LLMTrajGen        & 70\%   (1.5)   & 0\% (0.6)      & 0\% (0.6)      & 0\% (1.0)     & 0\% (2.6)      & 14\% (1.3)   \\
VoxPoser            & 70\% (1.4)      & 0\% (0.8)      & 0\% (0.8)      & 0\% (0.9)      & 10\% (4.4)      & 16\% (1.7)   \\
\textbf{Ours}                 & \textbf{90\% (1.9)}      & \textbf{80\% (3.7)}      & \textbf{90\% (3.9)}      & \textbf{60\% (4.7)}      & \textbf{60\% (7.2)}      & \textbf{76\% (4.3)}   \\
\bottomrule
\end{tabular}
}
\vspace{-5pt}
\caption{\small \textbf{Zero-shot Long Horizon Manipulation} We report task success rate and average number of stages completed per real world task. ManipGen outperforms all methods on each task, achieving 76\% with 4.28/4.8 stages completed on average.}
\label{table:main results}
\end{table}

\section{Discussion and Limitations}
\label{sec:conclusion}
We present ManipGen, a method for solving long-horizon manipulation tasks with unseen objects in unseen configurations by training generalist policies for sim2real transfer. 
We propose local policies, a novel policy class for sim2real transfer that is pose, skill order and scene configuration invariant, enabling it generalize broadly. For deployment, we take advantage of the broad generalization capabilities of foundation models for vision, language and motion planning to solve long-horizon manipulation tasks from text prompts. 
Across 50 real-world long-horizon manipulation tasks, our method achieves 76\% \textit{zero-shot} success, outperforming SOTA planning and imitation methods on every task.

ManipGen also leaves significant room for future work, for which we briefly describe limitations and outline potential solutions. First, as a modular framework, ManipGen is susceptible to cascading errors from any of the modules, particularly from the high-level planner providing incorrect plans and the pose-estimation latching onto the wrong object. Training value functions per module, incorporating closed-loop re-planning and double checking as well as online-adaptation of the modules (as they are all learnable) are all potential techniques that could be applied to diminish the impact of cascading errors. Next, due to the dependence on depth sensing, for pose-estimation, motion planning and local policy execution, ManipGen performs struggles with shiny, reflective and transparent objects. To that end, methods that learn depth from RGB~\cite{depthanything,Bochkovskii2024} could directly resolve this limitation. Finally, as interaction is learned in simulation, ManipGen is limited to learning behaviors that can be accurately simulated, limiting its direct applicability to tasks with state-changes and complex contacts. We note that the local policy framework described in this work, while uniquely suited for sim2real transfer, is amenable to training on real-world data as well.

\chapter{Conclusions}
\label{chap:conc}
\section{Summary}
In this thesis, we develop techniques for enabling generalization for robotics. Our key insight is that learning at scale, layered in modular, hierarchical systems can enable robots to generalize effectively to unseen tasks, situations and scenarios. We begin by first exploring how modularity can be integrated into learning algorithms for robotics and used to improve and accelerate learning performance. Key questions for doing so include how to construct the hierarchy and what to learn at each level. In Chapter~\ref{chap:raps}, we demonstrate that a modular structure in which an RL agent learns a high-level policy over pre-defined skills can dramatically improve learning performance over purely end-to-end approaches. Noting that in most robotics tasks, the most challenging behavior to acquire is interaction, in Chapter~\ref{chap:psl}, we reverse the prior decomposition, using planning for generalization in higher level task spaces while leveraging RL to learn reactive control for interaction. This approach produces demonstrable, state-of-the-art improvements over hierarchical, classical and learned approaches to long horizon robotic manipulation. Next, we turn to the question of scaling learning for generalization. We propose a simple recipe: procedural scene generalization at scale in simulation, using distilled planner experts for supervision. We first validate this approach purely in simulation in Chapter~\ref{chap:optimus}, training visuomotor policies to generalize manipulation skills across diverse object sets. We then extend the approach to the real world in Chapter~\ref{chap:nmp}, distilling motion planning at scale into a single, end-to-end neural network for motion planning. Finally, putting everything together, in Chapter~\ref{chap:manipgen}, we develop a generalist agent for long-horizon robotic manipulation, a modular system that leverages large scale learning in simulation to be able to solve long-horizon manipulation tasks purely from text prompt inputs.

\clearpage

\section{Takeaways}
This thesis covers a large body of work concerning accelerating, improving and generalizing robot learning. We now distill this thesis into three core lessons that can be applied to future work.

\noindent \textbf{Plan at the high level, learn at the low level}. For practitioners seeking to incorporate modularity into learning algorithms, consider using planning for higher level functions such as semantic reasoning and task planning while using learning for low-level control. In this thesis, we observe this phenomena in action, where learning low-level control and planning higher level information~\cite{dalal2024psl} significantly outperforms learning RL policies over parameterized skills~\cite{dalal2021accelerating}. Intuitively, this makes sense. Planning, particularly with ever-improving LLMs~\cite{openai2023gpt4}, has been shown to be capable of generalizing broadly over high-level abstract concepts, leveraging common sense and thinking ahead. Learning, on the other hand, is capable of acquiring complex low-level control behaviors that are difficult to manually engineer. By planning high-level and learning low-level control, we can leverage the strengths of each approach effectively.

\noindent \textbf{Incorporate learning at every step of a modular pipeline}. This point embodies the core insight of this thesis: Integrating learning with modularity can be done by first scaling models that can generalize within some specialty (\textit{e.g.} task planning, motion planning, control) and amplifying their generalization by combining their outputs. ManipGen~\cite{dalal2024manipgen} demonstrates this principle in action, acting as a fully learned TAMP system in which large scale models are used for task planning (VLMs), motion planning (Neural MP) and low-level control. These models talk to each other via limited interfaces, yet when combined appropriately can produce powerful generalization results. A critical advantage of this approach is that we do not require a vast corpus of training data for the entire robotics problem. Instead, dedicated datasets and models enable the per-module networks to be significantly more capable, simplifying the data gathering and learning burden.

\noindent \textbf{Separate behavior acquisition from visuomotor learning via distillation} Behavior acquisition is often extremely challenging when combined with perception, as the agent must learn both observation processing and low-level control. Separating the two enables control to be learned efficiently and then the second stage can simply learn perception while matching control. In this thesis, we demonstrate two different techniques for doing so, either using planning~\citep{dalal2023optimus,dalal2024neuralmp} and RL~\cite{dalal2024manipgen}. Planning generalizes automatically to wide distributions of tasks, simplifying the behavior acquisition problem significantly, but limiting the types of tasks that can be solved. RL on the other hand is capable of acquiring a larger, more complex set of behaviors with less engineering effort, but often requires significantly more resources to acquire behavior at scale. 

\section{Looking Forward}
This thesis proposes a variety of algorithms and techniques to enable robots to act over long horizons while generalizing to a wide variety of tasks and environments. That being said, many challenging open problems remain; we discuss these and propose potential solutions for future work to explore and tackle. 

\noindent \textbf{Closed-loop, reactive control}
Acquiring closed-loop, reactive behavior for manipulation remains an open and consequential challenge. Even simple skills such as picking and placing, become significantly more difficult to perform robustly when moving to the dexterous manipulation regime due to the high-dimensionality of the control problem. In such settings, it will be critical to learn policies that are capable of reactivity. Relatively simple strategies that solely predict grasp points often struggle to succeed. Behavior cloning, and other offline learning methods, while effective for learning quasi-static, open-loop behaviors, also struggle when applied to the reactive control regime due to compounding error. Online learning is critical to acquire these skills as it enables the agent to learn from its mistakes, correct errors and continually improve. Yet online learning faces a number of practical challenges, namely in the real world it is difficult to acquire the number of samples necessary for online learning to perform well, there is often little to no closed loop supervision (needed for algorithms such as DAgger~\cite{ross2011reduction}) and objectives are difficult to specify due to lack of state knowledge. 

Instead, in this thesis, we explored simulation as a means of acquiring closed loop reactive behavior as it is a paradigm that scales with compute, can produce as many samples as desired (30 billion trajectories in ManipGen~\cite{dalal2024manipgen} - an amount that would require years of data collection in the real world) and provides full state information to produce rewards. Yet with simulation, significant challenges remain: reward specification, procedural scene generation and the sim-to-real gap. That being said, sim-to-real transfer is one of the few methods in which the robotics community has seen significant progress in recent years: locomotion~\cite{kumar2021rma,agarwal2022legged}, making it an exciting prospect for similar results in manipulation. Future research should aim to push the limits of online learning for closed loop, reactive control, particularly on challenging tasks such as assembly~\cite{narang2022factory,tang2023industreal}. 

However, simulation is currently limited in the types of interactions it can express. Currently, simulation is best for rigid body interactions with limited contacts and no state changes. A promising paradigm that could eliminate this limitation entirely is world modeling. One reason why world modeling, in the long-term, may end up being the dominant paradigm is that it is in the end, a supervised learning problem. This lends it significantly stronger guarantees on performance than behavior cloning and other offline learning methods and would enable leveraging online learning to effectively learn closed loop control. Unfortunately, world models have traditionally been limited to solving simulated/simple control tasks~\cite{hafner2019dream} as there has not been significant effort in scaling these models with the right architecture. However, in recent years, video prediction/generation has improved significantly at scale~\cite{villegas2022phenaki} and with products such as MovieGen and Sora. These advances have begun to carry over to world modeling as well~\cite{bruce2024genie}. This lends credence to the hypothesis that in the long run, with sufficient compute and scale, accurately simulating trajectories over long horizons may become feasible, enabling online learning at scale for complex, diverse tasks beyond current capabilities.

\noindent \textbf{Planning over longer and more complex horizons}
Planning and long-horizon manipulation remain particularly challenging, requiring simultaneous reasoning across multiple future steps while executing dexterous physical interactions. As the horizon grows, the probability of success falls rapidly: for example with even $99\%$ success probability, after executing 100 skills the overall task success rate is approximately $36\%$. This problem is critical to solve for deploying robots: they must be able to operate continuously for long periods of time on temporally extended tasks without requiring human intervention. While this analysis suggests we require even more capable and robust skills, in practice for each additional degree of precision, we need all the more data and engineering effort. 

In this thesis, we explore the long-horizon problem from the perspective of simplifying the action abstraction~\cite{dalal2021accelerating} and modularizing the problem~\cite{dalal2024psl,dalal2024manipgen} to ease the learning burden and responsibility. By decomposing manipulation tasks into abstract reasoning (language), motion planning and control, we demonstrate generalization to a broad set of manipulation tasks. Nonetheless, this core problem of cascading error remains unsolved. Even in ManipGen, as the horizon grows, the success rate drops precipitously and this problem only becomes worse the more skills we try to chain. Reasons for why this is the case include open-loop deployment of plans, limited physical reasoning capabilities of current VLMs and motion planning failures.

Looking forward, there are several techniques of interest that future work should explore to maximize long horizon performance. The first step is simply improving the planning model itself, where models that reason at test time—such as OpenAI O1 and ~\cite{guo2025deepseek}—will be uniquely useful. This will enable longer horizon, coherent plans that take into account visual context, physical plausibility and the task objective. Next, for long-horizon tasks, it will likely be crucial to employ re-trying and re-planning behavior in the loop, as the first attempt may not always succeed. For this capability, we will require effective, robust closed loop controllers than can be re-deployed easily upon skill failure. This will also require much more capable VLM planners that can take in visual input, recognize that skills have failed to execute and retry in new ways to adapt.

\noindent \textbf{Continual learning and improvement}
In this thesis, we have strictly considered a fixed train-test split, where the agent learns during training and is deployed at test time without additional information. In reality, robots will be continuously dealing with new scenarios and information and must have the capability of adapting either on the fly, or over time as new data comes in. For example, consider a warehouse robot doing pick and place: over time, the item distributions will be constantly changing and it cannot expect to generalize to all future items after an initial training run. In such settings, the current paradigms of entirely offline and simulation-based online learning will not be sufficient. 

Continuously ingesting data, improving performance, and maintaining existing capabilities present a challenging problem requiring further investigation. Techniques such as offline RL will likely be relevant, as well as methods that do batched online RL. One potential paradigm may involve large behavior models pre-trained on vast corpi of data drawn from simulation, human-video and real-world data being deployed at scale. Such models will then need to be fine-tuned online based on their performance, using rewards estimated via learned reward functions and then improved offline in parallel before being deployed again.

\clearpage

\part{Appendix}
\appendix
\chapter{Chapter 2 Appendix}

\section{Table of Contents}
\begin{itemize}
    \item \textbf{Additional Experimental Results and Analysis} (Appendix~\ref{sec:additional-results}): Additional experimental results demonstrating RAPS\xspace's effectiveness on more tasks and additional baselines.
    \item \textbf{Ablations} (Appendix~\ref{sec:appendix-ablations}): Ablations and analyses of RAPS\xspace, demonstrating the effectiveness of our design decisions.
    \item \textbf{Environments} (Appendix~\ref{sec:appendix-envs}): Description of all the environments we use in this work.
    \item \textbf{Primitive Implementation Details} (Appendix~\ref{sec:appendix-prim-impl}): Details regarding how the primitives in RAPS\xspace are implemented.
    \item \textbf{RL Implementation Details} (Appendix~\ref{sec:appendix-rl-impl}): Full details on how RAPS\xspace is implemented, specifically the hyper-parameters used for training the RL agents and network architectures.
\end{itemize}

\newpage
\section{Additional Experimental Results}
\label{sec:additional-results}
\textbf{Cross Robot Transfer} $\quad$ Robot action primitives are agnostic to the exact geometry of the underlying robot, provided the robot is a manipulator arm. As a result, one could plausibly ask the question: is it possible to train RAPS\xspace on one robot and evaluate on a morphologically different robot for the same task? In order to answer this question, we train a higher level policy over RAPS from visual input to solve the door opening task in Robosuite using the xARM 7. We then directly transfer this policy (zero-shot) to an xARM 6 robot. The transferred policy is able to achieve \textbf{100\%} success rate on the door opening task with the 6DOF robot while trained on a 7DOF robot. To our knowledge such as result has not been shown before.

\begin{figure}[ht]
    \centering
    \begin{tabular}{cc}
    \includegraphics[width=.24\textwidth]{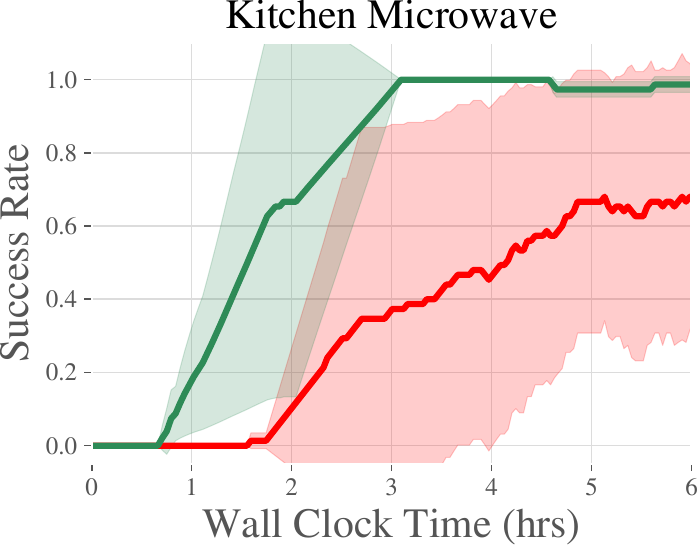}
    \includegraphics[width=.24\textwidth]{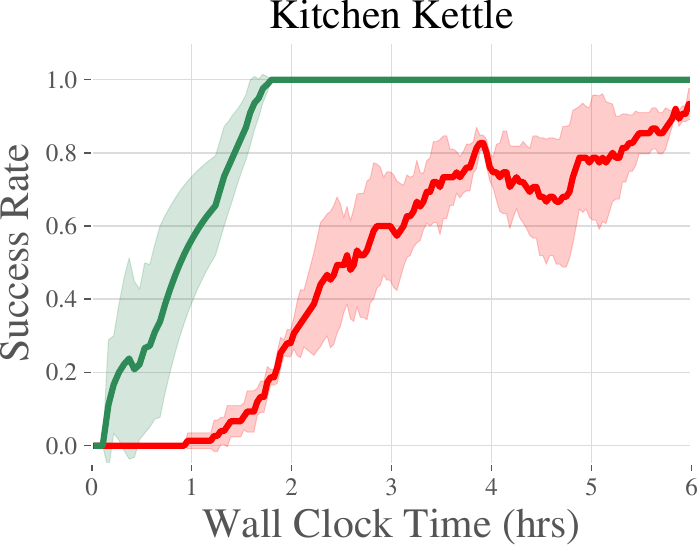} &
    \includegraphics[width=.24\textwidth]{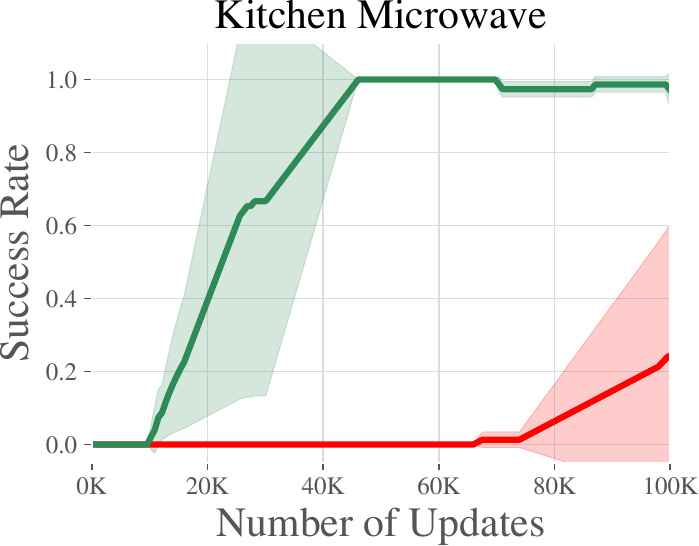}
    \includegraphics[width=.24\textwidth]{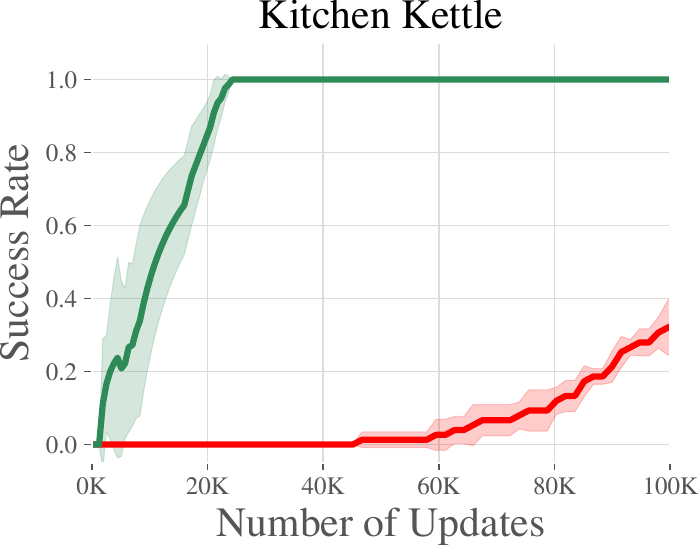}
    \vspace{.2cm}
    \\
    \includegraphics[width=.24\textwidth]{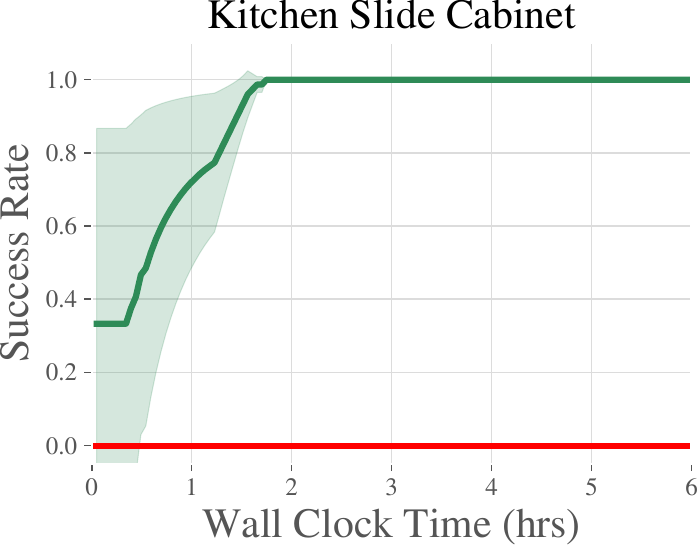}
    \includegraphics[width=.24\textwidth]{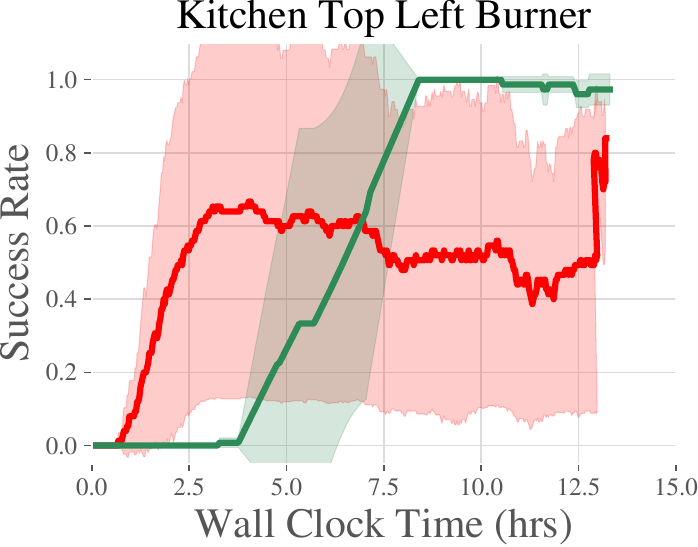} &
    \includegraphics[width=.24\textwidth]{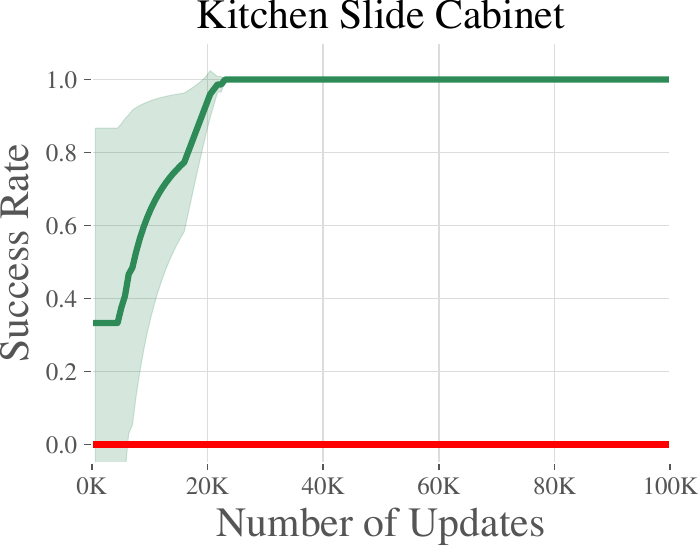}
    \includegraphics[width=.24\textwidth]{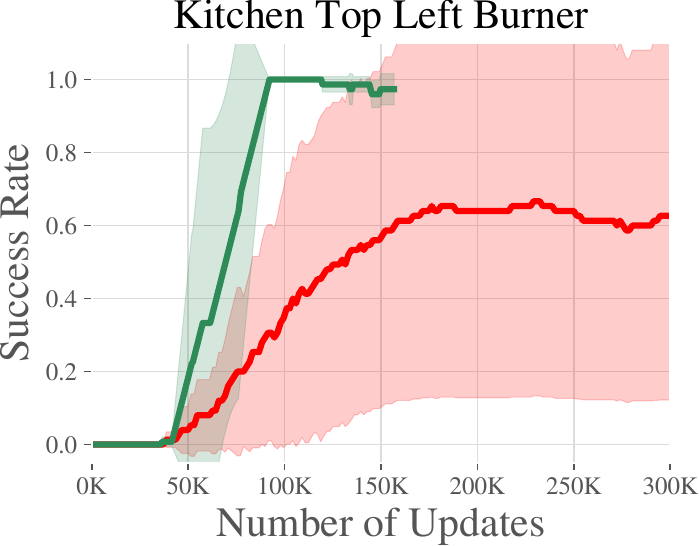}
    \vspace{.2cm}
    \\
    \includegraphics[width=.24\textwidth]{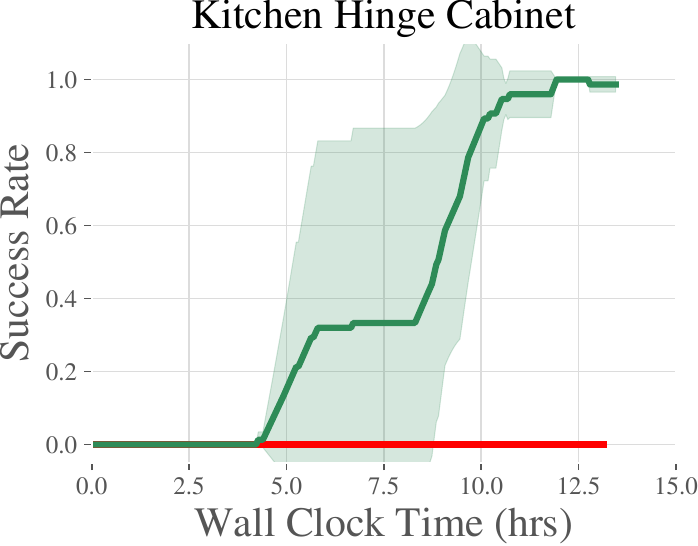}
    \includegraphics[width=.24\textwidth]{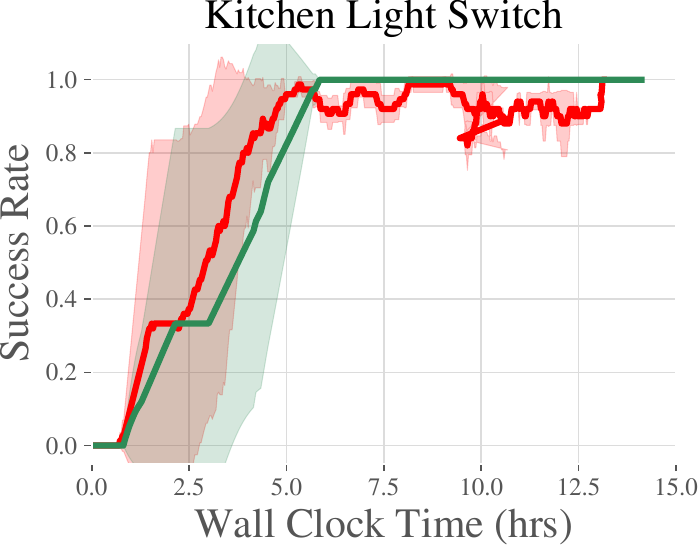} &
    \includegraphics[width=.24\textwidth]{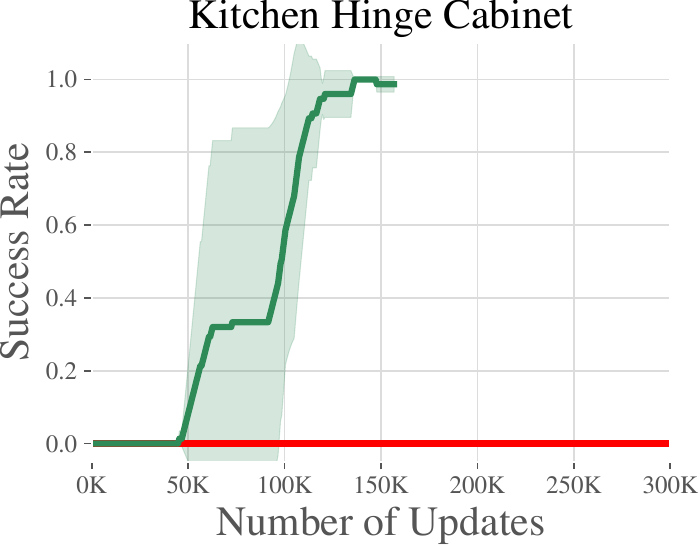}
    \includegraphics[width=.24\textwidth]{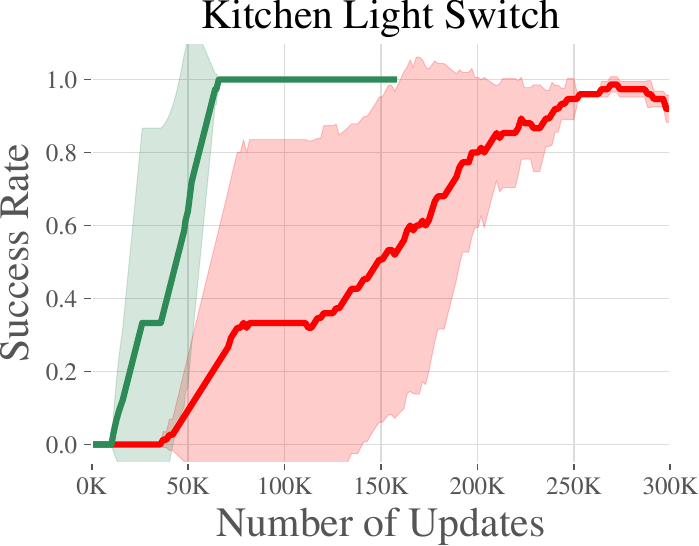}
    \end{tabular}
    \includegraphics[width=\textwidth]{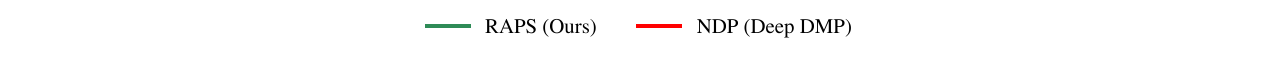}
    \vspace{-0.25in}
    \caption{Comparison of RAPS against NDP, a deep DMP method for RL. RAPS dramatically outperforms NDP on nearly every task from visual input, both in terms of wall-clock time and number of training steps. This result demonstrates the increased capability of RAPS over DMP-based methods. }
    \label{fig:ndp-comparison}
    \vspace{-0.05in}
\end{figure}

\textbf{Comparison against Dynamic Motion Primitives} $\quad$ 
As noted in the related works section, Dynamic Motion Primitives (DMP) are an alternative skill formulation that is common robotics literature. We compared RAPS\xspace with the latest state-of-the-art work that incorporates DMPs with Deep RL: Neural Dynamic Policies~\citep{bahl2020neural}. As seen in Figure \ref{fig:ndp-comparison}, across nearly every task in the Kitchen suite, RAPS outperforms NDP from visual input just as it outperforms all prior skill learning methods as well. 

\textbf{Real Robot Timing Results} $\quad$ To experimentally verify that RAPS\xspace runs faster than raw actions in the real wrold, we ran a randomly initialized deep RL agent with end-effector control and RAPS\xspace on a real xArm 6 robot and averaged the times of running ten trajectories. Each primitive ran 200 low-level actions with a path length of five high level actions, while the low-level path length was 500. Note that RAPS\xspace has double the number of low-level actions of the raw action space within a single trajectory. With raw actions, each episode took 16.49 seconds while with RAPS, each episode lasted an average of 0.51 seconds, a \textbf{32x} speed up.
\section{Ablations}
\label{sec:appendix-ablations}
\begin{figure}
    \centering
    \includegraphics[width=.24\textwidth]{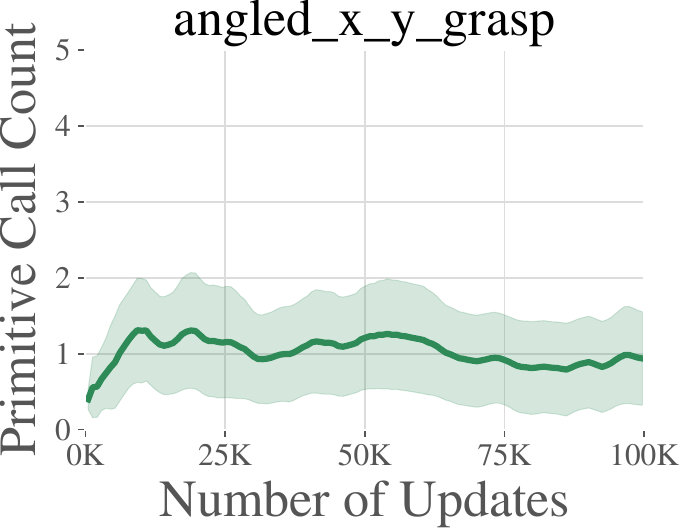}
    \includegraphics[width=.24\textwidth]{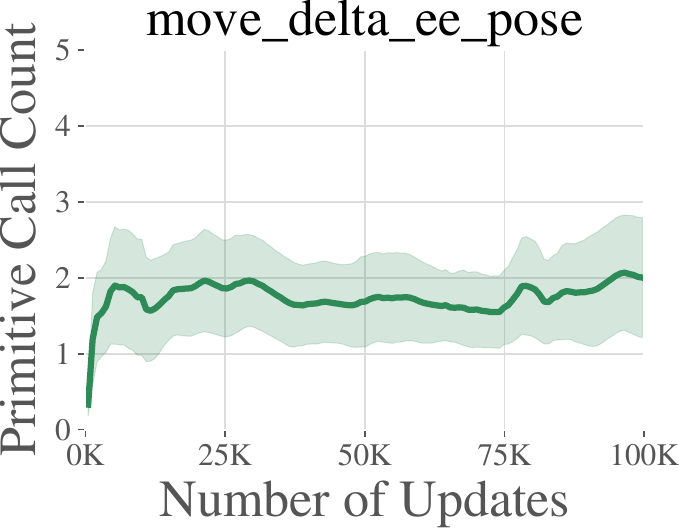}
    \includegraphics[width=.24\textwidth]{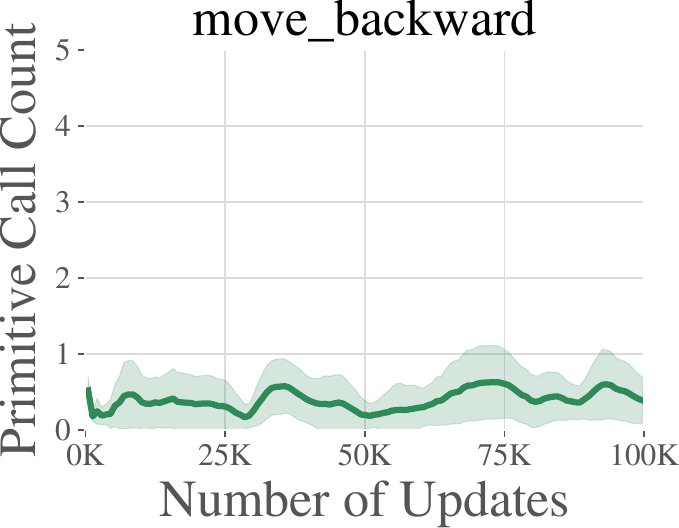}
    \includegraphics[width=.24\textwidth]{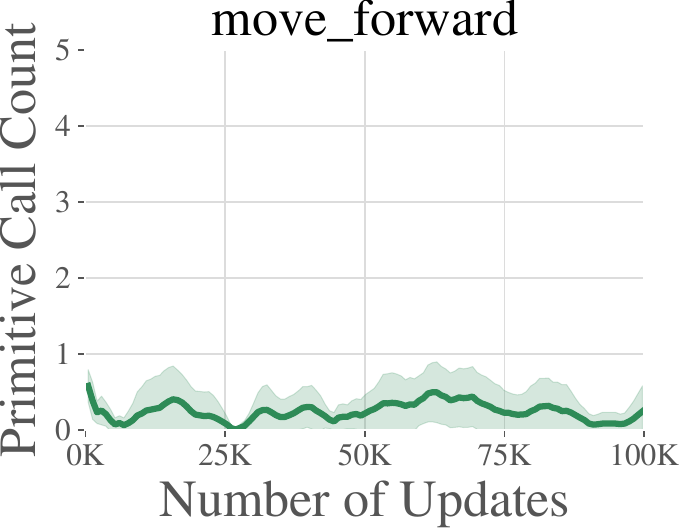}
    \vspace{.2cm}
    \\
    \includegraphics[width=.24\textwidth]{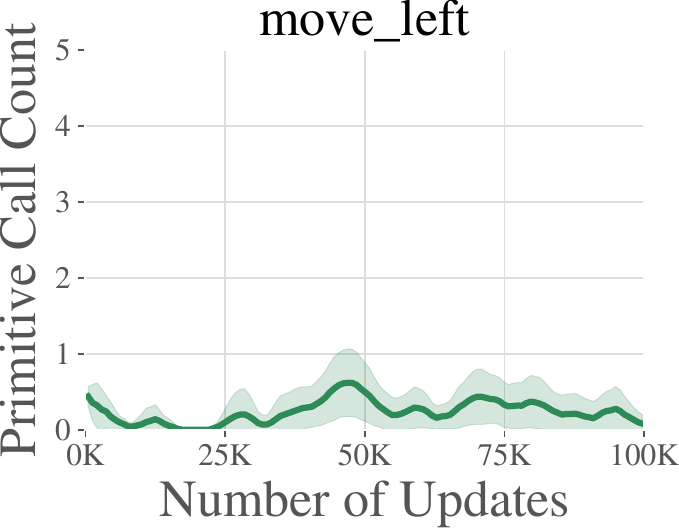}
    \includegraphics[width=.24\textwidth]{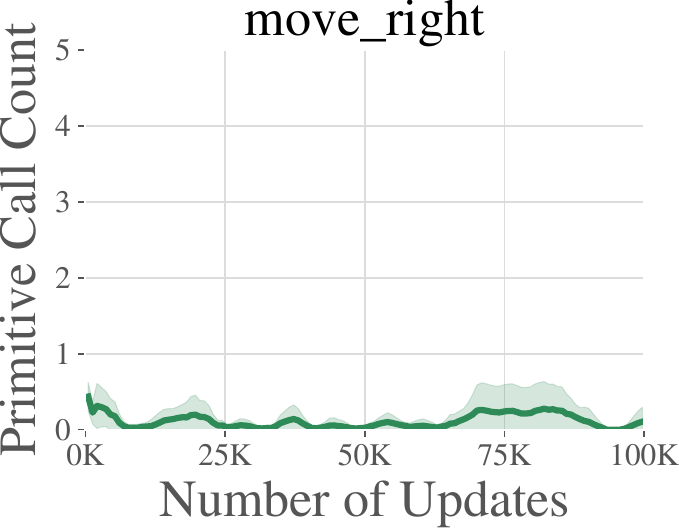}
    \includegraphics[width=.24\textwidth]{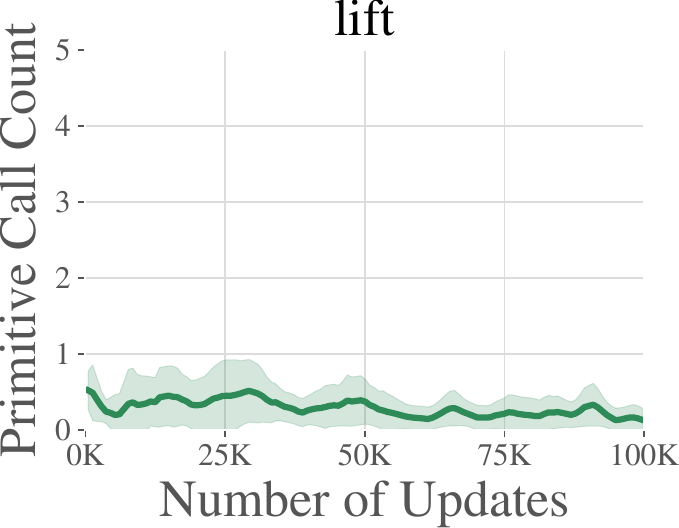}
    \includegraphics[width=.24\textwidth]{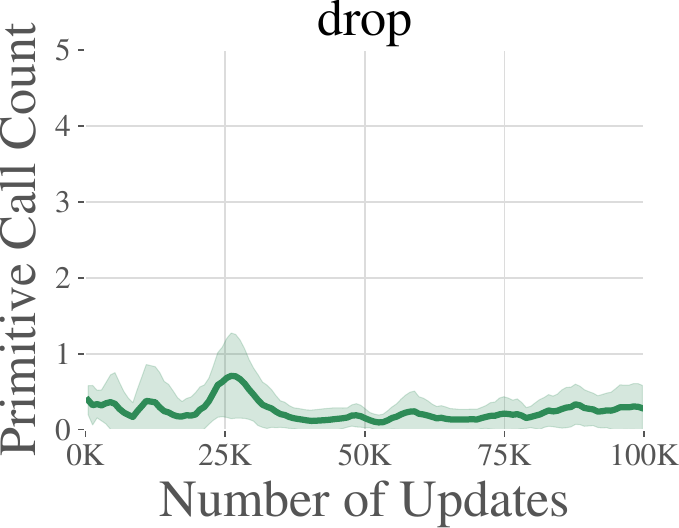}
    \vspace{.2cm}
    \\
    \includegraphics[width=.24\textwidth]{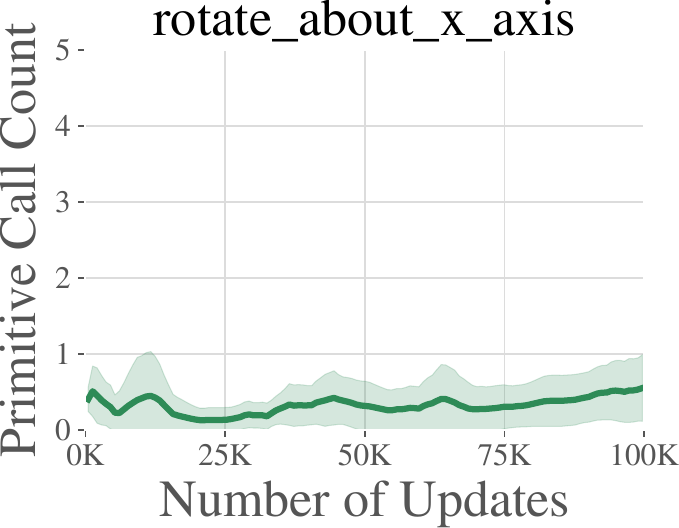}
    \includegraphics[width=.24\textwidth]{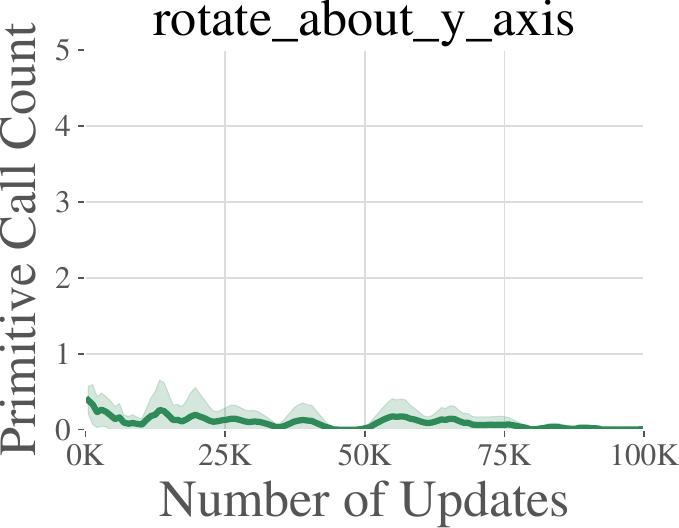}
    \includegraphics[width=.24\textwidth]{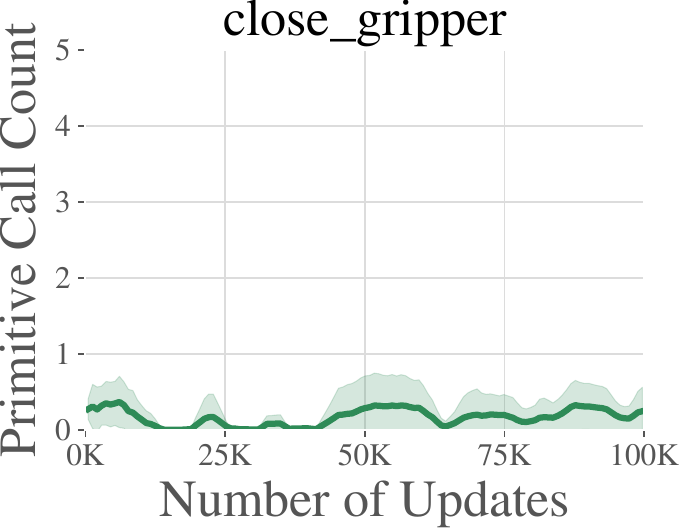}
    \includegraphics[width=.24\textwidth]{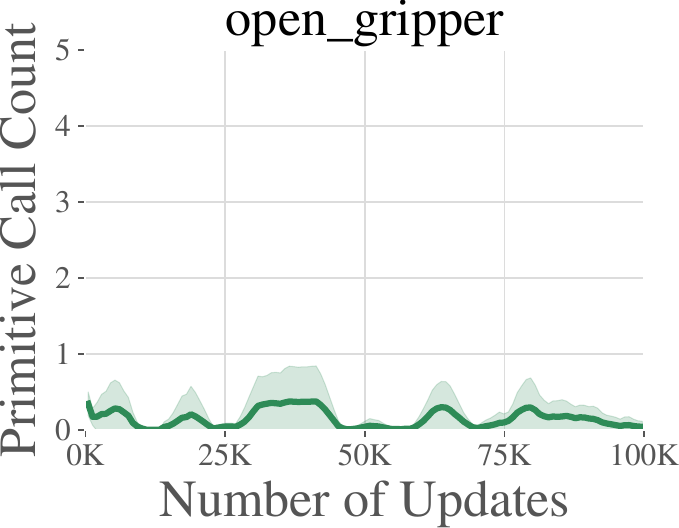}
    \caption{Primitive call counts for the evaluation policy averaged across all six kitchen tasks, plotted against number of training calls. In the beginning, each primitive is called at roughly the same frequency (uniformly at random), but over time the learned policies develop a preference for the dummy primitive and the angled xy grasp primitive, while still occasionally using the other primitives as necessary.}
    \label{fig:primitive-log-1}
    \vspace{-0.05in}
\end{figure}

\begin{figure}
    \centering
    \includegraphics[width=.3\textwidth]{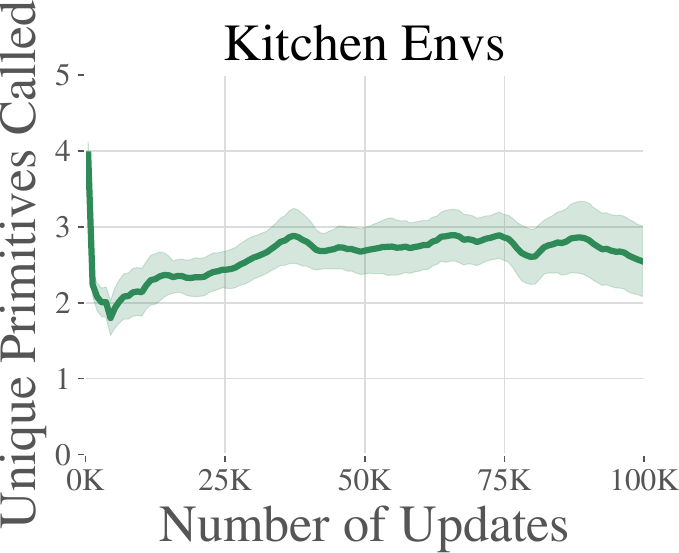}
    \caption{Number of unique primitives called by the evaluation policy averaged across all six Kitchen tasks, plotted against the number of training calls. Early on in training, the number of unique primitives called is four. With a path length of five this makes sense, on average it is calling unique primitives almost every time. At convergence, the number of unique primitives called is around 2.69. This suggests later on the policy learns to select certain primitives more often to optimally solve the task.}
    \label{fig:primitive-log-2}
\end{figure}

\textbf{Primitive Usage Experiments} $\quad$ We run an ablation to measure how often RAPS uses each primitive. In Figure \ref{fig:primitive-log-1}, we log the number of times each primitive is called at test time, averaged across all of the kitchen environments. It is clear from the figure that even at convergence, each primitive is called a non-zero amount of times, so each primitive is useful for some task. However, there are two primitives that are favored across all the tasks, \texttt{move-delta-ee-pose} and \texttt{angled-xy-grasp}. This is not surprising as these two primitives are easily applicable to many tasks. We evaluate the number of unique primitives selected by the test time policy over time (within a single episode) in Figure \ref{fig:primitive-log-2} and note that it converges to about $2.69$. To ground this number, the path length for these tasks is $5$. This means that on most tasks, the higher level policy ends up repeatedly applying certain primitives in order to achieve the task.

\begin{figure}
    \centering
    \includegraphics[width=.32\textwidth]{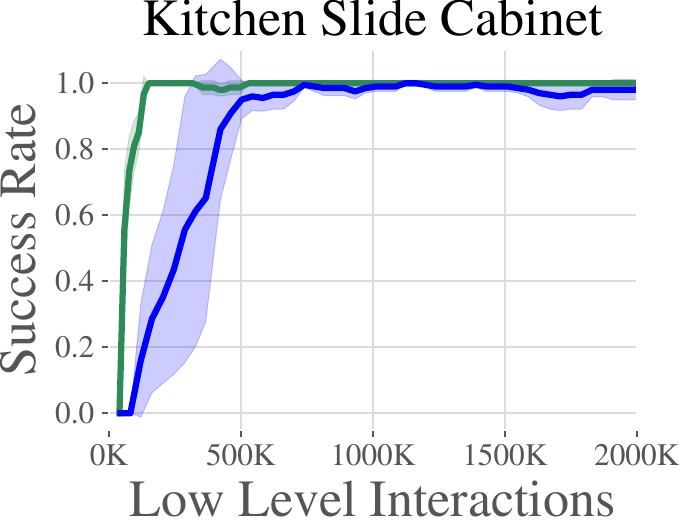}
    \includegraphics[width=.32\textwidth]{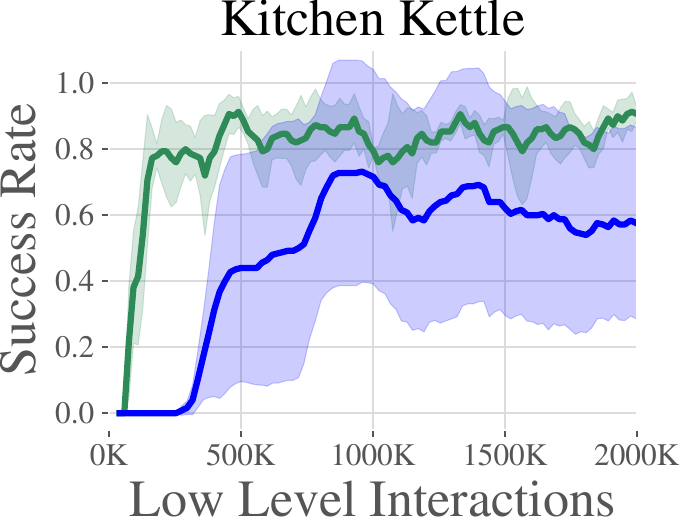}
    \includegraphics[width=.32\textwidth]{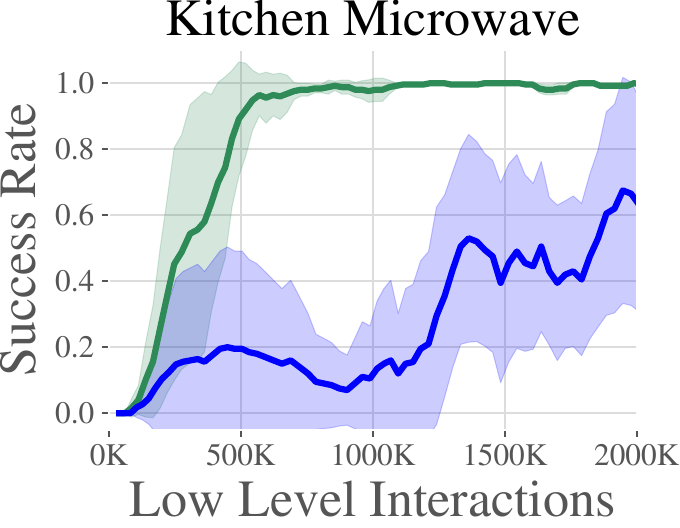}
    \vspace{.2cm}
    \\
    \includegraphics[width=.32\textwidth]{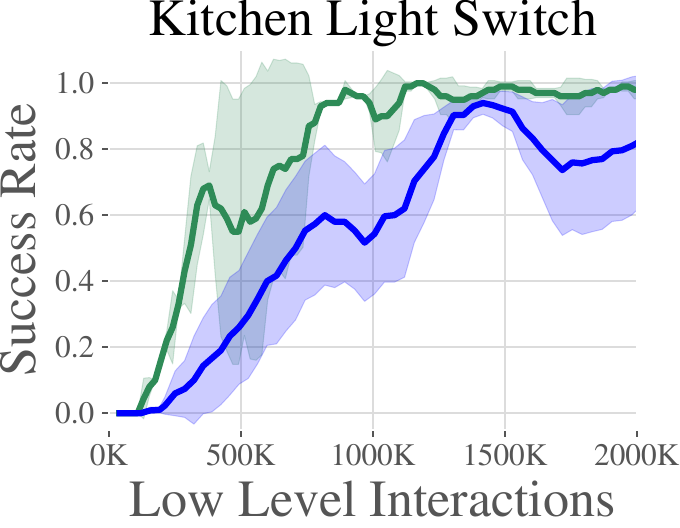}
    \includegraphics[width=.32\textwidth]{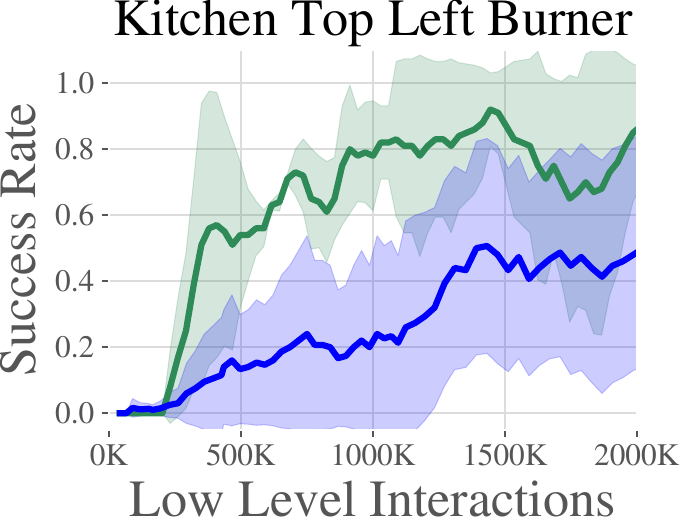}
    \includegraphics[width=.32\textwidth]{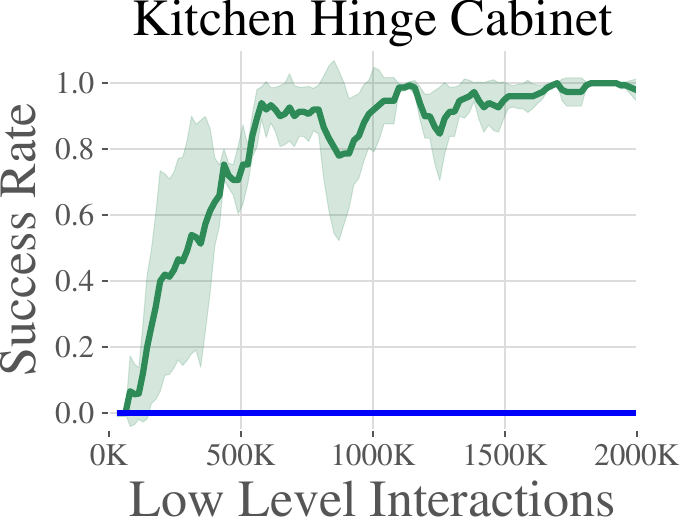}
    \includegraphics[width=\textwidth]{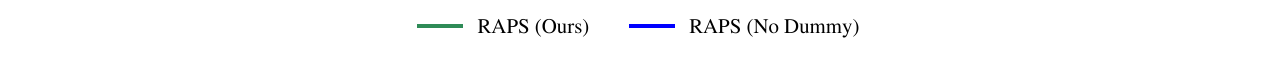}
    \caption{We run an ablation of RAPS in which we remove the dummy primitive, and we find that in general, this negatively impacts performance. Without the dummy primitive, RAPS is less stable and unable to solve the \texttt{hinge-cabinet} task.}
    \label{fig:no-dummy}
\end{figure}

\textbf{Evaluating the Dummy Primitive} $\quad$
The dummy primitive is one of the two most used primitives (also known as move delta ee pose), the other being angled xy grasp (also known as angled forward grasp in the appendix). One question that may arise is: How useful is the dummy primitive? We run an experiment with and without the dummy primitive in order to evaluate its impact, and find that the dummy primitive improves performance significantly. Based on the results in Figure \ref{fig:no-dummy}, hand-designed primitives are not always sufficient to solve the task.

\begin{figure}[ht]
    \centering
    \includegraphics[width=.32\textwidth]{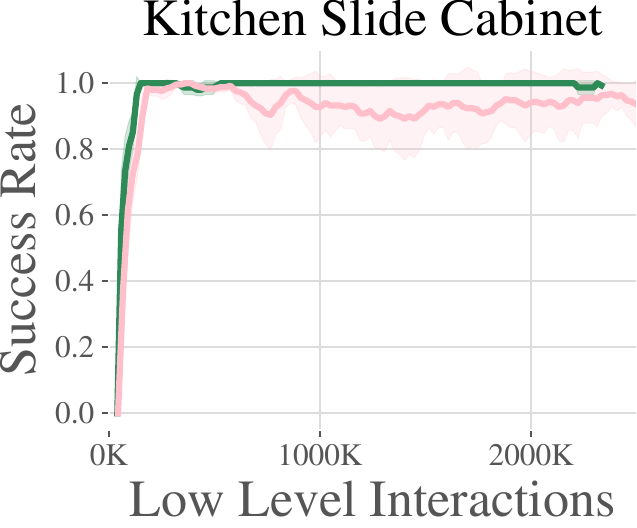}
    \includegraphics[width=.32\textwidth]{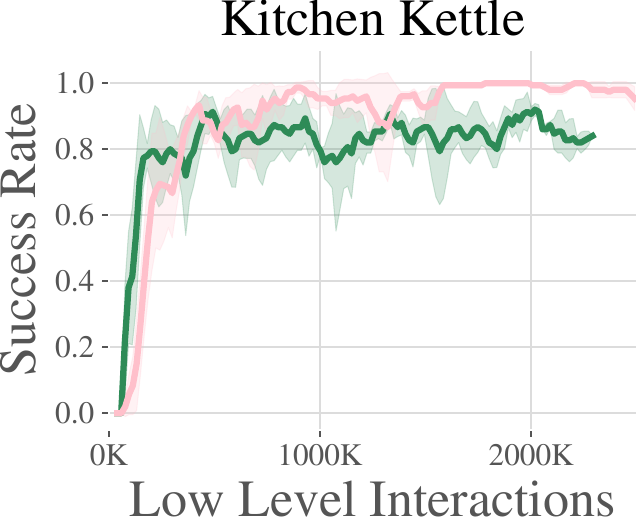}
    \includegraphics[width=.32\textwidth]{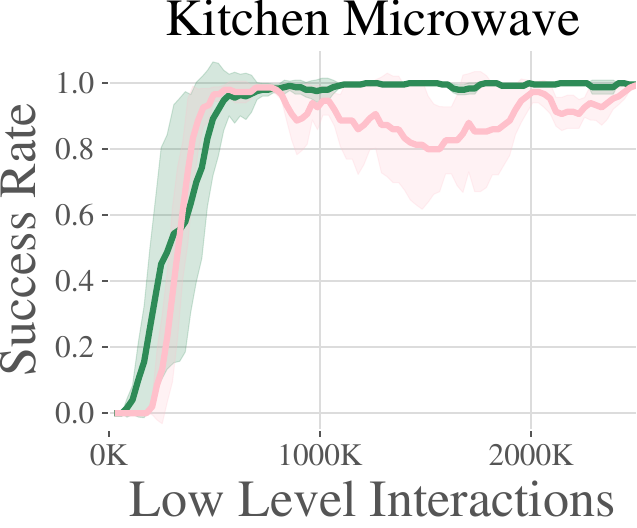}
    \vspace{.2cm}
    \\
    \includegraphics[width=.32\textwidth]{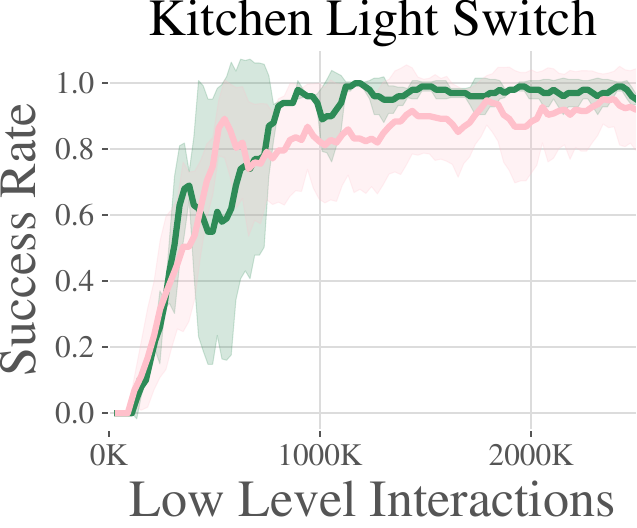}
    \includegraphics[width=.32\textwidth]{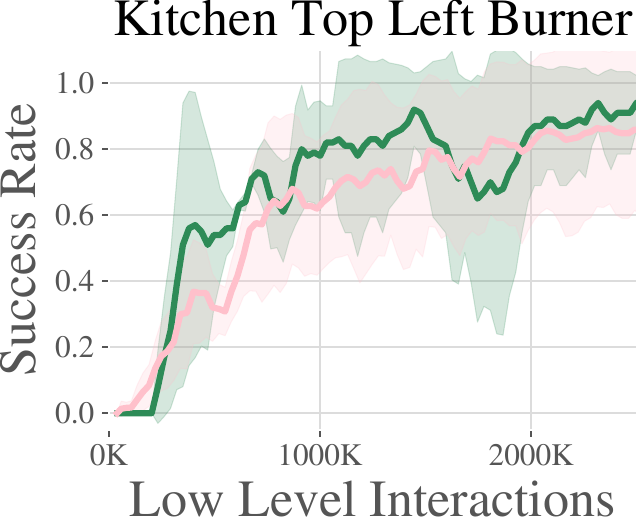}
    \includegraphics[width=.32\textwidth]{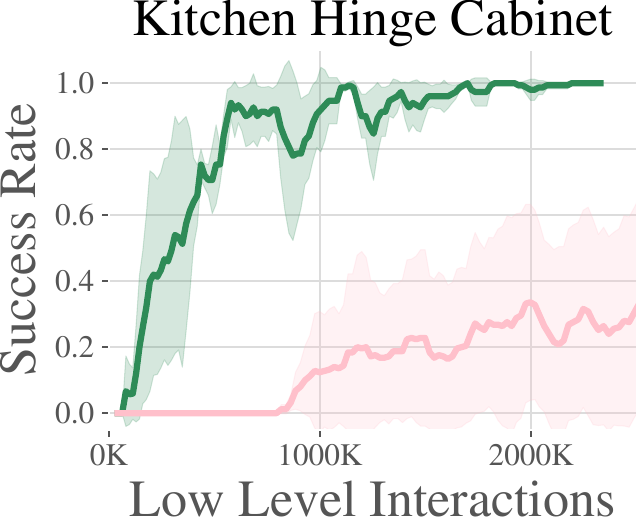}
    \includegraphics[width=\textwidth]{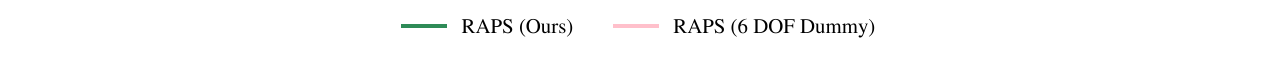}
    \caption{We plot the results of running RAPS with a 6DOF control dummy primitive, and find that in general, the performance is largely the same.}
    \label{fig:6-dof}
\end{figure}

\textbf{Using a 6DOF Control Dummy Primitive} $\quad$
The dummy primitive uses 3DOF (end-effector position) control in the experiments in the main paper, but we could just as easily do 6DOF control if desired. In fact, we ablate this exact design choice. If we change the dummy primitive to achieve any full 6-DOF pose (end-effector position as well as orientation expressed in roll-pitch-yaw), the overall performance of RAPS\xspace does not change. We plot the results of running RAPS\xspace on the Kitchen tasks against RAPS\xspace with a 6DOF dummy primitive in Figure \ref{fig:6-dof} and find that the performance is largely the same.

\section{Environments}
\label{sec:appendix-envs}
We provide detailed descriptions of each environment suite and the specific tasks each suite contains. All environments use the MuJoCo simulator~\citep{todorov12mujoco}.
\subsection{Kitchen}
\begin{figure}
    \centering
    \includegraphics[width=\textwidth]{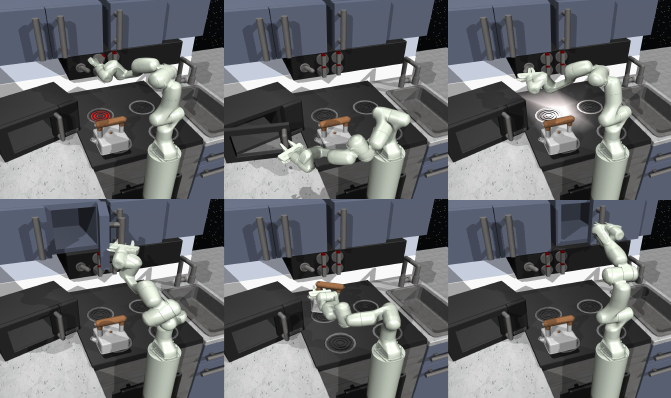}
    \caption{Visual depiction of the Kitchen environment; all tasks are contained within the same setup. Each image depicts the solution of one of the tasks, we omit the bottom burner task as it is the goal is the same as the top burner task, just with a different dial to turn. For the top row from the left: \texttt{top-left-burner}, \texttt{microwave}, \texttt{light-switch}. For the bottom row from the left: \texttt{hinge-cabinet}, \texttt{kettle}, \texttt{slide cabinet}.}
    \label{fig:kitchen}
\end{figure}

The Kitchen suite, introduced in \citep{gupta2019relay}, involves a set of different tasks in a kitchen setup with a single Franka Panda arm as visualized in Figure \ref{fig:kitchen}.
This domain contains 7 subtasks: \texttt{slide-cabinet} (shift right-side cabinet to the right), \texttt{microwave} (open the microwave door), \texttt{kettle} (place the kettle on the back burner), \texttt{hinge-cabinet} (open the hinge cabinet), \texttt{top-left-burner} (rotate the top stove dial), \texttt{bottom-left-burner} (rotate the bottom stove dial), and \texttt{light-switch} (flick the light switch to the left). 
The tasks are all defined in terms of a sparse reward, in which $+1$ reward is received when the norm of the joint position (qpos in MuJoCo) of the object is within $.3$ of the desired goal location and $0$ otherwise. See the appendix of the RPL~\citep{gupta2019relay} paper for the exact definition of the sparse and the dense reward functions in the kitchen environment. Since the rewards are defined simply in terms of distance of object to goal, the agent does not have to execute interpretable behavior in order to solve the task. For example, to solve the burner task, it is possible to push it to the right setting without grasping and turning it. The low level action space for this suite uses 6DOF end-effector control along with grasp control; we implement the primitives using this action space. 

For the sequential multi-task version of the environment, in a single episode, the goal is to complete four different subtasks. The agent receives reward once per sub-task completed with a maximum episode return (sum of rewards) of 4. In our case, we split the 7 tasks in the environment into two multi-task environments which are roughly split on difficulty.
We define the two multi-task environments in the kitchen setup: \texttt{Kitchen Multitask 1} which contains \texttt{microwave}, \texttt{kettle}, \texttt{light-switch} and \texttt{top-left-burner} while \texttt{Kitchen Multitask 2} contains the \texttt{hinge-cabinet}, \texttt{slide-cabinet}, \texttt{bottom-left-burner} and \texttt{light-switch}.
As mentioned in the experiments section, RL trained on joint velocity control is unable to solve almost any of the single task environments using image input from sparse rewards. Instead, we modify the environment to use 6DOF delta position control by adding a mocap constraint as implemented in Metaworld~\citep{yu2020meta}.

\subsection{Metaworld}
Metaworld~\citep{yu2020meta} consists of 50 different manipulation environments in which a simulated Sawyer Rethink robot is charged with solving tasks such as faucet opening/closing, pick and place, assembly/disassembly and many others. 
Due to computational considerations, we selected 6 tasks which range from easy to difficult: \texttt{drawer-close-v2} (push the drawer closed),
\texttt{hand-insert-v2} (place the hand inside the hole),
\texttt{soccer-v2} (hit the soccer ball to a specific location in the goal box), \texttt{sweep-into-v2} (push the block into the hole), \texttt{assembly-v2} (grasp the nut and place over the thin block), and  \texttt{disassembly-v2} (grasp the nut and remove from the thin block). 

In Metaworld, the raw actions are delta positions, while the end-effector orientation remains fixed. For fairness, we disabled the use of any rotation primitives for this suite. Metaworld has a hand designed dense reward per task which enables efficient learning, but is unrealistic for the real world in which it can be challenging to design dense rewards without access to the true state of the world. Instead, for more realistic evaluation, we run all methods with a sparse reward which uses the success metric emitted by the environment itself. The low level action space for these environments uses 3DOF end-effector control along with grasp control; we implement the primitives using this action space. 

\begin{figure}
    \centering
    \includegraphics[width=\textwidth]{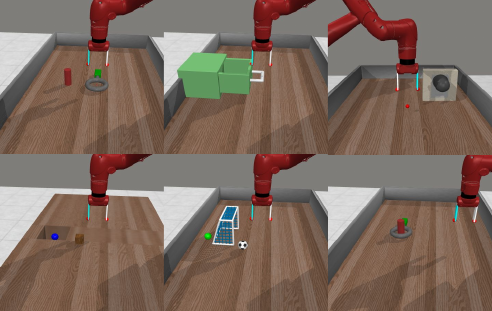}
    \caption{Visual depiction of the Metaworld environment suite. For the top row from the left: \texttt{assembly-v2}, \texttt{drawer-close-v2}, \texttt{peg-unplug-side-v2}. For the bottom row from the left: \texttt{sweep-into-v2}, \texttt{soccer-v2}, \texttt{disassemble-v2}.}
    \label{fig:metaworld}
\end{figure}

We run the environments in single task mode, meaning the target positions remain the same across experiments, in order to evaluate the basic effectiveness of RL across action spaces. This functionality is provided in the latest release of Metaworld. Additionally, we use the V2 versions of the tasks after correspondence with the current maintainers of the benchmark. The V2 environments have a more realistic visual appearance, improved reward functions and are now the primarily supported environments in Metaworld.
See Figure~\ref{fig:metaworld} for a visualization of the Metaworld tasks.

\subsection{Robosuite}
\begin{figure}
    \centering
    \includegraphics[width=.65\textwidth]{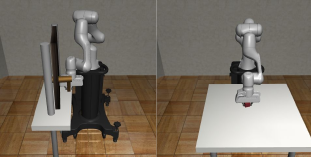}
    \caption{Visual depiction of the Robosuite environments. On the left we have the door opening task, and on the right we have the block lifting task.}
    \label{fig:robosuite}
\end{figure}

Robosuite is a benchmark of robotic manipulation tasks which emphasizes realistic simulation and control while containing several tasks existing RL algorithms struggle to solve, even when provided state based information and dense rewards. This suite contains a torque based end-effector position control implementation, Operational Space Control~\citep{khatib1987unified}. We select the \texttt{lift} and \texttt{door} tasks for evaluation, which we visualize in Figure \ref{fig:robosuite}. The lifting task involves accurately grasping a small red block and lifting it to a set height. The door task involves grasping the door handle, pushing it down to unlock it and pulling it open to a set position. These tasks contain initial state randomization; at each reset the position of the block or door is randomized within a small range. This property makes the Robosuite tasks more challenging than Kitchen and Metaworld, both of which are deterministic environments. For this environment, sparse rewards were already defined so we directly use them in our experiments. We made several changes to these environments to improve learning performance of the baselines as well as RAPS\xspace. Specifically, we included a workspace limit in a large area around the object, which improves exploration in the case of sparse rewards. For the lifting task, we increased the frequency of the default OSC controller to 40Hz from 20Hz, while for the door opening task we changed the max action magnitude to .1 from .05. We define the low level action space for this suite to use 3DOF end-effector control along with grasp control; we implement the primitives using this action space. 

\section{Primitive Implementation Details}
\label{sec:appendix-prim-impl}
In this section, we provide specific implementation details regarding the primitives we use in our experiments. In particular, we use an end-effector pose controller as $C_k$ for all $k$. We compute the target state $s^*$ using the components of the robot state which correspond to the input arguments of the primitive, $s_{\operatorname{args}}$. We compute $s^*$ using the formula $s^* = s_{\operatorname{args}} + \operatorname{args}$. The error metric is computed in a similar manner $e_k = s^* - s_{\operatorname{args}}$ across primitives. Returning to the lifting primitive example in the main text, $s_{\operatorname{args}}$ would be the z position of the end-effector, $s^*$ would be the target z position after lifting, and $e_k$ would be the difference between the target z position and the current z position of the end-effector. In Table \ref{table:primitives list} we provide additional details regarding each primitive including the search spaces, number of low-level actions and which environment it was used in. One primitive of note is go to pose (delta) which performs delta position control. Using this primitive alongside the grasp and release primitives corresponds closely to the raw action space for Metaworld and Robosuite, environment suites in which we do not use orientation control. 

We tuned the low-level actions per environment suite, but one could alternatively design a tolerance threshold and loop until it is achieved to avoid any tuning. We chose a fixed horizon which runs significantly faster and any inaccuracies in the primitives are accounted for by the learned policy. Finally, we do not use every primitive in every domain, yet across all tasks within a domain we use the same library. In Metaworld, the raw action space does not allow for orientation control so we do not either. Enabling orientation control with primitives can, in certain cases, make the task easier, but we do not include the x-axis and y-axis rotation primitives for fair comparison. In Robosuite, the default action space has orientation control. We found orientation control was unnecessary in order to solve the lifting and door opening tasks when we disabled orientation control for raw actions and for primitives. As a result, in this work we report results without orientation control in Robosuite.

\section{RL Implementation Details}
\label{sec:appendix-rl-impl}
Whenever possible, we use the original implementations of any method we compare against. We use standard implementations for each base RL algorithm except Dreamer, which we implement in PyTorch. We use the actor and model hyper-parameters from Dreamer-V2~\citep{hafner2020mastering} as we found it slightly improved the performance of Dreamer. For primitives, we made several hyper-parameter changes to better tailor Dreamer to hybrid discrete-continuous control. Specifically, instead of backpropagating the return through the dynamics, we use REINFORCE to train the actor in imagination. We additionally reduce the imagination trajectory length from 15 to 5 for the single task primitive experiments since the trajectory length is limited to 5 in any case. With the short trajectory lengths in RAPS\xspace, imagination often goes beyond the end of the episode, so we use a discount predictor to downweight imagined states beyond the end of the episode.
Finally since we cannot sample batch lengths of 50 from trajectories of length 5 or 15, we instead sample the full primitive trajectory and change the batch size to be $\frac{2500}{H}$, the primitive horizon. This results in an effective batch size of 2500, which is equal to the Dreamer batch size of 50 with a batch length of 50. 

In the case of SAC, we use the implementation of SAC~\citep{laskin2020reinforcement} but without data augmentation, which amounts to using their specific pixel encoder which we found to perform well. Finally for PPO, we use the following implementation:~\citet{pytorchrl}. See Tables \ref{tab:dreamer-hyper-params}, \ref{tab:sac-hyper-params}, \ref{tab:ppo-hyper-params} for the hyper-parameters used for each algorithm respectively. We use the same algorithm hyper-parameters across all the baselines. For primitives, we modify the discount factor in all experiments to $1-\frac{1}{H}$, in which $H$ is the primitive horizon. This encourages the agent to highly value near term rewards with short horizons. For single task experiments, we use a horizon of $5$, taking $5$ primitive actions in one episode, with a discount of $0.8$. For the hierarchical control experiments we use a horizon of 15 and a corresponding discount of $.93$. In practice, this method of computing the discount factor improves the performance and stability of RAPS\xspace. 

For each baseline we use the original implementation when possible as an underlying action space for each RL algorithm. For VICES, we take the impedance controller from the iros\_19\_vices branch and modify the environment action space to output the parameters for the controller. For PARROT, we use an unreleased version of the code provided by the original authors. For SPIRL, we use an improved version of the method which was released to the SPIRL code base recently. This version, SPIRL-CL, uses a closed loop decoder to map latents back to action trajectories which they find significantly improves performance on the Kitchen environment from state input. We use the authors' code for vision-based SPIRL-CL and still find that RAPS\xspace performs better.

\begin{table}[]
    \centering
    \begin{tabular}{c|c}
         Hyper Parameter & Value \\
         \hline
         Actor output distribution & Truncated Normal \\
         Discount factor & 0.99 \\
         $\lambda_{GAE}$ & 0.95 \\
         actor and value function learning rates & 8e-5\\
         world model learning rate & 3e-4 \\
         Imagination horizon & 15 \\ 
         Entropy coefficient & 1e-4 \\
         Predict discount & No \\ 
         Target value function update period & 100 \\
         reward loss scale & 2 \\
         Model hidden size & 400 \\
         Stochastic state size & 50 \\
         Deterministic state size & 200 \\
         Embedding size & 1024 \\
         RSSM hidden size & 200 \\
         Use GRU layer norm & Yes \\
         Actor hidden layers & 4 \\
         Value hidden layers & 3 \\
         batch size & 50 \\ 
         batch length & 50 \\
    \end{tabular}
    \caption{Dreamer hyper-parameters}
    \label{tab:dreamer-hyper-params}
\end{table}

\begin{table}[]
    \centering
    \begin{tabular}{c|c}
         Hyper Parameter & Value \\
         \hline
         Discount factor & 0.99 \\ 
         actor, critic, encoder learning rates & 2e-4 \\ 
         alpha learning rate & 1e-4 \\
         Target network update frequency & 2\\
         Polyak averaging constant & .01 \\ 
         Frame stack & 4\\
         Image size & 64 \\
         Random policy warm up steps & 2500 \\
         batch size & 512 \\ 
    \end{tabular}
    \caption{SAC hyper-parameters}
    \label{tab:sac-hyper-params}
\end{table}

\begin{table}[]
    \centering
    \begin{tabular}{c|c}
         Hyper Parameter & Value \\
         \hline
         Entropy coefficient & .01 \\
         Value loss coefficient & 0.5 \\
         Actor-value network learning rate & 3e-4\\
         Number of mini-batches per epoch & 10 \\
         PPO clip parameter & 0.2 \\ 
         Max gradient norm & 0.5 \\ 
         $\lambda_{GAE}$ & 0.95 \\ 
         Discount factor & 0.99 \\ 
         Number of parallel environments & 12 \\
         Frame stack & 4 \\
         Image size & 84 \\
    \end{tabular}
    \caption{PPO hyper-parameters}
    \label{tab:ppo-hyper-params}
\end{table}

\begin{table}[t]
    \centering
    \resizebox{\textwidth}{!}{%
    \begin{tabular}{c|c|c|c|c}
        \textbf{Primitive Skill} & \textbf{Parameters} & \textbf{Action Space} & \textbf{\# low-level actions} & \textbf{Environments} \\
        \hline
        grasp & d & [0,1] & 150-200 & Kitchen, Metaworld, Robosuite\\
        release & d & [-1,0] & 200-300 & Kitchen, Metaworld, Robosuite\\
        lift & z & [0, 1] & 40-300 & Kitchen, Metaworld, Robosuite\\
        drop & z & [-1, 0]& 40-300 & Kitchen, Metaworld, Robosuite\\
        push & y & [0, 1] &40-300 & Kitchen, Metaworld, Robosuite\\
        pull & y & [-1, 0] & 40-300& Kitchen, Metaworld, Robosuite\\
        shift right & x & [0, 1] & 40-300&Kitchen, Metaworld, Robosuite\\ 
        shift left & x & [-1, 0] & 40-300& Kitchen, Metaworld, Robosuite\\ 
        go to pose (delta) & x,y,z & [-1, 0]$^3$& 40-300& Kitchen, Metaworld, Robosuite\\
        x-axis twist & $\theta$ & [$-\pi$, $\pi$] & 300 & Kitchen \\
        y-axis twist & $\theta$ & [$-\pi$, $\pi$] & 300 & Kitchen \\ 
        angled forward grasp & $\theta$, x, y, d & [$-\pi$, $\pi$],  [-1, 0]$^3$ & 1100 & Kitchen\\ 
        top z grasp & z,d & [-1, 0]$^2$& 140-250 & Robosuite\\
        top grasp & x,y,z,d & [-1, 0]$^4$ & 1500 & Metaworld \\
        
    \end{tabular}}
    \vspace{.25cm}
    \caption{Description of skill parameters, search spaces, low-level actions and environment usage.}
    \label{table:primitives list}
\end{table}

\begin{figure}
    \centering
    \begin{tabular}{cc}
          \includegraphics[width=.24\textwidth]{chapters/raps/figures/final_plots/action_parameterizations/kitchen_single_task_Dreamer_kettle_time.pdf}
    \includegraphics[width=.24\textwidth]{chapters/raps/figures/final_plots/action_parameterizations/kitchen_single_task_Dreamer_microwave_time.pdf} &
    \includegraphics[width=.24\textwidth]{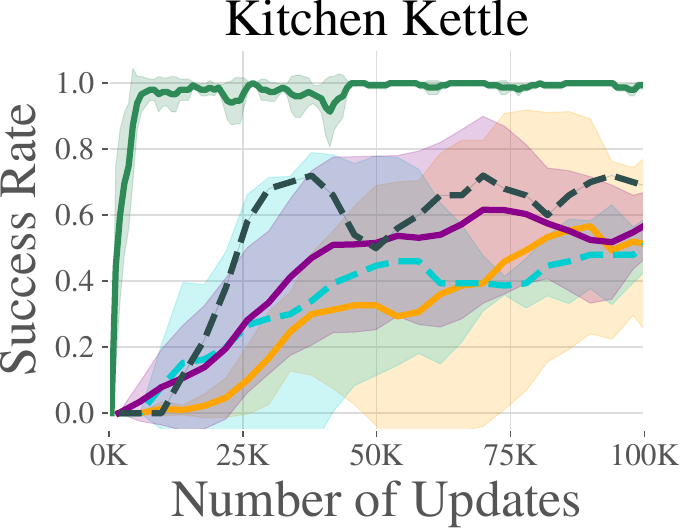}
    \includegraphics[width=.24\textwidth]{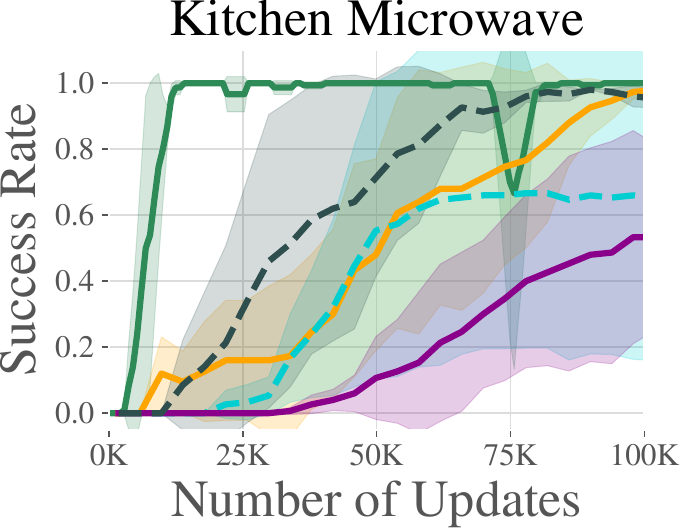}
    \vspace{.2cm}
    \\
    \includegraphics[width=.24\textwidth]{chapters/raps/figures/final_plots/action_parameterizations/kitchen_single_task_Dreamer_light_switch_time.pdf}
    \includegraphics[width=.24\textwidth]{chapters/raps/figures/final_plots/action_parameterizations/kitchen_single_task_Dreamer_top_left_burner_time.pdf} &
    \includegraphics[width=.24\textwidth]{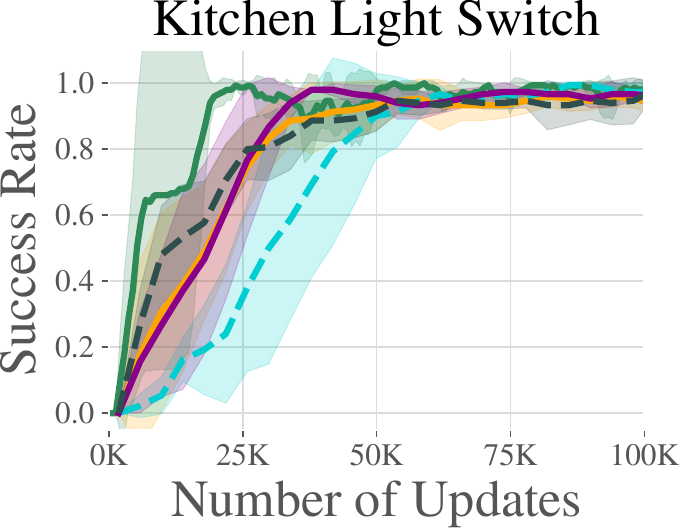}
    \includegraphics[width=.24\textwidth]{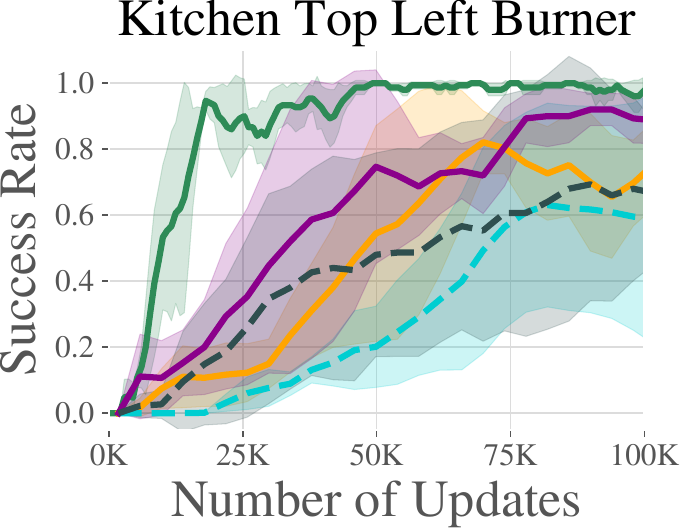}
    \vspace{.2cm}
    \\
    \includegraphics[width=.24\textwidth]{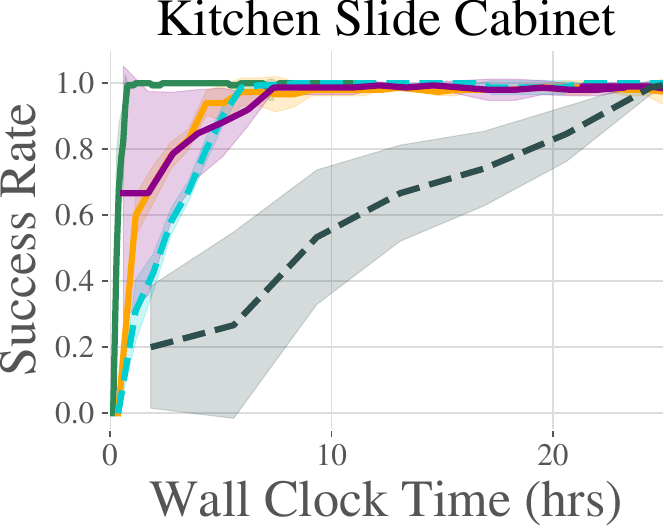}
    \includegraphics[width=.24\textwidth]{chapters/raps/figures/final_plots/action_parameterizations/kitchen_single_task_Dreamer_hinge_cabinet_time.pdf} &
    \includegraphics[width=.24\textwidth]{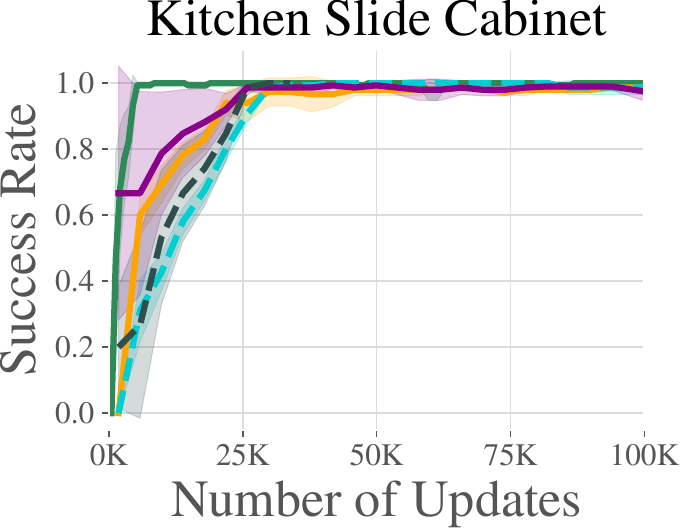}
    \includegraphics[width=.24\textwidth]{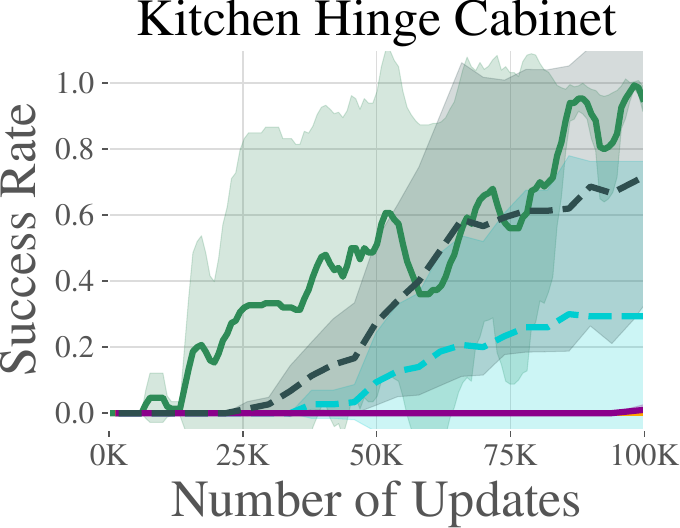}
    \vspace{.2cm}
    \\
    \includegraphics[width=.24\textwidth]{chapters/raps/figures/final_plots/action_parameterizations/metaworld_single_task_Dreamer_drawer-close-v2_time.pdf}
    \includegraphics[width=.24\textwidth]{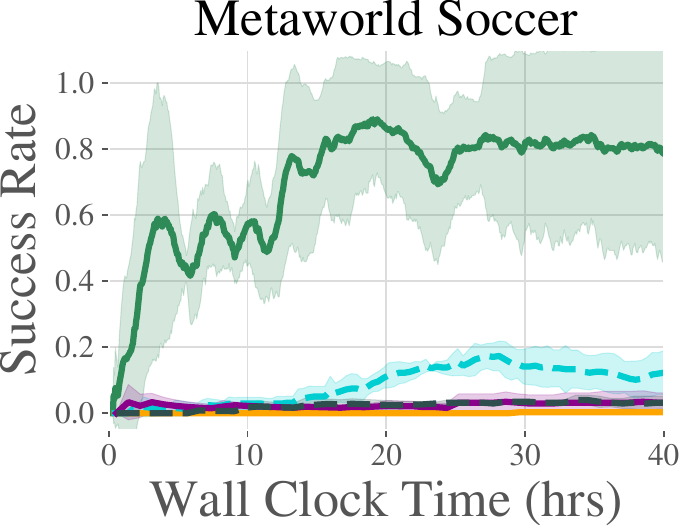} &
    \includegraphics[width=.24\textwidth]{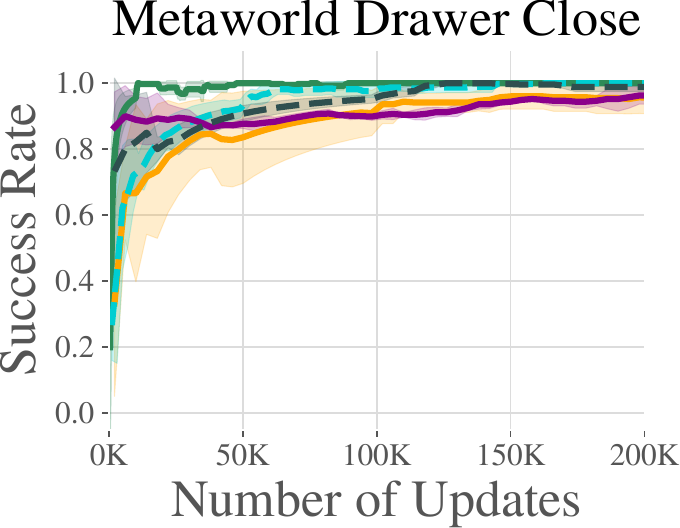}
    \includegraphics[width=.24\textwidth]{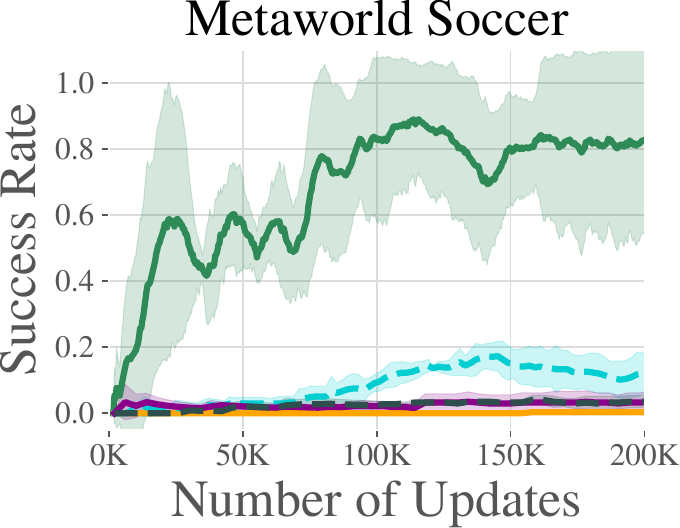}
    \vspace{.2cm}
    \\
    \includegraphics[width=.24\textwidth]{chapters/raps/figures/final_plots/action_parameterizations/metaworld_single_task_Dreamer_peg-unplug-side-v2_time.pdf}
    \includegraphics[width=.24\textwidth]{chapters/raps/figures/final_plots/action_parameterizations/metaworld_single_task_Dreamer_sweep-into-v2_time.pdf} &
     \includegraphics[width=.24\textwidth]{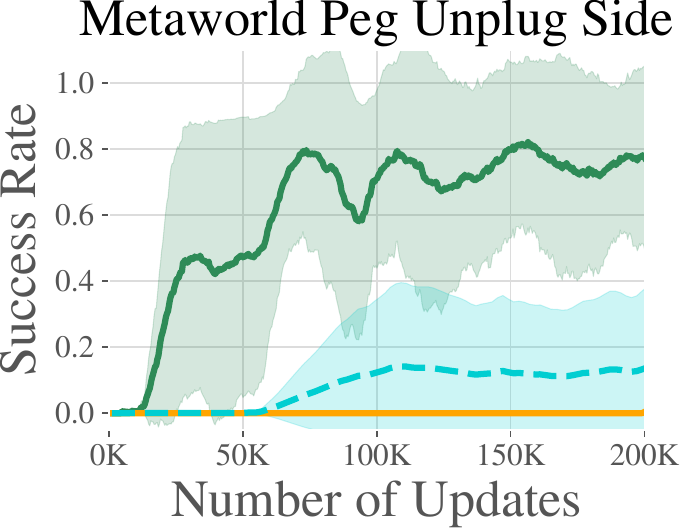}
    \includegraphics[width=.24\textwidth]{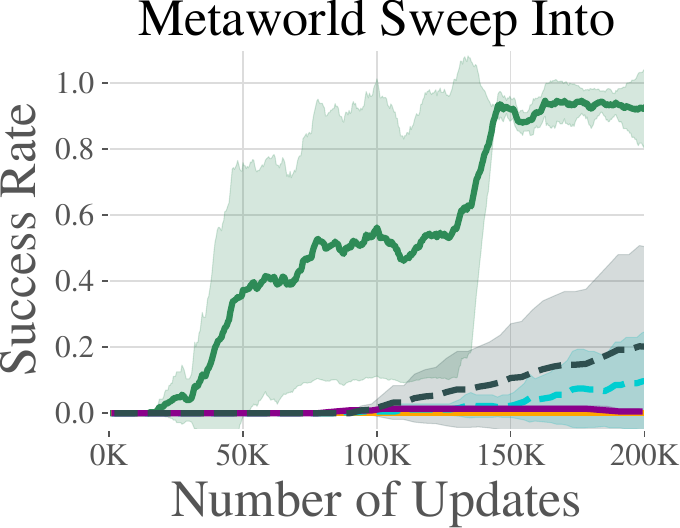}
    \vspace{.2cm}
    \\
    \includegraphics[width=.24\textwidth]{chapters/raps/figures/final_plots/action_parameterizations/metaworld_single_task_Dreamer_disassemble-v2_time.pdf}
    \includegraphics[width=.24\textwidth]{chapters/raps/figures/final_plots/action_parameterizations/metaworld_single_task_Dreamer_assembly-v2_time.pdf} &
    \includegraphics[width=.24\textwidth]{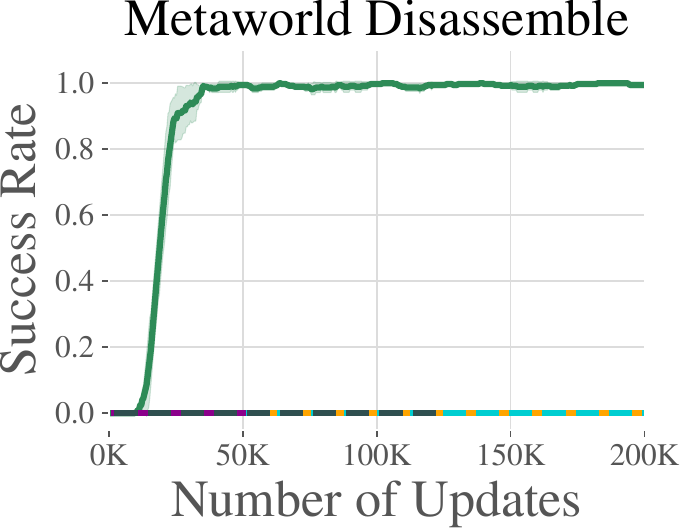}
    \includegraphics[width=.24\textwidth]{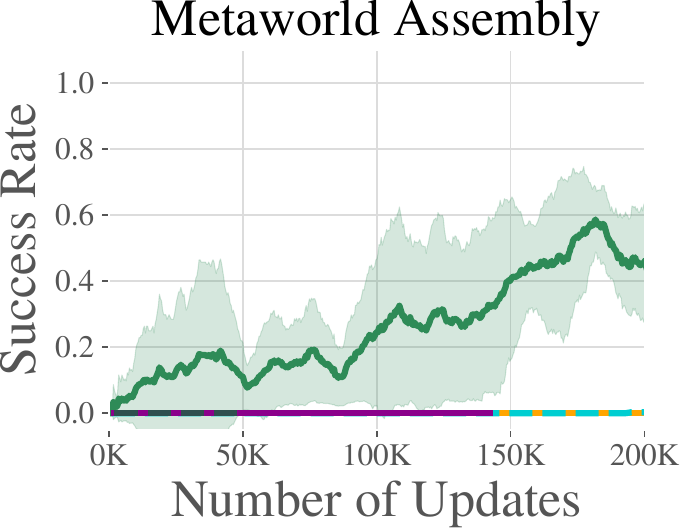}
    \vspace{.2cm}
    \\
    \includegraphics[width=.24\textwidth]{chapters/raps/figures/final_plots/action_parameterizations/robosuite_single_task_Dreamer_Door_time.pdf}
    \includegraphics[width=.24\textwidth]{chapters/raps/figures/final_plots/action_parameterizations/robosuite_single_task_Dreamer_Lift_time.pdf} &
    \includegraphics[width=.24\textwidth]{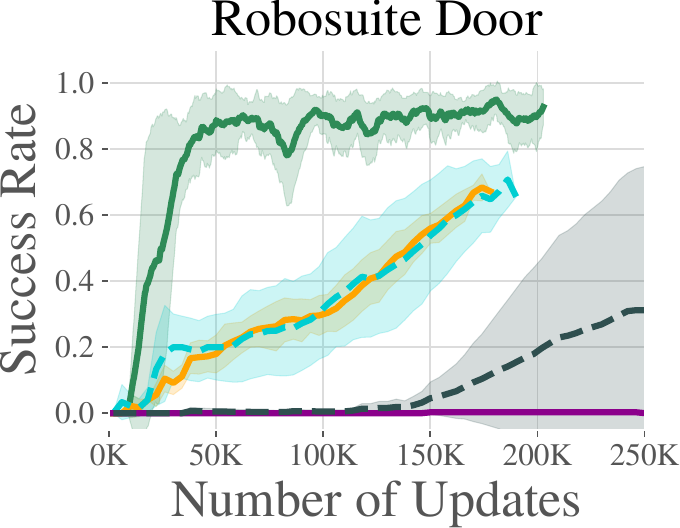}
    \includegraphics[width=.24\textwidth]{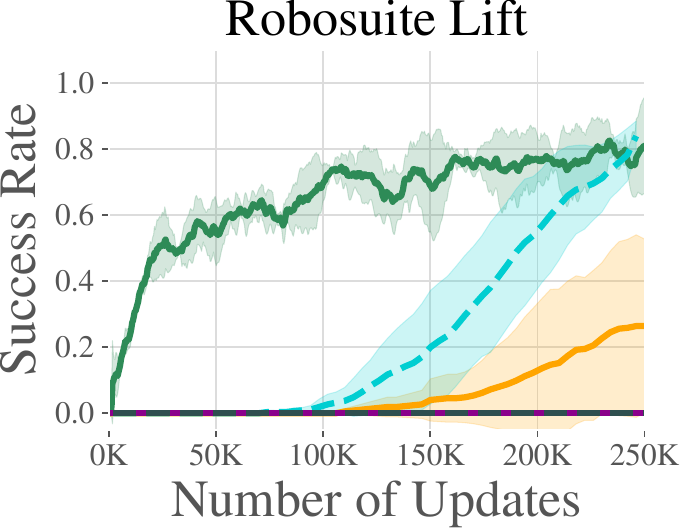}
    \end{tabular}
    \includegraphics[width=\textwidth]{chapters/raps/figures/final_plots/legends/action_param_legend.pdf}
    \vspace{-0.2in}
    \caption{Full version of Figure 3 with excluded environments (\texttt{slide-cabinet} and \texttt{soccer-v2}) and plots against number of updates (right two columns). RAPS\xspace outperforms all baselines against number of updates as well.}
    \label{fig:appendix-action-param-results}
    \vspace{-0.1in}
\end{figure}

\begin{figure}[t]
    \centering
    \begin{tabular}{cc}
         \includegraphics[width=.24\textwidth]{chapters/raps/figures/final_plots/offline_skill_learning/kitchen_single_task_SAC_kettle_time.pdf}
    \includegraphics[width=.24\textwidth]{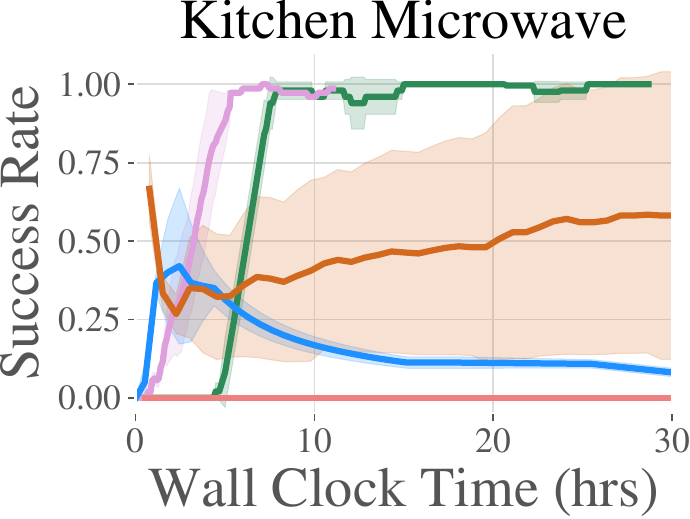} &
    \includegraphics[width=.24\textwidth]{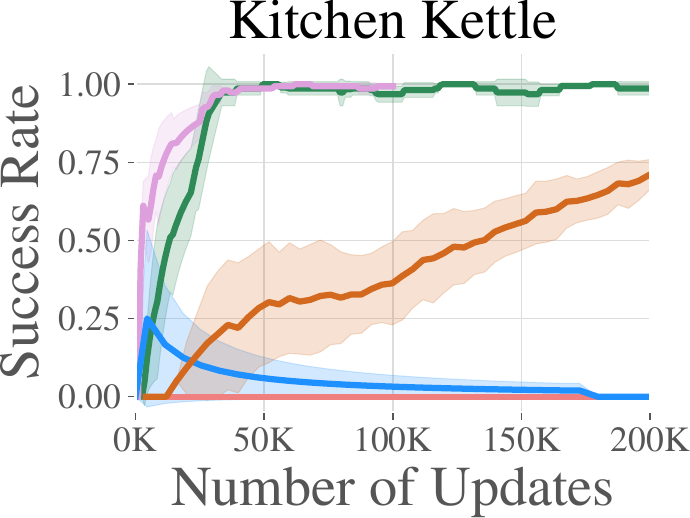}
    \includegraphics[width=.24\textwidth]{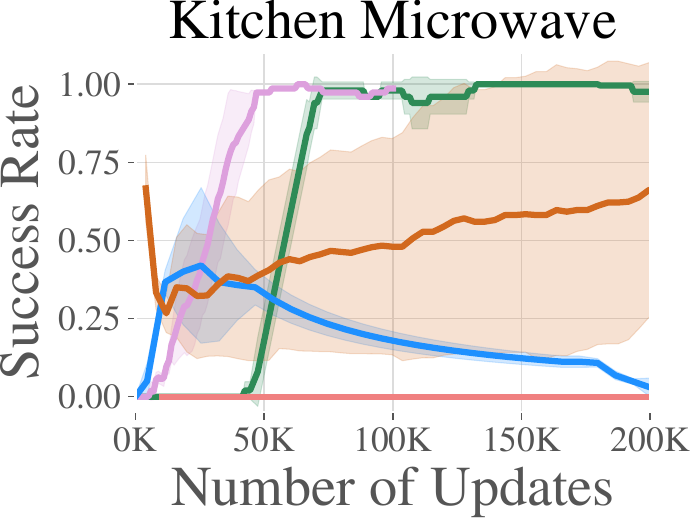}
    \vspace{.2cm}
    \\
    \includegraphics[width=.24\textwidth]{chapters/raps/figures/final_plots/offline_skill_learning/kitchen_single_task_SAC_slide_cabinet_time.pdf}
    \includegraphics[width=.24\textwidth]{chapters/raps/figures/final_plots/offline_skill_learning/kitchen_single_task_SAC_top_left_burner_time.pdf} &
    \includegraphics[width=.24\textwidth]{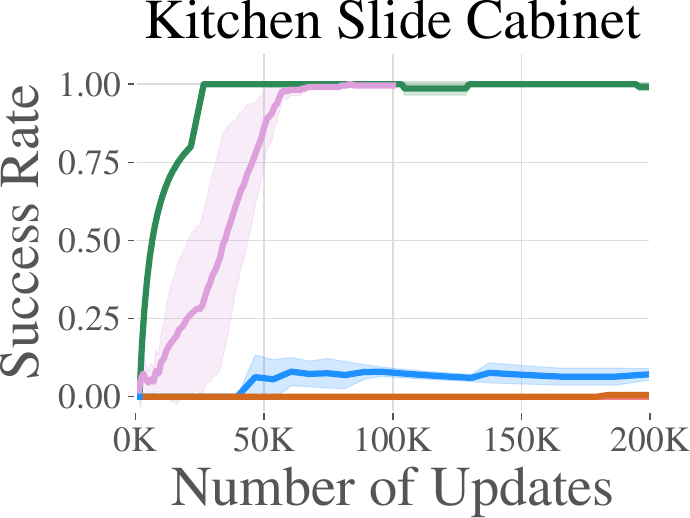}
    \includegraphics[width=.24\textwidth]{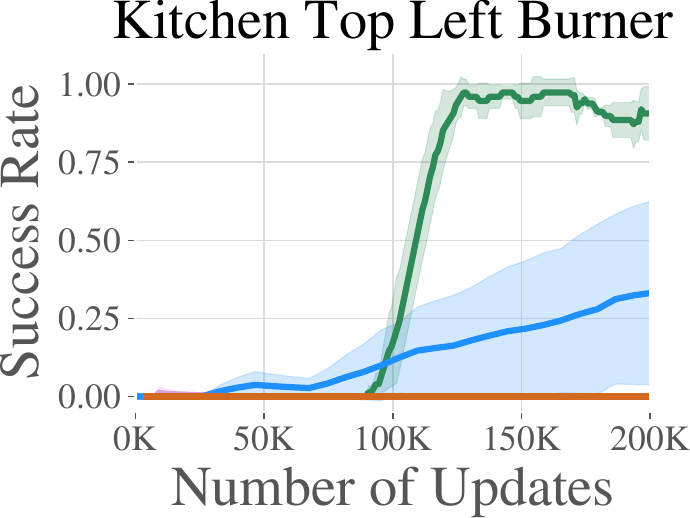}
    \vspace{.2cm}
    \\
    \includegraphics[width=.24\textwidth]{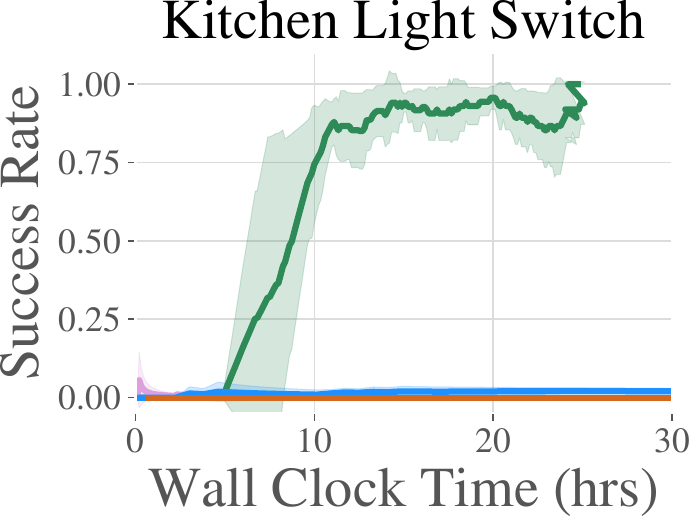}
    \includegraphics[width=.24\textwidth]{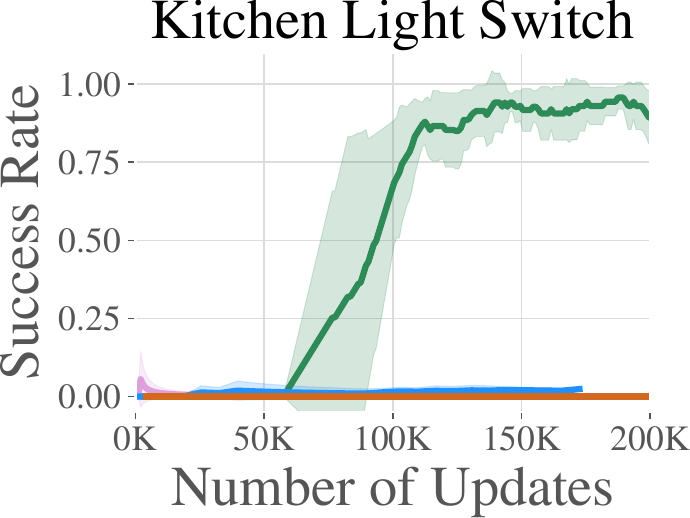} &
    \includegraphics[width=.24\textwidth]{chapters/raps/figures/final_plots/offline_skill_learning/kitchen_single_task_SAC_hinge_cabinet_time.pdf}
    \includegraphics[width=.24\textwidth]{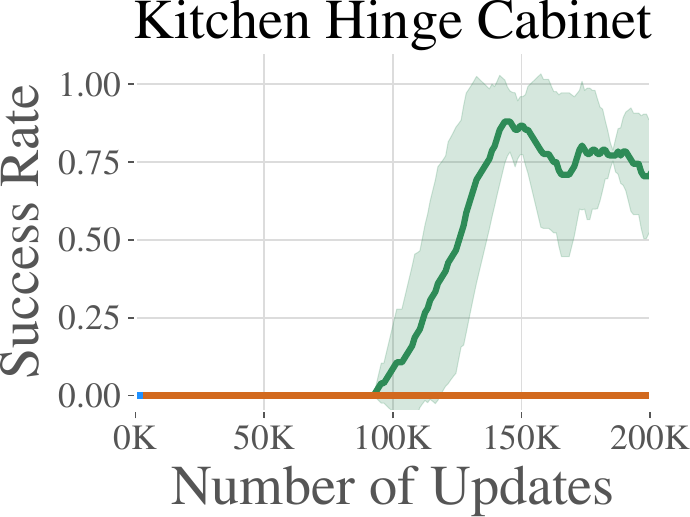}
    \end{tabular}
    \includegraphics[width=\textwidth]{chapters/raps/figures/final_plots/legends/offline_skill_extraction_legend.pdf}
    \vspace{-0.25in}
    \caption{Full version of Figure 4 with plots against number of updates (right column) and excluded environments (\texttt{light-switch}). While SPIRL is competitive with RAPS\xspace on the easier tasks, it fails to make progress on the more challenging tasks.}
    \label{fig:appendix-offline-skill-learn-results}
    \vspace{-0.05in}
\end{figure}

\begin{figure}
    \centering
\begin{tabular}{cc}
\includegraphics[width=.24\textwidth]{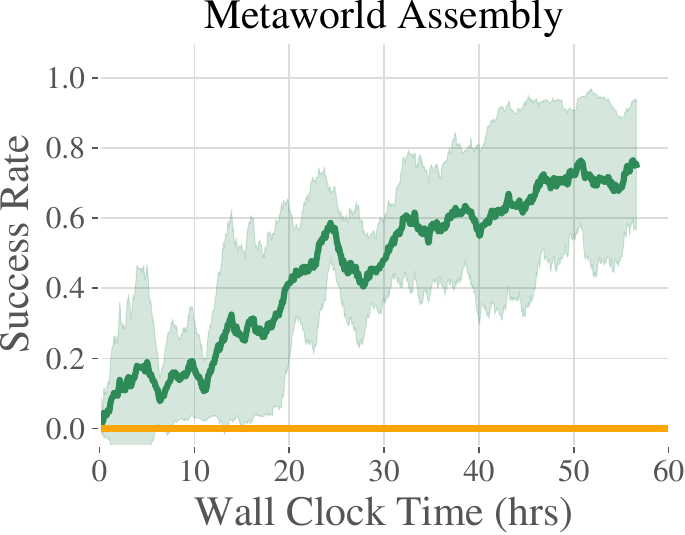}
\includegraphics[width=.24\textwidth]{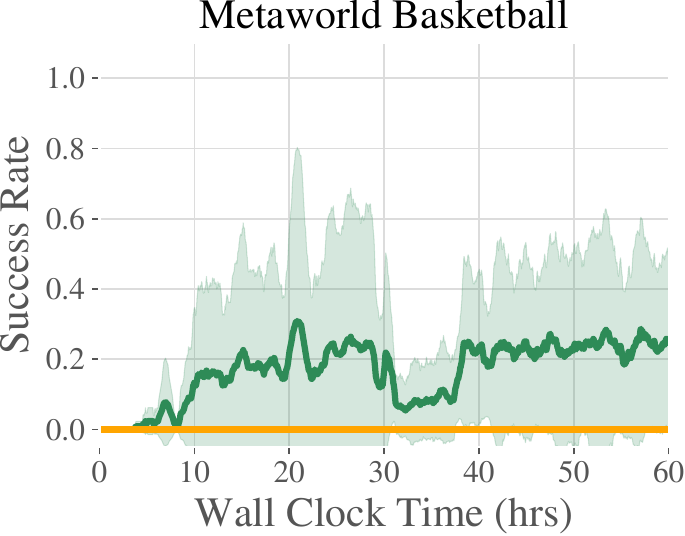} &
\includegraphics[width=.24\textwidth]{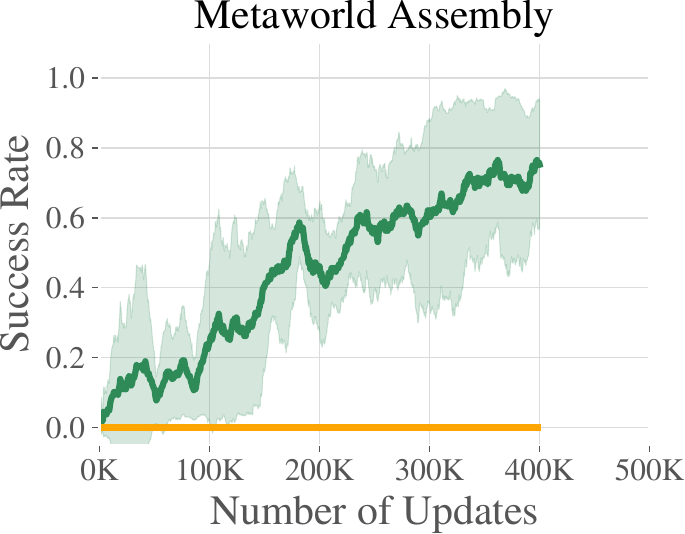}
\includegraphics[width=.24\textwidth]{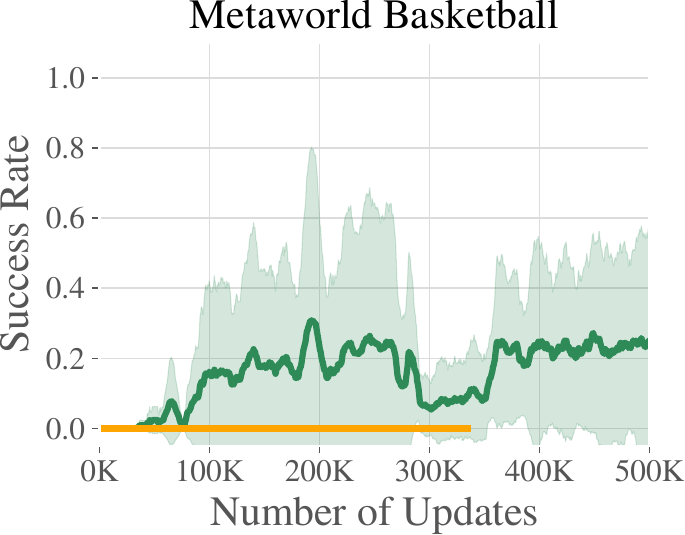}
\vspace{.2cm}
\\
\includegraphics[width=.24\textwidth]{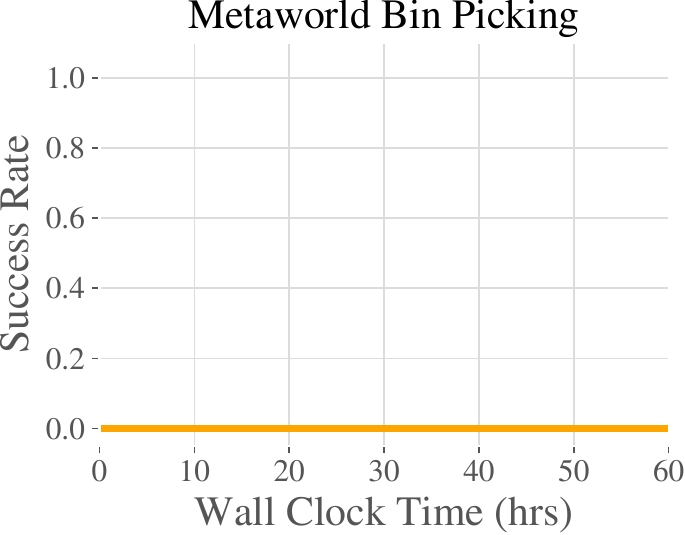}
\includegraphics[width=.24\textwidth]{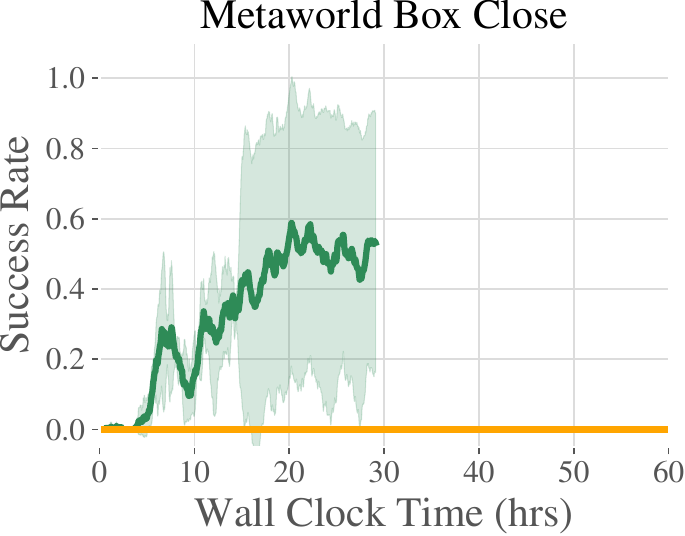} &
\includegraphics[width=.24\textwidth]{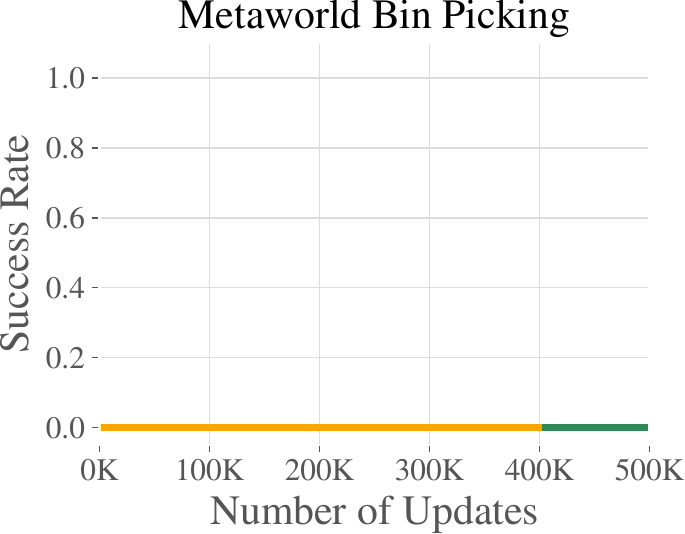}
\includegraphics[width=.24\textwidth]{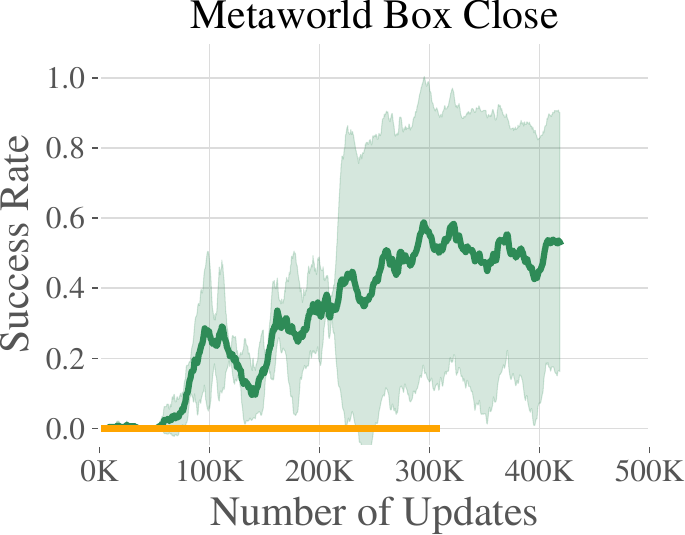}
\vspace{.2cm}
\\
\includegraphics[width=.24\textwidth]{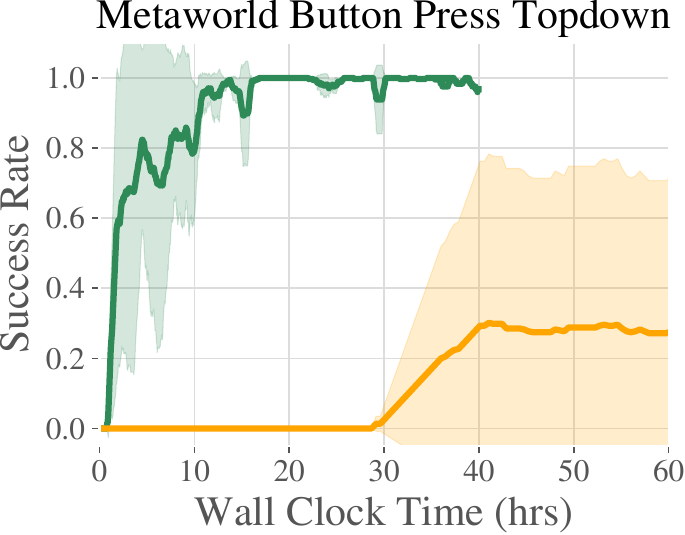}
\includegraphics[width=.24\textwidth]{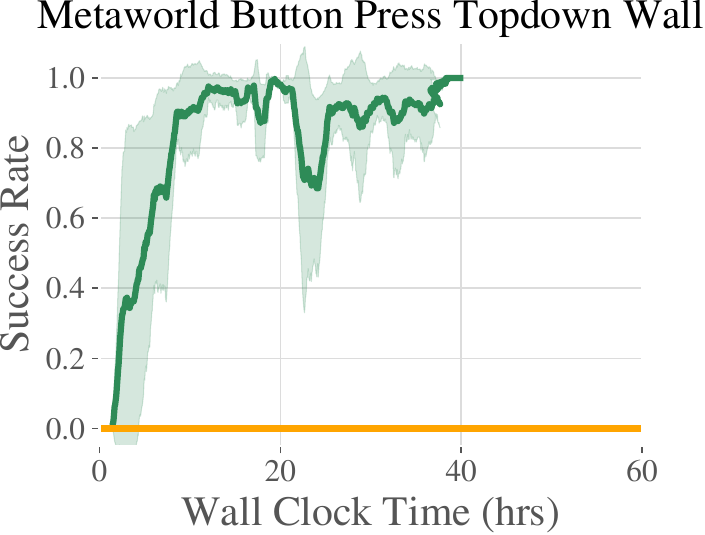} &
\includegraphics[width=.24\textwidth]{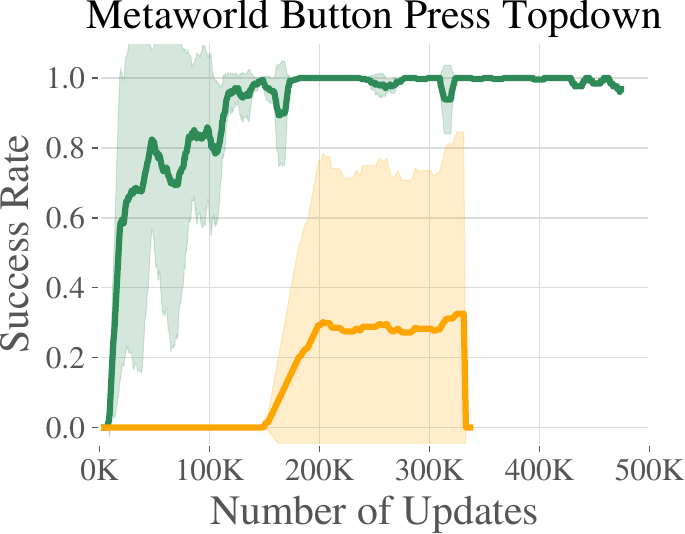}
\includegraphics[width=.24\textwidth]{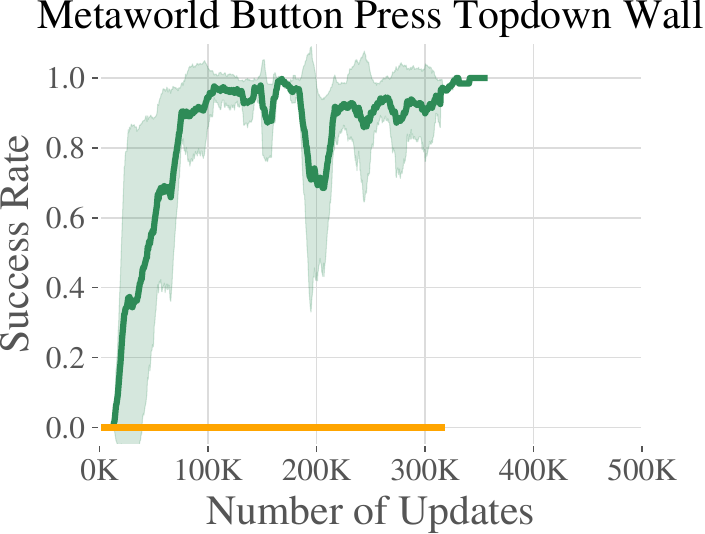}
\vspace{.2cm}
\\
\includegraphics[width=.24\textwidth]{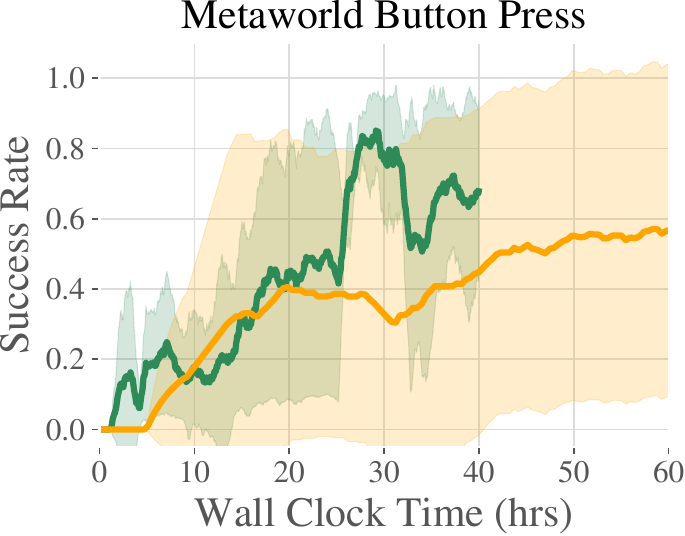}
\includegraphics[width=.24\textwidth]{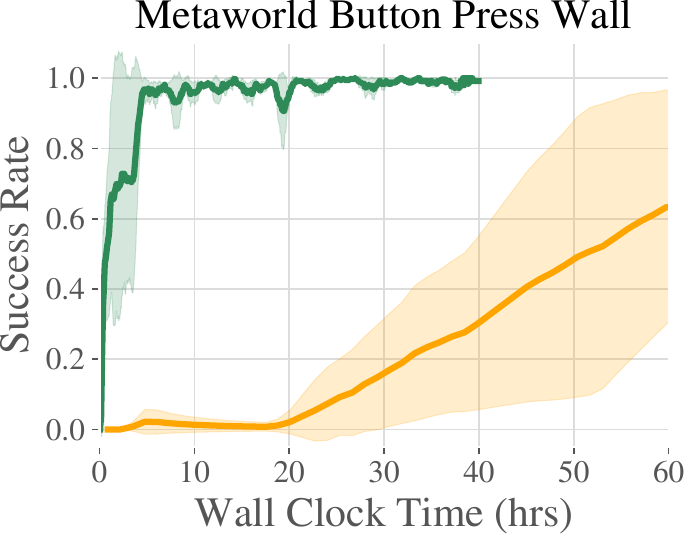} &
\includegraphics[width=.24\textwidth]{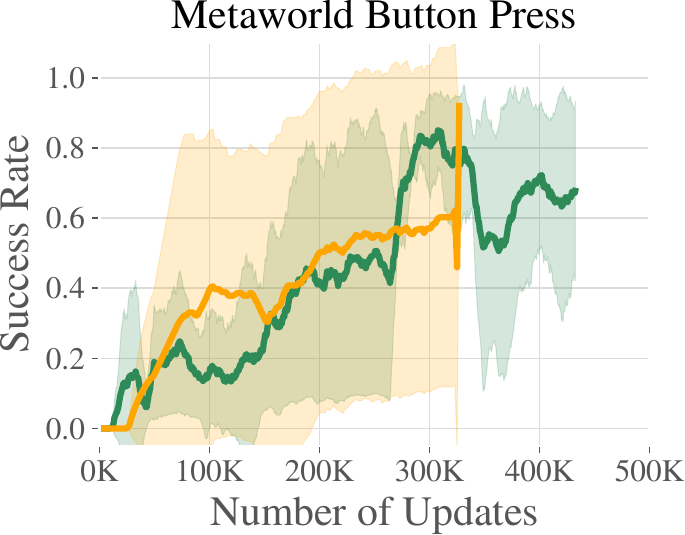}
\includegraphics[width=.24\textwidth]{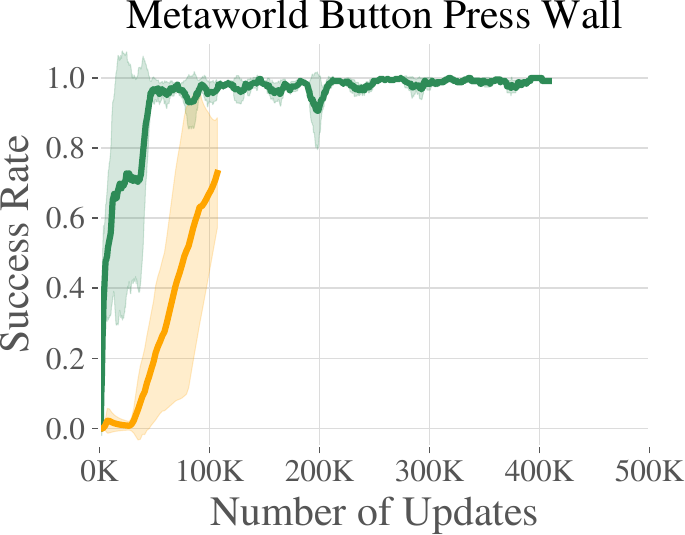}
\vspace{.2cm}
\\
\includegraphics[width=.24\textwidth]{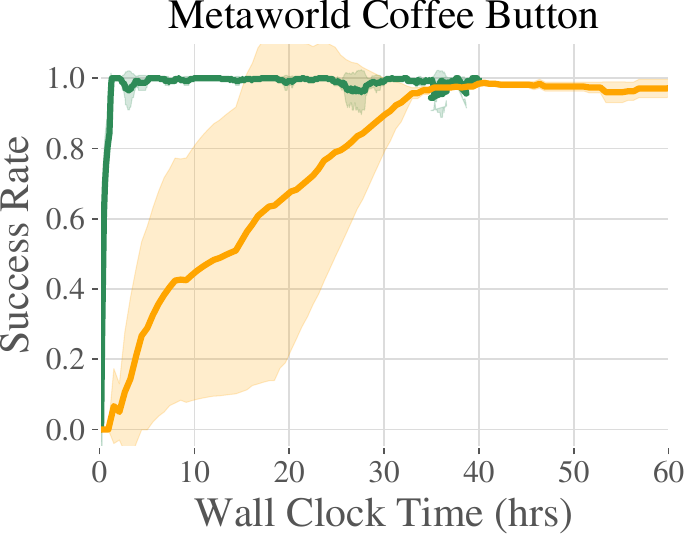}
\includegraphics[width=.24\textwidth]{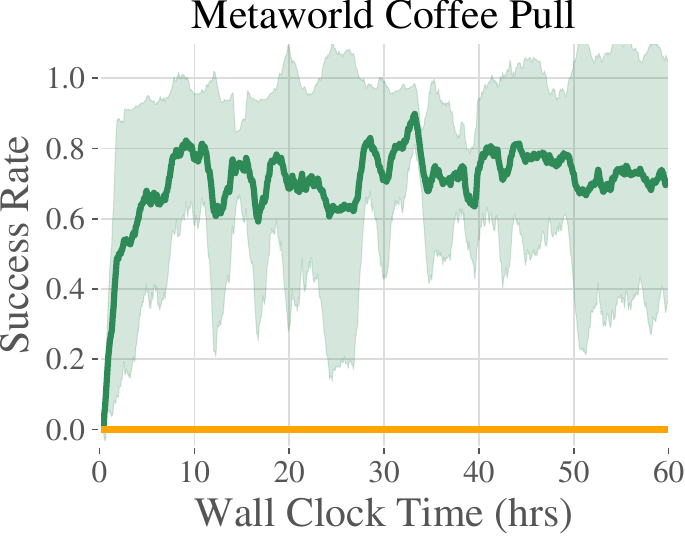} &
\includegraphics[width=.24\textwidth]{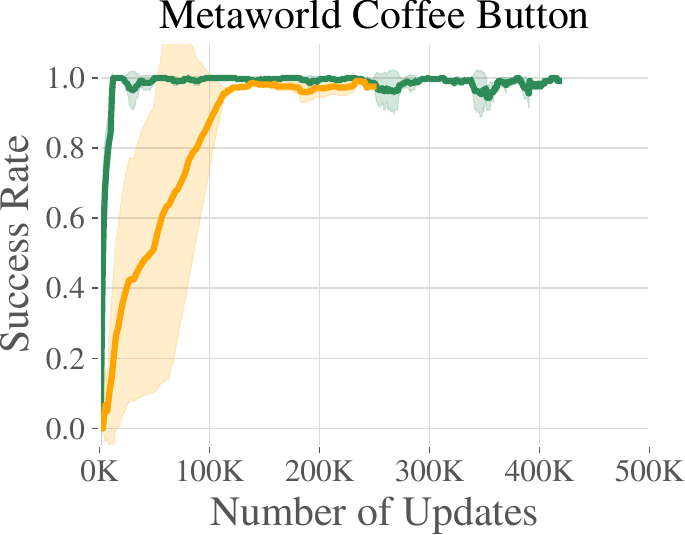}
\includegraphics[width=.24\textwidth]{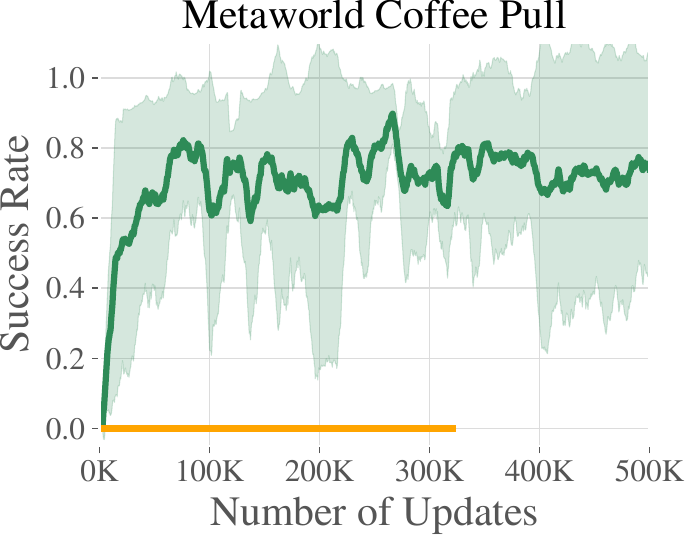}
\vspace{.2cm}
\\
\includegraphics[width=.24\textwidth]{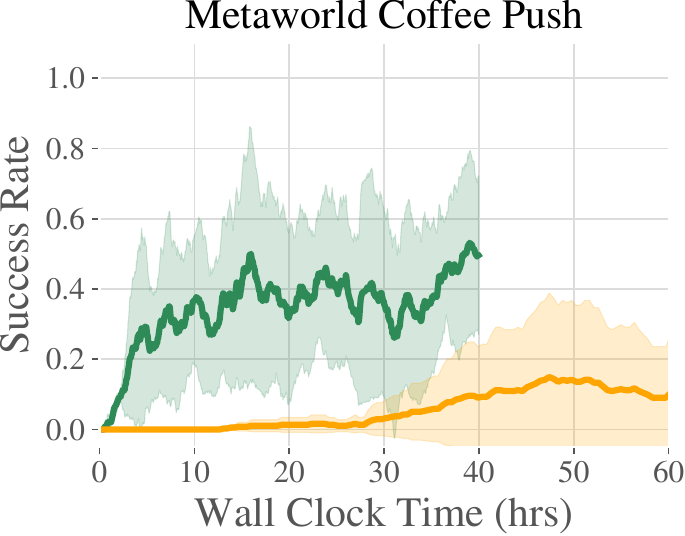}
\includegraphics[width=.24\textwidth]{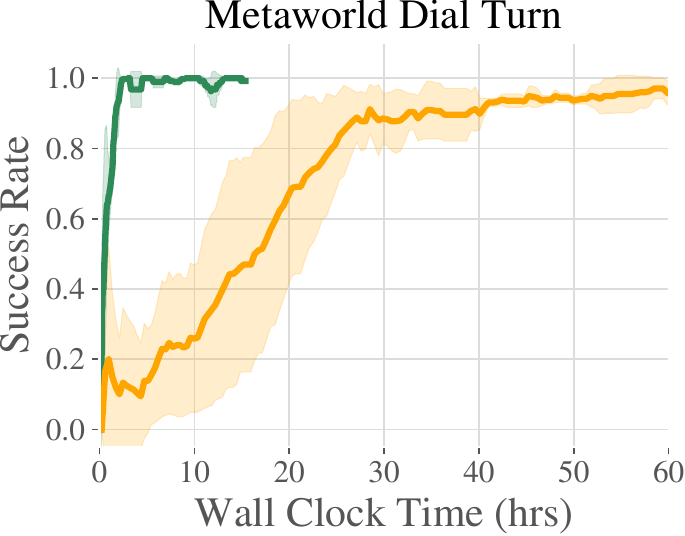} &
\includegraphics[width=.24\textwidth]{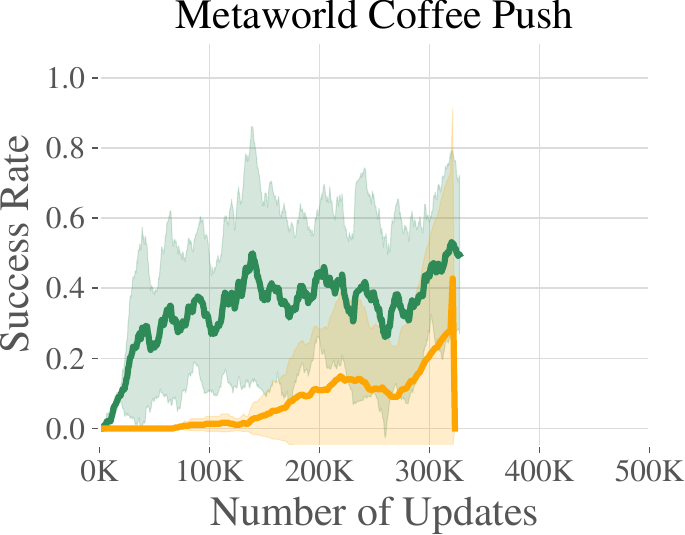}
\includegraphics[width=.24\textwidth]{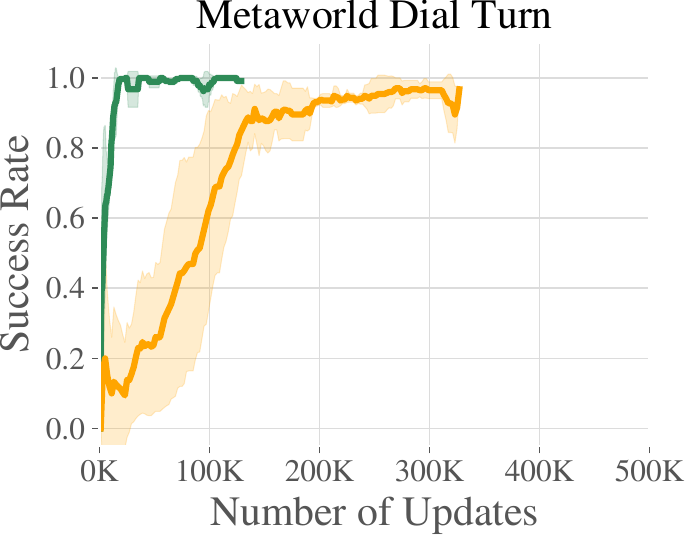}
\vspace{.2cm}
\\
\includegraphics[width=.24\textwidth]{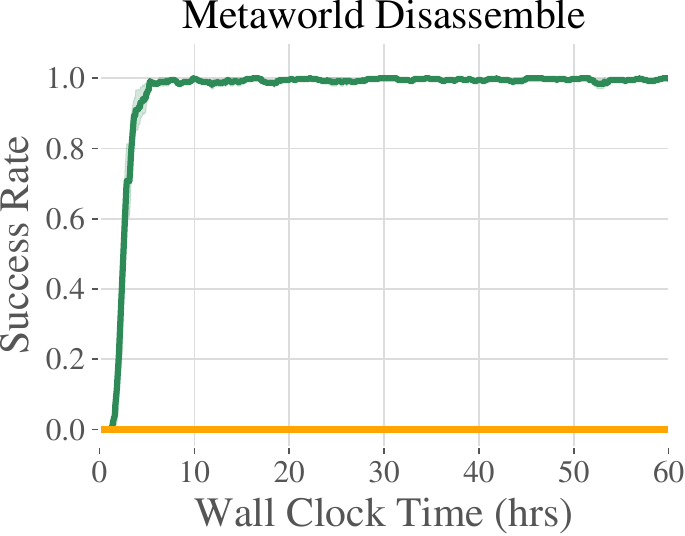}
\includegraphics[width=.24\textwidth]{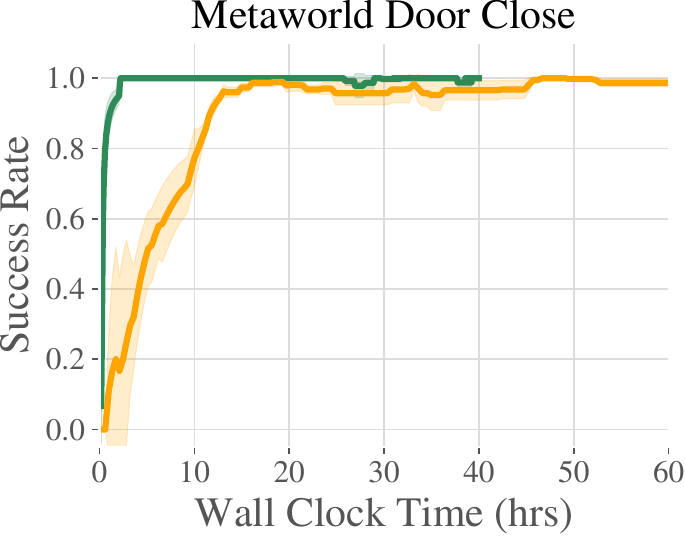} &
\includegraphics[width=.24\textwidth]{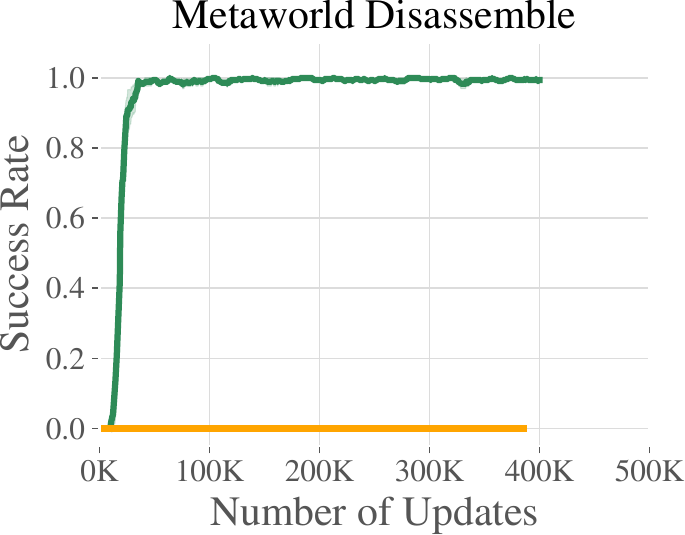}
\includegraphics[width=.24\textwidth]{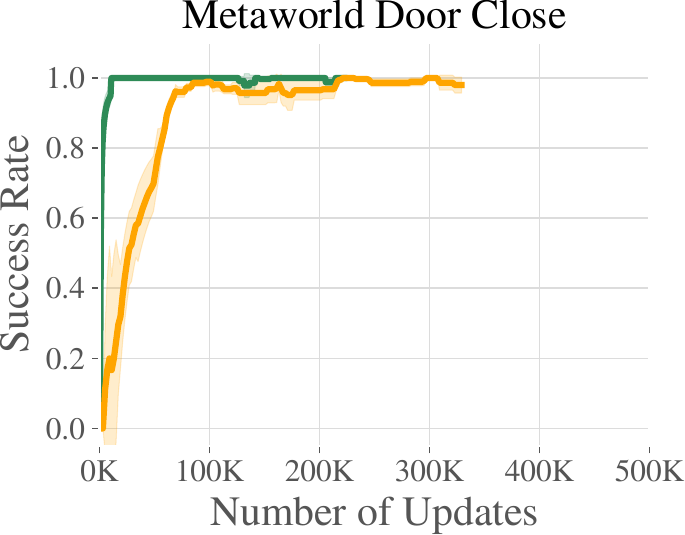}
\vspace{.2cm}
\\
\includegraphics[width=.24\textwidth]{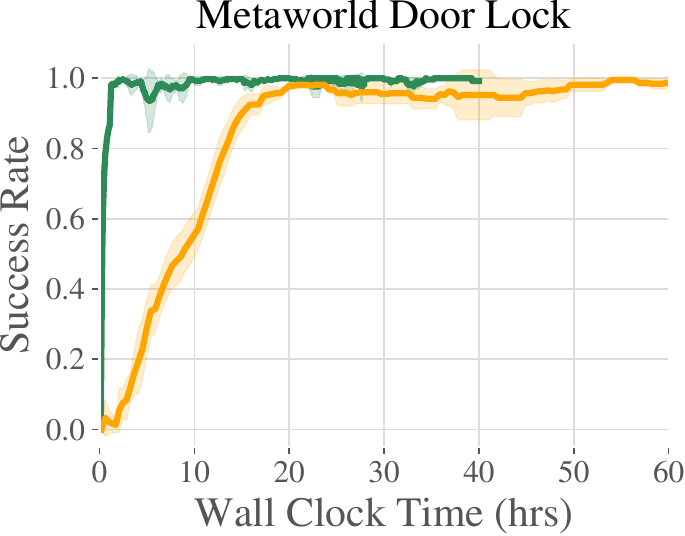}
\includegraphics[width=.24\textwidth]{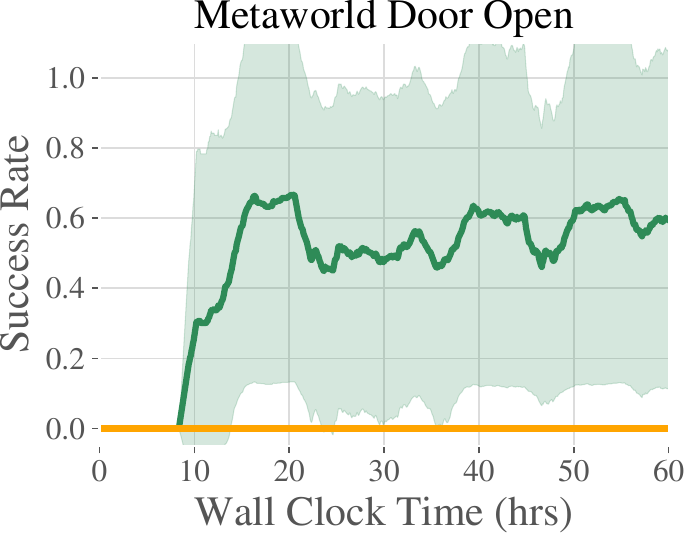} &
\includegraphics[width=.24\textwidth]{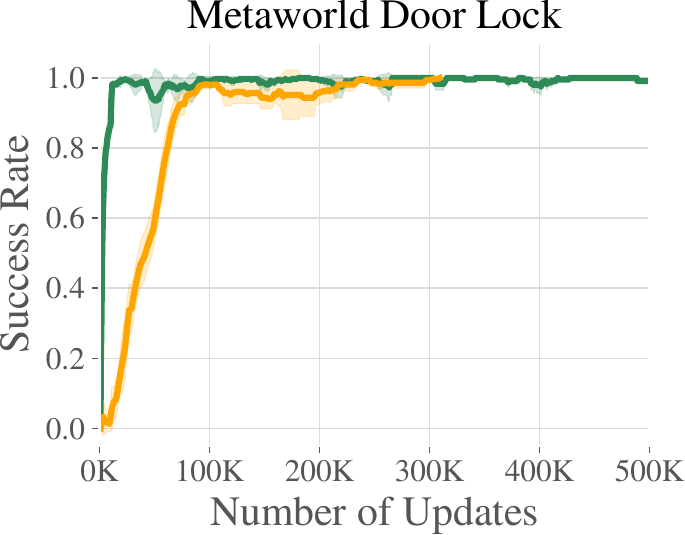}
\includegraphics[width=.24\textwidth]{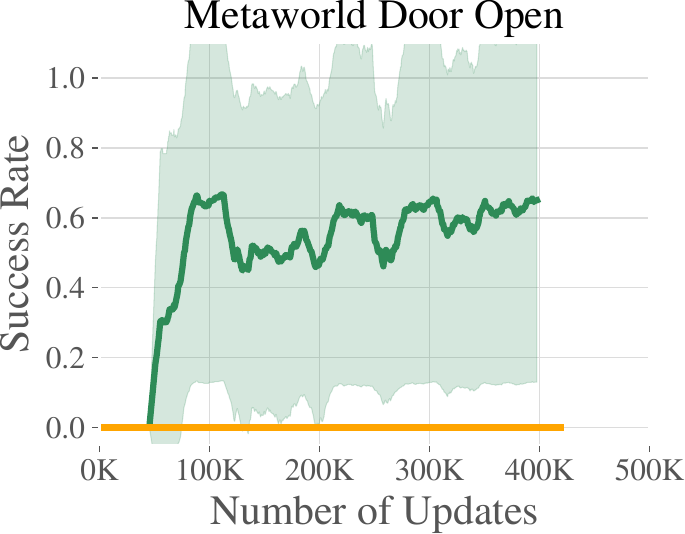}
    \end{tabular}

\end{figure}

\begin{figure}\ContinuedFloat
    \centering
    \begin{tabular}{cc}
\includegraphics[width=.24\textwidth]{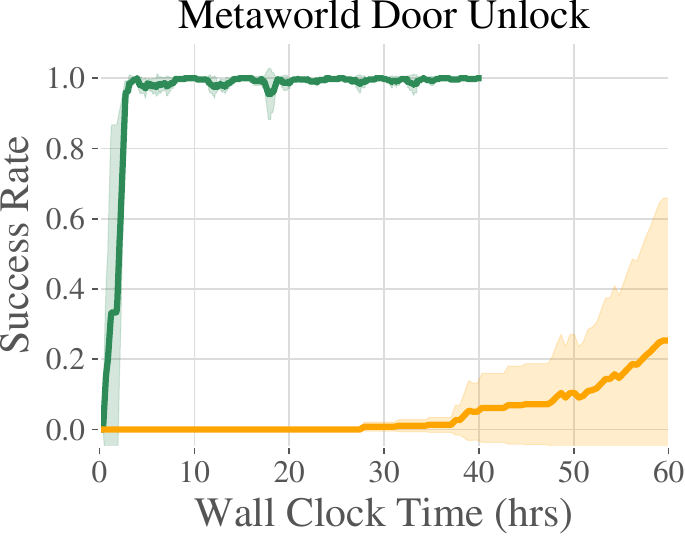}
\includegraphics[width=.24\textwidth]{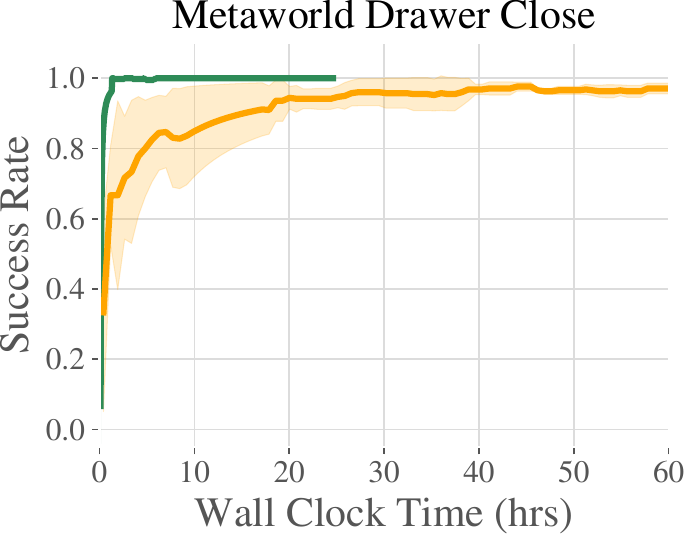} &
\includegraphics[width=.24\textwidth]{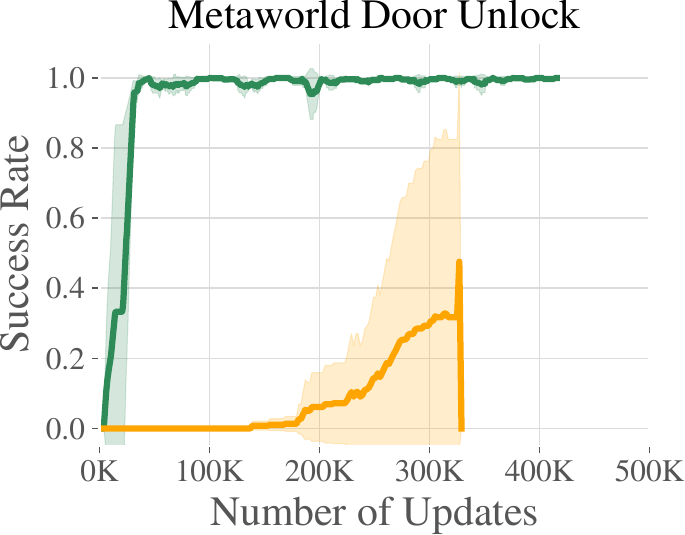}
\includegraphics[width=.24\textwidth]{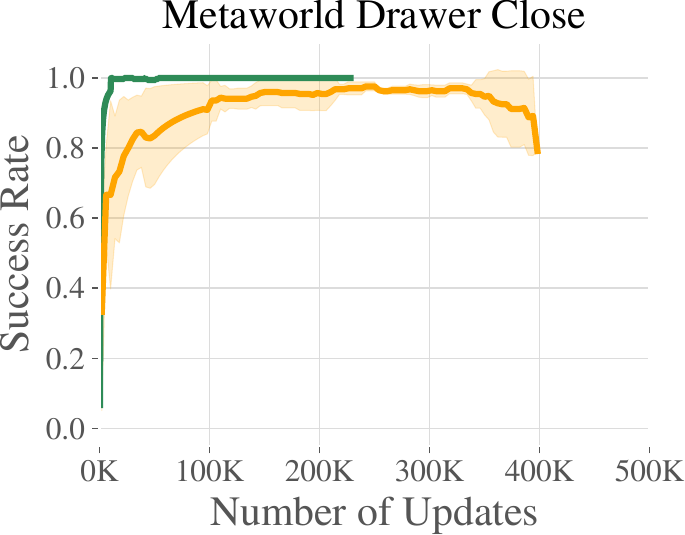}
\vspace{.2cm}
\\ 
\includegraphics[width=.24\textwidth]{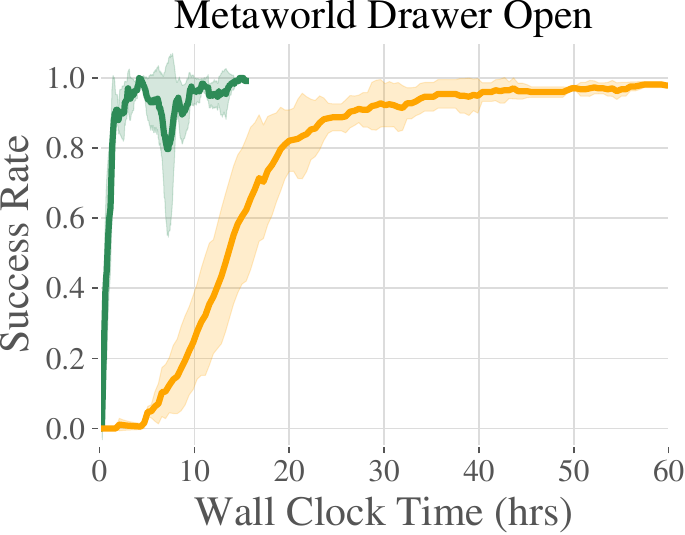}
\includegraphics[width=.24\textwidth]{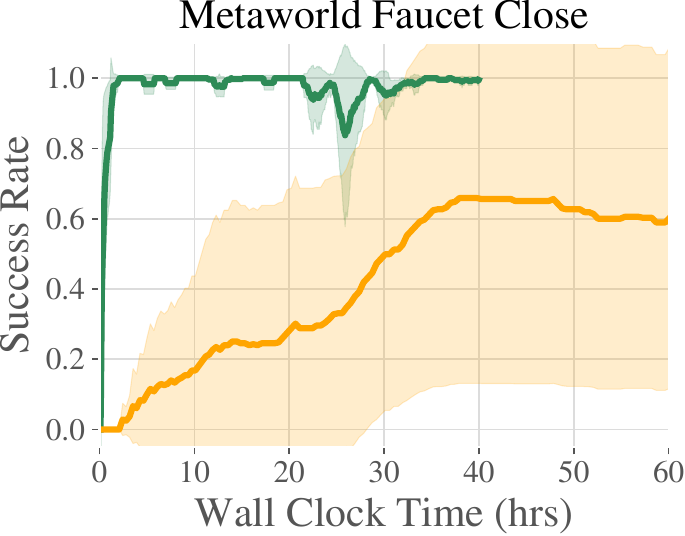} &
\includegraphics[width=.24\textwidth]{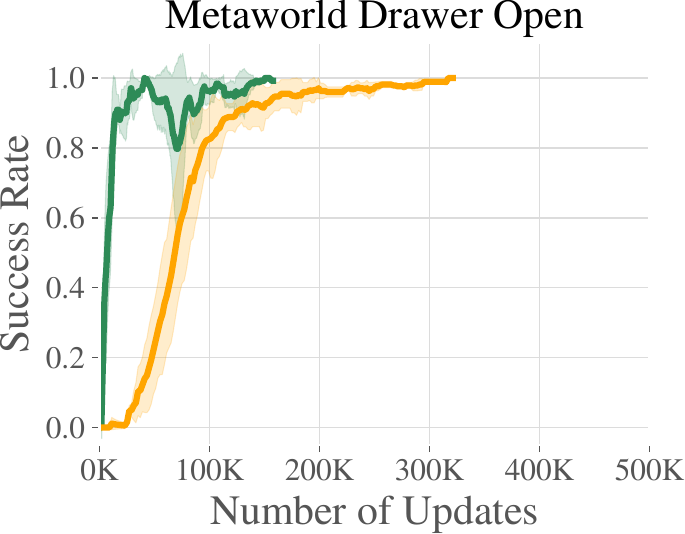}
\includegraphics[width=.24\textwidth]{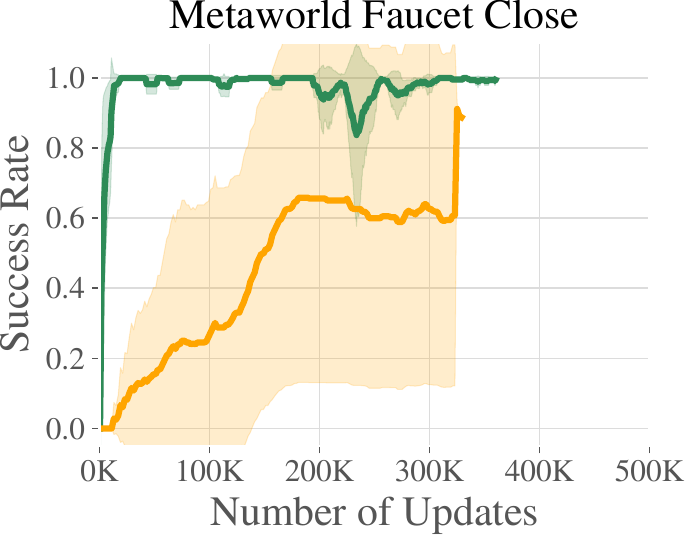}
\vspace{.2cm}
\\
\includegraphics[width=.24\textwidth]{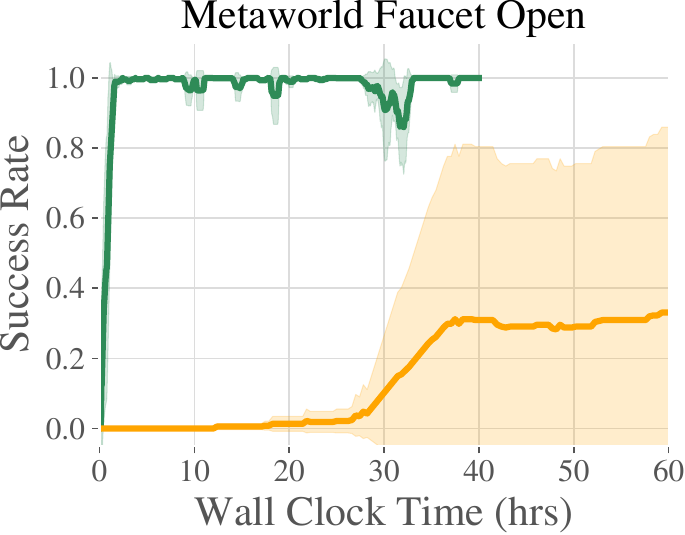}
\includegraphics[width=.24\textwidth]{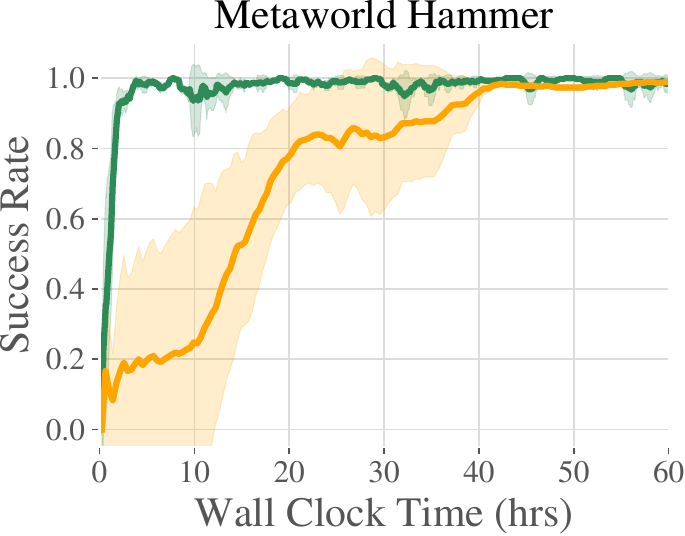} &
\includegraphics[width=.24\textwidth]{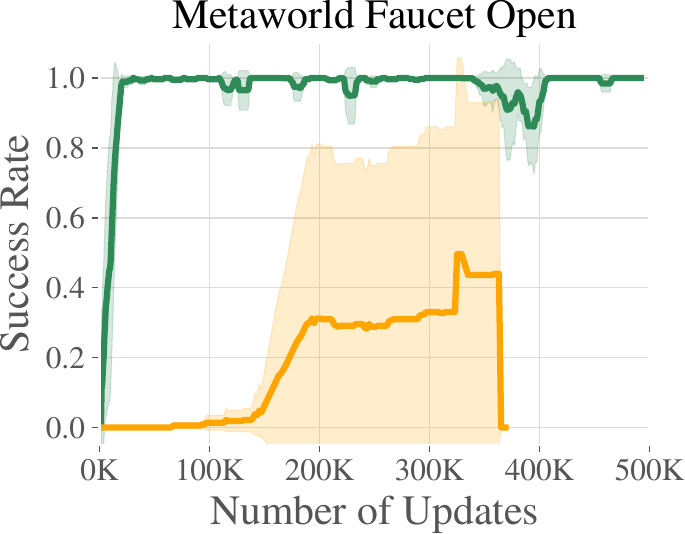}
\includegraphics[width=.24\textwidth]{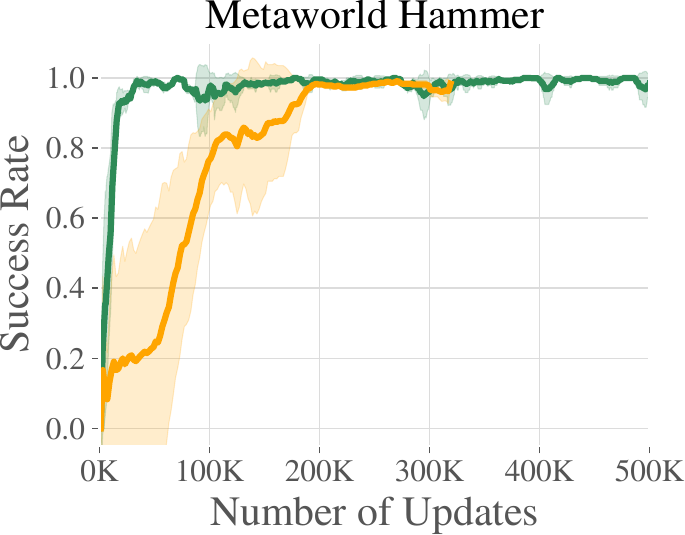}
\vspace{.2cm}
\\
\includegraphics[width=.24\textwidth]{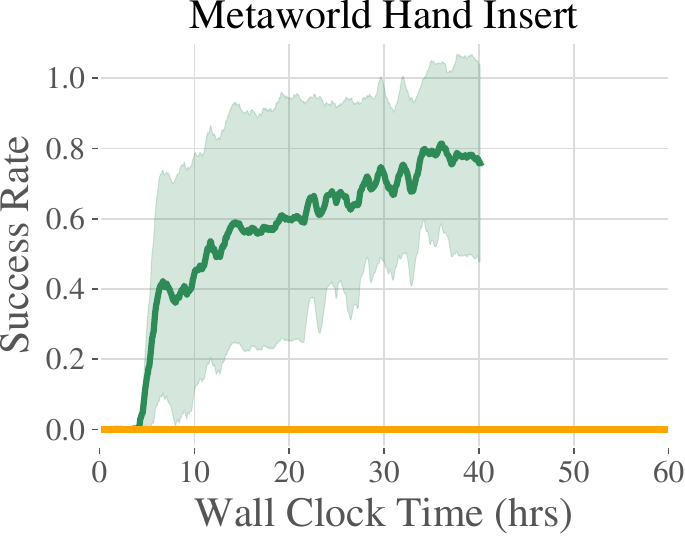}
\includegraphics[width=.24\textwidth]{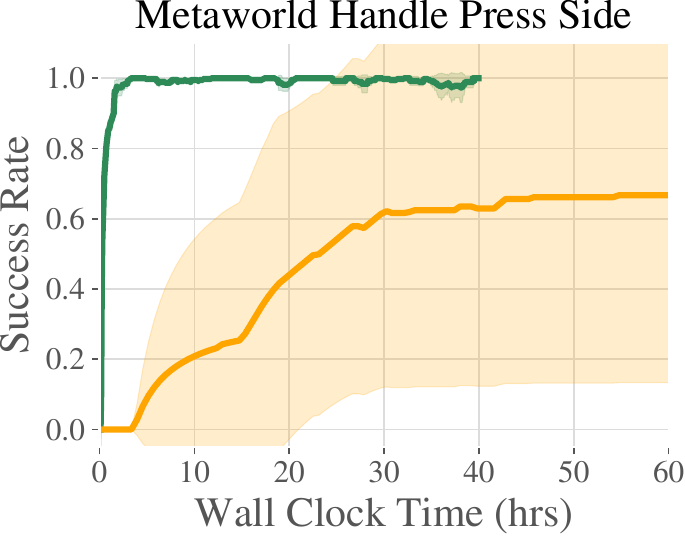} &
\includegraphics[width=.24\textwidth]{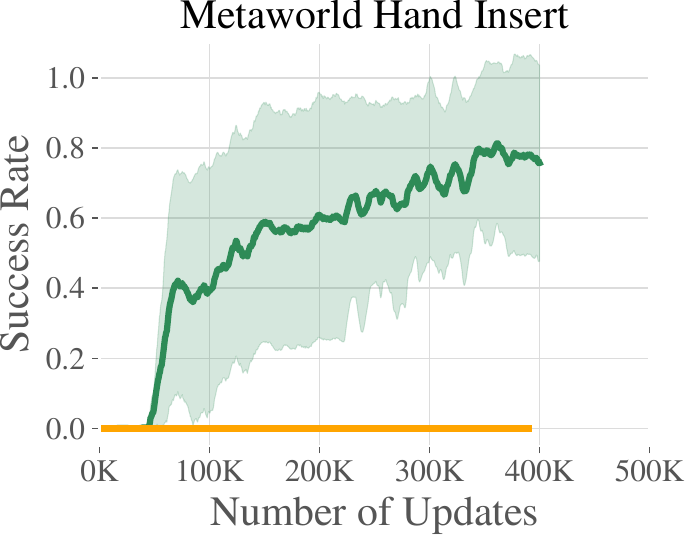}
\includegraphics[width=.24\textwidth]{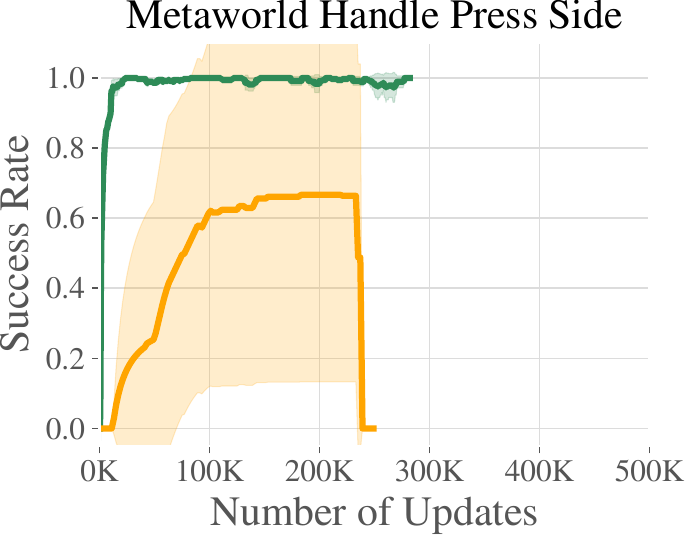} 
\vspace{.2cm}
\\
\includegraphics[width=.24\textwidth]{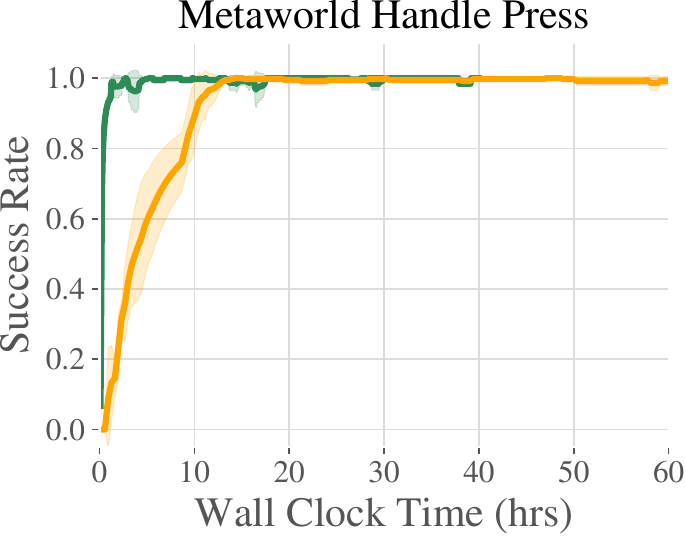}
\includegraphics[width=.24\textwidth]{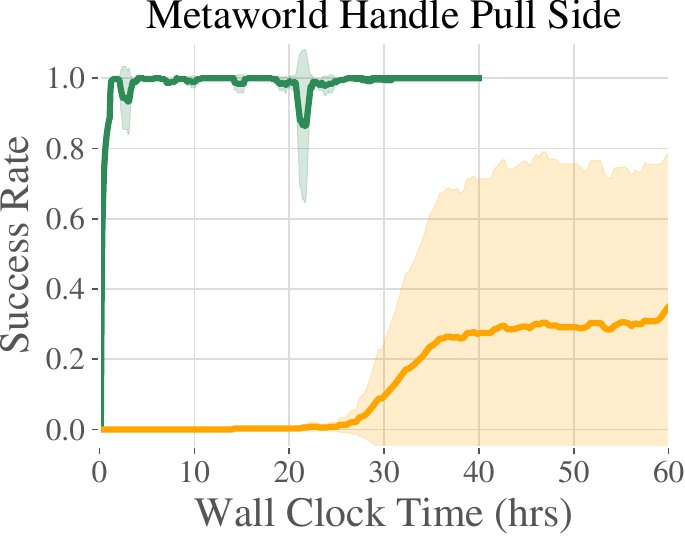} &
\includegraphics[width=.24\textwidth]{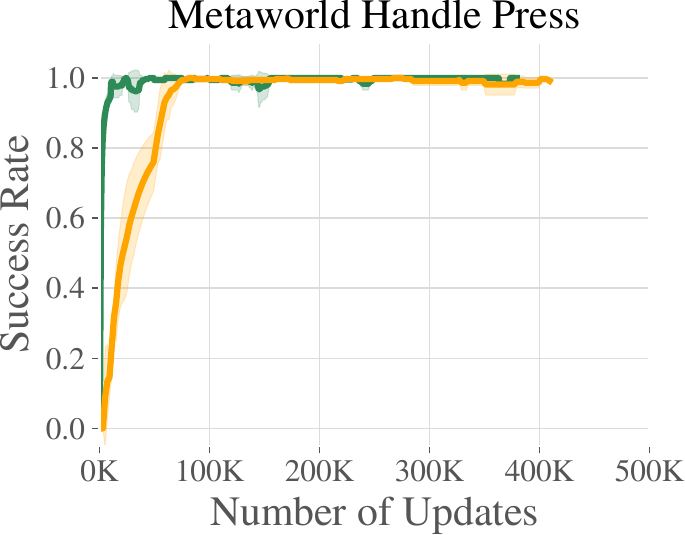}
\includegraphics[width=.24\textwidth]{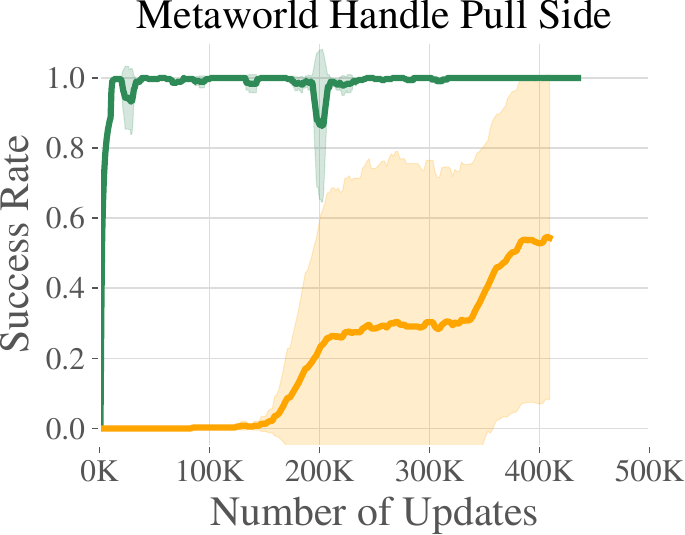}
\vspace{.2cm}
\\
\includegraphics[width=.24\textwidth]{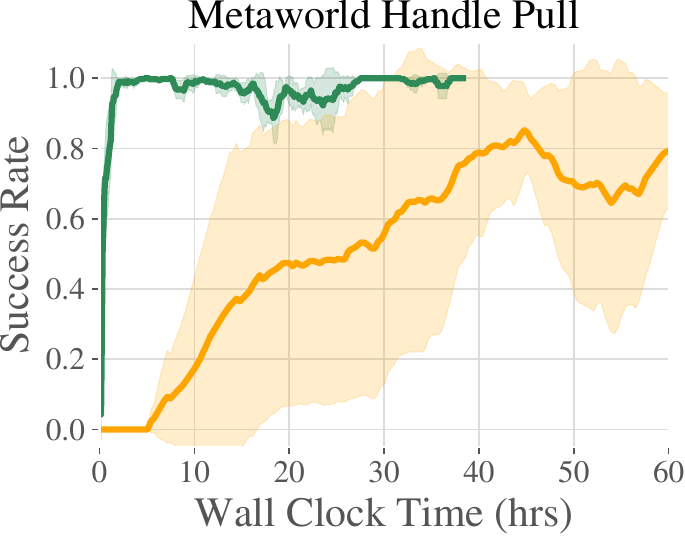}
\includegraphics[width=.24\textwidth]{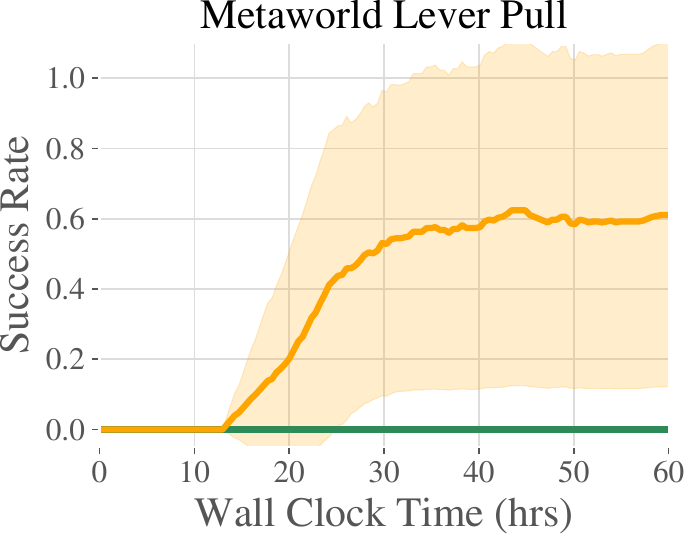} &
\includegraphics[width=.24\textwidth]{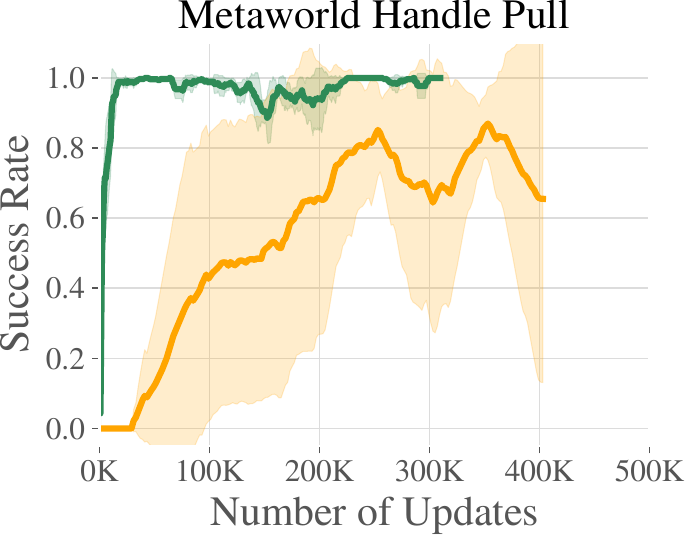}
\includegraphics[width=.24\textwidth]{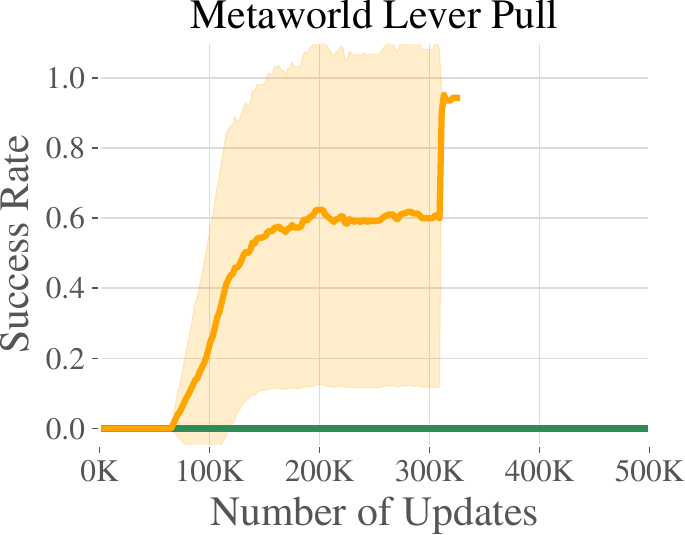}
\vspace{.2cm}
\\
\includegraphics[width=.24\textwidth]{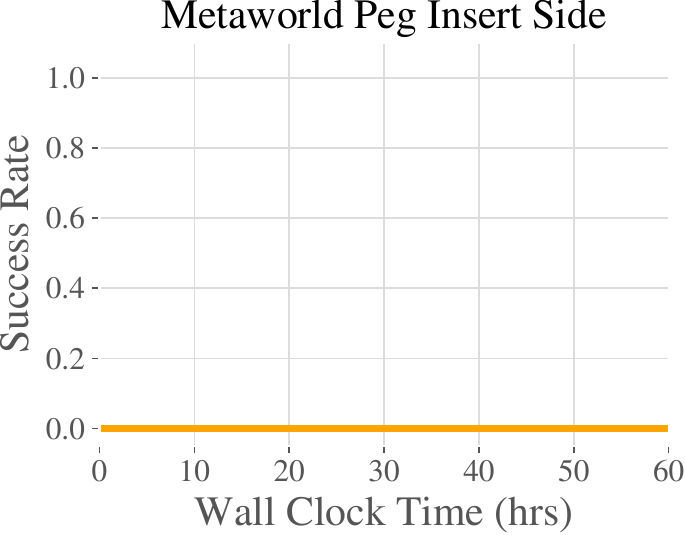}
\includegraphics[width=.24\textwidth]{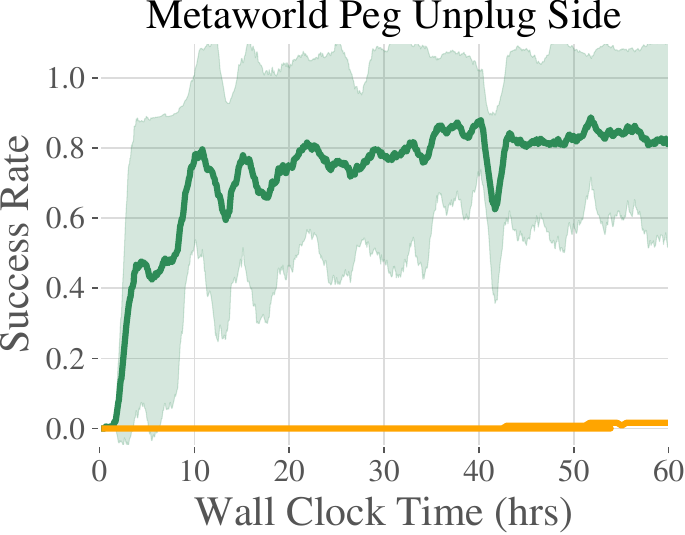} &
\includegraphics[width=.24\textwidth]{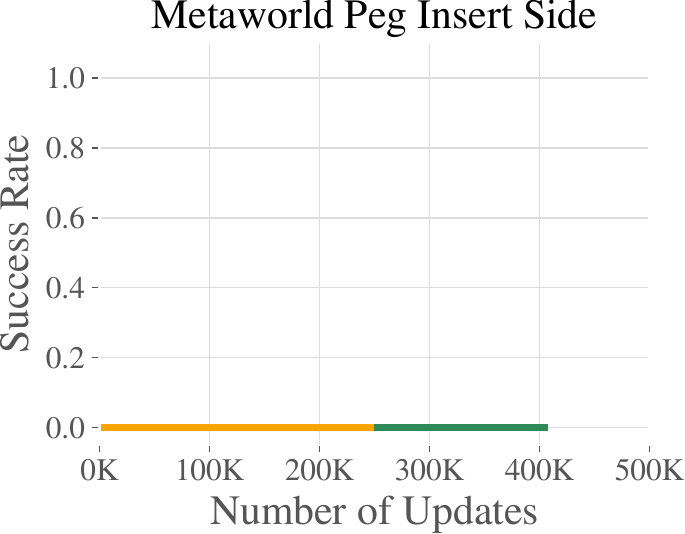}
\includegraphics[width=.24\textwidth]{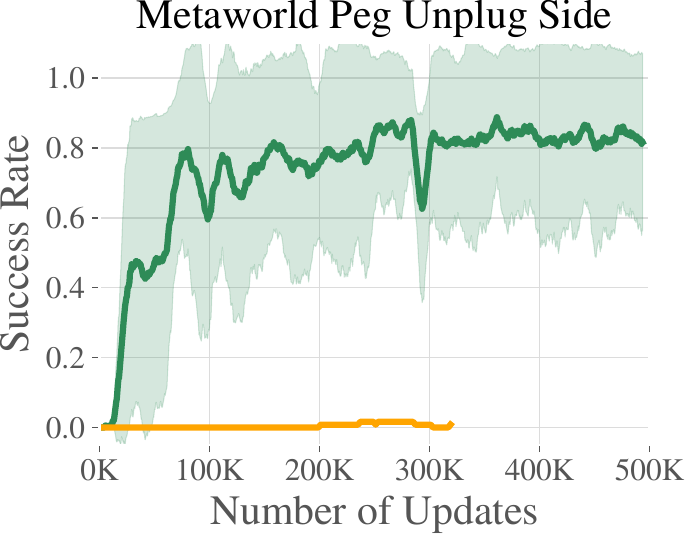}
\vspace{.2cm}
\\
\includegraphics[width=.24\textwidth]{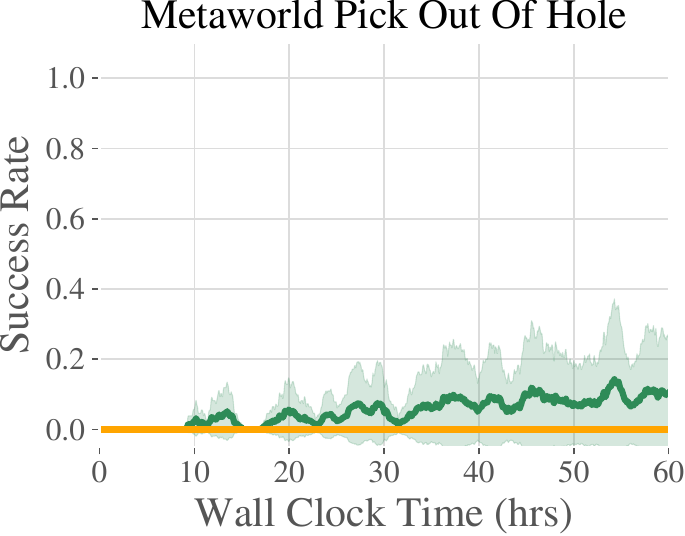}
\includegraphics[width=.24\textwidth]{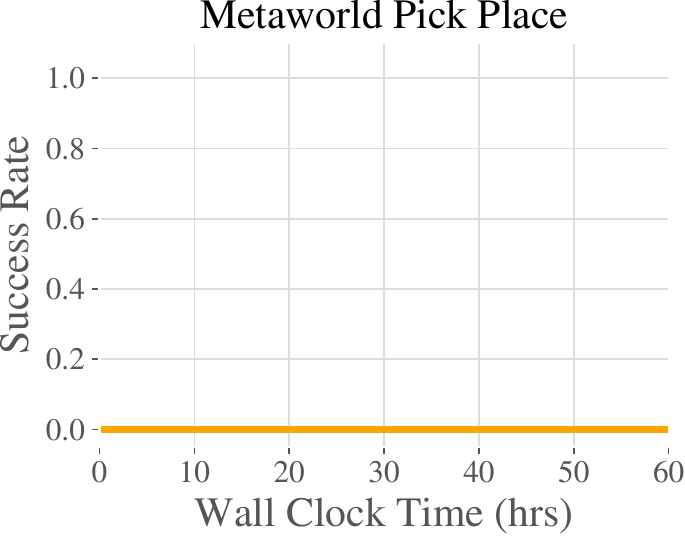} &
\includegraphics[width=.24\textwidth]{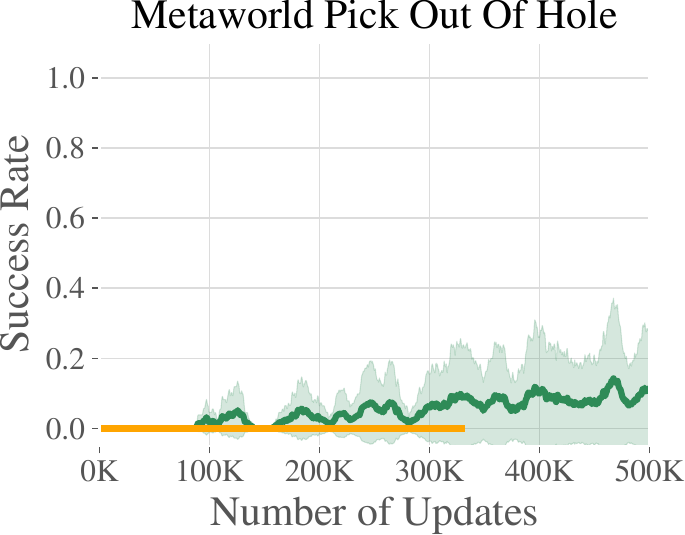}
\includegraphics[width=.24\textwidth]{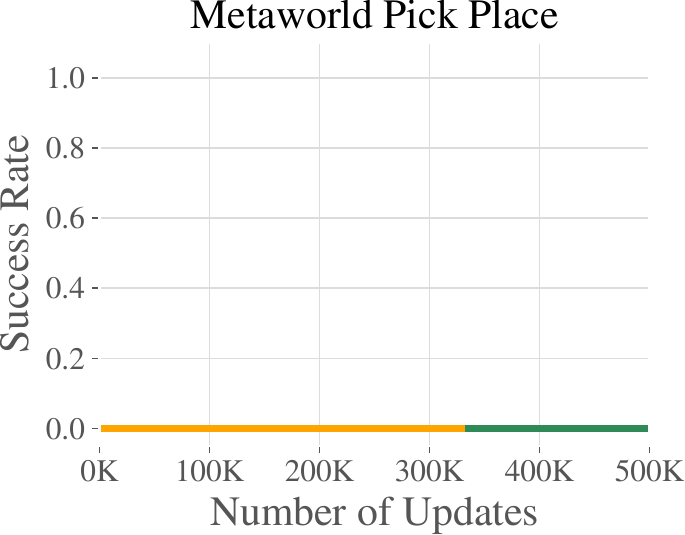}
    \end{tabular}

\end{figure}

\begin{figure}\ContinuedFloat
    \centering
    \begin{tabular}{cc}
\includegraphics[width=.24\textwidth]{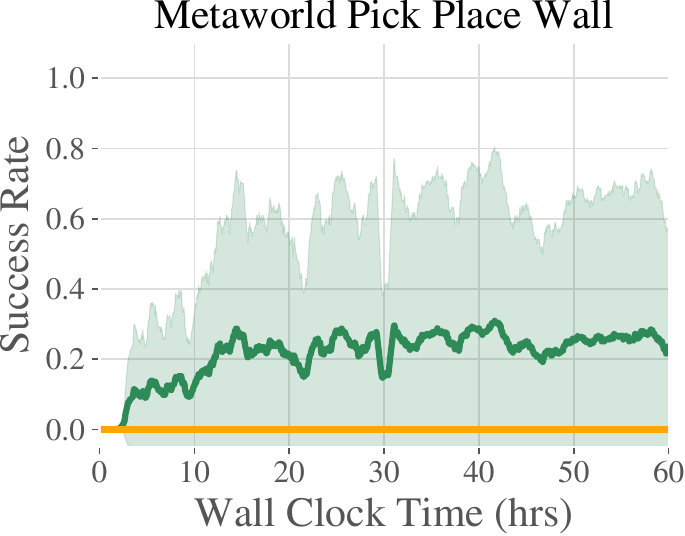}
\includegraphics[width=.24\textwidth]{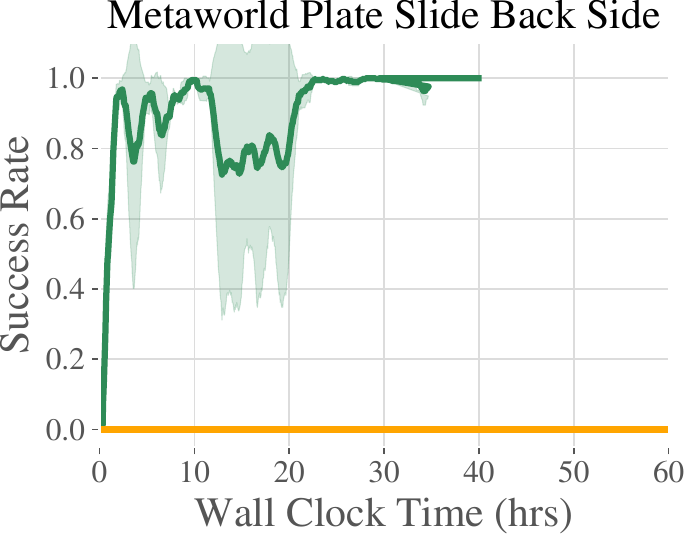} &
\includegraphics[width=.24\textwidth]{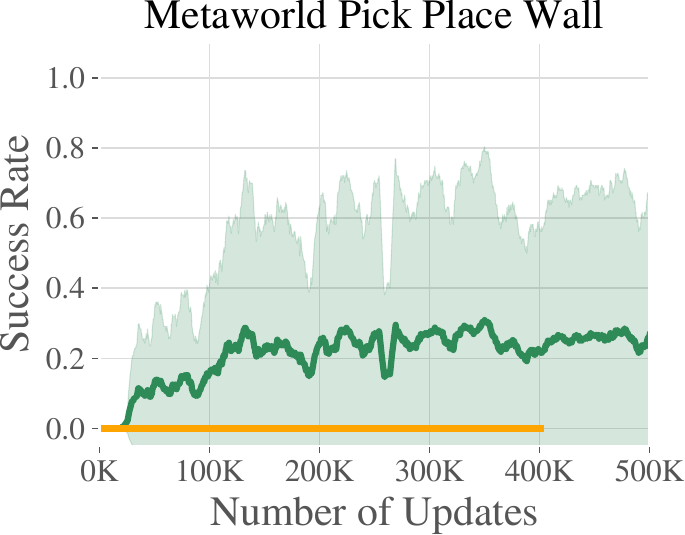}
\includegraphics[width=.24\textwidth]{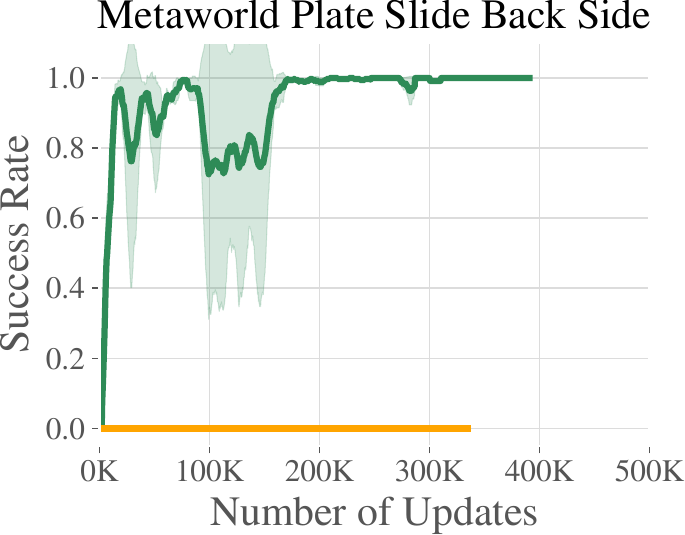}
\vspace{.2cm}
\\
\includegraphics[width=.24\textwidth]{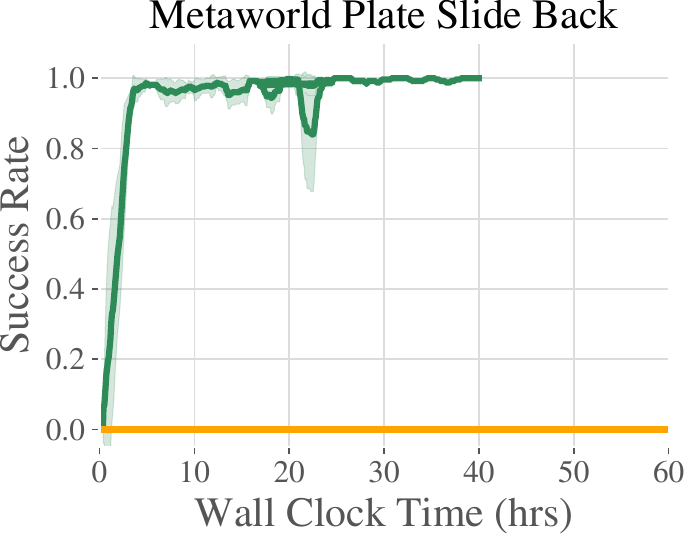}
\includegraphics[width=.24\textwidth]{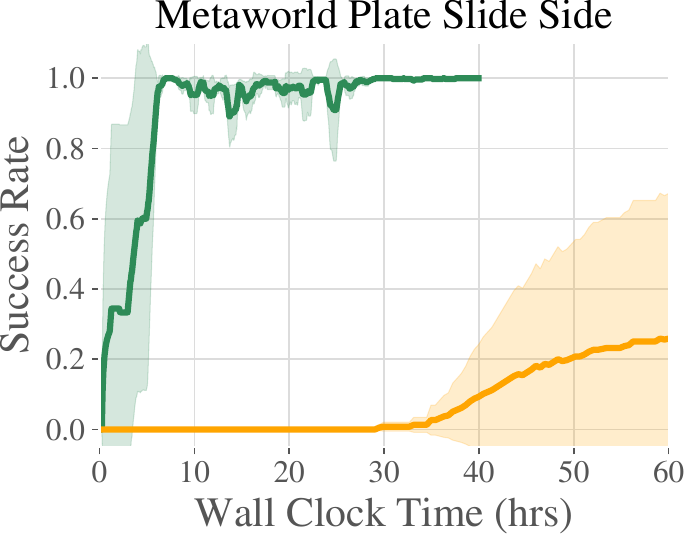} &
\includegraphics[width=.24\textwidth]{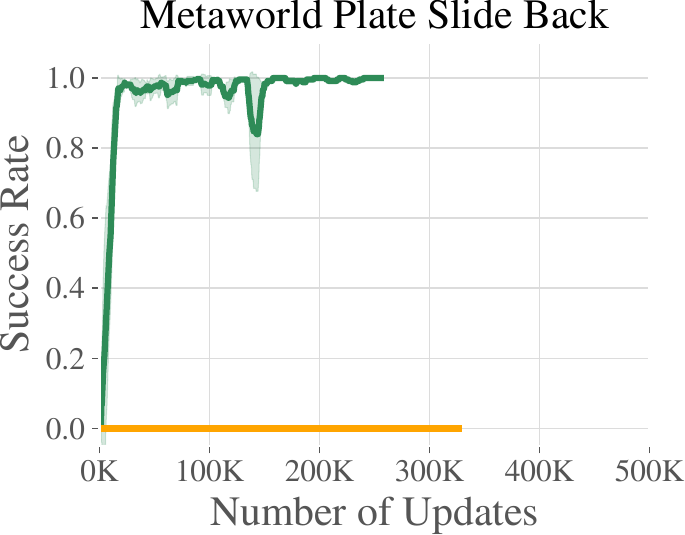}
\includegraphics[width=.24\textwidth]{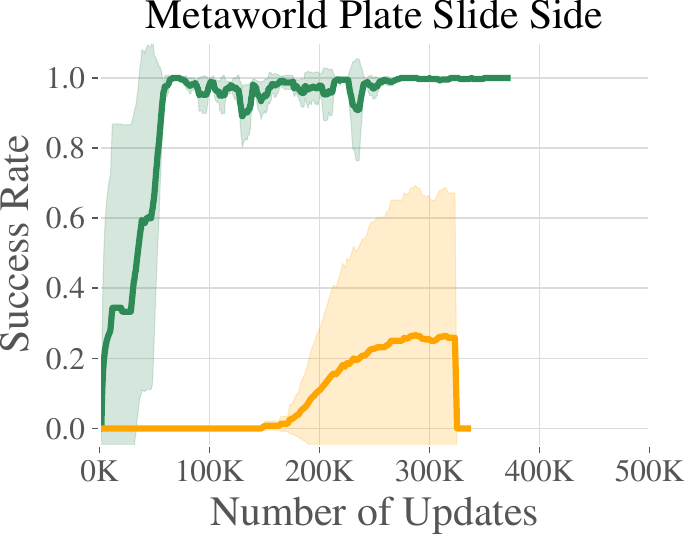}
\vspace{.2cm}
\\
\includegraphics[width=.24\textwidth]{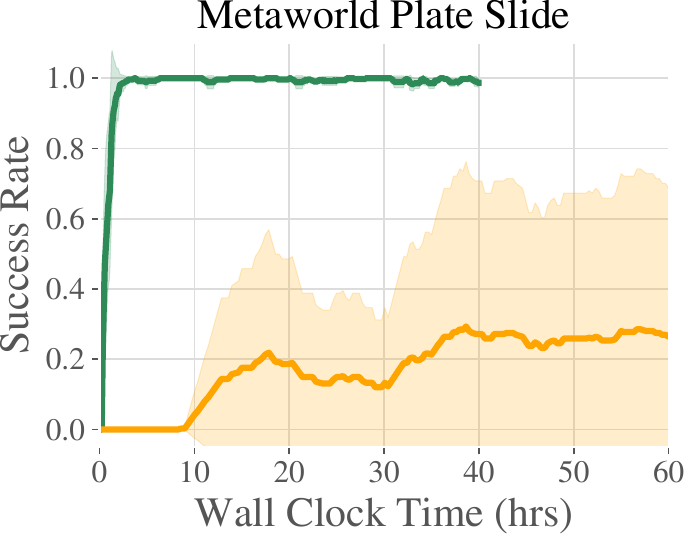}
\includegraphics[width=.24\textwidth]{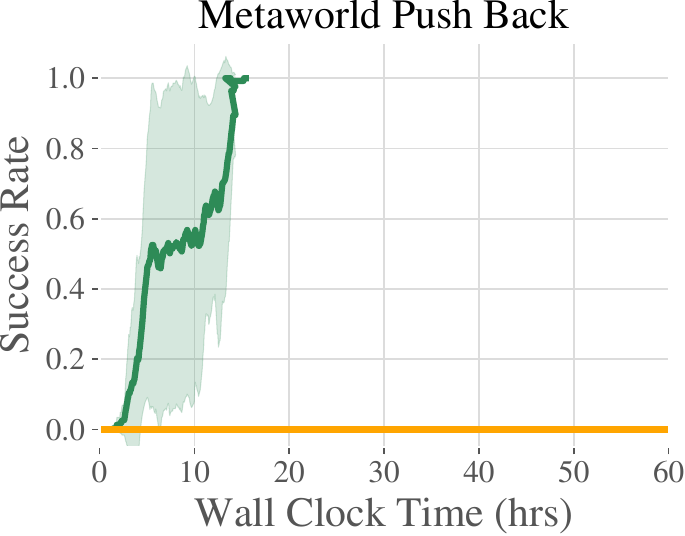} &
\includegraphics[width=.24\textwidth]{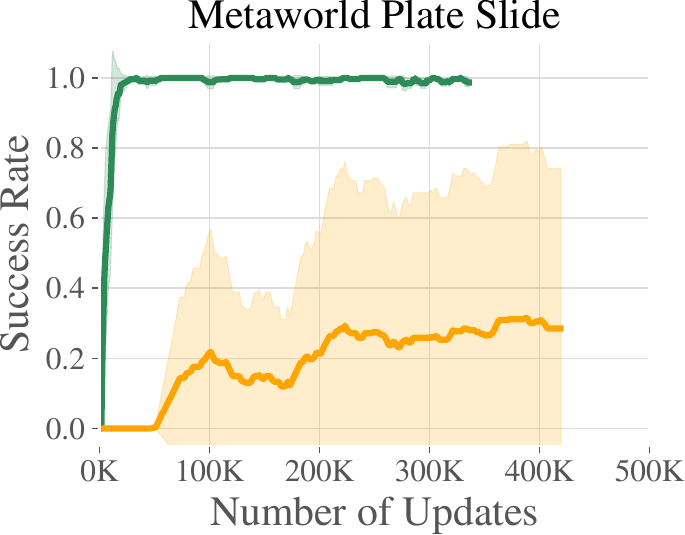}
\includegraphics[width=.24\textwidth]{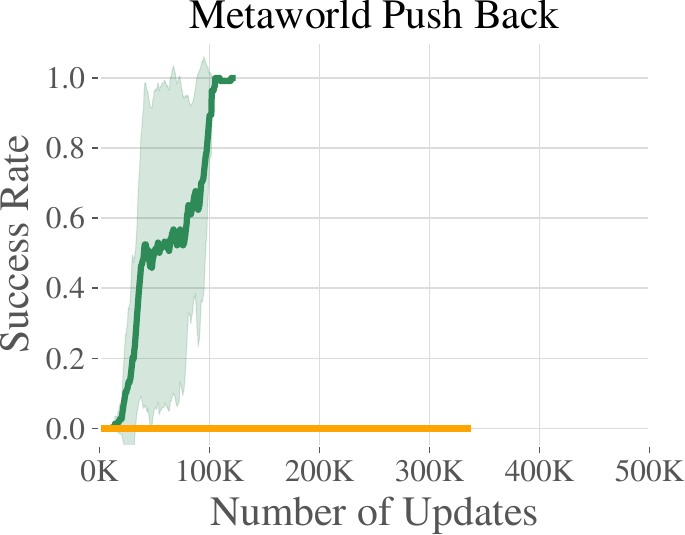} 
\vspace{.2cm}
\\
\includegraphics[width=.24\textwidth]{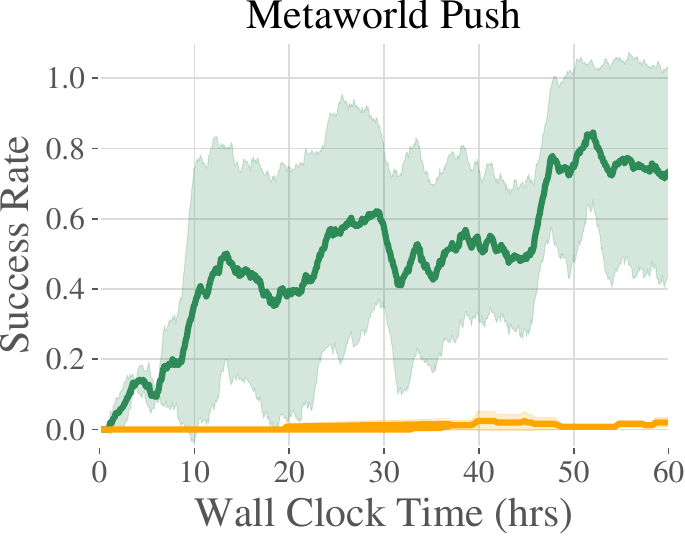}
\includegraphics[width=.24\textwidth]{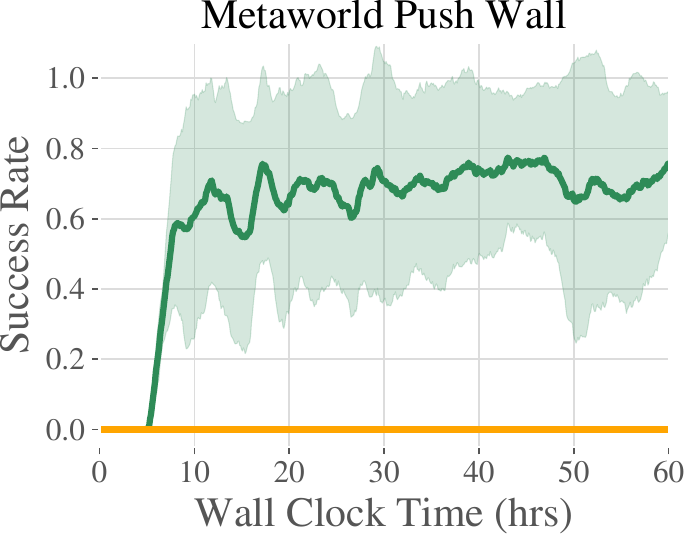} &
\includegraphics[width=.24\textwidth]{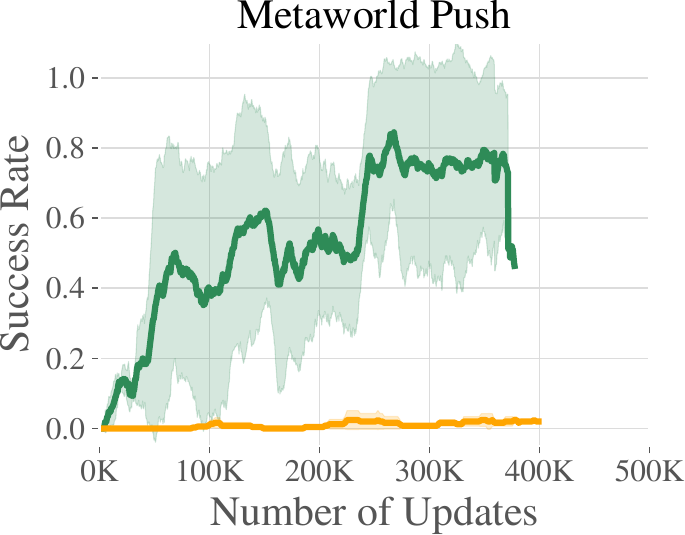}
\includegraphics[width=.24\textwidth]{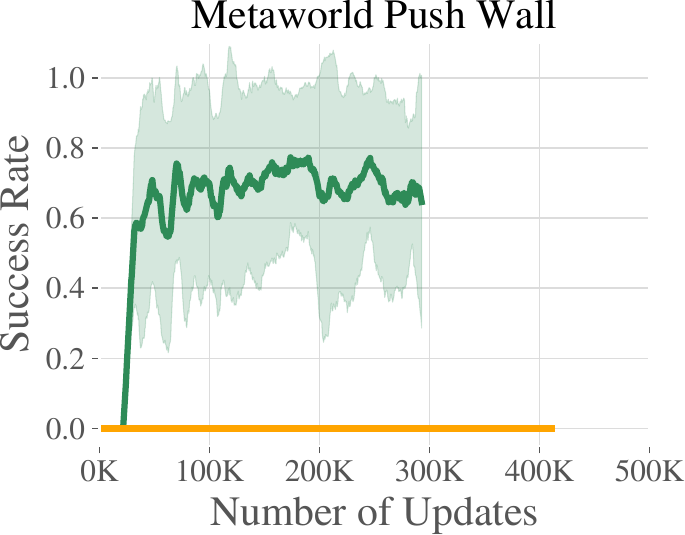}
\vspace{.2cm}
\\
\includegraphics[width=.24\textwidth]{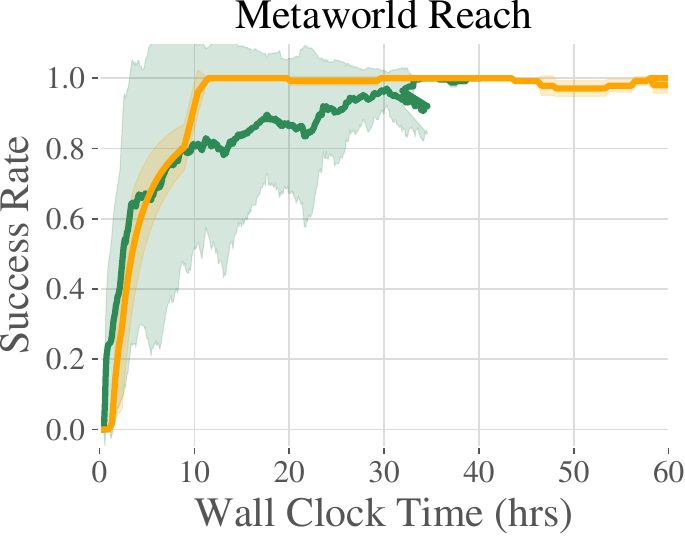}
\includegraphics[width=.24\textwidth]{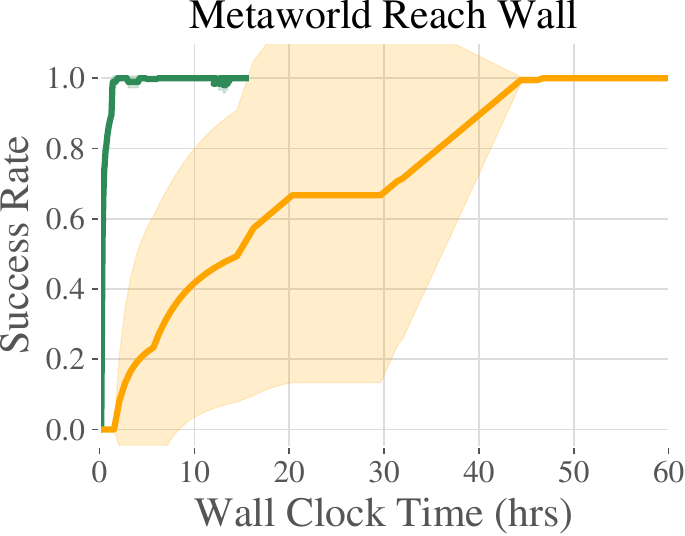} &
\includegraphics[width=.24\textwidth]{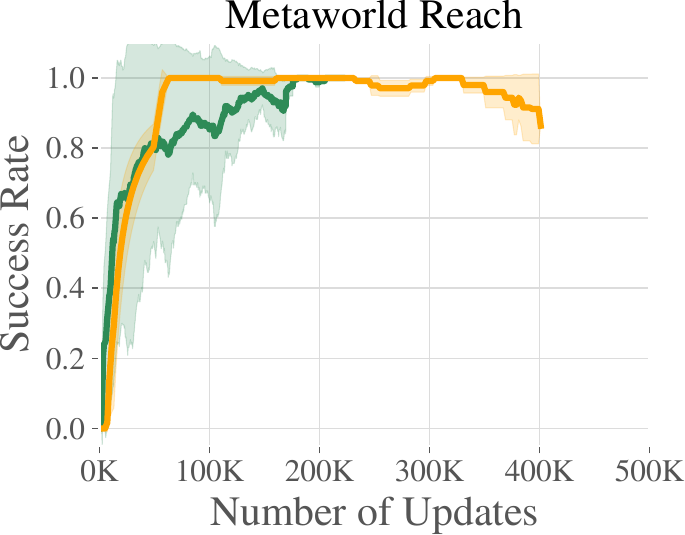}
\includegraphics[width=.24\textwidth]{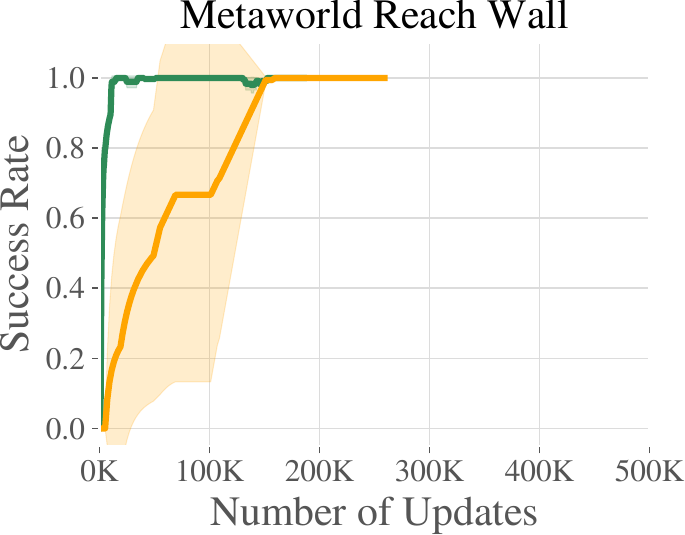}
\vspace{.2cm}
\\
\includegraphics[width=.24\textwidth]{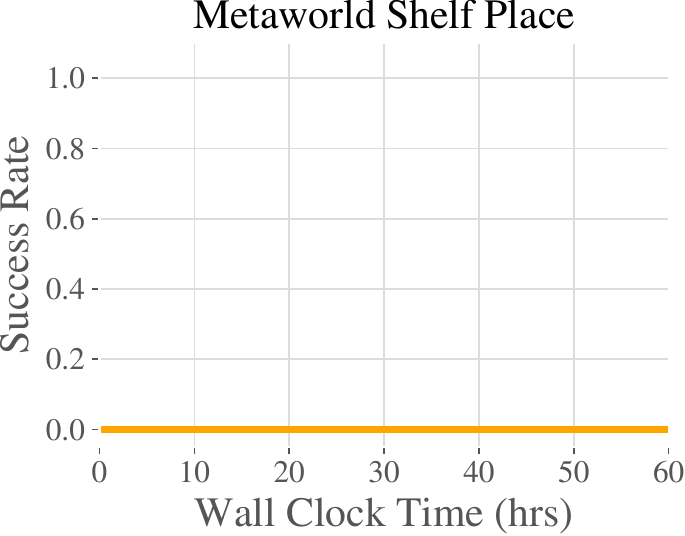}
\includegraphics[width=.24\textwidth]{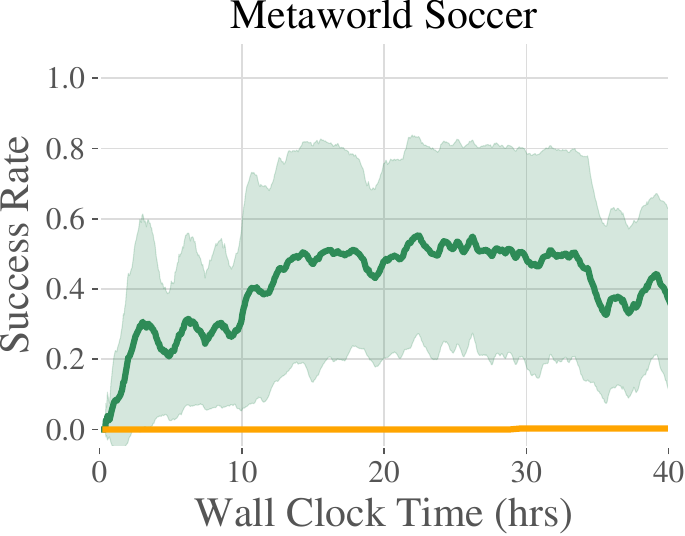} &
\includegraphics[width=.24\textwidth]{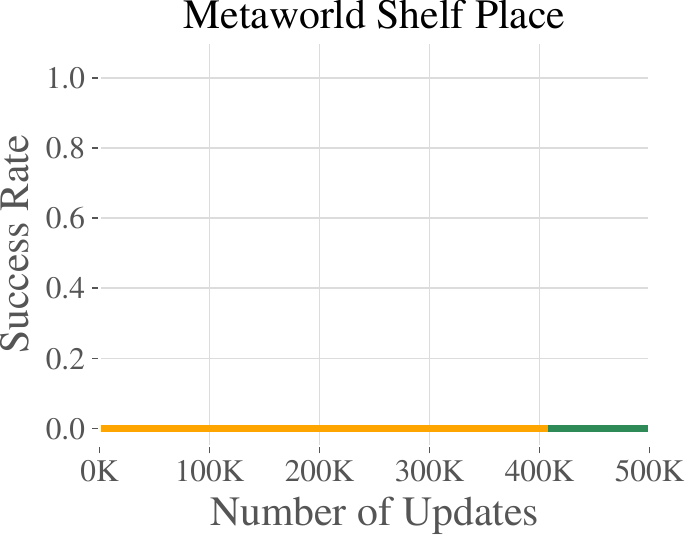}
\includegraphics[width=.24\textwidth]{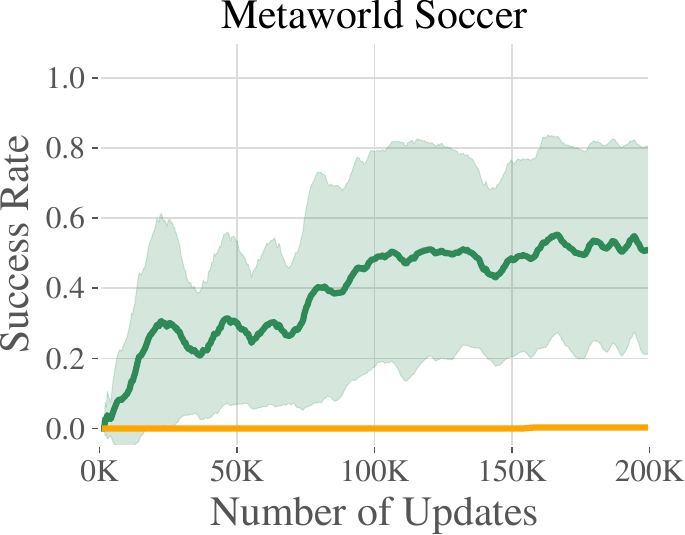} 
\vspace{.2cm}
\\
\includegraphics[width=.24\textwidth]{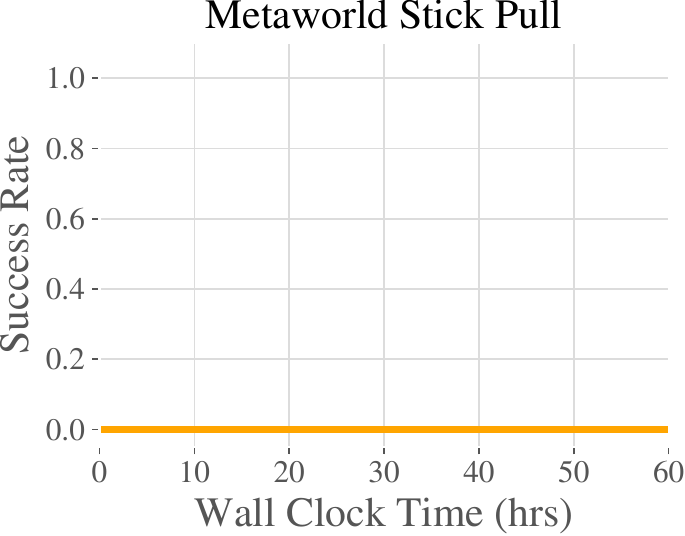}
\includegraphics[width=.24\textwidth]{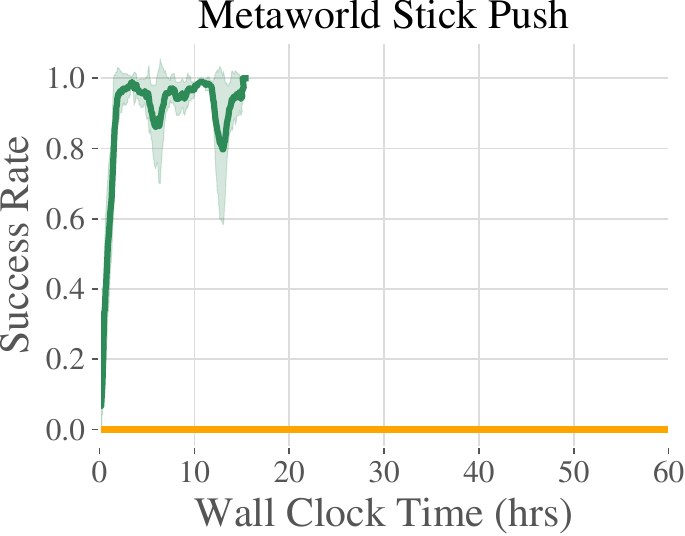} & 
\includegraphics[width=.24\textwidth]{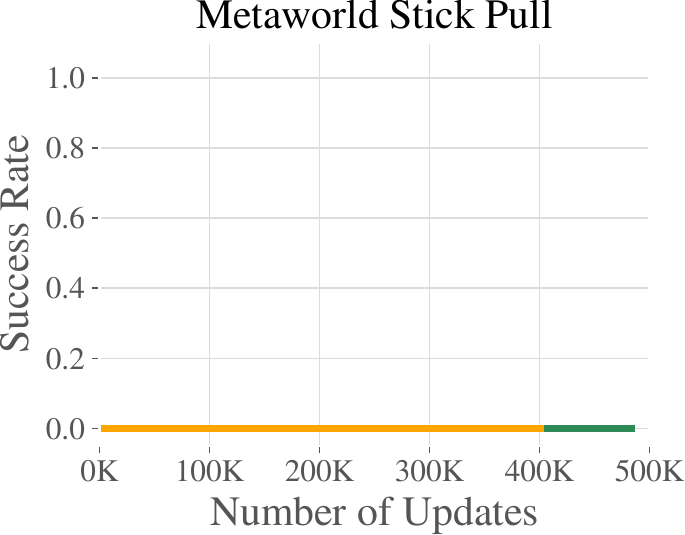}
\includegraphics[width=.24\textwidth]{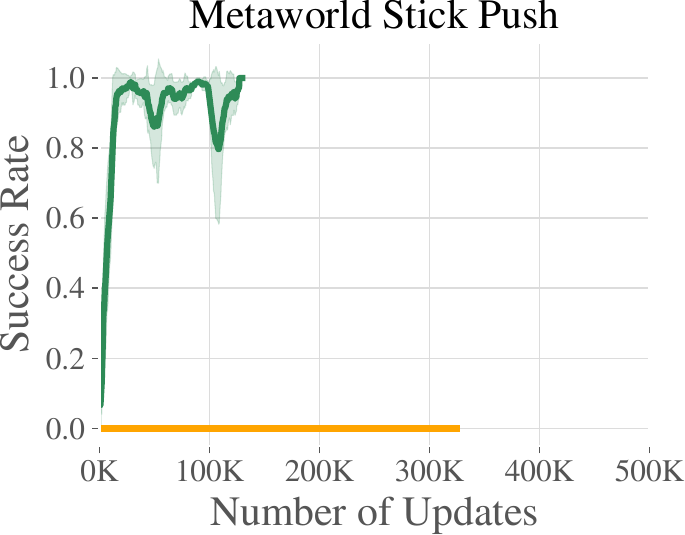}
    \end{tabular}

\end{figure}

\begin{figure}\ContinuedFloat
\centering
    \begin{tabular}{cc}
    \includegraphics[width=.24\textwidth]{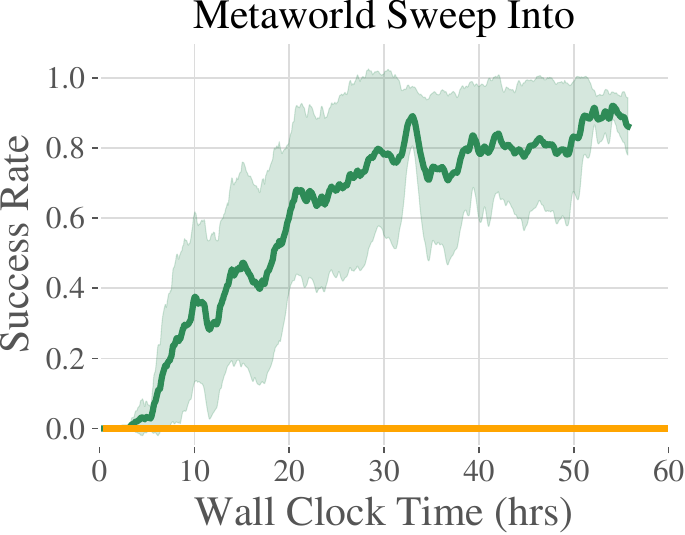}
    \includegraphics[width=.24\textwidth]{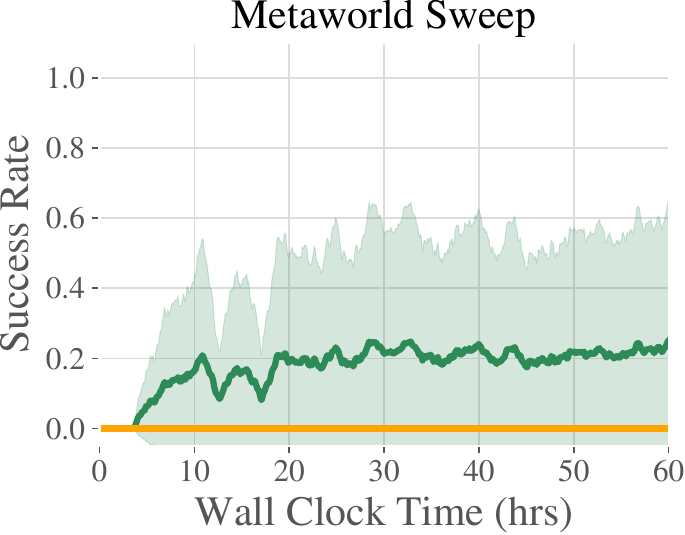} &
    \includegraphics[width=.24\textwidth]{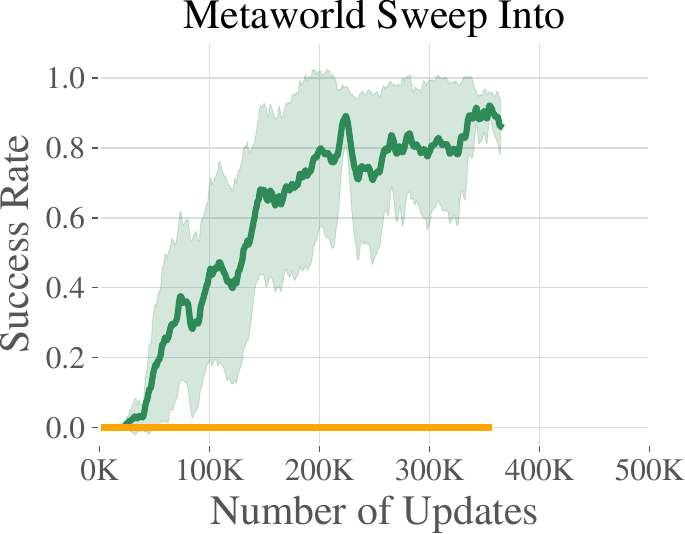}
    \includegraphics[width=.24\textwidth]{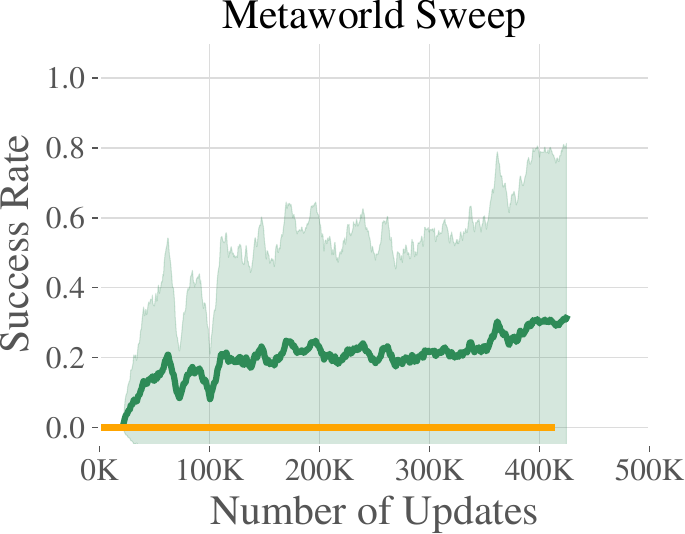}
    \vspace{.2cm}
    \\
    \includegraphics[width=.24\textwidth]{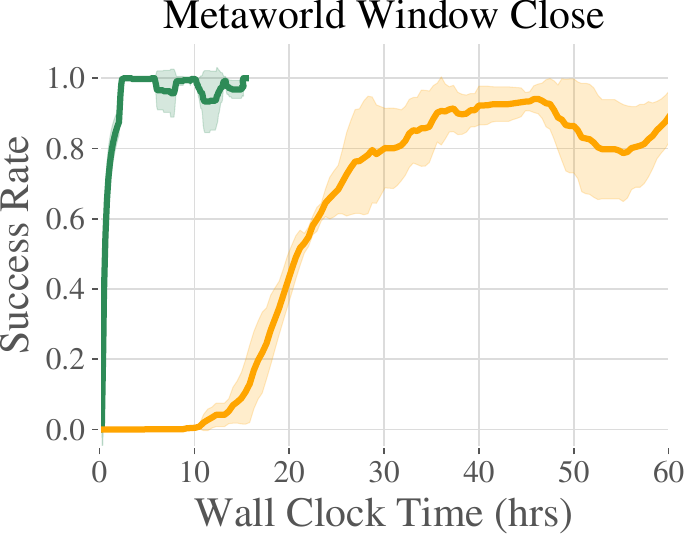}
    \includegraphics[width=.24\textwidth]{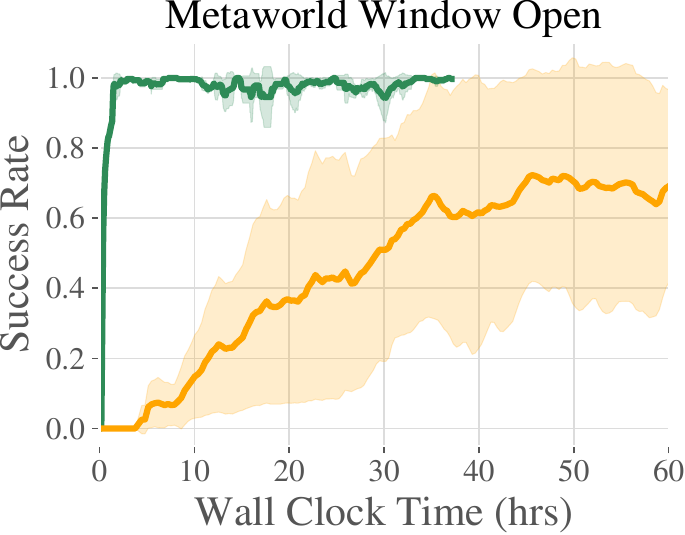} & \includegraphics[width=.24\textwidth]{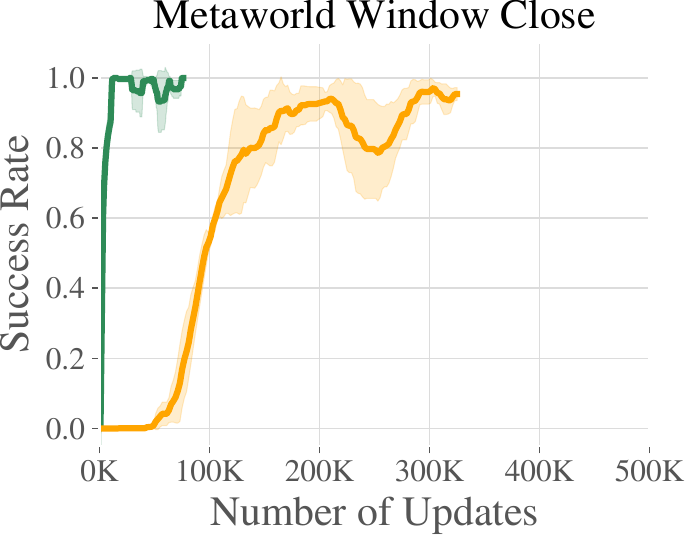}
    \includegraphics[width=.24\textwidth]{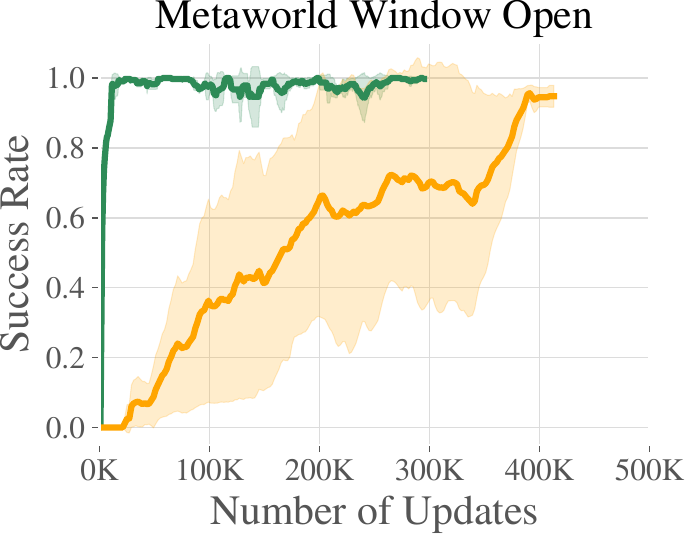}
\end{tabular}
\includegraphics[width=.3\textwidth]{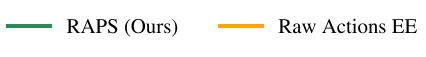}
\caption{Comparison of RAPS\xspace against raw actions across all 50 Metaworld tasks from sparse rewards. RAPS\xspace is able to outright solve or make progress on up to \textbf{43 tasks} while Raw Actions struggles to make progress on most environments.}
\label{fig:mt50-results}
\end{figure}

\chapter{Chapter 3 Appendix}

\section{Table of Contents}
\begin{itemize}
    \item \textbf{Additional Experimental Results} (Appendix~\ref{app:additional exps}): Additional ablations and analyses as well as learning curves for single-stage tasks and Meta-World.
    \item \textbf{PSL\xspace Implementation Details} (Appendix~\ref{app:our impl details}): Full details on how PSL\xspace is implemented, specifically the Sequencing Module.
    \item \textbf{Baseline Implementation Details} (Appendix~\ref{app:baseline impl details}): Full details regarding baseline implements (E2E, RAPS, MoPA-RL, TAMP, SayCan)
    \item \textbf{Tasks} (Appendix~\ref{app:tasks}): Visualizations of each task as well as descriptions of each environment suite.
    \item \textbf{LLM Prompts and Plans} (Appendix~\ref{app:llm prompts}): Prompts that we use for our method as well as generated plans by the LLM.
\end{itemize}
\clearpage

\section{Additional Experimental Results}
\label{app:additional exps}
We perform additional analyses of PSL\xspace in this section.

\begin{figure}[h]
    \centering
    \includegraphics[width=.4\linewidth]{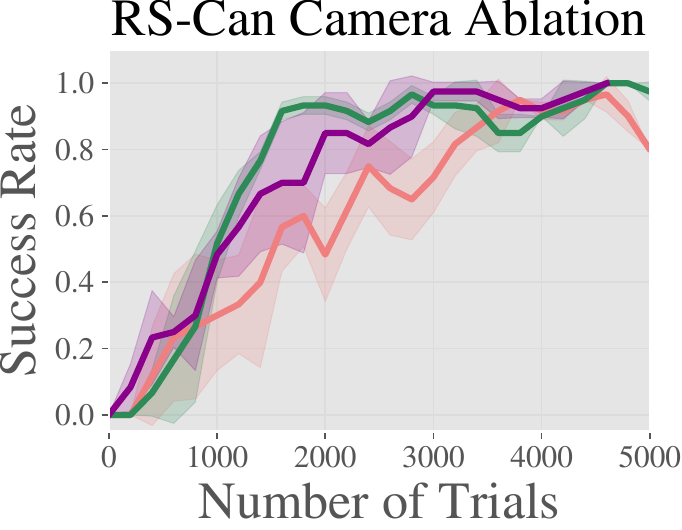}
    \vspace{5pt}\\
    \includegraphics[width=.5\linewidth]
    {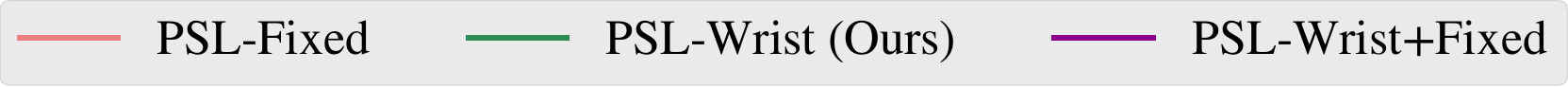}
    \vspace{5pt}
    \caption{\textbf{Camera View Learning Performance Ablation.} wrist camera views clearly accelerate learning performance, converging to near 100\% performance \textbf{4x} faster than using fixed-view and \textbf{3x} faster than using  wrist+fixed-view observations.} 
    \label{fig:camera-perf-abl}
\end{figure}

\textbf{Effect of camera view on policy learning performance:} As discussed in Sec.~\ref{sec:method}, for PSL\xspace we use local observations to improve learning performance and generalization to new poses. We validate this claim on the Robosuite Can task, in which we hypothesize that the local wrist camera view will accelerate policy learning performance. This is because the image of the can will be independent of the can's position in general since the Sequencing Module will initialize the RL agent as close to the can as possible. As observed in Fig.~\ref{fig:camera-perf-abl}, this is indeed the case - PSL\xspace learns \textbf{4x} faster than using a fixed view camera in terms of the number of trials. We additionally test if combining wrist and fixed view inputs (a common paradigm in robot learning) can alleviate the issue, however PSL\xspace with wrist cam is still \textbf{3x} faster at solving the task.

\textbf{Effect of camera view on chaining pre-trained policies:} In this ablation, we illustrate another important effect of using local views, such as wrist cameras: ease of chaining pre-trained policies. Since we leverage motion planning to sequence between policy executions, chaining pre-trained policies is relatively straightforward: simply execute the Sequencing Module to reach the first region of interest, execute the first pre-trained policy till its stage termination condition is triggered, then call the Sequencing Module on the next region, and so on. However, to do so, it is also crucial that the observations do not change significantly, so that the inputs to the pre-trained policies are not out of distribution (OOD). If we use a fixed, global view of the scene, the overall scene will change as multiple policies are executed, resulting in future policy executions failing due to OOD inputs. In Table~\ref{table:chain_policy_abl}, we observe this exact phenomenon, in which any version of PSL\xspace that is provided a fixed-view input fails to chain pre-trained policies effectively, while PSL\xspace with local (wrist) views only is able to chain pre-trained policies on every task, up to 5 stages.

\begin{figure}[ht]
    \centering
    \includegraphics[width=.24\textwidth]{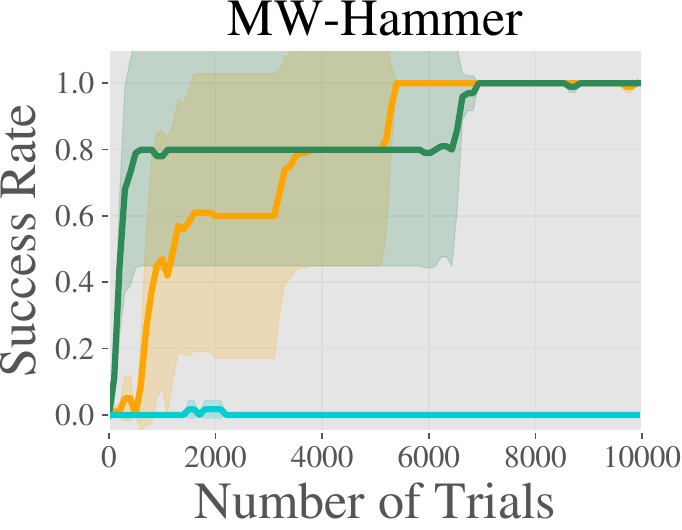}
    \includegraphics[width=.24\textwidth]{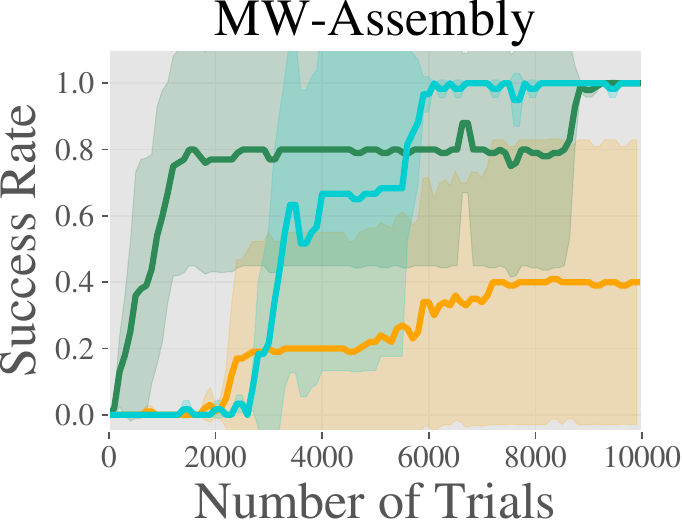}
    \includegraphics[width=.24\textwidth]{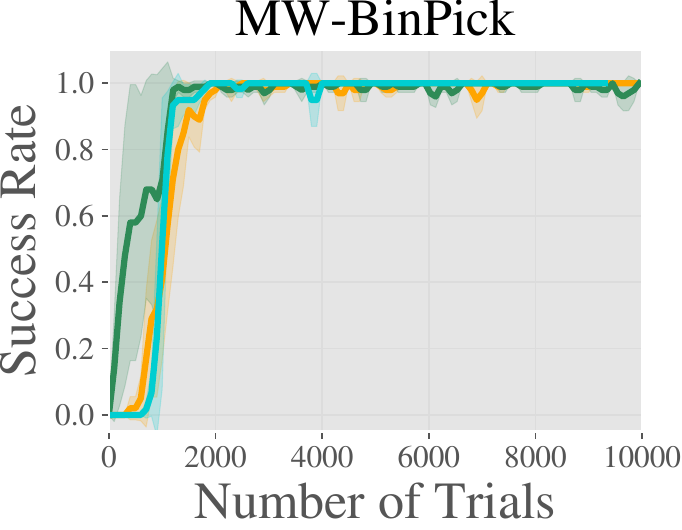}
    \\
    \vspace{2pt}
    \includegraphics[width=.4\textwidth]{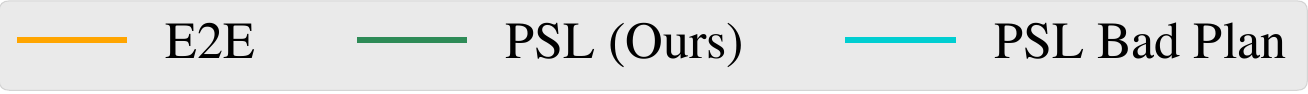}
    \caption{\small \textbf{PSL\xspace bad plans.} We plot task success rate as a function of the number of trials. Even when given the wrong high-level plan, PSL\xspace is able to learn to solve the task, albeit at a slower rate.}
    \label{fig:bad_plan}
    \vspace{-7pt}
\end{figure}

\textbf{Effect of incorrect plans on training policies using PSL\xspace:}
If the Plan Module or Sequence Module fail catastrophically (incorrect plan or moving to the wrong region in space), there is currently no concrete mechanism for the Learning Module to adapt. However, we run an experiment in which we train the agent using PSL\xspace using an incorrect high-level plan on two stage tasks (MW-Assembly, MW-Bin-Picking, MW-Hammer) and find that in some cases, the agent can still learn to solve the task, achieving performance close to E2E (Fig.~\ref{fig:bad_plan}. Intuitively, this is possible because in PSL\xspace, the high-level plan is not expressed as a hard constraint, but rather as a series of regions for the agent to visit and a set of exit conditions for those regions. In the end, however, only the task reward is used to train the RL policy so if the plan is wrong, the Learn Module must learn to solve the entire task end-to-end from sub-optimal initial states.

\begin{figure}[ht]
    \centering
    \includegraphics[width=.24\textwidth]{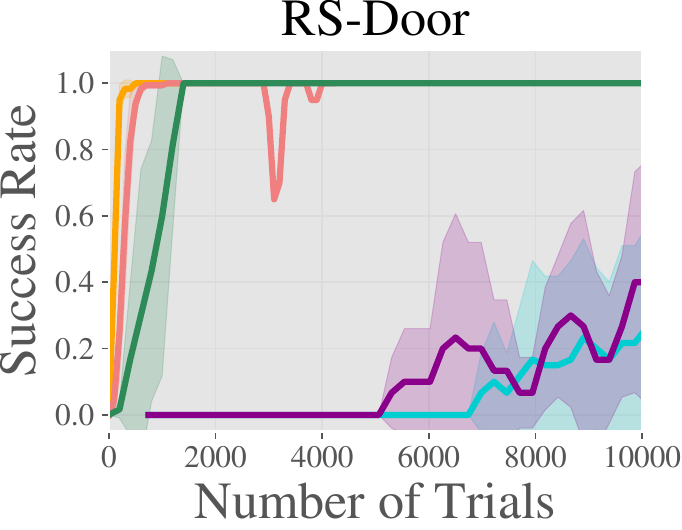}
    \includegraphics[width=.24\textwidth]{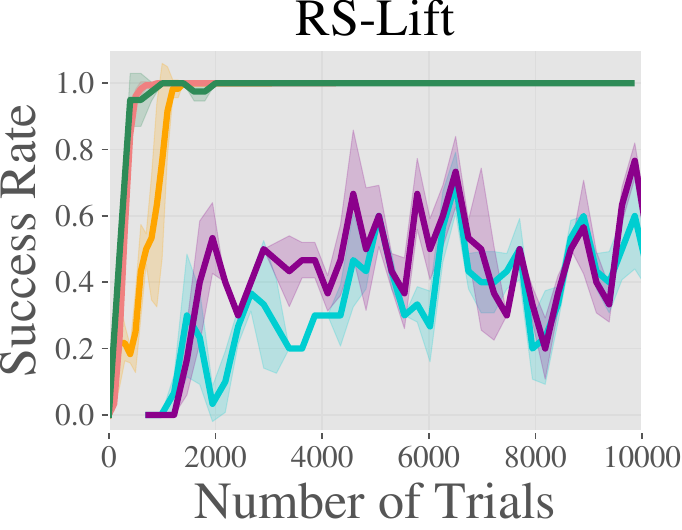}
    \includegraphics[width=.24\textwidth]{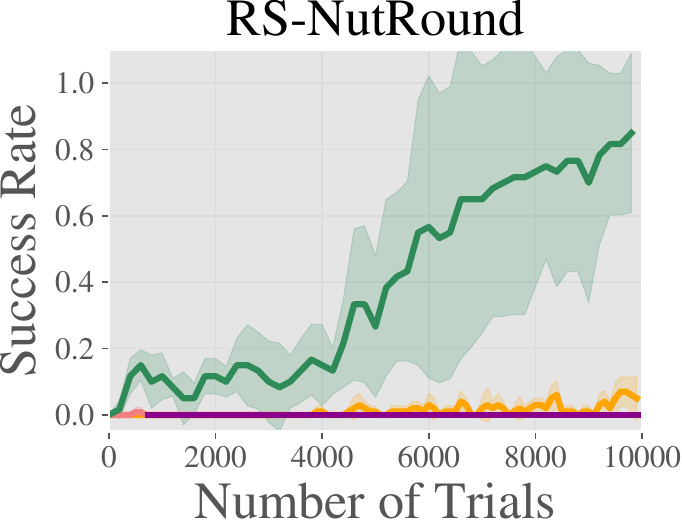}
    \includegraphics[width=.24\textwidth]{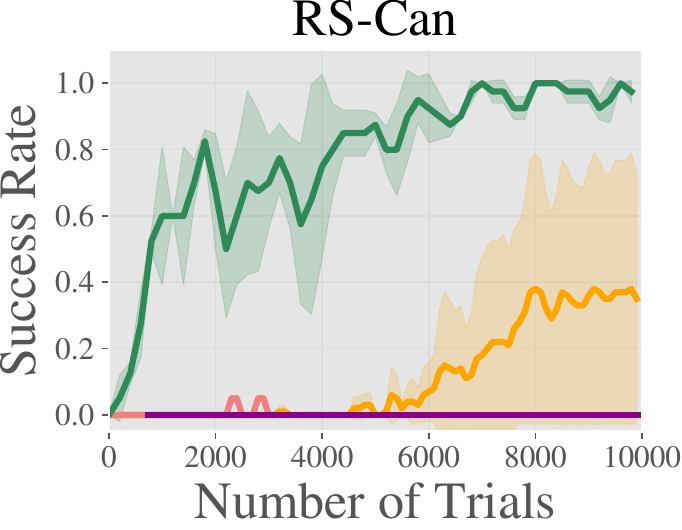}
    \\
    \vspace{2pt}
    \includegraphics[width=.9\textwidth]{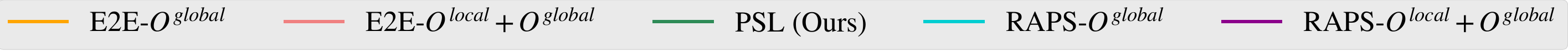}
    \caption{\small \textbf{Ablating Camera Views for Baselines.} We plot task success rate as a function of the number of trials. As seen in the figure above, including $O^{local}$ does not improve the performance of E2E or RAPS.}
    \label{fig:baseline camera abl}
    \vspace{-7pt}
\end{figure}

\textbf{Ablating Camera View choice for Baselines:}
We evaluate if including $O^{local}$ in addition to $O^{global}$ improves performance across four tasks (\texttt{RS-Lift}, \texttt{RS-Door}, \texttt{RS-Can}, \texttt{RS-NutRound}) and include the results in Fig.~\ref{fig:baseline camera abl}. In general there is little to no performance improvement for RAPS or E2E across the board. The additional local view marginally improves sample efficiency but it does not resolve the fundamental exploration problem for these tasks.

\begin{table}[ht]
\centering
\scriptsize
\begin{tabular}{@{}ccccc@{}}
\toprule
& K-Single-Task & K-MS-3 & K-MS-4 & K-MS-5 \\
\midrule 
\textbf{PSL\xspace}-Wrist & 1.0 $\pm$ 0.0  & 1.0 $\pm$ 0.0  & 1.0 $\pm$ 0.0 & 1.0 $\pm$ 0.0      \\ 
\rowcolor{Gray}
\textbf{PSL\xspace}-Fixed & 1.0 $\pm$ 0.0  & 0.0 $\pm$ 0.0  & 0.0 $\pm$ 0.0 & 0.0 $\pm$ 0.0      \\
\textbf{PSL\xspace}-Wrist+Fixed & 1.0 $\pm$ 0.0 & 0.0 $\pm$ 0.0   & 0.0 $\pm$ 0.0 & 0.0 $\pm$ 0.0      \\
\bottomrule 
\end{tabular}
\vspace{2pt}
\caption{\small \textbf{Chaining Pre-trained Policies Ablation.} PSL\xspace can leverage local views (wrist cameras) to chain together multiple pre-trained policies via motion-planning using the Sequencing Module. While PSL\xspace with each camera input is able to produce a capable single-task policy, chaining only works with wrist camera observations as the observations are kept in-distribution.}
\label{table:chain_policy_abl}
\vspace{-5pt}
\end{table}

\begin{figure}[ht]
    \centering
    \includegraphics[width=.24\textwidth]{chapters/psl/figures/single_stage/robosuite_Lift.pdf}
    \includegraphics[width=.24\textwidth]{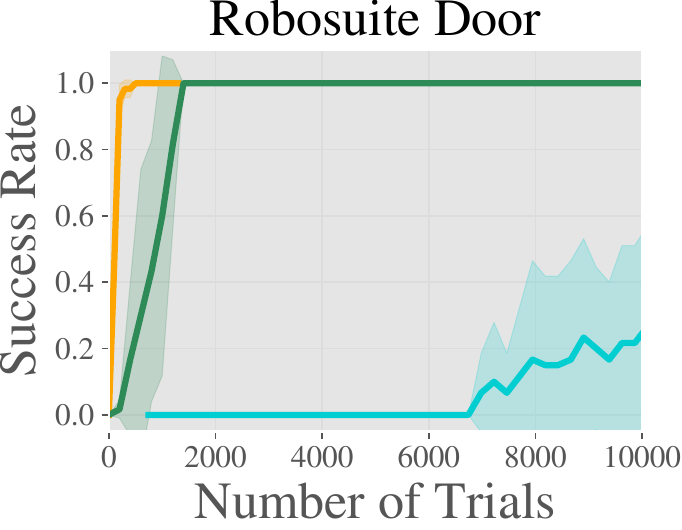}
    \includegraphics[width=.24\textwidth]{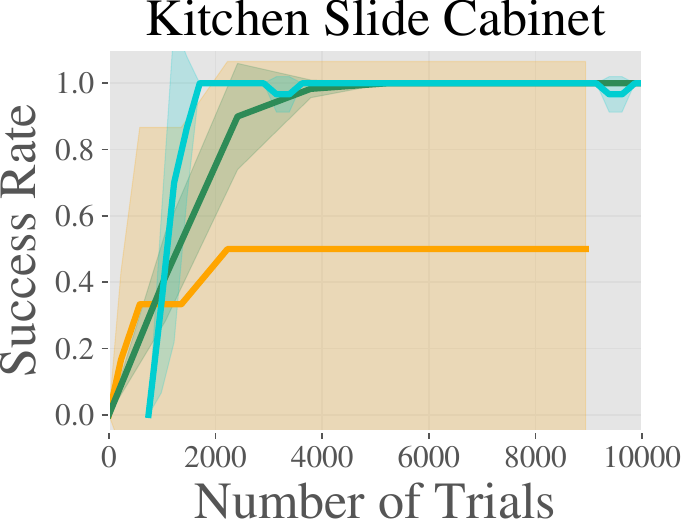}
    \includegraphics[width=.24\textwidth]{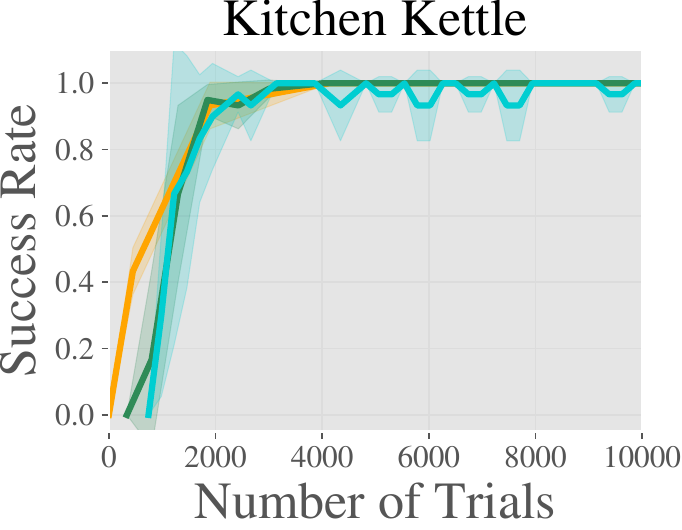}
    \includegraphics[width=.24\textwidth]{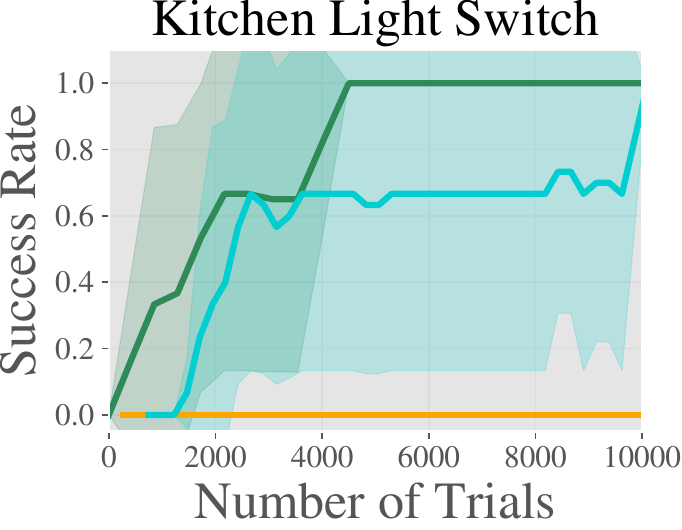}
    \includegraphics[width=.24\textwidth]{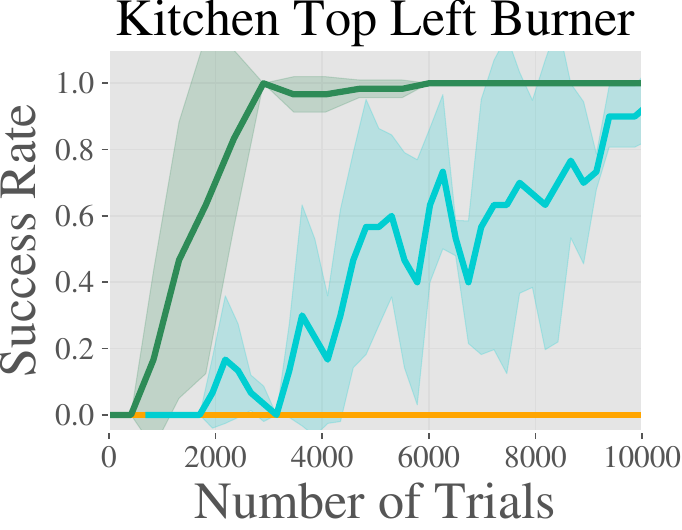}
    \includegraphics[width=.24\textwidth]{chapters/psl/figures/single_stage/kitchen_kitchen-microwave-v0.pdf}
    \includegraphics[width=.24\textwidth]{chapters/psl/figures/single_stage/metaworld_disassemble-v2.pdf}
    \includegraphics[width=.24\textwidth]{chapters/psl/figures/single_stage/mopa_SawyerAssemblyObstacle-v0.pdf}
    \includegraphics[width=.24\textwidth]{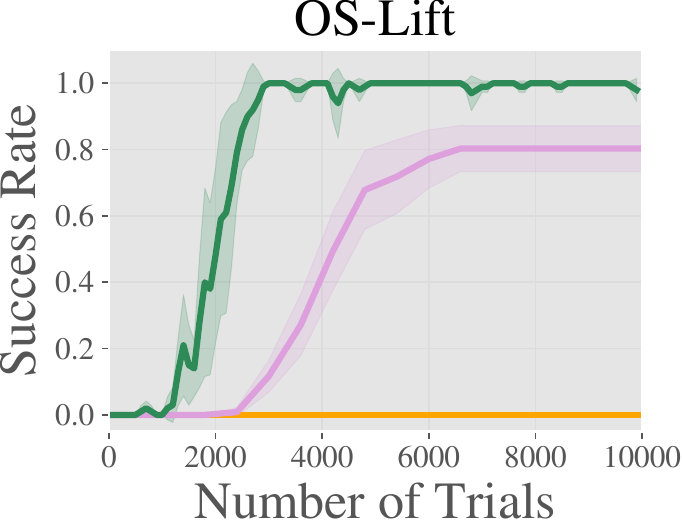}
    \includegraphics[width=.24\textwidth]{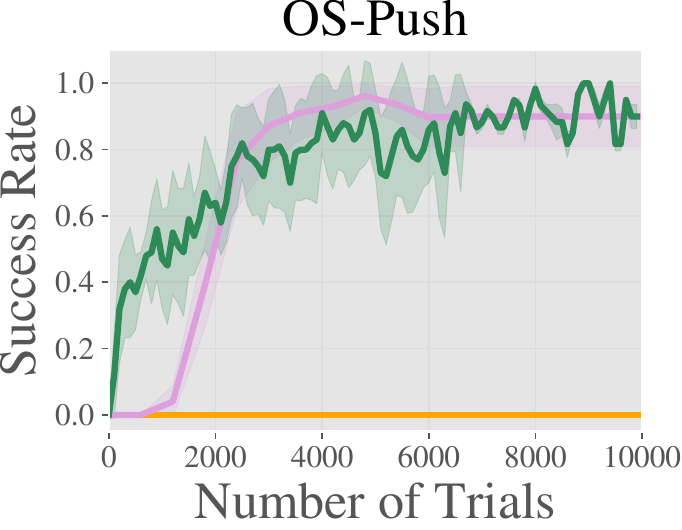}
    \\
    \includegraphics[width=.4\textwidth]{chapters/psl/figures/single_stage/legend.pdf}
    \vspace{10pt}
    \caption{\small \textbf{Single Stage Results.} We plot task success rate as a function of the number of trials. PSL\xspace improves on the efficiency of the baselines across single-stage tasks (\textit{plan length of 1}) in Robosuite, Kitchen, Meta-World, and Obstructed Suite, \textbf{achieving an asymptotic success rate of 100\% on all 11 tasks}.}
    \label{fig:single stage results}
    \vspace{-7pt}
\end{figure}

\begin{figure}[ht]
    \centering
    \includegraphics[width=.24\textwidth]{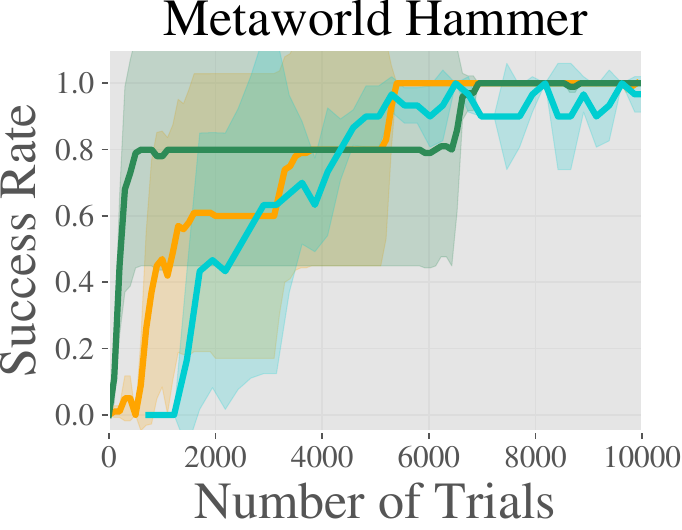}
    \includegraphics[width=.24\textwidth]{chapters/psl/figures/multi_stage/metaworld_assembly-v2.pdf}
    \includegraphics[width=.24\textwidth]{chapters/psl/figures/multi_stage/metaworld_bin-picking-v2.pdf}
    \\
    \includegraphics[width=.4\textwidth]{chapters/psl/figures/single_stage/legend.pdf}
    \vspace{10pt}
    \caption{\small \textbf{Meta-World Two Stage Learning Curves.} We plot task success rate as a function of the number of trials. PSL\xspace learns faster than the baselines by employing high-level planning to accelerate RL performance.}
    \label{fig:metaworld_two_stage_curves}
    \vspace{-7pt}
\end{figure}

\setlength{\tabcolsep}{2.4pt}

\begin{table}[ht]
\scriptsize
\centering
\begin{tabular}{ccccc}
\toprule
                                & \texttt{MW-BinPick} & \texttt{MW-Assembly}   & \texttt{MW-Hammer} &                       \\
\midrule
\textbf{E2E}                        & 1.0 $\pm$ 0.0 & 0.4 $\pm$ 0.5 & 0.0 $\pm$ 1.0                          \\
\rowcolor{Gray}
\textbf{RAPS}                       & 0.0 $\pm$ 0.0   & 0.3 $\pm$ .25    & 1.0 $\pm$ 0.0                              \\
\textbf{TAMP}                       & 1.0 $\pm$ 0.0   & 1.0 $\pm$ 0.0    & 0.0 $\pm$ 0.0                                \\
\rowcolor{Gray}
\textbf{SayCan}                    & 1.0 $\pm$ 0.0   & 0.5 $\pm$ .08 & 1.0 $\pm$ 0.0                        \\
\midrule
\textbf{PSL\xspace} & 1.0 $\pm$ 0.0   & 1.0 $\pm$ 0.0  & 1.0 $\pm$ 0.0     \\
\bottomrule
\end{tabular}
\vspace{2pt}
\caption{\small \textbf{Metaworld Two Stage Results.} While the baselines perform well on most of the tasks, only PSL\xspace is able to consistently solve every task. This is because the LLM planning and Sequencing modules ease the learning burden for the RL policy, enabling it to learn contact-rich, long-horizon behaviors.}
\label{table:metaworld two stage results}
\vspace{-10pt}
\end{table}

\clearpage

\section{PSL\xspace Implementation Details}
\label{app:our impl details}
\begin{algorithm}
    \footnotesize
    \caption{PSL\xspace Implementation}
    \begin{algorithmic}[1]
    \Require LLM, task description $g_l$, Motion Planner MP, low-level horizon $H_l$, segmentation model $\mathcal{S}$, RGB-D global cameras, RGB wrist camera, Camera Matrix $K^{global}$
    \State initialize RL: $\pi_{\theta}$, replay buffer $\mathcal{R}$
    \Statex \textit{Planning Module} 
    \State High-level plan $\mathcal{P}\leftarrow$ Prompt(LLM, $g_l$)
    \For{episode $1...N$}
        \For{$p \in \mathcal{P}$}
            \Statex \textit{Sequencing Module}
            \State target region ($t$), termination condition $\leftarrow p$
            \State $PC^{global}$ = Projection($O^{global}_1$, $O^{global}_2$, $K^{global}$)
            \State $M_{robot}, M_{obj}$ = Segmentation($O^{global}_1$, $O^{global}_2$, robot, object)
            \State $PC^{robot}$ , $PC^{object}$ = $M_{robot}*PC^{global}$, $M_{obj}*PC^{scene}$
            \State $PC^{scene} = PC^{global} - PC^{robot}$
            \State $ee_{target}$ = mean($PC^{obj}$)
            \State $q_{target}$ = IK($ee_{target}$)
            \State MotionPlan(MP, $q_{target}$, $PC^{scene}$)
            \Statex \textit{Learning Module}
            \For{$i=1,...,h$ low-level steps}
                \State Get action $a_t \sim \pi_{\theta}(O^{local}_t)$
                \State Get next state $O^{local}_{t+1} \sim p(· | s_t, a_t)$.
                \State Store $(O^{local}_t, a_t, O^{local}_{t+1}, r)$ into $\mathcal{R}$
                \State Sample $(O^{local}_k, a_t, O^{local}_{k+1}, r) \sim \mathcal{R}$
                \Comment k = random index
                \State update $\pi_{\theta}$ using RL
                \If{post-condition}
                    \State break
                \EndIf
            \EndFor
        \EndFor
    \EndFor
    \end{algorithmic}
    \label{alg:main-alg}
\end{algorithm}
\subsection{Planning Module}
Given a task description $g_l$, we prompt an LLM using the format described in Sec.~\ref{sec:plan} to produce a language plan. We experimented with a variety of publicly available and closed-source LLMs including LLAMA~\cite{touvron2023llama}, LLAMA-2~\cite{touvron2023llama2}, GPT-3~\cite{brown2020language}, Chat-GPT, and GPT-4~\cite{openai2023gpt4}. In initial experiments, we found that GPT-based models performed best, and GPT-4 in particularly most closely adhered to the prompt and produced the most accurate plans. As a result, in our experiments, we use GPT-4 as the LLM planner for all tasks. We sample from the model with temperature $0$ for determinism. Sometimes, the LLM hallucinates non-existent stage termination conditions or objects. As a result, we add a pre-processing step in which we delete components of the plan that contain such hallucinations.

\subsection{Sequencing Module}
\begin{figure}[ht]
    \centering
    \includegraphics[width=\linewidth]{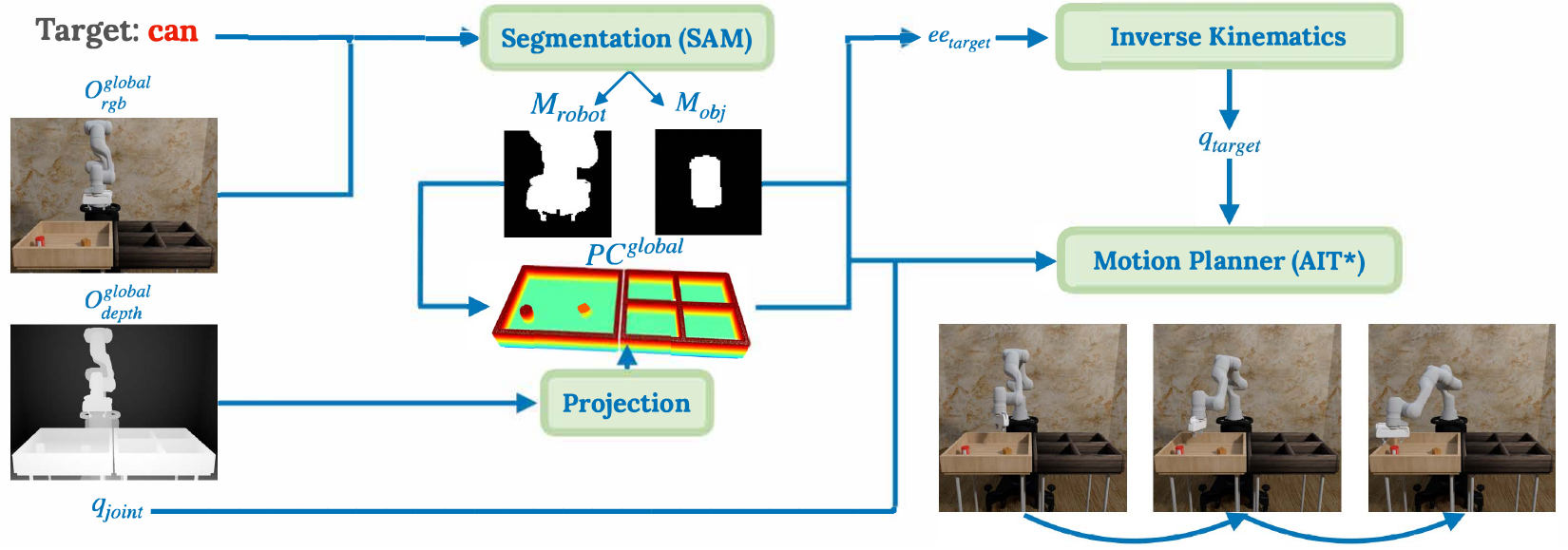}
    \vspace{-7pt}
    \caption{\small \textbf{Sequencing Module.} Inputs to the Sequencing Module are two calibrated RGB-D fixed views, $O^{global}$, the proprioception $q_{joint}$ and the target object. It performs visual motion planning to the target object by computing a scene point-cloud ($PC^{global}$), segmenting the target object ($M_{obj}$) to estimate its pose ($q_{target}$), segmenting the robot ($M_{robot}$) to remove it from $PC^{global}$ and motion planning using AIT*.}
    \label{fig:seq-module}
\end{figure}
The input to the Sequencing Module is $O^{global}$. In our experiments, we use two camera views, $O^{global}_1$ and $O^{global}_2$, which are RGB-D calibrated camera views of the scene, to obtain unoccluded views of the scene. We additionally provide the current robot configuration, which is joint angles for robot arms: $q_{joint}$ and the target region label around which the RL policy must perform environment interaction. From this information, the module must solve for a collision free path to a region near the target. This problem can be addressed by classical motion planning. We take advantage of sampling-based motion planning due to its minimal setup requirements (only collision-checking) and favorable performance on planning. In order to run the motion planner, we require a collision checker, which we implement using point-clouds. To compute the target object position, we use predicted segmentation along with calibrated depth, as opposed to a dedicated pose estimation network, primarily because state of the art segmentation models~\citep{kirillov2023segment,zhou2022detecting} have significant zero-shot capabilities across objects.

\textbf{Projection:} In this step, we project the depth map from each global view of the scene, $O^{global}_1$ and $O^{global}_2$ into a point-cloud $PC^{global}$ using their associated camera matrices $K^{global}_1$ and $K^{global}_2$. We perform the following processing steps to clean up $PC^{global}$: 1) cropping to remove all points outside the workspace 2) voxel down-sampling with a size of 0.005 $m^3$ to reduce the overall size of $PC^{global}$ 3) outlier removal, which prunes points that are farther from their 20 neighboring points than the average in the point-cloud as shown in Fig.~\ref{fig:seq-module}. 

\textbf{Segmentation:} We compute masks for the robot ($M_{robot})$ and the target object ($M_{obj}$) by using a segmentation model (SAM~\cite{kirillov2023segment}) $\mathcal{S}$ which segments the scene based on RGB input. We reduce noise in the masks by filling holes, computing contiguous mask clusters and selecting the largest mask. We use $M_{robot}$ to remove the robot from $PC^{global}$, in order to perform collision checking of the robot against the scene. Additionally, we use $M_{obj}$ along with $PC^{global}$ to compute the object point-cloud $PC^{obj}$, which we average to obtain an estimate of object position, which is the target position for the motion planner. For the manipulation tasks we consider in the paper, this is the target end-effector pose of the robot, $ee_{target}$.

\textbf{Visual Motion Planning:} Given the target end-effector pose $ee_{target}$, we use inverse kinematics (IK) to compute $q_{target}$ and pass $q_{joint}, q_{target}, PC^{global}$ into a joint-space motion planner. 
To that end, we use a sampling-based motion planner, AIT*~\cite{strub2020adaptively}, to perform motion planning. In order to implement collision checking from vision, for a sampled joint-configuration $q_{sample}$, we compute the corresponding position of the robot mesh and compute the occupancy of each point in the scene point-cloud against the robot mesh. If the object is detected as grasped, then we additionally remove the object from the scene pointcloud, compute its convex hull and use the signed distance function of the joint robot-object mesh for collision checking. As a result, the Sequencing Module operates entirely over visual input, and achieves a pose near the region of interest before handing off control to the local RL policy. We emphasize that the Sequencing Module \textit{does not need to be perfect}, it merely needs to produce a reasonable initialization for the Learning Module.

\subsection{Learning Module}
\subsubsection{Stage Termination Details}
As described in Section~\ref{sec:method}, we use stage termination conditions to determine when the Learning Module should hand control back to the Sequencing Module to continue to the next stage in the plan. For most of the tasks we consider, these stage termination conditions amount to checking for a grasp or placement for the target object in the stage. For example, for \texttt{RS-NutRound}, the plan for the first stage is (grasp, nut) and the plan for the second stage is (place, peg). Placements are straightforward to check: simply evaluate if the object being manipulated is within a small region near the target object. This can be computed using the estimated pose of the two objects (current and target). Grasps are more challenging to estimate and we employ a two stage pipeline to detecting a grasp. First, we estimate the object pose and then evaluate if the z value has increased from when the stage began. Second, in order to ensure the object is not simply tossed in the air, we check if the robot's gripper is tightly caging the object. We do so by collision checking the object point-cloud against the gripper mesh. We use the same collision checking procedure as outlined in Sec~\ref{sec:method} for checking collision between the scene point-cloud and robot mesh.

To estimate the turned condition, we compute the mask of the burner knob, evaluate its principal axis and measure its angle from vertical. If it is greater than X radians then the stage condition triggers. Checking the pushed condition is straightforward, the object needs to have moved by X distance forward from the start pose in xy. Finally, for checking the opened condition, we estimate the handle pose relative to the hinge and compute the angle of the door. If it is greater than X radians we consider the door opened. For the slide cabinet the handle pose itself can be used to check opening. Closing is likewise estimated as the inverse of opening. For all conditions, we take the threshold X from the environment success condition or reward function.

\subsubsection{Training Details}
For all tasks, we use the reward function defined by the environment, which may be dense or sparse depending on the task. We find that for PSL\xspace, it is crucial to use an action-repeat of 1, in general we found that increasing this harmed performance, in contrast to the E2E baseline which performs best with an action repeat of 2.
For training policies using DRQ-v2, we use the default hyper-parameters from the paper, held constant across all tasks. We train policies using 84x84 images. We use the "medium" difficult exploration schedule defined in~\cite{yarats2021mastering}, which anneals the exploration $\sigma$ from $1.0$ to $0.1$ over the course of 500K environment steps. Due to memory concerns, instead of using a replay buffer size of 1M as done in ~\citet{yarats2021mastering}, ours is of size 750K across each task. Finally, for path length, we use the standard benchmark path length for E2E and MoPA-RL, $5$ per stage for RAPS following ~\citet{dalal2021accelerating}, and $25$ per stage for PSL\xspace. 

\clearpage
\section{Baseline Implementation Details}
\label{app:baseline impl details}
\subsection{RAPS}
For this baseline, we simply take the results from the RAPS~\cite{dalal2021accelerating} paper as is, which use Dreamer~\cite{hafner2019dream} and sparse rewards. In initial experiments, we attempted to combine RAPS with DRQ-v2~\cite{yarats2021mastering} and found that Dreamer performed better, which is consistent with RAPS+Dreamer having the best results in \citet{dalal2021accelerating}. We additionally tried to run RAPS with dense rewards, but found that the method performed significantly worse. One potential reason for this is that it is not clear exactly how to aggregate the dense rewards across primitive executions - we tried simply taking the dense reward after executing a primitive as well as simply summing the rewards of intermediate primitive executions. Both performed worse than training RAPS with sparse rewards. Note that PSL\xspace outperforms RAPS even when both methods have only access to sparse rewards, e.g. the Kitchen environments. We observe clear benefits over RAPS on the single-stage (Fig.~\ref{fig:single stage results}) and multi-stage (Table~\ref{table:multi stage results}) tasks.

\subsection{MoPA-RL}
As described in the main paper, we take the results from MoPA-RL~\cite{yamada2021motion} as is on the Obstructed Suite of tasks. Those results were run from state-based input and leveraged the simulator for collision checking. We do so as we were unable to successfully combine MoPA-RL with DRQ-v2 based on the publicly released implementations of both methods. 

\subsection{TAMP}
We use PDDLStream~\cite{garrett2020pddlstream} as the TAMP algorithm of choice as it has been shown to have strong planning performance on long-horizon manipulation tasks in Robosuite~\citep{dalal2023optimus,mandlekarhitltamp}. The PDDLStream planning framework models the TAMP domain and uses the adaptive algorithm, a sampling based algorithm, to plan. This TAMP method uses samplers for grasp generation, placement sampling, inverse kinematics, and motion planning, making performance stochastic. Hence we average performance across 50 evaluations to reduce variance. We adapt the authors TAMP implementation (from ~\citep{dalal2023optimus,mandlekarhitltamp}) for our tasks. Note this method uses privileged access to the simulator, leveraging knowledge about the task (which must be explicitly specified in a problem file), the scene (from the domain file and access to collision checking) and 3D geometry of the environment objects. 

\subsection{SayCan}
As described in the main paper, we re-implement SayCan~\citet{ahn2022can} using GPT-4 (the same LLM we use in our methdo) and manually engineered pick/place skills that use pose-estimation and motion-planning. Following our Sequencing module: 1) we build a 3D scene point-cloud using camera calibration and depth images 2) we perform vision-based pose estimation using segmentation along with the scene point cloud and 3) we run motion planning using collision queries from the 3D point-cloud, which is used for collision queries. Finally, we use heuristically engineered pick and place primitives to perform interaction behavior which we describe as follows. We note that for our tasks of interest, the pick motion can be represented as a top-grasp. Once we position the robot near the object; we then simply lower the robot arm till the end-effector (not the grippers) come in contact with the object. We then close the gripper to execute the grasp. For place, we follow the implementation of ~\citet{ahn2022can} and lower the held object until contact with a surface, then release (open the gripper) and lift the robot arm. 
We set the affordance function for both skills to 1, following the design in~\citet{ahn2022can} for motion planned skills. 

For LLM planning, we specify the following prompt: 
\begin{framed}
Given the following library of robot skills: ...
Task description: ... 
Make sure to take into account object geometry. Formatting of output: a list of robot skills. Don't output anything else. 
\end{framed}
This prompt is the same as our prompt except we specify the robot skill library in terms of object centric behaviors, instead of stage termination conditions. 

\begin{framed}
    Given the following library of robot skills: ... Task description: ... Give me a simple plan to solve the task using only the provided skill library. Make sure the plan follows the formatting specified below and make sure to take into account object geometry. Formatting of output: a list of robot skills. Don't output anything else. 
\end{framed}

\texttt{Robosuite}
\begin{framed}
\textbf{Skill Library:} pick can, pick milk, pick cereal, pick bread slice, pick silver nut, pick gold nut, 
put can on/in X, put milk on/in X, put cereal on/in X, put bread slide on/in X, put silver nut on/in X, put gold nut on/in X, grasp door handle, turn door handle, pick cube
\end{framed}

\texttt{Kitchen}
\begin{framed}
\textbf{Skill Library:} grasp vertical door handle for slide cabinet, move left, move right, grasp hinge cabinet, grasp top left burner with red tip, rotate top left burner with red tip 90 degree clockwise, rotate top left burner with red tip 90 degrees counterclockwise, push light switch knob left, push light switch knob right, grasp kettle, lift kettle, place kettle on/in X, grasp microwave handle, pull microwave handle
\end{framed}

\texttt{Metaworld:}
\begin{framed}
\textbf{Skill Library:} grasp cube, place cube on/in X, grasp hammer, place hammer, hit nail with hammer, grasp wrench, lift wrench
\end{framed}

\texttt{Obstructed-Suite}
\begin{framed}
\textbf{Skill Library:} grasp can, place can in bin, insert table leg in X, move table leg, grasp cube, place cube on table, push cube
\end{framed}

\clearpage

\section{Tasks}
\label{app:tasks}
\begin{figure*}[ht!]
\centering
\begin{subfigure}[b]{0.24\linewidth}
    \includegraphics[width=\linewidth]{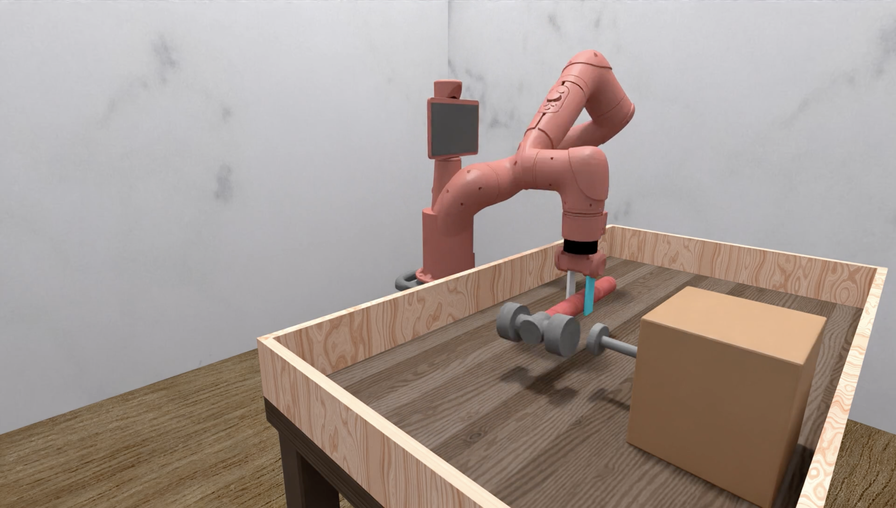}
    \caption{\small \texttt{MW-Hammer}}
\end{subfigure}
\vspace{.2in}
\begin{subfigure}[b]{0.24\linewidth}
    \includegraphics[width=\linewidth]{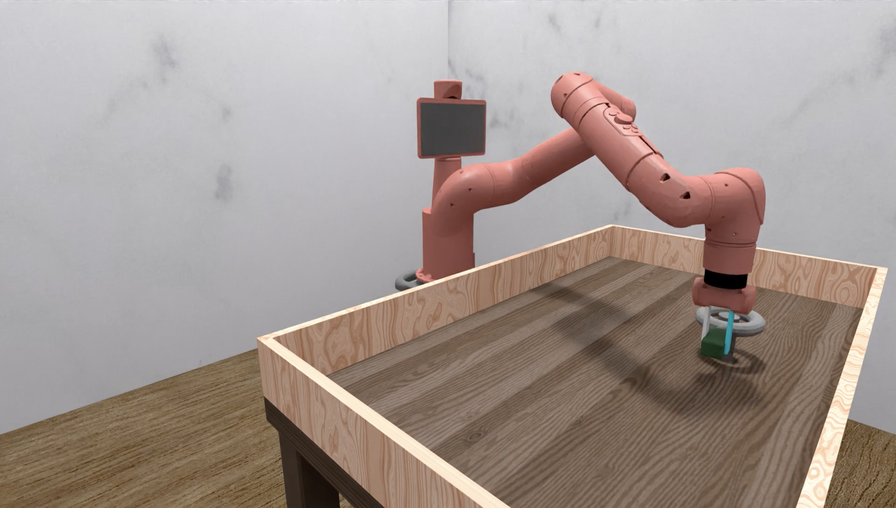}
    \caption{\small \texttt{MW-Assembly}}
\end{subfigure}
\begin{subfigure}[b]{0.24\linewidth}
    \includegraphics[width=\linewidth]{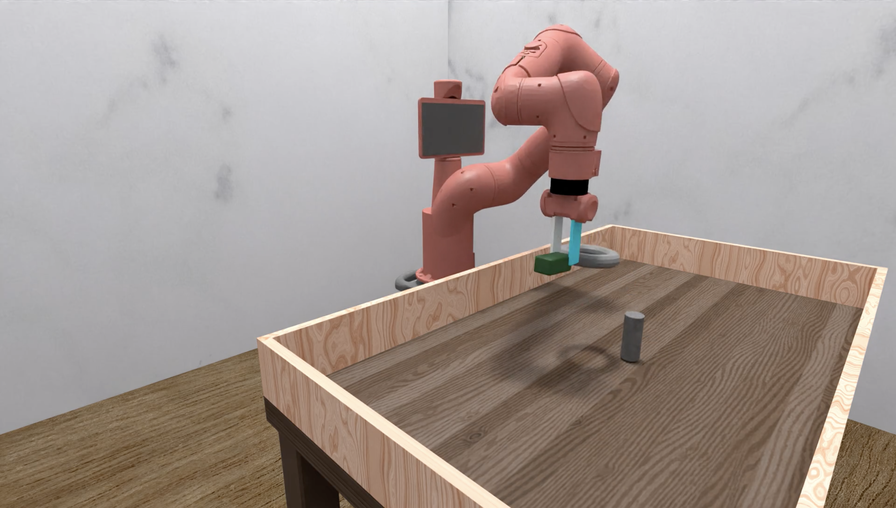}
    \caption{\small \texttt{MW-Disassemble}}
\end{subfigure}
\begin{subfigure}[b]{0.24\linewidth}
    \includegraphics[width=\linewidth]{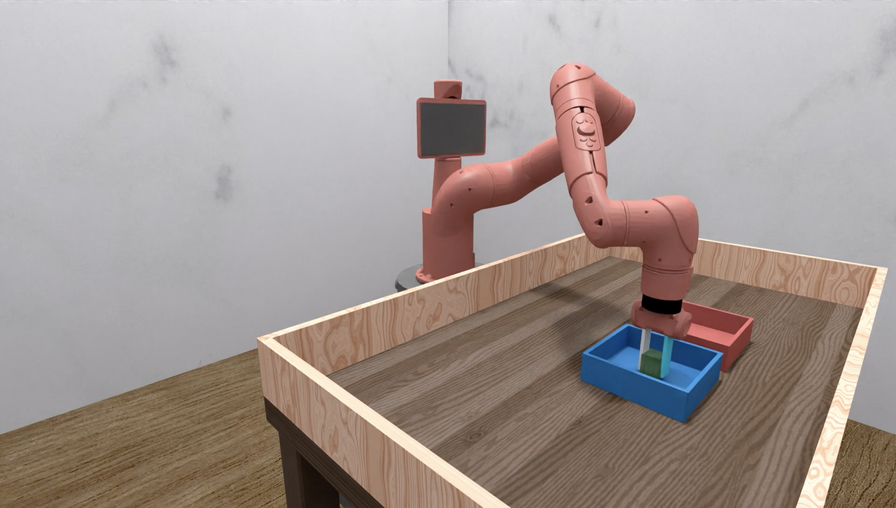}
    \caption{\small \texttt{MW-Bin-Picking}}
\end{subfigure}
\begin{subfigure}[b]{0.24\linewidth}
    \includegraphics[width=\linewidth]{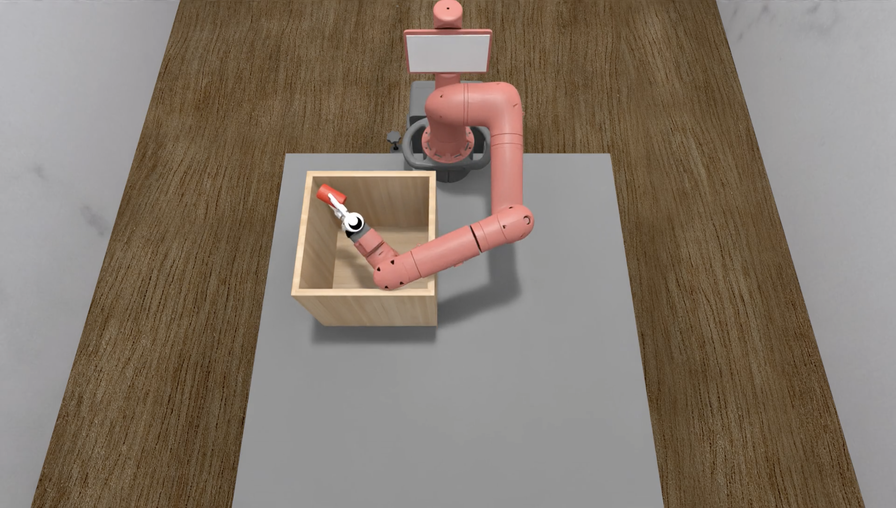}
    \caption{\small \texttt{OS-Lift}}
\end{subfigure}
\vspace{.2in}
\begin{subfigure}[b]{0.24\linewidth}
    \includegraphics[width=\linewidth]{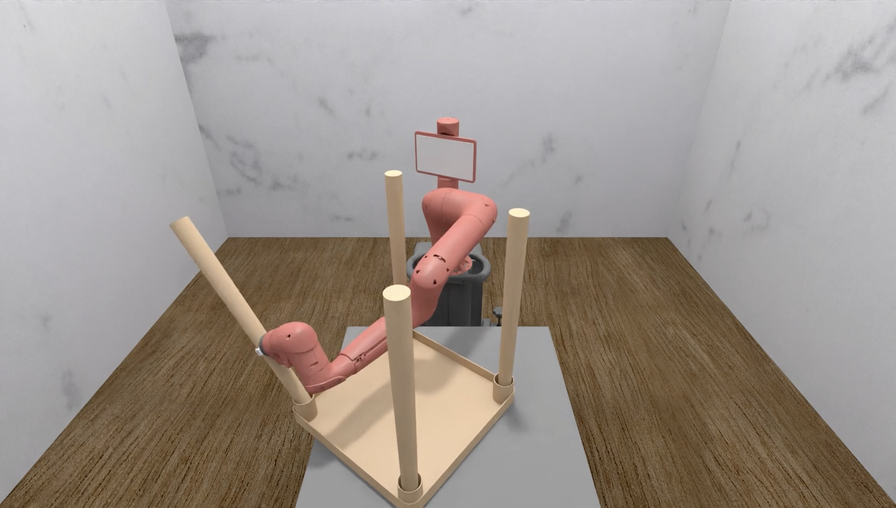}
    \caption{\small \texttt{OS-Assembly}}
\end{subfigure}
\begin{subfigure}[b]{0.24\linewidth}
    \includegraphics[width=\linewidth]{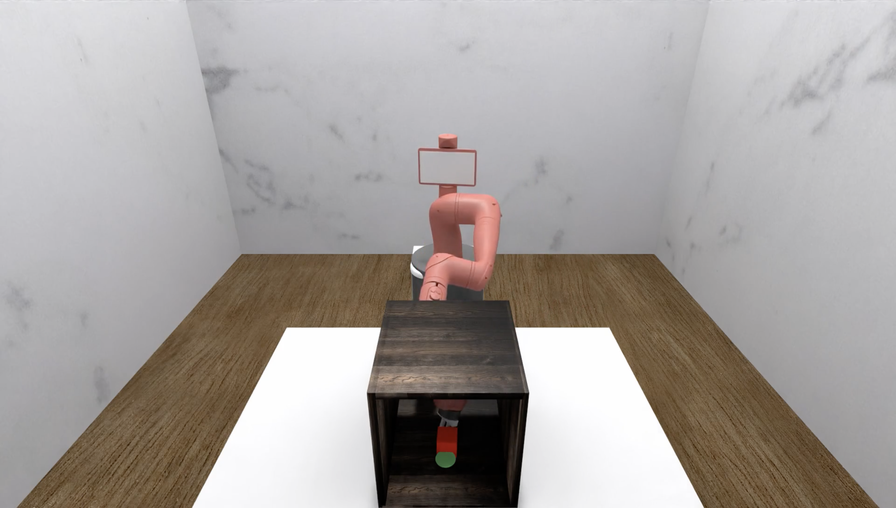}
    \caption{\small \texttt{OS-Push}}
\end{subfigure}
\begin{subfigure}[b]{0.24\linewidth}
    \includegraphics[width=\linewidth]{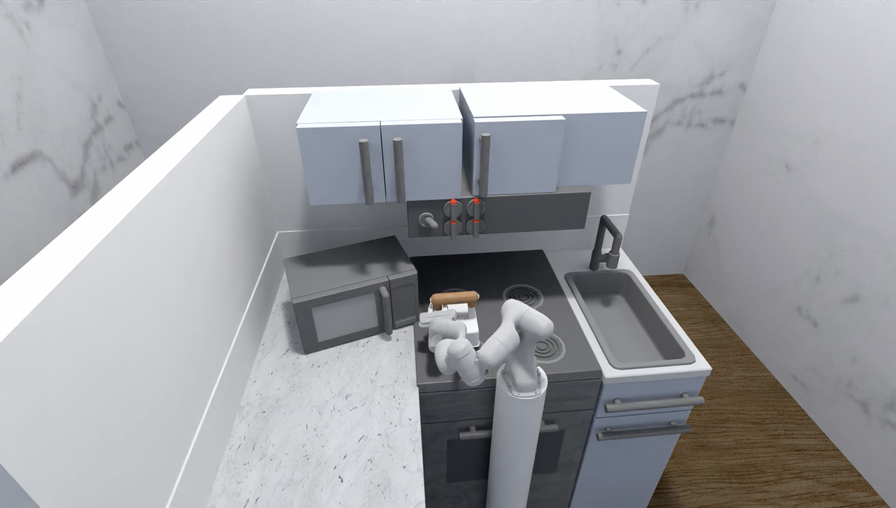}
    \caption{\small \texttt{K-Kettle}}
\end{subfigure}
\vspace{.2in}
\begin{subfigure}[b]{0.24\linewidth}
    \includegraphics[width=\linewidth]{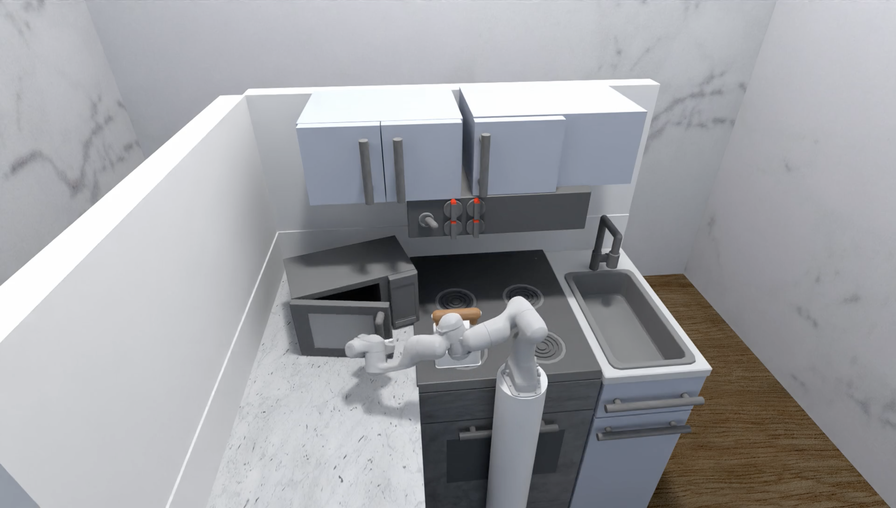}
    \caption{\small \texttt{K-Microwave}}
\end{subfigure}
\begin{subfigure}[b]{0.24\linewidth}
    \includegraphics[width=\linewidth]{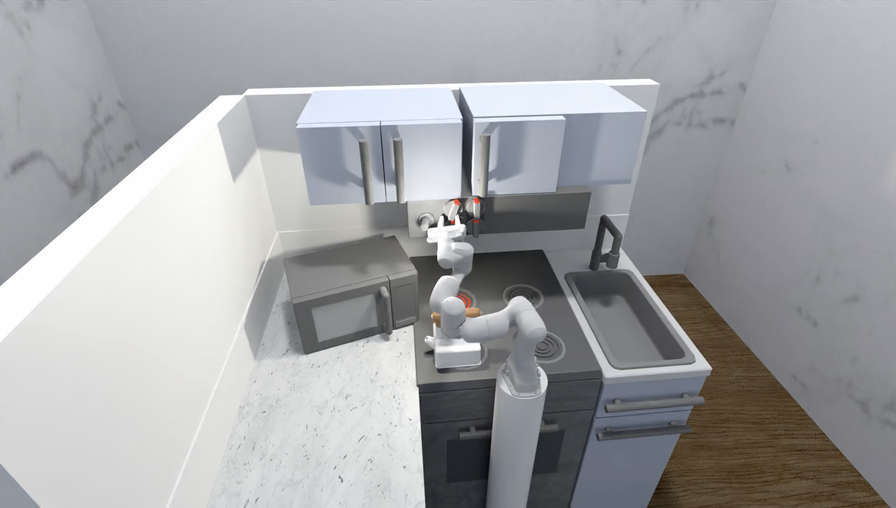}
    \caption{\small \texttt{K-Burner}}
\end{subfigure}
\begin{subfigure}[b]{0.24\linewidth}
    \includegraphics[width=\linewidth]{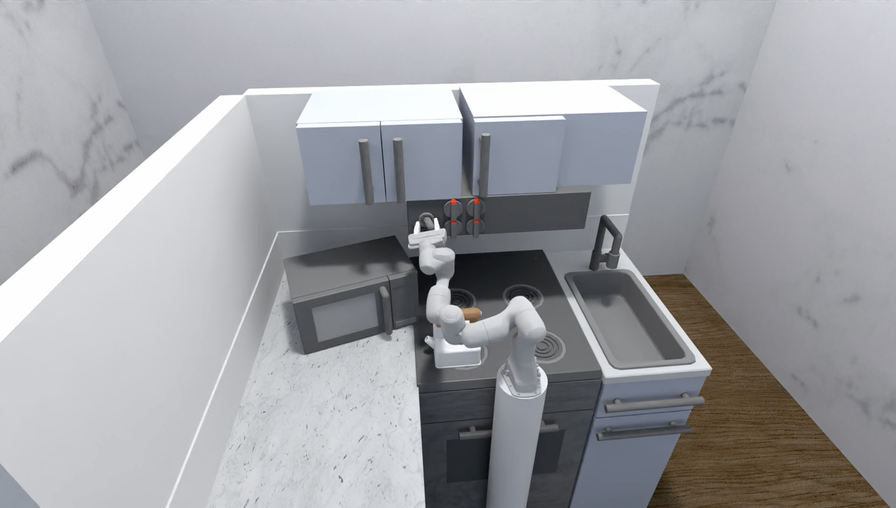}
    \caption{\small \texttt{K-Light}}
\end{subfigure}
\begin{subfigure}[b]{0.24\linewidth}
    \includegraphics[width=\linewidth]{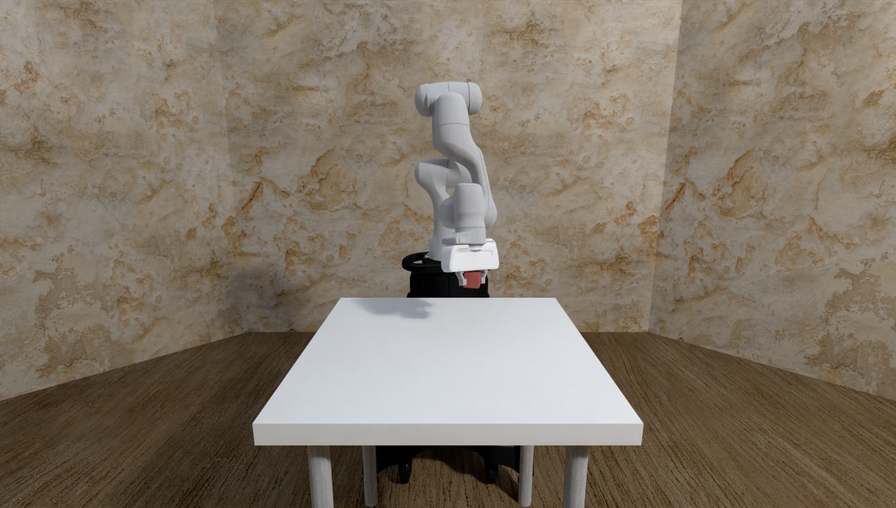}
    \caption{\small \texttt{RS-Lift}}
\end{subfigure}
\vspace{.2in}
\begin{subfigure}[b]{0.24\linewidth}
    \includegraphics[width=\linewidth]{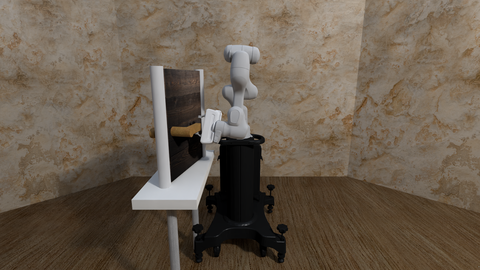}
    \caption{\small \texttt{RS-Door}}
\end{subfigure}
\begin{subfigure}[b]{0.24\linewidth}
    \includegraphics[width=\linewidth]{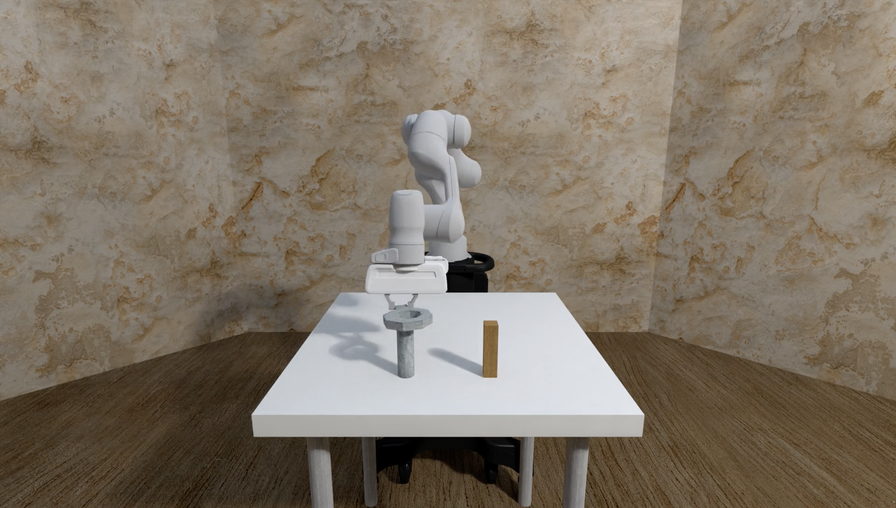}
    \caption{\small \texttt{RS-NutRound}}
\end{subfigure}
\begin{subfigure}[b]{0.24\linewidth}
    \includegraphics[width=\linewidth]{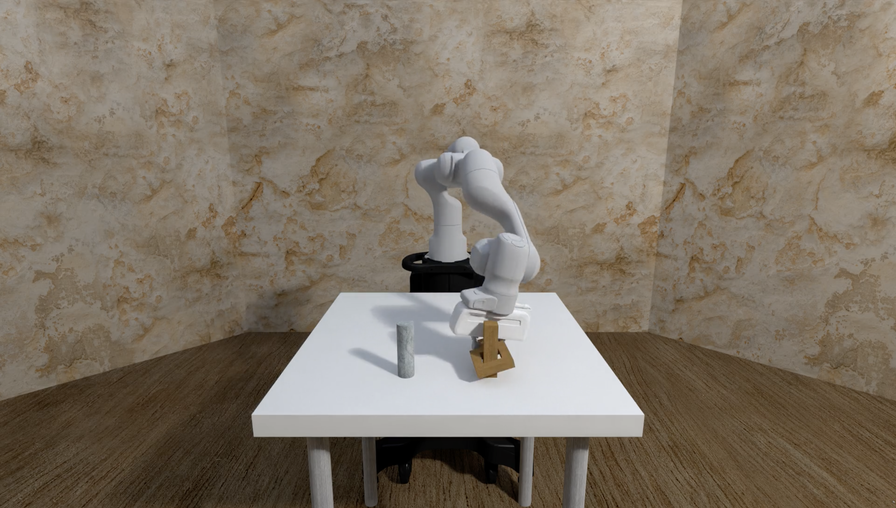}
    \caption{\small \texttt{RS-NutSquare}}
\end{subfigure}
\begin{subfigure}[b]{0.24\linewidth}
    \includegraphics[width=\linewidth]{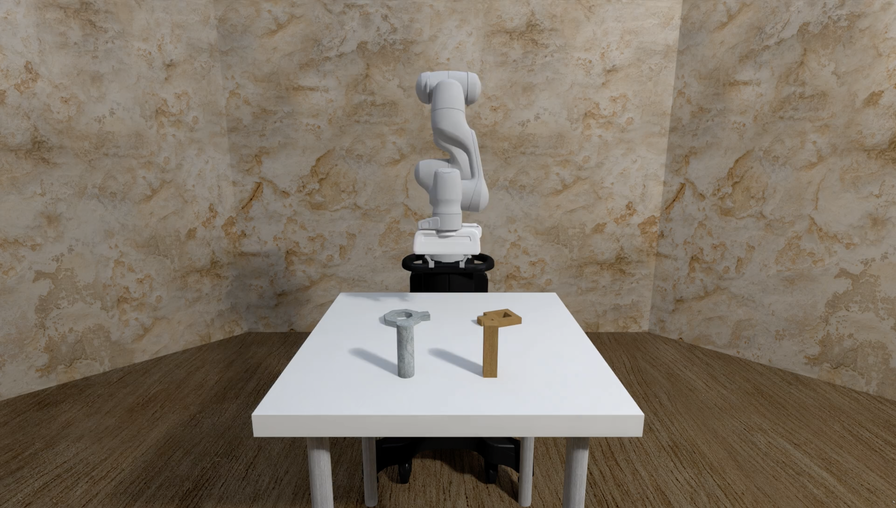}
    \caption{\small \texttt{RS-NutAssembly}}
\end{subfigure}
\vspace{.2in}
\begin{subfigure}[b]{0.24\linewidth}
    \includegraphics[width=\linewidth]{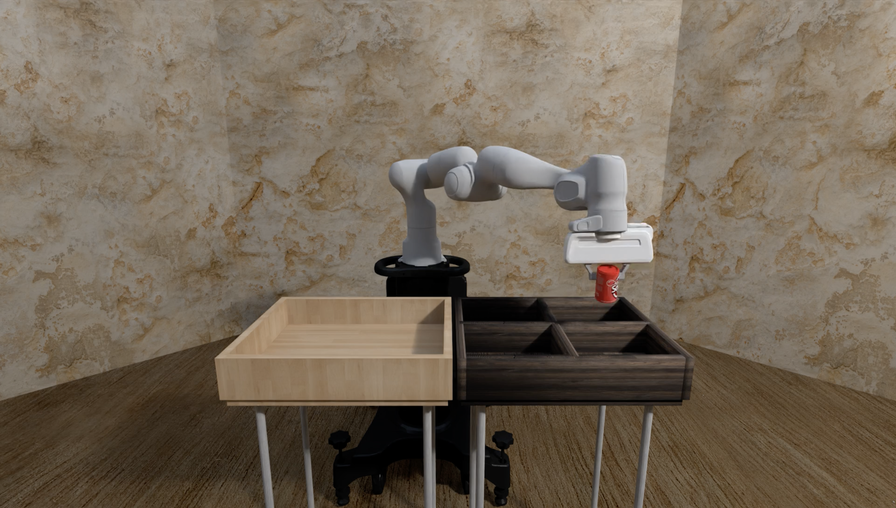}
    \caption{\small \texttt{RS-Can}}
\end{subfigure}
\begin{subfigure}[b]{0.24\linewidth}
    \includegraphics[width=\linewidth]{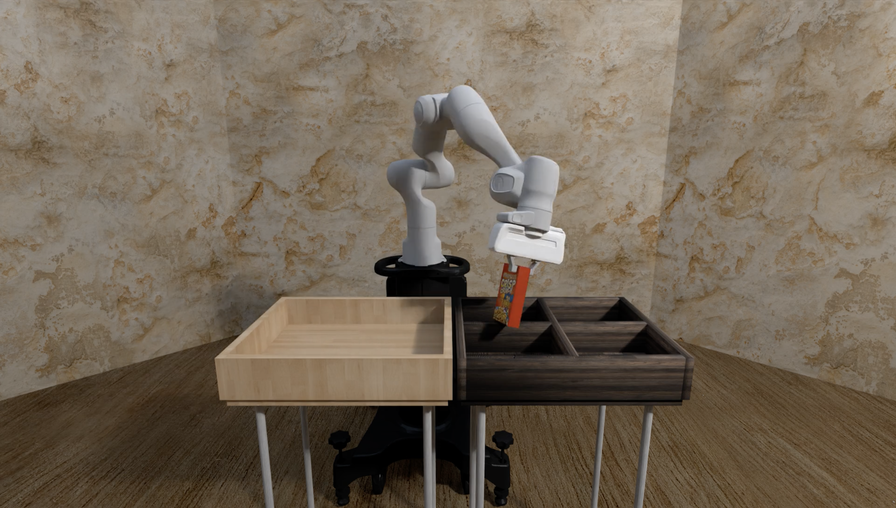}
    \caption{\small \texttt{RS-Cereal}}
\end{subfigure}
\begin{subfigure}[b]{0.24\linewidth}
    \includegraphics[width=\linewidth]{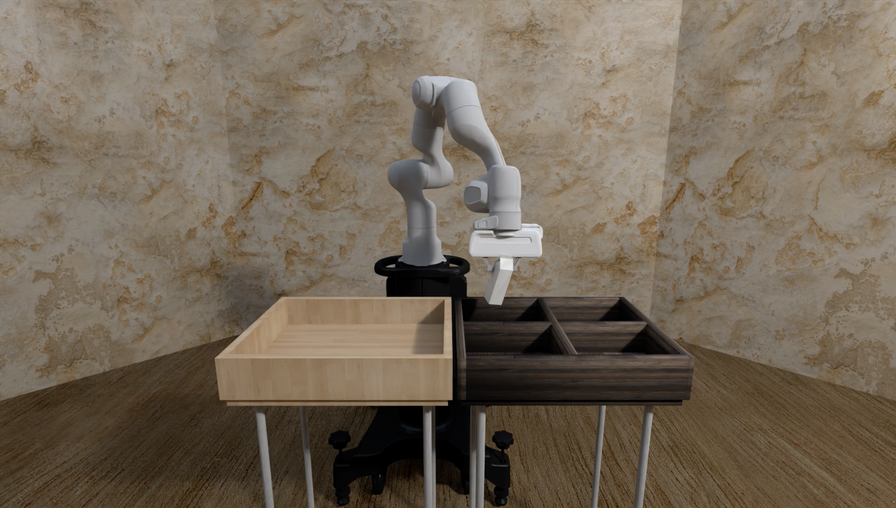}
    \caption{\small \texttt{RS-Milk}}
\end{subfigure}
\begin{subfigure}[b]{0.24\linewidth}
    \includegraphics[width=\linewidth]{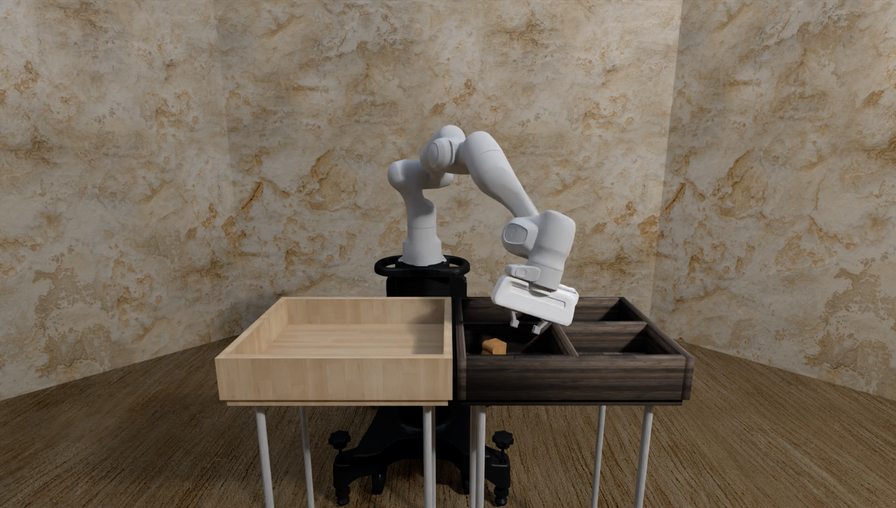}
    \caption{\small \texttt{RS-Bread}}
\end{subfigure}
\vspace{.2in}
\begin{subfigure}[b]{0.24\linewidth}
    \includegraphics[width=\linewidth]{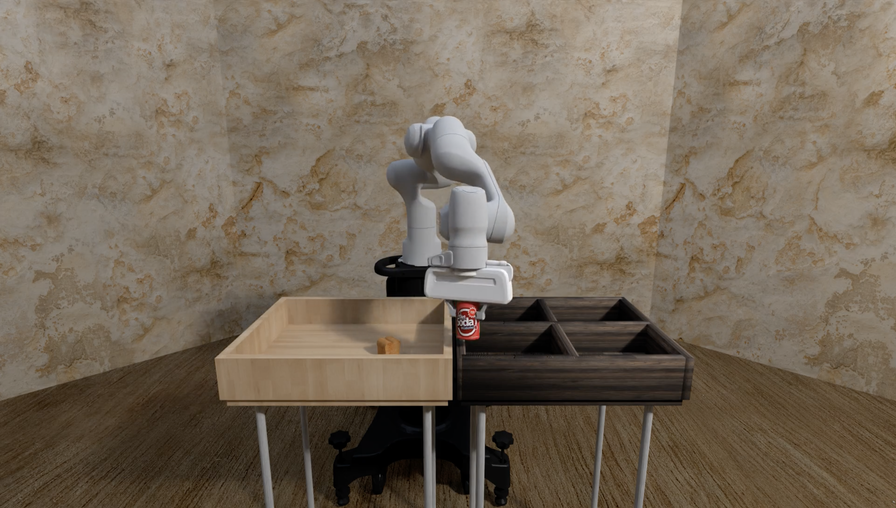}
    \caption{\small \texttt{RS-CanBread}}
\end{subfigure}
\begin{subfigure}[b]{0.24\linewidth}
    \includegraphics[width=\linewidth]{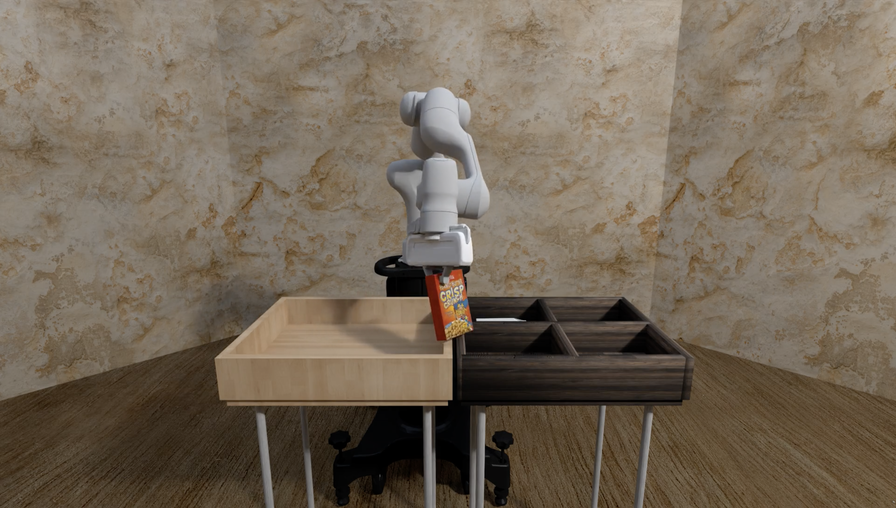}
    \caption{\small \texttt{RS-CerealMilk}}
\end{subfigure}
\vspace{.2in}
\caption{\small \textbf{Task Visualizations}. PSL\xspace is able to solve all tasks with at least 80\% success rate from purely visual input.}
\label{fig:task-list}
\end{figure*}

\clearpage
We discuss each of the environment suites that we evaluate using PSL\xspace. All environments are simulated using the MuJoCo simulator~\cite{todorov2012mujoco}.

\begin{enumerate}
    \item \textbf{Meta-World} (Row 1 of Fig.~\ref{fig:task-list}). Meta-World, introduced by~\citet{yu2020meta}, aims to offer a standardized suite for multi-task and meta-learning methods. The benchmark consists of 50 separate manipulation tasks with a Sawyer robot, well-shaped reward functions, involve manipulating a single object to a randomized goal position, or multiple objects to a deterministic goal position. We evaluate on the single-task, multi-goal, v2 variants of the Meta-World environments. All environments use end-effector position control - a 3DOF arm action space along with gripper control - orientation is fixed. In our evaluation we use the default environment task rewards, a fixed camera view for the baselines and a wrist camera for our local policies. We refer the reader to the Meta-World paper for additional details regarding the environment suite. 
    \item \textbf{Obstructed Suite} (Rows 1-2 of Fig.~\ref{fig:task-list}). The Obstructed Suite of tasks introduced by~\citet{yamada2021motion} are a challenging set of tasks requiring a Sawyer arm to perform obstacle avoidance while solving the task. The \texttt{OS-Lift} task requires the agent to pick up a can that is inside a tall box, requiring it to reach over the walls to grab the object and then lift it without making contact with the edges of the bin. The \texttt{OS-Push} environment tasks the agent with push a block to the goal in the present of a bin that forces the agent to adjust its motion in order to avoid being blocked by its upper joints. Finally, the \texttt{OS-Assembly} task involves moving the robot arm to a precise placement location while avoiding obstacles, then performing the table leg placement. Note that we evaluate our method on these environments from visual input, a more challenging setting than the one considered by~\citet{yamada2021motion}.
    \item \textbf{Kitchen} (Rows 2-3 of Fig.~\ref{fig:task-list}). The Kitchen manipulation suite introduced in the Relay Policy Learning paper~\cite{gupta2019relay} and maintained in D4RL~\cite{fu2020d4rl} is a set of challenging, sparse reward, joint-controlled manipulation tasks in a single kitchen. We modify the benchmark to use end-effector control as we find that this significantly improves learning performance. The tasks require the ability to explore efficiently whilst also being able to chain skills across long temporal horizons, to achieve behaviors such as opening the microwave, moving the kettle, flicking the light switch, turning the top left burner, and finally sliding the cabinet door (\texttt{K-MS-5}). Aside from the single-stage tasks described in Section~\ref{sec:setup}, we evaluate on three multi-stage tasks which require chaining the single-stage tasks in a particular order. \texttt{K-MS-3} involves moving the kettle, flicking the light switch and turning the top left burner, \texttt{K-MS-7} involves doing the same tasks as \texttt{K-MS-5} then turning the bottom right burner and opening the hinge cabinet and \texttt{K-MS-10} involves performing the tasks in \texttt{K-MS-7} and then opening the top right burner, opening the bottom left burner and then closing the microwave.
    \item \textbf{Robosuite} (Rows 3-6 of Fig.~\ref{fig:task-list}). The Robosuite benchmark from ~\citet{zhu2020robosuite} contains challenging, long-horizon manipulation tasks involving pick-place and nut assembly, as well as simpler tasks that involve lifting a cube and opening a door. The rewards are coarsely defined in terms of distances to targets as well as grasp/placement conditions, which, in fact, are straightforward to implement in the real world as well using pose estimation. This stands in contrast to Meta-World which spends considerable engineering effort defining well-shaped dense rewards often by taking advantage of object geometry. As a result, learning-based methods struggle to make any progress on Robosuite tasks that involve more than a single-stage - optimizing the reward function tends to leave the agent a local minima. The suite also contains a well-tuned, realistic Operation Space Control~\cite{khatib1987unified} implementation that we leverage to train policies in end-effector space.
\end{enumerate}

\clearpage

\section{LLM Prompts and Plans}
\label{app:llm prompts}
In this section, we list the LLM prompts per task.
\\
Overall prompt structure:
\begin{framed}
Stage termination conditions: (grasp, place, push, open, close, turn).
Task description: ... Give me a simple plan to solve the task using only the stage termination conditions. Make sure the plan follows the formatting specified below and make sure to take into account object geometry. Formatting of output: a list in which each element looks like: ($<$object/region$>$, $<$operator$>$). Don't output anything else.
\end{framed}

\subsection{Robosuite}

\texttt{RS-PickPlaceCan}:
\begin{framed}
\textbf{Task Description} can goes into bin 1.
\\
\textbf{Plan:} [(``can", ``grasp"), (``bin 1", ``place")])
\end{framed}

\texttt{RS-PickPlaceCereal}:
\begin{framed}
\textbf{Task Description:} cereal goes into bin 3.
\\
\textbf{Plan:} [(``cereal", ``grasp"), (``bin 3", ``place")])
\end{framed}

\texttt{RS-PickPlaceMilk}:
\begin{framed}
\textbf{Task Description}: milk goes into bin 2.
\\
\textbf{Plan:} [(``milk", ``grasp"), (``bin 2", ``place")])
\end{framed}

\texttt{RS-PickPlaceBread}:
\begin{framed}
\textbf{Task Description:} bread slice goes into bin 4. 
\\
\textbf{Plan:} [(``bread slice", ``grasp"), (``bin 4", "place")])
\end{framed}

\texttt{RS-PickPlaceCanBread}:
\begin{framed}
\textbf{Task Description:} can goes into bin 1, bread slice in bin 4.
\\
\textbf{Plan:} [(``can", ``grasp"), (``bin 1", ``place"), (``bread slice", "grasp"), (``bin 4", "place")])
\end{framed}

\texttt{RS-PickPlaceCerealMilk}:
\begin{framed}
\textbf{Task Description:} milk goes into in bin 2, cereal in bin 3. 
\\
\textbf{Plan:} [(``cereal", ``grasp"), (``bin 3", ``place"), (``milk", ``grasp"), (``bin 2", ``place")])
\end{framed}

\texttt{RS-NutAssembly:}
\begin{framed}
\textbf{Task Description:} The silver nut goes on the silver peg and the gold nut goes on the gold peg. 
\\
\textbf{Plan:} [(``silver nut", ``grasp"), (``silver peg", ``place"),(``gold nut", ``grasp"), (``gold peg", ``place")]
\end{framed}

\texttt{RS-NutAssemblySquare:}
\begin{framed}
\textbf{Task Description:} The gold nut goes on the gold peg.
\\
\textbf{Plan:} [(``gold nut", ``grasp"), (``gold peg", ``place")]
\end{framed}

\texttt{RS-NutAssemblyRound:}
\begin{framed}
\textbf{Task Description:} The silver nut goes on the silver peg.
\\
\textbf{Plan:} [(``silver nut", ``grasp"), (``silver peg", ``place")]
\end{framed}

\texttt{RS-Lift:}
\begin{framed}
\textbf{Task Description:} lift the red cube.
\\
\textbf{Plan:} [(``red cube", "grasp")]
\end{framed}

\texttt{RS-Door:}
\begin{framed}
\textbf{Task Description:} open the door.
\\
\textbf{Plan:} [(``door handle", ``grasp")]
\end{framed}

\subsection{Meta-World}

\texttt{MW-Assembly:}
\begin{framed}
\textbf{Task Description:} put the green wrench on the maroon peg.
\\
\textbf{Plan:} [(``green wrench'', ``grasp''), (``maroon peg'', ``place'')]
\end{framed}

\texttt{MW-Disassemble:}
\begin{framed}
\textbf{Task Description:} remove the green wrench from the peg.
\\
\textbf{Plan:} [(``green wrench'', ``grasp'')]
\end{framed}

\texttt{MW-Hammer:}
\begin{framed}
\textbf{Task Description:} use the red hammer to push in the nail.
\\
\textbf{Plan:} [(``red hammer'', ``grasp''), (``nail'', ``push'')]
\end{framed}

\texttt{MW-Bin-Picking:}
\begin{framed}
\textbf{Task Description:} move the cube in the red bin into the blue bin. 
\\
\textbf{Plan:} [(``cube in red bin'', ``grasp''), (``blue bin'', ``place'')]
\end{framed}

\subsection{Kitchen}
\texttt{Kitchen-Microwave:}
\begin{framed}
\textbf{Task Description:} pull the microwave door open.
\\
\textbf{Plan:} [(``microwave door handle'', ``open'')]
\end{framed}

\texttt{Kitchen-Slide}
\begin{framed}
\textbf{Task Description:} use the rightmost vertical bar to slide the door.
\\
\textbf{Plan:} [(``rightmost vertical bar'', ``open'')]
\end{framed}

\texttt{Kitchen-Light}
\begin{framed}
\textbf{Task Description:} use the round knob to flick the light switch. \\
\textbf{Plan:} [(``knob'', ``turn'')]
\end{framed}

\texttt{Kitchen-Burner}
\begin{framed}
\textbf{Task Description:} rotate the top left burner with the red tip. 
\\
\textbf{Plan:} [(``top left burner with the red tip'', ``turn'')]
\end{framed}

\texttt{Kitchen-Kettle}
\begin{framed}
\textbf{Task Description:} move the kettle forward.
\\
\textbf{Plan:} [(``kettle'', ``push'')]
\end{framed}

\subsection{Obstructed Suite}
\texttt{OS-Lift:}
\begin{framed}
\textbf{Task Description:} lift red can from wooden bin.
\\
\textbf{Plan:} [(``red can', ``grasp'')]
\end{framed}

\texttt{OS-Assembly:}
\begin{framed}
\textbf{Task Description:} move the table leg, which is already in your hand, into the empty hole.
\\
\textbf{Plan:} [(``empty hole', ``place'')]
\end{framed}

\texttt{OS-Push:}
\begin{framed}
\textbf{Task Description:} push the red block onto the green circle.
\\
\textbf{Plan:} [(``red block'', ``grasp'')]
\end{framed}

\chapter{Chapter 4 Appendix}

\section{Table of Contents}
\begin{itemize}
    \item \textbf{Additional Experimental Results} (Appendix~\ref{app:additional results}): Additional experimental results demonstrating OPTIMUS\xspace's effectiveness on more tasks and additional baselines.
    \item \textbf{Ablations} (Appendix~\ref{app:appendix ablations}): Ablations and analyses of OPTIMUS\xspace, demonstrating the effectiveness of our design decisions.
    \item \textbf{Environments} (Appendix~\ref{app:envs}): Description of all the environments we use in this work.
    \item \textbf{Agent Structure} (Appendix~\ref{app:agent structure}): details regarding the observation space and action space of the agent.
    \item \textbf{Experiment Details} (Appendix~\ref{app:optimus network details}): Full details on how OPTIMUS\xspace is implemented, specifically the hyper-parameters used for training and network architectures.
\end{itemize}

\clearpage
\section{Additional Experimental Results}
\label{app:additional results}

\textbf{OPTIMUS\xspace exhibits multi-task category control capabilities.}
We extend our multi-task results to the setting in which the task category can also vary by training a multi-task category model on a dataset of demonstrations from Pickplace, Shelf and Microwave. Across the tasks, the goal is implicitly communicated by the initial observation. In this setting, we use the same camera views across all tasks: the left/right shoulder views and the wrist camera. We build a dataset of 15K trajectories with 5 objects per task category and 1K demos per task. We include the results in Table~\ref{supp:multi_task_category}. Similar to our multitask results in the main text, we find that OPTIMUS\xspace is able to demonstrate multi-task category capabilities: a single Transformer is capable of learning to pick and place objects on a table, manipulate objects in a shelf, and open doors across a large set of objects at a success rate of 73\%. This experiment shows that OPTIMUS\xspace greatly outperforms the baselines on multi-task learning.
\begin{table}[h!]
\centering
\resizebox{.7\linewidth}{!}{
\begin{small}
\begin{tabular}{lcccc}
\toprule
\textbf{Dataset}                          & \textbf{BC-MLP} & \textbf{BC-RNN} & \textbf{BeT} & \textbf{OPTIMUS\xspace} \\
\midrule
PickPlace-Shelf-Microwave & 44     & 47     & 41  & \textbf{73}      \\
\bottomrule
\end{tabular}
\end{small}
}
\vspace{-.1in}
\caption{\small \textbf{Multi-task category results.}. By distilling TAMP demonstrations across three environments (PickPlace, Shelf, and Microwave), OPTIMUS\xspace is able to effectively manipulate a wide array of objects across diverse scenes, purely from image input.
}
\label{supp:multi_task_category}
\end{table}

\textbf{OPTIMUS\xspace can learn to adapt its behavior based on the scene configuration.} 
We evaluate OPTIMUS\xspace on two tasks that involve adapting the task plan based on the configuration of objects in the scene: StackAdapt and MicrowaveAdapt, and two that require adapting motions to randomized receptacle sizes: ShelfReceptacle and MicrowaveReceptacle. As shown in Table~\ref{table:task plan scene gen}, OPTIMUS\xspace is able to effectively leverage visual input to learn when additional stacking operations are needed (StackAdapt) or when the area in front of the microwave needs to be cleared (MicrowaveAdapt), achieving 96\% and 75\% respectively, compared to the best baseline (96\% and 40\%). Additionally, we demonstrate that OPTIMUS\xspace is able to effectively learn to generalize to unseen receptacle sizes with high success rates, achieves 80\% and 70\% on held out shelves and microwaves respectively. These results illustrate that OPTIMUS\xspace can distill  scene conditioned task plan adaptation and motion generalization across scene configurations from TAMP supervision.
\begin{table}[H]
\centering
\resizebox{.7\linewidth}{!}{
\begin{small}
\begin{tabular}{lcccc}
\toprule
\textbf{Dataset} & 
\begin{tabular}[c]{@{}c@{}}\textbf{BC-MLP}\end{tabular} & 
\begin{tabular}[c]{@{}c@{}}\textbf{BC-RNN}\end{tabular} &
\begin{tabular}[c]{@{}c@{}}\textbf{BeT}\end{tabular} &
\begin{tabular}[c]{@{}c@{}}\textbf{OPTIMUS\xspace}\end{tabular} \\ 
\midrule
StackAdapt & $\textbf{96}$ & $92$ & $81$ & $\textbf{96}$ \\
\rowcolor{Gray}
MicrowaveAdapt & $25$ & $40$ & $13$ & $\textbf{75}$ \\
ShelfReceptacle & $72$ & $71$ & $59$ & $\textbf{80}$  \\
\rowcolor{Gray}
MicrowaveReceptacle & $48$ & $55$ & $31$ & $\textbf{70}$\\
\bottomrule
\end{tabular}
\end{small}
}
\vspace{-.1in}
\caption{\small \textbf{Scene-based adaptation results}. OPTIMUS\xspace can learn to vary the task plan it executes based on the scene configuration \textit{(rows 1 and 2)} as well as adapt to unseen receptacles (\textit{rows 3 and 4}).} 

\label{table:task plan scene gen}
\end{table}

\textbf{OPTIMUS\xspace solves tasks that RL methods fail to make progress on.}
We perform a thorough comparison of OPTIMUS\xspace against modern deep RL methods across four benchmark tasks in Robosuite (Stack, PickPlaceCan, PickPlaceCereal, PickPlace), for which there exist dense rewards suitable for RL. We evaluate 3 algorithms: SAC~\cite{haarnoja2018soft}, a commonly used off-policy model free method, DRQ-v2~\cite{yarats2021mastering}, a state-of-the-art vision-based RL method, and MoDem~\cite{hansen2022MoDem}, an efficient visual model-based RL method. We train each RL method with up to 5 million samples with 5 seeds. We show the results in Table~\ref{supp:rl comparison}.
Across every task, the RL baselines struggle to learn the long-horizon behaviors, failing to achieve a greater than 10\% success rate on any given task. These environments pose a significant exploration challenge for RL agents, especially when trying to map high-dimensional observations such as images to low-level control actions.
\begin{table}[h!]
\centering
\resizebox{.7\linewidth}{!}{
\begin{small}
\begin{tabular}{lcccc}
\toprule
\textbf{Dataset}                & \textbf{SAC} & \textbf{Drq-v2} & \textbf{MoDem} & \textbf{OPTIMUS\xspace} \\
\midrule
Stack           & 0 & 6    & 3   & \textbf{100 }  \\
\rowcolor{Gray}
PickPlaceCan    & 0 & 10   & 0   & \textbf{100 }  \\
PickPlaceCereal & 0 & 5    & 0   & \textbf{100}   \\
\rowcolor{Gray}
PickPlace       & 0 & 0    & 0   & \textbf{90}   \\
\bottomrule
\end{tabular}
\end{small}
}
\vspace{-.1in}
\caption{\small \textbf{Comparison of OPTIMUS\xspace vs. RL methods}. OPTIMUS\xspace is able to solve each Robosuite task to a high success rate, while RL methods struggle to make progress due to exploration challenges.
}
\label{supp:rl comparison}
\end{table}

\textbf{OPTIMUS\xspace can outperform purely Transformer based architectures.} In this experiment, we integrate Transformer-in-Transformer~\cite{mao2022transformer}, a recently proposed Transformer architecture for control, into OPTIMUS\xspace and evaluate it across five tasks: Stack, Pickplace-1, Shelf-1, Microwave-1 and PickPlaceFour. We do so by modifying OPTIMUS\xspace to use the code released by the authors of~\cite{mao2022transformer} as the Transformer block. The default settings from~\cite{mao2022transformer} did not perform well on our tasks (20\% success rate on Stack), so we made the following modifications: We modify the backbone used in~\cite{mao2022transformer} by increasing the number of layers in the Vision Transformer backbone from 1 to 6, the number of heads from 1 to 4, the patch dimension from 84 to 19 (to obtain a 4x4 grid). With these settings we achieve 54\% success rate on Stack. We then perform one further modification: instead of using the class token output as the state representation in~\cite{mao2022transformer}, we reshape the tokens corresponding to each patch into 4x4 images and then pass them through a spatial softmax to obtain a keypoint representation of the image. Doing so improves the performance of Transformer-in-Transformer from 54\% to 86\% on the Stack task. We run Transformer-in-Transformer across all five tasks and include the results against OPTIMUS\xspace in Table~\ref{supp:transformer_in_transformer}. Across each task, we find that OPTIMUS\xspace is able to outperform Transformer-in-Transformer, with an average performance improvement of 16.8\%. One additional advantage of our architecture over the one proposed in~\cite{mao2022transformer} is that ours is 4-5x faster to execute. We hypothesize that a likely reason for this performance discrepancy is that on our visuomotor control tasks, ResNets~\cite{he2016deep} are a powerful inductive bias. They maintain spatial locality which allows the spatial softmax~\cite{levine2016end} to easily identify important key-points in the image.
\begin{table}[h!]
\centering
\resizebox{.7\linewidth}{!}{
\begin{small}
\begin{tabular}{lcc}
\toprule
\textbf{Dataset}              & \textbf{Transformer-in-Transformer} & \textbf{OPTIMUS\xspace} \\
\midrule
Stack         & 86                         & \textbf{100}     \\
\rowcolor{Gray}
PickPlace-1   & 82                         & \textbf{100}     \\
Shelf-1       & 73                         & \textbf{91}      \\
\rowcolor{Gray}
Microwave-1   & 67                         & \textbf{86}      \\
PickPlaceFour & 45                         & \textbf{60}      \\
\bottomrule
\end{tabular}
\end{small}
}
\vspace{-.1in}
\caption{\small \textbf{Comparison of OPTIMUS\xspace vs. Transformer-in-Transformer}. OPTIMUS\xspace is able to outperform purely Transformer based architectures such as Transformer-in-Transformer~\cite{mao2022transformer} by 16.8\% across 5 tasks, demonstrating that our architecture is well-suited to imitating TAMP data from visual input.
}
\label{supp:transformer_in_transformer}
\end{table}

We describe and empirically validate three advantages of the distilled policies over the TAMP system: 1) success rate improvement over the TAMP supervisor, 2) faster run-time, 3) operation from perceptual instead of state input. 

\textbf{OPTIMUS\xspace almost doubles the performance of the TAMP supervisor.} To evaluate TAMP, we execute 50 trials averaged over three random seeds on each single-task environment and record the performance in Table~\ref{supp:table:optimus v tamp}. We find that OPTIMUS\xspace is able to outperform the TAMP system by a wide margin, from 20\% on the easiest task, PickPlace, to 64\% on Microwave-1 and 44\% on the hardest task, PickPlaceFour. 
TAMP with joint space control has better performance on average than TAMP with task space control (52\% vs. 45\%), but still performs significantly worse than OPTIMUS\xspace (52\% vs. 87\%). We instead find that not all grasps execute perfectly every time, likely due to differences in simulation, planning and control schemes from the ACRONYM paper. 
As a result, we observe grasp execution failures and object slippage during placement motions. OPTIMUS\xspace avoids learning these failure cases by only distilling the successful trajectories, which enables it to successfully generalize to unseen configurations of the task.
\begin{table}[h!]
\centering
\resizebox{.7\linewidth}{!}{
\begin{small}
\begin{tabular}{lccc}
\toprule
\textbf{Dataset} & 
\begin{tabular}[c]{@{}c@{}}\textbf{TAMP-joint}\end{tabular} &
\begin{tabular}[c]{@{}c@{}}\textbf{TAMP-task}\end{tabular} &
\begin{tabular}[c]{@{}c@{}}\textbf{OPTIMUS\xspace}\end{tabular} \\ 
\midrule
PickPlace-1 & 82 & 82 &$\textbf{100}$ \\
\rowcolor{Gray}
PickPlaceTwo & 52 & 58 & $\textbf{96}$ \\
PickPlaceThree & 40 & 50 & $\textbf{91}$ \\
\rowcolor{Gray}
PickPlaceFour & 34 & 16 & $\textbf{60}$ \\
Shelf-1 & 58 & 44 & $\textbf{91}$\\
\rowcolor{Gray}
Microwave-1 &46 &22& $\textbf{86}$ \\
\midrule
Average & 52 & 45 & \textbf{87} \\
\bottomrule
\end{tabular}
\end{small}
}
\vspace{-.1in}
\caption{\small \textbf{Comparison of OPTIMUS\xspace vs. TAMP}. We plot percentage success on randomly chosen states from the environment. We find OPTIMUS\xspace greatly outperforms the TAMP supervisor, whether TAMP uses task space control or joint space.
}
\label{supp:table:optimus v tamp}
\end{table}

\textbf{OPTIMUS\xspace executes 5-7.5x faster than TAMP.} We evaluate the run-time of OPTIMUS\xspace against TAMP by computing the average time per step for both systems across 100 trials. We run the evaluation on a machine with an RTX 3090 GPU and Intel i9-10980XE CPU and include the results in Table~\ref{supp:table:timing}. TAMP takes 150ms per action on average while OPTIMUS\xspace (30M parameters) takes 21ms per action and OPTIMUS\xspace (100M parameters) takes 31ms per action. TAMP pays a high up-front cost of 2-5 seconds, and then executes a feedback controller to quickly track the planned way-points. In contrast, OPTIMUS\xspace spends a constant amount of time per action. Furthermore, it is possible to greatly improve the inference time performance of OPTIMUS\xspace by employing techniques such as FlashAttention~\cite{dao2022flashattention}, model compilation, and TensorRT.
\begin{table}[h!]
\centering
\resizebox{0.5\linewidth}{!}{
\begin{small}
\begin{tabular}{ccc}
\toprule
\begin{tabular}[c]{@{}c@{}}\textbf{TAMP}\end{tabular} & 
\begin{tabular}[c]{@{}c@{}}\textbf{OPTIMUS\xspace (30M)}\end{tabular} &
\begin{tabular}[c]{@{}c@{}}\textbf{OPTIMUS\xspace (100M)}\end{tabular} \\ 
\midrule
 $150$ms & $21$ms & $31$ms \\
\bottomrule
\end{tabular}
\end{small}
}
\vspace{-.1in}
\caption{\small \textbf{Timing Results.} We measure the average time taken per action (lower is better). On average, OPTIMUS\xspace is 5-7.5x faster to execute than TAMP.
}
\label{supp:table:timing}
\end{table}

\textbf{By distilling TAMP, we obtain a performant policy that executes high-frequency \textit{low-level} control from \textit{purely perceptual} input.} OPTIMUS\xspace produces policies that are fast to execute, reactive and perform visuomotor control at similar performance to policies that have access to state information (Fig.~\ref{fig:appendix ablations}) and out-performs the privileged TAMP expert (Table~\ref{supp:table:optimus v tamp}). 

\clearpage
\section{Ablations}
\label{app:appendix ablations}

In this section, we ablate components of OPTIMUS\xspace, low-level controller, data filtration scheme, gripper control scheme and data generation process, observation space design and loss function. 

\begin{figure}[h]
\begin{subfigure}[b]{0.55\textwidth}
    \includegraphics[width=\textwidth]{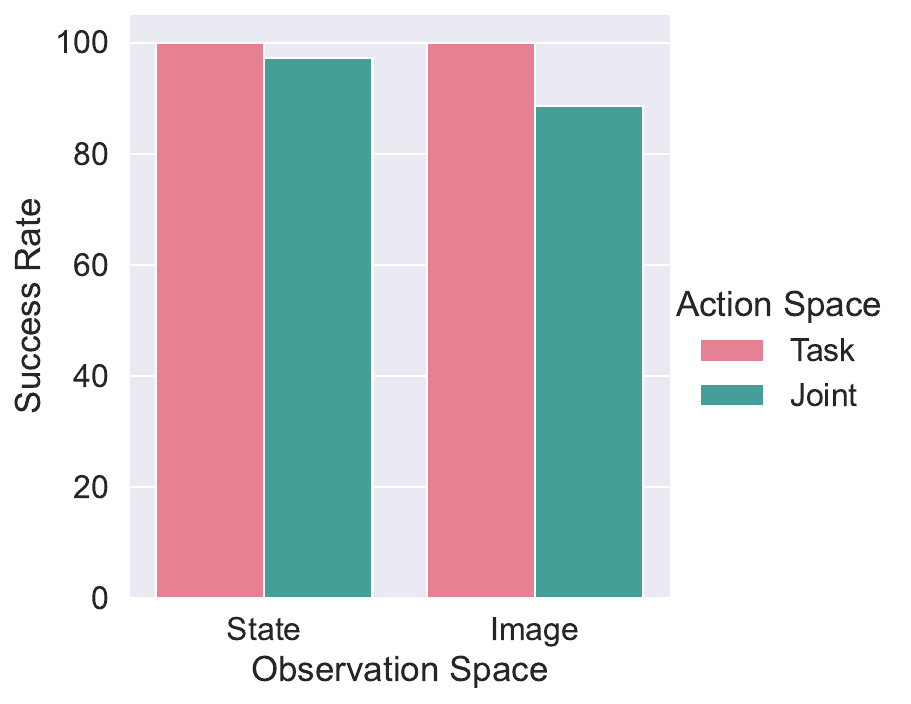}
    \caption{\small Arm Action Space}
\end{subfigure}
\begin{subfigure}[b]{0.43\textwidth}
    \includegraphics[width=\textwidth]{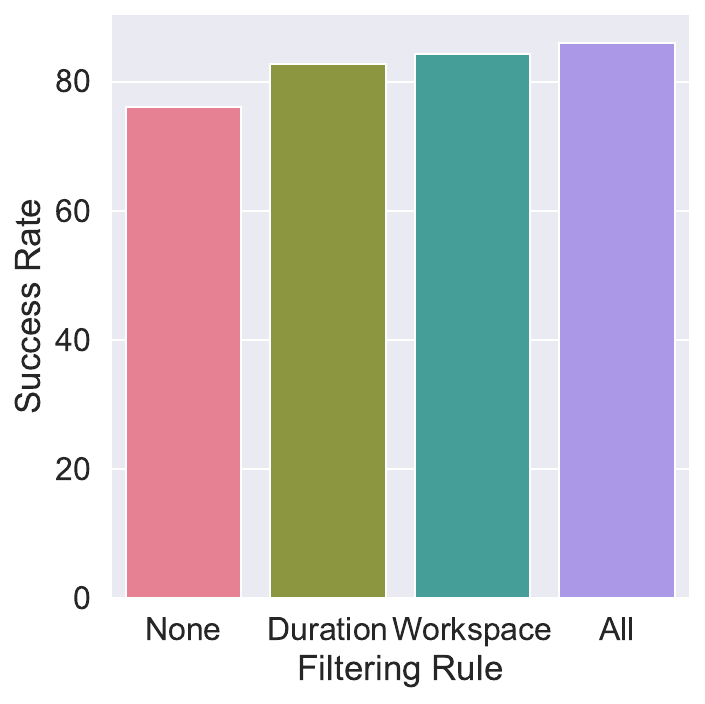}
    \caption{\small Data Curation}
\end{subfigure}
\caption{\small \textbf{Effect of Arm Action Space Choice and Data Filtering Rules.} (a) OPTIMUS\xspace task success rate improves with task-space over joint-space actions when using image observations. 
Image observation-space policies perform comparably to the privileged state-based policies when using task-space actions. 
(b) Performance is improved by filtering TAMP success trajectories based on the visible workspace and their duration.
}
\vspace{-8pt}
\label{fig:action and data curation ablations}
\end{figure}

\begin{figure*}[t]
\centering
\begin{subfigure}[b]{0.19\linewidth}
    \includegraphics[width=\linewidth]{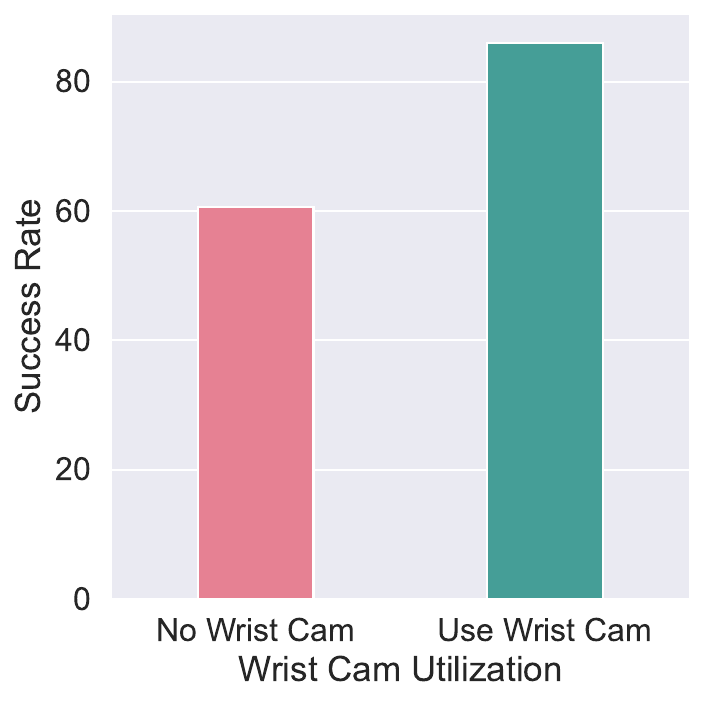}
    \caption{\small Wrist Camera}
\end{subfigure}
\begin{subfigure}[b]{0.19\linewidth}
    \includegraphics[width=\linewidth]{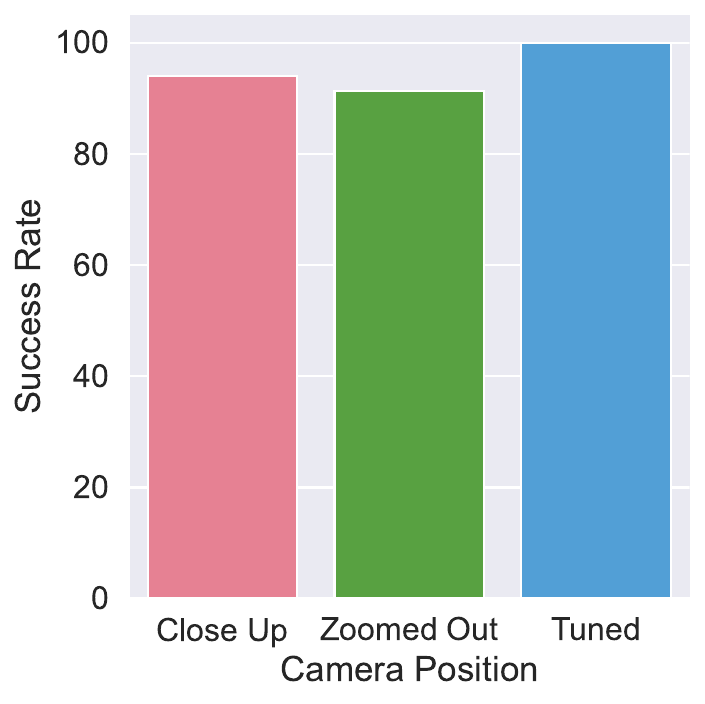}
    \caption{\small Camera Position}
\end{subfigure}
\begin{subfigure}[b]{0.19\linewidth}
    \includegraphics[width=\linewidth]{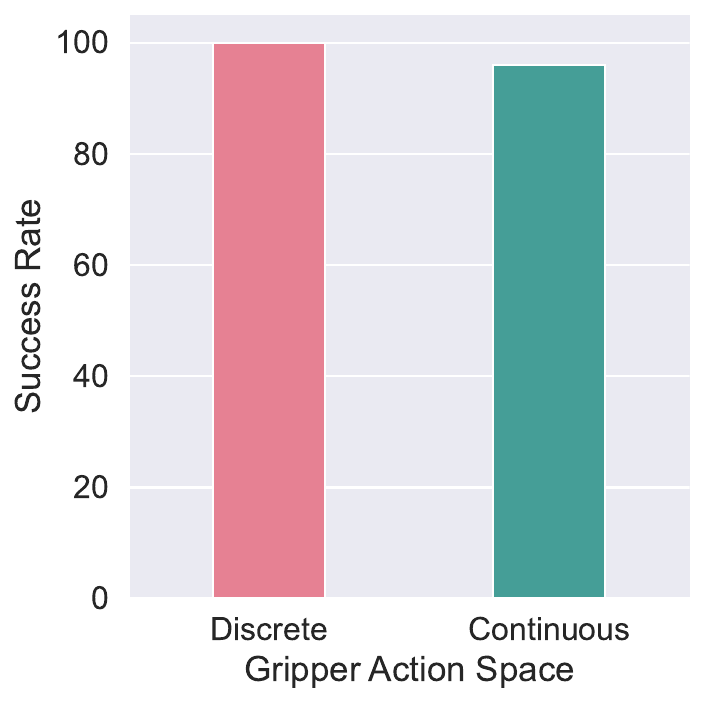}
    \caption{\small Gripper Control}
\end{subfigure}
\begin{subfigure}[b]{0.19\linewidth}
    \includegraphics[width=\linewidth]{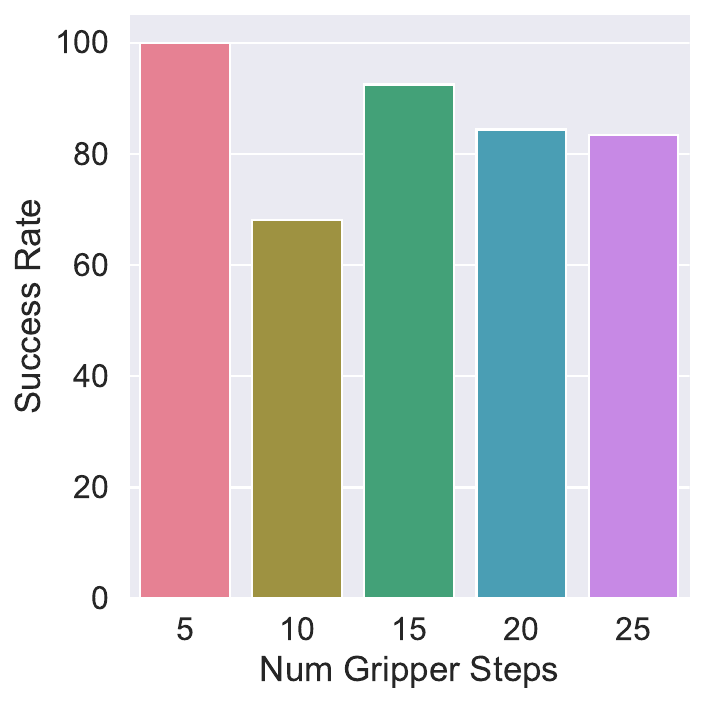}
    \caption{\small Gripper Steps}
\end{subfigure}
\begin{subfigure}[b]{0.19\linewidth}
    \includegraphics[width=\linewidth]{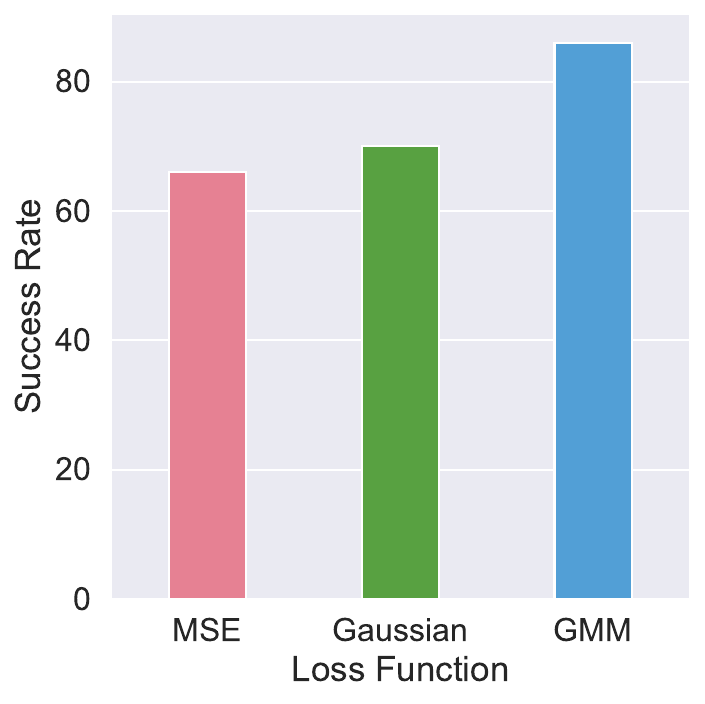}
    \caption{\small Loss}
\end{subfigure}
\caption{\small \textbf{Effect of Observation, Action and Loss Decisions.} We ablate a variety of design decisions in OPTIMUS\xspace and demonstrate that each produces a clear improvement.}
\label{fig:appendix ablations}
\end{figure*}

\textbf{Task-space control greatly improves visuomotor learning performance.}
We evaluate different controllers on the Microwave-1 task. For state-based learning, we find that the choice of action space makes little difference; both control schemes achieve high performance (98\% for joint space vs. 100\% for task space). However, when training with visual observations, we find that there is a large gap (86\% vs. 100\%) in performance between joint control and task-space control. We hypothesize that this is due to the difficulty of learning an inverse kinematics mapping from visual input, {\it i.e.} mapping 2D pixel locations to 7DOF joint angles.

\textbf{Data filtration results in a significant improvement in policy success rates.}
On the Microwave-1 task, we train four policies with different filtration schemes: 1) no filtering (None), 2) filtering based on trajectory length (Duration) 3) filtering based on visible workspace limits (Workspace), and 4) Duration and Workspace combined (Both). We find  (Fig.~\ref{fig:action and data curation ablations}) that policies trained on unfiltered data perform worse when compared to those trained on filtered data. Specifically, workspace filtering has a greater impact than Duration.
Combining both forms of filtering results in the greatest performance improvement of 10\% and demonstrates that filtering TAMP trajectories is crucial to obtaining high success rates for learned policies.

\textbf{Discrete gripper control and short "stall" regions directly impact the performance of TAMP imitation.}
We first analyze the impact of switching from continuous to discrete gripper control on the Stack task in Fig.~\ref{fig:appendix ablations}. By using discrete control, we can improve the success rate by 4\%, while qualitatively we observe smoother gripper control and decisive grasps. On the other hand, we find that the decision to tune the length of "stall" regions, namely TAMP grasp and release actions, is crucial to the performance of OPTIMUS\xspace. As observed in Fig.~\ref{fig:appendix ablations}, reducing the number of control actions per grasp and release action greatly improves performance, from 78\% at 25 steps to 100\% at 5 actions. This is likely due to two reasons, 1) we shorten the overall length of the roll-outs, easing the learning burden, and 2) we reduce the likelihood of the policy to encounter a series of states where the observations and actions do not change, which can result in freezing behavior in the policy.

\textbf{Camera view selection enables greatly improved visuomotor learning.} We evaluate two camera views on the Stack task. Both camera poses keep all objects as well as the robot in view; one is close up which hinders accurate estimation of scene geometry while the other is farther away which decreases the size of the objects in the frame, making it difficult for the policy to focus on them. As a result, we find in Fig.~\ref{fig:appendix ablations} that a well-tuned camera view that is angled and positioned appropriately performs best. We additionally evaluate the impact of using a wrist camera. For tasks with primitive objects such as blocks, we found that the wrist cam had little impact. However, moving to tasks such as Microwave, where close up views of the handle and target object enable improved perception of grasp geometries, the wrist camera affords a significant performance improvement as we show in Fig.~\ref{fig:appendix ablations}. 

\textbf{GMM loss enables OPTIMUS\xspace to better handle the multi-modality of TAMP supervision.}
TAMP generates highly multi-modal action distributions through randomized planning and non-deterministic IK. Therefore, as we note in Sec.~\ref{sec:imitation pipeline}, we use Gaussian Mixture Models to model the multi-modality. We experimentally validate that GMM output distributions greatly improve learning performance by comparing against MSE loss, which produces a deterministic, uni-modal output distribution, and Gaussian log-likelihood, which produces a non-deterministic, uni-modal output distribution. We find that GMM loss greatly out-performs both output distributions (86\% vs. 66\% and 70\%). While including a stochastic output distribution such as a Gaussian does improve performance by 4\%, the multi-modality of GMM produces a further improvement of 16\% performance. The results demonstrate that by providing the policy a more expressive output distribution, we can greatly improve how well the policy can model the TAMP expert.

\clearpage
\section{Environments}
\label{app:envs}
In this section, we provide a detailed description of the environments we use to evaluate OPTIMUS\xspace. We begin by describing settings which are common across environments. We then discuss each task individually. 

For all tasks, we use a Franka Panda 7-DOF manipulator with the default Franka gripper, though the TAMP system is capable of generating supervision using any manipulator, provided the robot URDF. For the Stack task, we use the block stacking environment from Robosuite~\cite{zhu2020robosuite}, modifying it to include up to 5 blocks and a larger workspace region. For all other tasks we use IsaacGym~\cite{makoviychuk2021isaac} with the PhysX~\cite{macklin2014unified} back-end. For each task, we use a fixed reset pose for the robot, while randomizing the positions of sampled objects. Object orientation about the z-axis is sampled uniformly at random from 0 to 360 degrees for all tasks.

For PickPlace, Multi-step PickPlace, Shelf and Microwave, we sample objects from ShapeNet~\cite{chang2015shapenet}. We select objects that have valid grasps in the Acronym~\cite{eppner2021acronym} dataset. We further refine our dataset by filtering out objects that do not simulate well in our IsaacGym environments. From the remaining objects, we form two datasets with 19 and 72 objects respectively. 

We next provide additional details for each task.

\textbf{Stack}: The goal is to stack the blocks in a fixed ordering. Each block is a different color. The block positions are sampled uniformly in an area of size 28cm x 28cm. The base block is of size 2.5$cm^3$; the rest are of size 2$cm^3$. The task is considered solved if all of the blocks are stacked in the correct ordering. 

\textbf{StackAdapt}: The task is the same as Stack, except there are two platforms, the blocks must be stacked on the target platform only. There is a 50/50 chance for the base block to be spawned on the target platform, in which the task simply involves stacking, and the base block to be spawned on the other platform, which requires the agent to first place the base block on the target platform then stack on top of it.

\textbf{PickPlace}: The task involves picking and placing ShapeNet objects from the left platform to the right platform. The platforms are of size .25 by .25 and are kept .5 apart. The object positions are sampled uniformly at random on the platform. The task success criteria is fulfilled if the object is placed anywhere on the target platform. 

\textbf{Multi-step PickPlace}: The task involves picking and placing ShapeNet objects from platforms on the left to bins on the right. Up to four objects: a basket, vase, magnet or cup are sampled on separate platforms. Each platform is of size .15x.15 and each bin is of size .2x.2m. Each object's position is sampled uniformly at random on its associated platform. The task is solved when all objects are in their associated bins. 

\textbf{Shelf}: The task involves moving ShapeNet objects from the lower rung of the shelf to the middle one. The shelf is 1m tall and has three rungs of size .5m x .25. The position and size of the shelf are constant. Object positions are sampled on the lowest rung, uniformly at random across the surface. The task is solved when the object is placed on the middle rung.

\textbf{ShelfReceptacle}: This task is the same as Shelf, but the shelf size is randomized within the following intervals: height (.8-1m), rungs: (.5x.25m - .4x.75m).

\textbf{Microwave}: The goal is to open the microwave by pulling open the handle, grasp a ShapeNet object, and place it inside the microwave. The microwave is .3m tall, 50cm wide and 20 cm deep. Microwave position and size are held fixed. The initial angle of the microwave door is 0, i.e. fully closed. Object positions are sampled on a platform of size .25x.25m. The agent has succeeded when the object is inside the microwave. 

\textbf{MicrowaveReceptacle}: This task is the same as Microwave, but the microwave size is randomized within the following intervals: height (.3-.4m), width: (.5-.6m), depth: (.2-.3m).

\textbf{MicrowaveAdapt}: The task is the same as the microwave task, except with 50\% probability an object is spawned in front of the microwave door, requiring the agent to first move the object aside then open the door and place the target object inside. 

\clearpage

\section{Agent Structure}
\label{app:agent structure}
\textbf{Observation spaces:} We use the same set of proprioceptive observations across all tasks: end-effector position, end-effector orientation (quaternion), gripper position. For each task, we select a different camera view that maximizes scene coverage. For Shelf and Microwave, we use two views, left and right shoulder views, whereas for the rest of the tasks we use a single forward facing view. Additionally, we use a wrist camera for every task, which greatly improves the performance. We use camera images of size 84x84. We empirically validate these decisions in Sec.~\ref{app:appendix ablations} and visualize the results in Fig.~\ref{fig:appendix ablations}.

\textbf{Action spaces:} As mentioned in the main text, we use task space control for moving the arm. In Robosuite, we use the built-in OSC controller~\cite{khatib1987unified}. In IsaacGym, we used a simple IK-based task-space controller. With regard to gripper control, we discuss and resolve two challenges related to TAMP. 1) Continuous gripper actions produced by the TAMP solver can be challenging for the network to fit, as the network does not fully commit to predicting grasps. To that end, we modify the gripper actions to be binary open and close motions which improves performance and reduces noise in policy execution. We validate that this results in a performance improvement in Appendix~\ref{app:appendix ablations}. 2) TAMP demonstrations can include``stall regions": segments of the trajectory in which the robot is not moving, such as when TAMP executes gripper-only actions for grasps and placements. This results in trained policies that may freeze after grasping an object, as the data does not contain cues for when to exit the stall region. To address this issue, we tune the length of stall regions during data collection against the agent's history length to ensure data collection success rate remains high while minimizing policy freezing behavior.

\clearpage
\section{Experiment Details}
\label{app:optimus network details}

\begin{table}[h!]
\caption{\small Hyper-parameters used during training.}
\resizebox{0.5\linewidth}{!}{
\centering
\begin{tabular}{@{}lc@{}}
\toprule
Hyper-parameter            & Value  \\ 
\midrule
Learning Rate             & 0.0001 \\
\rowcolor{Gray}
Batch Size & 16/512 \\
Warmup Steps              & 0     \\
\rowcolor{Gray}
Linear Scheduling Steps  & 100K    \\
Final Learning Rate & 0.00001 \\
\rowcolor{Gray}
Weight Decay              & 0.01      \\
Gradient Clip Threshold   & 1.0    \\ 
\rowcolor{Gray}
Number of Gradient Steps & 1M \\
Optimizer Type & AdamW \\
\rowcolor{Gray}
Loss Type & GMM \\
GMM Components & 5 \\
\rowcolor{Gray}
GMM Min. Std. Dev. & 0.0001 \\
GMM Std. Dev. Activation Fn. & SoftPlus \\
\bottomrule
\end{tabular}
}
\label{supp:table:train_hyperparams}
\end{table}
\begin{table*}[h!]
\caption{Model hyper-parameters.}
\resizebox{1\linewidth}{!}{
\centering
\begin{tabular}{@{}lcccc@{}}
\toprule
             &     OPTIMUS\xspace (30M/100M) & MLP (30M/100M) & RNN (30M/100M) & BeT (30M/100M) \\ 
\midrule
Num Layers     & 6/12  & 2/6 & 2/3 & 6/12 \\
\rowcolor{Gray}
Hidden Dimension & & 1024/1024 & 1000/2000 &  \\
Context Length & 8/8 & & 10/10 & 10/10 \\
\rowcolor{Gray}
Num Heads & 8/16  & & & 8/16 \\
Transformer Embed. Dim.   & 256/512 & & & 256/512\\ 
\rowcolor{Gray}
Embedding Dropout Prob. & 0.1/0.1 & & & 0.1/0.1\\
Attention Dropout Prob. & 0.1/0.1 & & & 0.1/0.1\\
\rowcolor{Gray}
Output Dropout Prob. & 0.1/0.1 & & & 0.1/0.1\\
Positional Embed. & Learned/Learned & & & Learned/Learned\\
\rowcolor{Gray}
Positional Embed. Type & Relative/Relative & & &  Relative/Relative\\
Num. Clusters & & & & 24/24 \\
\rowcolor{Gray}
Offset Loss Scale & & & & 100/100 \\
\bottomrule
\end{tabular}
}
\label{supp:table:transformer_hyperparams}
\end{table*}

\textbf{Network and Training Details:}
We include the model hyper-parameters for the 30M and 100M parameter variants of each method in Table~\ref{supp:table:transformer_hyperparams}. For the vision-backbone, as discussed in the main text, we use a Resnet-18~\cite{he2016deep} with a Spatial Softmax~\cite{levine2016end} output to encode each image separately. For details, please see the Robomimic paper~\cite{robomimic2021}. 
We include learned positional embeddings with each token and employ relative, rather than absolute, position embeddings to enable the network to adapt to longer horizons at test time. We use a linear annealing schedule that reduces the learning rate from $10^{-4}$ to $10^{-5}$ over 100K gradient steps and then keeps the learning rate constant. We train with the AdamW optimizer with a weight decay of $0.01$ and no learning rate warm-up. For single-task learning, we train with a batch size of $16$ on a single V100 GPU, while for multi-task learning we train using batch size of $512$ to $1024$ depending on the task, across 8 V100 GPUs. For visuomotor learning, we train with multiple camera views with image size 84x84, and we augment the data with random crops \citep{robomimic2021,laskin2020reinforcement,kostrikov2020image}. We additionally list the hyper-parameters used for training in Table~\ref{supp:table:train_hyperparams}. One note of interest: for multi-task training, we found that increasing the batch size greatly improved the results; hence we use a batch size of 512. 

For BeT, we tried using the original authors codebase, which we augmented with our vision backbone, but found that the performance was extremely low. Instead, we re-implemented BeT as a modification of OPTIMUS\xspace, using the same network structure but predicting a discrete cluster center and offset head instead and training using the focal and MT losses from the BeT paper. We found that the standard hyper-parameters for BeT did not perform well, and after significant hyper-parameter tuning found that the combination of 24 cluster centers and offset loss scale of 100 performed best.

\textbf{Evaluation Protocol:}
We note additional details regarding our evaluation protocol as follows. We split each dataset into a set of training and validation trajectories (using a 90/10 split). From the validation trajectories, we save the initial state of the demonstration. During evaluation, we reset the simulator state to an initial state from the validation set, and execute the policy from there. By comparing on the same set of validation states, we can better evaluate performance across seeds and algorithms. Note this means evaluation is performed from states that the TAMP solver is able to solve. As we note in Sec.~\ref{subsec:learning results}, in practice this distinction matters little, as the TAMP system does not have a systematic failure case which could be passed on to the policy. Therefore we observe similar success rates when evaluating on randomly sample poses from the environment. 

\clearpage

\chapter{Chapter 5 Appendix}
\section{Table of Contents}
\begin{itemize}
    \item \textbf{Additional Real World Experiment Analysis} (Appendix~\ref{app:analysis}): Additional analysis of our real world results, with more detailed experiments.
    \item \textbf{Ablations} (Appendix~\ref{app:ablations}): Ablations and analyses of Neural MP, demonstrating the effectiveness of our design decisions.
    \item \textbf{Procedural Scene Generation Details} (Appendix~\ref{app:procgen}): Description of all the environments we use in this work.
    \item \textbf{Network Training Details} (Appendix~\ref{app:network details}): Full details on how Neural MP is trained, specifically the hyper-parameters used for training and network architectures.
    \item \textbf{Real World Task Details} (Appendix~\ref{app:real world}): Details regarding the real world setup and analysis.
    
\end{itemize}

\clearpage
\section{Additional Real World Results and Analysis}
\label{app:analysis}

\subsection{Detailed Free Hand Motion Planning Results}
In this section we perform additional analysis of the free hand motion planning results from the main paper. We include a more detailed version of the main result table (Tab.~\ref{tab:main detailed results}). In this table, we additionally include the average (open loop) planning time per method and the average rate of safety violations. Safety violations are defined to occur where there are collisions, the robot hits its joint limits or there are torque limit errors. The open loop planning time for neural methods such as ours or M$\pi$Nets involves simply measuring the total time taken for rolling out the policy and test time optimization (TTO). We find that sampling-based planners in general never collide when executed. If they produce a safety violation, it is only because they find a trajectory that is infeasible for the robot to execute on the hardware, due to joint or torque limit errors. Neural motion planning methods have much higher collision rates, though Neural MP has a significantly lower collision rate than M$\pi$Nets, which we attribute to test-time optimization pruning out bad trajectories. We also note that not all collisions are created equal: some are slight, lightly grazing the environment objects while still achieving the goal, while others can be catastrophic, colliding heavily into the environment. In general, we found that our method tends to produce trajectories that may have slight collisions, though most of these are pruned out by TTO. With regards to planning time, M$\pi$Nets is the fastest method, as our method expends additional compute rolling out 100x more trajectories and then selecting the best one using SDF-based collision checking. 

\begin{table}[h!]
    \resizebox{\linewidth}{!}{%
        \begin{tabular}{lcccccc}
        \toprule
         & Bins ($\uparrow$) & Shelf ($\uparrow$) & Articulated ($\uparrow$) & Avg. Success Rate ($\uparrow$) & Avg. Planning Time ($\downarrow$) & Avg. Safety Viol. Rate ($\downarrow$)\\
        
        \multicolumn{2}{l}{\textit{Sampling-based Planning}:}\vspace{0.4em}\\
        \texttt{AIT*-80s~\cite{strub2020adaptively}} & 93.75 & 75 & 50.0 & 72.92 & 80 & \textbf{0}\\
        \texttt{AIT*-10s (fast)~\cite{strub2020adaptively}} & 75.0 & 37.5 & 25.0 & 45.83 & 10 & 2.1 \\
        \midrule
        
        \multicolumn{2}{l}{\textit{Neural}:}\vspace{0.4em}\\
        \texttt{M$\pi$Nets~\cite{fishman2023motion}} & 18.75 & 25.0 & 6.25 & 16.67 & \textbf{1.0} & 18.75  \\
        \midrule
        \textbf{\texttt{Ours}} & \textbf{100} & \textbf{100} & \textbf{87.5} & \textbf{95.83} & 3.9 & 4.2  \\ 
        \bottomrule
        \end{tabular}
        }
        \caption{\small 
        Neural MP performs best across tasks for free-hand motion planning, demonstrating greater improvement as the task complexity grows.}
        \label{tab:main detailed results}
\end{table}

\subsection{Detailed In-hand Motion Planning Results}
In this section, we extend the in-hand results shown in the main paper with additional baselines (AIT*-80s, AIT*-10s and M$\pi$Nets). For this evaluation (see Tab.~\ref{tab:detailed in hand results}, we consider two of the four in-hand motion planning objects, namely joystick and book. We find sampling-based methods are able to perform in-hand motion planning quite well, matching the performance of our base policy as well as our method without Objaverse data. We also see that M$\pi$Nets is unable to perform in-hand motion planning on any of the evaluated tasks. This is likely because that network was not trained on data with objects in-hand, demonstrating the importance of including in-hand data when training neural motion planners. Finally, there is a significant gap in performance between our method with and without test-time optimization; pruning out colliding trajectories at test time is crucial for achieving high success rates on motion planning tasks. 

\begin{table}[h!]
    \resizebox{\linewidth}{!}{%
        \begin{tabular}{lccccc}
        \toprule
        & Book ($\uparrow$) & Joystick ($\uparrow$) & Avg. Success Rate ($\uparrow$) & Avg. Planning Time ($\downarrow$) & Avg. Safety Viol. Rate ($\downarrow$)\\
        \multicolumn{2}{l}{\textit{Sampling-based Planning}:}\vspace{0.4em}\\
        \texttt{AIT*-80s~\cite{strub2020adaptively}} & 50 & 50		& 50	& 80	& \textbf{0}\\
        \texttt{AIT*-10s (fast)~\cite{strub2020adaptively}} 	& 25 & 50	& 37.5& 	10& 	\textbf{0} \\
        \midrule
        
        \multicolumn{2}{l}{\textit{Neural}:}\vspace{0.4em}\\
        \texttt{M$\pi$Nets~\cite{fishman2023motion}} & 0& 	0& 	0	& 1	& 37.5 \\
        \midrule
        \multicolumn{2}{l}{\textit{Ours}:}\vspace{0.4em} \\
        \texttt{Ours} (no TTO) 	& 25 & \textbf{75}	& 50	& \textbf{0.9} & 	50  \\ 
        \texttt{Ours} (no Objaverse) & 50 & 	50	& 50	& 3.9	& 50  \\ 
        \textbf{\texttt{Ours}} 	& \textbf{100} & \textbf{75}	& \textbf{87.5}	& 3.9 & 	12.5  \\ 
        \bottomrule
        \end{tabular}
        }
        \caption{\small 
        Neural MP performs best across tasks for in-hand motion planning, demonstrating greater improvement as the in-hand object becomes more challenging.}
        \label{tab:detailed in hand results}
\end{table}

\subsection{Test-time Optimization Analysis}

\begin{figure}[h!]
\centering
\includegraphics[width=\linewidth]{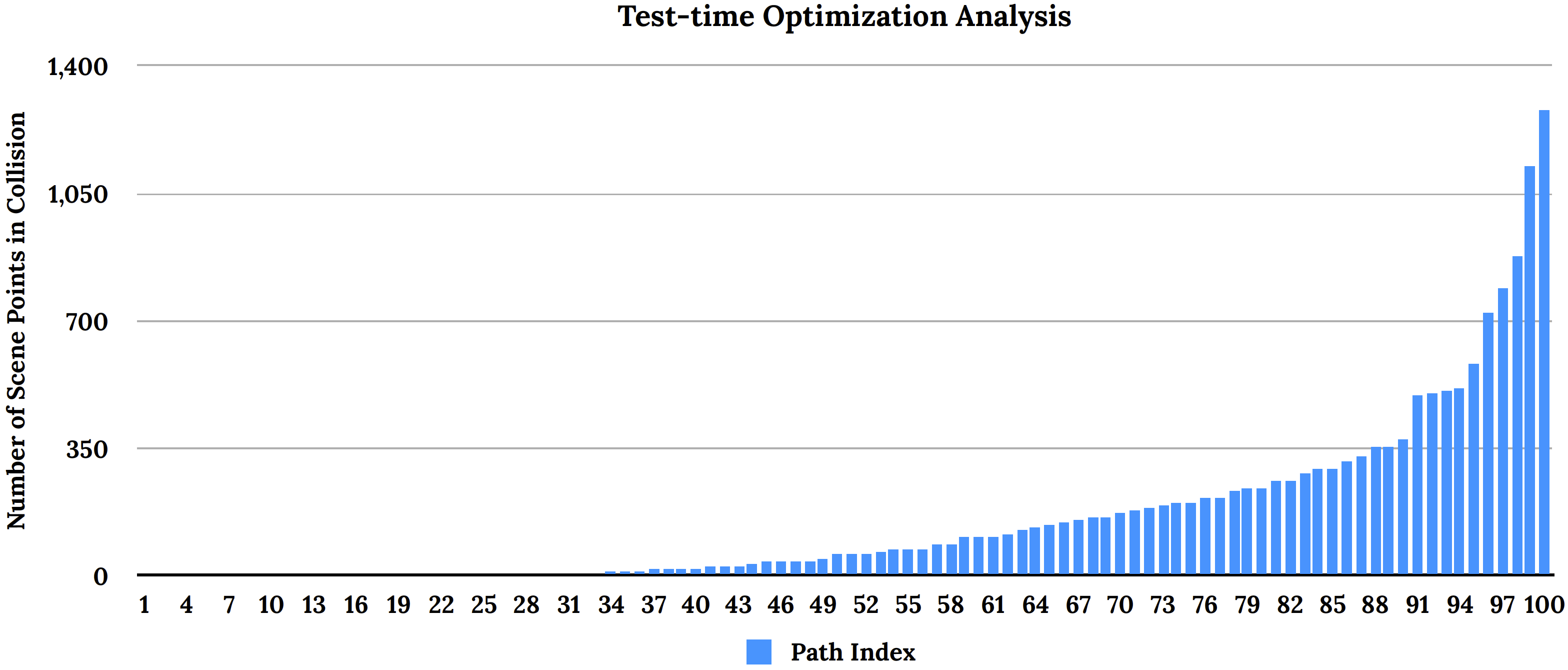}
\caption{\small \textbf{Test-time Optimization Analysis} For the Bins Scene 1 task, we plot the number of points in collision across 100 sampled trajectories from the model. 25\% of the trajectories are completely collision free and we select a trajectory execute from that subset.}
\label{fig:tto_analysis}
\end{figure}

To analyze what the test-time optimization procedure is doing, we first note that the base policy can sometimes produce slight collisions with the environment due to the imprecision of regression. As a result, when sampling from the policy, it is often likely that the policy will lightly graze objects which will count as failures when motion planning. We visualize a set of trajectories sampled from the policy here on our website for the real-world bins task. Observe that for some of the trajectories, the policy slightly intersects with the bin which would cause it to fail when executing in the real world, while for others it simply passes over the bin completely without colliding. We estimate the robot-scene intersection of all of these trajectories by comparing the robot SDF to the scene point-cloud and plot the range of values in Fig.~\ref{fig:tto_analysis}. We observe that 25\% of trajectories do not collide with the environment, and we select for those. In principle, one could further optimize by selecting the trajectory that is furthest from the scene (using the SDF). In practice, we did not find this necessary and that selecting the first trajectory among those with the fewest expected collisions performed quite well in our experiments.

\clearpage
\section{Ablations}
\label{app:ablations}

\begin{figure}[h!]
    \centering
    \includegraphics[height=50pt,width=\linewidth]{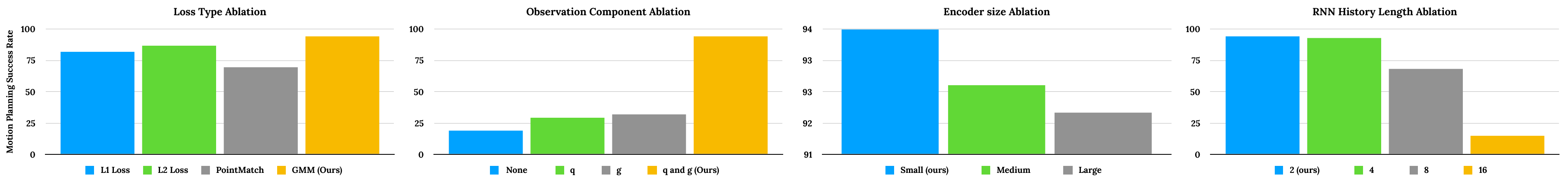}
    \caption{\small \textbf{Ablation Results} We evaluate four different components of Neural MP, loss type (\textit{left}), observation components (\textit{middle left}), encoder sizes (\textit{middle right}), and RNN history length (\textit{right}). We validate that our design decisions produce measurable improvements in motion planning success rates. }
    \label{fig:ablations}
\end{figure}
We run additional ablations analyzing components of our method in simulation using a subset of our dataset (100K trajectories) and include additional details for experiments discussed in the main paper.

\textbf{Loss Types}
For training objective, we evaluate 4 different options: GMM log likelihood (ours), MSE loss, L1 loss, and PointMatch loss (M$\pi$Nets). PointMatch loss involves computing the l2 distance between the goal and the predicted end-effector pose using 1024 key-points. We plot the results on held out scenes in Fig.~\ref{fig:ablations}. We find that GMM (ours) outperforms L2 loss, L1 loss, and PointMatch Loss (M$\pi$Nets) by (7\%, 12\%, and 24\%) respectively. One reason this may be the case is that sampling-based motion planners produce highly multi-modal trajectories: they can output entirely different trajectories for the same start and goal pair when sampled multiple times. Since Gaussian Mixture Models are generally more capable of capturing multi-modal distributions, they can hence fit our dataset well. At the same time, the PointMatch~\cite{fishman2023motion} loss struggles significantly on our data: it cannot distinguish between 0 and 180 degree flipped end-effector orientations, resulting in many failures due to incorrect end-effector orientations.

\textbf{Observation Components}
We evaluate whether our choice of observation components impacts the Neural MP's performance. In theory, the network should be able to learn as well from the point-cloud alone as when the proprioception is included, as the point-cloud contains a densely sampled point-cloud of the current and goal robot configurations. However, in practice, we find that this is not the case. Instead, removing either $q$ or $g$ or both severely harms performance as seen in Fig.~\ref{fig:ablations}. We hypothesize that including the proprioception provides a richer signal for the correct delta action to take.

\textbf{RNN History Length}
In our experiments, we chose a history length of 2 for the RNN, after sweeping over values of 2, 4, 8, 16 based on performance. From Fig.~\ref{fig:ablations} we see history length 2 achieves the best performance at 94\%, while using lengths 4, 8 and 16 achieve progressively decreasing success rates (92.67, 68, 14.67). One possible reason for this is that since point-clouds are already very dense representations that cover the scene quite well, the partial observability during training time is fairly low. A shorter history length also leads to faster training, due to smaller batches and fewer RNN unrolling steps.

\textbf{Encoder Size}
Finally, we briefly evaluate whether encoder size is important when training large-scale neural motion planners. We train 3 different size models: small (4M params), medium (8M params) and large (16M params). From the results in Fig.~\ref{fig:ablations}, we find that the encoder size does not affect performance by a significant margin (94\%, 93\%, 92\%) respectively and that the smallest model in fact performs best. Based on these results, we opt to use the small, 4M param model in our experiments.

\begin{table}[h]
\centering
\scriptsize
\resizebox{\linewidth}{!}{%
\begin{tabular}{@{}cccc@{}}
\toprule
\textbf{Neural MP-MLP} & \textbf{Neural MP-LSTM} & \textbf{Neural MP-Transformer} & \textbf{Neural MP-ACT} \\
\midrule 
65.0 & 82.5 & \textbf{85.0} & 47.5 \\
\bottomrule
\end{tabular}
}
\vspace{5pt}
\caption{\small Ablation of different architecture choices for the action decoder. We find that LSTMs and Transformers comparably while LSTMs boast faster inference times.}
\label{tab:arch abl}
\vspace{-10pt}
\end{table}

\textbf{Architecture Ablation}
In this experiment, we evaluate how different sequence modelling methods (Transformers and ACT~\cite{zhao2023learning}) and simpler action decoders such as MLPs compare against our design choice of using an LSTM. All methods are trained with the same dataset (of 1M trajectories), with the same encoder and GMM output distribution (with the exception of ACT which uses an L1 loss as per the ACT paper). We then evaluate them on held out motion planning tasks (Fig.~\ref{tab:arch abl} which are replicas of our real-world tasks (Bins and Shelf). We note several findings: 1) ACT performs poorly, largely due to its design choice of using an L1 loss which prevents it from handling planner multi-modality effectively, 2) Neural MP with an MLP action decoder also performs significantly worse than LSTMs and Transformers, as it is unable to use history information effectively to reason about the next action 3) Transformers and LSTMs perform similarly, with the Transformer variant performing marginally better, but with significantly slower inference time (2x). Hence we opt to use LSTM policies for our experimental evaluation, but certainly our method is amenable to any choice of sequence modeling architecture that performs well and has fast inference.

\begin{table}[h]
\centering
\scriptsize
\begin{tabular}{@{}ccc@{}}
\toprule
\textbf{Neural MP-MotionBenchMaker} & \textbf{Neural MP-M$\pi$Nets} & \textbf{Neural MP} \\
\midrule 
0 & 32.5 & 82.5 \\
\bottomrule
\end{tabular}
\vspace{5pt}
\caption{\small Comparing different methods for generating datasets for motion planning. We find that policies trained on our data generalize best to held out scenes.}
\label{tab:datagen abl}
\vspace{-10pt}
\end{table}
\textbf{Dataset Ablation}
Finally, we evaluate the quality of different dataset generation approaches for producing generalist neural motion planners. We do so by training policies on three different datasets (Neural MP, M$\pi$Nets~\cite{fishman2023motion}, and MotionBenchMaker~\cite{chamzas2021motionbenchmaker}) and evaluated on held out motion planning tasks in simulation. We train each model to convergence for 10K epochs and then execute trajectories on two held out tasks that mirror our real world tasks: RealBins and RealShelf. For fairness, we do not include any Objaverse meshes in these tasks, since MPiNets and MotionBenchMaker only have primitive objects. Still, we find that our dataset performs best by a wide margin (Tab.~\ref{tab:datagen abl}).
In general, we found that policies trained on MotionBenchMaker do not generalize well. As mentioned in the related works section, this dataset lacks the realism and diversity necessary to train policies that can generalize to held out motion planning scenes. 

\section{Procedural Scene Generation Details}
\label{app:procgen}

In this section we provide additional details regarding the data generation methods we develop for training large scale neural motion planners. 

\subsection{Procedural Scene Generation}

We formalize our procedural scene generation as a composition of randomly generated parameteric assets and sampled Objaverse meshes in Alg.~\ref{alg:scene_gen}

\textbf{Objaverse sampling details} 
The Objaverse are sampled in the task-relevant sampling location of the programmatic asset(s) in the scene, such as between shelf rungs, inside cubbies or within cabinets. Similar to the programmatic assets, these Objaverse assets are also sampled from a category generator $X_{obj}(\textbf{p})$. Here the parameter $p$ specifies the size, position, orientation of the object as well as task-relevant sampling location of the object in the scene, such as between shelf rungs, inside cubbies or within cabinets. 
As discussed in the main paper, we propose an approach that iteratively adds assets to a scene by adjusting their position using the effective collision normal vector, computed from the existing assets in the scene. We detail the steps for doing this in Alg.~\ref{alg:scene_gen}.

\subsection{Motion Planner Experts}
We use three techniques to improve the data generation throughput when imitating motion planners at scale.

\textbf{Hindsight Relabeling}
Tight-space to tight-space problems are the most challenging, particularly for sampling-based planners, often requiring significant planning time (up to 120 seconds) for the planner to find a solution. 
For some problems, the expert planner is unable to find an exact solution and instead produces approximate solutions. Instead of discarding these, note that we use a goal-conditioned imitation learning framework, where we can simply execute the trajectories in simulation and relabel the observed final state as the new goal.

\textbf{Reversibility}
We further improve our data generation throughput by observing that since motion planners inherently produce collision-free paths, the process is reversible, at least in simulation. This allows us to double our data throughput by reversing expert trajectories and re-calculating delta actions accordingly. Additionally, for a neural motion planner to be useful for practical manipulation tasks, it must be able to generate collision free plans for the robot even when it is holding objects. 
To enable such functionality, we augment our data generation process with trajectories 
where objects are spawned between the grippers of the robot end effector. There are transformed along with the end-effector during planning in simulation. We consider the object as part of the robot for collision checking and for the sake of our visual observations. 
In order to handle diverse objects that the robot might have to move with at inference time, we perform significant randomization of the in-hand object that we spawn in simulation. Specifically, we sample this object from the primitive categories of boxes, cylinders or spheres, or even from Objaverse meshes of everyday articles. We randomize the scale of the object between 3 and 30 cm along the longest dimension, and sample random starting locations within a 5cm cube around the end-effector mid-point between grippers.

\textbf{Smoothing}
Importantly, we found that naively imitating the output of the planner performs poorly in practice as the planner output is not well suited for learning. Specifically, plans produced by AIT* often result in way-points that are far apart, creating large action jumps and sparse data coverage, making it difficult to for networks to fit the data. To address this issue, we perform smoothing using cubic spline interpolation while enforcing velocity and acceleration limits. The implementation from M$\pi$Nets performs well in practice, smoothing to a fixed 50 timesteps with a max spacing of 0.1 radians. In general, we found that smoothing is crucial for learning performance as it ensures the maximum action size is small and thus easier for the network to fit to.

\subsection{Data Pipeline Parameters and Compute}
In Table~\ref{tab:datagen params}, we provide a detailed list of all the parameters used in generating the data to train our model. 

\textbf{Compute}
In order to collect a vast data of motion planning trajectories, we parallelize data collection across a cluster of 2K CPUs. It takes approximately 3.5 days to collect 1M trajectories. 

\section{Network Training Details}
\label{app:network details}

We first describe additional details regarding our neural policy, and then discuss how it is trained. 
Following the design decisions of M$\pi$Nets~\cite{fishman2023motion}, we construct a segmented point-cloud for the robot, consisting of the robot point-cloud, the target goal robot point-cloud and the obstacle point-cloud. Here we note two key differences from M$\pi$Nets: 1) our network conditioned on the target joint angles, while M$\pi$Nets only does so through the segmented point-cloud, 2) we condition on the target joint angles, not end-effector pose, decisions that we found improved adherence to the overall target configuration. For in-hand motion planning, we extend this representation by considering the object in-hand as part of the robot for the purpose of segmentation. 

We include a hyper-parameter list for our neural motion planner in Table~\ref{tab:model hyper params}. We train a 20M parameter neural network across our dataset of 1M trajectories. The PointNet++ encoder is 4M parameters and outputs an embedding of dimension 1024. We concatenate this embedding with the encoded $q_t$ and $g$ vectors and pass this into the 16M parameter LSTM decoder. The decoder outputs weights, means, and standard deviations of the 5 GMM modes. We then train the model with negative log likelihood loss for 4.5M gradient steps, which takes 2 days on a 4090 GPU with batch size of 16.

\clearpage
\onecolumn
\begin{longtable}{l|l}

\centering

\textbf{Hyper-parameter} & \textbf{Value} \\ \hline
\endfirsthead

\textbf{Hyper-parameter} & \textbf{Value} \\ \hline
\endhead
\multicolumn{2}{c}{\textbf{General Motion Planning Parameters}} \\ \hline
collision checking distance & 1cm \\
\rowcolor{Gray}
tight space configuration ratio & 50\% \\ 
dataset size & 1M trajectories \\ 
\rowcolor{Gray}
minimum motion planning time & 20s \\ 
maximum motion planning time & 80s \\
\hline
\multicolumn{2}{c}{\textbf{General Obstacle Parameters}} \\ \hline
in hand object ratio & 0.5 \\
\rowcolor{Gray}
in hand object size range & [[0.03, 0.03, 0.03], [0.3, 0.3, 0.3]] \\ 
in hand object xyz range & [[-0.05, -0.05, 0.], [0.05, 0.05, 0.05]] \\ 
\rowcolor{Gray}
min obstacle size & 0.1 \\ 
max obstacle size & 0.3 \\ 
\rowcolor{Gray}
table dim ranges & [[0.6, 1], [1.0, 1.5], [0.05, 0.15]] \\ 
table height range & [-0.3, 0.3] \\ 
\rowcolor{Gray}
num shelves range & [0, 3] \\ 
num open boxes range & [0, 3] \\ 
\rowcolor{Gray}
num cubbys range & [0, 1] \\ 
num microwaves range & [0, 3] \\ 
\rowcolor{Gray}
num dishwashers range & [0, 3] \\ 
num cabinets range & [0, 3] \\ 
\hline
\multicolumn{2}{c}{\textbf{Objaverse Mesh Parameters}} \\ \hline
scale range & [0.2, 0.4] \\ 
\rowcolor{Gray}
x pos range & [0.2, 0.4] \\ 
y pos range & [-0.4, 0.4] \\ 
\rowcolor{Gray}
number of mesh objects per programmatic asset & [0, 3] \\ 
number of mesh objects on the table & [0, 5] \\ \hline
\multicolumn{2}{c}{\textbf{Table Parameters}} \\ \hline
width range & [0.8, 1.2] \\ 
\rowcolor{Gray}
depth range & [0.4, 0.6] \\ 
height range & [0.35, 0.5] \\ 
\rowcolor{Gray}
thickness range & [0.03, 0.07] \\ 
leg thickness range & [0.03, 0.07] \\ 
\rowcolor{Gray}
leg margin range & [0.05, 0.15] \\ 
position range & [[0, 0.8], [-0.6, 0.6]] \\ 
\rowcolor{Gray}
z axis rotation range & [0, 3.14] \\ \hline
\multicolumn{2}{c}{\textbf{Shelf Parameters}} \\ \hline
width range & [0.5, 1] \\ 
\rowcolor{Gray}
depth range & [0.2, 0.5] \\ 
height range & [0.5, 1.2] \\ 
\rowcolor{Gray}
num boards range & [3, 5] \\ 
board thickness range & [0.02, 0.05] \\ 
\rowcolor{Gray}
backboard thickness range & [0.0, 0.05] \\ 
num vertical boards range & [0, 3] \\ 
\rowcolor{Gray}
num side columns range & [0, 4] \\ 
column thickness range & [0.02, 0.05] \\ 
\rowcolor{Gray}
position range & [[0, 0.8], [-0.6, 0.6]] \\ 
z axis rotation range & [-1.57, 0] \\ \hline
\multicolumn{2}{c}{\textbf{Open Box Parameters}} \\ \hline
width range & [0.2, 0.7] \\ 
\rowcolor{Gray}
depth range & [0.2, 0.7] \\ 
height range & [0.3, 0.5] \\
\rowcolor{Gray}
thickness range & [0.02, 0.06] \\ 
front scale range & [0.6, 1] \\ 
\rowcolor{Gray}
position range & [[0.0, 0.8], [-0.6, 0.6]] \\ 
z axis rotation range & [-1.57, 0.0] \\ \hline
\multicolumn{2}{c}{\textbf{Cubby Parameters}} \\ \hline
cubby left range & [0.4, 0.1] \\ 
\rowcolor{Gray}
cubby right range & [-0.4, 0.1] \\ 
cubby top range & [0.85, 0.35] \\ 
\rowcolor{Gray}
cubby bottom range & [0.0, 0.1] \\ 
cubby front range & [0.8, 0.1] \\ 
\rowcolor{Gray}
cubby width range & [0.35, 0.2] \\ 
cubby horizontal middle board z axis shift range & [0.45, 0.1] \\ 
\rowcolor{Gray}
cubby vertical middle board y axis shift range & [0.0, 0.1] \\ 
board thickness range & [0.02, 0.01] \\ 
\rowcolor{Gray}
external rotation range & [0, 1.57] \\ 
internal rotation range & [0, 0.5] \\ 
\rowcolor{Gray}
num shelves range & [3, 5] \\ \hline
\multicolumn{2}{c}{\textbf{Microwave Parameters}} \\ \hline
width range & [0.3, 0.6] \\ 
\rowcolor{Gray}
depth range & [0.3, 0.6] \\ 
height range & [0.3, 0.6] \\ 
\rowcolor{Gray}
thickness range & [0.01, 0.02] \\ 
display panel width range & [0.05, 0.15] \\ 
\rowcolor{Gray}
distance range & [0.5, 0.8] \\ 
external z axis rotation range & [-2.36, -0.79] \\ 
\rowcolor{Gray}
internal z axis rotation range & [-0.15, 0.15] \\ \hline
\multicolumn{2}{c}{\textbf{Dishwasher Parameters}} \\ \hline
width range & [0.4, 0.6] \\ 
\rowcolor{Gray}
depth range & [0.3, 0.4] \\ 
height range & [0.5, 0.7] \\ 
\rowcolor{Gray}
control panel height range & [0.1, 0.2] \\ 
foot panel height range & [0.1, 0.2] \\ 
\rowcolor{Gray}
wall thickness range & [0.01, 0.02] \\ 
opening angle range & [0.5, 1.57] \\ 
\rowcolor{Gray}
distance range & [0.6, 1.0] \\ 
external z axis rotation range & [-2.36, -0.79] \\ 
\rowcolor{Gray}
internal z axis rotation range & [-0.15, 0.15] \\ \hline
\multicolumn{2}{c}{\textbf{Cabinet Parameters}} \\ \hline
width range & [0.5, 0.8] \\ 
\rowcolor{Gray}
depth range & [0.25, 0.4] \\ 
height range & [0.6, 1.0] \\ 
\rowcolor{Gray}
wall thickness range & [0.01, 0.02] \\ 
left opening angle range & [0.7, 1.57] \\ 
\rowcolor{Gray}
right opening angle range & [0.7, 1.57] \\ 
distance range & [0.6, 1.0] \\ 
\rowcolor{Gray}
external z axis rotation range & [-2.36, -0.79] \\ 
internal z axis rotation range & [-0.15, 0.15] \\ 

\caption{\small \textbf{Data Generation Hyper-parameters} We provide a detailed list of hyper-parameters used to procedurally generate a vast variety of scenes in simulation. }
\label{tab:datagen params}
\end{longtable}
\clearpage

\begin{table}[h!]
\centering
\begin{tabular}{l|l}
\hline
\textbf{Hyper-parameter} & \textbf{Value} \\ \hline
PointNet++ Architecture & 
\begin{minipage}[t]{0.5\textwidth}
\begin{verbatim}
PointnetSAModule(
    npoint=128,
    radius=0.05,
    nsample=64,
    mlp=[1, 64, 64, 64],
)
PointnetSAModule(
    npoint=64,
    radius=0.3,
    nsample=64,
    mlp=[64, 128, 128, 256],
)
PointnetSAModule(
    nsample=64,
    mlp=[256, 512, 512],
)
MLP(
    Linear(512, 2048),
    GroupNorm(16, 2048),
    LeakyReLU,
    Linear(2048, 1024),
    GroupNorm(16, 1024),
    LeakyReLU,
    Linear(1024, 1024)
)
\end{verbatim}
\end{minipage} \\ 
\rowcolor{Gray}
LSTM & 1024 hidden dim, 2 layers \\ 
Inputs & $q_t$, $g$, $PCD_t$ \\ 
\rowcolor{Gray}
Batch Size & 16 \\ 
Learning Rate & $0.0001$ \\ 
\rowcolor{Gray}
GMM & 5 modes \\
Sequence Length (seq length) & 2 \\ \hline
\multicolumn{2}{c}{\textbf{Point Cloud Parameters}} \\ \hline
Number of Robot / Goal Point-cloud Points & 2048 \\ 
\rowcolor{Gray}
Number of Obstacle Point-cloud Points & 4096 \\ 
\end{tabular}
\caption{Hyper-parameters for the model}
\label{tab:model hyper params}
\end{table}

\clearpage
\section{Real World Setup Details}
\label{app:real world}
In this section, we describe our real world robot setup and tasks in detail and perform analysis on the perception used for operating our policies. 

\subsection{Real Robot Setup}
\begin{figure}
\begin{algorithm}[H]
\caption{Open-Loop Execution of Neural MP}
\label{alg:open loop execution}
\begin{algorithmic}[1]
\State \textbf{Input:} Neural MP $\pi_{\theta}$, segmentor $\mathcal{S}$, initial angles $q_0$, scene point-cloud $PCD_{full}$, goal $g$, horizon $H$
\State \textbf{Output:} Executed trajectory on the robot
\State \textbf{Initialize: } Timestep $t \leftarrow 0$
\State \textbf{Initialize: } Trajectory $\tau \leftarrow \{\}$
\State $PCD_0 \leftarrow \mathcal{S}(PCD_{full}) \cup PCD_{q_0} \cup PCD_{g}$
\While{goal $g$ not reached $\And t < H$}
    \State $a_t \sim \pi_{\theta}(q_{t-1}, PCD_{t-1}, g)$
    \State $q_t \leftarrow q_{t-1} + a_t$
    \State $PCD_t \leftarrow (PCD_{t_1} \setminus PCD_{q_{t-1}}) \cup PCD_{q_t}$
    \State $t \leftarrow t + 1$
    \State $\tau \leftarrow \tau + a_t$
\EndWhile
\State Execute the $\tau$ open loop on the robot.
\end{algorithmic}
\end{algorithm}
\end{figure}

\textbf{Hardware} For all of our experiments, we use a Franka Emika Panda Robot, which is a 7 degree of freedom manipulator arm. We control the robot using the manimo library (\href{https://github.com/AGI-Labs/manimo}{https://github.com/AGI-Labs/manimo}) and perform all of experiments using their joint position controller with the default PD gains. The robot is mounted to a fixed base pedestal behind a desk of size .762m by 1.22m with variable height. For sensing, we use four extrinsically calibrated depth cameras, Intel Realsense 435 / 435i, placed around the scene in order to accurately capture the environment. We project the depth maps from each camera into 3D and combine the individual point-clouds into a single scene representation. 
We then post-process the point-cloud by cropping it to the workspace, filtering outliers and denoising, and sub-sampling a set of 4096 points. This processed point-cloud is then used as input to the policy.

\begin{figure}[t]
    \centering
    \vspace{-.2in}
    \includegraphics[width=.45\linewidth]{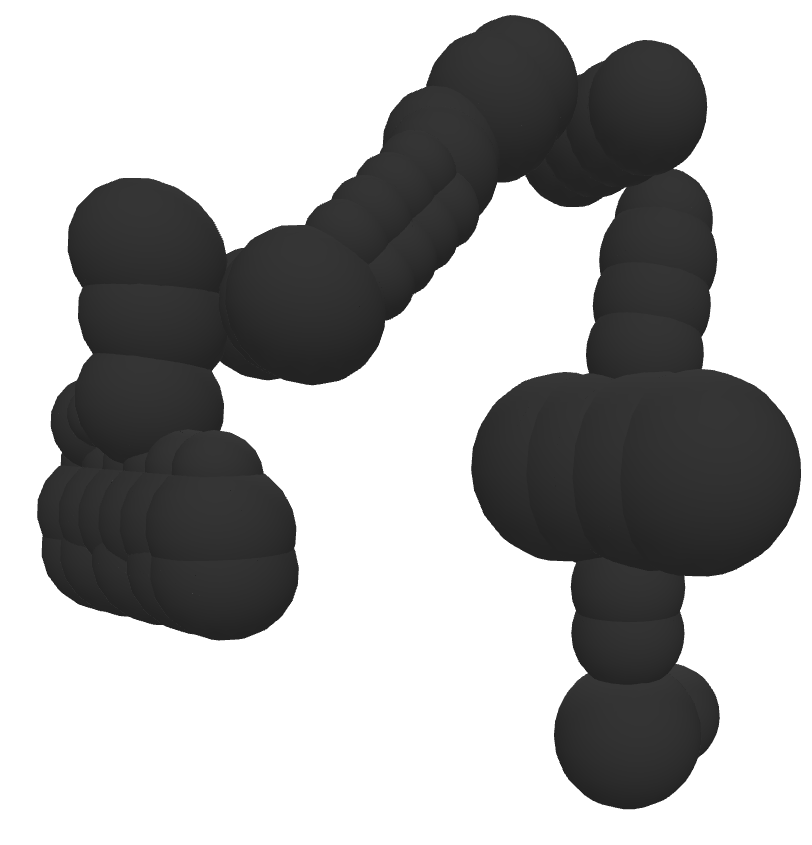}
    \includegraphics[width=.45\linewidth]{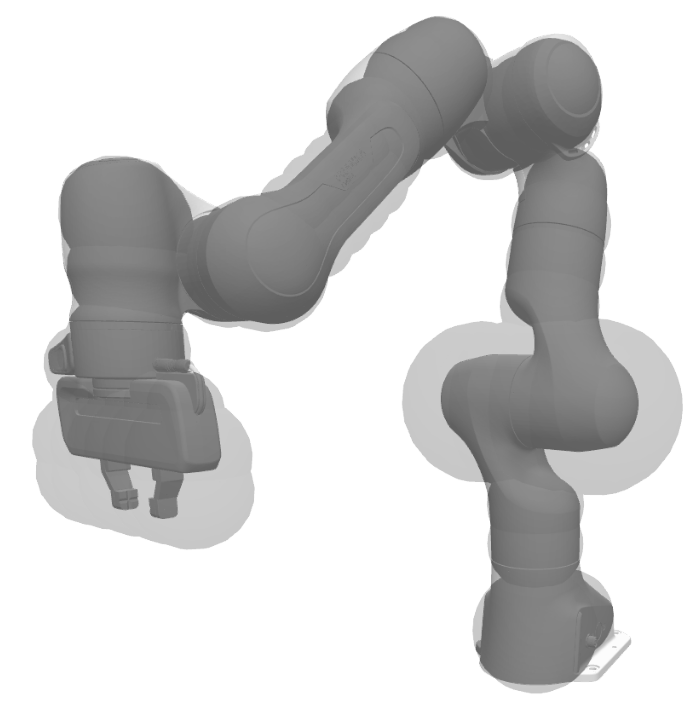}
    \caption{\small We visualize the spherical representation on the left and overlay it on the robot mesh on the right.}
    \label{fig:spherical repr vis}
    \vspace{-.1in}
\end{figure}

\textbf{Representation Collision Checking and Segmentation} In order to perform real world collision checking and robot point-cloud segmentation, we require a representation of the robot to check intersections with the scene (collision checking) and to filter out robot points from the scene point-cloud (segmentation). While the robot mesh is the ideal candidate for these operations, it is far too slow to run in real time. Instead, we approximate the robot mesh as spheres (visualized in Fig.~\ref{fig:spherical repr vis}) as we found this performs well in practice while operating an order of magnitude faster. We use 56 spheres in total to approximate the links of the robot as well as the end-effector and gripper. These have radii ranging from 2cm to 10cm and are defined relative to the center of mass of the link. This representation is a conservative one: it encapsulates the robot mesh, which is desirable for segmentation as this helps account for sensing errors which would place robot points outside of the robot mesh.

\textbf{Robot Segmentation} In order to perform robot segmentation in the real world, we use the spherical representation to filter out robot points in the scene, so only the obstacle point-cloud remains. Doing so requires computing the Signed Distance Function (SDF) of the robot representation and then checking the scene point-cloud against it, removing points from the point-cloud in which SDF value is less than threshold $\epsilon$. For the spherical representation, the SDF computation is efficient: for a sphere with center $C$ and radius $r$, the SDF of point x is simply $ ||x - C||_2 - r$. In our experiments, we use a threshold $\epsilon$ of 1cm. We then replace the removed points with points sampled from the robot mesh of the robot. This is done by pre-sampling a robot point-cloud from the robot mesh at the default configuration, then performing forward kinematics using the current joint angles $q_t$ and transforming the robot point-cloud accordingly. Replacing the real robot point-cloud with this sampled point-cloud ensures that the only difference between sim and real is the obstacle point-cloud. 

\textbf{Real-world Collision Checking} Given the SDF, collision checking is also straightforward, we denote the robot in collision if any point in the scene point-cloud (this is after robot segmentation) has SDF value less than 1cm. Note this means that first state is by definition collision free. Also, this technique will not hold if performing closed loop planning, in that case this method would always denote the state as collision free as the points with SDF value less than 1cm would be segmented out for each intermediate point-cloud.

\textbf{Open Loop Deployment} For open-loop execution of neural motion planners, we execute the following steps: 1) generate the segmented point-cloud at the first frame, 2) predict the next trajectory way-point by computing a forward pass through the network and sampling an action, 3) update the current robot point-cloud with mesh-sampled point-cloud at the predicted way-point, and 4) repeat until goal reaching success or maximum rollout length is reached. The entire trajectory is then executed on the robot after the rollout. Please see Alg.~\ref{alg:open loop execution} for a more detailed description of our open-loop deployment method.

\subsection{Tasks}

\begin{figure*}[ht!]
\centering
\begin{subfigure}[b]{0.24\linewidth}
    \includegraphics[width=\linewidth]{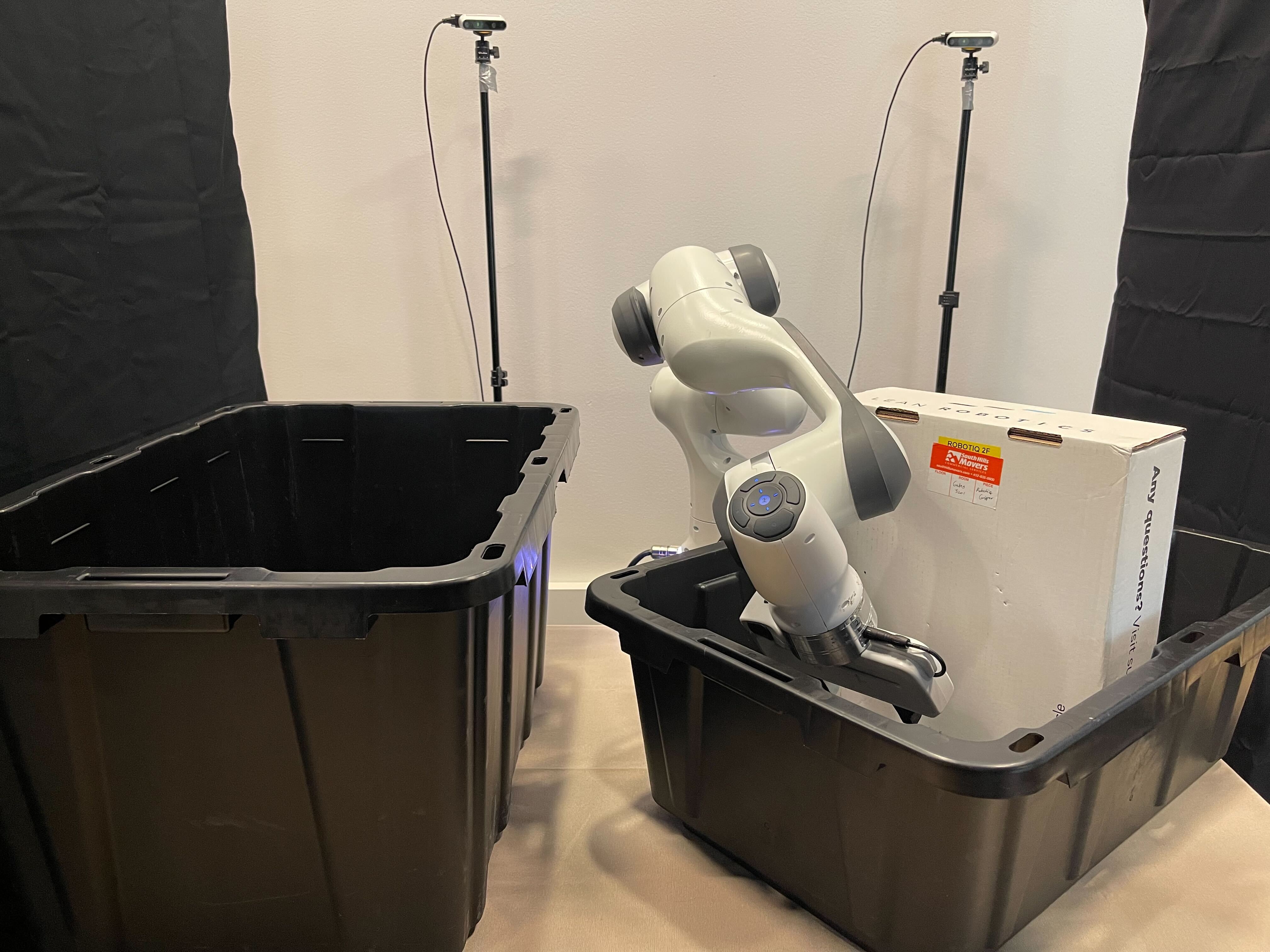}
    \caption{\small Bins Scene 1}
\end{subfigure}
\begin{subfigure}[b]{0.24\linewidth}
    \includegraphics[width=\linewidth]{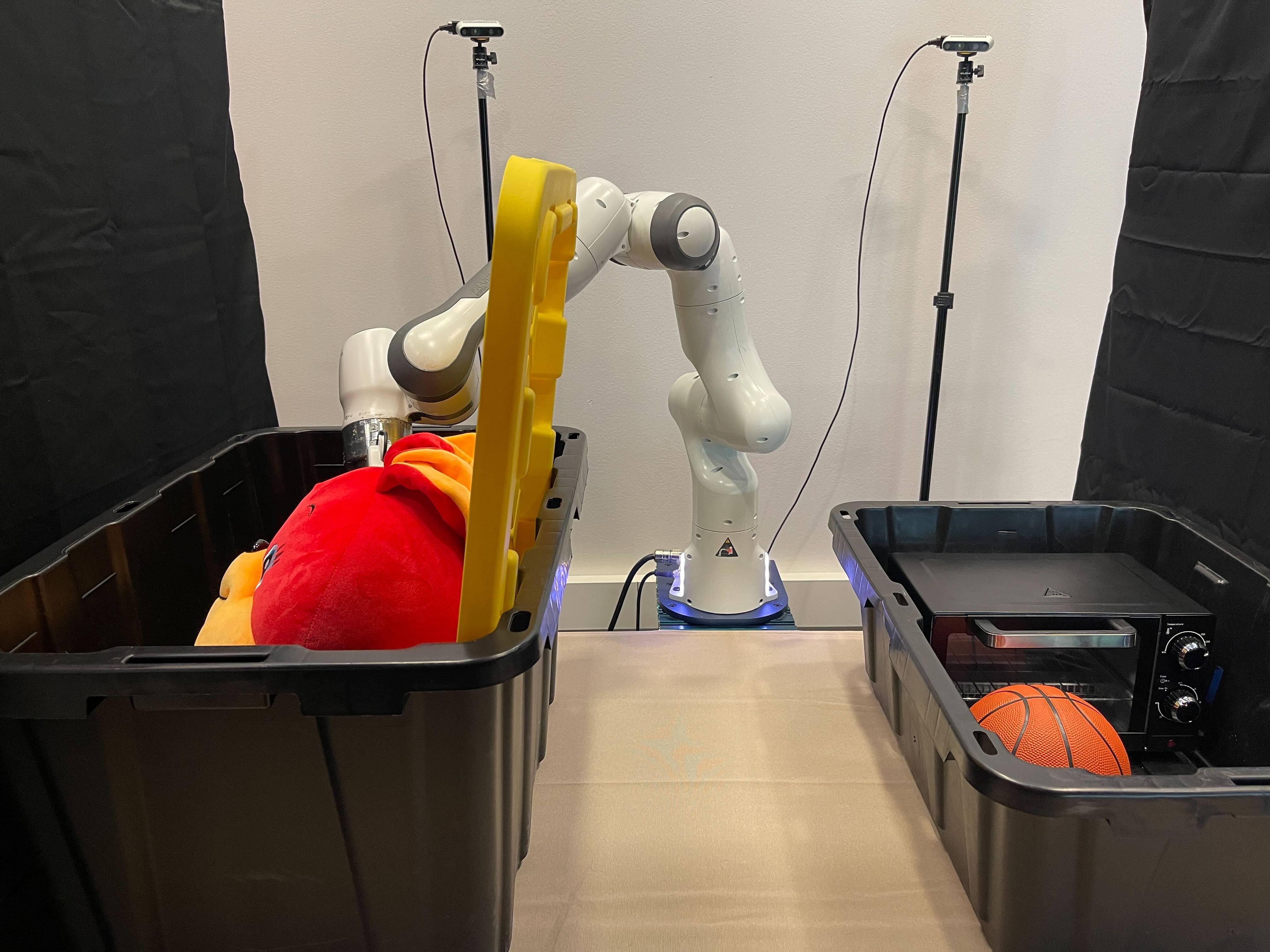}
    \caption{\small Bins Scene 2}
\end{subfigure}
\begin{subfigure}[b]{0.24\linewidth}
    \includegraphics[width=\linewidth]{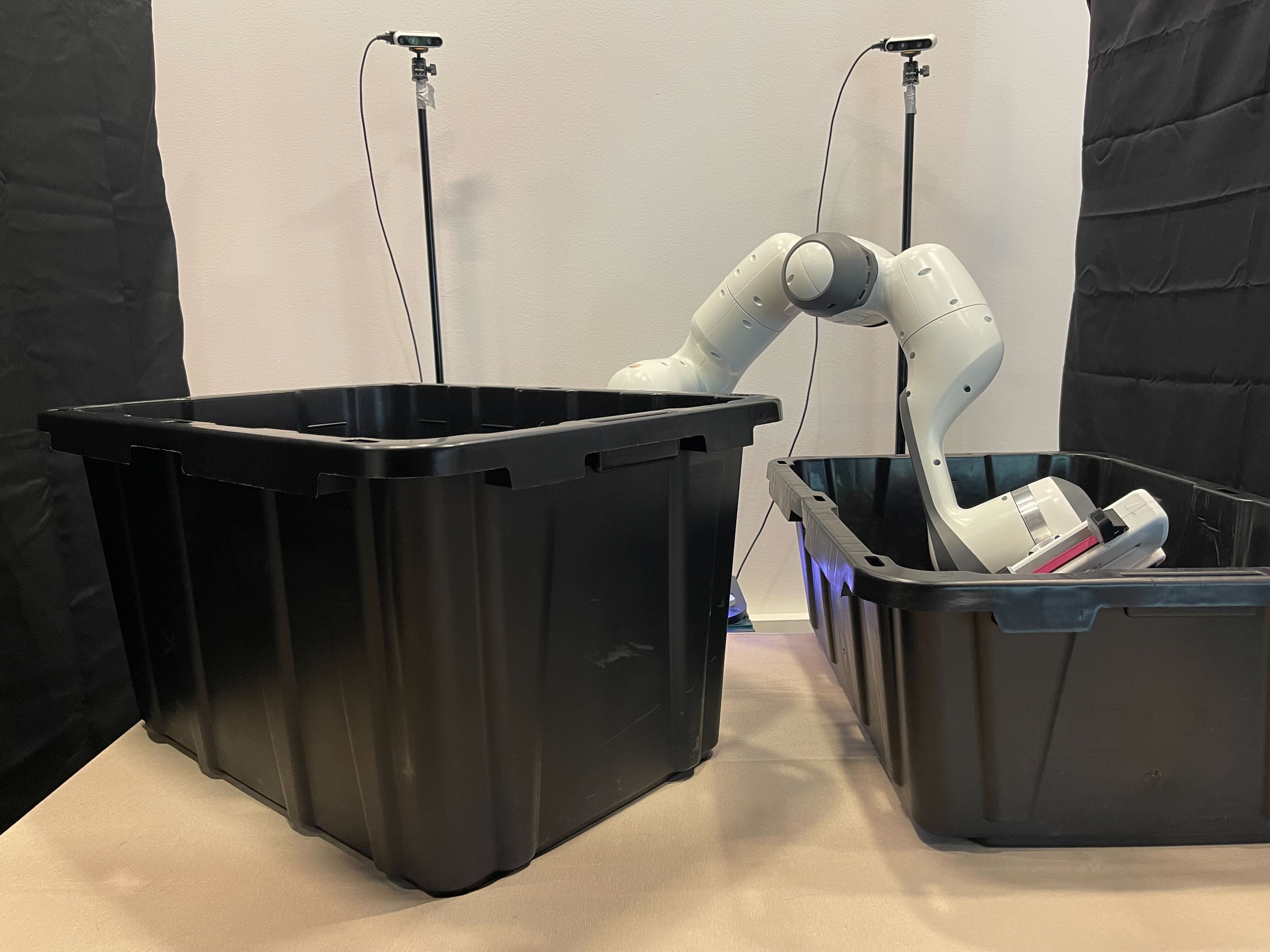}
    \caption{\small Bins Scene 3}
\end{subfigure}
\begin{subfigure}[b]{0.24\linewidth}
    \includegraphics[width=\linewidth]{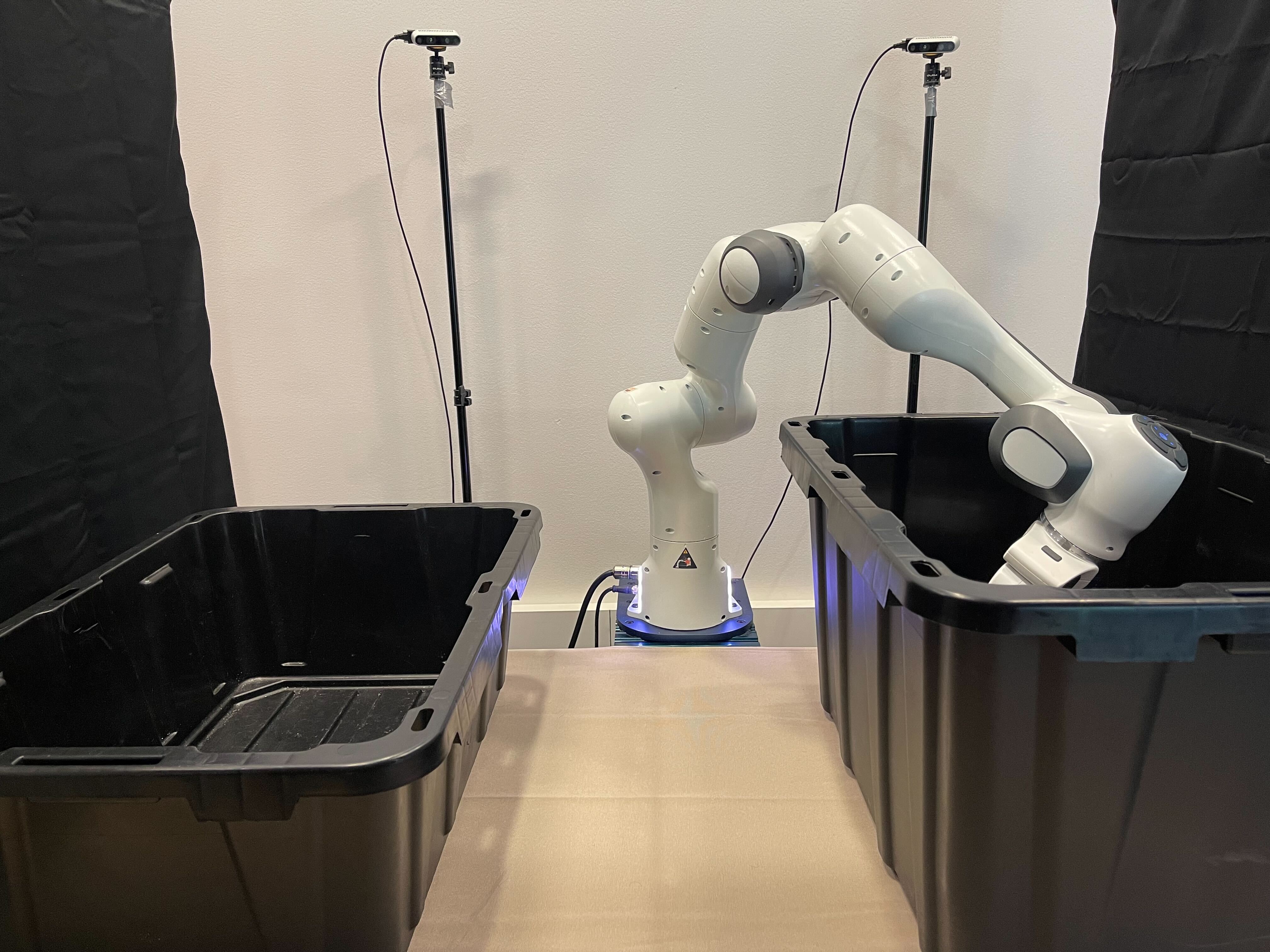}
    \caption{\small Bins Scene 4}
\end{subfigure}
\begin{subfigure}[b]{0.24\linewidth}
    \includegraphics[width=\linewidth]{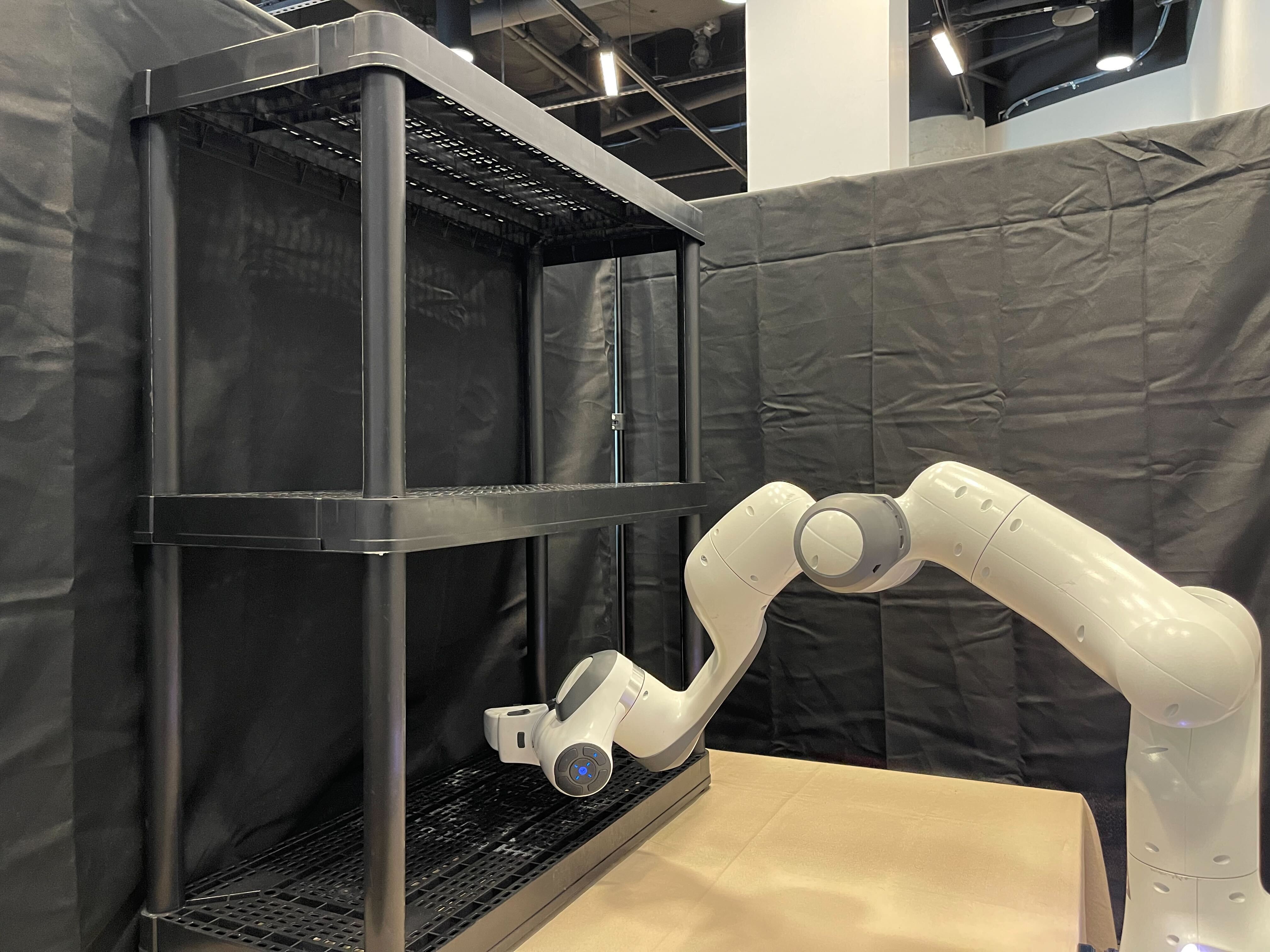}
    \caption{\small Shelf Scene 1}
\end{subfigure}
\begin{subfigure}[b]{0.24\linewidth}
    \includegraphics[width=\linewidth]{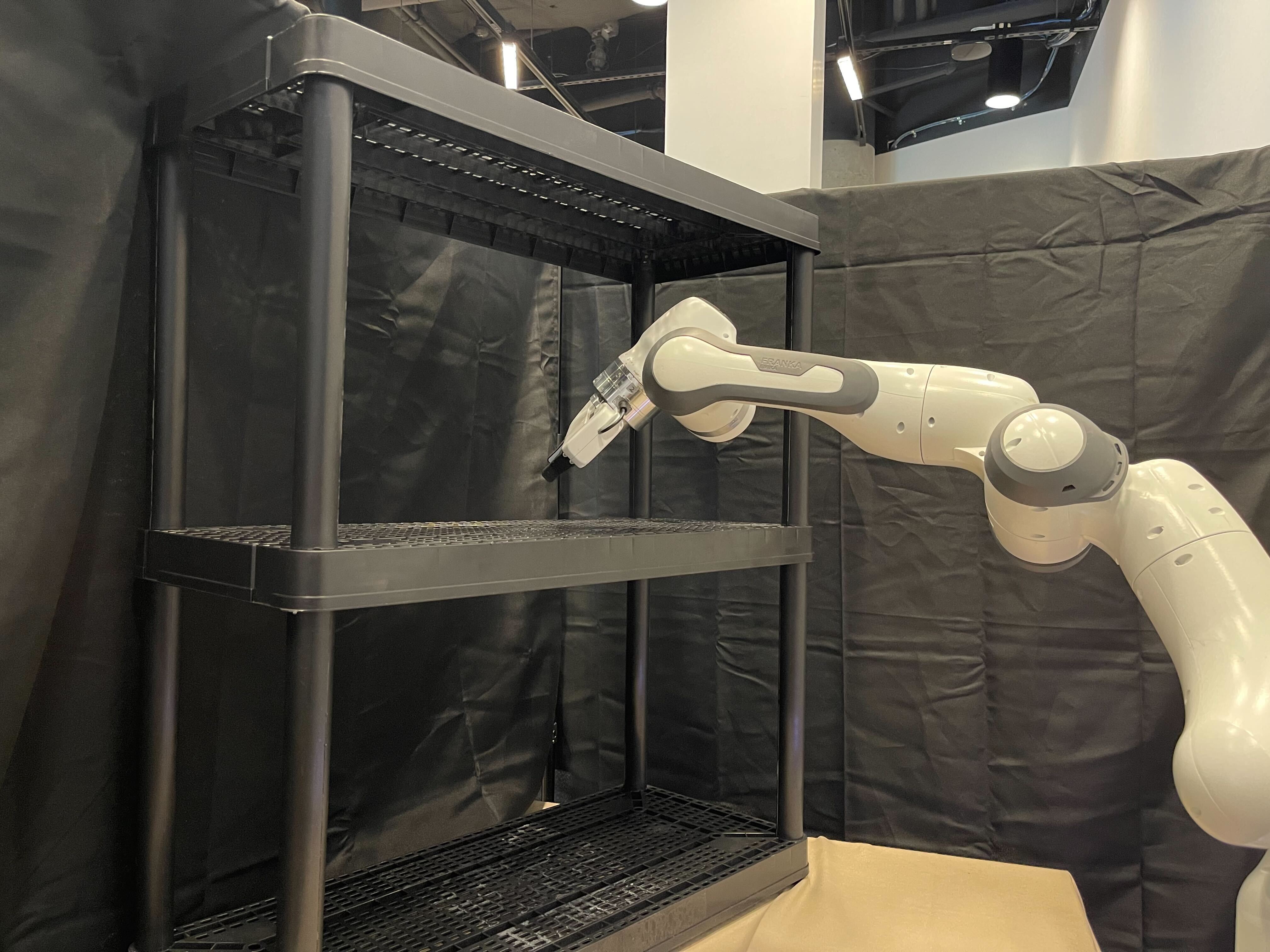}
    \caption{\small Shelf Scene 2}
\end{subfigure}
\begin{subfigure}[b]{0.24\linewidth}
    \includegraphics[width=\linewidth]{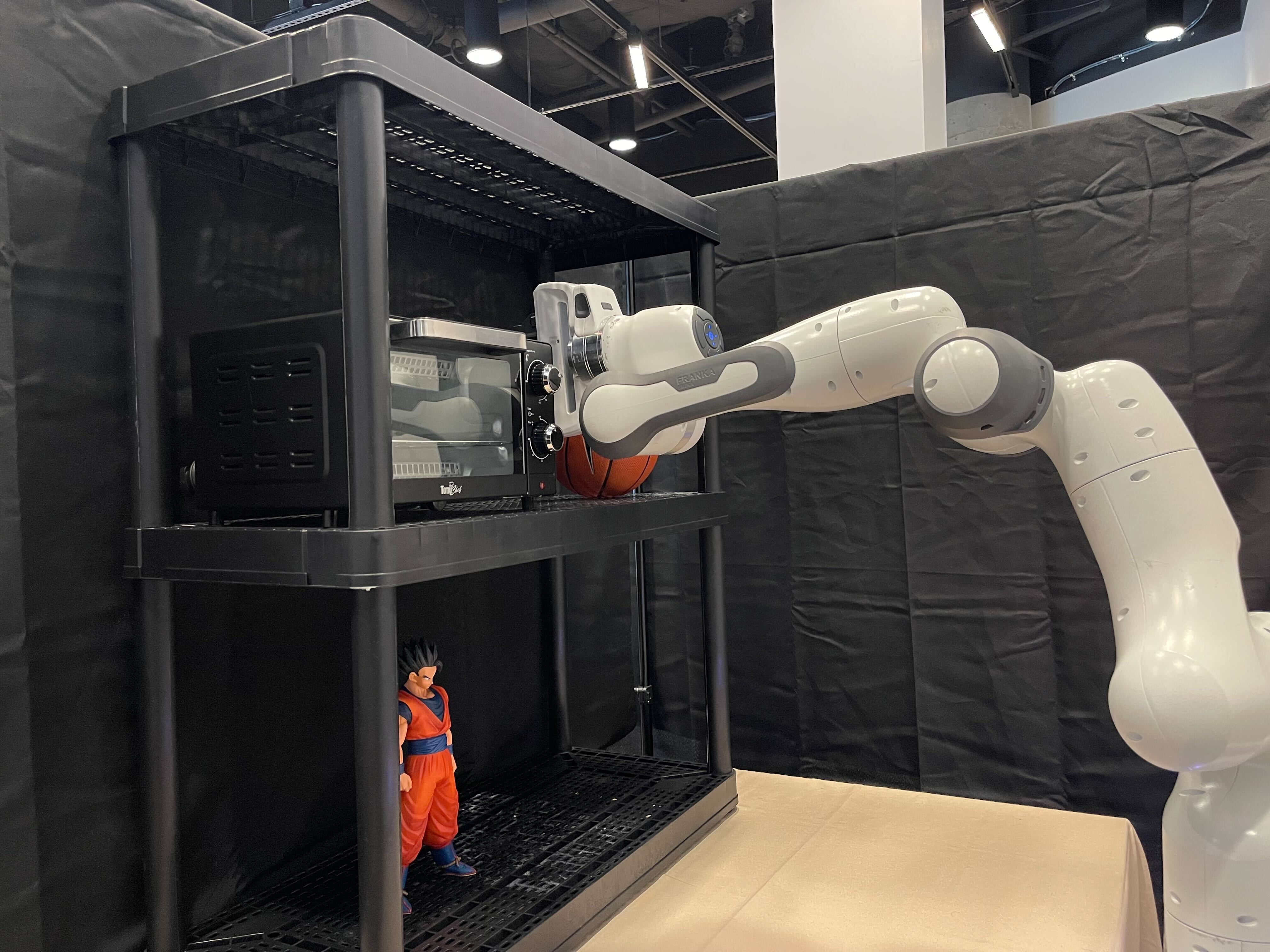}
    \caption{\small Shelf Scene 3}
\end{subfigure}
\begin{subfigure}[b]{0.24\linewidth}
    \includegraphics[width=\linewidth]{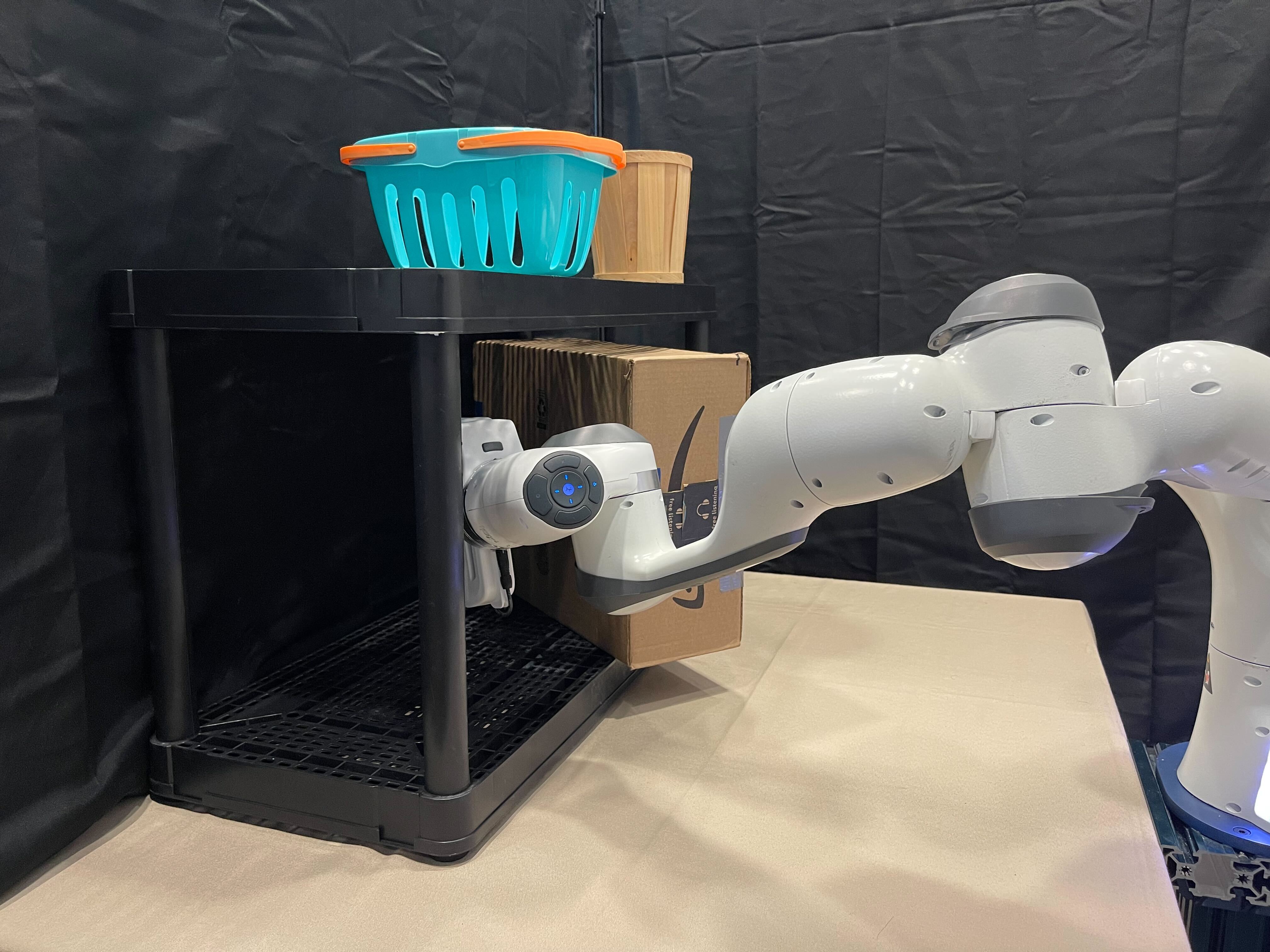}
    \caption{\small Shelf Scene 4}
\end{subfigure}
\begin{subfigure}[b]{0.24\linewidth}
    \includegraphics[width=\linewidth]{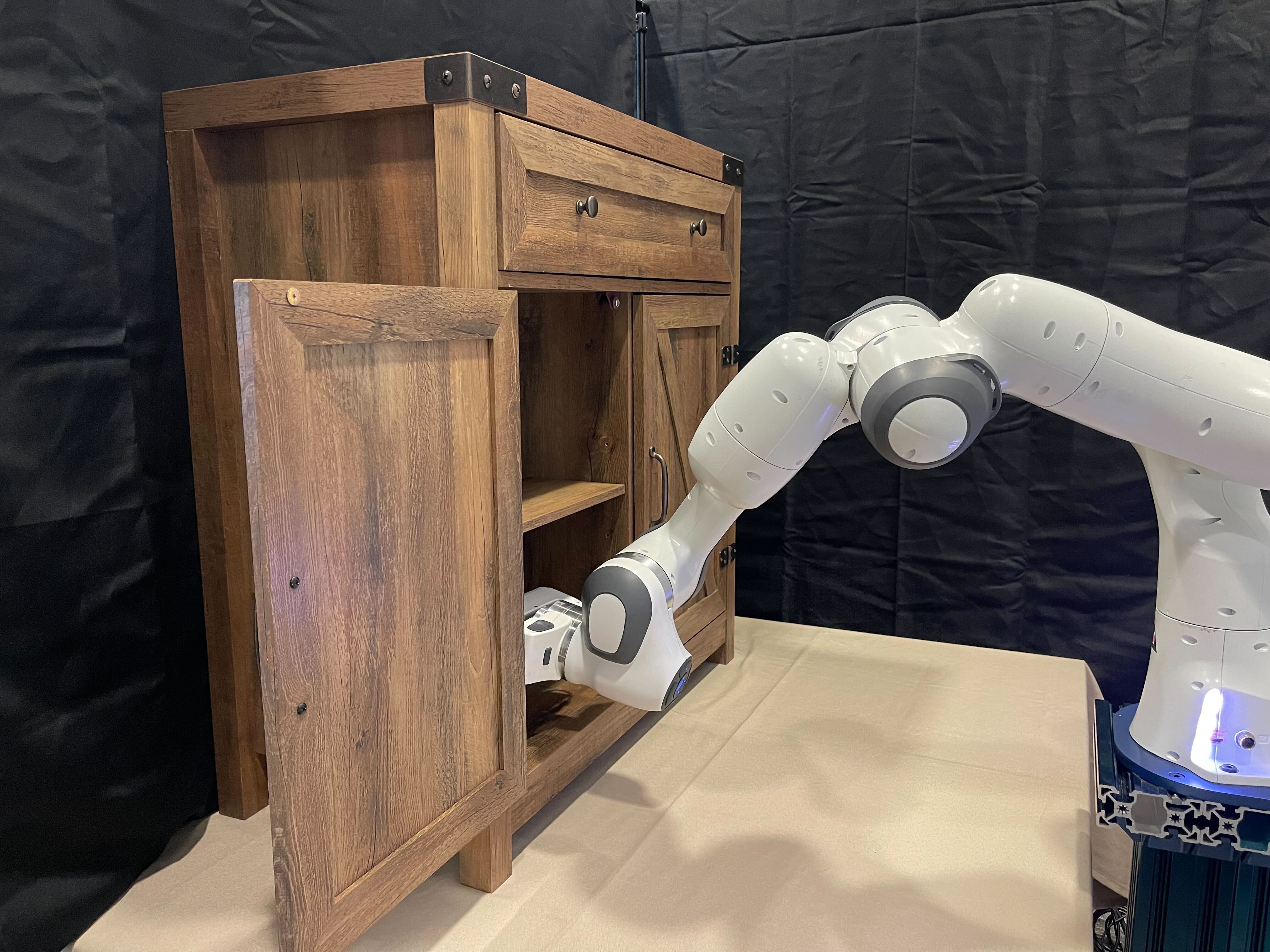}
    \caption{\small Articulated Scene 1}
\end{subfigure}
\begin{subfigure}[b]{0.24\linewidth}
    \includegraphics[width=\linewidth]{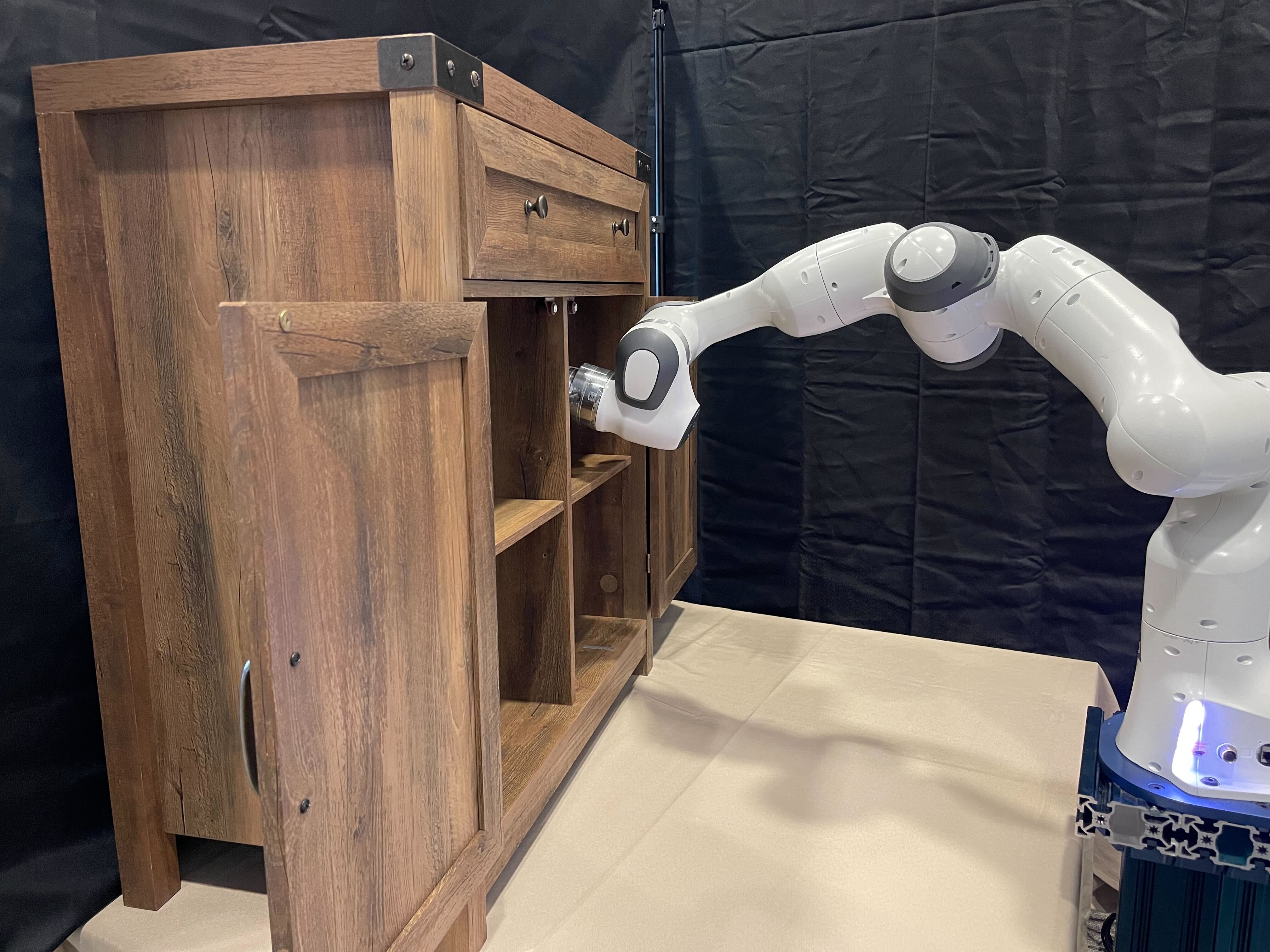}
    \caption{\small Articulated Scene 2}
\end{subfigure}
\begin{subfigure}[b]{0.24\linewidth}
    \includegraphics[width=\linewidth]{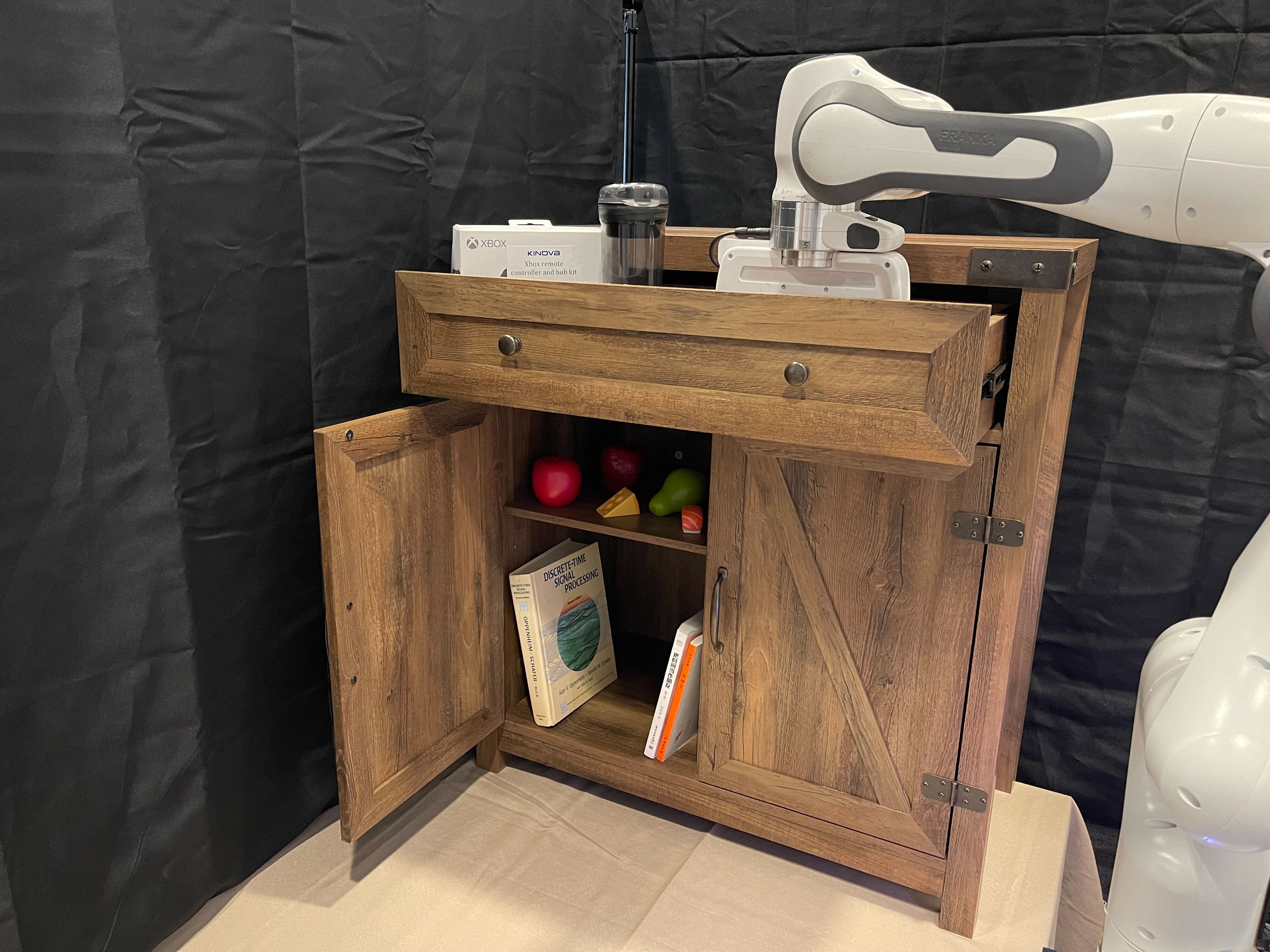}
    \caption{\small Articulated Scene 3}
\end{subfigure}
\begin{subfigure}[b]{0.24\linewidth}
    \includegraphics[width=\linewidth]{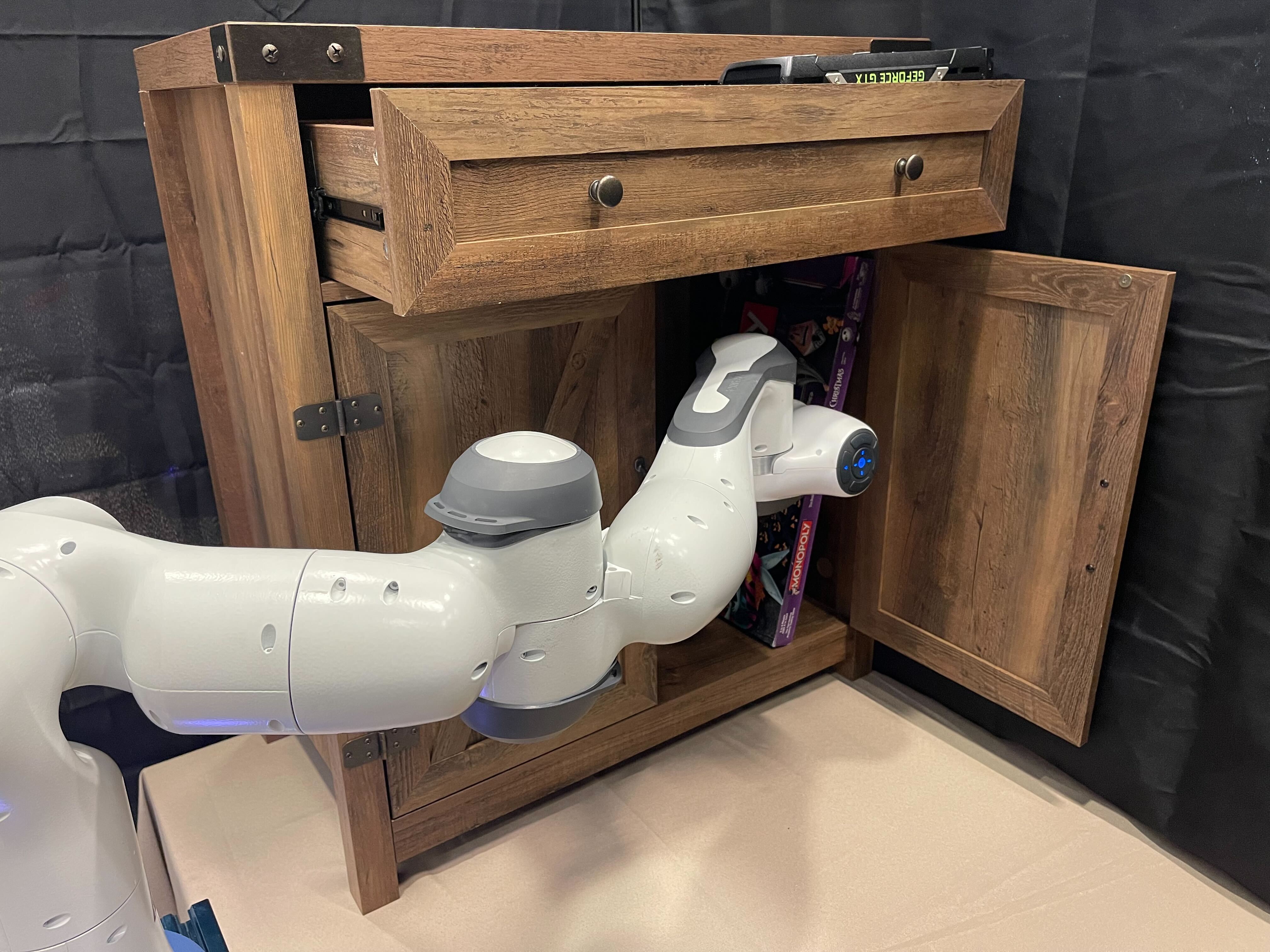}
    \caption{\small Articulated Scene 4}
\end{subfigure}
\begin{subfigure}[b]{0.24\linewidth}
    \includegraphics[width=\linewidth]{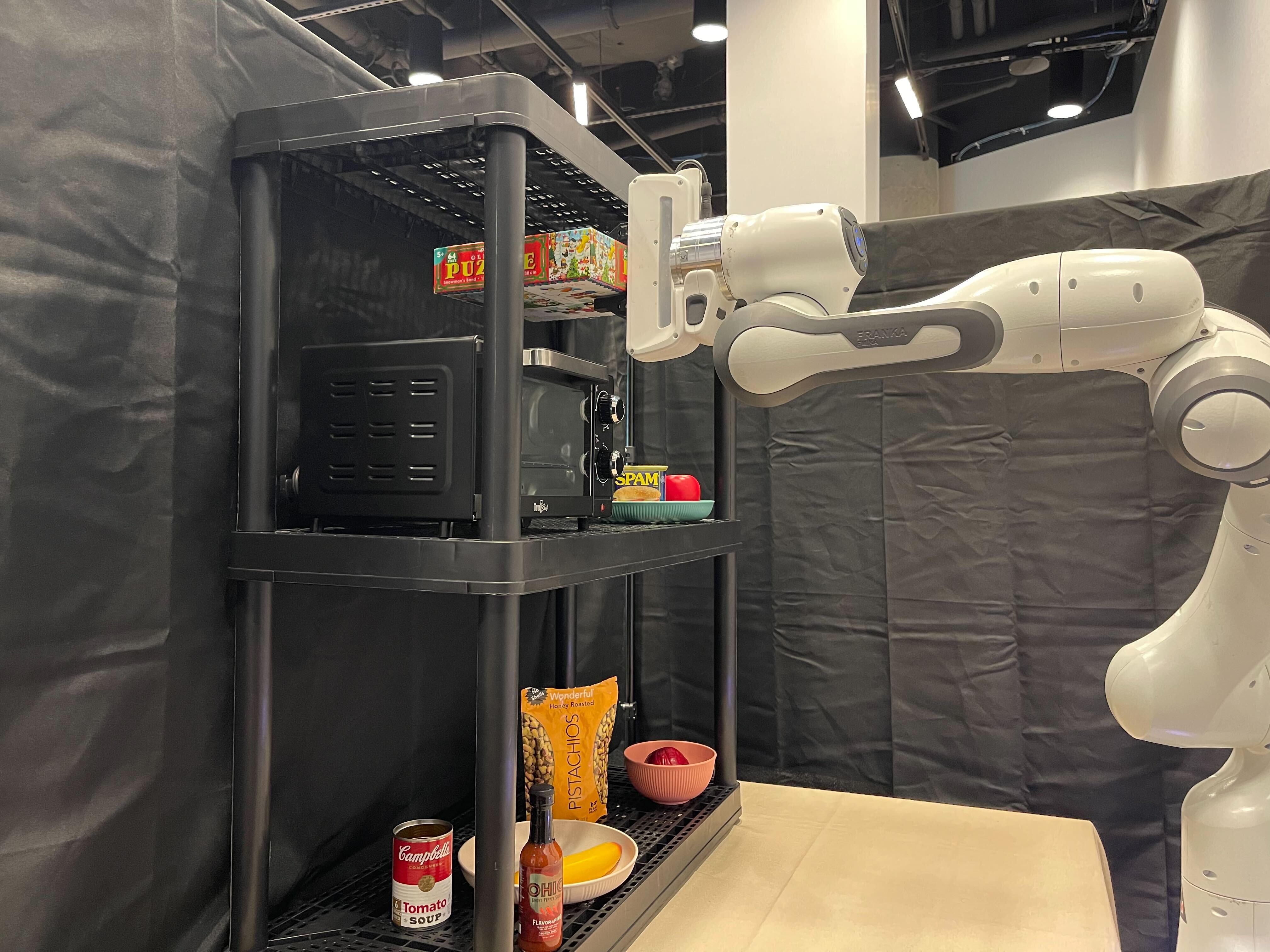}
    \caption{\small In Hand Object 1}
\end{subfigure}
\begin{subfigure}[b]{0.24\linewidth}
    \includegraphics[width=\linewidth]{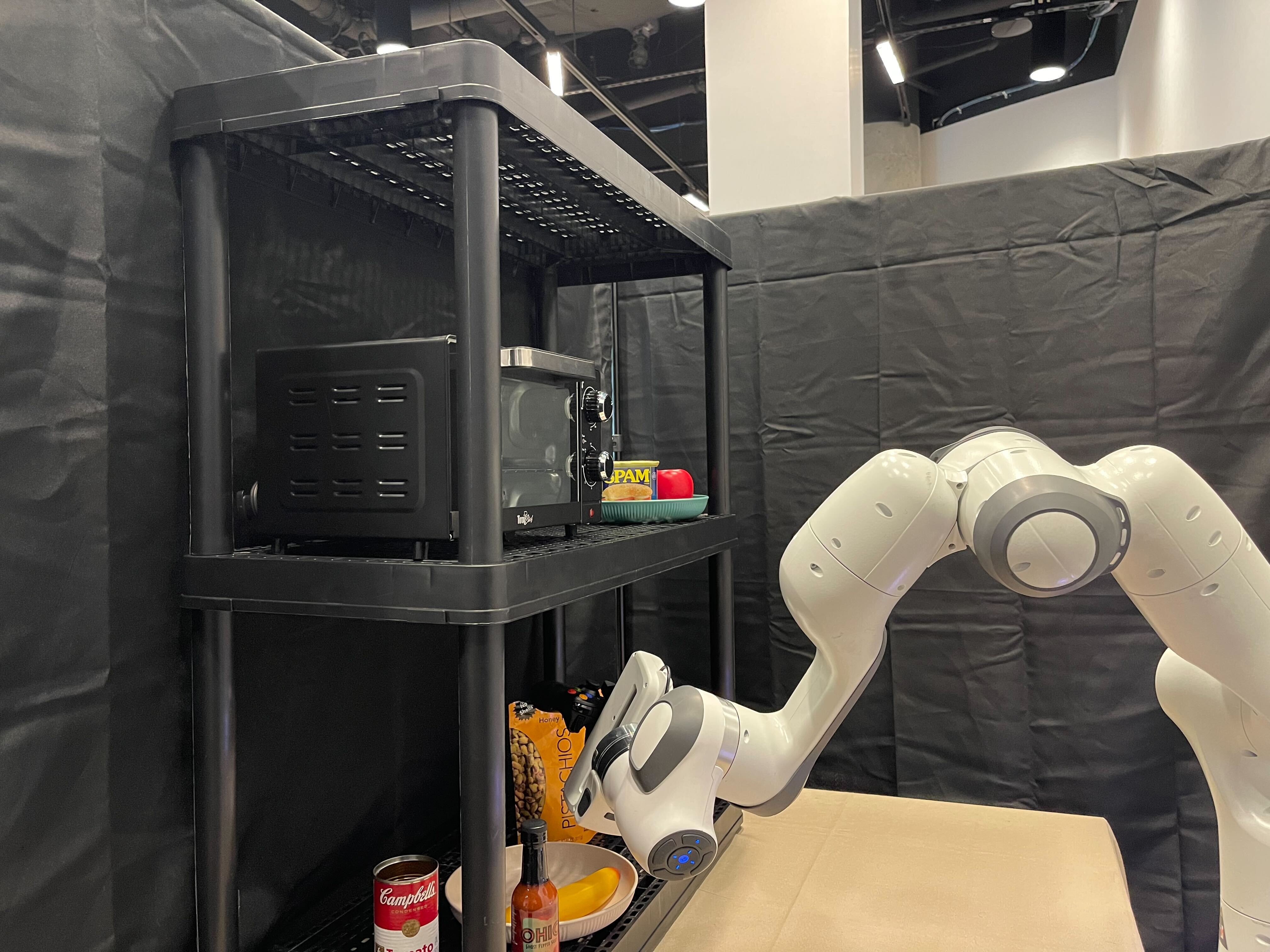}
    \caption{\small In Hand Object 2}
\end{subfigure}
\begin{subfigure}[b]{0.24\linewidth}
    \includegraphics[width=\linewidth]{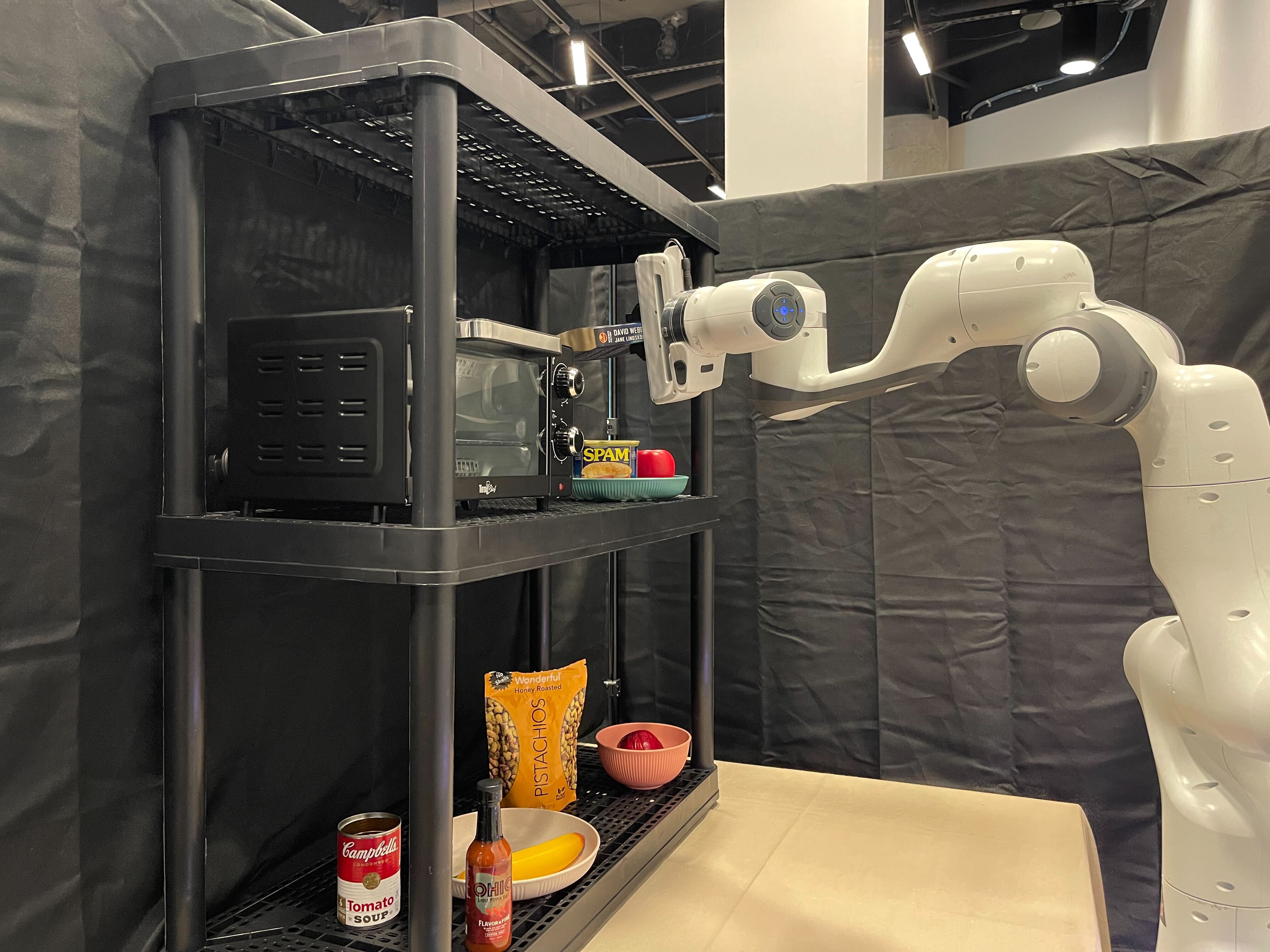}
    \caption{\small In Hand Object 3}
\end{subfigure}
\begin{subfigure}[b]{0.24\linewidth}
    \includegraphics[width=\linewidth]{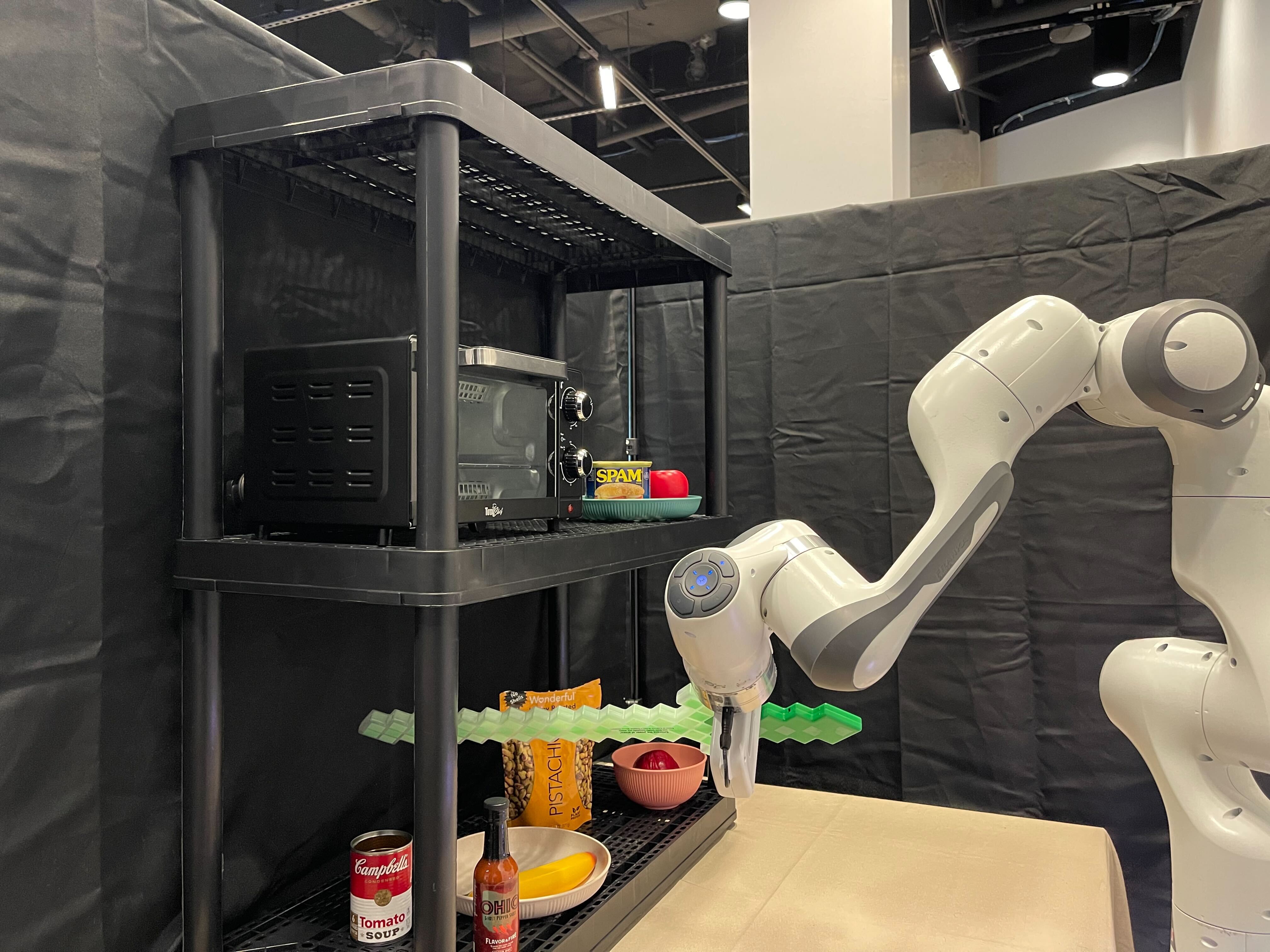}
    \caption{\small In Hand Object 4}
\end{subfigure}

\caption{\small Images of our 16 evaluation scenes.}
\vspace{-0.1in}
\label{fig:detailed setups}
\end{figure*}

\textbf{Bins}
This task requires the neural planner to perform collision avoidance when moving in-between, around and inside two different industrial bins pictured in the first row of Fig.~\ref{fig:detailed setups}. We randomize the position and orientation of the bins over the table and include the following objects as additional obstacles for the robot to avoid: toaster, doll, basketball, bin cap, and white box. The small bin is of size 70cm x 50cm x 25cm. The larger bin is of size 70cm x 50cm x 37cm. The bins are placed at two sides of the table. Between tasks, we randomize the orientation of the bins between 0 and 45 degrees and we swap the bin ordering (which bin is on the left vs. the right). The bins are placed 45cm in front of the robot, and shifted 60cm left/right.

\textbf{Shelf}
This task tests the agent's ability to handle horizontal obstacles (the rungs of the shelf) while maneuvering in tighter spaces (row two in Fig.~\ref{fig:detailed setups}). We randomize the size of the shelf (by changing the number of layers in the shelf from 3 to 2) as well as the position and orientation (anywhere at least .8m away from the robot) with 0 or 30 degrees orientation. The obstacles for this task include the toaster, basketball, baskets, an amazon box and an action figure which increase the difficulty. The shelf obstacle itself is of size 35cm x 80cm x 95cm. 

\textbf{Articulated}
We extend our evaluation to a more complex primary obstacle, the cabinet, which contains one drawer and two doors and tight internal spaces with small cubby holes (row three of Fig.~\ref{fig:detailed setups}). We randomize the position of the entire cabinet over the table, the joint positions of the drawer and doors and the sizes of the cubby holes. The obstacles for this task are xbox controller box, gpu, action figure, food toy, books and board game box. The size of the cabinet is 40cm x 75cm x 80cm. The size of the top drawer is 30cm x 65cm x 12cm. The size of the cubbies is 35cm x 35cm x 25cm. The drawer has an opening range of 0-30cm and the doors open between 0 and 180 degrees.  

\textbf{In-Hand Motion Planning}
In this task (shown in row four of Fig.~\ref{fig:detailed setups}), the planner needs to reason about collisions with not only the robot and the environment, but the held object too. We initialize the robot with an object grasped in-hand and run motion planning to reach a target configuration. For this task, we fix the obstacle (shelf) and its position (directly 80cm in front of the robot), instead randomizing across in-hand objects and configurations. We select four objects that vary significantly in size and shape: Xbox controller (18cm x 15cm x 8cm), book (17cm x 23cm x 5cm), toy sword (65cm x 10cm x 2cm), and board game (25cm x 25cm x 6cm). For this evaluation, we assume the object is already grasped by the robot, and the robot must just move with the object in-hand while maintaining its grasp. 

\subsection{Perception Visualization and Analysis}

\begin{figure}
    \vspace{-0.3in}
    \centering
    \includegraphics[width=0.45\linewidth]{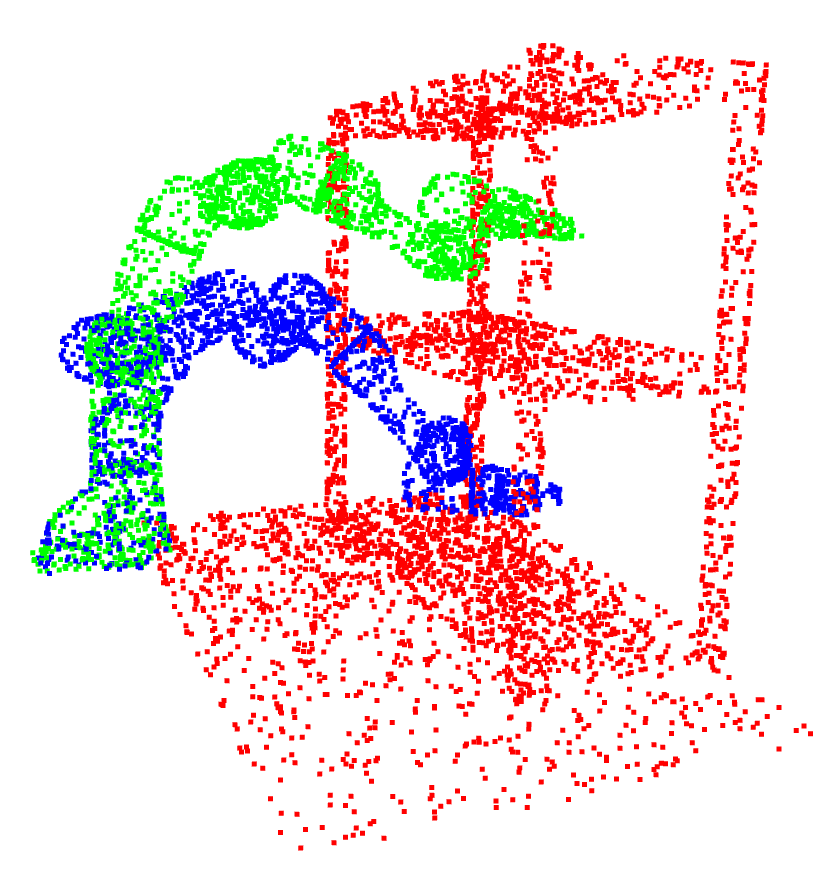} 
    \includegraphics[width=0.49\linewidth]{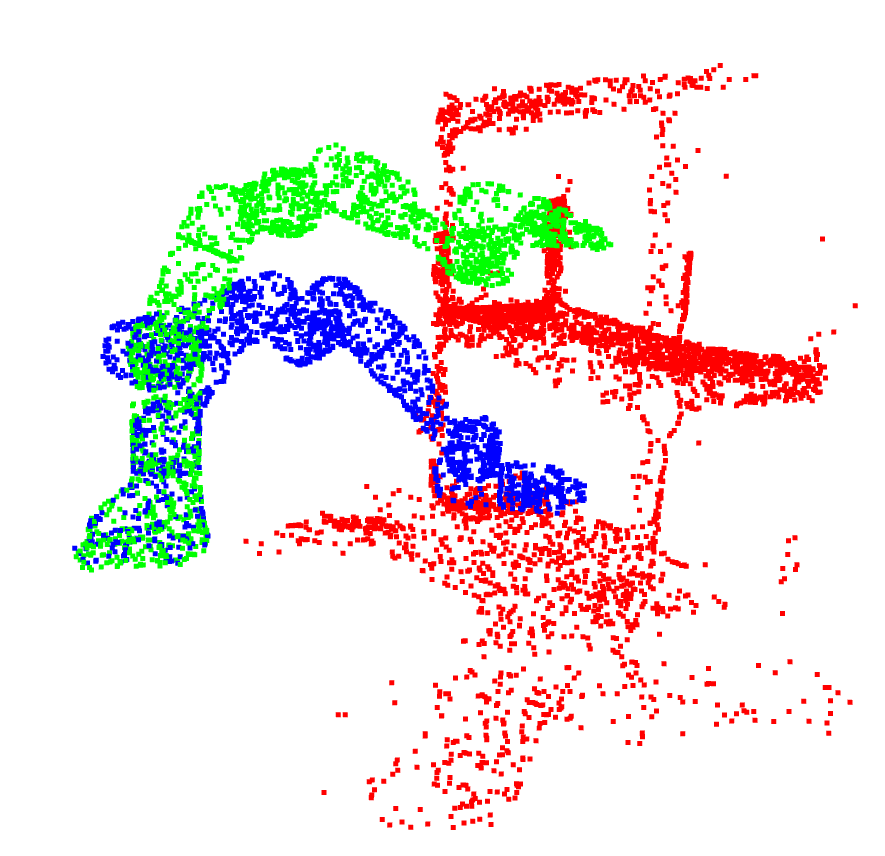} 
    \includegraphics[width=0.49\linewidth]{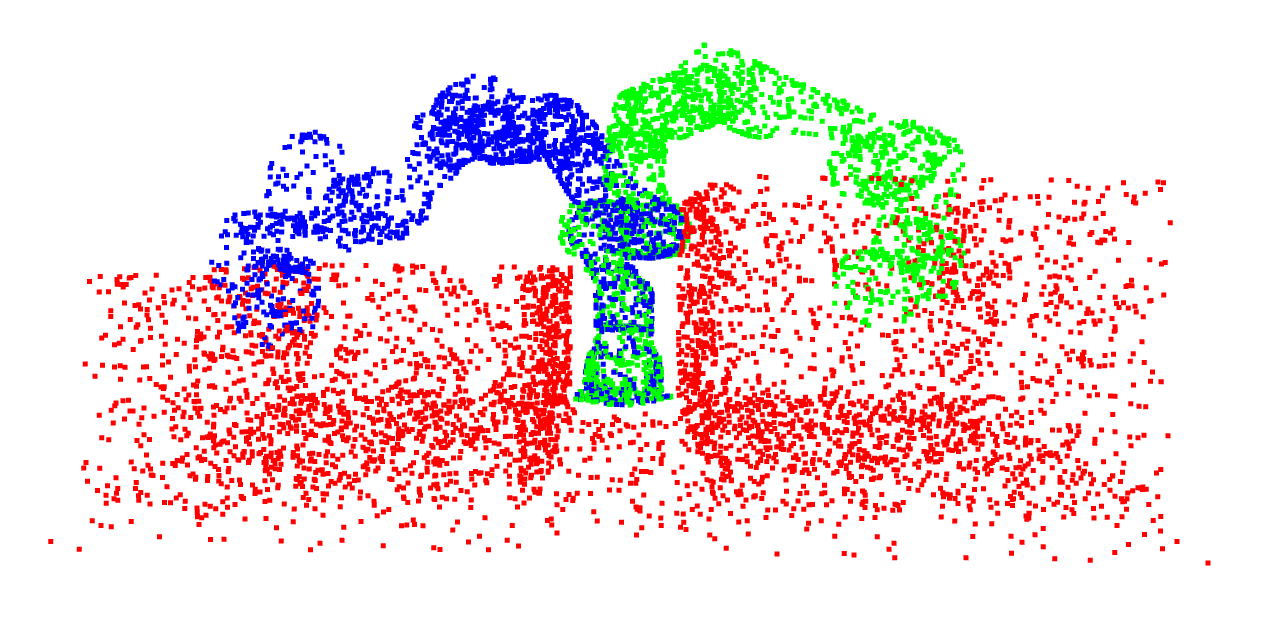}
    \includegraphics[width=0.49\linewidth]{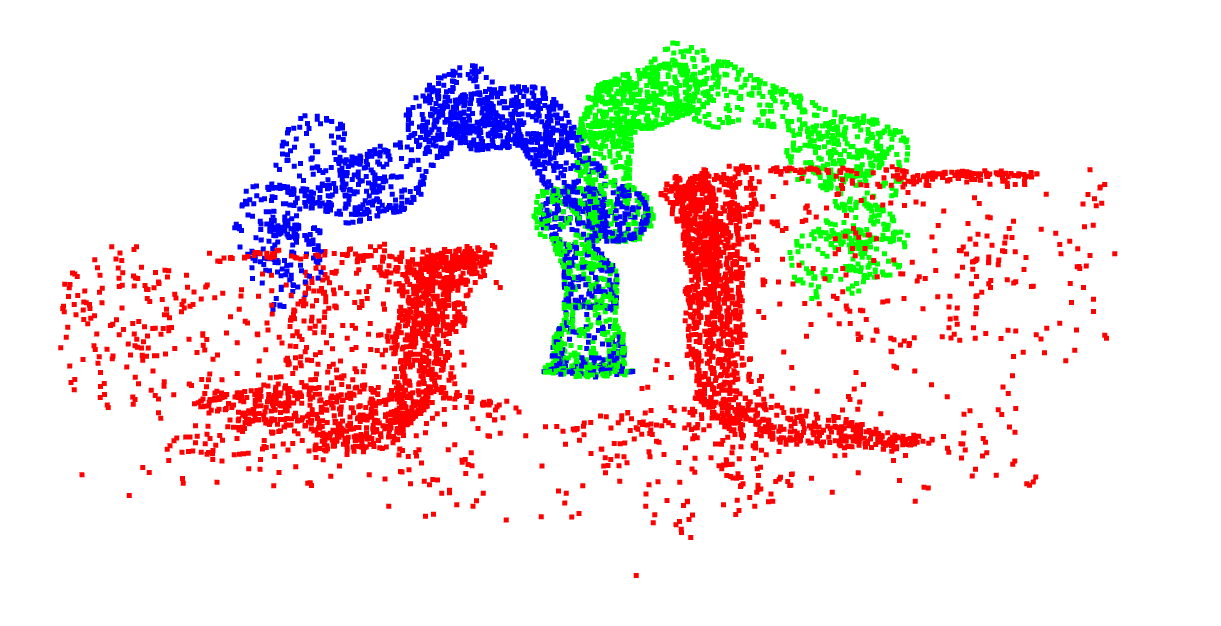}
    \caption{\small \textbf{Visualization of Sim and Real point-clouds}: We visualize point-clouds of the Bins and Shelf task in sim and real, in the same poses. Due to noise in depth sensing, the real world point-clouds have significantly more deformations, yet our policy generalizes well to these tasks.
    }
    \vspace{-0.12in}
    \label{fig:sim real point-cloud vis}
\end{figure}

We compare point-clouds from simulation and the real world for the Bins and Shelf task and analyze their properties. We replicate Bins Scene 4 and Shelf Scene 1 in simulation: simply measure the dimensions and positions of the real world objects and set those dimensions in simulation using the OpenBox and Shelf procedural assets. As seen in Fig.~\ref{fig:sim real point-cloud vis}, simulated point-clouds are far cleaner than those in the real world, which are noisy and perhaps more importantly, partial. The real-world point-clouds often have portions missing due to camera coverage as for large objects it is challenging to cover the scene well while remaining within the depth camera operating range. However, we find that our policy is still able to able operate well in these scenes, as PointNet++ is capable of handling partial point-clouds and is trained on a diverse dataset containing many variations of boxes and shelves with different types and number of components as well as sizes, which may enable the policy to generalize to partial boxes and shelves observed in the real world.

\chapter{Chapter 6 Appendix}
\section{Table of Contents}
\begin{itemize}
    \item \textbf{RL Training Details} (Appendix~\ref{app:rl details}): Additional analysis of our real world results, with more detailed experiments.
    \item \textbf{Distillation Training Details} (Appendix~\ref{app:distillation details}): Ablations and analyses of Neural MP, demonstrating the effectiveness of our design decisions.
    \item \textbf{Experiment Details} (Appendix~\ref{app:experiment details}): Description of all the environments we use in this work.
\end{itemize}

\clearpage

\section{RL Training Details}
\label{app:rl details}
In this section, we provide a detailed description of the data generation process, the exact reward definitions and the specific hyper-parameters we use to train our skills.

\subsection{Data Generation}

For training generalist pick and place skills, we require a large dataset of common objects that can simulate well with contact. As a result, we train policies using the UnidexGrasp dataset~\cite{xu2023unidexgrasp} which contains 3.5K objects of 133 categories such as bowls, cups, bottles, cameras, remotes, etc. Since our policies are local, we can simply generate scenes with a single object spawned on a table top. However, such an agent may not generalize well to manipulation in tight spaces and among clutter, when it needs to perform local obstacle avoidance and constrained manipulation. For robustness, we train with randomly sampled clutter objects (UnidexGrasp) and obstacles (cuboids) that we spawn in the scene. 

For picking, to define an initial state distributions which ensures locality (within $\epsilon$ of the target object), we sample poses in a half-sphere above the table with radius $\epsilon=0.08$ and an additional error tolerance of 0.05 around the target object that are always pointing towards the object (ensuring the object is visible from the wrist camera). For placing, we execute the pick policy and then sample initial poses in a cuboid of with side length $\epsilon=0.2$ around the picking pose. 
In both cases we ensure that the sampled poses are not in contact with anything in the environment (aside from the in-hand object if present).

For local articulated object manipulation of objects such as doors and drawers, the design and global structure of the asset is not important. In fact, the only component of interest to the local policy is the handle. Accordingly, we sample from a dataset of 2.6K door and drawer handles from the PartNet dataset~\cite{mo2019partnet} and build door and drawer assets out of cuboids (Fig.~\ref{fig:sim vis}) as they are straightforward to randomize. We define drawers as boxes to be pulled straight out and doors as boxes to be opened using a vertical hinge joint. We randomize the size, shape, position, orientation, articulated joint range, friction and damping coefficients of the articulated objects, which covers a wide set of real world articulated objects. Detailed randomization distributions are presented in Tab.~\ref{tab:randomization_details}. For the grasp handle skill, we sample initial poses pointing toward the handle in a half sphere (in this case vertical half-sphere, away from the door) with radius $\epsilon=0.08$ and an additional error tolerance of 0.05. For opening and closing, after sampling a random initial pose of the articulated joint, we execute the grasp handle policy and add a small noise to ensure diversity of the final end-effector pose.

Finally, we collect valid pre-grasp and rest poses in simulation to help train our local policies. Specifically, we randomly sample grasp poses on the object mesh using antipodal sampling~\cite{sundermeyer2021contact} (1K per rest pose for UniDexGrasp objects and 2.5K for PartNet objects). We then move the Franka arm to pick/grasp the handle of the object using the pre-sampled grasp poses and save the successful poses. We also utilize the success rate of this scheme to filter out rest object poses that are not graspable (\textit{e.g.}, an upside down bowl). We also  generate a wide set of object rest poses for training the placing policy using the initial poses from the UnidexGrasp dataset, and augmenting them by rotating about the z-axis with 8 different angles and testing to ensure the objects remain at rest.

\begin{table}[H]
    \centering
    \begin{tabular}{c|c}
    \toprule[1.2pt]
       Parameter & Range / Distribution \\
       \midrule
        \multicolumn{2}{c}{\textit{UniDexGrasp objects}} \\
        Object size & $[0.06, 0.30]$\\
        Initial object position (XY) & $\mathcal{U}(0.3, 0.7) \times \mathcal{U}(-0.2, 0.2)$\\
        Initial object rotation (Z-axis) & $\mathcal{U}(-\pi, \pi)$ \\
        \midrule
        \multicolumn{2}{c}{\textit{Articulated objects}} \\
        Door size & $\mathcal{U}(0.25, 0.40) \times \mathcal{U}(0.20, 0.50)$\\
        Door damping & $\mathcal{U}(0.01, 0.02)$\\
        Door friction & $\mathcal{U}(0.025, 0.050)$\\
        Door joint range & $[0, \pi/2 ]$ \\
        Drawer size & $\mathcal{U}(0.25, 0.50) \times \mathcal{U}(0.08, 0.25)$\\
        Drawer damping & $\mathcal{U}(0.10, 0.20)$\\
        Drawer friction & $\mathcal{U}(0.25, 0.50)$\\
        Drawer joint range & $[0, 0.3]$ \\
        Distance to robot base & $\mathcal{U}(0.65, 0.75)$\\
        Object Orientation & $\mathcal{U}(-\pi/2, \pi/2)$\\
    \bottomrule[1.2pt]
    \end{tabular}
    \caption{Randomization for data generation.}
    \label{tab:randomization_details}
\end{table}

\subsection{Rewards}
We provide additional details on how to specify rewards for each skill. Specifically, we define $r_{ee}, r_{obj}, r_{ee,obj}$ and $r_{action}$.

\noindent \textbf{Pick} involves moving the gripper so the object can be easily grasped. 
Instead of encouraging the agent to move towards the overall object pose, which is not necessarily the pose to achieve for grasping, we provide dense signal for learning to grasp using pre-sampled grasp poses ($\{X_{targ}\}$). We encourage the agent to minimize the key-point distance~\cite{allshire2022transferring} to the nearest grasp pose:
\begin{align*}
    r_{ee,grasp} = \sum_{i=1}^Ne^{\min_{\{X_{targ}\}}||X_{ee}^i-X_{targ}^i||^2}
\end{align*}

Another challenge is that of picking in tight spaces, in which the policy changing directions and too frequently may cause damage and failure to complete the task. Thus, we encourage the agent to minimize its change in gripper orientation while interacting, by adding a penalty term on the angle between the current and previous gripper pose along the gripper's central axis ($\varv$), 
\begin{align*}
r_{ee, init} = -\arccos((R_{ee}^t\varv)(R_{ee}^{t-1}\varv))
\end{align*}
We also add a term to minimize the contact force on the gripper which discourages contact with any part of the scene. 
\begin{align*}
    r_{ee,obj} = \max (\max(f_{left}, f_{right}), 0)
\end{align*}
Finally, when picking, the agent should try to minimize moving the target object: we set 
\begin{align*}
r_{obj} = e^{||X_{obj,xy}^t - X_{obj,xy}^0||^2}
\end{align*}
 to penalize changes in object pose. All other reward components have a constant set to 0.
 
\noindent \textbf{Place} requires the agent to carefully set the object down near the initial pose. 
To provide dense signal for placing and ensure stability, we use a key-point distance reward on the object pose to encourage it to reach a stable absolute rest pose ($r_{obj}$) 
\begin{align*}
    r_{obj} = e^{\sum_{i=1}^{8}||X_{obj,FPS}^i - X_{rest,FPS}^i||^2}
\end{align*}
and a key-point distance reward to the nearest grasp pose to encourage the agent to maintain a stable pose relative to the object itself while moving ($r_{ee}$). This reward is the same as in pick.

\noindent \textbf{Grasp Handle} enables the agent to be able to grasp the handle of any door or drawer, a necessary skill to interact with articulated objects. This skill uses similar rewards to pick, but adapted for articulation: 1) key-point-based rewards for reaching pre-grasp poses ($r_{ee}$) 2) restricting the gripper orientation in task irrelevant directions ($r_{ee, z}$) 3) restricting the gripper orientation relative to the handle in the x-axis ($r_{ee,obj}$) 4) discouraging the agent from moving the object by minimizing the motion of the joint ($r_{obj}$).

Same as in pick, we wish to use pre-grasp poses in order to provide dense signal to the agent and ensure that it does not change orientation too much. We use the same key-point-based reward $r_{ee,grasp}$, but instead restrict the policy's movement in the z-direction, an axis along which motion is not beneficial to solving the task 
\begin{align*}
    r_{ee,z} = -|X_{ee,z}^t-X_{ee,z}^0|
\end{align*}
We further encourage the agent to only move in task relevant directions, by encouraging the agent to minimize the angle between the gripper and handle, in the x-axis of the door frame:
\begin{align*}
    r_{ee,obj} = e^{\min((R_{handle}^tR_{ee}^t\varv)_x + thresh, 0)}
\end{align*}
Similar to pick, we discourage the agent from moving the object, in this case by minimizing the motion of the joint 
\begin{align*}
 r_{obj} = -|q^t - q^0|
\end{align*}

\noindent \textbf{Open and Close} involve opening and closing articulated objects, having already grasped the handle. We can solve the task with a dense absolute difference reward ($r_{obj}$) between the current and target joint angles (0 for close, and $q_{limit}$, the joint limit of the object for open). To ensure the agent maintains its grasp while smoothly moving the articulated to joint to the desired configuration, we use $r_{ee,z}$ from Grasp handle, penalize movement relative to the door/drawer handle ($r_{ee,obj}$), and discourage ($r_{action}$) taking actions that cause the handle to slip out (moving in the y or z axes).

These skills utilize a dense reward for solving the task 
\begin{align*}
    r_{obj} = |q^t - q_{targ}|
\end{align*} where $q_{targ}$ is 0 for close and is $q_{limit}$ the joint limit of the object for open. As with grasp handle, we restrict the policy's movement along the z-axis using $r_{ee,z}$. We additionally penalize any movement of the end-effector in the handle frame, 
\begin{align*}
    r_{ee,obj} = ||R_{handle}^tX_{ee}^t - R_{handle}^{t-1}X_{ee}^{t-1}||
\end{align*} and any sampled actions that cause the agent to move in the y or z axes: 
\begin{align*}
    r_{action} = -||(R_{handle}^t(a))_{yz}||^2
\end{align*}
These rewards ensure that the agent maintains its grasp while smoothly moving the articulated to joint to the desired configuration.

We train all RL policies with PPO~\cite{schulman2017proximal}. Detailed hyper-parameters are presented in Tab.~\ref{tab:rl_training_details} and Tab.~\ref{tab:rl_training_epochs}.
\begin{table}[H]
    \centering
    \begin{tabular}{c|c}
    \toprule[1.2pt]
       Hyperparameter & Value \\\hline
        Num. envs (Isaac Gym, state-based) & 8192\\
        Num. rollout steps per policy update & 32\\
        Num. learning epochs & 5\\
        Episode length & 120\\
        Discount factor & 0.99\\
        GAE parameter & 0.95\\
        Entropy coeff. & 0.0\\
        PPO clip range & 0.2\\
        Learning rate & 0.0005\\
        KL threshould for adaptive schedule & 0.16 \\
        Value loss coeff. & 4.0\\
        Max gradient norm & 1.0 \\
    \bottomrule[1.2pt]
    \end{tabular}
    \caption{Hyper-parameters for PPO.}
    \label{tab:rl_training_details}
\end{table}

\begin{table}[h]
    \centering
    \begin{tabular}{c|ccc}
    \toprule[1.2pt]
       Skill & \#Max epoch & \#Save best & \#Early stop  \\\hline
        Pick & 500 & 100 & 200 \\
        Place & 500 & 100 & 200 \\
        Grasp Handle & 500 & 100 & 100 \\
        Open & 500 & 100 & 100 \\
        Close & 500 & 100 & 100 \\
    \bottomrule[1.2pt]
    \end{tabular}
    \caption{Training epochs and early stop criterion for each task. \#Max epoch is the maximum number of iterations to run. \#Save best is the first iteration to begin saving best checkpoints. \# Early stop suggests terminating early if there is no improvement after certain number of iterations.}
    \label{tab:rl_training_epochs}
\end{table}

\section{Distillation Training Details}
\label{app:distillation details}
\subsection{Multitask DAgger}
In our multitask DAgger implementation, we incorporate a replay buffer of size $K$ that holds the last $K\times B$ trajectories in memory. The training process alternates between updating the policy for a single epoch on this buffer and collecting a batched set of trajectories (size $B$) from the environment for the current object. 
In practice, we find that K=100, B=32 performs well, which means for a single object we collect 32 simultaneous trajectories at a time, and we can hold data for up to 100 objects in our buffer which is constantly refreshed as we collect new data. A practical issue is loading all objects from the dataset into simulation simultaneously is unfeasible. To address this, we split the dataset into batches of 100 objects and sequentially launch training on each batch for 100 epochs. Detailed multitask DAgger parameters are presented in Tab.~\ref{tab:dagger_training_details}.

\begin{table}[H]
    \centering
    \begin{tabular}{c|c}
    \toprule[1.2pt]
       Hyperparameter & Value \\\hline
        Num. envs (Isaac Gym, vision) & 128\\
        Episode length & 120\\
        Num. rollout steps per policy epoch & 120\\
        Num. learning epochs & 1\\
        Buffer size & 100 * 128 \\
        Learning rate & 0.0001\\
        Batch size & 2048 \\
    \bottomrule[1.2pt]
    \end{tabular}
    \caption{Hyper-parameters for Multitask DAgger.}
    \label{tab:dagger_training_details}
\end{table}

\subsection{Data Augmentation}
In addition to random camera cropping, we also apply \textit{edge noise} and \textit{random holes} to enhance robustness to real world observations.

\textbf{Edge artifacts} To model the noisiness along object edges, we use the correlated depth noise via bi-linear interpolation of shifted depth.  Given a depth map of size $H\times W$, we construct a grid $\{0,\cdots, H-1\}\times \{0,\cdots, W-1\}$. For each node on the grid, we apply a random shift $\mathcal{N}(0, 0.5)$ with probability $0.8$. We then perform bilinear interpolation between the original depth values and the adjusted grid to generate a new depth map.

\textbf{Random holes} We observe that, even after hole filling, the real world depth maps still contain irregular holes (especially for reflective surfaces and dim environments). To model these holes, we create a random pixel-level mask from $\mathcal U(0, 1)$. This mask is smoothed with Gaussian blur and normalized to the range $[0, 1]$. Based on the mask, we zero out pixels with mask values exceeding a threshold randomly sampled from $\mathcal U(0.6, 0.9)$. The randomization is applied to a depth map with probability $0.5$.

\begin{figure}[ht!]
    \centering
    \includegraphics[width=1.0\linewidth]{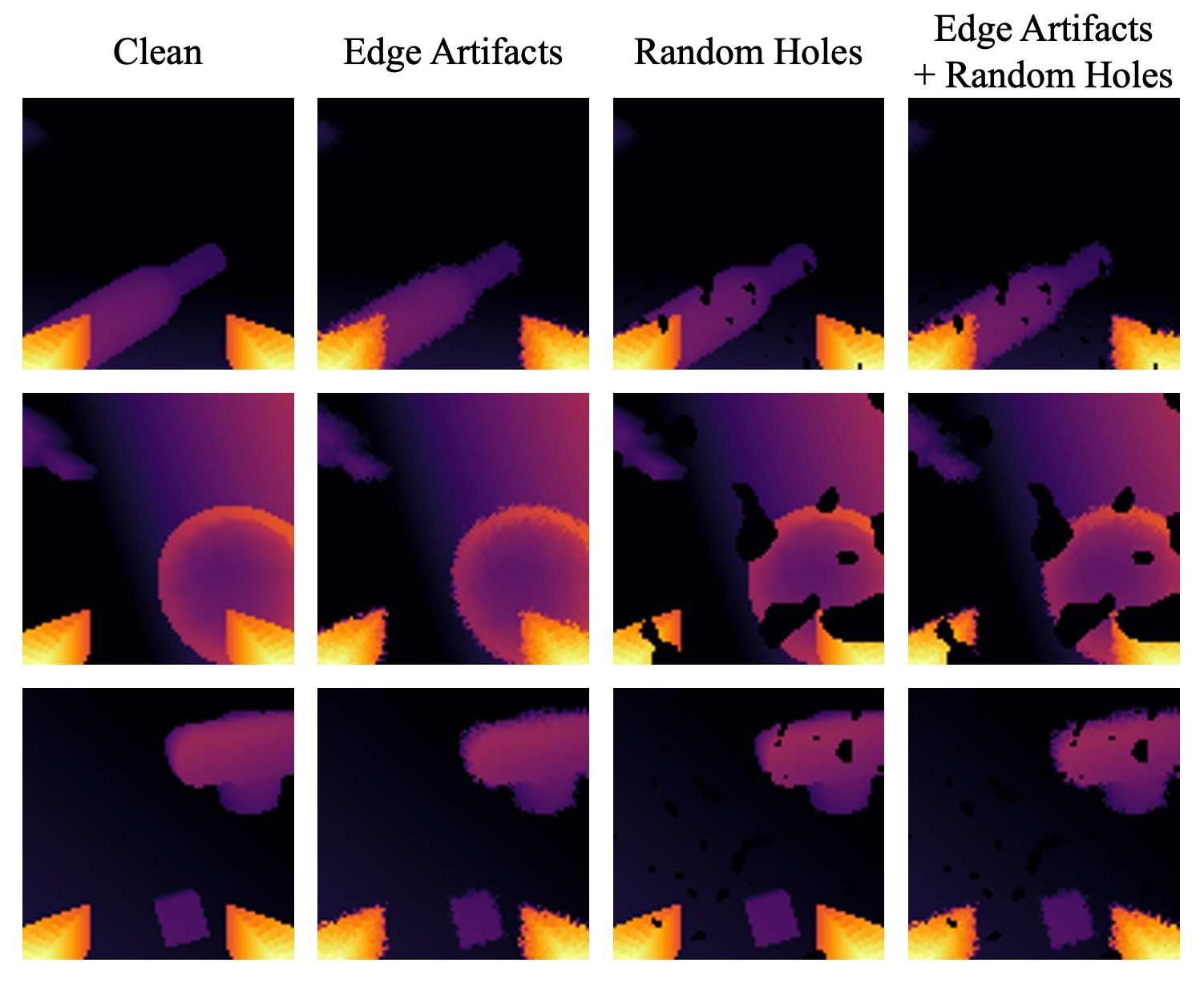}
    \caption{\textbf{Depth Augmentation} Visualization of edge artifacts and random holes on depth maps.}
    \label{fig:vis_depth_aug}
\end{figure}

We summarize hyper-parameters for DAgger data augmentation in Tab.~\ref{tab:dagger data aug} and visualize them in Fig.~\ref{fig:vis_depth_aug}. Note that the resolution of our depth map is $84\times 84$. The visual effect with these hyper-parameters will change on different resolutions.

\begin{table}[H]
    \centering
    \begin{tabular}{c|c}
    \toprule[1.2pt]
       Hyper-parameter & Value \\\hline
       \textit{Edge Noise} & \\
       Gaussian Noise Std & 0.5 \\
       Noise Accept Prob & 0.8 \\\hline
       \textit{Random Holes} & \\
       Gaussian Blur Kernel Size & [3, 27] \\
       Gaussian Blur Std & [1, 7] \\
       Mask Threshold & [0.7, 0.9] \\
       Hole Keep Prob & 0.5 \\
    \bottomrule[1.2pt]
    \end{tabular}
    \caption{Hyper-parameters for DAgger data augmentation.}
    \label{tab:dagger data aug}
\end{table}

\section{Deployment Details}
In this section, we describe our real-world deployment system in detail. 
We begin by providing a high-level overview of ManipGen deployment in pseudocode below:

\begin{figure}
    \centering
    \includegraphics[width=\linewidth]{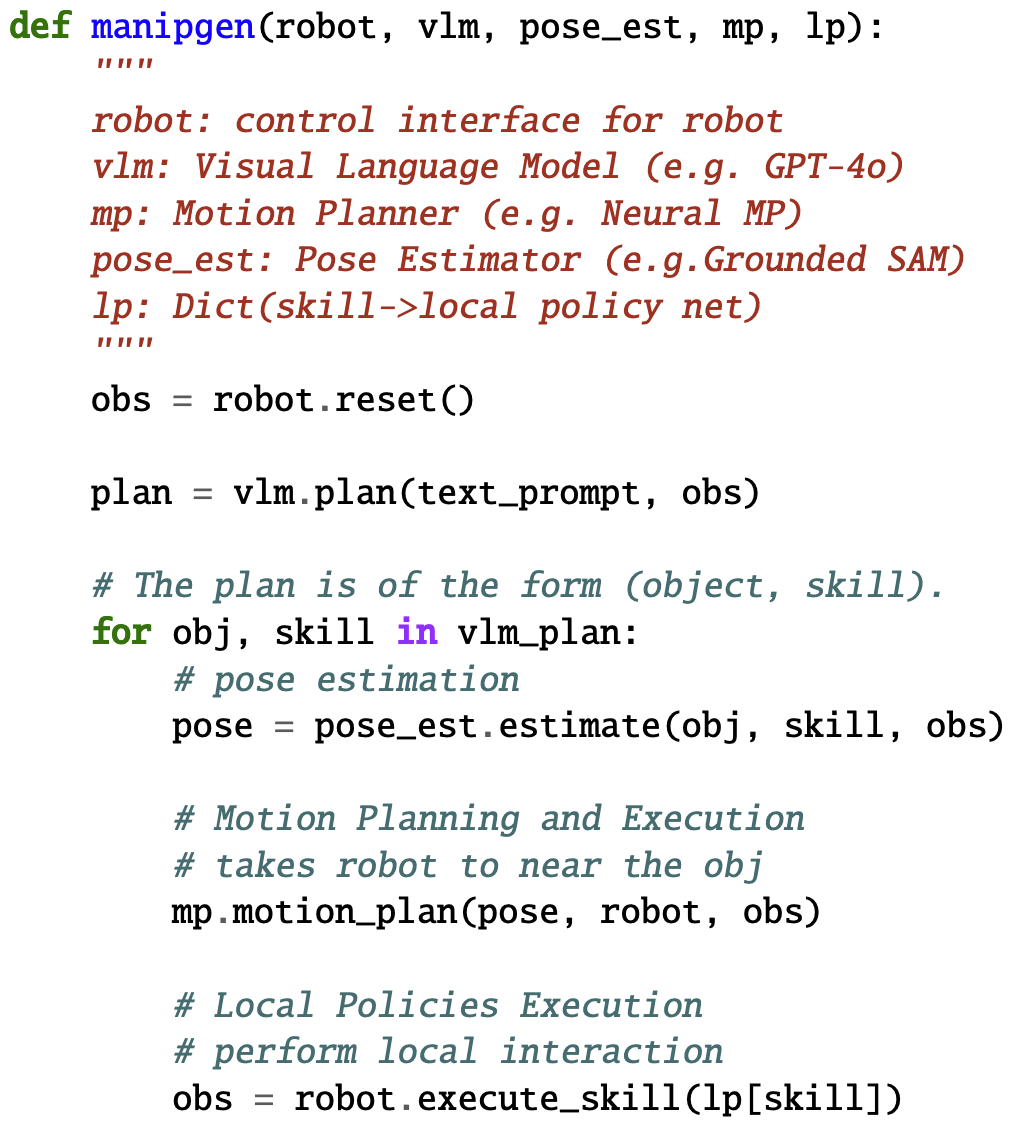}
\end{figure}

\textbf{Tagging} One challenge with Grounded SAM is prompting: the Grounding Dino module is quite sensitive to the input prompt and does not always detect the correct object unless the prompt is formatted well (descriptive, includes colors and texture and shape). Since the VLM will be used to prompt SAM with the target object for each skill, we need the VLM to output tags for objects such that Grounded SAM will trigger on the correct object. As a result, we include an initial tagging phase, in which we have the VLM list all objects that are present in the scene, which we then pass into Grounded SAM to get a list of tags and associated segmentation masks based on how Grounded SAM interprets the scene. We then pass the tagged image, as well as the original image back into the VLM for planning

\textbf{Planning} For the VLM to output a plan (of the same format as in PSL), we provide a system prompt to the VLM that provides it a detailed description of the skill library, their effects on the environment and when they can be used. We also provide it a list of hints as to what consistitutes reasonable plans, this is necessary as existing VLMs still lack strong spatial reasoning capabilities inherently, though prompting seems to alleviate this issue to a certain degree. We include several in-context examples that allow the model to understand the output format, prompt the model to justify its decisions (which helped produce better plans) and format the output as a JSON string (which resolved most of the parsing/formatting issues). 

After planning, ManipGen loops through the plan, estimating the pose of the object/region of interest, motion planning there, and then executing the appropriate local policy. 

\textbf{Pose Estimation} 
We begin by using Grounded-SAM to segment the object of interest, averaging the 3D points of the segmented pixels to determine its position. Orientation is then estimated depending on specific cases. Based on whether there is obstacle above the object, we leverage VLM to classify pick and place tasks into two scenarios: open-space (\textit{e.g.}, table surfaces) and tight-space (\textit{e.g.}, microwaves). We let the robot gripper point downwards in open-space setting. For open-space picking, we project the object’s points onto the XY plane and apply damped least square to fit the points, estimating the object’s longest axis to get gripper's rotation along Z-axis. For open-space placing, we simplify the problem by selecting a fixed orientation pointing down.
In tight-space pick and place, we first sample a set of robot poses around the object. We then capture point cloud of the current scene (excluding the robot) and evaluate the number of points in collision with each sampled pose. To bias towards poses that are further from obstacles, we apply Gaussian noise $\mathcal{N}(0.0, 0.1)$ to the points, and select the pose with minimal collision. For articulated objects, we estimate position of the handle and whether it is vertical or horizontal. Then we sample a set of target poses around the handle and use the same collision-checking method as in tight-space scenario to identify the target pose.

\textbf{Motion Planning}
We use the released Neural MP code and checkpoint to perform motion planning given the target pose and current point-cloud. We perform open-loop motion planning with test-time optimization batch size of 64 (the paper used 100) and max path length of 100. Neural MP can produce paths that are jerky and non-smooth at times. As a result, we perform EMA smoothing with $\alpha=0.9$ to reduce the jerkiness of the trajectories. 

\textbf{Local Policies}
Once initialized near the object of interest, we segment the target object in the first frame using Grounded SAM). We store this mask along with the depth map of the first frame and then we deploy the local policy. At each step, we pass in the segmentation mask of the target object, the first frame depth map, the current frame depth map and the proprioception to the policy. We run the Task Space Impedance Controller on the robot at 60Hz for a fixed, skill specific duration (max 8s). Then, depending on the skill, we either open or close the gripper and begin executing the next stage.

\section{Experiment Details}
\label{app:experiment details}
\textbf{Hardware} For all of our experiments, we use a Franka Emika Panda Robot, which is a 7 degree of freedom manipulator arm. We control the robot using the Industreallib library (\href{https://github.com/NVlabs/industreallib}{https://github.com/NVlabs/industreallib}) using the Task Space Impedance Controller. For deployment, we use Leaky PLAI with action scales, thresholds and skill deployment durations chosen per skill (Table~\ref{tab:real skill configs}). For all skills, we used a position gain of 1000 and rotation gain of 50.

\begin{table}[H]
    \centering
    \resizebox{\linewidth}{!}{%
    \begin{tabular}{c|c|c|c|c|c}
    \toprule[1.2pt]
        & Pick & Place & Grasp Handle & Open & Close \\\hline
       Duration & 5s & 4s & 6s & 8s & 4.5s \\
       Action Scale Pos & .002 & .002 & .002 & .003 & .005 \\
       Action Scale Rot & .05  & .05 & .004 & .0005 & .0005 \\ 
       Leaky PLAI Thresh X, Y & .02 & .02 & .02 & 0.02 & 0.03 \\
       Leaky PLAI Thresh Z & .02 & .02 & .02 & .003 & .003 \\
       Leaky PLAI Thresh Rot (Deg) & 4 & 4 & 2 & .005 & .005 \\
    \bottomrule[1.2pt]
    \end{tabular}
    }
    \caption{Configurations for skill deployment.}
    \label{tab:real skill configs}
\end{table}

The robot is mounted to a fixed base pedestal behind a desk of size .762m by 1.22m with variable height. For global views (used for the VLM, pose estimation and Neural MP), we use four calibrated depth cameras, Intel Realsense 455, placed around the scene in order to accurately capture the environment. 
We project the depth maps from each camera into 3D and combine the individual point-clouds into a single scene representation for Neural MP and pose estimation.
For input to Neural MP, we further process the point-clouds according to the paper~\cite{dalal2024neuralmp}. While the VLM and the pose estimators can take in multiple views in principle, in practice we found that their results (predicted plans, pose estimations) were significantly less reliable and consistent when using more than a single camera. As a result, for each task, we select one out of the four global view cameras to use for plan prediction and pose estimation. Finally, for local views, we use the d405 camera mounted on the wrist, and pass its depth maps (after clamping and normalization) as input to the policy.

\subsection{Simulated Comparison (Robosuite) Details}
We zero-shot transfer our local policies (trained in IsaacGym) to the robosuite tasks. Note, we do not train on the Robosuite objects at all, we use our data generation and training pipeline to train local policies in IsaacGym and transfer the policies to Robosuite. To deploy our method, we modify the environment to use the UMI~\cite{chi2024universal} gripper and Task Space Impedance Control. We use the same LLM-planning and motion planning infrastructure as used in the PSL~\cite{dalal2024psl} paper and evaluate our policies using 100 trials per task. All other numbers for baselines are taken from the PSL paper.

\subsection{Furniture Bench Experiment Details}
We replicate the exact task setup from the Transic~\cite{jiang2024transic} paper: 3D printed objects from FurnitureBench~\cite{heo2023furniturebench} in the exact poses specified in their paper. In this experiment, we train local policies for these specific tasks and then deploy them. As a result, this experiment compares our method of sim2real transfer (local policies) with end-to-end sim2real transfer (w/ and w/o data aug) as well as Transic (which uses real-world data). 

\subsection{Real World Long-horizon Manipulation Details}

\begin{figure*} [t!]
    \centering
    \subfloat[Cook]{
        \includegraphics[width=0.24\linewidth]{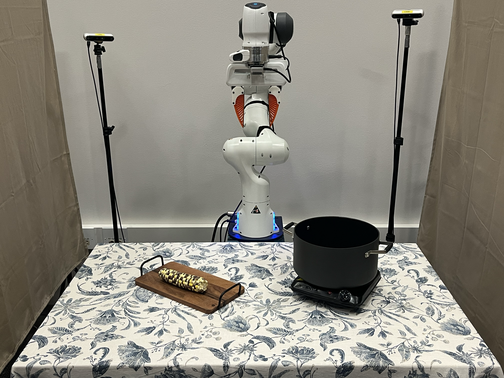}
        \includegraphics[width=0.24\linewidth]{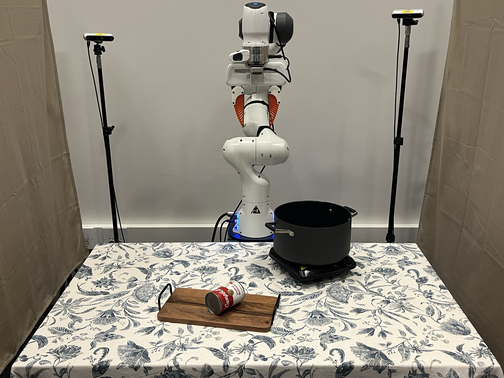}
        \includegraphics[width=0.24\linewidth]{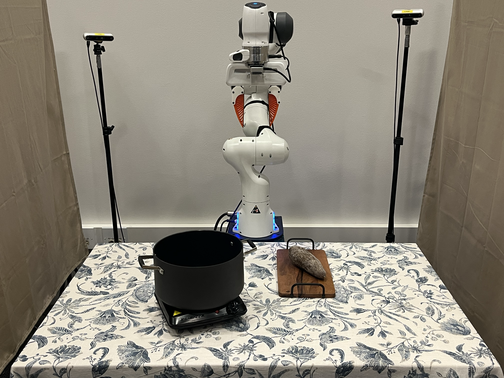}
        \includegraphics[width=0.24\linewidth]{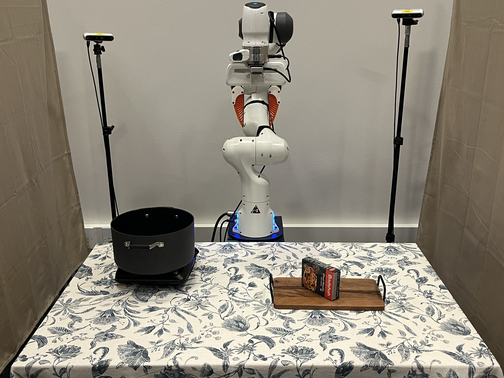}
    }
    \vspace{0.4em}
    \\
    \subfloat[Replace]{
        \includegraphics[width=0.24\linewidth]{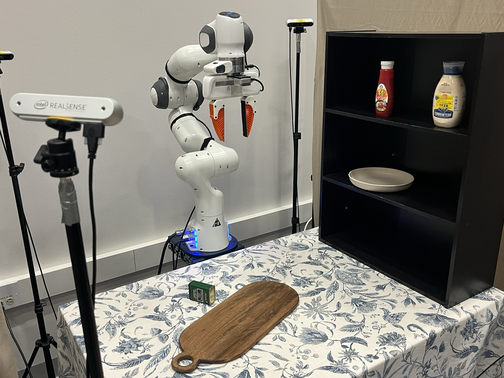}
        \includegraphics[width=0.24\linewidth]{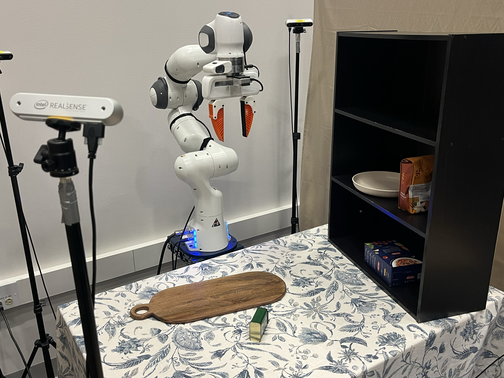}
        \includegraphics[width=0.24\linewidth]{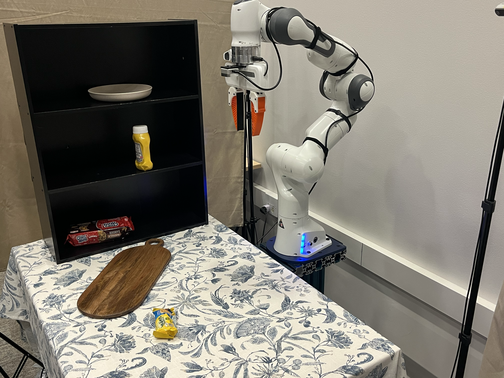}
        \includegraphics[width=0.24\linewidth]{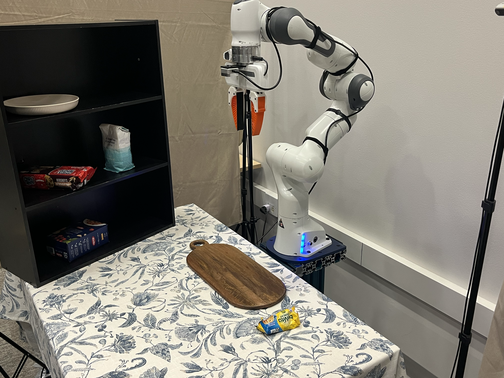}
    }
    \vspace{0.4em}
    \\
    \subfloat[Cabinet Store]{
        \includegraphics[width=0.24\linewidth]{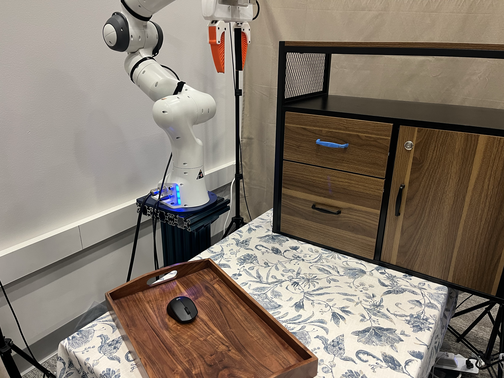}
        \includegraphics[width=0.24\linewidth]{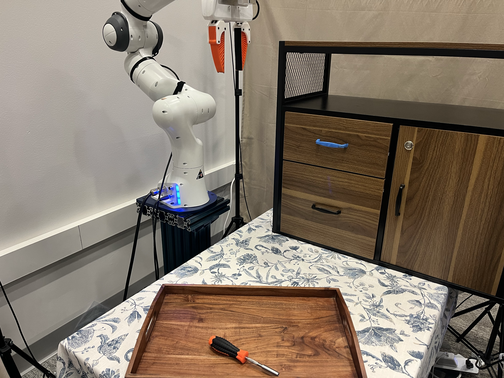}
        \includegraphics[width=0.24\linewidth]{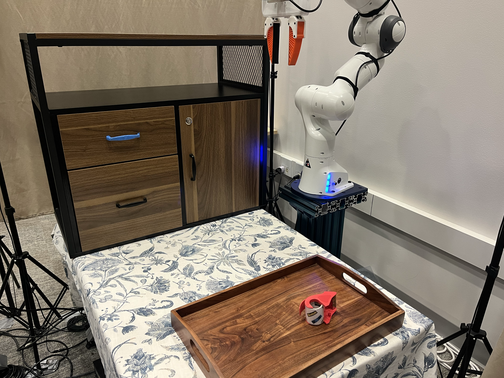}
        \includegraphics[width=0.24\linewidth]{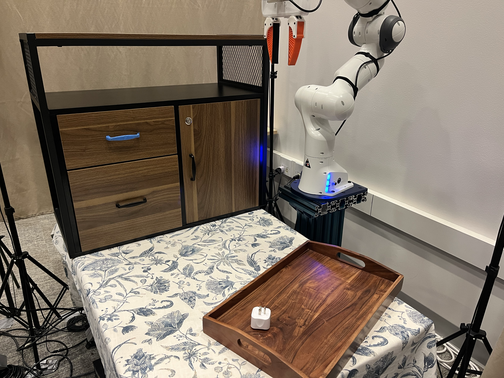}
    }
    \vspace{0.4em}
    \\
    \subfloat[Drawer Store]{
        \includegraphics[width=0.24\linewidth]{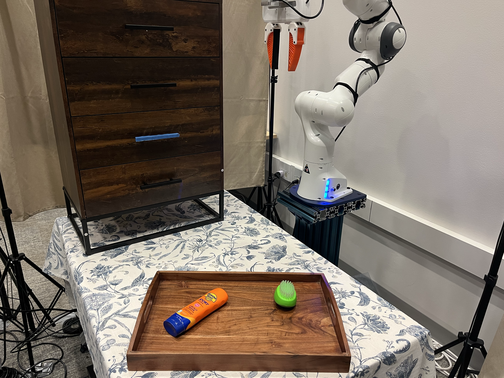}
        \includegraphics[width=0.24\linewidth]{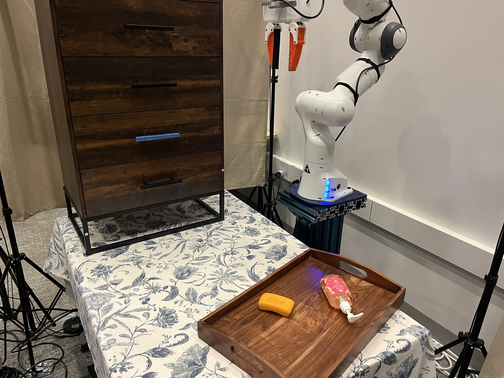}
        \includegraphics[width=0.24\linewidth]{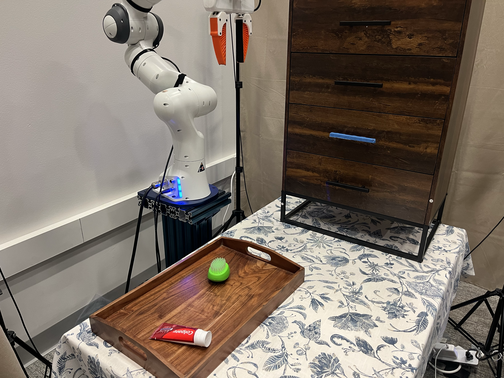}
        \includegraphics[width=0.24\linewidth]{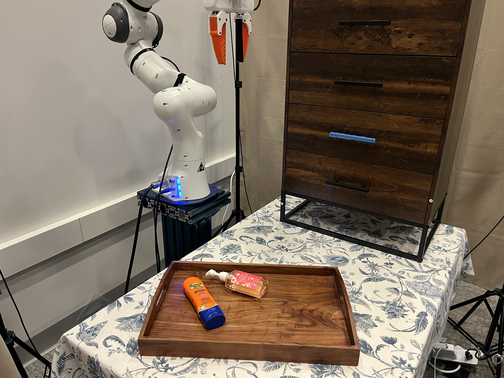}
    }
    \vspace{0.4em}
    \\
    \subfloat[Tidy]{
        \includegraphics[width=0.24\linewidth]{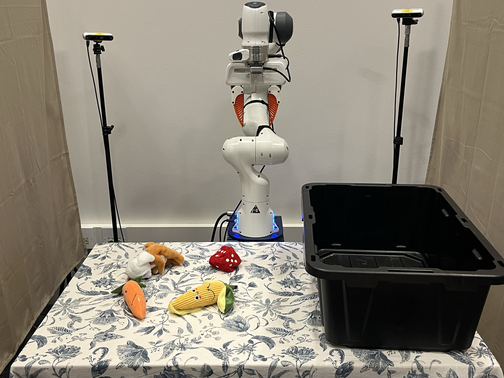}
        \includegraphics[width=0.24\linewidth]{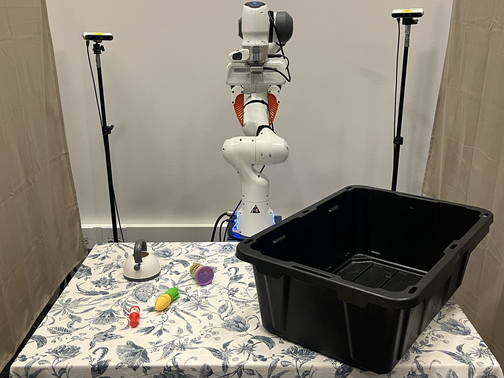}
        \includegraphics[width=0.24\linewidth]{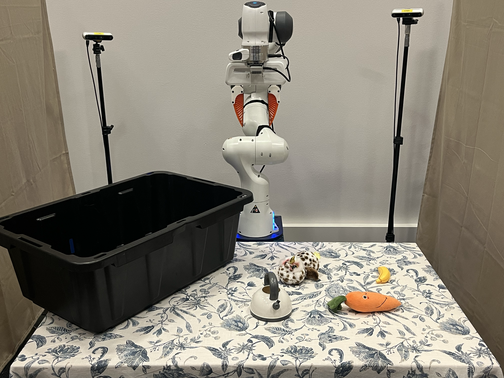}
        \includegraphics[width=0.24\linewidth]{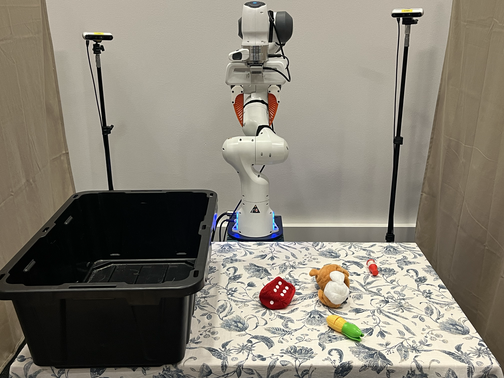}
    }
    \caption{
    Example scene layouts for real world evaluation.
    }
    \label{fig:real_eval_images} 
\end{figure*}

For each of the following tasks, we specify the large object (if present) in each task (as a receptacle) as well as the list of objects we randomize, the task description in detail, and the task prompt provided to ManipGen and the baselines.
Unless otherwise stated, all objects (handles for articulated objects) to interact with are positioned within 0.8 meters of the robot base to ensure they are within the gripper’s reach.
Some example scene arrangements are presented in Fig.~\ref{fig:real_eval_images}.

\subsubsection{Cook (2 stages)} Pick up a food item on the cutting board and put it in a pot on the stove.

Task prompt: \textcolor{ForestGreen}{Put [OBJ] in the black pot.}

Large object: black pot (39cm $\times$ 32cm $\times$ 14cm)

Randomized objects: carrot (17.0cm $\times $ 2.5cm $\times $ 2.5cm), cassava (20.0cm $\times$ 6.1cm $\times$ 6.0cm), corn (17.7cm $\times$ 4.2cm $\times$ 4.3cm), spice box (12.8cm $\times$ 9.0cm $\times$ 3.0cm), soup can (10.0cm  $\times$ 6.7cm $\times$ 6.7cm)

Randomization: We put the food item on a cutting board, the pot on a stove, and randomize their poses across the table within the gripper's reach.

\subsubsection{Replace (4 stages)} Fetch a pantry item from the shelf, put it on a wooden board on the table and take an object from the table, put it on a white plate.

Task prompt: \textcolor{ForestGreen}{Place [OBJ A] on the wooden board, and put [OBJ B] on the white plate in the shelf.}

Large objects: wooden board (51cm $\times$ 18.8cm $\times$ 1.2cm), shelf (80cm $\times$ 60cm $\times$ 23cm)

Randomized objects: [OBJ A] brown coffee package (15.7cm $\times$ 8.0cm $\times$ 6.0cm), blue coffee package (15.7cm $\times$ 8.0cm $\times$ 6.0cm), ketchup bottle (17.5cm $\times$ 9.3cm $\times$ 5.5cm), mustard bottle (17.5cm $\times$ 9.3cm $\times$ 5.5cm), gochujang bottle (17.4cm $\times$ 6.6cm $\times$ 4.1cm); [OBJ B] spice jar (8.0cm $\times$ 4.0cm $\times$ 4.0cm), pepper container (8.2cm $\times$ 5.8cm $\times$ 3.1cm), biscuit pack (10.5cm $\times$ 5.6cm $\times$ 2.5cm)

Randomization: The shelf is placed on the left or right end of the table with randomized orientation between 0 and 30 degrees. [OBJ A] and the white plate are randomly placed on the second or third level of the shelf. The wooden board and [OBJ B] are randomly placed on the other half of the table. We ensure that center of the board and [OBJ B] are at least 30cm away from base of the shelf.

\subsubsection{CabinetStore (4 stages)} Open the drawer with blue handle in the cabinet, put
an office supply inside, and close the drawer.

Task Prompt: \textcolor{ForestGreen}{Store [OBJ] in the drawer with blue handle.}

Large objects: cabinet (80cm $\times$ 71cm $\times$ 40cm, \url{https://www.amazon.com/gp/product/B0CHHTZ52F/ref=ewc_pr_img_12?smid=A38QU35WKLBIKI&psc=1})

Randomized objects: computer mouse (10.5cm $\times$ 6.5cm  $\times$ 3.8cm), tape (10.6cm $\times$ 6.4cm $\times$ 6.3cm), screw driver (18.8cm $\times$ 3.2cm $\times$ 3.2cm), plug (4.9cm $\times$ 4.9cm $\times$ 5.4cm), staple container (10.1cm $\times$ 5.9cm $\times$ 4.0cm)

Randomization: The cabinet is placed on the left or right end of the table, with its back aligned to the table edge. The office supply is randomly placed on a wooden tray on the other half of the table. We ensure that after fully opening the drawer, the office supply is at least 20cm from edge of the drawer.

\subsubsection{DrawerStore (6 stages)} Open a drawer with blue handle, put two personal care items inside, and close the drawer.

Task Prompt: \textcolor{ForestGreen}{Arrange [OBJ A] and [OBJ B] in the drawer with blue handle.}

Large objects: drawer (80cm x 60cm x 23cm, \url{https://www.amazon.com/gp/product/B0BJPLBSHQ/ref=ewc_pr_img_1?smid=A2XKE81PYMCHT4&psc=1})

Randomized objects: brush (8.4cm $\times$ 6.4cm $\times$ 5.4cm), sunscreen bottle (18.9cm $\times$ 5.7cm $\times$  3.8cm), soap (9.4cm $\times$ 5.4cm $\times$ 2.5cm), toothpaste (14.1cm $\times$ 5.7cm $\times$ 3.6cm), sanitizer bottle (17.0cm $\times$ 6.8cm $\times$ 4.5cm)

Randomization: The drawer is placed on the left or right end of the table, with its back aligned to the table edge. We randomly select two personal care items and place them on a wooden tray on the other half of the table. We ensure that after fully opening the drawer, the personal care items are at least 20cm from edge of the drawer.

\subsubsection{Tidy (8 stages)} Clean up the table by putting 4 toys on the table into a bin.

Task Prompt: \textcolor{ForestGreen}{Sort all the toys into the black bin.}

Large objects: black bin

Randomized objects: stuffed carrot (17.2cm $\times$ 4.5cm $\times$ 4.5cm),  stuffed corn (21.3cm $\times$ 5.7cm $\times$ 6.0cm),  stuffed owl (15.0cm $\times$ 7.6cm  $\times$ 6.6cm),  stuffed dog (18.7cm $\times$ 7.8cm $\times$ 8.4cm),  stuffed dice (6.6cm $\times$ 6.6cm  $\times$  6.6cm),  tiny bottle (7.4cm $\times$ 2.9cm $\times$ 2.9cm),  toy teapot (15.1cm  $\times$  12.5cm  $\times$  12.5cm),  toy banana (8.8cm $\times$ 2.9cm $\times$ 4.0cm),  toy corn (12.0cm $\times$ 3.8cm $\times$ 3.8cm), play-doh container (7.6cm $\times$ 6.3cm $\times$ 6.3cm)

Randomization: The bin is placed on the left, right, or front end of the table with randomized orientation between 0 and 180 degrees. The toys are scattered on the other half of the table. We ensure the toys are at least 20cm away from edge of the bin.

\clearpage

{\small \bibliography{ref}}

\begin{thebibliography}{302}
\providecommand{\natexlab}[1]{#1}
\providecommand{\url}[1]{\texttt{#1}}
\expandafter\ifx\csname urlstyle\endcsname\relax
  \providecommand{\doi}[1]{doi: #1}\else
  \providecommand{\doi}{doi: \begingroup \urlstyle{rm}\Url}\fi

\bibitem[Achiam et~al.(2018)Achiam, Edwards, Amodei, and Abbeel]{achiam2018variational}
Joshua Achiam, Harrison Edwards, Dario Amodei, and Pieter Abbeel.
\newblock Variational option discovery algorithms.
\newblock \emph{arXiv preprint arXiv:1807.10299}, 2018.

\bibitem[Agarwal et~al.(2022)Agarwal, Kumar, Malik, and Pathak]{agarwal2022legged}
Ananye Agarwal, Ashish Kumar, Jitendra Malik, and Deepak Pathak.
\newblock Legged locomotion in challenging terrains using egocentric vision.
\newblock 2022.

\bibitem[Agarwal et~al.(2023{\natexlab{a}})Agarwal, Kumar, Malik, and Pathak]{agarwal2023legged}
Ananye Agarwal, Ashish Kumar, Jitendra Malik, and Deepak Pathak.
\newblock Legged locomotion in challenging terrains using egocentric vision.
\newblock In \emph{Conference on Robot Learning}, pages 403--415. PMLR, 2023{\natexlab{a}}.

\bibitem[Agarwal et~al.(2023{\natexlab{b}})Agarwal, Uppal, Shaw, and Pathak]{agarwal2023dexterous}
Ananye Agarwal, Shagun Uppal, Kenneth Shaw, and Deepak Pathak.
\newblock Dexterous functional grasping.
\newblock In \emph{7th Annual Conference on Robot Learning}, 2023{\natexlab{b}}.

\bibitem[Ahn et~al.(2022{\natexlab{a}})Ahn, Brohan, Brown, Chebotar, Cortes, David, Finn, Fu, Gopalakrishnan, Hausman, Herzog, Ho, Hsu, Ibarz, Ichter, Irpan, Jang, Ruano, Jeffrey, Jesmonth, Joshi, Julian, Kalashnikov, Kuang, Lee, Levine, Lu, Luu, Parada, Pastor, Quiambao, Rao, Rettinghouse, Reyes, Sermanet, Sievers, Tan, Toshev, Vanhoucke, Xia, Xiao, Xu, Xu, Yan, and Zeng]{saycan2022arxiv}
Michael Ahn, Anthony Brohan, Noah Brown, Yevgen Chebotar, Omar Cortes, Byron David, Chelsea Finn, Chuyuan Fu, Keerthana Gopalakrishnan, Karol Hausman, Alex Herzog, Daniel Ho, Jasmine Hsu, Julian Ibarz, Brian Ichter, Alex Irpan, Eric Jang, Rosario~Jauregui Ruano, Kyle Jeffrey, Sally Jesmonth, Nikhil Joshi, Ryan Julian, Dmitry Kalashnikov, Yuheng Kuang, Kuang-Huei Lee, Sergey Levine, Yao Lu, Linda Luu, Carolina Parada, Peter Pastor, Jornell Quiambao, Kanishka Rao, Jarek Rettinghouse, Diego Reyes, Pierre Sermanet, Nicolas Sievers, Clayton Tan, Alexander Toshev, Vincent Vanhoucke, Fei Xia, Ted Xiao, Peng Xu, Sichun Xu, Mengyuan Yan, and Andy Zeng.
\newblock Do as i can and not as i say: Grounding language in robotic affordances.
\newblock In \emph{arXiv preprint arXiv:2204.01691}, 2022{\natexlab{a}}.

\bibitem[Ahn et~al.(2022{\natexlab{b}})Ahn, Brohan, Brown, Chebotar, Cortes, David, Finn, Gopalakrishnan, Hausman, Herzog, Ho, Hsu, Ibarz, Ichter, Irpan, Jang, Ruano, Jeffrey, Jesmonth, Joshi, Julian, Kalashnikov, Kuang, Lee, Levine, Lu, Luu, Parada, Pastor, Quiambao, Rao, Rettinghouse, Reyes, Sermanet, Sievers, Tan, Toshev, Vanhoucke, Xia, Xiao, Xu, Xu, and Yan]{ahn2022say}
Michael Ahn, Anthony Brohan, Noah Brown, Yevgen Chebotar, Omar Cortes, Byron David, Chelsea Finn, Keerthana Gopalakrishnan, Karol Hausman, Alex Herzog, Daniel Ho, Jasmine Hsu, Julian Ibarz, Brian Ichter, Alex Irpan, Eric Jang, Rosario~Jauregui Ruano, Kyle Jeffrey, Sally Jesmonth, Nikhil~J Joshi, Ryan Julian, Dmitry Kalashnikov, Yuheng Kuang, Kuang-Huei Lee, Sergey Levine, Yao Lu, Linda Luu, Carolina Parada, Peter Pastor, Jornell Quiambao, Kanishka Rao, Jarek Rettinghouse, Diego Reyes, Pierre Sermanet, Nicolas Sievers, Clayton Tan, Alexander Toshev, Vincent Vanhoucke, Fei Xia, Ted Xiao, Peng Xu, Sichun Xu, and Mengyuan Yan.
\newblock Do as i can, not as i say: Grounding language in robotic affordances.
\newblock \emph{arXiv preprint arXiv: Arxiv-2204.01691}, 2022{\natexlab{b}}.

\bibitem[Ahn et~al.(2022{\natexlab{c}})Ahn, Brohan, Brown, Chebotar, Cortes, David, Finn, Gopalakrishnan, Hausman, Herzog, et~al.]{ahn2022can}
Michael Ahn, Anthony Brohan, Noah Brown, Yevgen Chebotar, Omar Cortes, Byron David, Chelsea Finn, Keerthana Gopalakrishnan, Karol Hausman, Alex Herzog, et~al.
\newblock Do as i can, not as i say: Grounding language in robotic affordances.
\newblock \emph{arXiv preprint arXiv:2204.01691}, 2022{\natexlab{c}}.

\bibitem[Ajay et~al.(2021)Ajay, Kumar, Agrawal, Levine, and Nachum]{ajay2021opal}
Anurag Ajay, Aviral Kumar, Pulkit Agrawal, Sergey Levine, and Ofir Nachum.
\newblock Opal: Offline primitive discovery for accelerating offline reinforcement learning, 2021.

\bibitem[Akgun et~al.(2012)Akgun, Cakmak, Yoo, and Thomaz]{akgun2012trajectories}
Baris Akgun, Maya Cakmak, Jae~Wook Yoo, and Andrea~Lockerd Thomaz.
\newblock Trajectories and keyframes for kinesthetic teaching: A human-robot interaction perspective.
\newblock In \emph{Proceedings of the seventh annual ACM/IEEE international conference on Human-Robot Interaction}, pages 391--398, 2012.

\bibitem[Akkaya et~al.(2019)Akkaya, Andrychowicz, Chociej, Litwin, McGrew, Petron, Paino, Plappert, Powell, Ribas, et~al.]{akkaya2019solving}
Ilge Akkaya, Marcin Andrychowicz, Maciek Chociej, Mateusz Litwin, Bob McGrew, Arthur Petron, Alex Paino, Matthias Plappert, Glenn Powell, Raphael Ribas, et~al.
\newblock Solving rubik's cube with a robot hand.
\newblock \emph{arXiv preprint arXiv:1910.07113}, 2019.

\bibitem[Allshire et~al.(2021)Allshire, Mart{\'\i}n-Mart{\'\i}n, Lin, Manuel, Savarese, and Garg]{allshire2021laser}
Arthur Allshire, Roberto Mart{\'\i}n-Mart{\'\i}n, Charles Lin, Shawn Manuel, Silvio Savarese, and Animesh Garg.
\newblock Laser: Learning a latent action space for efficient reinforcement learning.
\newblock \emph{arXiv preprint arXiv:2103.15793}, 2021.

\bibitem[Allshire et~al.(2022)Allshire, MittaI, Lodaya, Makoviychuk, Makoviichuk, Widmaier, W{\"u}thrich, Bauer, Handa, and Garg]{allshire2022transferring}
Arthur Allshire, Mayank MittaI, Varun Lodaya, Viktor Makoviychuk, Denys Makoviichuk, Felix Widmaier, Manuel W{\"u}thrich, Stefan Bauer, Ankur Handa, and Animesh Garg.
\newblock Transferring dexterous manipulation from gpu simulation to a remote real-world trifinger.
\newblock In \emph{2022 IEEE/RSJ International Conference on Intelligent Robots and Systems (IROS)}, pages 11802--11809. IEEE, 2022.

\bibitem[Andrychowicz et~al.(2020)Andrychowicz, Baker, Chociej, Jozefowicz, McGrew, Pachocki, Petron, Plappert, Powell, Ray, et~al.]{andrychowicz2020learning}
OpenAI:~Marcin Andrychowicz, Bowen Baker, Maciek Chociej, Rafal Jozefowicz, Bob McGrew, Jakub Pachocki, Arthur Petron, Matthias Plappert, Glenn Powell, Alex Ray, et~al.
\newblock Learning dexterous in-hand manipulation.
\newblock \emph{The International Journal of Robotics Research}, 39\penalty0 (1):\penalty0 3--20, 2020.

\bibitem[Argall et~al.(2009)Argall, Chernova, Veloso, and Browning]{argall2009survey}
Brenna~D Argall, Sonia Chernova, Manuela Veloso, and Brett Browning.
\newblock A survey of robot learning from demonstration.
\newblock \emph{Robotics and autonomous systems}, 57\penalty0 (5):\penalty0 469--483, 2009.

\bibitem[Bacon et~al.(2017)Bacon, Harb, and Precup]{bacon2017option}
Pierre-Luc Bacon, Jean Harb, and Doina Precup.
\newblock The option-critic architecture.
\newblock In \emph{Proceedings of the AAAI Conference on Artificial Intelligence}, volume~31, 2017.

\bibitem[Bahl et~al.(2020)Bahl, Mukadam, Gupta, and Pathak]{bahl2020neural}
Shikhar Bahl, Mustafa Mukadam, Abhinav Gupta, and Deepak Pathak.
\newblock Neural dynamic policies for end-to-end sensorimotor learning.
\newblock \emph{Advances in Neural Information Processing Systems}, 33:\penalty0 5058--5069, 2020.

\bibitem[Bahl et~al.(2023)Bahl, Mendonca, Chen, Jain, and Pathak]{bahl2023affordances}
Shikhar Bahl, Russell Mendonca, Lili Chen, Unnat Jain, and Deepak Pathak.
\newblock Affordances from human videos as a versatile representation for robotics.
\newblock 2023.

\bibitem[Barron and Malik(2013)]{barron2013intrinsic}
Jonathan~T Barron and Jitendra Malik.
\newblock Intrinsic scene properties from a single rgb-d image.
\newblock In \emph{Proceedings of the IEEE Conference on Computer Vision and Pattern Recognition}, pages 17--24, 2013.

\bibitem[Beeson and Ames(2015)]{Beeson-humanoids-15}
Patrick Beeson and Barrett Ames.
\newblock \{TRAC-IK\}: An open-source library for improved solving of generic inverse kinematics.
\newblock 11 2015.

\bibitem[Bemporad and Morari(1999)]{bemporad1999control}
Alberto Bemporad and Manfred Morari.
\newblock Control of systems integrating logic, dynamics, and constraints.
\newblock \emph{Automatica}, 35\penalty0 (3):\penalty0 407--427, 1999.

\bibitem[Bhardwaj et~al.(2017)Bhardwaj, Choudhury, and Scherer]{bhardwaj2017learning}
Mohak Bhardwaj, Sanjiban Choudhury, and Sebastian Scherer.
\newblock Learning heuristic search via imitation.
\newblock In \emph{Conference on Robot Learning}, pages 271--280. PMLR, 2017.

\bibitem[Billard et~al.(2008)Billard, Calinon, Dillmann, and Schaal]{Billard2008RobotPB}
Aude Billard, Sylvain Calinon, R{\"u}diger Dillmann, and Stefan Schaal.
\newblock Robot programming by demonstration.
\newblock In \emph{Springer Handbook of Robotics}, 2008.

\bibitem[Bochkovskii et~al.(2024)Bochkovskii, Delaunoy, Germain, Santos, Zhou, Richter, and Koltun]{Bochkovskii2024}
Aleksei Bochkovskii, Ama\"{e}l Delaunoy, Hugo Germain, Marcel Santos, Yichao Zhou, Stephan~R. Richter, and Vladlen Koltun.
\newblock Depth pro: Sharp monocular metric depth in less than a second.
\newblock \emph{arXiv}, 2024.
\newblock URL \url{https://arxiv.org/abs/2410.02073}.

\bibitem[Bohlin and Kavraki(2000)]{lazyprm}
R.~Bohlin and L.E. Kavraki.
\newblock Path planning using lazy prm.
\newblock In \emph{Proceedings 2000 ICRA. Millennium Conference. IEEE International Conference on Robotics and Automation. Symposia Proceedings (Cat. No.00CH37065)}, volume~1, pages 521--528 vol.1, 2000.
\newblock \doi{10.1109/ROBOT.2000.844107}.

\bibitem[Branicky et~al.(1998)Branicky, Borkar, and Mitter]{branicky1998unified}
Michael~S Branicky, Vivek~S Borkar, and Sanjoy~K Mitter.
\newblock A unified framework for hybrid control: Model and optimal control theory.
\newblock \emph{IEEE transactions on automatic control}, 43\penalty0 (1):\penalty0 31--45, 1998.

\bibitem[Brohan et~al.(2022{\natexlab{a}})Brohan, Brown, Carbajal, Chebotar, Dabis, Finn, Gopalakrishnan, Hausman, Herzog, Hsu, Ibarz, Ichter, Irpan, Jackson, Jesmonth, Joshi, Julian, Kalashnikov, Kuang, Leal, Lee, Levine, Lu, Malla, Manjunath, Mordatch, Nachum, Parada, Peralta, Perez, Pertsch, Quiambao, Rao, Ryoo, Salazar, Sanketi, Sayed, Singh, Sontakke, Stone, Tan, Tran, Vanhoucke, Vega, Vuong, Xia, Xiao, Xu, Xu, Yu, and Zitkovich]{rt12022arxiv}
Anthony Brohan, Noah Brown, Justice Carbajal, Yevgen Chebotar, Joseph Dabis, Chelsea Finn, Keerthana Gopalakrishnan, Karol Hausman, Alex Herzog, Jasmine Hsu, Julian Ibarz, Brian Ichter, Alex Irpan, Tomas Jackson, Sally Jesmonth, Nikhil Joshi, Ryan Julian, Dmitry Kalashnikov, Yuheng Kuang, Isabel Leal, Kuang-Huei Lee, Sergey Levine, Yao Lu, Utsav Malla, Deeksha Manjunath, Igor Mordatch, Ofir Nachum, Carolina Parada, Jodilyn Peralta, Emily Perez, Karl Pertsch, Jornell Quiambao, Kanishka Rao, Michael Ryoo, Grecia Salazar, Pannag Sanketi, Kevin Sayed, Jaspiar Singh, Sumedh Sontakke, Austin Stone, Clayton Tan, Huong Tran, Vincent Vanhoucke, Steve Vega, Quan Vuong, Fei Xia, Ted Xiao, Peng Xu, Sichun Xu, Tianhe Yu, and Brianna Zitkovich.
\newblock Rt-1: Robotics transformer for real-world control at scale.
\newblock In \emph{arXiv preprint arXiv:2204.01691}, 2022{\natexlab{a}}.

\bibitem[Brohan et~al.(2022{\natexlab{b}})Brohan, Brown, Carbajal, Chebotar, Dabis, Finn, Gopalakrishnan, Hausman, Herzog, Hsu, et~al.]{brohan2022rt}
Anthony Brohan, Noah Brown, Justice Carbajal, Yevgen Chebotar, Joseph Dabis, Chelsea Finn, Keerthana Gopalakrishnan, Karol Hausman, Alex Herzog, Jasmine Hsu, et~al.
\newblock Rt-1: Robotics transformer for real-world control at scale.
\newblock \emph{arXiv preprint arXiv:2212.06817}, 2022{\natexlab{b}}.

\bibitem[Brohan et~al.(2023)Brohan, Brown, Carbajal, Chebotar, Chen, Choromanski, Ding, Driess, Dubey, Finn, et~al.]{brohan2023rt}
Anthony Brohan, Noah Brown, Justice Carbajal, Yevgen Chebotar, Xi~Chen, Krzysztof Choromanski, Tianli Ding, Danny Driess, Avinava Dubey, Chelsea Finn, et~al.
\newblock Rt-2: Vision-language-action models transfer web knowledge to robotic control.
\newblock \emph{arXiv preprint arXiv:2307.15818}, 2023.

\bibitem[Brown et~al.(2020)Brown, Mann, Ryder, Subbiah, Kaplan, Dhariwal, Neelakantan, Shyam, Sastry, Askell, et~al.]{brown2020language}
Tom Brown, Benjamin Mann, Nick Ryder, Melanie Subbiah, Jared~D Kaplan, Prafulla Dhariwal, Arvind Neelakantan, Pranav Shyam, Girish Sastry, Amanda Askell, et~al.
\newblock Language models are few-shot learners.
\newblock \emph{Advances in neural information processing systems}, 33:\penalty0 1877--1901, 2020.

\bibitem[Bruce et~al.(2024)Bruce, Dennis, Edwards, Parker-Holder, Shi, Hughes, Lai, Mavalankar, Steigerwald, Apps, et~al.]{bruce2024genie}
Jake Bruce, Michael~D Dennis, Ashley Edwards, Jack Parker-Holder, Yuge Shi, Edward Hughes, Matthew Lai, Aditi Mavalankar, Richie Steigerwald, Chris Apps, et~al.
\newblock Genie: Generative interactive environments.
\newblock In \emph{Forty-first International Conference on Machine Learning}, 2024.

\bibitem[Calinon et~al.(2010)Calinon, D'halluin, Sauser, Caldwell, and Billard]{Calinon2010LearningAR}
Sylvain Calinon, Florent D'halluin, Eric~L. Sauser, Darwin~G. Caldwell, and Aude Billard.
\newblock Learning and reproduction of gestures by imitation.
\newblock \emph{IEEE Robotics and Automation Magazine}, 17:\penalty0 44--54, 2010.

\bibitem[Cambon et~al.(2009)Cambon, Alami, and Gravot]{cambon2009hybrid}
Stephane Cambon, Rachid Alami, and Fabien Gravot.
\newblock A hybrid approach to intricate motion, manipulation and task planning.
\newblock \emph{The International Journal of Robotics Research}, 28\penalty0 (1):\penalty0 104--126, 2009.

\bibitem[Carvalho et~al.(2023)Carvalho, Le, Baierl, Koert, and Peters]{carvalho2023motion}
Joao Carvalho, An~T Le, Mark Baierl, Dorothea Koert, and Jan Peters.
\newblock Motion planning diffusion: Learning and planning of robot motions with diffusion models.
\newblock In \emph{2023 IEEE/RSJ International Conference on Intelligent Robots and Systems (IROS)}, pages 1916--1923. IEEE, 2023.

\bibitem[Center(1984)]{center1984shakey}
Artificial~Intellgence Center.
\newblock Shakey the robot.
\newblock 1984.

\bibitem[Chamzas et~al.(2021)Chamzas, Quintero-Pena, Kingston, Orthey, Rakita, Gleicher, Toussaint, and Kavraki]{chamzas2021motionbenchmaker}
Constantinos Chamzas, Carlos Quintero-Pena, Zachary Kingston, Andreas Orthey, Daniel Rakita, Michael Gleicher, Marc Toussaint, and Lydia~E Kavraki.
\newblock Motionbenchmaker: A tool to generate and benchmark motion planning datasets.
\newblock \emph{IEEE Robotics and Automation Letters}, 7\penalty0 (2):\penalty0 882--889, 2021.

\bibitem[Chang et~al.(2014)Chang, Savva, and Manning]{chang2014learning}
Angel Chang, Manolis Savva, and Christopher~D Manning.
\newblock Learning spatial knowledge for text to 3d scene generation.
\newblock In \emph{Proceedings of the 2014 conference on empirical methods in natural language processing (EMNLP)}, pages 2028--2038, 2014.

\bibitem[Chang et~al.(2015{\natexlab{a}})Chang, Monroe, Savva, Potts, and Manning]{chang2015text}
Angel Chang, Will Monroe, Manolis Savva, Christopher Potts, and Christopher~D Manning.
\newblock Text to 3d scene generation with rich lexical grounding.
\newblock \emph{arXiv preprint arXiv:1505.06289}, 2015{\natexlab{a}}.

\bibitem[Chang et~al.(2015{\natexlab{b}})Chang, Funkhouser, Guibas, Hanrahan, Huang, Li, Savarese, Savva, Song, Su, et~al.]{chang2015shapenet}
Angel~X Chang, Thomas Funkhouser, Leonidas Guibas, Pat Hanrahan, Qixing Huang, Zimo Li, Silvio Savarese, Manolis Savva, Shuran Song, Hao Su, et~al.
\newblock Shapenet: An information-rich 3d model repository.
\newblock \emph{arXiv preprint arXiv:1512.03012}, 2015{\natexlab{b}}.

\bibitem[Chen et~al.(2022)Chen, Tippur, Wu, Kumar, Adelson, and Agrawal]{chen2022visual}
Tao Chen, Megha Tippur, Siyang Wu, Vikash Kumar, Edward Adelson, and Pulkit Agrawal.
\newblock Visual dexterity: In-hand dexterous manipulation from depth.
\newblock \emph{arXiv preprint arXiv:2211.11744}, 2022.

\bibitem[Chen et~al.(2023)Chen, Tippur, Wu, Kumar, Adelson, and Agrawal]{chen2023visual}
Tao Chen, Megha Tippur, Siyang Wu, Vikash Kumar, Edward Adelson, and Pulkit Agrawal.
\newblock Visual dexterity: In-hand reorientation of novel and complex object shapes.
\newblock \emph{Science Robotics}, 8\penalty0 (84):\penalty0 eadc9244, 2023.

\bibitem[Cheng and Xu(2022)]{cheng2022guided}
Shuo Cheng and Danfei Xu.
\newblock Guided skill learning and abstraction for long-horizon manipulation.
\newblock \emph{arXiv preprint arXiv:2210.12631}, 2022.

\bibitem[Cheng et~al.(2023{\natexlab{a}})Cheng, Shi, Agarwal, and Pathak]{cheng2023extreme}
Xuxin Cheng, Kexin Shi, Ananye Agarwal, and Deepak Pathak.
\newblock Extreme parkour with legged robots.
\newblock \emph{arXiv preprint arXiv:2309.14341}, 2023{\natexlab{a}}.

\bibitem[Cheng et~al.(2023{\natexlab{b}})Cheng, Shi, Agarwal, and Pathak]{cheng2023parkour}
Xuxin Cheng, Kexin Shi, Ananye Agarwal, and Deepak Pathak.
\newblock Extreme parkour with legged robots.
\newblock \emph{arXiv preprint arXiv:2309.14341}, 2023{\natexlab{b}}.

\bibitem[Chi et~al.(2023)Chi, Feng, Du, Xu, Cousineau, Burchfiel, and Song]{chi2023diffusion}
Cheng Chi, Siyuan Feng, Yilun Du, Zhenjia Xu, Eric Cousineau, Benjamin Burchfiel, and Shuran Song.
\newblock Diffusion policy: Visuomotor policy learning via action diffusion.
\newblock \emph{arXiv preprint arXiv:2303.04137}, 2023.

\bibitem[Chi et~al.(2024)Chi, Xu, Pan, Cousineau, Burchfiel, Feng, Tedrake, and Song]{chi2024universal}
Cheng Chi, Zhenjia Xu, Chuer Pan, Eric Cousineau, Benjamin Burchfiel, Siyuan Feng, Russ Tedrake, and Shuran Song.
\newblock Universal manipulation interface: In-the-wild robot teaching without in-the-wild robots.
\newblock \emph{arXiv preprint arXiv:2402.10329}, 2024.

\bibitem[Chitnis et~al.(2020)Chitnis, Tulsiani, Gupta, and Gupta]{chitnis2020efficient}
Rohan Chitnis, Shubham Tulsiani, Saurabh Gupta, and Abhinav Gupta.
\newblock Efficient bimanual manipulation using learned task schemas.
\newblock In \emph{2020 IEEE International Conference on Robotics and Automation (ICRA)}, pages 1149--1155. IEEE, 2020.

\bibitem[Chowdhery et~al.(2022)Chowdhery, Narang, Devlin, Bosma, Mishra, Roberts, Barham, Chung, Sutton, Gehrmann, et~al.]{chowdhery2022palm}
Aakanksha Chowdhery, Sharan Narang, Jacob Devlin, Maarten Bosma, Gaurav Mishra, Adam Roberts, Paul Barham, Hyung~Won Chung, Charles Sutton, Sebastian Gehrmann, et~al.
\newblock Palm: Scaling language modeling with pathways.
\newblock \emph{arXiv preprint arXiv:2204.02311}, 2022.

\bibitem[Christen et~al.(2023)Christen, Yang, Pérez-D'Arpino, Hilliges, Fox, and Chao]{christen2023handoversim2real}
Sammy Christen, Wei Yang, Claudia Pérez-D'Arpino, Otmar Hilliges, Dieter Fox, and Yu-Wei Chao.
\newblock Learning human-to-robot handovers from point clouds.
\newblock In \emph{Proceedings of the IEEE/CVF Conference on Computer Vision and Pattern Recognition (CVPR)}, 2023.

\bibitem[Cohen et~al.(2010)Cohen, Chitta, and Likhachev]{Cohen2010}
Benjamin Cohen, Sachin Chitta, and Maxim Likhachev.
\newblock Search-based planning for manipulation with motion primitives.
\newblock In \emph{International Conference on Robotics and Automation}, 2010.

\bibitem[Colas et~al.(2020)Colas, Karch, Lair, Dussoux, Moulin-Frier, Dominey, and Oudeyer]{colas2020language}
C{\'e}dric Colas, Tristan Karch, Nicolas Lair, Jean-Michel Dussoux, Cl{\'e}ment Moulin-Frier, Peter Dominey, and Pierre-Yves Oudeyer.
\newblock Language as a cognitive tool to imagine goals in curiosity driven exploration.
\newblock \emph{Advances in Neural Information Processing Systems}, 33:\penalty0 3761--3774, 2020.

\bibitem[Collaboration et~al.(2023)Collaboration, Padalkar, Pooley, Jain, Bewley, Herzog, Irpan, Khazatsky, Rai, Singh, et~al.]{collaboration2023open}
Open-X~Embodiment Collaboration, A~Padalkar, A~Pooley, A~Jain, A~Bewley, A~Herzog, A~Irpan, A~Khazatsky, A~Rai, A~Singh, et~al.
\newblock Open x-embodiment: Robotic learning datasets and rt-x models.
\newblock \emph{arXiv preprint arXiv:2310.08864}, 2023.

\bibitem[Cui et~al.(2022)Cui, Wang, Muhammad, Pinto, et~al.]{cui2022play}
Zichen~Jeff Cui, Yibin Wang, Nur Muhammad, Lerrel Pinto, et~al.
\newblock From play to policy: Conditional behavior generation from uncurated robot data.
\newblock \emph{arXiv preprint arXiv:2210.10047}, 2022.

\bibitem[Curtis et~al.(2022)Curtis, Fang, Kaelbling, Lozano-P{\'{e}}rez, and Garrett]{curtis2022long}
Aidan Curtis, Xiaolin Fang, Leslie~Pack Kaelbling, Tom{\'{a}}s Lozano-P{\'{e}}rez, and Caelan~Reed Garrett.
\newblock {Long-horizon manipulation of unknown objects via task and motion planning with estimated affordances}.
\newblock In \emph{2022 International Conference on Robotics and Automation (ICRA)}, pages 1940--1946. IEEE, 2022.

\bibitem[Dalal et~al.(2021)Dalal, Pathak, and Salakhutdinov]{dalal2021accelerating}
Murtaza Dalal, Deepak Pathak, and Russ~R Salakhutdinov.
\newblock Accelerating robotic reinforcement learning via parameterized action primitives.
\newblock \emph{Advances in Neural Information Processing Systems}, 34:\penalty0 21847--21859, 2021.

\bibitem[Dalal et~al.(2023)Dalal, Mandlekar, Garrett, Handa, Salakhutdinov, and Fox]{dalal2023optimus}
Murtaza Dalal, Ajay Mandlekar, Caelan Garrett, Ankur Handa, Ruslan Salakhutdinov, and Dieter Fox.
\newblock Imitating task and motion planning with visuomotor transformers.
\newblock 2023.

\bibitem[Dalal et~al.(2024{\natexlab{a}})Dalal, Chiruvolu, Chaplot, and Salakhutdinov]{dalal2024psl}
Murtaza Dalal, Tarun Chiruvolu, Devendra Chaplot, and Ruslan Salakhutdinov.
\newblock Plan-seq-learn: Language model guided rl for solving long horizon robotics tasks.
\newblock In \emph{International Conference on Learning Representations (ICLR)}, 2024{\natexlab{a}}.

\bibitem[Dalal et~al.(2024{\natexlab{b}})Dalal, Yang, Mendonca, Khaky, Salakhutdinov, and Pathak]{dalal2024neuralmp}
Murtaza Dalal, Jiahui Yang, Russell Mendonca, Youssef Khaky, Ruslan Salakhutdinov, and Deepak Pathak.
\newblock Data driven neural motion planning.
\newblock In \emph{In Submission}, 2024{\natexlab{b}}.

\bibitem[Dalal et~al.(2025)Dalal, Liu, Talbott, Chen, Pathak, Zhang, and Salakhutdinov]{dalal2024manipgen}
Murtaza Dalal, Min Liu, Walter Talbott, Chen Chen, Deepak Pathak, Jian Zhang, and Ruslan Salakhutdinov.
\newblock Local policies enable zero-shot long-horizon manipulation.
\newblock 2025.

\bibitem[Daniel et~al.(2016)Daniel, Neumann, Kroemer, Peters, et~al.]{daniel2016hierarchical}
Christian Daniel, Gerhard Neumann, Oliver Kroemer, Jan Peters, et~al.
\newblock Hierarchical relative entropy policy search.
\newblock \emph{Journal of Machine Learning Research}, 17:\penalty0 1--50, 2016.

\bibitem[Dao et~al.(2022)Dao, Fu, Ermon, Rudra, and R{\'e}]{dao2022flashattention}
Tri Dao, Daniel~Y Fu, Stefano Ermon, Atri Rudra, and Christopher R{\'e}.
\newblock Flashattention: Fast and memory-efficient exact attention with io-awareness.
\newblock \emph{arXiv preprint arXiv:2205.14135}, 2022.

\bibitem[Dasari and Gupta(2021)]{dasari2021transformers}
Sudeep Dasari and Abhinav Gupta.
\newblock Transformers for one-shot visual imitation.
\newblock In \emph{Conference on Robot Learning}, pages 2071--2084. PMLR, 2021.

\bibitem[Dass et~al.(2022)Dass, Pertsch, Zhang, Lee, Lim, and Nikolaidis]{dass2022pato}
Shivin Dass, Karl Pertsch, Hejia Zhang, Youngwoon Lee, Joseph~J Lim, and Stefanos Nikolaidis.
\newblock Pato: Policy assisted teleoperation for scalable robot data collection.
\newblock \emph{arXiv preprint arXiv:2212.04708}, 2022.

\bibitem[Deitke et~al.(2022)Deitke, VanderBilt, Herrasti, Weihs, Ehsani, Salvador, Han, Kolve, Kembhavi, and Mottaghi]{deitke2022️}
Matt Deitke, Eli VanderBilt, Alvaro Herrasti, Luca Weihs, Kiana Ehsani, Jordi Salvador, Winson Han, Eric Kolve, Aniruddha Kembhavi, and Roozbeh Mottaghi.
\newblock Procthor: Large-scale embodied ai using procedural generation.
\newblock \emph{Advances in Neural Information Processing Systems}, 35:\penalty0 5982--5994, 2022.

\bibitem[Deitke et~al.(2023)Deitke, Schwenk, Salvador, Weihs, Michel, VanderBilt, Schmidt, Ehsani, Kembhavi, and Farhadi]{deitke2023objaverse}
Matt Deitke, Dustin Schwenk, Jordi Salvador, Luca Weihs, Oscar Michel, Eli VanderBilt, Ludwig Schmidt, Kiana Ehsani, Aniruddha Kembhavi, and Ali Farhadi.
\newblock Objaverse: A universe of annotated 3d objects.
\newblock In \emph{Proceedings of the IEEE/CVF Conference on Computer Vision and Pattern Recognition}, pages 13142--13153, 2023.

\bibitem[Dosovitskiy et~al.(2020)Dosovitskiy, Beyer, Kolesnikov, Weissenborn, Zhai, Unterthiner, Dehghani, Minderer, Heigold, Gelly, et~al.]{dosovitskiy2020image}
Alexey Dosovitskiy, Lucas Beyer, Alexander Kolesnikov, Dirk Weissenborn, Xiaohua Zhai, Thomas Unterthiner, Mostafa Dehghani, Matthias Minderer, Georg Heigold, Sylvain Gelly, et~al.
\newblock An image is worth 16x16 words: Transformers for image recognition at scale.
\newblock \emph{arXiv preprint arXiv:2010.11929}, 2020.

\bibitem[Dragan et~al.(2011)Dragan, Ratliff, and Srinivasa]{dragan2011manipulation}
Anca~D Dragan, Nathan~D Ratliff, and Siddhartha~S Srinivasa.
\newblock Manipulation planning with goal sets using constrained trajectory optimization.
\newblock In \emph{2011 IEEE International Conference on Robotics and Automation}, pages 4582--4588. IEEE, 2011.

\bibitem[Driess et~al.(2022)Driess, Ha, Toussaint, and Tedrake]{driess2022learning}
Danny Driess, Jung-Su Ha, Marc Toussaint, and Russ Tedrake.
\newblock Learning models as functionals of signed-distance fields for manipulation planning.
\newblock In \emph{Conference on Robot Learning}, pages 245--255. PMLR, 2022.

\bibitem[Du et~al.(2023)Du, Watkins, Wang, Colas, Darrell, Abbeel, Gupta, and Andreas]{du2023guiding}
Yuqing Du, Olivia Watkins, Zihan Wang, C{\'e}dric Colas, Trevor Darrell, Pieter Abbeel, Abhishek Gupta, and Jacob Andreas.
\newblock Guiding pretraining in reinforcement learning with large language models.
\newblock \emph{arXiv preprint arXiv:2302.06692}, 2023.

\bibitem[Ebert et~al.(2021)Ebert, Yang, Schmeckpeper, Bucher, Georgakis, Daniilidis, Finn, and Levine]{ebert2021bridge}
Frederik Ebert, Yanlai Yang, Karl Schmeckpeper, Bernadette Bucher, Georgios Georgakis, Kostas Daniilidis, Chelsea Finn, and Sergey Levine.
\newblock Bridge data: Boosting generalization of robotic skills with cross-domain datasets.
\newblock \emph{arXiv preprint arXiv:2109.13396}, 2021.

\bibitem[Eppner et~al.(2021)Eppner, Mousavian, and Fox]{eppner2021acronym}
Clemens Eppner, Arsalan Mousavian, and Dieter Fox.
\newblock Acronym: A large-scale grasp dataset based on simulation.
\newblock In \emph{2021 IEEE International Conference on Robotics and Automation (ICRA)}, pages 6222--6227. IEEE, 2021.

\bibitem[Eysenbach et~al.(2018)Eysenbach, Gupta, Ibarz, and Levine]{eysenbach2018diversity}
Benjamin Eysenbach, Abhishek Gupta, Julian Ibarz, and Sergey Levine.
\newblock Diversity is all you need: Learning skills without a reward function.
\newblock \emph{arXiv preprint arXiv:1802.06070}, 2018.

\bibitem[Fan et~al.(2019)Fan, Su, Zhang, and Yu]{fan2019hybrid}
Zhou Fan, Rui Su, Weinan Zhang, and Yong Yu.
\newblock Hybrid actor-critic reinforcement learning in parameterized action space.
\newblock \emph{arXiv preprint arXiv:1903.01344}, 2019.

\bibitem[Fikes and Nilsson(1971)]{fikes1971strips}
Richard~E Fikes and Nils~J Nilsson.
\newblock Strips: A new approach to the application of theorem proving to problem solving.
\newblock \emph{Artificial intelligence}, 2\penalty0 (3-4):\penalty0 189--208, 1971.

\bibitem[Finn et~al.(2016)Finn, Tan, Duan, Darrell, Levine, and Abbeel]{finn2016deep}
Chelsea Finn, Xin~Yu Tan, Yan Duan, Trevor Darrell, Sergey Levine, and Pieter Abbeel.
\newblock Deep spatial autoencoders for visuomotor learning.
\newblock In \emph{2016 IEEE International Conference on Robotics and Automation (ICRA)}, pages 512--519. IEEE, 2016.

\bibitem[Finn et~al.(2017)Finn, Yu, Zhang, Abbeel, and Levine]{Finn2017OneShotVI}
Chelsea Finn, Tianhe Yu, Tianhao Zhang, Pieter Abbeel, and Sergey Levine.
\newblock One-shot visual imitation learning via meta-learning.
\newblock In \emph{Conference in Robot Learning}, volume abs/1709.04905, 2017.

\bibitem[Fishman et~al.(2022)Fishman, Murali, Eppner, Peele, Boots, and Fox]{fishman2022motion}
Adam Fishman, Adithyavairan Murali, Clemens Eppner, Bryan Peele, Byron Boots, and Dieter Fox.
\newblock Motion policy networks.
\newblock \emph{arXiv preprint arXiv:2210.12209}, 2022.

\bibitem[Fishman et~al.(2023)Fishman, Murali, Eppner, Peele, Boots, and Fox]{fishman2023motion}
Adam Fishman, Adithyavairavan Murali, Clemens Eppner, Bryan Peele, Byron Boots, and Dieter Fox.
\newblock Motion policy networks.
\newblock In \emph{Conference on Robot Learning}, pages 967--977. PMLR, 2023.

\bibitem[Frans et~al.(2017)Frans, Ho, Chen, Abbeel, and Schulman]{frans2017meta}
Kevin Frans, Jonathan Ho, Xi~Chen, Pieter Abbeel, and John Schulman.
\newblock Meta learning shared hierarchies.
\newblock \emph{arXiv preprint arXiv:1710.09767}, 2017.

\bibitem[Fu et~al.(2019)Fu, Tang, Hao, Lei, Chen, and Fan]{fu2019deep}
Haotian Fu, Hongyao Tang, Jianye Hao, Zihan Lei, Yingfeng Chen, and Changjie Fan.
\newblock Deep multi-agent reinforcement learning with discrete-continuous hybrid action spaces.
\newblock \emph{arXiv preprint arXiv:1903.04959}, 2019.

\bibitem[Fu et~al.(2020)Fu, Kumar, Nachum, Tucker, and Levine]{fu2020d4rl}
Justin Fu, Aviral Kumar, Ofir Nachum, George Tucker, and Sergey Levine.
\newblock D4rl: Datasets for deep data-driven reinforcement learning.
\newblock \emph{arXiv preprint arXiv:2004.07219}, 2020.

\bibitem[Fu et~al.(2021)Fu, Kumar, Nachum, Tucker, and Levine]{fu2021d4rl}
Justin Fu, Aviral Kumar, Ofir Nachum, George Tucker, and Sergey Levine.
\newblock D4rl: Datasets for deep data-driven reinforcement learning, 2021.

\bibitem[Fu et~al.(2024)Fu, Zhao, and Finn]{fu2024mobile}
Zipeng Fu, Tony~Z. Zhao, and Chelsea Finn.
\newblock Mobile aloha: Learning bimanual mobile manipulation with low-cost whole-body teleoperation.
\newblock \emph{arXiv preprint arXiv: Arxiv-2401.02117}, 2024.

\bibitem[Gammell et~al.(2015)Gammell, Srinivasa, and Barfoot]{bit*}
Jonathan~D. Gammell, Siddhartha~S. Srinivasa, and Timothy~D. Barfoot.
\newblock Batch informed trees (bit*): Sampling-based optimal planning via the heuristically guided search of implicit random geometric graphs.
\newblock In \emph{2015 IEEE International Conference on Robotics and Automation (ICRA)}, pages 3067--3074, 2015.
\newblock \doi{10.1109/ICRA.2015.7139620}.

\bibitem[Garrett et~al.(2020{\natexlab{a}})Garrett, Lozano-P{\'e}rez, and Kaelbling]{garrett2020pddlstream}
Caelan~Reed Garrett, Tom{\'a}s Lozano-P{\'e}rez, and Leslie~Pack Kaelbling.
\newblock Pddlstream: Integrating symbolic planners and blackbox samplers via optimistic adaptive planning.
\newblock In \emph{Proceedings of the International Conference on Automated Planning and Scheduling}, volume~30, pages 440--448, 2020{\natexlab{a}}.

\bibitem[Garrett et~al.(2020{\natexlab{b}})Garrett, Paxton, Lozano-P{\'e}rez, Kaelbling, and Fox]{garrett2020online}
Caelan~Reed Garrett, Chris Paxton, Tom{\'a}s Lozano-P{\'e}rez, Leslie~Pack Kaelbling, and Dieter Fox.
\newblock Online replanning in belief space for partially observable task and motion problems.
\newblock In \emph{2020 IEEE International Conference on Robotics and Automation (ICRA)}, pages 5678--5684. IEEE, 2020{\natexlab{b}}.

\bibitem[Garrett et~al.(2021)Garrett, Chitnis, Holladay, Kim, Silver, Kaelbling, and Lozano-P´erez]{Garrett2021}
Caelan~Reed Garrett, Rohan Chitnis, Rachel Holladay, Beomjoon Kim, Tom Silver, Leslie~Pack Kaelbling, and Tom´as Lozano-P´erez.
\newblock {Integrated Task and Motion Planning}.
\newblock \emph{Annual review of control, robotics, and autonomous systems}, 4, 2021.

\bibitem[Guhur et~al.(2022)Guhur, Chen, Garcia, Tapaswi, Laptev, and Schmid]{guhur2022instruction}
Pierre-Louis Guhur, Shizhe Chen, Ricardo Garcia, Makarand Tapaswi, Ivan Laptev, and Cordelia Schmid.
\newblock Instruction-driven history-aware policies for robotic manipulations.
\newblock 2022.

\bibitem[Guo et~al.(2025)Guo, Yang, Zhang, Song, Zhang, Xu, Zhu, Ma, Wang, Bi, et~al.]{guo2025deepseek}
Daya Guo, Dejian Yang, Haowei Zhang, Junxiao Song, Ruoyu Zhang, Runxin Xu, Qihao Zhu, Shirong Ma, Peiyi Wang, Xiao Bi, et~al.
\newblock Deepseek-r1: Incentivizing reasoning capability in llms via reinforcement learning.
\newblock \emph{arXiv preprint arXiv:2501.12948}, 2025.

\bibitem[Gupta et~al.(2019)Gupta, Kumar, Lynch, Levine, and Hausman]{gupta2019relay}
Abhishek Gupta, Vikash Kumar, Corey Lynch, Sergey Levine, and Karol Hausman.
\newblock Relay policy learning: Solving long-horizon tasks via imitation and reinforcement learning, 2019.

\bibitem[Ha et~al.(2023)Ha, Florence, and Song]{ha2023scalingup}
Huy Ha, Pete Florence, and Shuran Song.
\newblock Scaling up and distilling down: Language-guided robot skill acquisition.
\newblock In \emph{Proceedings of the 2023 Conference on Robot Learning}, 2023.

\bibitem[Haarnoja et~al.(2018)Haarnoja, Zhou, Abbeel, and Levine]{haarnoja2018soft}
Tuomas Haarnoja, Aurick Zhou, Pieter Abbeel, and Sergey Levine.
\newblock Soft actor-critic: Off-policy maximum entropy deep reinforcement learning with a stochastic actor, 2018.

\bibitem[Haarnoja et~al.(2024)Haarnoja, Moran, Lever, Huang, Tirumala, Humplik, Wulfmeier, Tunyasuvunakool, Siegel, Hafner, et~al.]{haarnoja2024learning}
Tuomas Haarnoja, Ben Moran, Guy Lever, Sandy~H Huang, Dhruva Tirumala, Jan Humplik, Markus Wulfmeier, Saran Tunyasuvunakool, Noah~Y Siegel, Roland Hafner, et~al.
\newblock Learning agile soccer skills for a bipedal robot with deep reinforcement learning.
\newblock \emph{Science Robotics}, 9\penalty0 (89):\penalty0 eadi8022, 2024.

\bibitem[Hafner et~al.(2019)Hafner, Lillicrap, Ba, and Norouzi]{hafner2019dream}
Danijar Hafner, Timothy Lillicrap, Jimmy Ba, and Mohammad Norouzi.
\newblock Dream to control: Learning behaviors by latent imagination.
\newblock \emph{arXiv preprint arXiv:1912.01603}, 2019.

\bibitem[Hafner et~al.(2020{\natexlab{a}})Hafner, Lillicrap, Ba, and Norouzi]{hafner2020dream}
Danijar Hafner, Timothy Lillicrap, Jimmy Ba, and Mohammad Norouzi.
\newblock Dream to control: Learning behaviors by latent imagination, 2020{\natexlab{a}}.

\bibitem[Hafner et~al.(2020{\natexlab{b}})Hafner, Lillicrap, Norouzi, and Ba]{hafner2020mastering}
Danijar Hafner, Timothy Lillicrap, Mohammad Norouzi, and Jimmy Ba.
\newblock Mastering atari with discrete world models.
\newblock \emph{arXiv preprint arXiv:2010.02193}, 2020{\natexlab{b}}.

\bibitem[Handa et~al.(2022)Handa, Allshire, Makoviychuk, Petrenko, Singh, Liu, Makoviichuk, Van~Wyk, Zhurkevich, Sundaralingam, et~al.]{handa2022dextreme}
Ankur Handa, Arthur Allshire, Viktor Makoviychuk, Aleksei Petrenko, Ritvik Singh, Jingzhou Liu, Denys Makoviichuk, Karl Van~Wyk, Alexander Zhurkevich, Balakumar Sundaralingam, et~al.
\newblock Dextreme: Transfer of agile in-hand manipulation from simulation to reality.
\newblock \emph{arXiv preprint arXiv:2210.13702}, 2022.

\bibitem[Hansen et~al.(2022)Hansen, Lin, Su, Wang, Kumar, and Rajeswaran]{hansen2022MoDem}
Nicklas Hansen, Yixin Lin, Hao Su, Xiaolong Wang, Vikash Kumar, and Aravind Rajeswaran.
\newblock Modem: Accelerating visual model-based reinforcement learning with demonstrations.
\newblock \emph{arXiv preprint arXiv:2212.05698}, 2022.

\bibitem[Hart et~al.(1968)Hart, Nilsson, and Raphael]{hart1968formal}
Peter~E Hart, Nils~J Nilsson, and Bertram Raphael.
\newblock A formal basis for the heuristic determination of minimum cost paths.
\newblock \emph{IEEE transactions on Systems Science and Cybernetics}, 4\penalty0 (2):\penalty0 100--107, 1968.

\bibitem[Hauser and Ng-Thow-Hing(2010)]{hauser2010fast}
Kris Hauser and Victor Ng-Thow-Hing.
\newblock Fast smoothing of manipulator trajectories using optimal bounded-acceleration shortcuts.
\newblock pages 2493--2498, 2010.

\bibitem[Hausknecht and Stone(2015)]{hausknecht2015deep}
Matthew Hausknecht and Peter Stone.
\newblock Deep reinforcement learning in parameterized action space.
\newblock \emph{arXiv preprint arXiv:1511.04143}, 2015.

\bibitem[Hausman et~al.(2018)Hausman, Springenberg, Wang, Heess, and Riedmiller]{hausman2018learning}
Karol Hausman, Jost~Tobias Springenberg, Ziyu Wang, Nicolas Heess, and Martin Riedmiller.
\newblock Learning an embedding space for transferable robot skills.
\newblock In \emph{International Conference on Learning Representations}, 2018.

\bibitem[He et~al.(2016)He, Zhang, Ren, and Sun]{he2016deep}
Kaiming He, Xiangyu Zhang, Shaoqing Ren, and Jian Sun.
\newblock Deep residual learning for image recognition.
\newblock In \emph{Proceedings of the IEEE conference on computer vision and pattern recognition}, pages 770--778, 2016.

\bibitem[He et~al.(2024)He, Luo, Xiao, Zhang, Kitani, Liu, and Shi]{he2024learning}
Tairan He, Zhengyi Luo, Wenli Xiao, Chong Zhang, Kris Kitani, Changliu Liu, and Guanya Shi.
\newblock Learning human-to-humanoid real-time whole-body teleoperation.
\newblock \emph{arXiv preprint arXiv: Arxiv-2403.04436}, 2024.

\bibitem[Heo et~al.(2023)Heo, Lee, Lee, and Lim]{heo2023furniturebench}
Minho Heo, Youngwoon Lee, Doohyun Lee, and Joseph~J. Lim.
\newblock Furniturebench: Reproducible real-world benchmark for long-horizon complex manipulation.
\newblock In Kostas~E. Bekris, Kris Hauser, Sylvia~L. Herbert, and Jingjin Yu, editors, \emph{Robotics: Science and Systems XIX, Daegu, Republic of Korea, July 10-14, 2023}, 2023.
\newblock \doi{10.15607/RSS.2023.XIX.041}.
\newblock URL \url{https://doi.org/10.15607/RSS.2023.XIX.041}.

\bibitem[Hersch et~al.(2008)Hersch, Guenter, Calinon, and Billard]{hersch2008dynamical}
Micha Hersch, Florent Guenter, Sylvain Calinon, and Aude Billard.
\newblock Dynamical system modulation for robot learning via kinesthetic demonstrations.
\newblock \emph{IEEE Transactions on Robotics}, 24\penalty0 (6):\penalty0 1463--1467, 2008.

\bibitem[Herzog* et~al.(2023)Herzog*, Rao*, Hausman*, Lu*, Wohlhart*, Yan, Lin, Arenas, Xiao, Kappler, Ho, Rettinghouse, Chebotar, Lee, Gopalakrishnan, Julian, Li, Fu, Wei, Ramesh, Holden, Kleiven, Rendleman, Kirmani, Bingham, Weisz, Xu, Lu, Bennice, Fong, Do, Lam, Brown, Kalakrishnan, Ibarz, Pastor, and Levine]{rlscale2023arxiv}
Alexander Herzog*, Kanishka Rao*, Karol Hausman*, Yao Lu*, Paul Wohlhart*, Mengyuan Yan, Jessica Lin, Montserrat~Gonzalez Arenas, Ted Xiao, Daniel Kappler, Daniel Ho, Jarek Rettinghouse, Yevgen Chebotar, Kuang-Huei Lee, Keerthana Gopalakrishnan, Ryan Julian, Adrian Li, Chuyuan~Kelly Fu, Bob Wei, Sangeetha Ramesh, Khem Holden, Kim Kleiven, David Rendleman, Sean Kirmani, Jeff Bingham, Jon Weisz, Ying Xu, Wenlong Lu, Matthew Bennice, Cody Fong, David Do, Jessica Lam, Noah Brown, Mrinal Kalakrishnan, Julian Ibarz, Peter Pastor, and Sergey Levine.
\newblock Deep rl at scale: Sorting waste in office buildings with a fleet of mobile manipulators.
\newblock In \emph{arXiv preprint arXiv:2305.03270}, 2023.

\bibitem[Hochreiter and Schmidhuber(1997)]{hochreiter1997long}
Sepp Hochreiter and J{\"u}rgen Schmidhuber.
\newblock Long short-term memory.
\newblock \emph{Neural computation}, 9\penalty0 (8):\penalty0 1735--1780, 1997.

\bibitem[Hoeller et~al.(2024)Hoeller, Rudin, Sako, and Hutter]{hoeller2024anymal}
David Hoeller, Nikita Rudin, Dhionis Sako, and Marco Hutter.
\newblock Anymal parkour: Learning agile navigation for quadrupedal robots.
\newblock \emph{Science Robotics}, 9\penalty0 (88):\penalty0 eadi7566, 2024.

\bibitem[Hokayem and Spong(2006)]{hokayem2006bilateral}
Peter~F Hokayem and Mark~W Spong.
\newblock Bilateral teleoperation: An historical survey.
\newblock \emph{Automatica}, 42\penalty0 (12):\penalty0 2035--2057, 2006.

\bibitem[Hoque et~al.(2021{\natexlab{a}})Hoque, Balakrishna, Novoseller, Wilcox, Brown, and Goldberg]{hoque2021thriftydagger}
Ryan Hoque, Ashwin Balakrishna, Ellen Novoseller, Albert Wilcox, Daniel~S Brown, and Ken Goldberg.
\newblock Thriftydagger: Budget-aware novelty and risk gating for interactive imitation learning.
\newblock \emph{arXiv preprint arXiv:2109.08273}, 2021{\natexlab{a}}.

\bibitem[Hoque et~al.(2021{\natexlab{b}})Hoque, Balakrishna, Putterman, Luo, Brown, Seita, Thananjeyan, Novoseller, and Goldberg]{hoque2021lazydagger}
Ryan Hoque, Ashwin Balakrishna, Carl Putterman, Michael Luo, Daniel~S Brown, Daniel Seita, Brijen Thananjeyan, Ellen Novoseller, and Ken Goldberg.
\newblock Lazydagger: Reducing context switching in interactive imitation learning.
\newblock In \emph{2021 IEEE 17th International Conference on Automation Science and Engineering (CASE)}, pages 502--509. IEEE, 2021{\natexlab{b}}.

\bibitem[Hsu et~al.(2022)Hsu, Kim, Rafailov, Wu, and Finn]{hsu2022vision}
Kyle Hsu, Moo~Jin Kim, Rafael Rafailov, Jiajun Wu, and Chelsea Finn.
\newblock Vision-based manipulators need to also see from their hands.
\newblock \emph{arXiv preprint arXiv:2203.12677}, 2022.

\bibitem[Huang et~al.(2024)Huang, Lin, Hu, Wang, and Gao]{huang2024copa}
Haoxu Huang, Fanqi Lin, Yingdong Hu, Shengjie Wang, and Yang Gao.
\newblock Copa: General robotic manipulation through spatial constraints of parts with foundation models.
\newblock \emph{arXiv preprint arXiv:2403.08248}, 2024.

\bibitem[Huang et~al.(2023{\natexlab{a}})Huang, Wang, Li, Jia, Liu, Zhu, Liang, and Zhu]{huang2023diffusion}
Siyuan Huang, Zan Wang, Puhao Li, Baoxiong Jia, Tengyu Liu, Yixin Zhu, Wei Liang, and Song-Chun Zhu.
\newblock Diffusion-based generation, optimization, and planning in 3d scenes.
\newblock In \emph{Proceedings of the IEEE/CVF Conference on Computer Vision and Pattern Recognition}, pages 16750--16761, 2023{\natexlab{a}}.

\bibitem[Huang et~al.(2022{\natexlab{a}})Huang, Abbeel, Pathak, and Mordatch]{huang2022language}
Wenlong Huang, Pieter Abbeel, Deepak Pathak, and Igor Mordatch.
\newblock Language models as zero-shot planners: Extracting actionable knowledge for embodied agents.
\newblock In \emph{International Conference on Machine Learning}, pages 9118--9147. PMLR, 2022{\natexlab{a}}.

\bibitem[Huang et~al.(2022{\natexlab{b}})Huang, Xia, Xiao, Chan, Liang, Florence, Zeng, Tompson, Mordatch, Chebotar, et~al.]{huang2022inner}
Wenlong Huang, Fei Xia, Ted Xiao, Harris Chan, Jacky Liang, Pete Florence, Andy Zeng, Jonathan Tompson, Igor Mordatch, Yevgen Chebotar, et~al.
\newblock Inner monologue: Embodied reasoning through planning with language models.
\newblock \emph{arXiv preprint arXiv:2207.05608}, 2022{\natexlab{b}}.

\bibitem[Huang et~al.(2023{\natexlab{b}})Huang, Wang, Zhang, Li, Wu, and Fei-Fei]{huang2023voxposer}
Wenlong Huang, Chen Wang, Ruohan Zhang, Yunzhu Li, Jiajun Wu, and Li~Fei-Fei.
\newblock Voxposer: Composable 3d value maps for robotic manipulation with language models.
\newblock \emph{arXiv preprint arXiv:2307.05973}, 2023{\natexlab{b}}.

\bibitem[Ichter et~al.(2020)Ichter, Sermanet, and Lynch]{ichter2020broadly}
Brian Ichter, Pierre Sermanet, and Corey Lynch.
\newblock Broadly-exploring, local-policy trees for long-horizon task planning.
\newblock \emph{arXiv preprint arXiv:2010.06491}, 2020.

\bibitem[Ijspeert et~al.(2002{\natexlab{a}})Ijspeert, Nakanishi, and Schaal]{Ijspeert2002MovementIW}
Auke~Jan Ijspeert, Jun Nakanishi, and Stefan Schaal.
\newblock Movement imitation with nonlinear dynamical systems in humanoid robots.
\newblock \emph{Proceedings 2002 IEEE International Conference on Robotics and Automation}, 2:\penalty0 1398--1403 vol.2, 2002{\natexlab{a}}.

\bibitem[Ijspeert et~al.(2002{\natexlab{b}})Ijspeert, Nakanishi, and Schaal]{ijspeert2002learning}
Auke~Jan Ijspeert, Jun Nakanishi, and Stefan Schaal.
\newblock Learning attractor landscapes for learning motor primitives.
\newblock Technical report, 2002{\natexlab{b}}.

\bibitem[James and Davison(2022)]{james2022q}
Stephen James and Andrew~J Davison.
\newblock Q-attention: Enabling efficient learning for vision-based robotic manipulation.
\newblock \emph{IEEE Robotics and Automation Letters}, 7\penalty0 (2):\penalty0 1612--1619, 2022.

\bibitem[James et~al.(2022)James, Wada, Laidlow, and Davison]{james2022coarse}
Stephen James, Kentaro Wada, Tristan Laidlow, and Andrew~J Davison.
\newblock Coarse-to-fine q-attention: Efficient learning for visual robotic manipulation via discretisation.
\newblock In \emph{Proceedings of the IEEE/CVF Conference on Computer Vision and Pattern Recognition}, pages 13739--13748, 2022.

\bibitem[Jang et~al.(2022)Jang, Irpan, Khansari, Kappler, Ebert, Lynch, Levine, and Finn]{jang2022bc}
Eric Jang, Alex Irpan, Mohi Khansari, Daniel Kappler, Frederik Ebert, Corey Lynch, Sergey Levine, and Chelsea Finn.
\newblock Bc-z: Zero-shot task generalization with robotic imitation learning.
\newblock In \emph{Conference on Robot Learning}, pages 991--1002. PMLR, 2022.

\bibitem[Jiang et~al.(2022)Jiang, Gupta, Zhang, Wang, Dou, Chen, Fei-Fei, Anandkumar, Zhu, and Fan]{jiang2022vima}
Yunfan Jiang, Agrim Gupta, Zichen Zhang, Guanzhi Wang, Yongqiang Dou, Yanjun Chen, Li~Fei-Fei, Anima Anandkumar, Yuke Zhu, and Linxi Fan.
\newblock Vima: General robot manipulation with multimodal prompts.
\newblock \emph{arXiv preprint arXiv: Arxiv-2210.03094}, 2022.

\bibitem[Jiang et~al.(2024)Jiang, Wang, Zhang, Wu, and Fei-Fei]{jiang2024transic}
Yunfan Jiang, Chen Wang, Ruohan Zhang, Jiajun Wu, and Li~Fei-Fei.
\newblock Transic: Sim-to-real policy transfer by learning from online correction.
\newblock \emph{arXiv preprint arXiv: Arxiv-2405.10315}, 2024.

\bibitem[Johnson et~al.(2020)Johnson, Li, Liu, Qureshi, and Yip]{johnson2020dynamically}
Jacob~J Johnson, Linjun Li, Fei Liu, Ahmed~H Qureshi, and Michael~C Yip.
\newblock Dynamically constrained motion planning networks for non-holonomic robots.
\newblock In \emph{2020 IEEE/RSJ International Conference on Intelligent Robots and Systems (IROS)}, pages 6937--6943. IEEE, 2020.

\bibitem[Kaelbling and Lozano-P{\'e}rez(2011)]{kaelbling2011hierarchical}
Leslie~Pack Kaelbling and Tom{\'a}s Lozano-P{\'e}rez.
\newblock Hierarchical task and motion planning in the now.
\newblock In \emph{2011 IEEE International Conference on Robotics and Automation}, pages 1470--1477. IEEE, 2011.

\bibitem[Kaelbling and Lozano-P{\'e}rez(2013)]{kaelbling2013integrated}
Leslie~Pack Kaelbling and Tom{\'a}s Lozano-P{\'e}rez.
\newblock Integrated task and motion planning in belief space.
\newblock \emph{The International Journal of Robotics Research}, 32\penalty0 (9-10):\penalty0 1194--1227, 2013.

\bibitem[Kaelbling and Lozano-P{\'e}rez(2017)]{kaelbling2017learning}
Leslie~Pack Kaelbling and Tom{\'a}s Lozano-P{\'e}rez.
\newblock Learning composable models of parameterized skills.
\newblock In \emph{2017 IEEE International Conference on Robotics and Automation (ICRA)}, pages 886--893. IEEE, 2017.

\bibitem[Kalashnikov et~al.(2018{\natexlab{a}})Kalashnikov, Irpan, Pastor, Ibarz, Herzog, Jang, Quillen, Holly, Kalakrishnan, Vanhoucke, et~al.]{kalashnikov2018qt}
Dmitry Kalashnikov, Alex Irpan, Peter Pastor, Julian Ibarz, Alexander Herzog, Eric Jang, Deirdre Quillen, Ethan Holly, Mrinal Kalakrishnan, Vincent Vanhoucke, et~al.
\newblock Qt-opt: Scalable deep reinforcement learning for vision-based robotic manipulation.
\newblock \emph{arXiv preprint arXiv:1806.10293}, 2018{\natexlab{a}}.

\bibitem[Kalashnikov et~al.(2018{\natexlab{b}})Kalashnikov, Irpan, Pastor, Ibarz, Herzog, Jang, Quillen, Holly, Kalakrishnan, Vanhoucke, et~al.]{kalashnikov2018scalable}
Dmitry Kalashnikov, Alex Irpan, Peter Pastor, Julian Ibarz, Alexander Herzog, Eric Jang, Deirdre Quillen, Ethan Holly, Mrinal Kalakrishnan, Vincent Vanhoucke, et~al.
\newblock Scalable deep reinforcement learning for vision-based robotic manipulation.
\newblock In \emph{Conference on Robot Learning}, pages 651--673. PMLR, 2018{\natexlab{b}}.

\bibitem[Kalashnikov et~al.(2021)Kalashnikov, Varley, Chebotar, Swanson, Jonschkowski, Finn, Levine, and Hausman]{kalashnikov2021mt}
Dmitry Kalashnikov, Jacob Varley, Yevgen Chebotar, Benjamin Swanson, Rico Jonschkowski, Chelsea Finn, Sergey Levine, and Karol Hausman.
\newblock Mt-opt: Continuous multi-task robotic reinforcement learning at scale.
\newblock \emph{arXiv preprint arXiv:2104.08212}, 2021.

\bibitem[Kappler et~al.(2018)Kappler, Meier, Issac, Mainprice, Cifuentes, W{\"u}thrich, Berenz, Schaal, Ratliff, and Bohg]{kappler2018real}
Daniel Kappler, Franziska Meier, Jan Issac, Jim Mainprice, Cristina~Garcia Cifuentes, Manuel W{\"u}thrich, Vincent Berenz, Stefan Schaal, Nathan Ratliff, and Jeannette Bohg.
\newblock Real-time perception meets reactive motion generation.
\newblock \emph{IEEE Robotics and Automation Letters}, 3\penalty0 (3):\penalty0 1864--1871, 2018.

\bibitem[Karaman and Frazzoli(2011)]{karaman2011sampling}
Sertac Karaman and Emilio Frazzoli.
\newblock Sampling-based algorithms for optimal motion planning.
\newblock \emph{The international journal of robotics research}, 30\penalty0 (7):\penalty0 846--894, 2011.

\bibitem[Katara et~al.(2023)Katara, Xian, and Fragkiadaki]{katara2023gen2sim}
Pushkal Katara, Zhou Xian, and Katerina Fragkiadaki.
\newblock Gen2sim: Scaling up robot learning in simulation with generative models, 2023.

\bibitem[Kavraki et~al.(1996)Kavraki, Svestka, Latombe, and Overmars]{kavraki1996probabilistic}
Lydia~E Kavraki, Petr Svestka, J-C Latombe, and Mark~H Overmars.
\newblock Probabilistic roadmaps for path planning in high-dimensional configuration spaces.
\newblock \emph{IEEE transactions on Robotics and Automation}, 12\penalty0 (4):\penalty0 566--580, 1996.

\bibitem[Kelly et~al.(2018)Kelly, Sidrane, Driggs-Campbell, and Kochenderfer]{kelly2018hgdagger}
Michael Kelly, Chelsea Sidrane, Katherine Driggs-Campbell, and Mykel~J. Kochenderfer.
\newblock Hg-dagger: Interactive imitation learning with human experts.
\newblock \emph{arXiv preprint arXiv: Arxiv-1810.02890}, 2018.

\bibitem[Kelly et~al.(2019)Kelly, Sidrane, Driggs-Campbell, and Kochenderfer]{kelly2019hg}
Michael Kelly, Chelsea Sidrane, Katherine Driggs-Campbell, and Mykel~J Kochenderfer.
\newblock Hg-dagger: Interactive imitation learning with human experts.
\newblock In \emph{2019 International Conference on Robotics and Automation (ICRA)}, pages 8077--8083. IEEE, 2019.

\bibitem[Khatib(1986)]{khatib1986real}
Oussama Khatib.
\newblock Real-time obstacle avoidance for manipulators and mobile robots.
\newblock \emph{The international journal of robotics research}, 5\penalty0 (1):\penalty0 90--98, 1986.

\bibitem[Khatib(1987)]{khatib1987unified}
Oussama Khatib.
\newblock A unified approach for motion and force control of robot manipulators: The operational space formulation.
\newblock \emph{IEEE Journal on Robotics and Automation}, 3\penalty0 (1):\penalty0 43--53, 1987.

\bibitem[Khazatsky et~al.(2024)Khazatsky, Pertsch, Nair, Balakrishna, Dasari, Karamcheti, Nasiriany, Srirama, Chen, Ellis, et~al.]{khazatsky2024droid}
Alexander Khazatsky, Karl Pertsch, Suraj Nair, Ashwin Balakrishna, Sudeep Dasari, Siddharth Karamcheti, Soroush Nasiriany, Mohan~Kumar Srirama, Lawrence~Yunliang Chen, Kirsty Ellis, et~al.
\newblock Droid: A large-scale in-the-wild robot manipulation dataset.
\newblock \emph{arXiv preprint arXiv:2403.12945}, 2024.

\bibitem[Kim et~al.(2024)Kim, Pertsch, Karamcheti, Xiao, Balakrishna, Nair, Rafailov, Foster, Lam, Sanketi, et~al.]{kim2024openvla}
Moo~Jin Kim, Karl Pertsch, Siddharth Karamcheti, Ted Xiao, Ashwin Balakrishna, Suraj Nair, Rafael Rafailov, Ethan Foster, Grace Lam, Pannag Sanketi, et~al.
\newblock Openvla: An open-source vision-language-action model.
\newblock \emph{arXiv preprint arXiv:2406.09246}, 2024.

\bibitem[Kirillov et~al.(2023)Kirillov, Mintun, Ravi, Mao, Rolland, Gustafson, Xiao, Whitehead, Berg, Lo, et~al.]{kirillov2023segment}
Alexander Kirillov, Eric Mintun, Nikhila Ravi, Hanzi Mao, Chloe Rolland, Laura Gustafson, Tete Xiao, Spencer Whitehead, Alexander~C Berg, Wan-Yen Lo, et~al.
\newblock Segment anything.
\newblock \emph{arXiv preprint arXiv:2304.02643}, 2023.

\bibitem[Kober and Peters(2009)]{kober2009learning}
Jens Kober and Jan Peters.
\newblock Learning motor primitives for robotics.
\newblock In \emph{2009 IEEE International Conference on Robotics and Automation}, pages 2112--2118. IEEE, 2009.

\bibitem[Koenig and Likhachev(2006)]{koenig2006new}
Sven Koenig and Maxim Likhachev.
\newblock A new principle for incremental heuristic search: Theoretical results.
\newblock In \emph{ICAPS}, pages 402--405, 2006.

\bibitem[Kormushev et~al.(2011)Kormushev, Calinon, and Caldwell]{kormushev2011imitation}
Petar Kormushev, Sylvain Calinon, and Darwin~G Caldwell.
\newblock Imitation learning of positional and force skills demonstrated via kinesthetic teaching and haptic input.
\newblock \emph{Advanced Robotics}, 25\penalty0 (5):\penalty0 581--603, 2011.

\bibitem[Kostrikov(2018)]{pytorchrl}
Ilya Kostrikov.
\newblock Pytorch implementations of reinforcement learning algorithms.
\newblock \url{https://github.com/ikostrikov/pytorch-a2c-ppo-acktr-gail}, 2018.

\bibitem[Kostrikov et~al.(2020)Kostrikov, Yarats, and Fergus]{kostrikov2020image}
Ilya Kostrikov, Denis Yarats, and Rob Fergus.
\newblock Image augmentation is all you need: Regularizing deep reinforcement learning from pixels.
\newblock \emph{arXiv preprint arXiv:2004.13649}, 2020.

\bibitem[Kostrikov et~al.(2021)Kostrikov, Nair, and Levine]{kostrikov2021offline}
Ilya Kostrikov, Ashvin Nair, and Sergey Levine.
\newblock Offline reinforcement learning with implicit q-learning.
\newblock \emph{arXiv preprint arXiv: Arxiv-2110.06169}, 2021.

\bibitem[Kuffner and LaValle(2000)]{kuffner2000rrt}
James~J Kuffner and Steven~M LaValle.
\newblock Rrt-connect: An efficient approach to single-query path planning.
\newblock In \emph{Proceedings 2000 ICRA. Millennium Conference. IEEE International Conference on Robotics and Automation. Symposia Proceedings (Cat. No. 00CH37065)}, volume~2, pages 995--1001. IEEE, 2000.

\bibitem[{Kuffner Jr.} and LaValle(2000{\natexlab{a}})]{KuffnerLaValle}
James~J {Kuffner Jr.} and Steven~M LaValle.
\newblock {RRT-Connect: An efficient approach to single-query path planning}.
\newblock In \emph{IEEE International Conference on Robotics and Automation (ICRA)}, 2000{\natexlab{a}}.

\bibitem[{Kuffner Jr.} and LaValle(2000{\natexlab{b}})]{KuffnerLaValleRRT}
James~J {Kuffner Jr.} and Steven~M LaValle.
\newblock {RRT-Connect: An efficient approach to single-query path planning}.
\newblock In \emph{IEEE International Conference on Robotics and Automation (ICRA)}, 2000{\natexlab{b}}.

\bibitem[Kumar et~al.(2021)Kumar, Fu, Pathak, and Malik]{kumar2021rma}
Ashish Kumar, Zipeng Fu, Deepak Pathak, and Jitendra Malik.
\newblock Rma: Rapid motor adaptation for legged robots.
\newblock 2021.

\bibitem[Kwon et~al.(2023)Kwon, Xie, Bullard, and Sadigh]{kwon2023reward}
Minae Kwon, Sang~Michael Xie, Kalesha Bullard, and Dorsa Sadigh.
\newblock Reward design with language models.
\newblock \emph{arXiv preprint arXiv:2303.00001}, 2023.

\bibitem[Kwon et~al.(2024)Kwon, Di~Palo, and Johns]{kwon2024language}
Teyun Kwon, Norman Di~Palo, and Edward Johns.
\newblock Language models as zero-shot trajectory generators.
\newblock \emph{IEEE Robotics and Automation Letters}, 2024.

\bibitem[Labb{\'e} et~al.(2022)Labb{\'e}, Manuelli, Mousavian, Tyree, Birchfield, Tremblay, Carpentier, Aubry, Fox, and Sivic]{labbe2022megapose}
Yann Labb{\'e}, Lucas Manuelli, Arsalan Mousavian, Stephen Tyree, Stan Birchfield, Jonathan Tremblay, Justin Carpentier, Mathieu Aubry, Dieter Fox, and Josef Sivic.
\newblock Megapose: 6d pose estimation of novel objects via render \& compare.
\newblock \emph{arXiv preprint arXiv:2212.06870}, 2022.

\bibitem[Laskin et~al.(2020)Laskin, Lee, Stooke, Pinto, Abbeel, and Srinivas]{laskin2020reinforcement}
Michael Laskin, Kimin Lee, Adam Stooke, Lerrel Pinto, Pieter Abbeel, and Aravind Srinivas.
\newblock Reinforcement learning with augmented data, 2020.

\bibitem[LaValle and Kuffner(2001)]{lavalle2001rapidly}
Steven~M LaValle and James~J Kuffner.
\newblock Rapidly-exploring random trees: Progress and prospects: Steven m. lavalle, iowa state university, a james j. kuffner, jr., university of tokyo, tokyo, japan.
\newblock \emph{Algorithmic and computational robotics}, pages 303--307, 2001.

\bibitem[Lee et~al.(2020{\natexlab{a}})Lee, Hwangbo, Wellhausen, Koltun, and Hutter]{lee2020learning}
Joonho Lee, Jemin Hwangbo, Lorenz Wellhausen, Vladlen Koltun, and Marco Hutter.
\newblock Learning quadrupedal locomotion over challenging terrain.
\newblock \emph{Science robotics}, 5\penalty0 (47):\penalty0 eabc5986, 2020{\natexlab{a}}.

\bibitem[Lee et~al.(2020{\natexlab{b}})Lee, Florensa, Tremblay, Ratliff, Garg, Ramos, and Fox]{lee2020guided}
Michelle~A Lee, Carlos Florensa, Jonathan Tremblay, Nathan Ratliff, Animesh Garg, Fabio Ramos, and Dieter Fox.
\newblock Guided uncertainty-aware policy optimization: Combining learning and model-based strategies for sample-efficient policy learning.
\newblock In \emph{2020 IEEE International Conference on Robotics and Automation (ICRA)}, pages 7505--7512. IEEE, 2020{\natexlab{b}}.

\bibitem[Levine et~al.(2016)Levine, Finn, Darrell, and Abbeel]{levine2016end}
Sergey Levine, Chelsea Finn, Trevor Darrell, and Pieter Abbeel.
\newblock End-to-end training of deep visuomotor policies.
\newblock \emph{The Journal of Machine Learning Research}, 17\penalty0 (1):\penalty0 1334--1373, 2016.

\bibitem[Li et~al.(2019)Li, Florensa, Clavera, and Abbeel]{li2019sub}
Alexander~C Li, Carlos Florensa, Ignasi Clavera, and Pieter Abbeel.
\newblock Sub-policy adaptation for hierarchical reinforcement learning.
\newblock \emph{arXiv preprint arXiv:1906.05862}, 2019.

\bibitem[Likhachev et~al.(2003)Likhachev, Gordon, and Thrun]{likhachev2003ara}
Maxim Likhachev, Geoffrey~J Gordon, and Sebastian Thrun.
\newblock Ara*: Anytime a* with provable bounds on sub-optimality.
\newblock \emph{Advances in neural information processing systems}, 16, 2003.

\bibitem[Lin et~al.(2023)Lin, Agia, Migimatsu, Pavone, and Bohg]{lin2023text2motion}
Kevin Lin, Christopher Agia, Toki Migimatsu, Marco Pavone, and Jeannette Bohg.
\newblock Text2motion: From natural language instructions to feasible plans.
\newblock \emph{arXiv preprint arXiv:2303.12153}, 2023.

\bibitem[Liu et~al.(2023{\natexlab{a}})Liu, Jiang, Zhang, Liu, Zhang, Biswas, and Stone]{liu2023llm+p}
Bo~Liu, Yuqian Jiang, Xiaohan Zhang, Qiang Liu, Shiqi Zhang, Joydeep Biswas, and Peter Stone.
\newblock Llm+ p: Empowering large language models with optimal planning proficiency.
\newblock \emph{arXiv preprint arXiv:2304.11477}, 2023{\natexlab{a}}.

\bibitem[Liu et~al.(2022)Liu, Uppal, Sukhatme, Lim, Englert, and Lee]{liu2022distilling}
I-Chun~Arthur Liu, Shagun Uppal, Gaurav~S Sukhatme, Joseph~J Lim, Peter Englert, and Youngwoon Lee.
\newblock Distilling motion planner augmented policies into visual control policies for robot manipulation.
\newblock In \emph{Conference on Robot Learning}, pages 641--650. PMLR, 2022.

\bibitem[Liu et~al.(2023{\natexlab{b}})Liu, Zeng, Ren, Li, Zhang, Yang, Li, Yang, Su, Zhu, et~al.]{liu2023grounding}
Shilong Liu, Zhaoyang Zeng, Tianhe Ren, Feng Li, Hao Zhang, Jie Yang, Chunyuan Li, Jianwei Yang, Hang Su, Jun Zhu, et~al.
\newblock Grounding dino: Marrying dino with grounded pre-training for open-set object detection.
\newblock \emph{arXiv preprint arXiv:2303.05499}, 2023{\natexlab{b}}.

\bibitem[Liu et~al.(2021)Liu, Lin, Cao, Hu, Wei, Zhang, Lin, and Guo]{liu2021swin}
Ze~Liu, Yutong Lin, Yue Cao, Han Hu, Yixuan Wei, Zheng Zhang, Stephen Lin, and Baining Guo.
\newblock Swin transformer: Hierarchical vision transformer using shifted windows.
\newblock In \emph{Proceedings of the IEEE/CVF International Conference on Computer Vision}, pages 10012--10022, 2021.

\bibitem[Lozano-Perez et~al.(1984)Lozano-Perez, Mason, and Taylor]{lozano1984automatic}
Tomas Lozano-Perez, Matthew~T Mason, and Russell~H Taylor.
\newblock Automatic synthesis of fine-motion strategies for robots.
\newblock \emph{The International Journal of Robotics Research}, 3\penalty0 (1):\penalty0 3--24, 1984.

\bibitem[Lum et~al.(2024)Lum, Matak, Makoviychuk, Handa, Allshire, Hermans, Ratliff, and Van~Wyk]{lum2024dextrah}
Tyler Ga~Wei Lum, Martin Matak, Viktor Makoviychuk, Ankur Handa, Arthur Allshire, Tucker Hermans, Nathan~D Ratliff, and Karl Van~Wyk.
\newblock Dextrah-g: Pixels-to-action dexterous arm-hand grasping with geometric fabrics.
\newblock \emph{arXiv preprint arXiv:2407.02274}, 2024.

\bibitem[Lynch and Sermanet(2020{\natexlab{a}})]{lynch2020grounding}
Corey Lynch and Pierre Sermanet.
\newblock Grounding language in play.
\newblock \emph{arXiv preprint arXiv:2005.07648}, 2020{\natexlab{a}}.

\bibitem[Lynch and Sermanet(2020{\natexlab{b}})]{lynch2020language}
Corey Lynch and Pierre Sermanet.
\newblock Language conditioned imitation learning over unstructured data.
\newblock \emph{arXiv preprint arXiv:2005.07648}, 2020{\natexlab{b}}.

\bibitem[Lynch et~al.(2020)Lynch, Khansari, Xiao, Kumar, Tompson, Levine, and Sermanet]{lynch2020learning}
Corey Lynch, Mohi Khansari, Ted Xiao, Vikash Kumar, Jonathan Tompson, Sergey Levine, and Pierre Sermanet.
\newblock Learning latent plans from play.
\newblock In \emph{Conference on Robot Learning}, pages 1113--1132. PMLR, 2020.

\bibitem[Macklin et~al.(2014)Macklin, M{\"u}ller, Chentanez, and Kim]{macklin2014unified}
Miles Macklin, Matthias M{\"u}ller, Nuttapong Chentanez, and Tae-Yong Kim.
\newblock Unified particle physics for real-time applications.
\newblock \emph{ACM Transactions on Graphics (TOG)}, 33\penalty0 (4):\penalty0 1--12, 2014.

\bibitem[Mahler et~al.(2016)Mahler, Pokorny, Hou, Roderick, Laskey, Aubry, Kohlhoff, Kr{\"o}ger, Kuffner, and Goldberg]{mahler2016dex}
Jeffrey Mahler, Florian~T Pokorny, Brian Hou, Melrose Roderick, Michael Laskey, Mathieu Aubry, Kai Kohlhoff, Torsten Kr{\"o}ger, James Kuffner, and Ken Goldberg.
\newblock Dex-net 1.0: A cloud-based network of 3d objects for robust grasp planning using a multi-armed bandit model with correlated rewards.
\newblock In \emph{IEEE International Conference on Robotics and Automation (ICRA)}, pages 1957--1964. IEEE, 2016.

\bibitem[Makoviychuk et~al.(2021)Makoviychuk, Wawrzyniak, Guo, Lu, Storey, Macklin, Hoeller, Rudin, Allshire, Handa, et~al.]{makoviychuk2021isaac}
Viktor Makoviychuk, Lukasz Wawrzyniak, Yunrong Guo, Michelle Lu, Kier Storey, Miles Macklin, David Hoeller, Nikita Rudin, Arthur Allshire, Ankur Handa, et~al.
\newblock Isaac gym: High performance gpu-based physics simulation for robot learning.
\newblock \emph{arXiv preprint arXiv:2108.10470}, 2021.

\bibitem[Mandlekar et~al.(2018)Mandlekar, Zhu, Garg, Booher, Spero, Tung, Gao, Emmons, Gupta, Orbay, et~al.]{mandlekar2018roboturk}
Ajay Mandlekar, Yuke Zhu, Animesh Garg, Jonathan Booher, Max Spero, Albert Tung, Julian Gao, John Emmons, Anchit Gupta, Emre Orbay, et~al.
\newblock Roboturk: A crowdsourcing platform for robotic skill learning through imitation.
\newblock In \emph{Conference on Robot Learning}, pages 879--893. PMLR, 2018.

\bibitem[Mandlekar et~al.(2019)Mandlekar, Booher, Spero, Tung, Gupta, Zhu, Garg, Savarese, and Fei-Fei]{mandlekar2019scaling}
Ajay Mandlekar, Jonathan Booher, Max Spero, Albert Tung, Anchit Gupta, Yuke Zhu, Animesh Garg, Silvio Savarese, and Li~Fei-Fei.
\newblock Scaling robot supervision to hundreds of hours with roboturk: Robotic manipulation dataset through human reasoning and dexterity.
\newblock \emph{arXiv preprint arXiv:1911.04052}, 2019.

\bibitem[Mandlekar et~al.(2020{\natexlab{a}})Mandlekar, Xu, Mart{\'i}n-Mart{\'i}n, Savarese, and Fei-Fei]{mandlekar2020learning}
Ajay Mandlekar, Danfei Xu, Roberto Mart{\'i}n-Mart{\'i}n, Silvio Savarese, and Li~Fei-Fei.
\newblock Learning to generalize across long-horizon tasks from human demonstrations.
\newblock \emph{arXiv preprint arXiv:2003.06085}, 2020{\natexlab{a}}.

\bibitem[Mandlekar et~al.(2020{\natexlab{b}})Mandlekar, Xu, Mart{\'\i}n-Mart{\'\i}n, Zhu, Fei-Fei, and Savarese]{mandlekar2020human}
Ajay Mandlekar, Danfei Xu, Roberto Mart{\'\i}n-Mart{\'\i}n, Yuke Zhu, Li~Fei-Fei, and Silvio Savarese.
\newblock Human-in-the-loop imitation learning using remote teleoperation.
\newblock \emph{arXiv preprint arXiv:2012.06733}, 2020{\natexlab{b}}.

\bibitem[Mandlekar et~al.(2020{\natexlab{c}})Mandlekar, Xu, Martín-Martín, Zhu, Fei-Fei, and Savarese]{mandlekar2020humanintheloop}
Ajay Mandlekar, Danfei Xu, Roberto Martín-Martín, Yuke Zhu, Li~Fei-Fei, and Silvio Savarese.
\newblock Human-in-the-loop imitation learning using remote teleoperation.
\newblock \emph{arXiv preprint arXiv: Arxiv-2012.06733}, 2020{\natexlab{c}}.

\bibitem[Mandlekar et~al.(2021{\natexlab{a}})Mandlekar, Xu, Wong, Nasiriany, Wang, Kulkarni, Fei-Fei, Savarese, Zhu, and Mart\'{i}n-Mart\'{i}n]{robomimic2021}
Ajay Mandlekar, Danfei Xu, Josiah Wong, Soroush Nasiriany, Chen Wang, Rohun Kulkarni, Li~Fei-Fei, Silvio Savarese, Yuke Zhu, and Roberto Mart\'{i}n-Mart\'{i}n.
\newblock What matters in learning from offline human demonstrations for robot manipulation.
\newblock In \emph{arXiv preprint arXiv:2108.03298}, 2021{\natexlab{a}}.

\bibitem[Mandlekar et~al.(2021{\natexlab{b}})Mandlekar, Xu, Wong, Nasiriany, Wang, Kulkarni, Fei-Fei, Savarese, Zhu, and Martín-Martín]{mandlekar2021matters}
Ajay Mandlekar, Danfei Xu, Josiah Wong, Soroush Nasiriany, Chen Wang, Rohun Kulkarni, Li~Fei-Fei, Silvio Savarese, Yuke Zhu, and Roberto Martín-Martín.
\newblock What matters in learning from offline human demonstrations for robot manipulation.
\newblock \emph{arXiv preprint arXiv: Arxiv-2108.03298}, 2021{\natexlab{b}}.

\bibitem[Mandlekar et~al.(2023{\natexlab{a}})Mandlekar, Garret, Xu, and Fox]{mandlekarhitltamp}
Ajay Mandlekar, Caelan Garret, Danfei Xu, and Dieter Fox.
\newblock Human-in-the-loop task and motion planning for imitation learning.
\newblock \emph{Conference on Robot Learning}, 2023{\natexlab{a}}.

\bibitem[Mandlekar et~al.(2023{\natexlab{b}})Mandlekar, Garrett, Xu, and Fox]{mandlekar2023hitltamp}
Ajay Mandlekar, Caelan Garrett, Danfei Xu, and Dieter Fox.
\newblock Human-in-the-loop task and motion planning for imitation learning.
\newblock \emph{Workshop on effective Representations, Abstractions, and Priors for Robot Learning (RAP4Robots)}, 2023{\natexlab{b}}.

\bibitem[Mao et~al.(2022)Mao, Zhao, Chen, Hao, Chen, Li, Zhang, and Xiao]{mao2022transformer}
Hangyu Mao, Rui Zhao, Hao Chen, Jianye Hao, Yiqun Chen, Dong Li, Junge Zhang, and Zhen Xiao.
\newblock Transformer in transformer as backbone for deep reinforcement learning.
\newblock \emph{arXiv preprint arXiv:2212.14538}, 2022.

\bibitem[Mart{\'\i}n-Mart{\'\i}n et~al.(2019)Mart{\'\i}n-Mart{\'\i}n, Lee, Gardner, Savarese, Bohg, and Garg]{martin2019variable}
Roberto Mart{\'\i}n-Mart{\'\i}n, Michelle~A Lee, Rachel Gardner, Silvio Savarese, Jeannette Bohg, and Animesh Garg.
\newblock Variable impedance control in end-effector space: An action space for reinforcement learning in contact-rich tasks.
\newblock In \emph{2019 IEEE/RSJ international conference on intelligent robots and systems (IROS)}, pages 1010--1017. IEEE, 2019.

\bibitem[Martín-Martín et~al.(2019)Martín-Martín, Lee, Gardner, Savarese, Bohg, and Garg]{martinvices}
Roberto Martín-Martín, Michelle~A. Lee, Rachel Gardner, Silvio Savarese, Jeannette Bohg, and Animesh Garg.
\newblock Variable impedance control in end-effector space: An action space for reinforcement learning in contact-rich tasks, 2019.

\bibitem[Mason(2001)]{mason2001mechanics}
Matthew~T Mason.
\newblock \emph{Mechanics of robotic manipulation}.
\newblock MIT press, 2001.

\bibitem[Masson et~al.(2016)Masson, Ranchod, and Konidaris]{masson2016reinforcement}
Warwick Masson, Pravesh Ranchod, and George Konidaris.
\newblock Reinforcement learning with parameterized actions.
\newblock In \emph{Proceedings of the AAAI Conference on Artificial Intelligence}, volume~30, 2016.

\bibitem[McDonald and Hadfield-Menell(2022)]{mcdonald2022guided}
Michael~James McDonald and Dylan Hadfield-Menell.
\newblock Guided imitation of task and motion planning.
\newblock In \emph{Conference on Robot Learning}, pages 630--640. PMLR, 2022.

\bibitem[Miller and Allen(2004)]{miller2004graspit}
Andrew~T Miller and Peter~K Allen.
\newblock Graspit! a versatile simulator for robotic grasping.
\newblock \emph{IEEE Robotics \& Automation Magazine}, 11\penalty0 (4):\penalty0 110--122, 2004.

\bibitem[Mo et~al.(2019)Mo, Zhu, Chang, Yi, Tripathi, Guibas, and Su]{mo2019partnet}
Kaichun Mo, Shilin Zhu, Angel~X Chang, Li~Yi, Subarna Tripathi, Leonidas~J Guibas, and Hao Su.
\newblock Partnet: A large-scale benchmark for fine-grained and hierarchical part-level 3d object understanding.
\newblock In \emph{Proceedings of the IEEE/CVF conference on computer vision and pattern recognition}, pages 909--918, 2019.

\bibitem[Mousavian et~al.(2019)Mousavian, Eppner, and Fox]{mousavian20196}
Arsalan Mousavian, Clemens Eppner, and Dieter Fox.
\newblock 6-dof graspnet: Variational grasp generation for object manipulation.
\newblock In \emph{Proceedings of the IEEE/CVF International Conference on Computer Vision}, pages 2901--2910, 2019.

\bibitem[Murphy(2019)]{murphy2019introduction}
Robin~R Murphy.
\newblock \emph{Introduction to AI robotics}.
\newblock MIT press, 2019.

\bibitem[Nachum et~al.(2018)Nachum, Gu, Lee, and Levine]{nachum2018data}
Ofir Nachum, Shixiang Gu, Honglak Lee, and Sergey Levine.
\newblock Data-efficient hierarchical reinforcement learning.
\newblock \emph{arXiv preprint arXiv:1805.08296}, 2018.

\bibitem[Nagabandi et~al.(2020)Nagabandi, Konolige, Levine, and Kumar]{nagabandi2020deep}
Anusha Nagabandi, Kurt Konolige, Sergey Levine, and Vikash Kumar.
\newblock Deep dynamics models for learning dexterous manipulation.
\newblock In \emph{Conference on Robot Learning}, pages 1101--1112. PMLR, 2020.

\bibitem[Narang et~al.(2022)Narang, Storey, Akinola, Macklin, Reist, Wawrzyniak, Guo, Morav{\'{a}}nszky, State, Lu, Handa, and Fox]{narang2022factory}
Yashraj~S. Narang, Kier Storey, Iretiayo Akinola, Miles Macklin, Philipp Reist, Lukasz Wawrzyniak, Yunrong Guo, {\'{A}}d{\'{a}}m Morav{\'{a}}nszky, Gavriel State, Michelle Lu, Ankur Handa, and Dieter Fox.
\newblock Factory: Fast contact for robotic assembly.
\newblock In Kris Hauser, Dylan~A. Shell, and Shoudong Huang, editors, \emph{Robotics: Science and Systems XVIII, New York City, NY, USA, June 27 - July 1, 2022}, 2022.
\newblock \doi{10.15607/RSS.2022.XVIII.035}.
\newblock URL \url{https://doi.org/10.15607/RSS.2022.XVIII.035}.

\bibitem[Nasiriany et~al.(2021)Nasiriany, Liu, and Zhu]{nasiriany2021augmenting}
Soroush Nasiriany, Huihan Liu, and Yuke Zhu.
\newblock Augmenting reinforcement learning with behavior primitives for diverse manipulation tasks, 2021.

\bibitem[Niekum et~al.(2013)Niekum, Chitta, Barto, Marthi, and Osentoski]{niekum2013incremental}
Scott Niekum, Sachin Chitta, Andrew~G Barto, Bhaskara Marthi, and Sarah Osentoski.
\newblock Incremental semantically grounded learning from demonstration.
\newblock In \emph{Robotics: Science and Systems}, volume~9, pages 10--15607. Berlin, Germany, 2013.

\bibitem[OpenAI(2023)]{openai2023gpt4}
R~OpenAI.
\newblock Gpt-4 technical report.
\newblock \emph{arXiv}, pages 2303--08774, 2023.

\bibitem[Padalkar et~al.(2023)Padalkar, Pooley, Jain, Bewley, Herzog, Irpan, Khazatsky, Rai, Singh, Brohan, et~al.]{padalkar2023open}
Abhishek Padalkar, Acorn Pooley, Ajinkya Jain, Alex Bewley, Alex Herzog, Alex Irpan, Alexander Khazatsky, Anant Rai, Anikait Singh, Anthony Brohan, et~al.
\newblock Open x-embodiment: Robotic learning datasets and rt-x models.
\newblock \emph{arXiv preprint arXiv:2310.08864}, 2023.

\bibitem[Parr and Russell(1997)]{parr1997reinforcement}
Ronald Parr and Stuart Russell.
\newblock Reinforcement learning with hierarchies of machines.
\newblock \emph{Advances in neural information processing systems}, 10, 1997.

\bibitem[Paul(1981)]{paul1981robot}
Richard~P Paul.
\newblock \emph{Robot manipulators: mathematics, programming, and control: the computer control of robot manipulators}.
\newblock Richard Paul, 1981.

\bibitem[Peng et~al.(2017)Peng, Andrychowicz, Zaremba, and Abbeel]{peng2017simtoreal}
Xue~Bin Peng, Marcin Andrychowicz, Wojciech Zaremba, and Pieter Abbeel.
\newblock Sim-to-real transfer of robotic control with dynamics randomization.
\newblock \emph{arXiv preprint arXiv: Arxiv-1710.06537}, 2017.

\bibitem[Pertsch et~al.(2020)Pertsch, Lee, and Lim]{pertsch2020accelerating}
Karl Pertsch, Youngwoon Lee, and Joseph~J. Lim.
\newblock Accelerating reinforcement learning with learned skill priors, 2020.

\bibitem[Peters et~al.(2010)Peters, Mulling, and Altun]{peters2010relative}
Jan Peters, Katharina Mulling, and Yasemin Altun.
\newblock Relative entropy policy search.
\newblock In \emph{Proceedings of the AAAI Conference on Artificial Intelligence}, volume~24, 2010.

\bibitem[Pomerleau(1988)]{NIPS1988_812b4ba2}
Dean~A. Pomerleau.
\newblock Alvinn: An autonomous land vehicle in a neural network.
\newblock In D.~Touretzky, editor, \emph{Advances in Neural Information Processing Systems}, volume~1. Morgan-Kaufmann, 1988.
\newblock URL \url{https://proceedings.neurips.cc/paper_files/paper/1988/file/812b4ba287f5ee0bc9d43bbf5bbe87fb-Paper.pdf}.

\bibitem[Pomerleau(1989)]{pomerleau1989alvinn}
Dean~A Pomerleau.
\newblock Alvinn: An autonomous land vehicle in a neural network.
\newblock In \emph{Advances in neural information processing systems}, pages 305--313, 1989.

\bibitem[Pong et~al.(2019)Pong, Dalal, Lin, Nair, Bahl, and Levine]{pong2019skew}
Vitchyr~H Pong, Murtaza Dalal, Steven Lin, Ashvin Nair, Shikhar Bahl, and Sergey Levine.
\newblock Skew-fit: State-covering self-supervised reinforcement learning.
\newblock \emph{arXiv preprint arXiv:1903.03698}, 2019.

\bibitem[Poole et~al.(2022)Poole, Jain, Barron, and Mildenhall]{poole2022dreamfusion}
Ben Poole, Ajay Jain, Jonathan~T Barron, and Ben Mildenhall.
\newblock Dreamfusion: Text-to-3d using 2d diffusion.
\newblock \emph{arXiv preprint arXiv:2209.14988}, 2022.

\bibitem[Qi et~al.(2017)Qi, Yi, Su, and Guibas]{qi2017pointnet++}
Charles~Ruizhongtai Qi, Li~Yi, Hao Su, and Leonidas~J Guibas.
\newblock Pointnet++: Deep hierarchical feature learning on point sets in a metric space.
\newblock \emph{Advances in neural information processing systems}, 30, 2017.

\bibitem[Quinlan and Khatib(1993)]{quinlan1993elastic}
Sean Quinlan and Oussama Khatib.
\newblock Elastic bands: Connecting path planning and control.
\newblock In \emph{[1993] Proceedings IEEE International Conference on Robotics and Automation}, pages 802--807. IEEE, 1993.

\bibitem[Qureshi et~al.(2019)Qureshi, Simeonov, Bency, and Yip]{qureshi2019motion}
Ahmed~H Qureshi, Anthony Simeonov, Mayur~J Bency, and Michael~C Yip.
\newblock Motion planning networks.
\newblock In \emph{2019 International Conference on Robotics and Automation (ICRA)}, pages 2118--2124. IEEE, 2019.

\bibitem[Qureshi et~al.(2020)Qureshi, Dong, Choe, and Yip]{qureshi2020neural}
Ahmed~H Qureshi, Jiangeng Dong, Austin Choe, and Michael~C Yip.
\newblock Neural manipulation planning on constraint manifolds.
\newblock \emph{IEEE Robotics and Automation Letters}, 5\penalty0 (4):\penalty0 6089--6096, 2020.

\bibitem[Qureshi et~al.(2021)Qureshi, Mousavian, Paxton, Yip, and Fox]{qureshi2021nerp}
Ahmed~H Qureshi, Arsalan Mousavian, Chris Paxton, Michael~C Yip, and Dieter Fox.
\newblock Nerp: Neural rearrangement planning for unknown objects.
\newblock \emph{arXiv preprint arXiv:2106.01352}, 2021.

\bibitem[Rana et~al.(2023)Rana, Haviland, Garg, Abou-Chakra, Reid, and Suenderhauf]{rana2023sayplan}
Krishan Rana, Jesse Haviland, Sourav Garg, Jad Abou-Chakra, Ian Reid, and Niko Suenderhauf.
\newblock Sayplan: Grounding large language models using 3d scene graphs for scalable task planning.
\newblock \emph{arXiv preprint arXiv:2307.06135}, 2023.

\bibitem[Ratliff et~al.(2009)Ratliff, Zucker, Bagnell, and Srinivasa]{ratliff2009chomp}
Nathan Ratliff, Matt Zucker, J~Andrew Bagnell, and Siddhartha Srinivasa.
\newblock Chomp: Gradient optimization techniques for efficient motion planning.
\newblock In \emph{2009 IEEE international conference on robotics and automation}, pages 489--494. IEEE, 2009.

\bibitem[Reed et~al.(2022)Reed, Zolna, Parisotto, Colmenarejo, Novikov, Barth-Maron, Gimenez, Sulsky, Kay, Springenberg, et~al.]{reed2022generalist}
Scott Reed, Konrad Zolna, Emilio Parisotto, Sergio~Gomez Colmenarejo, Alexander Novikov, Gabriel Barth-Maron, Mai Gimenez, Yury Sulsky, Jackie Kay, Jost~Tobias Springenberg, et~al.
\newblock A generalist agent.
\newblock \emph{arXiv preprint arXiv:2205.06175}, 2022.

\bibitem[Ren et~al.(2024)Ren, Liu, Zeng, Lin, Li, Cao, Chen, Huang, Chen, Yan, Zeng, Zhang, Li, Yang, Li, Jiang, and Zhang]{ren2024grounded}
Tianhe Ren, Shilong Liu, Ailing Zeng, Jing Lin, Kunchang Li, He~Cao, Jiayu Chen, Xinyu Huang, Yukang Chen, Feng Yan, Zhaoyang Zeng, Hao Zhang, Feng Li, Jie Yang, Hongyang Li, Qing Jiang, and Lei Zhang.
\newblock Grounded sam: Assembling open-world models for diverse visual tasks, 2024.

\bibitem[Ritchie et~al.(2019)Ritchie, Wang, and Lin]{ritchie2019fast}
Daniel Ritchie, Kai Wang, and Yu-an Lin.
\newblock Fast and flexible indoor scene synthesis via deep convolutional generative models.
\newblock In \emph{Proceedings of the IEEE/CVF Conference on Computer Vision and Pattern Recognition}, pages 6182--6190, 2019.

\bibitem[Rosete-Beas et~al.(2022)Rosete-Beas, Mees, Kalweit, Boedecker, and Burgard]{rosete2022latent}
Erick Rosete-Beas, Oier Mees, Gabriel Kalweit, Joschka Boedecker, and Wolfram Burgard.
\newblock Latent plans for task-agnostic offline reinforcement learning.
\newblock \emph{arXiv preprint arXiv:2209.08959}, 2022.

\bibitem[Ross et~al.(2011)Ross, Gordon, and Bagnell]{ross2011reduction}
St{\'e}phane Ross, Geoffrey Gordon, and Drew Bagnell.
\newblock A reduction of imitation learning and structured prediction to no-regret online learning.
\newblock In \emph{Proceedings of the fourteenth international conference on artificial intelligence and statistics}, pages 627--635. JMLR Workshop and Conference Proceedings, 2011.

\bibitem[Saha et~al.(2023)Saha, Mandadi, Reddy, Srikanth, Agarwal, Sen, Singh, and Krishna]{saha2023edmp}
Kallol Saha, Vishal Mandadi, Jayaram Reddy, Ajit Srikanth, Aditya Agarwal, Bipasha Sen, Arun Singh, and Madhava Krishna.
\newblock Edmp: Ensemble-of-costs-guided diffusion for motion planning.
\newblock \emph{arXiv preprint arXiv:2309.11414}, 2023.

\bibitem[Saharia et~al.(2022)Saharia, Chan, Saxena, Li, Whang, Denton, Ghasemipour, Ayan, Mahdavi, Lopes, et~al.]{saharia2022photorealistic}
Chitwan Saharia, William Chan, Saurabh Saxena, Lala Li, Jay Whang, Emily Denton, Seyed Kamyar~Seyed Ghasemipour, Burcu~Karagol Ayan, S~Sara Mahdavi, Rapha~Gontijo Lopes, et~al.
\newblock Photorealistic text-to-image diffusion models with deep language understanding.
\newblock \emph{arXiv preprint arXiv:2205.11487}, 2022.

\bibitem[Schaal(2006)]{schaal2006dynamic}
Stefan Schaal.
\newblock Dynamic movement primitives-a framework for motor control in humans and humanoid robotics.
\newblock In \emph{Adaptive motion of animals and machines}, pages 261--280. Springer, 2006.

\bibitem[Schuhmann et~al.(2022)Schuhmann, Beaumont, Vencu, Gordon, Wightman, Cherti, Coombes, Katta, Mullis, Wortsman, et~al.]{schuhmann2022laion}
Christoph Schuhmann, Romain Beaumont, Richard Vencu, Cade Gordon, Ross Wightman, Mehdi Cherti, Theo Coombes, Aarush Katta, Clayton Mullis, Mitchell Wortsman, et~al.
\newblock Laion-5b: An open large-scale dataset for training next generation image-text models.
\newblock \emph{arXiv preprint arXiv:2210.08402}, 2022.

\bibitem[Schulman et~al.(2013)Schulman, Ho, Lee, Awwal, Bradlow, and Abbeel]{schulman2013finding}
John Schulman, Jonathan Ho, Alex~X Lee, Ibrahim Awwal, Henry Bradlow, and Pieter Abbeel.
\newblock Finding locally optimal, collision-free trajectories with sequential convex optimization.
\newblock In \emph{Robotics: science and systems}, volume~9, pages 1--10. Berlin, Germany, 2013.

\bibitem[Schulman et~al.(2014)Schulman, Duan, Ho, Lee, Awwal, Bradlow, Pan, Patil, Goldberg, and Abbeel]{schulman2014motion}
John Schulman, Yan Duan, Jonathan Ho, Alex Lee, Ibrahim Awwal, Henry Bradlow, Jia Pan, Sachin Patil, Ken Goldberg, and Pieter Abbeel.
\newblock Motion planning with sequential convex optimization and convex collision checking.
\newblock \emph{The International Journal of Robotics Research}, 33\penalty0 (9):\penalty0 1251--1270, 2014.

\bibitem[Schulman et~al.(2017)Schulman, Wolski, Dhariwal, Radford, and Klimov]{schulman2017proximal}
John Schulman, Filip Wolski, Prafulla Dhariwal, Alec Radford, and Oleg Klimov.
\newblock Proximal policy optimization algorithms, 2017.

\bibitem[Sekar et~al.(2020)Sekar, Rybkin, Daniilidis, Abbeel, Hafner, and Pathak]{sekar2020planning}
Ramanan Sekar, Oleh Rybkin, Kostas Daniilidis, Pieter Abbeel, Danijar Hafner, and Deepak Pathak.
\newblock Planning to explore via self-supervised world models, 2020.

\bibitem[Shafiullah et~al.(2022)Shafiullah, Cui, Altanzaya, and Pinto]{shafiullah2022behavior}
Nur Muhammad~Mahi Shafiullah, Zichen~Jeff Cui, Ariuntuya Altanzaya, and Lerrel Pinto.
\newblock Behavior transformers: Cloning $ k $ modes with one stone.
\newblock \emph{arXiv preprint arXiv:2206.11251}, 2022.

\bibitem[Shankar and Gupta(2020)]{shankar2020learning}
Tanmay Shankar and Abhinav Gupta.
\newblock Learning robot skills with temporal variational inference.
\newblock In \emph{International Conference on Machine Learning}, pages 8624--8633. PMLR, 2020.

\bibitem[Shankar et~al.(2019)Shankar, Tulsiani, Pinto, and Gupta]{shankar2019discovering}
Tanmay Shankar, Shubham Tulsiani, Lerrel Pinto, and Abhinav Gupta.
\newblock Discovering motor programs by recomposing demonstrations.
\newblock In \emph{International Conference on Learning Representations}, 2019.

\bibitem[Sharma et~al.(2019)Sharma, Gu, Levine, Kumar, and Hausman]{sharma2019dynamics}
Archit Sharma, Shixiang Gu, Sergey Levine, Vikash Kumar, and Karol Hausman.
\newblock Dynamics-aware unsupervised discovery of skills.
\newblock \emph{arXiv preprint arXiv:1907.01657}, 2019.

\bibitem[Sharma et~al.(2020)Sharma, Liang, Zhao, LaGrassa, and Kroemer]{sharma2020learning}
Mohit Sharma, Jacky Liang, Jialiang Zhao, Alex LaGrassa, and Oliver Kroemer.
\newblock Learning to compose hierarchical object-centric controllers for robotic manipulation.
\newblock \emph{arXiv preprint arXiv:2011.04627}, 2020.

\bibitem[Shridhar et~al.(2022)Shridhar, Manuelli, and Fox]{shridhar2022peract}
Mohit Shridhar, Lucas Manuelli, and Dieter Fox.
\newblock Perceiver-actor: A multi-task transformer for robotic manipulation.
\newblock In \emph{Proceedings of the 6th Conference on Robot Learning (CoRL)}, 2022.

\bibitem[Shridhar et~al.(2023)Shridhar, Manuelli, and Fox]{shridhar2023perceiver}
Mohit Shridhar, Lucas Manuelli, and Dieter Fox.
\newblock Perceiver-actor: A multi-task transformer for robotic manipulation.
\newblock In \emph{Conference on Robot Learning}, pages 785--799. PMLR, 2023.

\bibitem[Simeonov et~al.(2020)Simeonov, Du, Kim, Hogan, Tenenbaum, Agrawal, and Rodriguez]{simeonov2020long}
Anthony Simeonov, Yilun Du, Beomjoon Kim, Francois~R Hogan, Joshua Tenenbaum, Pulkit Agrawal, and Alberto Rodriguez.
\newblock A long horizon planning framework for manipulating rigid pointcloud objects.
\newblock \emph{arXiv preprint arXiv:2011.08177}, 2020.

\bibitem[Singh et~al.(2020)Singh, Liu, Zhou, Yu, Rhinehart, and Levine]{singh2020parrot}
Avi Singh, Huihan Liu, Gaoyue Zhou, Albert Yu, Nicholas Rhinehart, and Sergey Levine.
\newblock Parrot: Data-driven behavioral priors for reinforcement learning.
\newblock \emph{arXiv preprint arXiv:2011.10024}, 2020.

\bibitem[Singh et~al.(2023)Singh, Blukis, Mousavian, Goyal, Xu, Tremblay, Fox, Thomason, and Garg]{singh2023progprompt}
Ishika Singh, Valts Blukis, Arsalan Mousavian, Ankit Goyal, Danfei Xu, Jonathan Tremblay, Dieter Fox, Jesse Thomason, and Animesh Garg.
\newblock Progprompt: Generating situated robot task plans using large language models.
\newblock In \emph{2023 IEEE International Conference on Robotics and Automation (ICRA)}, pages 11523--11530. IEEE, 2023.

\bibitem[Sohl-Dickstein et~al.(2015)Sohl-Dickstein, Weiss, Maheswaranathan, and Ganguli]{sohl2015deep}
Jascha Sohl-Dickstein, Eric Weiss, Niru Maheswaranathan, and Surya Ganguli.
\newblock Deep unsupervised learning using nonequilibrium thermodynamics.
\newblock In \emph{International Conference on Machine Learning}, pages 2256--2265. PMLR, 2015.

\bibitem[Song et~al.(2023)Song, Wu, Washington, Sadler, Chao, and Su]{song2023llm}
Chan~Hee Song, Jiaman Wu, Clayton Washington, Brian~M Sadler, Wei-Lun Chao, and Yu~Su.
\newblock Llm-planner: Few-shot grounded planning for embodied agents with large language models.
\newblock In \emph{Proceedings of the IEEE/CVF International Conference on Computer Vision}, pages 2998--3009, 2023.

\bibitem[Strub and Gammell(2020)]{strub2020adaptively}
Marlin~P Strub and Jonathan~D Gammell.
\newblock Adaptively informed trees (ait): Fast asymptotically optimal path planning through adaptive heuristics.
\newblock In \emph{2020 IEEE International Conference on Robotics and Automation (ICRA)}, pages 3191--3198. IEEE, 2020.

\bibitem[Sun et~al.(2017)Sun, Shrivastava, Singh, and Gupta]{sun2017revisiting}
Chen Sun, Abhinav Shrivastava, Saurabh Singh, and Abhinav Gupta.
\newblock Revisiting unreasonable effectiveness of data in deep learning era.
\newblock In \emph{Proceedings of the IEEE international conference on computer vision}, pages 843--852, 2017.

\bibitem[Sundaralingam et~al.(2023)Sundaralingam, Hari, Fishman, Garrett, Van~Wyk, Blukis, Millane, Oleynikova, Handa, Ramos, et~al.]{sundaralingam2023curobo}
Balakumar Sundaralingam, Siva Kumar~Sastry Hari, Adam Fishman, Caelan Garrett, Karl Van~Wyk, Valts Blukis, Alexander Millane, Helen Oleynikova, Ankur Handa, Fabio Ramos, et~al.
\newblock Curobo: Parallelized collision-free robot motion generation.
\newblock In \emph{2023 IEEE International Conference on Robotics and Automation (ICRA)}, pages 8112--8119. IEEE, 2023.

\bibitem[Sundermeyer et~al.(2021)Sundermeyer, Mousavian, Triebel, and Fox]{sundermeyer2021contact}
Martin Sundermeyer, Arsalan Mousavian, Rudolph Triebel, and Dieter Fox.
\newblock Contact-graspnet: Efficient 6-dof grasp generation in cluttered scenes.
\newblock In \emph{2021 IEEE International Conference on Robotics and Automation (ICRA)}, pages 13438--13444. IEEE, 2021.

\bibitem[Sutton et~al.(1999{\natexlab{a}})Sutton, Precup, and Singh]{SUTTON1999181}
Richard~S. Sutton, Doina Precup, and Satinder Singh.
\newblock Between mdps and semi-mdps: A framework for temporal abstraction in reinforcement learning.
\newblock \emph{Artificial Intelligence}, 112\penalty0 (1):\penalty0 181--211, 1999{\natexlab{a}}.
\newblock ISSN 0004-3702.
\newblock \doi{https://doi.org/10.1016/S0004-3702(99)00052-1}.
\newblock URL \url{https://www.sciencedirect.com/science/article/pii/S0004370299000521}.

\bibitem[Sutton et~al.(1999{\natexlab{b}})Sutton, Precup, and Singh]{sutton1999between}
Richard~S Sutton, Doina Precup, and Satinder Singh.
\newblock Between mdps and semi-mdps: A framework for temporal abstraction in reinforcement learning.
\newblock \emph{Artificial intelligence}, 112\penalty0 (1-2):\penalty0 181--211, 1999{\natexlab{b}}.

\bibitem[Tang et~al.(2023{\natexlab{a}})Tang, Lin, Akinola, Handa, Sukhatme, Ramos, Fox, and Narang]{tang2023industreal}
Bingjie Tang, Michael~A Lin, Iretiayo Akinola, Ankur Handa, Gaurav~S Sukhatme, Fabio Ramos, Dieter Fox, and Yashraj Narang.
\newblock Industreal: Transferring contact-rich assembly tasks from simulation to reality.
\newblock \emph{arXiv preprint arXiv:2305.17110}, 2023{\natexlab{a}}.

\bibitem[Tang et~al.(2024)Tang, Akinola, Xu, Wen, Handa, Van~Wyk, Fox, Sukhatme, Ramos, and Narang]{tang2024automate}
Bingjie Tang, Iretiayo Akinola, Jie Xu, Bowen Wen, Ankur Handa, Karl Van~Wyk, Dieter Fox, Gaurav~S Sukhatme, Fabio Ramos, and Yashraj Narang.
\newblock Automate: Specialist and generalist assembly policies over diverse geometries.
\newblock \emph{arXiv preprint arXiv:2407.08028}, 2024.

\bibitem[Tang et~al.(2023{\natexlab{b}})Tang, Yu, Tan, Zen, Faust, and Harada]{tang2023saytap}
Yujin Tang, Wenhao Yu, Jie Tan, Heiga Zen, Aleksandra Faust, and Tatsuya Harada.
\newblock Saytap: Language to quadrupedal locomotion.
\newblock \emph{arXiv preprint arXiv:2306.07580}, 2023{\natexlab{b}}.

\bibitem[Tanneberg et~al.(2021)Tanneberg, Ploeger, Rueckert, and Peters]{tanneberg2021skid}
Daniel Tanneberg, Kai Ploeger, Elmar Rueckert, and Jan Peters.
\newblock Skid raw: Skill discovery from raw trajectories.
\newblock \emph{IEEE Robotics and Automation Letters}, 6\penalty0 (3):\penalty0 4696--4703, 2021.

\bibitem[Taylor et~al.(1987)Taylor, Mason, and Goldberg]{taylor1987sensor}
Russ~H Taylor, Matthew~T Mason, and Kenneth~Y Goldberg.
\newblock Sensor-based manipulation planning as a game with nature.
\newblock In \emph{Fourth International Symposium on Robotics Research}, pages 421--429, 1987.

\bibitem[Thoppilan et~al.(2022)Thoppilan, De~Freitas, Hall, Shazeer, Kulshreshtha, Cheng, Jin, Bos, Baker, Du, et~al.]{thoppilan2022lamda}
Romal Thoppilan, Daniel De~Freitas, Jamie Hall, Noam Shazeer, Apoorv Kulshreshtha, Heng-Tze Cheng, Alicia Jin, Taylor Bos, Leslie Baker, Yu~Du, et~al.
\newblock Lamda: Language models for dialog applications.
\newblock \emph{arXiv preprint arXiv:2201.08239}, 2022.

\bibitem[Todorov et~al.(2012{\natexlab{a}})Todorov, Erez, and Tassa]{todorov12mujoco}
Emanuel Todorov, Tom Erez, and Yuval Tassa.
\newblock {MuJoCo: A physics engine for model-based control}.
\newblock In \emph{The IEEE/RSJ International Conference on Intelligent Robots and Systems}, 2012{\natexlab{a}}.

\bibitem[Todorov et~al.(2012{\natexlab{b}})Todorov, Erez, and Tassa]{todorov2012mujoco}
Emanuel Todorov, Tom Erez, and Yuval Tassa.
\newblock Mujoco: A physics engine for model-based control.
\newblock In \emph{2012 IEEE/RSJ international conference on intelligent robots and systems}, pages 5026--5033. IEEE, 2012{\natexlab{b}}.

\bibitem[Toussaint et~al.(2018)Toussaint, Allen, Smith, and Tenenbaum]{toussaint2018differentiable}
Marc~A Toussaint, Kelsey~Rebecca Allen, Kevin~A Smith, and Joshua~B Tenenbaum.
\newblock Differentiable physics and stable modes for tool-use and manipulation planning.
\newblock 2018.

\bibitem[Touvron et~al.(2023{\natexlab{a}})Touvron, Lavril, Izacard, Martinet, Lachaux, Lacroix, Rozi{\`e}re, Goyal, Hambro, Azhar, et~al.]{touvron2023llama}
Hugo Touvron, Thibaut Lavril, Gautier Izacard, Xavier Martinet, Marie-Anne Lachaux, Timoth{\'e}e Lacroix, Baptiste Rozi{\`e}re, Naman Goyal, Eric Hambro, Faisal Azhar, et~al.
\newblock Llama: Open and efficient foundation language models.
\newblock \emph{arXiv preprint arXiv:2302.13971}, 2023{\natexlab{a}}.

\bibitem[Touvron et~al.(2023{\natexlab{b}})Touvron, Martin, Stone, Albert, Almahairi, Babaei, Bashlykov, Batra, Bhargava, Bhosale, et~al.]{touvron2023llama2}
Hugo Touvron, Louis Martin, Kevin Stone, Peter Albert, Amjad Almahairi, Yasmine Babaei, Nikolay Bashlykov, Soumya Batra, Prajjwal Bhargava, Shruti Bhosale, et~al.
\newblock Llama 2: Open foundation and fine-tuned chat models.
\newblock \emph{arXiv preprint arXiv:2307.09288}, 2023{\natexlab{b}}.

\bibitem[Tung et~al.(2020)Tung, Wong, Mandlekar, Mart{\'\i}n-Mart{\'\i}n, Zhu, Fei-Fei, and Savarese]{tung2020learning}
Albert Tung, Josiah Wong, Ajay Mandlekar, Roberto Mart{\'\i}n-Mart{\'\i}n, Yuke Zhu, Li~Fei-Fei, and Silvio Savarese.
\newblock Learning multi-arm manipulation through collaborative teleoperation.
\newblock \emph{arXiv preprint arXiv:2012.06738}, 2020.

\bibitem[Uppal et~al.(2024)Uppal, Agarwal, Xiong, Shaw, and Pathak]{uppal2024spin}
Shagun Uppal, Ananye Agarwal, Haoyu Xiong, Kenneth Shaw, and Deepak Pathak.
\newblock Spin: Simultaneous perception, interaction and navigation.
\newblock \emph{arXiv preprint arXiv:2405.07991}, 2024.

\bibitem[Vaswani et~al.(2017)Vaswani, Shazeer, Parmar, Uszkoreit, Jones, Gomez, Kaiser, and Polosukhin]{vaswani2017attention}
Ashish Vaswani, Noam Shazeer, Niki Parmar, Jakob Uszkoreit, Llion Jones, Aidan~N Gomez, {\L}ukasz Kaiser, and Illia Polosukhin.
\newblock Attention is all you need.
\newblock \emph{Advances in neural information processing systems}, 30, 2017.

\bibitem[Vega-Brown and Roy(2016)]{vega2016asymptotically}
William Vega-Brown and Nicholas Roy.
\newblock Asymptotically optimal planning under piecewise-analytic constraints.
\newblock 2016.
\newblock URL \url{http://www.wafr.org/papers/WAFR_2016_paper_11.pdf}.

\bibitem[Vezhnevets et~al.(2017)Vezhnevets, Osindero, Schaul, Heess, Jaderberg, Silver, and Kavukcuoglu]{vezhnevets2017feudal}
Alexander~Sasha Vezhnevets, Simon Osindero, Tom Schaul, Nicolas Heess, Max Jaderberg, David Silver, and Koray Kavukcuoglu.
\newblock Feudal networks for hierarchical reinforcement learning.
\newblock In \emph{International Conference on Machine Learning}, pages 3540--3549. PMLR, 2017.

\bibitem[Villegas et~al.(2022)Villegas, Babaeizadeh, Kindermans, Moraldo, Zhang, Saffar, Castro, Kunze, and Erhan]{villegas2022phenaki}
Ruben Villegas, Mohammad Babaeizadeh, Pieter-Jan Kindermans, Hernan Moraldo, Han Zhang, Mohammad~Taghi Saffar, Santiago Castro, Julius Kunze, and Dumitru Erhan.
\newblock Phenaki: Variable length video generation from open domain textual descriptions.
\newblock In \emph{International Conference on Learning Representations}, 2022.

\bibitem[Vukobratovi{\'c} and Potkonjak(1982)]{vukobratovic1982dynamics}
Miomir Vukobratovi{\'c} and Veljko Potkonjak.
\newblock \emph{Dynamics of manipulation robots: theory and application}.
\newblock Springer, 1982.

\bibitem[Walke et~al.(2023)Walke, Black, Lee, Kim, Du, Zheng, Zhao, Hansen-Estruch, Vuong, He, Myers, Fang, Finn, and Levine]{walke2023bridgedata}
Homer Walke, Kevin Black, Abraham Lee, Moo~Jin Kim, Max Du, Chongyi Zheng, Tony Zhao, Philippe Hansen-Estruch, Quan Vuong, Andre He, Vivek Myers, Kuan Fang, Chelsea Finn, and Sergey Levine.
\newblock Bridgedata v2: A dataset for robot learning at scale.
\newblock In \emph{Conference on Robot Learning (CoRL)}, 2023.

\bibitem[Wang et~al.(2021{\natexlab{a}})Wang, Wang, Xu, Mandlekar, Fei-Fei, and Savarese]{wang2021generalization}
Chen Wang, Rui Wang, Danfei Xu, Ajay Mandlekar, Li~Fei-Fei, and Silvio Savarese.
\newblock Generalization through hand-eye coordination: An action space for learning spatially-invariant visuomotor control.
\newblock \emph{arXiv preprint arXiv:2103.00375}, 2021{\natexlab{a}}.

\bibitem[Wang et~al.(2023{\natexlab{a}})Wang, Du, Li, Yeh, and Shakhnarovich]{wang2023score}
Haochen Wang, Xiaodan Du, Jiahao Li, Raymond~A Yeh, and Greg Shakhnarovich.
\newblock Score jacobian chaining: Lifting pretrained 2d diffusion models for 3d generation.
\newblock In \emph{Proceedings of the IEEE/CVF Conference on Computer Vision and Pattern Recognition}, pages 12619--12629, 2023{\natexlab{a}}.

\bibitem[Wang et~al.(2021{\natexlab{b}})Wang, Yeshwanth, and Nie{\ss}ner]{wang2021sceneformer}
Xinpeng Wang, Chandan Yeshwanth, and Matthias Nie{\ss}ner.
\newblock Sceneformer: Indoor scene generation with transformers.
\newblock In \emph{2021 International Conference on 3D Vision (3DV)}, pages 106--115. IEEE, 2021{\natexlab{b}}.

\bibitem[Wang et~al.(2023{\natexlab{b}})Wang, Zhang, Chen, and Sreenath]{wang2023prompt}
Yen-Jen Wang, Bike Zhang, Jianyu Chen, and Koushil Sreenath.
\newblock Prompt a robot to walk with large language models.
\newblock \emph{arXiv preprint arXiv:2309.09969}, 2023{\natexlab{b}}.

\bibitem[Wang et~al.(2023{\natexlab{c}})Wang, Xian, Chen, Wang, Wang, Fragkiadaki, Erickson, Held, and Gan]{wang2023robogen}
Yufei Wang, Zhou Xian, Feng Chen, Tsun-Hsuan Wang, Yian Wang, Katerina Fragkiadaki, Zackory Erickson, David Held, and Chuang Gan.
\newblock Robogen: Towards unleashing infinite data for automated robot learning via generative simulation.
\newblock \emph{arXiv preprint arXiv:2311.01455}, 2023{\natexlab{c}}.

\bibitem[Wang et~al.(2021{\natexlab{c}})Wang, Garrett, Kaelbling, and Lozano-P{\'e}rez]{wang2021learning}
Zi~Wang, Caelan~Reed Garrett, Leslie~Pack Kaelbling, and Tom{\'a}s Lozano-P{\'e}rez.
\newblock Learning compositional models of robot skills for task and motion planning.
\newblock \emph{The International Journal of Robotics Research}, 40\penalty0 (6-7):\penalty0 866--894, 2021{\natexlab{c}}.

\bibitem[Warren(1989)]{warren1989global}
Charles~W Warren.
\newblock Global path planning using artificial potential fields.
\newblock In \emph{1989 IEEE International Conference on Robotics and Automation}, pages 316--317. IEEE Computer Society, 1989.

\bibitem[Wei et~al.(2018)Wei, Wicke, and Luke]{wei2018hierarchical}
Ermo Wei, Drew Wicke, and Sean Luke.
\newblock Hierarchical approaches for reinforcement learning in parameterized action space.
\newblock \emph{arXiv preprint arXiv:1810.09656}, 2018.

\bibitem[Whitney(1972)]{whitney1972mathematics}
Daniel~E Whitney.
\newblock The mathematics of coordinated control of prosthetic arms and manipulators.
\newblock 1972.

\bibitem[Whitney(2004)]{whitney2004mechanical}
Daniel~E Whitney.
\newblock \emph{Mechanical assemblies: their design, manufacture, and role in product development}, volume~1.
\newblock Oxford university press New York, 2004.

\bibitem[Whitney et~al.(2020)Whitney, Agarwal, Cho, and Gupta]{whitney2020dynamicsaware}
William Whitney, Rajat Agarwal, Kyunghyun Cho, and Abhinav Gupta.
\newblock Dynamics-aware embeddings, 2020.

\bibitem[Wong et~al.(2022)Wong, Tung, Kurenkov, Mandlekar, Fei-Fei, Savarese, and Mart{\'\i}n-Mart{\'\i}n]{wong2022error}
Josiah Wong, Albert Tung, Andrey Kurenkov, Ajay Mandlekar, Li~Fei-Fei, Silvio Savarese, and Roberto Mart{\'\i}n-Mart{\'\i}n.
\newblock Error-aware imitation learning from teleoperation data for mobile manipulation.
\newblock In \emph{Conference on Robot Learning}, pages 1367--1378. PMLR, 2022.

\bibitem[Wu et~al.(2023{\natexlab{a}})Wu, Bi, Pfrommer, Cebulla, Mangold, and Beyerer]{wu2023sim2real}
Chengzhi Wu, Xuelei Bi, Julius Pfrommer, Alexander Cebulla, Simon Mangold, and J{\"u}rgen Beyerer.
\newblock Sim2real transfer learning for point cloud segmentation: An industrial application case on autonomous disassembly.
\newblock In \emph{Proceedings of the IEEE/CVF Winter Conference on Applications of Computer Vision}, pages 4531--4540, 2023{\natexlab{a}}.

\bibitem[Wu et~al.(2023{\natexlab{b}})Wu, Antonova, Kan, Lepert, Zeng, Song, Bohg, Rusinkiewicz, and Funkhouser]{wu2023tidybot}
Jimmy Wu, Rika Antonova, Adam Kan, Marion Lepert, Andy Zeng, Shuran Song, Jeannette Bohg, Szymon Rusinkiewicz, and Thomas Funkhouser.
\newblock Tidybot: Personalized robot assistance with large language models.
\newblock \emph{arXiv preprint arXiv:2305.05658}, 2023{\natexlab{b}}.

\bibitem[Xia et~al.(2020)Xia, Li, Mart{\'\i}n-Mart{\'\i}n, Litany, Toshev, and Savarese]{xia2020relmogen}
Fei Xia, Chengshu Li, Roberto Mart{\'\i}n-Mart{\'\i}n, Or~Litany, Alexander Toshev, and Silvio Savarese.
\newblock Relmogen: Leveraging motion generation in reinforcement learning for mobile manipulation.
\newblock \emph{arXiv preprint arXiv:2008.07792}, 2020.

\bibitem[Xie et~al.(2020)Xie, Bharadhwaj, Hafner, Garg, and Shkurti]{xie2020latent}
Kevin Xie, Homanga Bharadhwaj, Danijar Hafner, Animesh Garg, and Florian Shkurti.
\newblock Latent skill planning for exploration and transfer.
\newblock In \emph{International Conference on Learning Representations}, 2020.

\bibitem[Xiong et~al.(2018)Xiong, Wang, Yang, Sun, Han, Zheng, Fu, Zhang, Liu, and Liu]{xiong2018parametrized}
Jiechao Xiong, Qing Wang, Zhuoran Yang, Peng Sun, Lei Han, Yang Zheng, Haobo Fu, Tong Zhang, Ji~Liu, and Han Liu.
\newblock Parametrized deep q-networks learning: Reinforcement learning with discrete-continuous hybrid action space.
\newblock \emph{arXiv preprint arXiv:1810.06394}, 2018.

\bibitem[Xu et~al.(2023)Xu, Wan, Zhang, Liu, Shan, Shen, Wang, Geng, Weng, Chen, et~al.]{xu2023unidexgrasp}
Yinzhen Xu, Weikang Wan, Jialiang Zhang, Haoran Liu, Zikang Shan, Hao Shen, Ruicheng Wang, Haoran Geng, Yijia Weng, Jiayi Chen, et~al.
\newblock Unidexgrasp: Universal robotic dexterous grasping via learning diverse proposal generation and goal-conditioned policy.
\newblock In \emph{Proceedings of the IEEE/CVF Conference on Computer Vision and Pattern Recognition}, pages 4737--4746, 2023.

\bibitem[Yamada et~al.(2021)Yamada, Lee, Salhotra, Pertsch, Pflueger, Sukhatme, Lim, and Englert]{yamada2021motion}
Jun Yamada, Youngwoon Lee, Gautam Salhotra, Karl Pertsch, Max Pflueger, Gaurav Sukhatme, Joseph Lim, and Peter Englert.
\newblock Motion planner augmented reinforcement learning for robot manipulation in obstructed environments.
\newblock In \emph{Conference on Robot Learning}, pages 589--603. PMLR, 2021.

\bibitem[Yang et~al.(2024)Yang, Kang, Huang, Xu, Feng, and Zhao]{depthanything}
Lihe Yang, Bingyi Kang, Zilong Huang, Xiaogang Xu, Jiashi Feng, and Hengshuang Zhao.
\newblock Depth anything: Unleashing the power of large-scale unlabeled data.
\newblock In \emph{CVPR}, 2024.

\bibitem[Yarats et~al.(2021)Yarats, Fergus, Lazaric, and Pinto]{yarats2021mastering}
Denis Yarats, Rob Fergus, Alessandro Lazaric, and Lerrel Pinto.
\newblock Mastering visual continuous control: Improved data-augmented reinforcement learning.
\newblock \emph{arXiv preprint arXiv:2107.09645}, 2021.

\bibitem[Ye et~al.(2023)Ye, Li, Gupta, De~Mello, Birchfield, Song, Tulsiani, and Liu]{ye2023affordance}
Yufei Ye, Xueting Li, Abhinav Gupta, Shalini De~Mello, Stan Birchfield, Jiaming Song, Shubham Tulsiani, and Sifei Liu.
\newblock Affordance diffusion: Synthesizing hand-object interactions.
\newblock In \emph{Proceedings of the IEEE/CVF Conference on Computer Vision and Pattern Recognition}, pages 22479--22489, 2023.

\bibitem[Yu et~al.(2020)Yu, Quillen, He, Julian, Hausman, Finn, and Levine]{yu2020meta}
Tianhe Yu, Deirdre Quillen, Zhanpeng He, Ryan Julian, Karol Hausman, Chelsea Finn, and Sergey Levine.
\newblock Meta-world: A benchmark and evaluation for multi-task and meta reinforcement learning.
\newblock In \emph{Conference on Robot Learning}, pages 1094--1100. PMLR, 2020.

\bibitem[Yu et~al.(2023)Yu, Gileadi, Fu, Kirmani, Lee, Gonzalez~Arenas, Lewis~Chiang, Erez, Hasenclever, Humplik, Ichter, Xiao, Xu, Zeng, Zhang, Heess, Sadigh, Tan, Tassa, and Xia]{yu2023language}
Wenhao Yu, Nimrod Gileadi, Chuyuan Fu, Sean Kirmani, Kuang-Huei Lee, Montse Gonzalez~Arenas, Hao-Tien Lewis~Chiang, Tom Erez, Leonard Hasenclever, Jan Humplik, Brian Ichter, Ted Xiao, Peng Xu, Andy Zeng, Tingnan Zhang, Nicolas Heess, Dorsa Sadigh, Jie Tan, Yuval Tassa, and Fei Xia.
\newblock Language to rewards for robotic skill synthesis.
\newblock \emph{Arxiv preprint arXiv:2306.08647}, 2023.

\bibitem[Zhang et~al.(2023)Zhang, Zhang, Pertsch, Liu, Ren, Chang, Sun, and Lim]{zhangboss}
Jesse Zhang, Jiahui Zhang, Karl Pertsch, Ziyi Liu, Xiang Ren, Minsuk Chang, Shao-Hua Sun, and Joseph~J Lim.
\newblock Bootstrap your own skills: Learning to solve new tasks with large language model guidance.
\newblock \emph{Conference on Robot Learning}, 2023.

\bibitem[Zhang and Cho(2016)]{zhang2016query}
Jiakai Zhang and Kyunghyun Cho.
\newblock Query-efficient imitation learning for end-to-end autonomous driving.
\newblock \emph{arXiv preprint arXiv:1605.06450}, 2016.

\bibitem[Zhang et~al.(2017)Zhang, McCarthy, Jow, Lee, Goldberg, and Abbeel]{zhang2017deep}
Tianhao Zhang, Zoe McCarthy, Owen Jow, Dennis Lee, Ken Goldberg, and Pieter Abbeel.
\newblock Deep imitation learning for complex manipulation tasks from virtual reality teleoperation.
\newblock \emph{arXiv preprint arXiv:1710.04615}, 2017.

\bibitem[Zhang et~al.(2018)Zhang, McCarthy, Jow, Lee, Chen, Goldberg, and Abbeel]{zhang2018deep}
Tianhao Zhang, Zoe McCarthy, Owen Jow, Dennis Lee, Xi~Chen, Ken Goldberg, and Pieter Abbeel.
\newblock Deep imitation learning for complex manipulation tasks from virtual reality teleoperation.
\newblock In \emph{2018 IEEE International Conference on Robotics and Automation (ICRA)}, pages 5628--5635. IEEE, 2018.

\bibitem[Zhao et~al.(2023)Zhao, Kumar, Levine, and Finn]{zhao2023learning}
Tony~Z Zhao, Vikash Kumar, Sergey Levine, and Chelsea Finn.
\newblock Learning fine-grained bimanual manipulation with low-cost hardware.
\newblock \emph{arXiv preprint arXiv:2304.13705}, 2023.

\bibitem[Zhou et~al.(2020)Zhou, Bajracharya, and Held]{zhou2020plas}
Wenxuan Zhou, Sujay Bajracharya, and David Held.
\newblock Plas: Latent action space for offline reinforcement learning.
\newblock \emph{arXiv preprint arXiv:2011.07213}, 2020.

\bibitem[Zhou et~al.(2022)Zhou, Girdhar, Joulin, Kr{\"a}henb{\"u}hl, and Misra]{zhou2022detecting}
Xingyi Zhou, Rohit Girdhar, Armand Joulin, Philipp Kr{\"a}henb{\"u}hl, and Ishan Misra.
\newblock Detecting twenty-thousand classes using image-level supervision.
\newblock In \emph{Computer Vision--ECCV 2022: 17th European Conference, Tel Aviv, Israel, October 23--27, 2022, Proceedings, Part IX}, pages 350--368. Springer, 2022.

\bibitem[Zhu et~al.(2020{\natexlab{a}})Zhu, Yu, Gupta, Shah, Hartikainen, Singh, Kumar, and Levine]{zhu2020ingredients}
Henry Zhu, Justin Yu, Abhishek Gupta, Dhruv Shah, Kristian Hartikainen, Avi Singh, Vikash Kumar, and Sergey Levine.
\newblock The ingredients of real-world robotic reinforcement learning.
\newblock \emph{arXiv preprint arXiv:2004.12570}, 2020{\natexlab{a}}.

\bibitem[Zhu et~al.(2020{\natexlab{b}})Zhu, Wong, Mandlekar, and Martín-Martín]{zhu2020robosuite}
Yuke Zhu, Josiah Wong, Ajay Mandlekar, and Roberto Martín-Martín.
\newblock robosuite: A modular simulation framework and benchmark for robot learning, 2020{\natexlab{b}}.

\bibitem[Zhuang et~al.(2023)Zhuang, Fu, Wang, Atkeson, Schwertfeger, Finn, and Zhao]{zhuang2023robot}
Ziwen Zhuang, Zipeng Fu, Jianren Wang, Christopher Atkeson, Soeren Schwertfeger, Chelsea Finn, and Hang Zhao.
\newblock Robot parkour learning.
\newblock \emph{arXiv preprint arXiv:2309.05665}, 2023.

\end{thebibliography}
\bibliographystyle{plainnat}

\end{document}